\PassOptionsToPackage{dvipsnames}{xcolor}     %
\documentclass{article} %
\usepackage{colm2024_conference}

\usepackage{microtype}
\usepackage{caption}
\usepackage{graphicx}
\usepackage{wrapfig}
\usepackage{tocloft}
\usepackage{hyperref}
\newcommand\myshade{85}
\colorlet{mylinkcolor}{RoyalBlue}
\colorlet{mycitecolor}{violet}
\colorlet{myurlcolor}{YellowOrange}
\usepackage{makecell}
\hypersetup{
  linkcolor  = mylinkcolor!\myshade!black,
  citecolor  = mycitecolor!\myshade!black,
  urlcolor   = myurlcolor!\myshade!black,
  colorlinks = true,
}

\usepackage{multicol}
\usepackage{url}
\usepackage{booktabs}
\usepackage{multirow}
\usepackage{makecell}
\usepackage{arydshln}
\usepackage{xcolor}
\usepackage{pdfpages}
\usepackage{CJKutf8}
\usepackage{xspace}

\newcommand{\cmmmu}{\textit{CMMMU}\xspace}
\usepackage{tcolorbox}
\usepackage{xcolor} %
\usepackage{soulutf8}
\definecolor{softblue}{rgb}{0.88, 0.95, 1.0} %
\definecolor{softyellow}{rgb}{0.98, 0.98, 0.82} %

\newcommand{\highlightblue}[1]{\sethlcolor{softblue}\hl{#1}}

\newcommand{\listcasestudyfiguresname}{\normalsize{List of Case Study Figures}}
\newlistof{casestudyfigures}{csf}{\listcasestudyfiguresname}

\newcommand{\casestudyfigure}[4]{%
  \clearpage
  \begin{figure}[ht]
    \centering
    \refstepcounter{casestudyfigures}%
    \addcontentsline{csf}{casestudyfigures}{\protect\numberline{\thecasestudyfigures}#2}%
    \includegraphics[width=0.98\textwidth]{#1}
    \caption{#3}
    \label{#4}
    \hyperlink{listofcasestudyfigures}{Back to List of figures}
\end{figure}}

\title{\cmmmu: \\ A Chinese Massive Multi-discipline Multimodal Understanding Benchmark}

\newcommand*\samethanks[1][\value{footnote}]{\footnotemark[#1]}
\author{
\hspace{-2mm}
\small
Ge Zhang\textsuperscript{\includegraphics[scale=0.03]{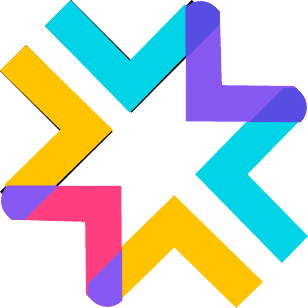},1,2}\thanks{These authors contribute equally to the work.}\ \ 
Xinrun Du\textsuperscript{\includegraphics[scale=0.03]{fig/map-logo-c.pdf}}\samethanks[1]\ \ 
Bei Chen\textsuperscript{9}\samethanks[1]
\\\small 
\textbf{Yiming Liang}\textsuperscript{3,4}\ 
\textbf{Tongxu Luo}\textsuperscript{1} \ 
\textbf{Tianyu Zheng}\textsuperscript{\includegraphics[scale=0.03]{fig/map-logo-c.pdf},9}\ 
\textbf{Kang Zhu}\textsuperscript{\includegraphics[scale=0.03]{fig/map-logo-c.pdf}}\ 
\textbf{Yuyang Cheng}\textsuperscript{1,5}\ 
\textbf{Chunpu Xu}\textsuperscript{6}
\\\small 
\textbf{Shuyue Guo}\textsuperscript{9}\ 
\textbf{Haoran Zhang}\textsuperscript{1}\ 
\textbf{Xingwei Qu}\textsuperscript{\includegraphics[scale=0.03]{fig/map-logo-c.pdf}}\ 
\textbf{Junjie Wang}\textsuperscript{1,7}\ 
\textbf{Ruibin Yuan}\textsuperscript{\includegraphics[scale=0.03]{fig/map-logo-c.pdf},1}\ 
\textbf{Yizhi Li}\textsuperscript{\includegraphics[scale=0.03]{fig/map-logo-c.pdf},8}\ 
\\\small 
\textbf{Zekun Wang}\textsuperscript{\includegraphics[scale=0.03]{fig/map-logo-c.pdf},9}\
\textbf{Yudong Liu}\textsuperscript{9}\ 
\textbf{Yu-Hsuan Tsai}\textsuperscript{9}\ 
\textbf{Fengji Zhang}\textsuperscript{9}\ 
\\\small 
\textbf{Chenghua Lin}\textsuperscript{\includegraphics[scale=0.03]{fig/map-logo-c.pdf},8}\ \ \ 
\textbf{Wenhao Huang}\textsuperscript{\includegraphics[scale=0.03]{fig/map-logo-c.pdf},9}\thanks{Corresponding Authors.}\ \ \ 
\textbf{Jie Fu}\textsuperscript{1}\samethanks[2] 
\\
\scriptsize
    \textsuperscript{\includegraphics[scale=0.03]{fig/map-logo-c.pdf}}Multimodal Art Projection Research Community\quad
    \textsuperscript{1}Hong Kong University of Science and Technology\quad
\\ 
\scriptsize
    \textsuperscript{2}University of Waterloo\quad
    \textsuperscript{3}Institute of Automation, Chinese Academy of Sciences\quad
\\
\scriptsize
    \textsuperscript{4}School of Artificial Intelligence, University of Chinese Academy of Sciences\quad
    \textsuperscript{5}Peking University \quad
\\
\scriptsize
    \textsuperscript{6}The Hong Kong Polytechnic University \quad
    \textsuperscript{7}Waseda University \quad
    \textsuperscript{8}University of Manchester\quad
    \textsuperscript{9}01.AI \quad
\\
\small
\texttt{\{zhangge,huangwenhao,duxinrun,chenbei\}@01.ai}\quad 
\texttt{jiefu@ust.hk}
\\
\url{https://cmmmu-benchmark.github.io/}
}
\colmfinalcopy

\begin{document}

\maketitle

\begin{abstract}

As the capabilities of large multimodal models (\textbf{LMMs}) continue to advance, evaluating the performance of LMMs emerges as an increasing need. Additionally, there is an even larger gap in evaluating the advanced knowledge and reasoning abilities of LMMs in non-English contexts such as Chinese. We introduce \textbf{\cmmmu}, a new Chinese Massive Multi-discipline Multimodal Understanding benchmark designed to evaluate LMMs on tasks demanding college-level subject knowledge and deliberate reasoning in a Chinese context. \cmmmu is inspired by and strictly follows the annotation and analysis pattern of \textit{MMMU}~\citep{yue2023mmmu}. 
\cmmmu includes 12k manually collected multimodal questions from college exams, quizzes, and textbooks, covering six core disciplines. %
These questions span 30 subjects and comprise 39 highly heterogeneous image types, such as charts, diagrams, maps, tables, music sheets, and chemical structures.
\cmmmu focuses on complex perception and reasoning with domain-specific knowledge in the Chinese context. We evaluate 11 open-source LLMs and one proprietary GPT-4V(ision). Even GPT-4V only achieves accuracy of 43\%, indicating a large space for improvement. \cmmmu aims to enhance the development of next-generation LMMs for expert AI and support LMM democratization through offering varied language contexts.

\end{abstract}

\section{Introduction}
Large Multimodal Models (\textbf{LMMs}) have exhibited impressive problem-solving skills in many tasks, \textit{e.g.}, zero-shot image/video classification, zero-shot image/video-text retrieval, and multimodal question answering. %
But \citet{yue2023mmmu, lu2023mathvista, deng2023mind2web} reveals a significant gap between advanced LMMs and multimodal expert AI, notably in complex perception and reasoning within specialized knowledge areas.
To close this gap, college-level exams for different disciplines are a natural starting point for evaluating whether a Large Language Model (\textbf{LLM}) or an LMM can perform like an expert adult 
\cite{yue2023mmmu,hendrycks2020measuring,zhong2023agieval,zhang2023evaluating}.

Additionally, with benchmarks a void, the development of bilingual LMMs has no sense of direction.
We fill the gap by proposing \cmmmu, a new comprehensive Chinese benchmark designed to evaluate LMMs on massive multi-discipline tasks, guiding the development of bilingual LMMs towards a path toward expert-level artificial intelligence.

\begin{wrapfigure}{r}{0.5\textwidth}
  \centering
      \includegraphics[width=0.5\textwidth]{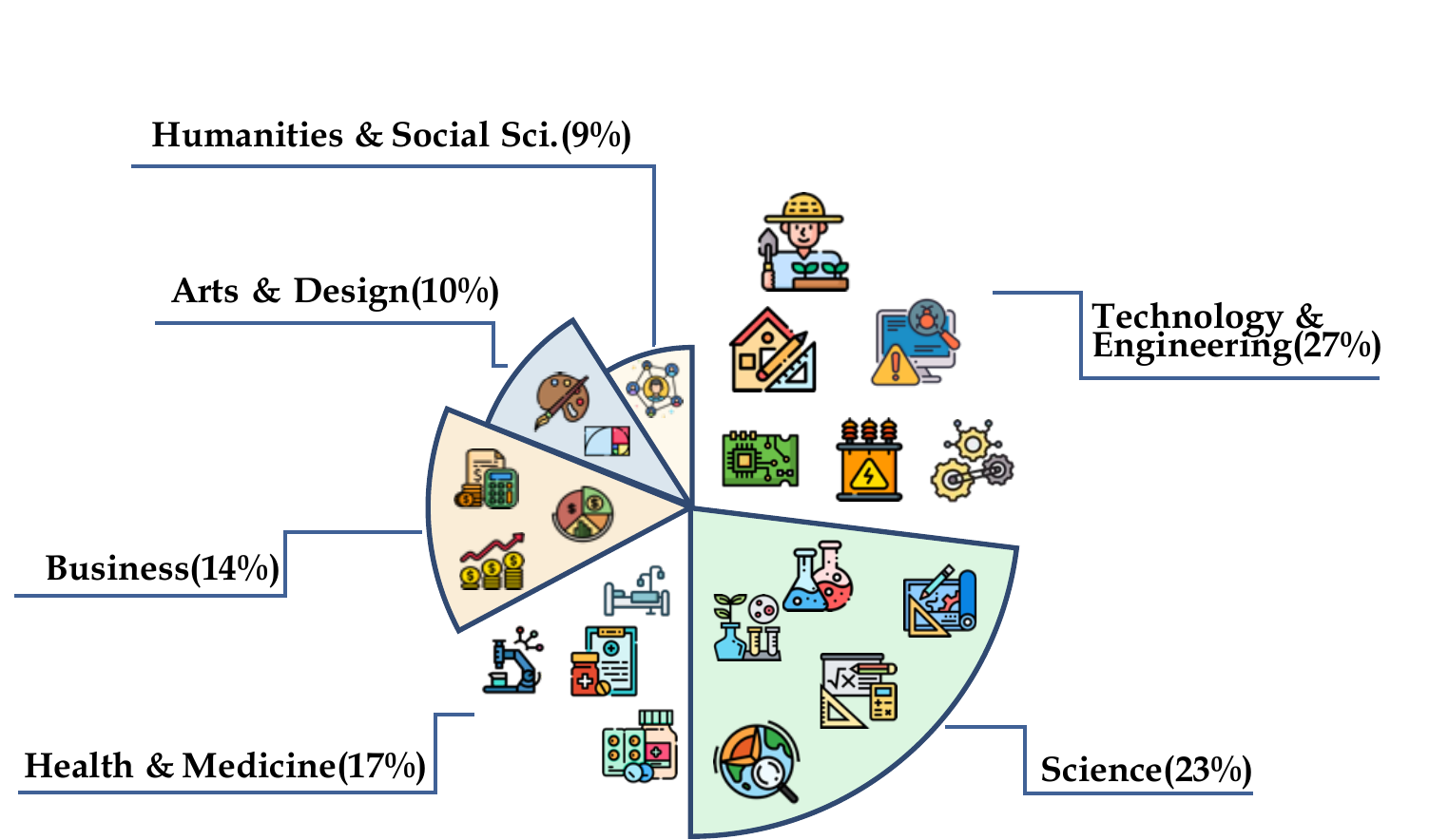} 
      \vspace{-0.15in}
  \caption{Disciplines of \cmmmu.}
  \label{fig:CMMMU-discipline}
\end{wrapfigure}

As in Fig. \ref{fig:CMMMU-discipline}, \cmmmu, including 12k manually collected Chinese multimodal questions from college exams, quizzes, and textbooks, covering six core disciplines: Art \& Design, Business, Science, Health \& Medicine, Humanities \& Social Science, and Tech \& Engineering, is one of the most comprehensive benchmarks for evaluating LMMs' complex reasoning and perception abilities.
Each question in \cmmmu is further annotated with detailed subfields and image types to investigate which types of questions are difficult for LMMs.

We provide a comprehensive error analysis of 150 samples, which GPT-4V(ision) answers incorrectly, evenly distributed among 30 subjects, and covering most cases leading the most advanced LMMs to astray. By evaluating top-performing LMMs, \textit{e.g.}, Qwen-VL-Plus and GPT-4V, on \cmmmu, we argue that there is still a long way to go towards an expert-level bilingual LMM. Even the most advanced closed-source LMMs, GPT-4V and Qwen-VL-Plus, only achieve accuracies of 43\% and 36\%, respectively, indicating significant room for improvement. We further reveal that the gap between LMMs released by the open-source community and the most powerful closed-source LMMs in a Chinese context is much smaller than in English, as demonstrated in MMMU. 
For example, the most powerful open-source LMM, \textit{i.e.}, Yi-VL-34B, achieves an accuracy of 36\%, with a 7\% gap compared to GPT-4V, while the gap in English is 11\%. In light of the insights obtained while developing \cmmmu and benchmarking existing open-source LMMs, we observe that only Yi-VL-6B\footnote{\url{https://huggingface.co/01-ai/Yi-VL-6B}}, Yi-VL-34B\footnote{\url{https://huggingface.co/01-ai/Yi-VL-34B}}, and Qwen-VL-Chat perform notably better compared to a random choice setting and are close to GPT-4V, while other open-source LMMs perform similarly to the random choice setting. %
Surprisingly, Yi-VL-34B even narrows the gap between open-source LMMs and GPT-4V on \cmmmu to 7\%. 

We believe \cmmmu can benefit the ongoing LMM research and development efforts, and promote the democratization of LMMs.
Our contributions are summarized as follows:
\begin{itemize}
    \item We introduce \cmmmu, the first Chinese Massive Multi-discipline Multimodal Understanding benchmark.
    \item We reveal that existing LMMs, even including GPT-4V, perform poorly on complex reasoning and understanding in a Chinese context.
    \item We examine the gap between open-source bilingual LMMs and closed-source LMMs in Chinese, finding it notably narrower than in English contexts.
\end{itemize}

\section{Related Work}
\subsection{Multimodal Benchmark}

Traditionally, multimodal benchmarks are task-oriented, thus not designed to evaluate LMMs.and benchmarking relies on tasks that align and utilize representations from various modalities, such as visual question answering (VQA)~\citep{antol2015vqa}, image captioning~\citep{vinyals2014show}, and information retrieval~\citep{wei2023uniir,wu2024scimmir}. The success of building such multimodal tasks and benchmarks heavily relies on large-scale annotated datasets like MSCOCO~\citep{lin2014mscoco} and Flickr30k~\citep{plummer2015flickr30k}. Some work also evaluates the cross-modal alignment ability with VQA data derived from general knowledge bases~\citep{marino2019okvqa, schwenk2022okvqa}.

A recent line of research attempts to design benchmarks tailored to evaluating LMMs. For example, we can examine the models by requiring them to perceive and learn the complicated knowledge from the given data distribution, \textit{e.g.}, in the scientific domain~\citep{lu2022learn, wu2024scimmir}. To construct benchmarks compatible with generative LMMs, MME \citep{fu2023mme} uses yes-no problems, and MMBench~\citep{liu2023mmbench} is based on the multi-choice format. Some recent studies propose examining whether models can perceive and interpret information produced in more challenging scenarios like math reasoning~\citep{lu2023mathvista}, website interaction~\cite{deng2023mind2web}, or comprehensive college-level knowledge reasoning~\citep{yue2023mmmu}. Though promising progress in this field of multimodal benchmarking has been made, a dominant ratio of the dataset is in English, which makes it an urgent gap to build a comprehensive and challenging benchmark in other frequently used languages like Chinese.

\begin{figure*}[t]
    \centering
    \includegraphics[width=0.9\textwidth]{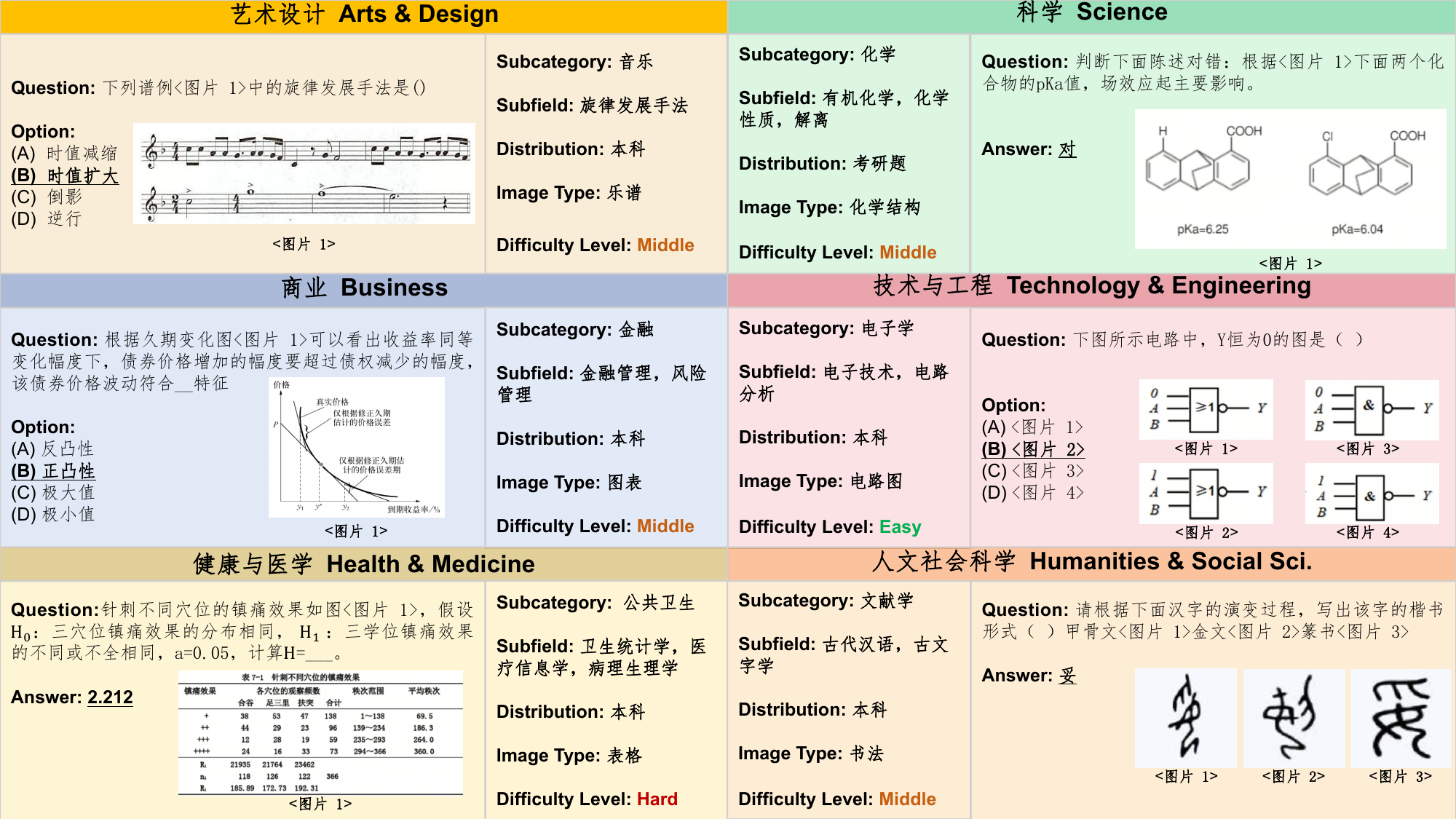}
    \caption{\cmmmu examples sampled from each discipline. The pictures include music scores, tables, chemical structures, curves, circuit diagrams and other types of pictures, and the difficulty of the questions requires expert-level knowledge to understand and reason.}
    \label{fig:CMMMU-sample}
\end{figure*}

\subsection{Bilingual Large Multimodal Models}

Different from the development trace of the benchmarks, many of the existing multimodal models support both English and Chinese due to the integrated bilingual large language models(\textbf{LLMs}). Although such a statement is established in different models, the cases may vary on nuanced features. While multimodal models aim to go beyond the textual data by adapting the language models with cross-modality alignment methods, some language models pre-trained with Chinese-English bilingual corpus are selected as the component for text modeling~\citep{hu2023viscpm, Bai2023Qwen-VL, ding2021cogview,du2022glm, linksoulai2023chinesellava}. Although some interesting insights are explored, only a few of the models are evaluated on Chinese multimodal tasks. For instance, it is revealed by \citet{hu2023viscpm} that multimodal models trained only with English instruction tuning data work well in Chinese even in the zero-shot setting. Another set of models selects the language models adapted to Chinese with efficient tuning~\citep{chinese-llama-alpaca}. Given the proper alignment architecture designs and training data selection, these models still show strong performances on bilingual multimodal tasks~\citep{ye2023mplugowl2, sun2023Emu2, chen2023internvl, wang2023cogvlm, hong2023cogagent, linksoulai2023chinesellava}. Moreover, even though the closed-source GPT-4~\citep{achiam2023gpt4} does not provide architecture-relevant details, it is a worth mentioning baseline for %
Chinese multimodal benchmarks, given it achieves visual understanding tasks in English close to human-level.

Regardless of the choice of language models and the training data, many multimodal models show the capability for Chinese tasks at a certain level in practical use. In this work, we aim to quantitatively measure the ability boundaries of the models with comprehensive and challenging Chinese multimodal tasks, as most of them have been only assessed with English tasks.

\section{The \cmmmu Benchmark}

We introduce the Chinese Massive Multi-discipline Multimodal Understanding (\textbf{\textit{CMMMU}}) benchmark, a manually curated benchmark covering college-level knowledge to evaluate LMMs' expert-level multimodal understanding capability across a broad scope of tasks. \cmmmu is the first multimodal question-answering benchmark in a Chinese context and one of the few existing multimodal benchmarks investigating LMMs' complex understanding and reasoning capacities.

\begin{CJK}{UTF8}{gbsn}
\begin{table*}[ht]
\centering
\small
\resizebox{\linewidth}{!}{
\begin{tabular}{@{}cccccc@{}}
\toprule
\textbf{Image Type} & \textbf{\#Num} & \textbf{Image Type} & \textbf{\#Num} & \textbf{Image Type} & \textbf{\#Num} \\
\midrule
\makecell{广告\\	\textit{Advertisement}} & 4 & \makecell{历史时间线\\ \textit{Historical Timeline}} & 6 & \makecell{人体扫描\\ \textit{Body Scan}} & 9 \\
\midrule
\makecell{电力学符号\\ \textit{Electrical Symbols}} & 10 & \makecell{DNA序列\\ \textit{DNA Sequence}} & 13 & \makecell{数学符号\\ \textit{Mathematical Symbols}} & 21 \\
\midrule
\makecell{标志和品牌形象\\ \textit{Logos and Brand Identity}} & 22 & \makecell{风景画\\ \textit{Landscape Painting}} & 23 & \makecell{3D渲染图\\ \textit{3D Rendering}} & 24 \\
\midrule
\makecell{天文图像\\ \textit{Astronomical Images}} & 31 & \makecell{图标和符号\\ \textit{Icons and Symbols}} & 31 & \makecell{其他\\ \textit{Other}} & 39 \\ 
\midrule
\makecell{海报\\ \textit{Poster}} & 47 & \makecell{树形图\\ \textit{Tree Diagram}} & 54 & \makecell{雕塑\\ \textit{Sculpture}} & 67 \\
\midrule
\makecell{书法\\ \textit{Calligraphy}} & 72 & \makecell{有向图\\ \textit{Directed Graph}} & 82 & \makecell{地图\\ \textit{Map}} & 85 \\
\midrule
\makecell{建筑设计图\\ \textit{Architectural Design Drawing}} & 94 & \makecell{病理图像\\ \textit{Pathology Images}} & 99 & \makecell{机械工程图\\ \textit{Mechanical Engineering Drawings}} & 107 \\
\midrule
\makecell{流程图\\ \textit{Flowchart}} & 128 & \makecell{乐谱\\ \textit{Sheet Music}} & 137 & \makecell{系统框图\\ \textit{System Diagram}} & 174 \\
\midrule
\makecell{漫画和卡通\\ \textit{Cartoons and Comics}} & 209 & \makecell{肖像\\ \textit{Portrait}} & 235 & \makecell{绘画作品\\ \textit{Artwork}} & 286 \\
\midrule
\makecell{屏幕截图\\ \textit{Screenshot}} & 301 & \makecell{机械结构图\\ \textit{Mechanical Structure Diagram}} & 339 & \makecell{几何形状\\ \textit{Geometric Shapes}} & 346 \\
\midrule
\makecell{显微镜图像\\ \textit{Microscope Image}} & 416 & \makecell{医学图像\\ \textit{Medical Images}} & 491 & \makecell{工程结构图\\ \textit{Engineering Structural Diagram}} & 517 \\
\midrule
\makecell{电路图\\ \textit{Circuit Diagram}} & 557 & \makecell{化学结构\\ \textit{Chemical Structures}} & 676 & \makecell{图表\\ \textit{Charts}} & 851 \\
\midrule
\makecell{照片\\ \textit{Photographs}} & 1680 & \makecell{表格\\ \textit{Table}} & 2480 & \makecell{草图\\ \textit{Sketches}} & 3180 \\
\bottomrule
\end{tabular}
}
\caption{Image type and corresponding number.}
\label{tab:images type}
\end{table*}
\end{CJK}
\vspace{-0.2cm}
\subsection{Data Curation Process}
\textbf{Data Collection:} We carefully design a three-stage data collection procedure.
In \textbf{Stage 1}, annotator organizers (mainly the authors) collect sources satisfying license requirements in the format of website links or book titles.
The annotator organizers are well instructed to adhere to copyright and license regulations, avoiding data from sites prohibiting copying and redistribution.
We collect at least 20 annotation sources, \textit{i.e.}, websites or books, for each subject in each discipline.
In \textbf{Stage 2}, annotator organizers forward the annotation sources to the crowdsourcing annotators for further annotation.
All annotators are undergraduate students or have higher degrees to ensure they can verify the annotated questions and related explanations.
During the annotation process, we ask the annotators to strictly follow several key principles to filter out unqualified questions with images: (1) Questions that can be answered without the images should be filtered out. (2) Questions that use the same image should be filtered out as much as possible. (3) Questions not requiring expert knowledge to answer should be filtered out as much as possible. (4) The number of questions that are about the same specific knowledge point and have similar question angles should not exceed 10. 
We also ask annotators to follow the data annotation protocol in the Appendix.G of \citet{yue2023mmmu}.
In \textbf{Stage 3}, annotator organizers additionally supplement questions to subjects that lack questions, \textit{e.g.}, Arts, Diagnostics, and Economics, to balance the datasets.

\begin{table}
\centering
\resizebox{1\columnwidth}{!}
{
\begin{tabular}{@{}lccccc@{}}

  \toprule
  Dataset & Size & Images & Format & Source & Answer \\
  \midrule
  VQA ~\citep{agrawal2015vqa} & $>1$M & V & I+T & Annotated & Open \\
  GQA ~\citep{hudson2019gqa} & $>1$M & V & I+T & Synthesized & Open \\
  VizWiz ~\citep{gurari2018vizwiz} & 32K & V & I+T & Annotated & Open \\
  TextVQA ~\citep{ganz2023towards} & 45K & OC & I+T & Annotated & MC \\
  OKVQA ~\citep{marino2019okvqa} & 14K & V+OC & I+T & Annotated & Open \\
  SEED ~\citep{li2023seedbench} & 19K & V+OC & I+T & Annotated & MC \\
  MMBench ~\citep{liu2023mmbench} & 3K & V+OC & I+T & Repurposed & MC \\
  MM-Vet ~\citep{yu2023mmvet} & 0.2K & V+OC & I+T & Annotated & Open \\
  ScienceQA ~\citep{lu2022learn} & 6K & 5 Types & I+T & Textbooks & MC \\
  MathVista ~\citep{lu2023mathvista} & 6K & V+OC & I+T & Synthesized & MC/Open \\
  \midrule
  MMMU ~\citep{yue2023mmmu} & 11.5K & 30 Types & Interleaved & \makecell{Textbooks \\ Internet \\ Annotated} & \makecell{Open \\ MC} \\
  \midrule
  \cmmmu & 12K & 39 Types & Interleaved & \makecell{Textbooks \\ Internet \\ Annotated} & \makecell{Open \\ MC \\ T/F} \\
  \bottomrule
\end{tabular}
}
\caption{Comparison with other benchmarks. V: visual input, OC: optical characters, I+T: images and text, Open: open questions, MC: multiple choice questions, FIB: fill in the blank questions, T/F: true or false questions.}\label{tab:comparison with multimodal benchmarks}
\vspace{-0.2cm}
\end{table}

\textbf{Data Quality Control:} To further improve the data quality of \cmmmu, we follow a strict data quality control protocol.
\textbf{First}, each question is manually verified by at least one of the paper's authors.
We carefully filter out questions with answers that are too hard to extract from the responses generated by LMMs.
During the process, we also carefully filter out all the questions that are not up to college-level examinations.
\textbf{Second}, given the concern of data contamination, we filter out all the questions that can be correctly solved by GPT-4, Qwen-7B, Deepseek-7B, and Yi-7B simultaneously without the assistance of OCR. Some example questions are shown in Fig. \ref{fig:CMMMU-sample}.

\subsection{Comparison with Existing Benchmarks}

We compare \cmmmu with existing multimodal benchmarks in Tab. \ref{tab:comparison with multimodal benchmarks}.From the input image type, the common image formats in the benchmark can be roughly divided into three simple categories: visual input (V), optical characters (OC) and V+OC. 
In addition, there are 5 types of image formats in the ScienceQA benchmark. 
\cmmmu benchmark has 39 types as in Tab. \ref{tab:images type}, involving charts, tables, diagrams, chemical structures, photos, paintings, geometric shapes, musical scores, and medical images. 
Concerning the input format, existing benchmarks generally exhibit a relatively independent relationship between the input images and text(I+T). In the \cmmmu benchmark, images and text are interleaved, establishing a markedly tighter connection.
In terms of question types, most of the common benchmarks are open questions (Open) or multiple choice questions (MC). 
\cmmmu not only contains open-ended questions and multiple choice questions, but also adds judgment questions to enrich the question types. 
In terms of knowledge depth, previous benchmarks typically require common sense or simple physical or temporal reasoning. In contrast, our proposed \cmmmu benchmark requires thoughtful reasoning with university-level subject knowledge.

\begin{table}[h]
\hspace*{-0.1\linewidth}
\centering
\small
\begin{minipage}[t]{0.4\linewidth}
\centering
\begin{tabular}{@{}lc@{}}
\toprule
Statistics                               & Number          \\ 
\midrule
Total Questions                          & 12012          \\
Disciplines/Subjects/Subfields           & 6/30/4165       \\
Image Types                              & 39              \\
\midrule
Dev:Validation:Test                      & 112:900:11000   \\
Easy: Medium: Hard                       & 30\%:58\%:12\%  \\
\midrule
Average question length                  & 51.12           \\
Average option length                    & 8.76            \\
Average explanation length               & 78.29          \\
\bottomrule
\end{tabular}
\end{minipage}%
\hspace{0.1\linewidth}
\begin{minipage}[t]{0.40\linewidth}
\centering
\begin{tabular}{@{}lc@{}}
\toprule
Statistics                            & Number \\
\addlinespace[2pt]
\midrule
Multiple-choice Questions                & 7738 (64.41\%) \\
\addlinespace[2pt]
Fill in the blank Questions              & 2998 (24.95\%) \\
\addlinespace[2pt]
True or false Questions                  & 1276 (10.62\%) \\
\addlinespace[2pt]
\midrule
Questions with an Explanation            & 247 (2.05\%)  \\
\addlinespace[2pt]
Image in the Question                    & 11760 (84.42\%) \\
\addlinespace[2pt]
Image in Options                         & 2169 (15.57\%) \\
\addlinespace[2pt]
Example with Multiple Images             & 597 (4.97\%) \\
\addlinespace[2pt]
\bottomrule
\end{tabular}
\end{minipage}
\caption{Statistics of \cmmmu}
\vspace{-0.4cm}
\label{tab:statistics}
\end{table}

\subsection{Statistics of \cmmmu}

\cmmmu covers 6 disciplines, including Art \& Design, Business, Science, Health \& Medicine, Humanities \& Social Science, and Tech \& Engineering, 
spanning over 30 subjects. As Fig. \ref{tab:statistics}, \cmmmu consists of 12K questions, divided into few-shot development set, validation set, and test set. The few-shot development set comprises 5 questions for each topic, the validation set aids in hyperparameter selection with 900 questions, and the test set includes 11K questions. 

The pictures include 39 types such as pathological diagrams, musical scores, circuit diagrams, and chemical structure diagrams. We categorized the data as Easy (30\%), Medium (58\%), and Hard (12\%) by logical difficulty rather than intellectual difficulty.
According to the question type, there are 7738 multiple choice questions, 2998 fill-in-the-blank questions, and 1276 judgment questions. Of these examples, 11,760 are images in the question, 2169 are images in the option, and 597 are images with multiple images. %
The average question length is approximately 51 words, the average option length is about 9 words, and the average explanation length is around 78 words.

\begin{table}[ht]
\centering
\resizebox{\linewidth}{!}{
\begin{tabular}{@{}lcccccccc@{}}
\toprule
& \textbf{\makecell{Validation \\ Overall}} & \textbf{\makecell{Test \\ Overall}} & \textbf{\makecell{Art \& \\ Design }} & \textbf{\makecell{Business }} & \textbf{\makecell{Science }} & \textbf{\makecell{Health \& \\ Medicine}} & \textbf{\makecell{Human. \& \\Social Sci.}} & \textbf{\makecell{Tech \& \\ Eng.}} \\
& (900) & (11,000) & (1,091) & (1,538) & (2,494) & (1,865) & (1,038) & (2,974) \\
\midrule
Random Choice & 21.6 & 21.6 & 32.9 & 9.1 & 18.8 & 23.8 & 23.8 & 23.9 \\
Frequent Choice & 24.1 & 26.0 & 36.2 & 11.8 & 23.9 & 30.2 & 28.5 & 27.7 \\
\midrule
\multicolumn{9}{c}{\textbf{LMMs: Text + Image as Input}} \\
\midrule
mPLUG-Owl2 & 20.8 & 22.2 & 30.4 & 13.3 & 19.6 & 25.2 & 24.7 & 23.4 \\
VisCPM  & 25.2 & 22.7 & 37.7 & 11.3 & 19.1 & 26.1 & 24.0 & 23.7 \\
Chinese-LLaVA & 25.5 & 23.4 & 34.4 & 11.7 & 21.6 & 25.5 & 26.3 & 24.7 \\
Emu2-Chat & 23.8 & 24.5 & 35.3 & 11.7 & 22.1 & 25.5 & 28.0 & 27.1 \\
CogAgent-Chat & 24.6 & 23.6 & 33.8 & 14.1 & 20.6 & 26.3 & 24.8 & 25.3 \\ 
Qwen-VL-Chat & 30.7 & 31.3 & 52.6 & 18.5 & 26.9 & 33.4 & 34.1 & 31.4 \\
InternVL-Chat-V1.1 & 34.7 & 34.0 & 56.7 & 19.7 & 28.6 & 39.2 & 39.6 & 32.3 \\
Yi-VL-6B & 35.8 & 35.0 & 58.0 & \highlightblue{19.9} & \highlightblue{32.3} & 39.3 & 40.6 & 32.1 \\
Yi-VL-34B & \highlightblue{36.2} & \highlightblue{36.5} & \highlightblue{\textbf{62.9}} & 19.1 & 31.5 & \highlightblue{42.1} & \highlightblue{42.5} & \highlightblue{34.5} \\
\midrule
Qwen-VL-Plus & 39.5 & 36.8 & 61.5 & 23.2 & 32.8 & 40.5 & 43.4 & 33.3 \\
GPT-4V & \textbf{42.5} & \textbf{43.7} & 61.0 & \textbf{36.3} & \textbf{40.9} & \textbf{46.8} & \textbf{44.2} & \textbf{41.5} \\
\midrule
\multicolumn{9}{c}{\textbf{LLMs: Only Text as Input}} \\
\midrule
DeepSeek-7B & 22.3 & 21.9 & 41.3 & 11.2 & 18.3  & 23.5 & 24.7 & 21.3 \\
Baichuan-7B & 26.0 & 24.3 & 42.7 & 12.6 & 19.6 & 28.0 & 27.8 & 23.9 \\
Qwen-7B & 24.7 & 25.1 & 43.8 & 12.6 & 20.7 & 30.5 & 26.9 & 24.5 \\
Yi-6B & 25.6 & 24.2 & 26.3 & 15.0 & 23.4 & 29.1 & 27.0 & 24.7 \\
\midrule
DeepSeek-7B + OCR & 25.2 & 23.2 & 41.2 & 13.2 & 19.4 & 26.1 & 26.5 & 21.8 \\
Baichuan-7B + OCR & 25.3 & 24.7 & 40.2 & 15.2 & 21.0 & 27.9 & 30.7 & 22.8 \\
Qwen-7B + OCR & 27.0 & 26.1 & 44.6 & 14.3 & 22.1 & 29.3 & 29.8 & 25.4 \\
Yi-6B + OCR & 28.4 & 26.8 & 33.4 & 16.9 & 24.8 & 32.3 & 33.2 & 25.5 \\
\bottomrule
\end{tabular}
}
\caption{Overall results of open-source and closed-source models on the \cmmmu validation and test set. \textbf{bold results} in LMMs indicate the best results for all models, and the \highlightblue{blue results} indicate the best results among the open-source models.}
\label{tab:performance}
\end{table}
\vspace{-0.2cm}
\section{Experiments}
We perform a comprehensive evaluation of various models, including LLMs and LMMs, with considering both closed-source and open-source implementations. 
The evaluation process employs zero-shot settings, rather than fine-tuning or few-shot settings, to examine the raw ability of the model to generate accurate answers on multimodal tasks. 
For models with corresponding task prompts, we use the default prompts for either multiple-choice or open-ended question-answering tasks. 
As for models without corresponding task prompts, we use the same task prompts, which are hand-picked on the validation set. 
In addition, we also test the results of some models on few-shot settings, which are documented in the Appendix. 
All these experiments are performed on NVIDIA A100 GPUs.
\subsection{Baselines}
\textbf{LMMs.} We consider the current mainstream Chinese-English bilingual large multimodal models. We use each model's official API (closed-source) or official checkpoint (open-source) published on the huggingface website. Baselines includes: 
(1) mPLUG-Owl2 ~\citep{ye2023mplugowl2} employs a modular network design with a language decoder as a common interface for managing different modalities, effectively exploiting modal collaboration to improve performance in textual and multimodal tasks. 
(2) VisCPM ~\citep{hu2023viscpm} is trained based on the large language model CPM-Bee with 10B parameters, fusing visual encoder (Q-Former) and visual decoder (Diffusion-UNet) to support visual inputs and outputs. 
(3) Chinese-LLaVA ~\citep{linksoulai2023chinesellava} uses Chinese Llama2 as the language model base, plus image understanding capabilities. The work follows the structure of LLaVA with a two-stage training using Chinese data. %
(4) Emu2 ~\citep{sun2023Emu2} is a generative multimodal model with 37 billion parameters that performs well in few-shot Settings. 
(5) CogAgent ~\citep{hong2023cogagent} is a 180 billion-parameter Vision-Language Model designed for GUI comprehension and navigation. 
(6) Qwen-VL ~\citep{Bai2023Qwen-VL} uses Qwen-7B as the initialization of the LLM, and Openclip ViT-bigG as the initialization of the visual encoder. And connects them with a randomly initialized cross-attention layer. We choose QWen-VL-Chat and QWen-VL-plus. %
(7) InternVL ~\citep{chen2023internvl} scales up the Vision Transformer (ViT) to 6B parameters and aligns it with LLM. There are multimodal models with varying sizes of language models within the InternVL series, including InternVL-Chat-vit-6B-Vicuna-7B, InternVL-Chat-vit-6B-Vicuna-13B, InternVL-Chat-vit-6B-Llama2-13B, and InternVL-Chat-V1.1. 
(8) GPT-4V \footnote{\url{https://openai.com/research/gpt-4v-system-card}} is a closed-source large multimodal model from OpenAI that accepts image and text inputs and emits text outputs, demonstrating human-level performance on a variety of professional and academic benchmarks. 
(9) Yi-VL-6B and Yi-VL-34B are our multimodal models, providing image understanding capabilities to large language models. In these models, Vit is the Openclip 224, and the language model is either Yi-6B-Chat or Yi-34B-Chat.

\begin{table}[ht]
\centering
\small
\resizebox{\linewidth}{!}{
\begin{tabular}{@{}lcccccccccc@{}}
\toprule
\multirow{2}{*}{Models} && \multicolumn{3}{c}{Question Type} & \multicolumn{1}{c}{} & \multicolumn{3}{c}{Question Difficulty} & & \multirow{2}{*}{Overall} \\
\cmidrule(lr){3-5} \cmidrule(lr){7-9}
                        && MC & FIB & T/F & \quad\quad & Easy & Medium & Hard && \\
\midrule
mPLUG-Owl2              && 22.9        & 7.0       & 53.8       & & 25.5  & 20.8  & 20.7  && 22.2 \\
VisCPM                  && 24.5        & 5.4       & 52.8       & & 26.8  & 21.1  & 20.1  && 22.7 \\
Chinese-LLaVA           && 25.6        & 5.4       & 52.7       & & 25.5  & 26.3  & 24.7  && 23.4 \\
Emu2                    && 28.4    & 2.9       & 51.4       & & 28.0  & 22.4  & 25.1  && 24.5 \\
CogAgent                && 25.9    & 5.9       & 51.9       & & 27.7  & 21.7  & 22.7  && 23.6 \\
InternVL-Chat-V1.1           && 36.7    & \highlightblue{14.4}       & \highlightblue{63.5}       & & 41.8  & 30.8  & 29.4  && 34.0 \\
Yi-VL-6B                && 40.8    & 11.7       & 54.9       & & 43.3  & 31.6  & 30.3  && 35.0 \\
Yi-VL-34B               && \highlightblue{42.5}    & 10.4       & 61.6       & & \highlightblue{45.6}  &\highlightblue{32.6}  & \highlightblue{31.9}  && \highlightblue{36.5} \\
Qwen-VL-Plus            && 42.9    & 15.7       & 49.4       & & 46.7  & 32.9  & 29.9  && 36.8 \\
GPT-4V                  && \textbf{46.4}       & \textbf{27.4}       & \textbf{66.0}       & & \textbf{51.5}  & \textbf{40.7}  & \textbf{38.3}  && \textbf{43.7} \\
\bottomrule
\end{tabular}
}
\caption{Combined result decomposition across question type and difficulty level. MC: multiple choice questions, FIB: fill in the blank questions, T/F: true or false questions.}
\label{tab:combined}
\vspace{-0.4cm}
\end{table}

\textbf{Text-only LLMs.} We evaluate the performance of LLMs (\textit{e.g.}, GPT4\footnote{\url{https://openai.com/research/gpt-4}}, Qwen-7B~\citep{bai2023qwen}, Deepseek-7B ~\citep{deepseek-ai2024deepseek}, Yi-6B\footnote{\url{https://huggingface.co/01-ai/Yi-6B-Chat}}) when dealing with plain text, and Baichuan-7B on multimodal data. 
In addition, to verify whether external image tools can enhance the performance of LLMs on multimodal data, we deploy OCR by Mathpix \footnote{\url{https://mathpix.com/}} processing images to convert certain image information into textual forms.

\textbf{Evaluation.} We build a systematic and rule-based evaluation pipeline. 
Robust regular expressions are built to extract answers from the model responses. 
Specifically, for multiple-choice questions, we directly use options as keywords to extract model responses, and take the one with the highest number of options in the model response as the answer. 
If there is no valid answer in the model's response, random selection is performed for multiple-choice questions. 
For the judgment and open-ended question answering questions, we utilize specific rules to extract some segments where the answer may occur, and then detect whether the answer occurs in them. 
We add random selection and frequent selection as baselines: the former randomly selects an option, while the latter selects the most frequent option for each specific topic in the validation set based on its frequency of occurrence in that topic. 
Finally, we adopt micro-average accuracy as the evaluation metric. The prompts we use are in Appendix \ref{sec:appendix-prompt}.

\vspace{-0.2cm}
\subsection{Results of \cmmmu}
In this section, we present the main result and detailed ablation studies of different LMMs' and their performances on the \cmmmu benchmark.
Results are shown in Tab. \ref{tab:performance}, \ref{tab:combined} and \ref{tab:image type}. We emphasize our key observations as follows:
\vspace{-0.2cm}
\begin{itemize}
    \item \textbf{\cmmmu is much more challenging than MMMU, while MMMU is already very challenging.} GPT-4V only achieves an accuracy of \textbf{41.7\%} while it achieves an accuracy of \textbf{55.7\%} in an English context. It reveals that existing cross-linguistic generalization is not good enough even for the most advanced closed-source LMMs.
    \item \textbf{The disparity between representative open-source models and GPT-4V is relatively smaller in a Chinese context compared to MMMU.} The disparity between Qwen-VL-Chat and GPT-4V on \cmmmu is \textbf{13.3\%} while the disparity between BLIP2-FLAN-T5-
XXL and GPT-4V on MMMU is \textbf{21.9\%}. Surprisingly, Yi-VL-34B even shortens the disparity between open-source bilingual LMMs and GPT-4V on \cmmmu to 7.5\%, meaning that \textit{\textbf{open-source bilingual LMMs hold a candle to GPT-4V in a Chinese context}}, which is a promising progress in the open-source community.
    \item \textbf{The key disparity between open-source LMMs and GPT-4V is the capacity to calculate and reason given complex conditions.} Notably, the performance disparity between Open-source LMMs and GPT-4V of Business, Science, and Tech \& Eng is larger compared to other disciplines. More questions require complex reasoning in the three disciplines, which reveals that open-source LMMs cannot calculate and reason given complex conditions.
    \item \textbf{The game of pursuing expert Chinese Multimodal Artificial General Intelligence (AGI) has just begun in the open-source community.} We point out that all the bilingual LMMs from the open-source community only achieve comparable accuracies with the frequent choice setting referring to MMMU, except recently released Qwen-VL-Chat, Yi-VL-6B, and Yi-VL-34B. These three LMMs trigger the first shot for the race of Chinese Multimodal AGI. 
\end{itemize}
\vspace{-0.2cm}
We conduct result decomposition across question difficulties, as shown in Tab.~\ref{tab:difficulty}.
Notably, there is a larger gap between the best open-source LMM, \textit{i.e.} Yi-VL-34B, and GPT-4V when facing the medium and hard questions.
This is further strong evidence of the observation that the key disparity between open-source LMMs and GPT-4V is the capacity to calculate and reason given complex conditions.

We conduct result decomposition across question types, as shown in Tab.~\ref{tab:question type}.
We notice that Qwen-VL-Plus does not well on True or False questions, indicating that Qwen-VL-Plus may not understand the prompt for answering True or False questions.
It might be a free lunch for Qwen-VL-Plus to improve its performance on \cmmmu.
We further point out that the disparity between Yi-VL Series, Qwen-VL-Plus, and GPT-4V is mainly because of their capacity difference for answering Multiple-choice questions.

\subsection{Error Analysis}

\begin{wrapfigure}{r}{0.5\textwidth}
  \centering
      \includegraphics[width=0.47\textwidth]{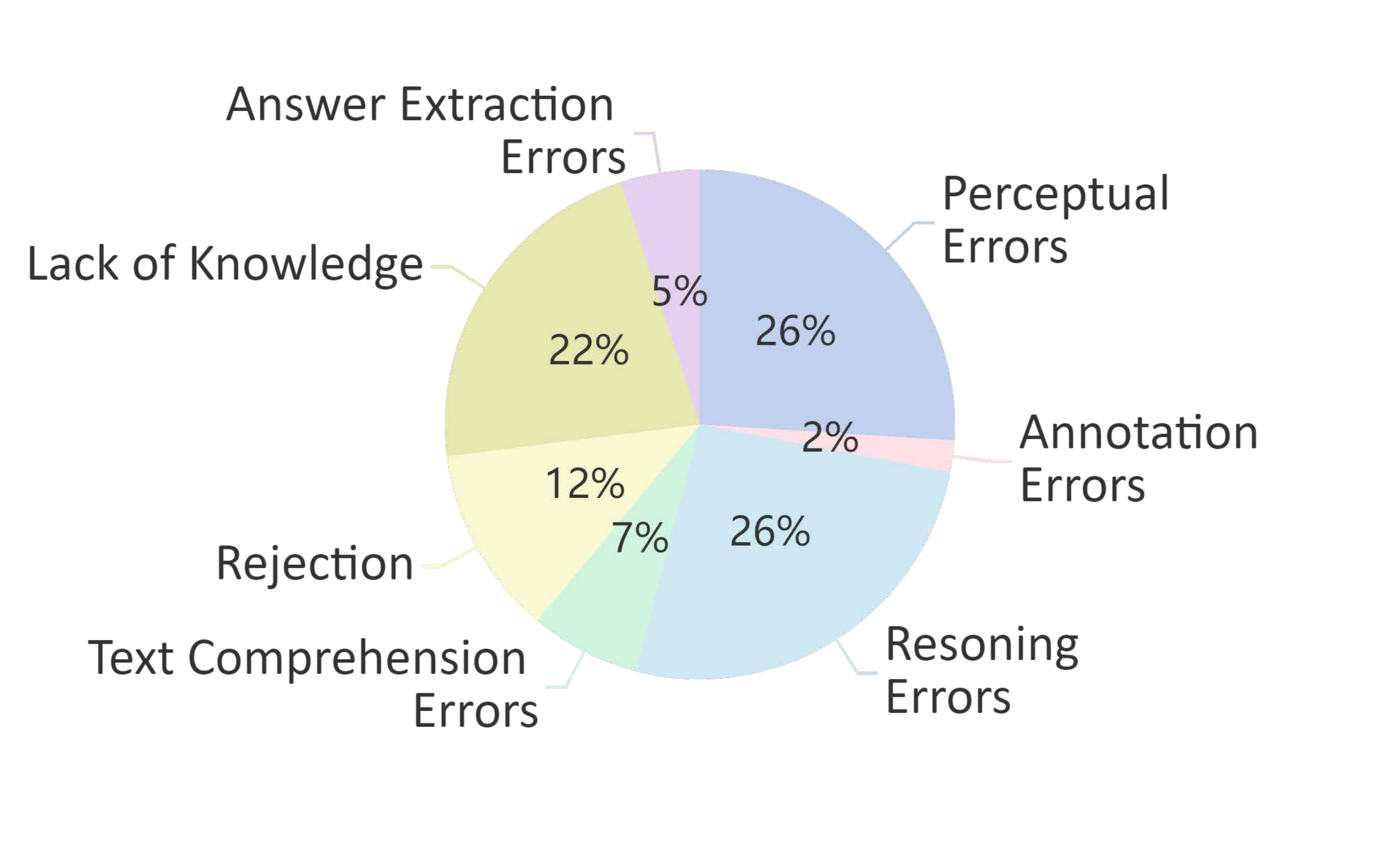}
  \caption{GPT-4V error response distribution.}
  \label{fig:figpie}
\end{wrapfigure}

This section carefully analyzes over 150 examples of GPT-4V's incorrect answers.
As shown in the error distribution  Fig.~\ref{fig:figpie}, several main types of errors are found, such as perceptual errors, lack of knowledge, reasoning errors, rejection to answer, and annotation errors. 
Analyzing these error types is key to understanding the capabilities and limitations of current LMMs, and can also guide future improvements in designing and training models. The 75 examples of correct responses and 150 examples of incorrect responses are detailed in Appendix \ref{sec:appendix-case study}, and the characteristics of each error type are described next.

\textbf{Perceptual Errors (26\%):} Perceptual errors are one of the primary reasons for the generation of erroneous examples by GPT-4V. On one hand, when the model fails to comprehend arrows and symbols in the image, misinterprets the sequence from top to bottom and left to right, it introduces deviations in the basic perception of the image, leading to incorrect responses. On the other hand, when the model encounters ambiguity in domain-specific knowledge, hidden meanings, or unclear formulas, it tends to exhibit perceptual errors specific to that domain. In such cases, GPT-4V tends to rely more on answering based on textual information (\textit{i.e.}, the question and options), prioritizing textual information over visual input, causing a bias in understanding multimodal data.

\textbf{Resoning Errors (26\%):} Reasoning Error is another major factor contributing to the generation of erroneous examples by GPT-4V. On the one hand, reasoning errors arise when the model receives incorrect information, often stemming from the perceptual errors mentioned earlier, such as in the illustration of Fig. \ref{fig:error_analysis_case_215}, where the model fails to perceive the hidden meaning of symbols, leading to erroneous inferences and outputs. On the other hand, even if the model correctly perceives the meaning conveyed by the image and text, errors in the reasoning process can occur when solving problems that require complex logical and mathematical reasoning. Typically, such errors result from the model's weaker logical and mathematical reasoning capabilities.

\textbf{Lack of Knowledge (22\%):} The lack of expertise is also one of the reasons why GPT-4V generates erroneous responses.  The example in Fig. \ref{fig:error_analysis_case_214} shows GPT-4V producing incorrect answers due to the lack of corresponding physics knowledge. Since \cmmmu is for evaluating expert AGI of LMMs, expert-level knowledge in different disciplines and subfields is required. So, injecting expert-level knowledge into LMMs is also one of the directions that can be worked towards AGI.

\textbf{Rejection (12\%):} The phenomenon of the model refusing to answer, resulting in incorrect responses, is also a common occurrence. Through analysis, we have identified several reasons for the model's refusal to answer: \textit{(i)} The model fails to perceive information from the image, and the textual information in the question is insufficient, causing the model to wait for more information. 
\textit{(ii)} Questions involving religious matters or personal real-life information lead the model to refrain from answering, adhering to human values. \textit{(iii)} When questions involve gender and subjective matters, the model avoids providing accurate responses.

\textbf{Other Errors:} The remaining errors are text comprehension errors (7\%), annotation errors (2\%), and answer extraction errors (5\%). These errors stem from factors like complex instruction comprehension, intricate text logic understanding, response generation limits, data annotation inaccuracies, and issues in extracting answer matches.
\vspace{-0.2cm}

\section{Conclusion}
\cmmmu represents a significant stride in developing AGI. 
The CMMMU's design is tailored to rigorously evaluating the latest LMMs,  and testing elementary perceptual skills, intricate logical reasoning, and profound expertise in specific domains. 
We reveal the disparity between the reasoning capacity of the most advanced bilingual LMMs in a Chinese context and an English context by comparing LMMs' performance on \cmmmu and MMMU.
Such an exhaustive assessment is pivotal for delineating the trajectory towards achieving AGI that parallels the proficiency of seasoned professionals in various fields.

\section*{Ethics Policy}
In developing the \cmmmu benchmark, we strictly adhere to ethical and legal standards, ensuring that our data collection and usage comply fully with pertinent ethical guidelines and legal regulations. Our dedication to promoting fairness, inclusivity, and diversity in our dataset is critical, aiming to reduce biases that might exacerbate societal disparities. We emphasize the importance of protecting privacy and intellectual property rights, highlighting our commitment to responsible and lawful data management. This methodology reflects our steadfast commitment to ethical integrity and legal compliance in the pursuit of advancing research in multimodal understanding and reasoning.

\section*{Limitations}
We recognize the inherent limitations of our study. Although the \cmmmu benchmark is comprehensive, it does not encompass the entire range of human knowledge and cognitive skills. It is primarily focused on college-level content, which, despite its complexity, covers only a portion of human expertise. Additionally, our evaluation metrics, despite their robustness, might not completely grasp the sophisticated understanding and reasoning abilities of advanced AI systems. These limitations highlight the iterative process of our research, motivating ongoing refinement and expansion of our benchmarks to more accurately reflect the complexities of human cognition and learning. Moreover, the challenges of conducting manual experiments with contracted experts for question-answering tasks have precluded us from offering a comprehensive score for human experiments to date. Nonetheless, we are contemplating the inclusion of such scores in future updates.

\bibliography{colm2024_conference}

\begin{thebibliography}{37}
\providecommand{\natexlab}[1]{#1}
\providecommand{\url}[1]{\texttt{#1}}
\expandafter\ifx\csname urlstyle\endcsname\relax
  \providecommand{\doi}[1]{doi: #1}\else
  \providecommand{\doi}{doi: \begingroup \urlstyle{rm}\Url}\fi

\bibitem[Achiam et~al.(2023)Achiam, Adler, Agarwal, Ahmad, Akkaya, Aleman, Almeida, Altenschmidt, Altman, Anadkat, et~al.]{achiam2023gpt4}
Josh Achiam, Steven Adler, Sandhini Agarwal, Lama Ahmad, Ilge Akkaya, Florencia~Leoni Aleman, Diogo Almeida, Janko Altenschmidt, Sam Altman, Shyamal Anadkat, et~al.
\newblock Gpt-4 technical report.
\newblock \emph{ArXiv preprint}, abs/2303.08774, 2023.
\newblock URL \url{https://arxiv.org/abs/2303.08774}.

\bibitem[Antol et~al.(2015{\natexlab{a}})Antol, Agrawal, Lu, Mitchell, Batra, Zitnick, and Parikh]{agrawal2015vqa}
Stanislaw Antol, Aishwarya Agrawal, Jiasen Lu, Margaret Mitchell, Dhruv Batra, C.~Lawrence Zitnick, and Devi Parikh.
\newblock {VQA:} visual question answering.
\newblock In \emph{2015 {IEEE} International Conference on Computer Vision, {ICCV} 2015, Santiago, Chile, December 7-13, 2015}, pp.\  2425--2433. {IEEE} Computer Society, 2015{\natexlab{a}}.
\newblock \doi{10.1109/ICCV.2015.279}.
\newblock URL \url{https://doi.org/10.1109/ICCV.2015.279}.

\bibitem[Antol et~al.(2015{\natexlab{b}})Antol, Agrawal, Lu, Mitchell, Batra, Zitnick, and Parikh]{antol2015vqa}
Stanislaw Antol, Aishwarya Agrawal, Jiasen Lu, Margaret Mitchell, Dhruv Batra, C.~Lawrence Zitnick, and Devi Parikh.
\newblock {VQA:} visual question answering.
\newblock In \emph{2015 {IEEE} International Conference on Computer Vision, {ICCV} 2015, Santiago, Chile, December 7-13, 2015}, pp.\  2425--2433. {IEEE} Computer Society, 2015{\natexlab{b}}.
\newblock \doi{10.1109/ICCV.2015.279}.
\newblock URL \url{https://doi.org/10.1109/ICCV.2015.279}.

\bibitem[Bai et~al.(2023{\natexlab{a}})Bai, Bai, Chu, Cui, Dang, Deng, Fan, Ge, Han, Huang, et~al.]{bai2023qwen}
Jinze Bai, Shuai Bai, Yunfei Chu, Zeyu Cui, Kai Dang, Xiaodong Deng, Yang Fan, Wenbin Ge, Yu~Han, Fei Huang, et~al.
\newblock Qwen technical report.
\newblock \emph{ArXiv preprint}, abs/2309.16609, 2023{\natexlab{a}}.
\newblock URL \url{https://arxiv.org/abs/2309.16609}.

\bibitem[Bai et~al.(2023{\natexlab{b}})Bai, Bai, Yang, Wang, Tan, Wang, Lin, Zhou, and Zhou]{Bai2023Qwen-VL}
Jinze Bai, Shuai Bai, Shusheng Yang, Shijie Wang, Sinan Tan, Peng Wang, Junyang Lin, Chang Zhou, and Jingren Zhou.
\newblock Qwen-vl: A versatile vision-language model for understanding, localization, text reading, and beyond.
\newblock \emph{ArXiv preprint}, abs/2308.12966, 2023{\natexlab{b}}.
\newblock URL \url{https://arxiv.org/abs/2308.12966}.

\bibitem[Chen et~al.(2023)Chen, Wu, Wang, Su, Chen, Xing, Zhong, Zhang, Zhu, Lu, Li, Luo, Lu, Qiao, and Dai]{chen2023internvl}
Zhe Chen, Jiannan Wu, Wenhai Wang, Weijie Su, Guo Chen, Sen Xing, Muyan Zhong, Qinglong Zhang, Xizhou Zhu, Lewei Lu, Bin Li, Ping Luo, Tong Lu, Yu~Qiao, and Jifeng Dai.
\newblock Internvl: Scaling up vision foundation models and aligning for generic visual-linguistic tasks.
\newblock \emph{ArXiv preprint}, abs/2312.14238, 2023.
\newblock URL \url{https://arxiv.org/abs/2312.14238}.

\bibitem[Cui et~al.(2023)Cui, Yang, and Yao]{chinese-llama-alpaca}
Yiming Cui, Ziqing Yang, and Xin Yao.
\newblock Efficient and effective text encoding for chinese llama and alpaca.
\newblock \emph{ArXiv preprint}, abs/2304.08177, 2023.
\newblock URL \url{https://arxiv.org/abs/2304.08177}.

\bibitem[DeepSeek-AI et~al.(2024)DeepSeek-AI, :, Bi, Chen, Chen, Chen, Dai, Deng, Ding, Dong, Du, Fu, Gao, Gao, Gao, Ge, Guan, Guo, Guo, Hao, Hao, He, Hu, Huang, Li, Li, Li, Li, Li, Liang, Lin, Liu, Liu, Liu, Liu, Liu, Liu, Lu, Lu, Luo, Ma, Nie, Pei, Piao, Qiu, Qu, Ren, Ren, Ruan, Sha, Shao, Song, Su, Sun, Sun, Tang, Wang, Wang, Wang, Wang, Wang, Wu, Wu, Xie, Xie, Xie, Xiong, Xu, Xu, Xu, Yang, You, Yu, Yu, Zhang, Zhang, Zhang, Zhang, Zhang, Zhang, Zhang, Zhang, Zhao, Zhao, Zhou, Zhou, Zhu, and Zou]{deepseek-ai2024deepseek}
DeepSeek-AI, :, Xiao Bi, Deli Chen, Guanting Chen, Shanhuang Chen, Damai Dai, Chengqi Deng, Honghui Ding, Kai Dong, Qiushi Du, Zhe Fu, Huazuo Gao, Kaige Gao, Wenjun Gao, Ruiqi Ge, Kang Guan, Daya Guo, Jianzhong Guo, Guangbo Hao, Zhewen Hao, Ying He, Wenjie Hu, Panpan Huang, Erhang Li, Guowei Li, Jiashi Li, Yao Li, Y.~K. Li, Wenfeng Liang, Fangyun Lin, A.~X. Liu, Bo~Liu, Wen Liu, Xiaodong Liu, Xin Liu, Yiyuan Liu, Haoyu Lu, Shanghao Lu, Fuli Luo, Shirong Ma, Xiaotao Nie, Tian Pei, Yishi Piao, Junjie Qiu, Hui Qu, Tongzheng Ren, Zehui Ren, Chong Ruan, Zhangli Sha, Zhihong Shao, Junxiao Song, Xuecheng Su, Jingxiang Sun, Yaofeng Sun, Minghui Tang, Bingxuan Wang, Peiyi Wang, Shiyu Wang, Yaohui Wang, Yongji Wang, Tong Wu, Y.~Wu, Xin Xie, Zhenda Xie, Ziwei Xie, Yiliang Xiong, Hanwei Xu, R.~X. Xu, Yanhong Xu, Dejian Yang, Yuxiang You, Shuiping Yu, Xingkai Yu, B.~Zhang, Haowei Zhang, Lecong Zhang, Liyue Zhang, Mingchuan Zhang, Minghua Zhang, Wentao Zhang, Yichao Zhang, Chenggang Zhao, Yao Zhao, Shangyan Zhou, Shunfeng
  Zhou, Qihao Zhu, and Yuheng Zou.
\newblock Deepseek llm: Scaling open-source language models with longtermism.
\newblock \emph{arXiv preprint arXiv: 2401.02954}, 2024.

\bibitem[Deng et~al.(2023)Deng, Gu, Zheng, Chen, Stevens, Wang, Sun, and Su]{deng2023mind2web}
Xiang Deng, Yu~Gu, Boyuan Zheng, Shijie Chen, Samuel Stevens, Boshi Wang, Huan Sun, and Yu~Su.
\newblock Mind2web: Towards a generalist agent for the web.
\newblock \emph{ArXiv preprint}, abs/2306.06070, 2023.
\newblock URL \url{https://arxiv.org/abs/2306.06070}.

\bibitem[Ding et~al.(2021)Ding, Yang, Hong, Zheng, Zhou, Yin, Lin, Zou, Shao, Yang, and Tang]{ding2021cogview}
Ming Ding, Zhuoyi Yang, Wenyi Hong, Wendi Zheng, Chang Zhou, Da~Yin, Junyang Lin, Xu~Zou, Zhou Shao, Hongxia Yang, and Jie Tang.
\newblock Cogview: Mastering text-to-image generation via transformers.
\newblock In Marc'Aurelio Ranzato, Alina Beygelzimer, Yann~N. Dauphin, Percy Liang, and Jennifer~Wortman Vaughan (eds.), \emph{Advances in Neural Information Processing Systems 34: Annual Conference on Neural Information Processing Systems 2021, NeurIPS 2021, December 6-14, 2021, virtual}, pp.\  19822--19835, 2021.
\newblock URL \url{https://proceedings.neurips.cc/paper/2021/hash/a4d92e2cd541fca87e4620aba658316d-Abstract.html}.

\bibitem[Du et~al.(2022)Du, Qian, Liu, Ding, Qiu, Yang, and Tang]{du2022glm}
Zhengxiao Du, Yujie Qian, Xiao Liu, Ming Ding, Jiezhong Qiu, Zhilin Yang, and Jie Tang.
\newblock {GLM}: General language model pretraining with autoregressive blank infilling.
\newblock In \emph{Proceedings of the 60th Annual Meeting of the Association for Computational Linguistics (Volume 1: Long Papers)}, pp.\  320--335, Dublin, Ireland, 2022. Association for Computational Linguistics.
\newblock \doi{10.18653/v1/2022.acl-long.26}.
\newblock URL \url{https://aclanthology.org/2022.acl-long.26}.

\bibitem[Fu et~al.(2023)Fu, Chen, Shen, Qin, Zhang, Lin, Yang, Zheng, Li, Sun, et~al.]{fu2023mme}
Chaoyou Fu, Peixian Chen, Yunhang Shen, Yulei Qin, Mengdan Zhang, Xu~Lin, Jinrui Yang, Xiawu Zheng, Ke~Li, Xing Sun, et~al.
\newblock Mme: A comprehensive evaluation benchmark for multimodal large language models.
\newblock \emph{ArXiv preprint}, abs/2306.13394, 2023.
\newblock URL \url{https://arxiv.org/abs/2306.13394}.

\bibitem[Ganz et~al.(2023)Ganz, Nuriel, Aberdam, Kittenplon, Mazor, and Litman]{ganz2023towards}
Roy Ganz, Oren Nuriel, Aviad Aberdam, Yair Kittenplon, Shai Mazor, and Ron Litman.
\newblock Towards models that can see and read.
\newblock \emph{IEEE International Conference on Computer Vision}, 2023.
\newblock \doi{10.1109/ICCV51070.2023.01985}.

\bibitem[Gurari et~al.(2018)Gurari, Li, Stangl, Guo, Lin, Grauman, Luo, and Bigham]{gurari2018vizwiz}
Danna Gurari, Qing Li, Abigale~J. Stangl, Anhong Guo, Chi Lin, Kristen Grauman, Jiebo Luo, and Jeffrey~P. Bigham.
\newblock Vizwiz grand challenge: Answering visual questions from blind people.
\newblock In \emph{2018 {IEEE} Conference on Computer Vision and Pattern Recognition, {CVPR} 2018, Salt Lake City, UT, USA, June 18-22, 2018}, pp.\  3608--3617. {IEEE} Computer Society, 2018.
\newblock \doi{10.1109/CVPR.2018.00380}.
\newblock URL \url{http://openaccess.thecvf.com/content\_cvpr\_2018/html/Gurari\_VizWiz\_Grand\_Challenge\_CVPR\_2018\_paper.html}.

\bibitem[Hendrycks et~al.(2021)Hendrycks, Burns, Basart, Zou, Mazeika, Song, and Steinhardt]{hendrycks2020measuring}
Dan Hendrycks, Collin Burns, Steven Basart, Andy Zou, Mantas Mazeika, Dawn Song, and Jacob Steinhardt.
\newblock Measuring massive multitask language understanding.
\newblock In \emph{9th International Conference on Learning Representations, {ICLR} 2021, Virtual Event, Austria, May 3-7, 2021}. OpenReview.net, 2021.
\newblock URL \url{https://openreview.net/forum?id=d7KBjmI3GmQ}.

\bibitem[Hong et~al.(2023)Hong, Wang, Lv, Xu, Yu, Ji, Wang, Wang, Dong, Ding, and Tang]{hong2023cogagent}
Wenyi Hong, Weihan Wang, Qingsong Lv, Jiazheng Xu, Wenmeng Yu, Junhui Ji, Yan Wang, Zihan Wang, Yuxiao Dong, Ming Ding, and Jie Tang.
\newblock Cogagent: A visual language model for gui agents, 2023.

\bibitem[Hu et~al.(2023)Hu, Yao, Wang, Wang, Pan, Chen, Yu, Wu, Zhao, Zhang, Han, Lin, Xue, Li, Liu, and Sun]{hu2023viscpm}
Jinyi Hu, Yuan Yao, Chongyi Wang, Shan Wang, Yinxu Pan, Qianyu Chen, Tianyu Yu, Hanghao Wu, Yue Zhao, Haoye Zhang, Xu~Han, Yankai Lin, Jiao Xue, Dahai Li, Zhiyuan Liu, and Maosong Sun.
\newblock Large multilingual models pivot zero-shot multimodal learning across languages.
\newblock 2023.

\bibitem[Hudson \& Manning(2019)Hudson and Manning]{hudson2019gqa}
Drew~A. Hudson and Christopher~D. Manning.
\newblock {GQA:} {A} new dataset for real-world visual reasoning and compositional question answering.
\newblock In \emph{{IEEE} Conference on Computer Vision and Pattern Recognition, {CVPR} 2019, Long Beach, CA, USA, June 16-20, 2019}, pp.\  6700--6709. Computer Vision Foundation / {IEEE}, 2019.
\newblock \doi{10.1109/CVPR.2019.00686}.
\newblock URL \url{http://openaccess.thecvf.com/content\_CVPR\_2019/html/Hudson\_GQA\_A\_New\_Dataset\_for\_Real-World\_Visual\_Reasoning\_and\_Compositional\_CVPR\_2019\_paper.html}.

\bibitem[Li et~al.(2023)Li, Wang, Wang, Ge, Ge, and Shan]{li2023seedbench}
Bohao Li, Rui Wang, Guangzhi Wang, Yuying Ge, Yixiao Ge, and Ying Shan.
\newblock Seed-bench: Benchmarking multimodal llms with generative comprehension.
\newblock \emph{arXiv preprint arXiv: 2307.16125}, 2023.

\bibitem[Lin et~al.(2014)Lin, Maire, Belongie, Hays, Perona, Ramanan, Doll{\'a}r, and Zitnick]{lin2014mscoco}
Tsung-Yi Lin, Michael Maire, Serge Belongie, James Hays, Pietro Perona, Deva Ramanan, Piotr Doll{\'a}r, and C~Lawrence Zitnick.
\newblock Microsoft coco: Common objects in context.
\newblock In \emph{Computer Vision--ECCV 2014: 13th European Conference, Zurich, Switzerland, September 6-12, 2014, Proceedings, Part V 13}, pp.\  740--755. Springer, 2014.

\bibitem[LinkSoul-AI(2023)]{linksoulai2023chinesellava}
LinkSoul-AI.
\newblock Chinese llava.
\newblock \url{https://github.com/LinkSoul-AI/Chinese-LLaVA}, 2023.

\bibitem[Liu et~al.(2023)Liu, Duan, Zhang, Li, Zhang, Zhao, Yuan, Wang, He, Liu, et~al.]{liu2023mmbench}
Yuan Liu, Haodong Duan, Yuanhan Zhang, Bo~Li, Songyang Zhang, Wangbo Zhao, Yike Yuan, Jiaqi Wang, Conghui He, Ziwei Liu, et~al.
\newblock Mmbench: Is your multi-modal model an all-around player?
\newblock \emph{ArXiv preprint}, abs/2307.06281, 2023.
\newblock URL \url{https://arxiv.org/abs/2307.06281}.

\bibitem[Lu et~al.(2022)Lu, Mishra, Xia, Qiu, Chang, Zhu, Tafjord, Clark, and Kalyan]{lu2022learn}
Pan Lu, Swaroop Mishra, Tanglin Xia, Liang Qiu, Kai-Wei Chang, Song-Chun Zhu, Oyvind Tafjord, Peter Clark, and Ashwin Kalyan.
\newblock Learn to explain: Multimodal reasoning via thought chains for science question answering.
\newblock \emph{Advances in Neural Information Processing Systems}, 35:\penalty0 2507--2521, 2022.

\bibitem[Lu et~al.(2023)Lu, Bansal, Xia, Liu, Li, Hajishirzi, Cheng, Chang, Galley, and Gao]{lu2023mathvista}
Pan Lu, Hritik Bansal, Tony Xia, Jiacheng Liu, Chunyuan Li, Hannaneh Hajishirzi, Hao Cheng, Kai-Wei Chang, Michel Galley, and Jianfeng Gao.
\newblock Mathvista: Evaluating mathematical reasoning of foundation models in visual contexts.
\newblock \emph{ArXiv preprint}, abs/2310.02255, 2023.
\newblock URL \url{https://arxiv.org/abs/2310.02255}.

\bibitem[Marino et~al.(2019)Marino, Rastegari, Farhadi, and Mottaghi]{marino2019okvqa}
Kenneth Marino, Mohammad Rastegari, Ali Farhadi, and Roozbeh Mottaghi.
\newblock {OK-VQA:} {A} visual question answering benchmark requiring external knowledge.
\newblock In \emph{{IEEE} Conference on Computer Vision and Pattern Recognition, {CVPR} 2019, Long Beach, CA, USA, June 16-20, 2019}, pp.\  3195--3204. Computer Vision Foundation / {IEEE}, 2019.
\newblock \doi{10.1109/CVPR.2019.00331}.
\newblock URL \url{http://openaccess.thecvf.com/content\_CVPR\_2019/html/Marino\_OK-VQA\_A\_Visual\_Question\_Answering\_Benchmark\_Requiring\_External\_Knowledge\_CVPR\_2019\_paper.html}.

\bibitem[Plummer et~al.(2015)Plummer, Wang, Cervantes, Caicedo, Hockenmaier, and Lazebnik]{plummer2015flickr30k}
Bryan~A. Plummer, Liwei Wang, Chris~M. Cervantes, Juan~C. Caicedo, Julia Hockenmaier, and Svetlana Lazebnik.
\newblock Flickr30k entities: Collecting region-to-phrase correspondences for richer image-to-sentence models.
\newblock In \emph{2015 {IEEE} International Conference on Computer Vision, {ICCV} 2015, Santiago, Chile, December 7-13, 2015}, pp.\  2641--2649. {IEEE} Computer Society, 2015.
\newblock \doi{10.1109/ICCV.2015.303}.
\newblock URL \url{https://doi.org/10.1109/ICCV.2015.303}.

\bibitem[Schwenk et~al.(2022)Schwenk, Khandelwal, Clark, Marino, and Mottaghi]{schwenk2022okvqa}
Dustin Schwenk, Apoorv Khandelwal, Christopher Clark, Kenneth Marino, and Roozbeh Mottaghi.
\newblock A-okvqa: A benchmark for visual question answering using world knowledge.
\newblock In \emph{European Conference on Computer Vision}, pp.\  146--162. Springer, 2022.

\bibitem[Sun et~al.(2023)Sun, Cui, Zhang, Zhang, Yu, Luo, Wang, Rao, Liu, Huang, and Wang]{sun2023Emu2}
Quan Sun, Yufeng Cui, Xiaosong Zhang, Fan Zhang, Qiying Yu, Zhengxiong Luo, Yueze Wang, Yongming Rao, Jingjing Liu, Tiejun Huang, and Xinlong Wang.
\newblock Generative multimodal models are in-context learners.
\newblock \emph{ArXiv preprint}, abs/2312.13286, 2023.
\newblock URL \url{https://arxiv.org/abs/2312.13286}.

\bibitem[Vinyals et~al.(2014)Vinyals, Toshev, Bengio, and Erhan]{vinyals2014show}
Oriol Vinyals, Alexander Toshev, Samy Bengio, and Dumitru Erhan.
\newblock Show and tell: A neural image caption generator. corr abs/1411.4555 (2014).
\newblock \emph{arXiv preprint arXiv:1411.4555}, 2014.

\bibitem[Wang et~al.(2023)Wang, Lv, Yu, Hong, Qi, Wang, Ji, Yang, Zhao, Song, Xu, Xu, Li, Dong, Ding, and Tang]{wang2023cogvlm}
Weihan Wang, Qingsong Lv, Wenmeng Yu, Wenyi Hong, Ji~Qi, Yan Wang, Junhui Ji, Zhuoyi Yang, Lei Zhao, Xixuan Song, Jiazheng Xu, Bin Xu, Juanzi Li, Yuxiao Dong, Ming Ding, and Jie Tang.
\newblock Cogvlm: Visual expert for pretrained language models, 2023.

\bibitem[Wei et~al.(2023)Wei, Chen, Chen, Hu, Zhang, Fu, Ritter, and Chen]{wei2023uniir}
Cong Wei, Yang Chen, Haonan Chen, Hexiang Hu, Ge~Zhang, Jie Fu, Alan Ritter, and Wenhu Chen.
\newblock Uniir: Training and benchmarking universal multimodal information retrievers.
\newblock \emph{ArXiv preprint}, abs/2311.17136, 2023.
\newblock URL \url{https://arxiv.org/abs/2311.17136}.

\bibitem[Wu et~al.(2024)Wu, Li, Zhu, Zhang, Liang, Ma, Xiao, Zhang, Yang, Chen, Huang, Moubayed, Fu, and Lin]{wu2024scimmir}
Siwei Wu, Yizhi Li, Kang Zhu, Ge~Zhang, Yiming Liang, Kaijing Ma, Chenghao Xiao, Haoran Zhang, Bohao Yang, Wenhu Chen, Wenhao Huang, Noura~Al Moubayed, Jie Fu, and Chenghua Lin.
\newblock Scimmir: Benchmarking scientific multi-modal information retrieval, 2024.

\bibitem[Ye et~al.(2023)Ye, Xu, Ye, Yan, Liu, Qian, Zhang, Huang, and Zhou]{ye2023mplugowl2}
Qinghao Ye, Haiyang Xu, Jiabo Ye, Ming Yan, Haowei Liu, Qi~Qian, Ji~Zhang, Fei Huang, and Jingren Zhou.
\newblock mplug-owl2: Revolutionizing multi-modal large language model with modality collaboration.
\newblock \emph{ArXiv preprint}, abs/2311.04257, 2023.
\newblock URL \url{https://arxiv.org/abs/2311.04257}.

\bibitem[Yu et~al.(2023)Yu, Yang, Li, Wang, Lin, Liu, Wang, and Wang]{yu2023mmvet}
Weihao Yu, Zhengyuan Yang, Linjie Li, Jianfeng Wang, Kevin Lin, Zicheng Liu, Xinchao Wang, and Lijuan Wang.
\newblock Mm-vet: Evaluating large multimodal models for integrated capabilities.
\newblock \emph{arXiv preprint arXiv: 2308.02490}, 2023.

\bibitem[Yue et~al.(2023)Yue, Ni, Zhang, Zheng, Liu, Zhang, Stevens, Jiang, Ren, Sun, et~al.]{yue2023mmmu}
Xiang Yue, Yuansheng Ni, Kai Zhang, Tianyu Zheng, Ruoqi Liu, Ge~Zhang, Samuel Stevens, Dongfu Jiang, Weiming Ren, Yuxuan Sun, et~al.
\newblock Mmmu: A massive multi-discipline multimodal understanding and reasoning benchmark for expert agi.
\newblock \emph{ArXiv preprint}, abs/2311.16502, 2023.
\newblock URL \url{https://arxiv.org/abs/2311.16502}.

\bibitem[Zhang et~al.(2023)Zhang, Li, Zong, Ying, He, and Qiu]{zhang2023evaluating}
Xiaotian Zhang, Chunyang Li, Yi~Zong, Zhengyu Ying, Liang He, and Xipeng Qiu.
\newblock Evaluating the performance of large language models on gaokao benchmark.
\newblock \emph{ArXiv preprint}, abs/2305.12474, 2023.
\newblock URL \url{https://arxiv.org/abs/2305.12474}.

\bibitem[Zhong et~al.(2023)Zhong, Cui, Guo, Liang, Lu, Wang, Saied, Chen, and Duan]{zhong2023agieval}
Wanjun Zhong, Ruixiang Cui, Yiduo Guo, Yaobo Liang, Shuai Lu, Yanlin Wang, Amin Saied, Weizhu Chen, and Nan Duan.
\newblock Agieval: A human-centric benchmark for evaluating foundation models.
\newblock \emph{ArXiv preprint}, abs/2304.06364, 2023.
\newblock URL \url{https://arxiv.org/abs/2304.06364}.

\end{thebibliography}
\bibliographystyle{colm2024_conference}

\newpage
\appendix
\section{Appendix}

\captionsetup[figure]{labelformat=simple, labelsep=colon, name=Figure}
\renewcommand{\thefigure}{A\arabic{figure}}
\setcounter{figure}{0}
\captionsetup[table]{labelformat=simple, labelsep=colon, name=Table}
\renewcommand{\thetable}{A\arabic{table}}
\setcounter{table}{0}

\label{sec:appendix-prompt}
In experiments, the prompts we use and their corresponding question types are as follows:

\begin{CJK}{UTF8}{gbsn}
\textbf{Multiple-choice questions: }\textit{请回答以下多项选择题，并选出正确选项。这些题目可能包括单选和多选题型。如果所提供的信息不足以确定一个明确的答案，那么请根据可用的数据和你的判断来选择最可能正确的选项。} (Please answer the following multiple-choice questions and select the correct options. These questions may include both single-choice and multiple-choice formats. If the provided information is not sufficient to determine a definite answer, please choose the option that is most likely correct based on the available data and your judgment.)\footnote{The English version is not part of the input to the models.}

\textbf{True/False questions: }\textit{请回答以下判断题，并根据题目描述和所给的信息来判断问题中陈述的对错。如果信息不完整或不足以作出绝对判断，请运用你的逻辑推理和现有信息来做出最可能的判断。} (Please answer the following true/false questions and determine the correctness of the statements based on the question descriptions and the provided information. If the information is incomplete or insufficient for an absolute judgment, please use your logical reasoning and available information to make the most likely judgment.)

\textbf{Fill-in-the-blank questions: }\textit{请回答以下填空题，并根据题目的要求和所提供的信息来给出最恰当的答案。如果信息不足以确切回答，那么请依据现有的数据和你的推理能力来填写最合理的答案。} (Please answer the following fill-in-the-blank questions and provide the most appropriate answer based on the question requirements and the provided information. If the information is insufficient for an exact answer, please fill in the most reasonable response based on the available data and your reasoning abilities.)
\end{CJK}

\begin{figure*}[hb]
    \centering
    \includegraphics[width=1\textwidth]{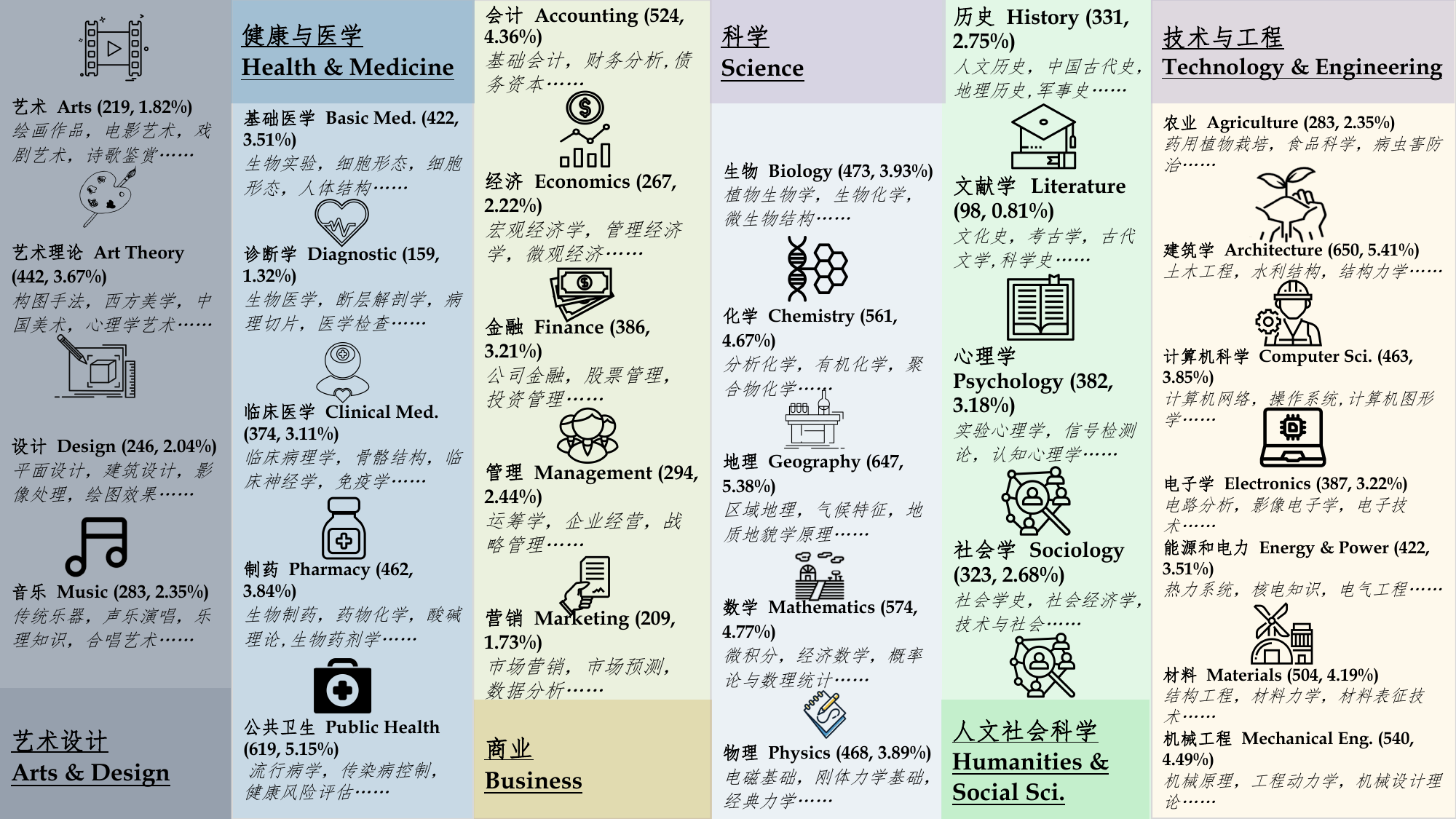}
    \caption{The proportion of 6 disciplines and 30 subjects in the \cmmmu. The multimodal samples in 30 subjects uniformly cover the relevant expert-level domain knowledge.}
    \label{fig:CMMMU-statics2}
\end{figure*}

Fig. \ref{fig:CMMMU-statics2} shows the proportion of 6 disciplines and 30 subjects in \cmmmu, and Tab. \ref{tab:image type} shows the result decomposition across image types.

\begin{table}[ht]
\centering
\resizebox{\linewidth}{!}{
\begin{tabular}{@{}lcccccccccccc@{}}
\toprule
\multirow{2}{*}{Models} & \multicolumn{1}{c}{Sketches} & \multicolumn{1}{c}{Table} & \multicolumn{1}{c}{Photos} & \multicolumn{1}{c}{Charts} & \multicolumn{1}{c}{\makecell{Chemical \\ Structures}} & \multicolumn{1}{c}{\makecell{Circuit \\ Diagram}} & \multicolumn{1}{c}{\makecell{Engineering \\ Diagram}} & \multicolumn{1}{c}{\makecell{Medical \\ Images}} & \multicolumn{1}{c}{\makecell{Microsc. \\ Images}} & \multicolumn{1}{c}{Overall}               \\
                        \midrule
mPLUG-Owl2              & 21.8        & 15.7        & 29.0        & 22.5        & 24.6        & 14.3        & 20.2        & 26.7        & 21.4       & 22.2              \\
VisCPM                  & 22.4        & 14.0        & 31.2        & 23.8        & 20.6        & 16.2        & 21.7        & 28.9        & 21.2       & 22.7              \\
Chinese-LLaVA           & 24.2        & 15.5        & 29.8        & 21.8        & 24.1        & 15.0        & 24.0        & 26.9        & 20.9       & 23.4              \\
Emu2                    & 25.6        & 15.8        & 30.3        & 22.9        & 27.3        & 16.6        & 26.1        & 26.9        & 21.2       & 24.5   \\
CogAgent                & 23.7        & 16.1        & 30.5        & 24.5        & 22.4        & 20.3        & 22.9        & 28.0        & 23.5       & 23.6   \\
InternVL-Chat-V1.1                     & 29.9        & 21.3        & 51.0        & 33.1        & 31.1        & 20.3        & 23.9        & 44.0        & 35.7       & 34.0   \\
Yi-VL-6B                    & 30.6        & \highlightblue{21.3}        & 52.6        & \highlightblue{35.1}        & 34.8        & 19.9        & \highlightblue{26.5}        & 42.3        & 36.0       & 35.0   \\
Yi-VL-34B                    & \highlightblue{31.2}        & 20.8        & \highlightblue{55.5}        & 35.0        & \highlightblue{34.9}        & \highlightblue{26.2}        & 22.3        & \highlightblue{\textbf{47.4}}        & \highlightblue{36.8}       & \highlightblue{36.5}   \\
\midrule
Qwen-VL-Plus                    & 30.2        & 24.2        & 53.1        & 41.1        & 35.1        & 23.0        & \textbf{27.8}        & 43.4        & \textbf{37.6}       & 36.8   \\
GPT-4V                  & \textbf{35.7}        & \textbf{38.8}        & \textbf{56.8}        & \textbf{44.9}        & \textbf{44.0}        & \textbf{37.7}        & 25.1        & 44.3        & 35.2       & \textbf{43.7}              \\ \bottomrule
\end{tabular}
}
\caption{Result decomposition across images type. \textbf{bold results} in LMMs indicate the best results for all models, and the \highlightblue{Blue results} indicate the best results among the open-source models.}
\label{tab:image type}
\end{table}

\clearpage
\newpage
\section{Case Study}
\captionsetup[figure]{labelformat=simple, labelsep=colon, name=Figure}
\renewcommand{\thefigure}{B\arabic{figure}}
\setcounter{figure}{0}
\label{sec:appendix-case study}
The appendix is our sample analysis of GPT-4V, including an analysis of 150 error examples and 75 correct examples.

\hypertarget{listofcasestudyfigures}{}
\listofcasestudyfigures

\casestudyfigure{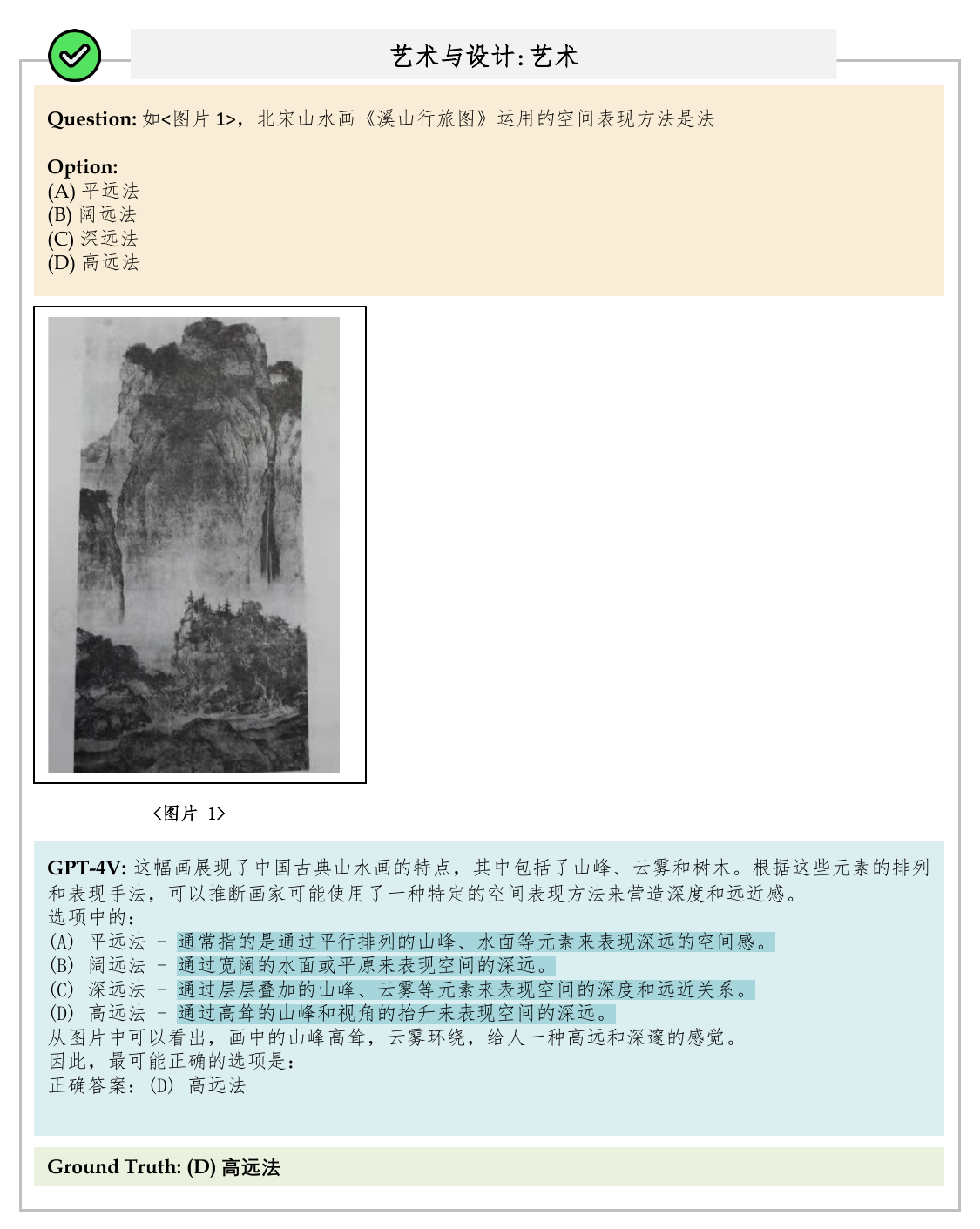}{Art 1: Correct Case}{A sample correct case of Art and Design (subcategory: Art).}{fig:error_analysis_case_31}

\casestudyfigure{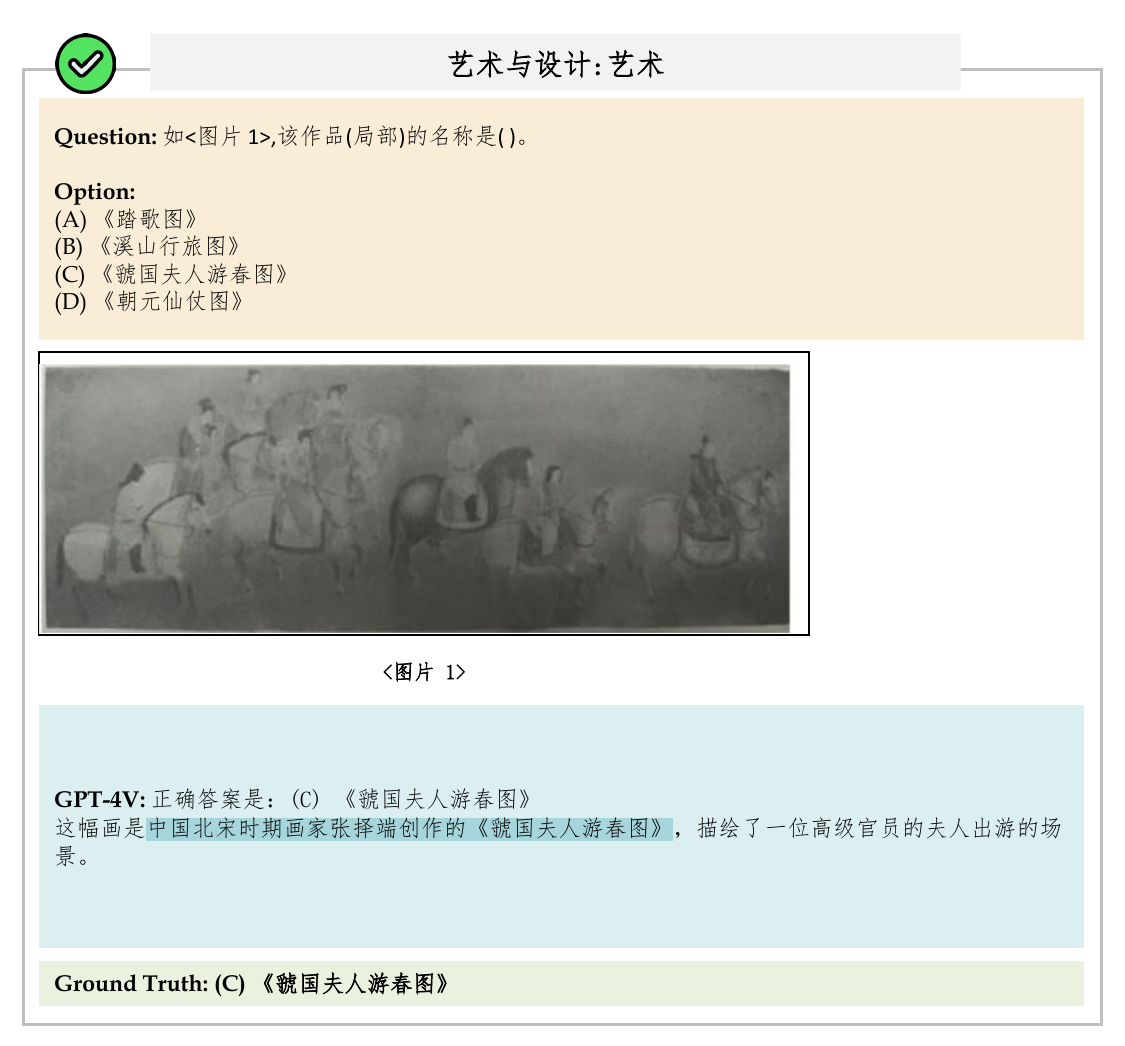}{Art 2: Correct Case}{A sample correct case of Art and Design (subcategory: Art).}{fig:error_analysis_case_32}

\casestudyfigure{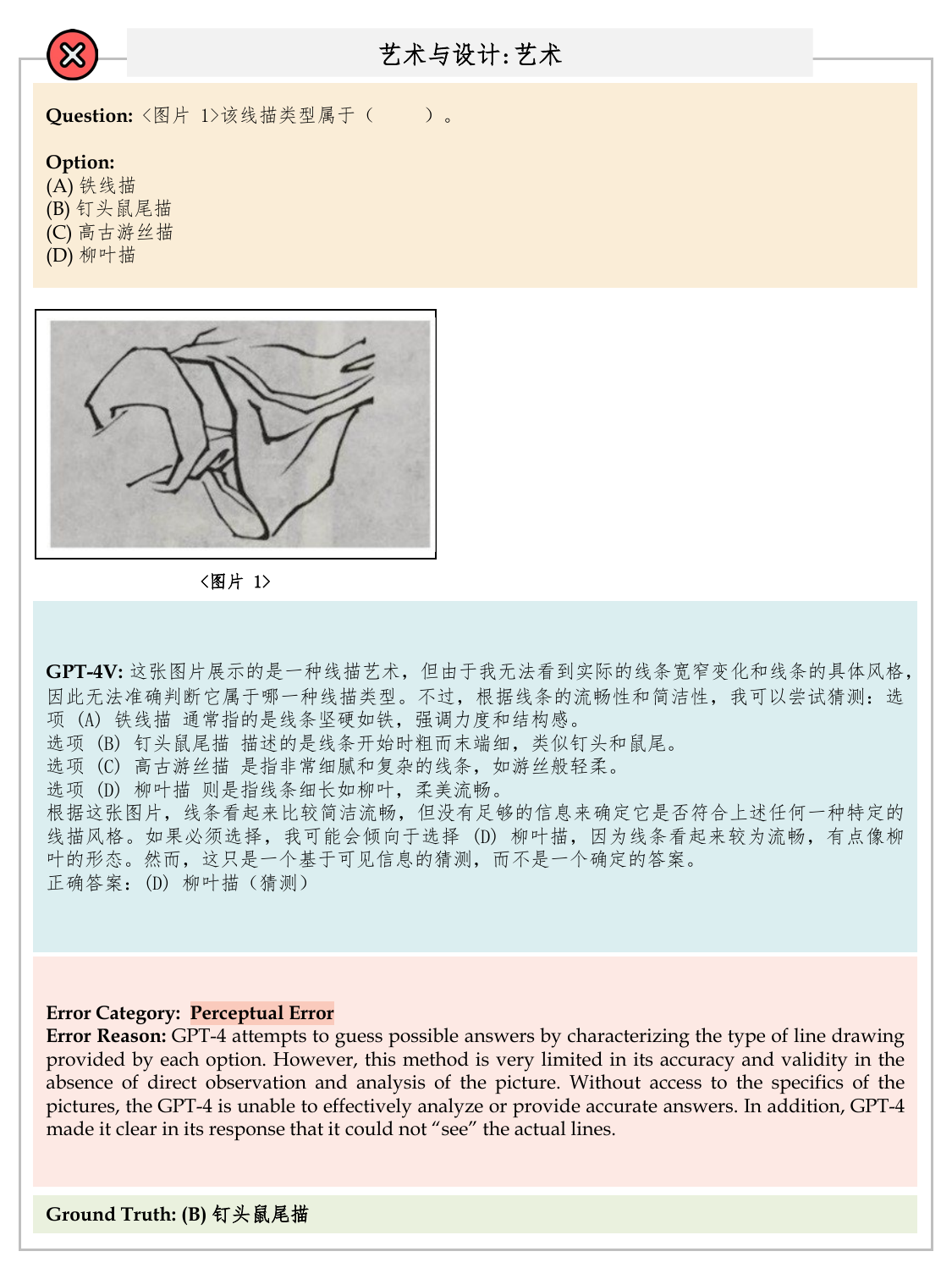}{Art 3: Perceptual Error}{A sample error case of Art and Design (subcategory: Art).}{fig:error_analysis_case_154}

\casestudyfigure{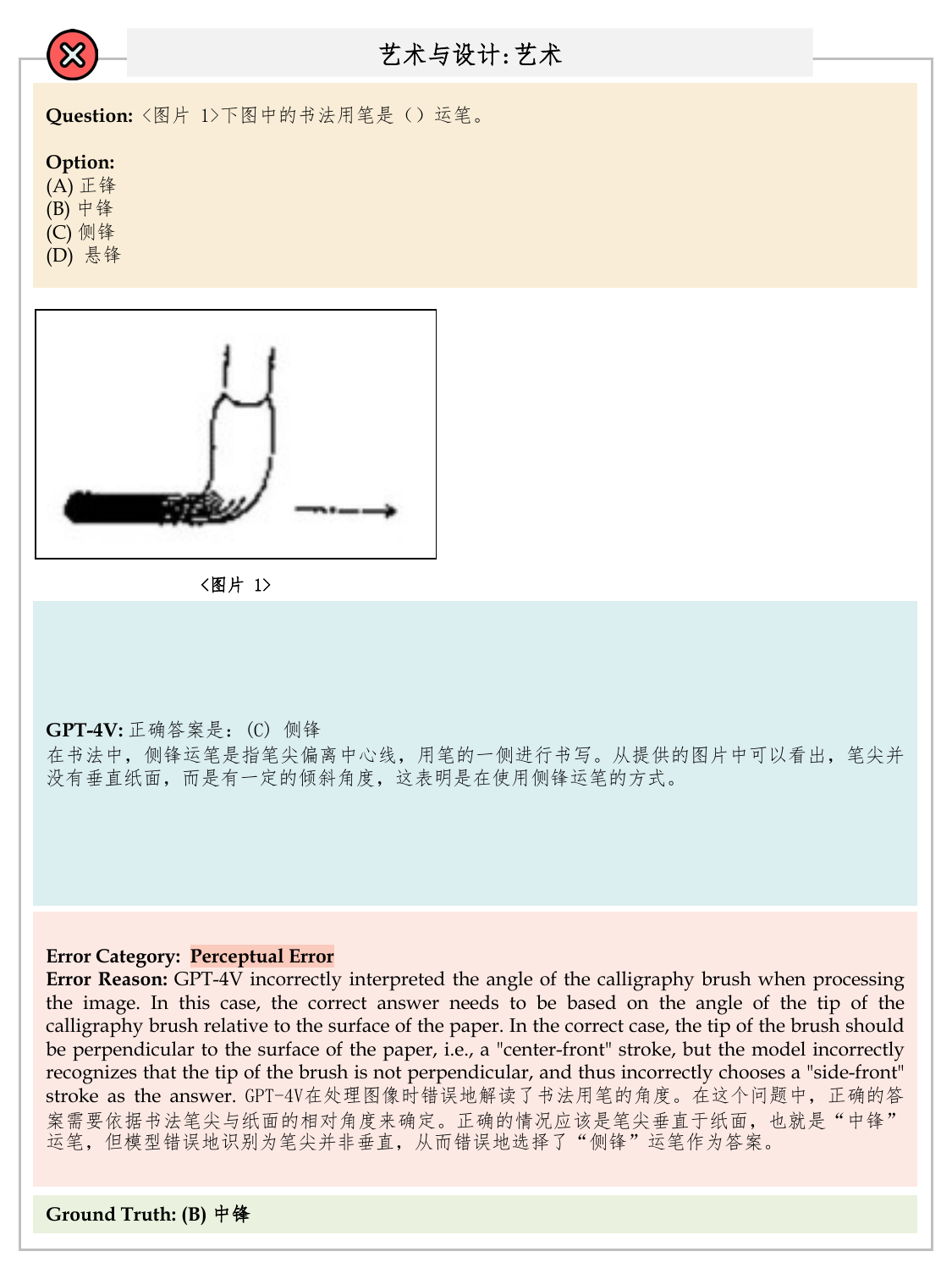}{Art 4: Perceptual Error}{A sample error case of Art and Design (subcategory: Art).}{fig:error_analysis_case_155}

\casestudyfigure{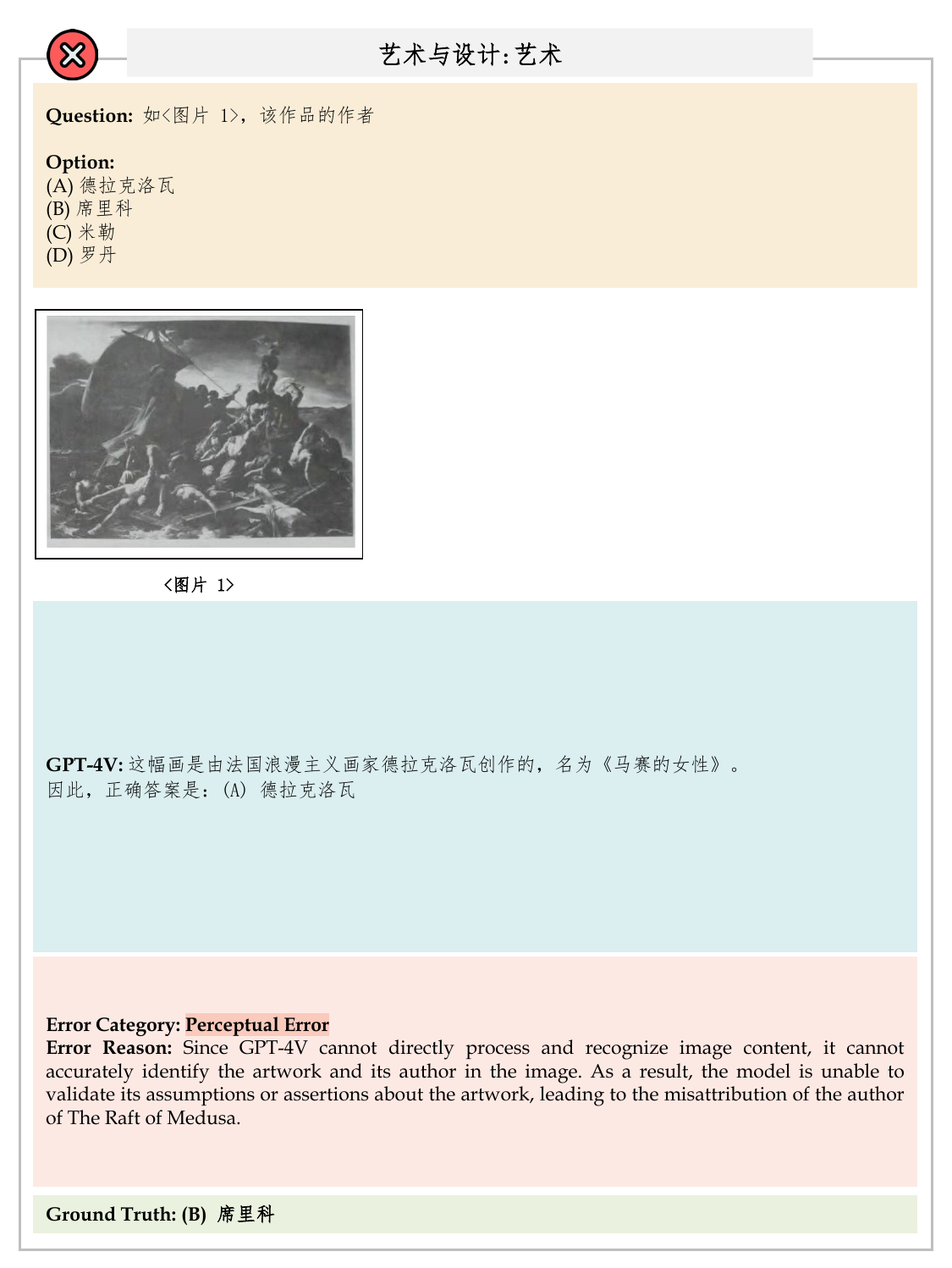}{Art 5: Perceptual Error}{A sample error case of Art and Design (subcategory: Art).}{fig:error_analysis_case_157}

\casestudyfigure{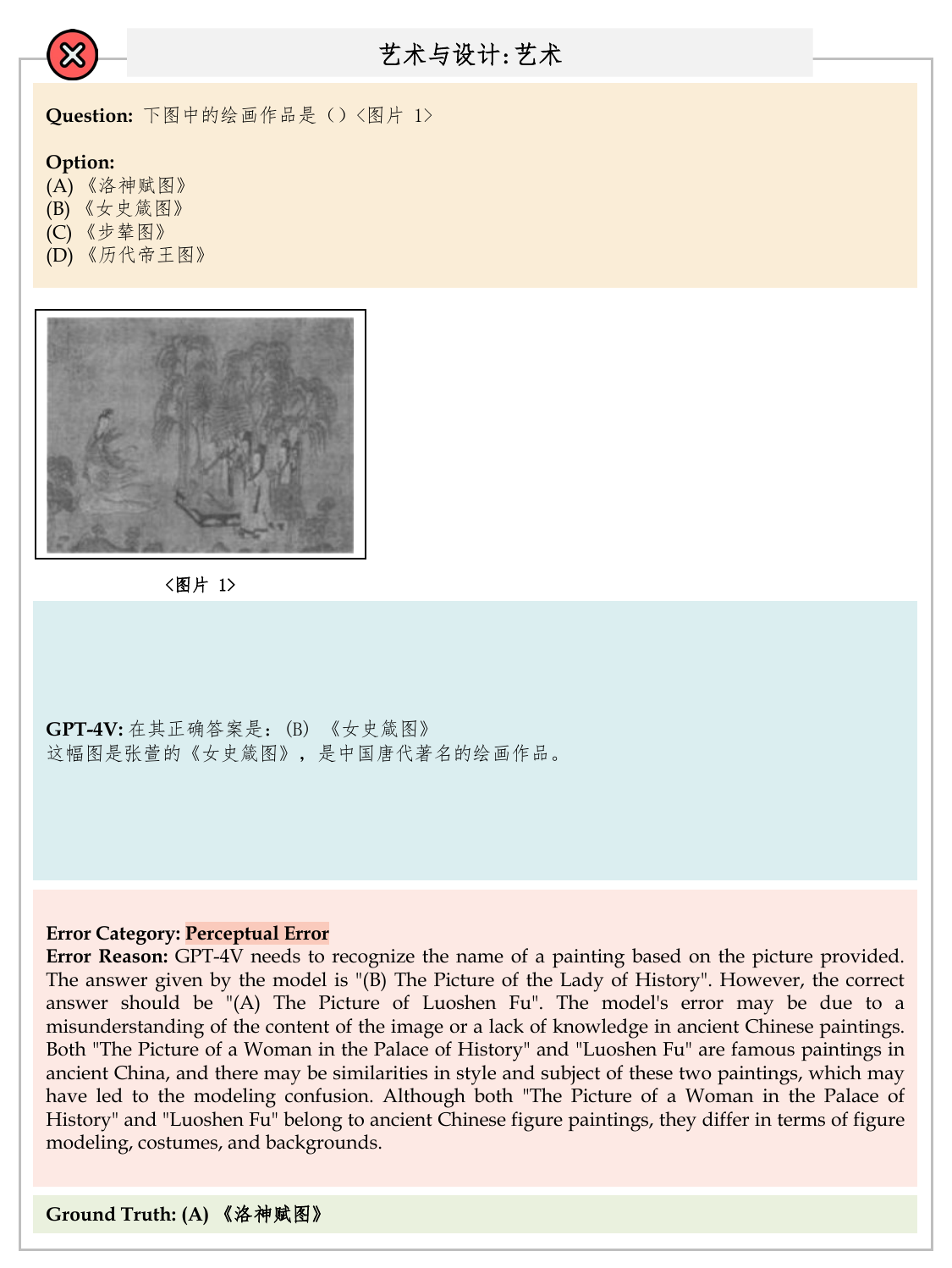}{Art 6: Perceptual Error}{A sample error case of Art and Design (subcategory: Art).}{fig:error_analysis_case_159}

\casestudyfigure{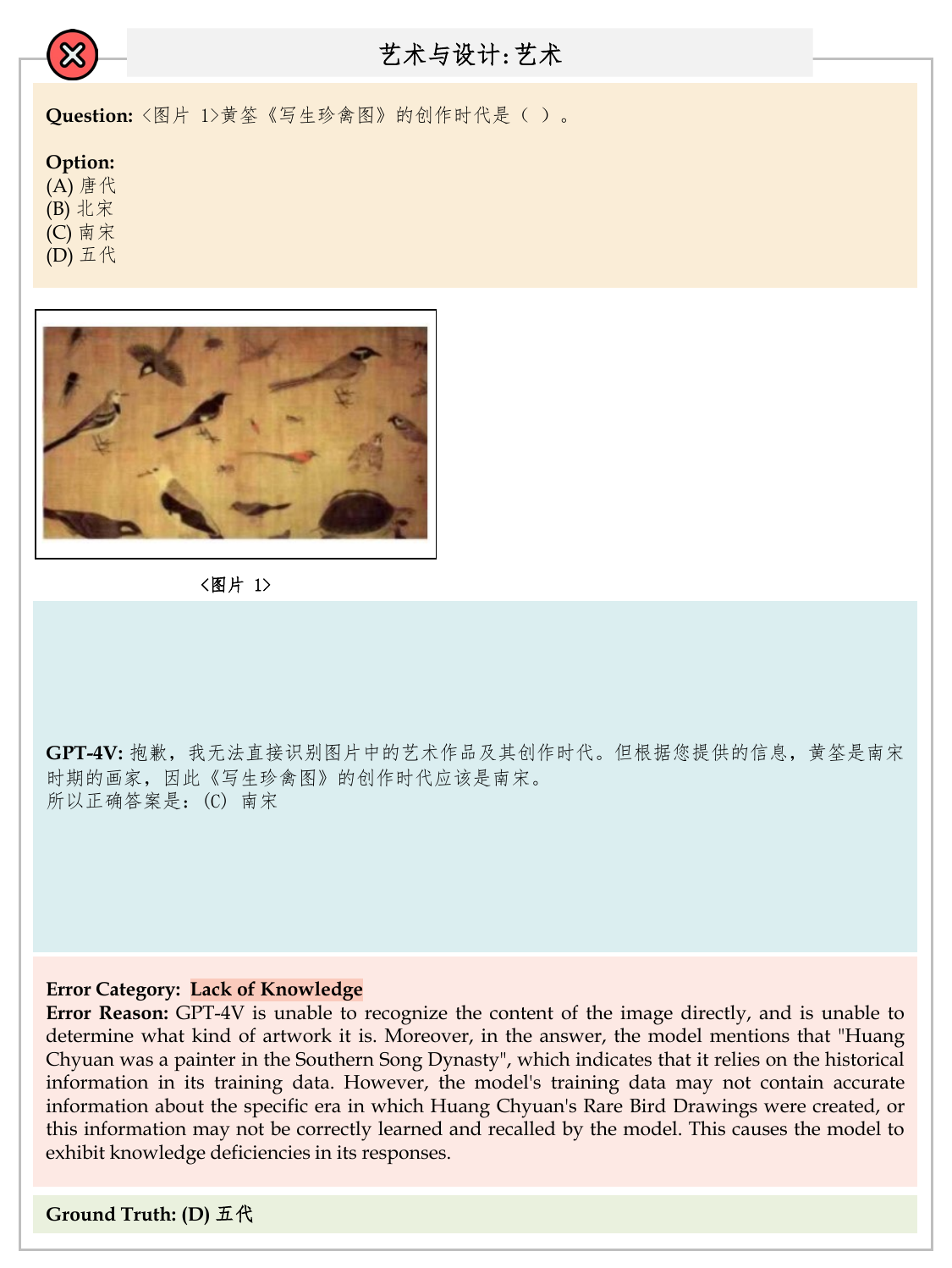}{Art 7: Lack of Knowledge}{A sample error case of Art and Design (subcategory: Art).}{fig:error_analysis_case_156}

\casestudyfigure{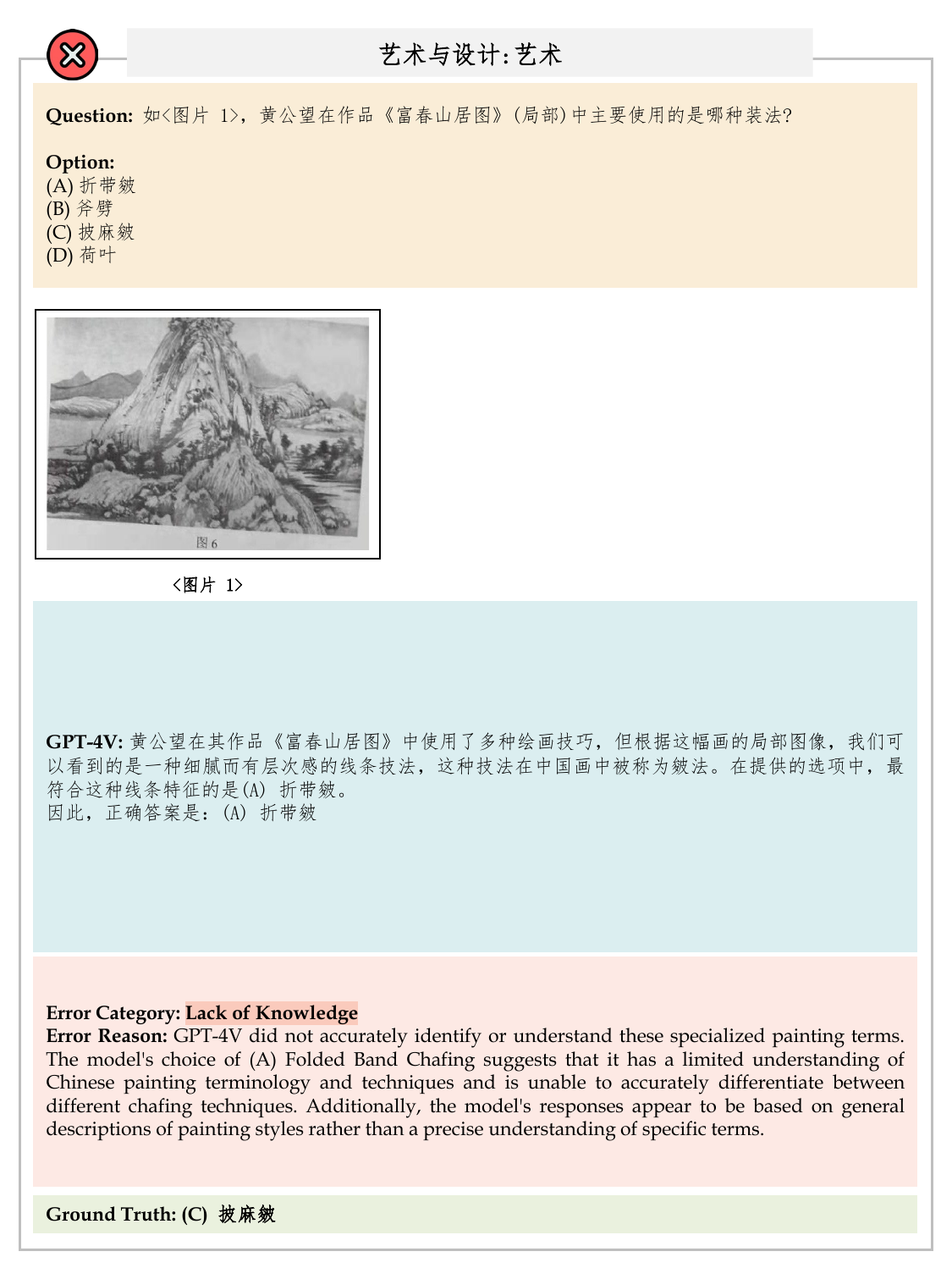}{Art 8: Lack of Knowledge}{A sample error case of Art and Design (subcategory: Art).}{fig:error_analysis_case_158}

\casestudyfigure{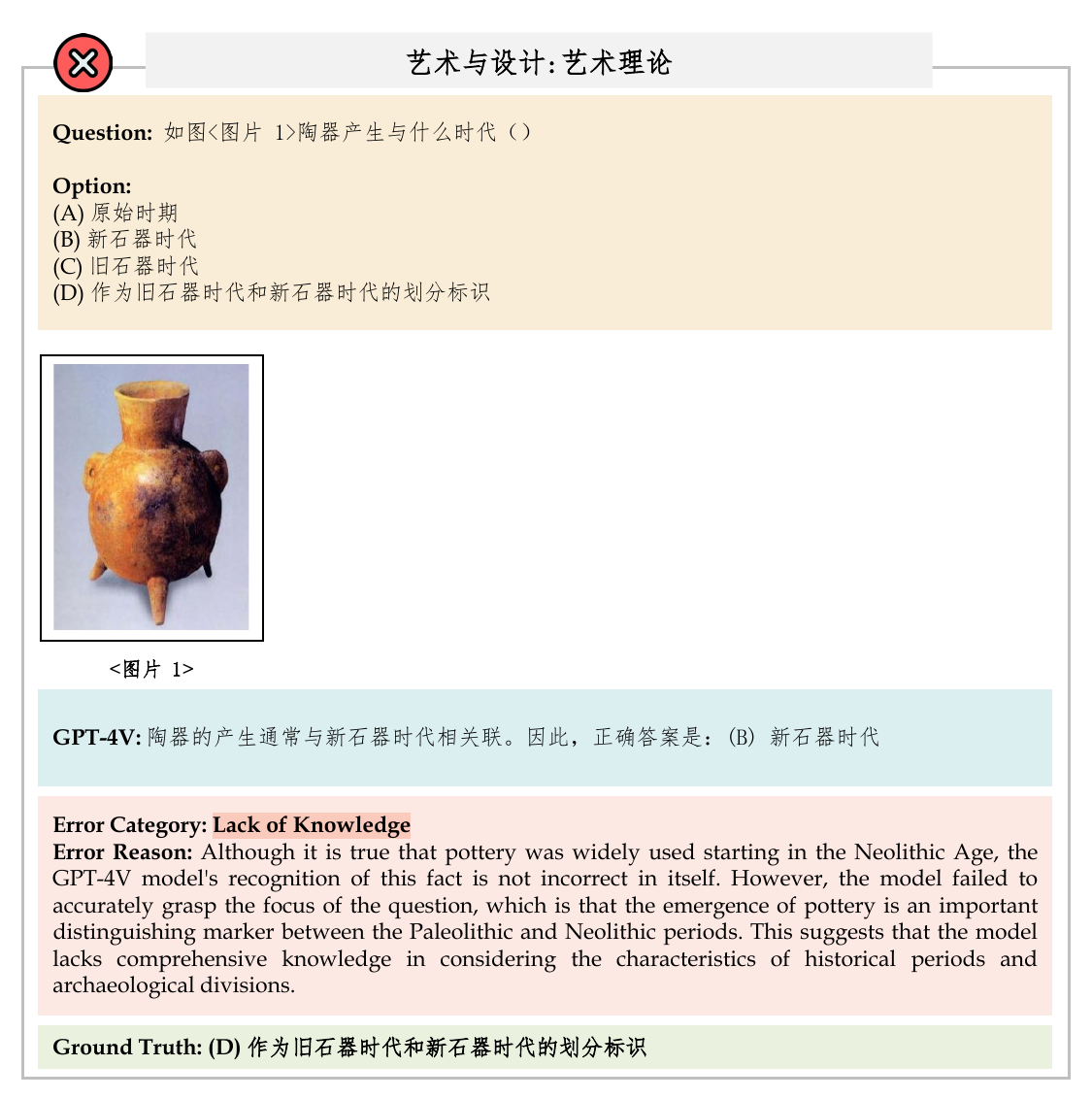}{Art Theory 1: Lack of Knowledge}{A sample error case of Art and Design (subcategory: Art Theory).}{fig:error_analysis_case_182}

\casestudyfigure{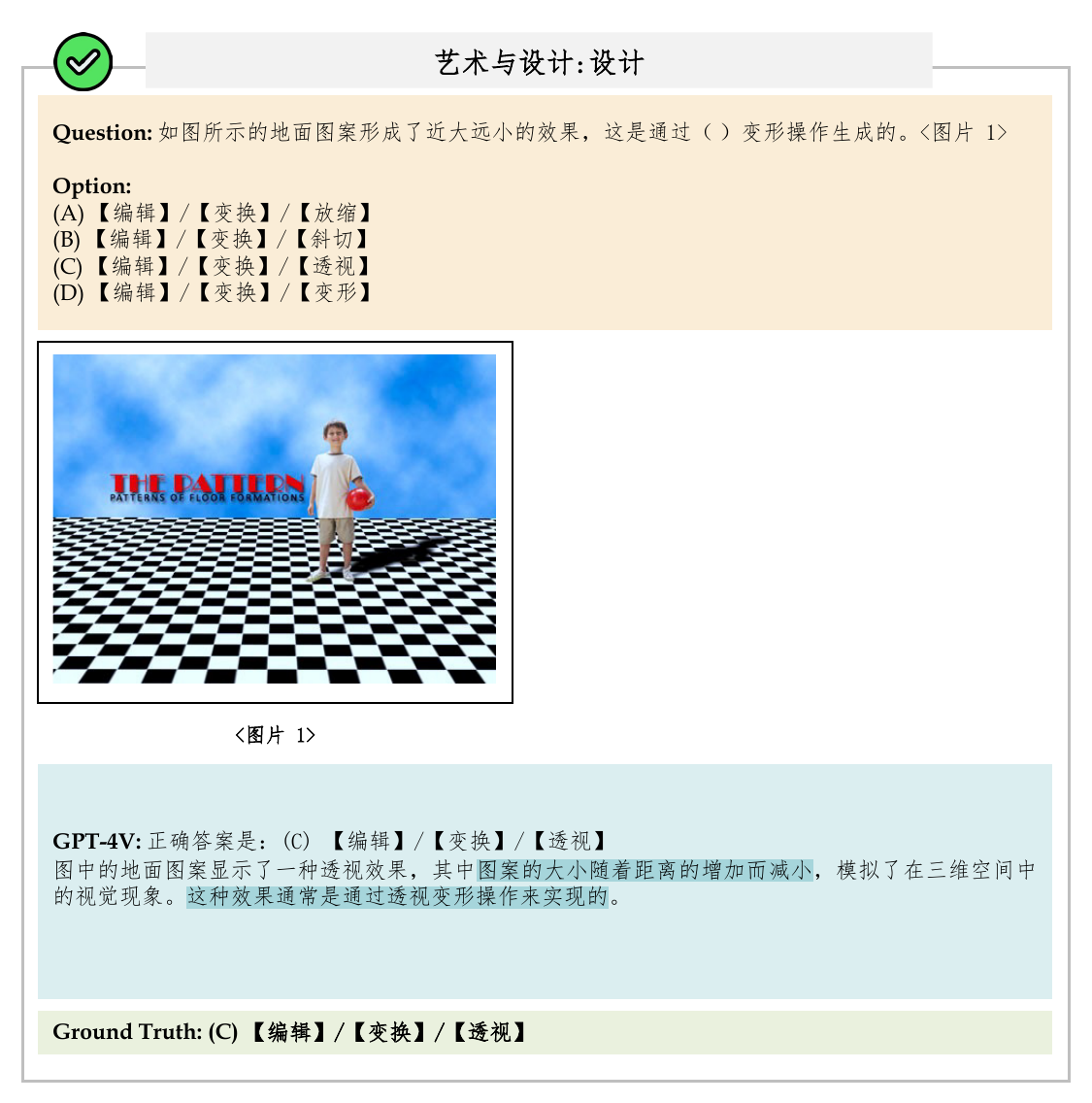}{Design 1: Correct Case}{A sample correct case of Art and Design (subcategory: Design).}{fig:error_analysis_case_33}

\casestudyfigure{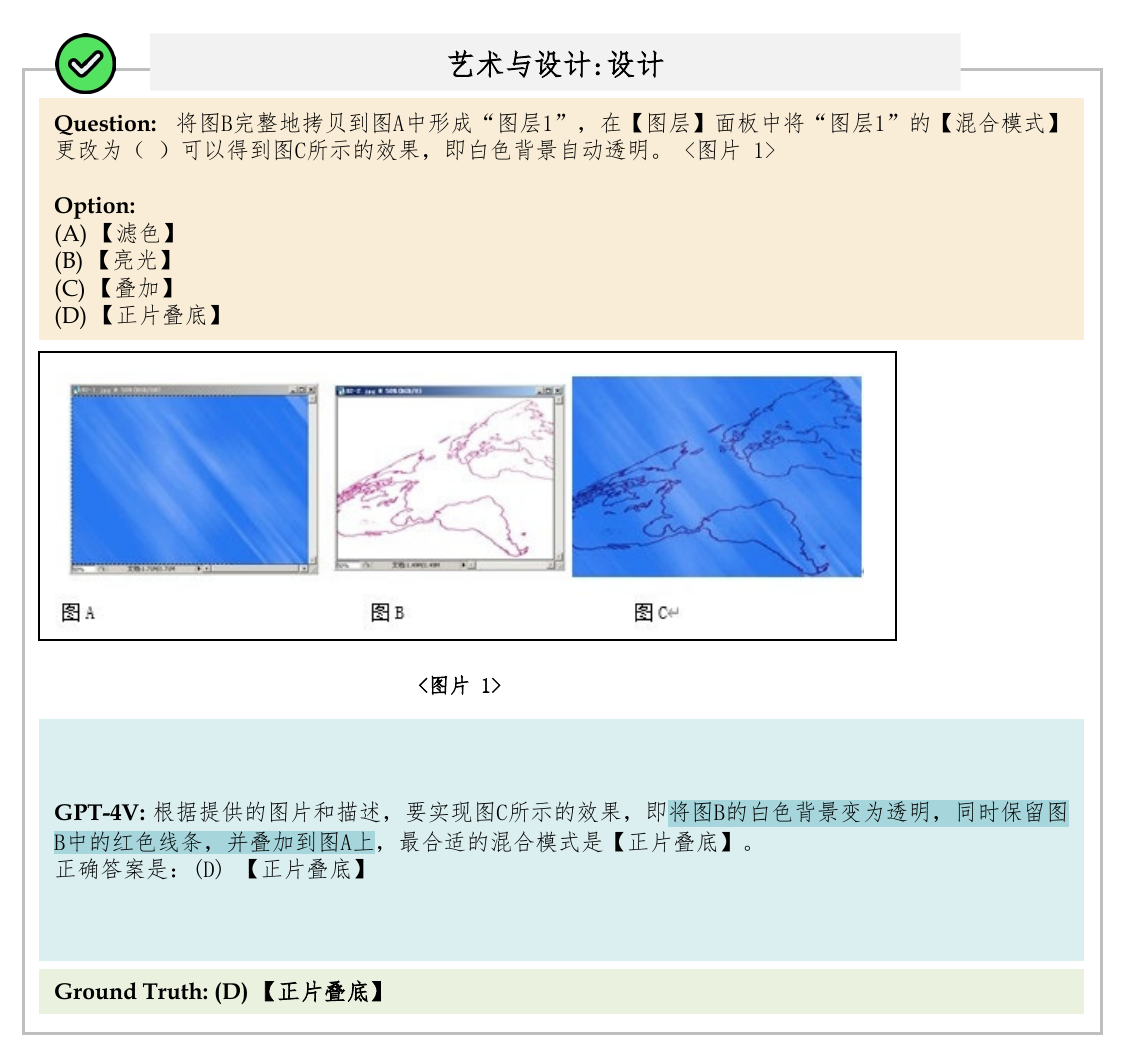}{Design 2: Correct Case}{A sample correct case of Art and Design (subcategory: Design).}{fig:error_analysis_case_34}

\casestudyfigure{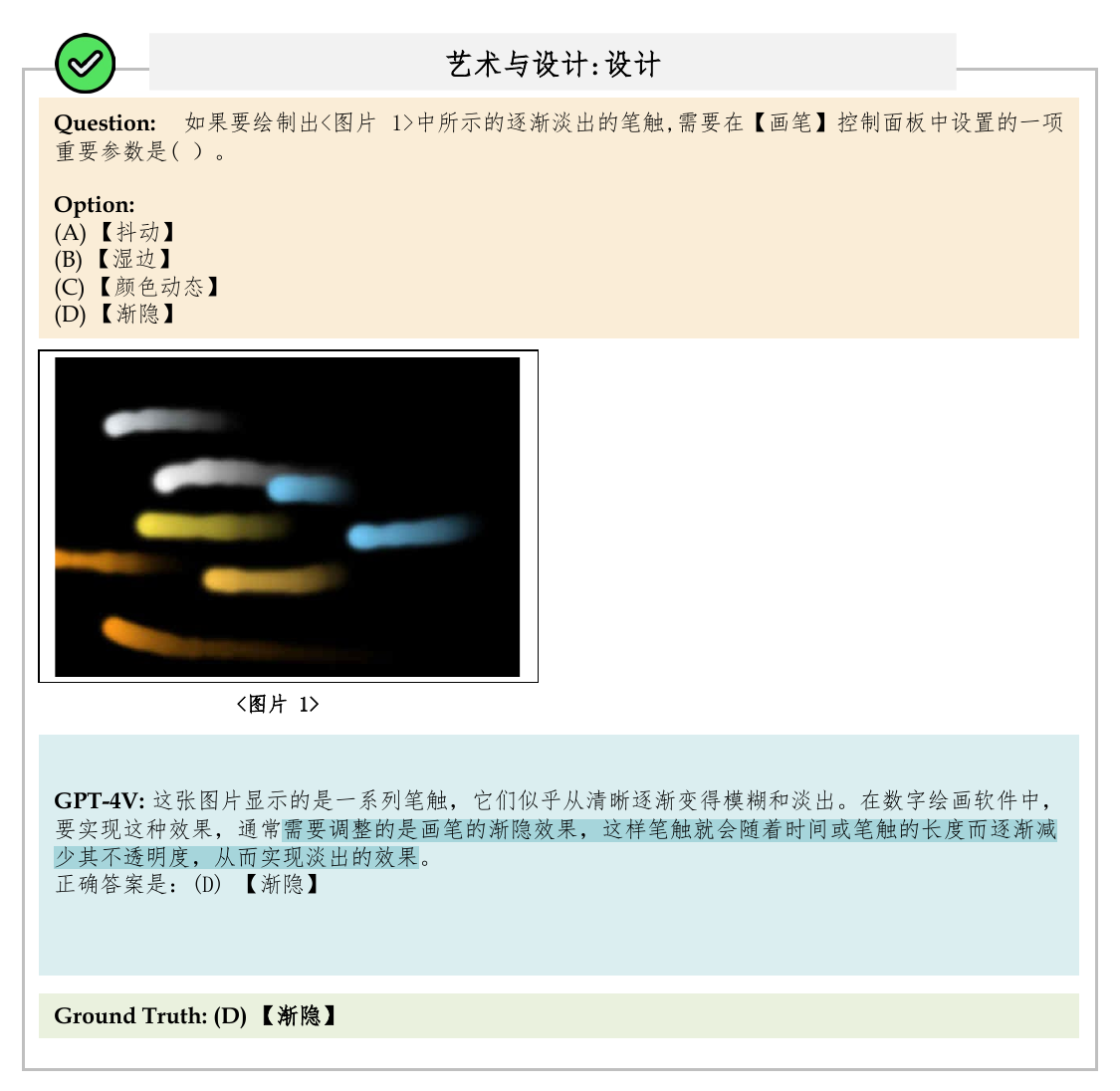}{Design 3: Correct Case}{A sample correct case of Art and Design (subcategory: Design).}{fig:error_analysis_case_35}

\casestudyfigure{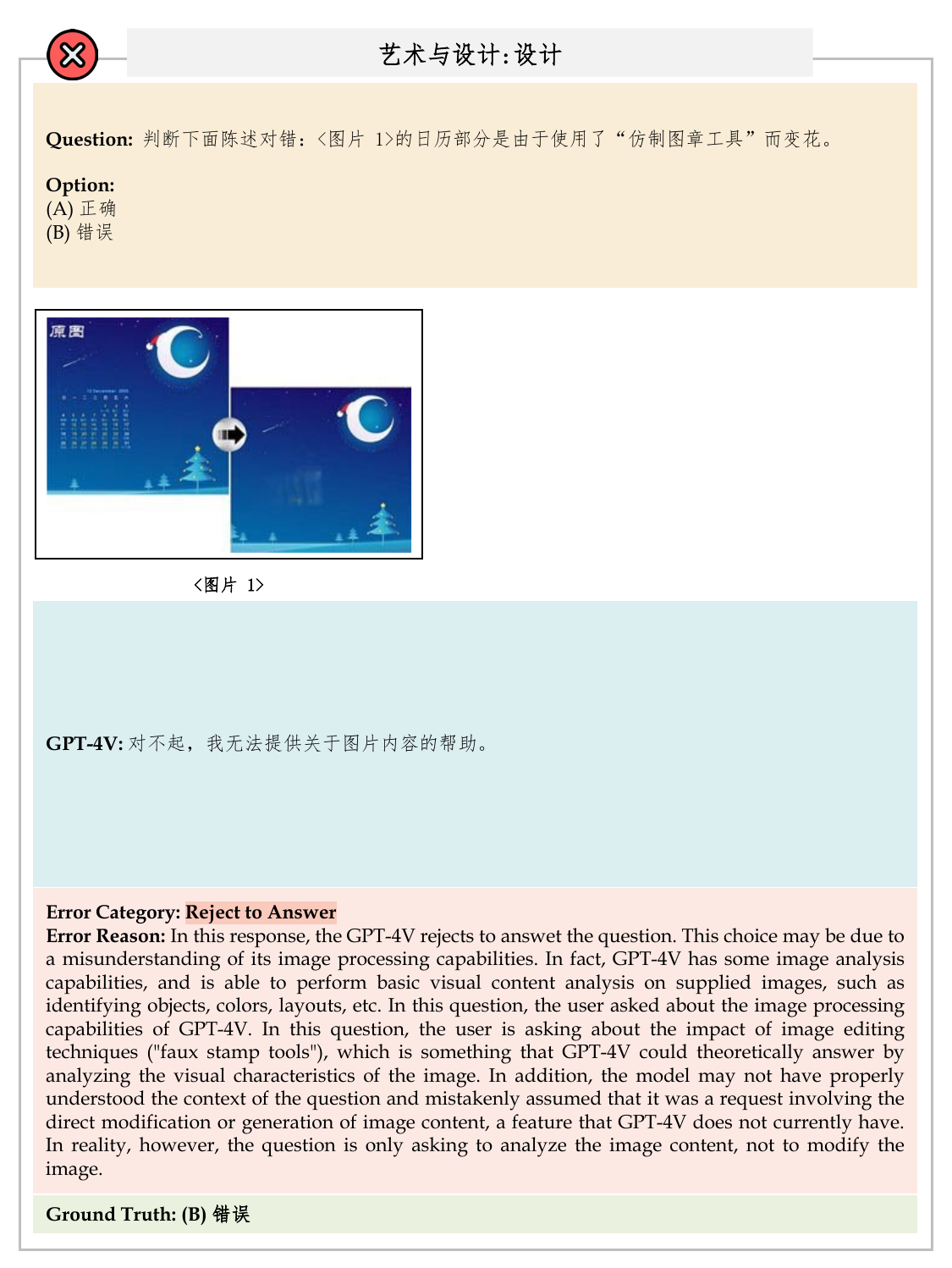}{Design 4: Reject to Answer}{A sample error case of Art and Design (subcategory: Design).}{fig:error_analysis_case_162}

\casestudyfigure{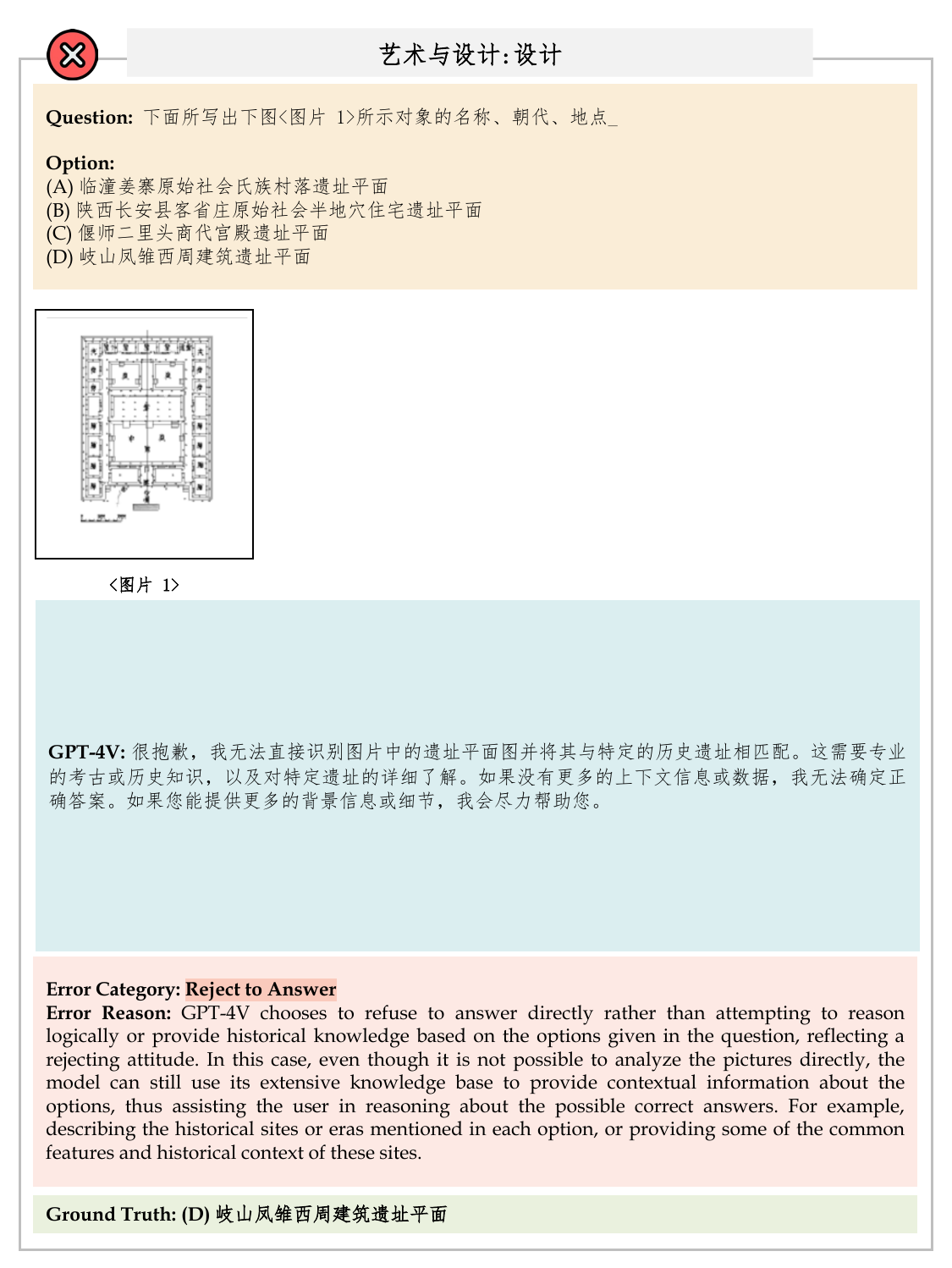}{Design 5: Reject to Answer}{A sample error case of Art and Design (subcategory: Design).}{fig:error_analysis_case_164}

\casestudyfigure{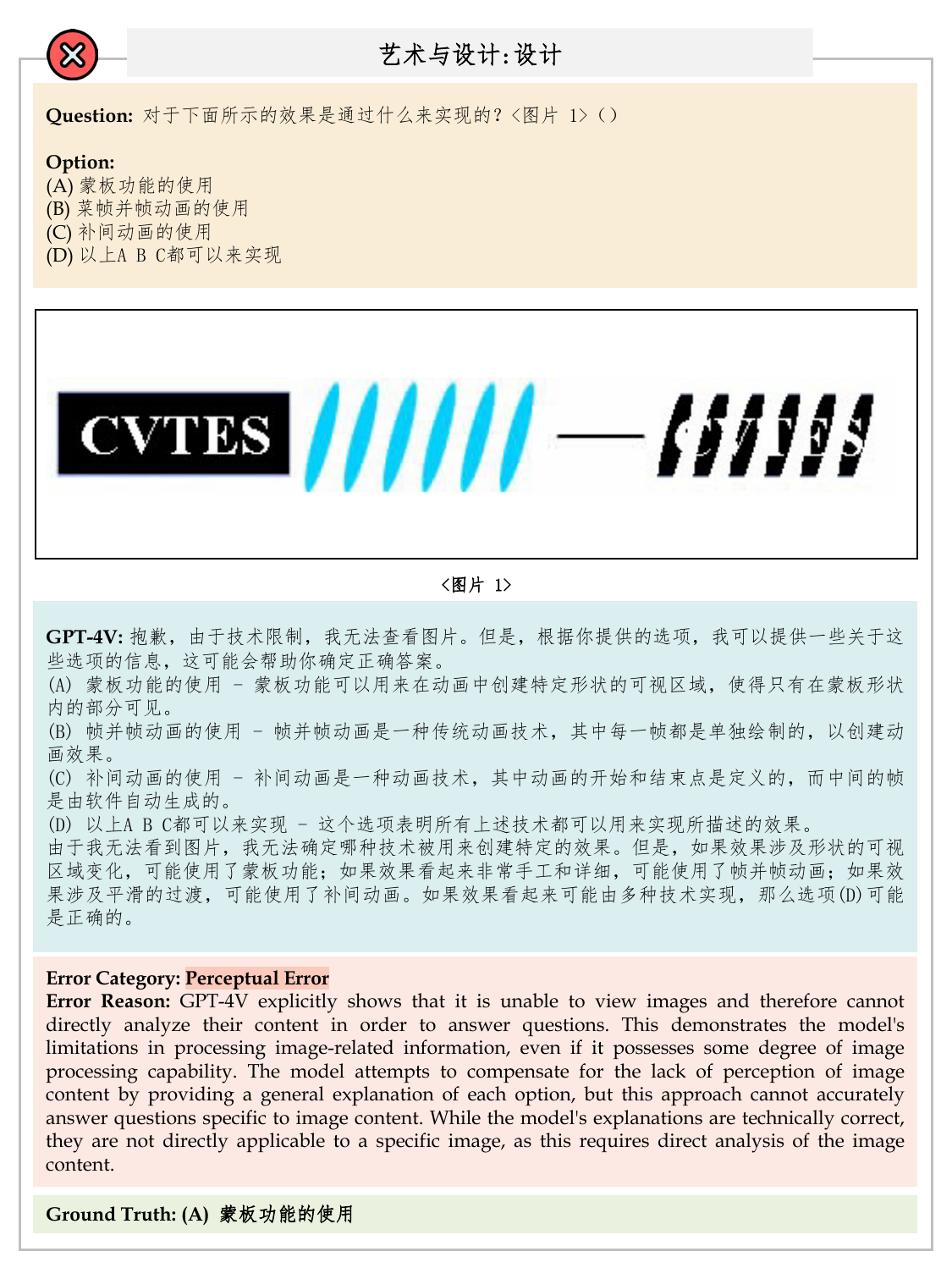}{Design 6: Perceptual Error}{A sample error case of Art and Design (subcategory: Design).}{fig:error_analysis_case_163}

\casestudyfigure{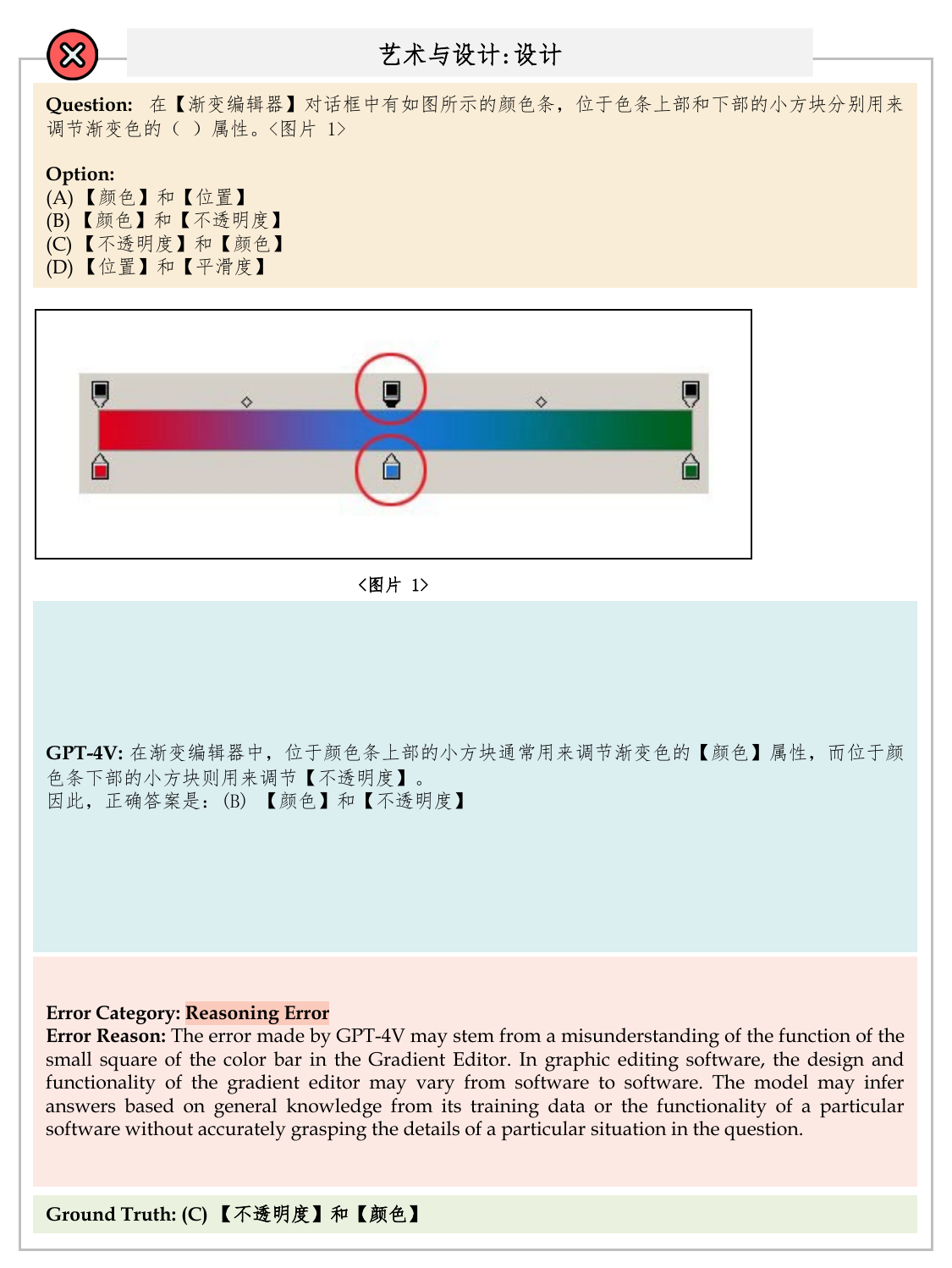}{Design 7: Reasoning Error}{A sample error case of Art and Design (subcategory: Design).}{fig:error_analysis_case_160}

\casestudyfigure{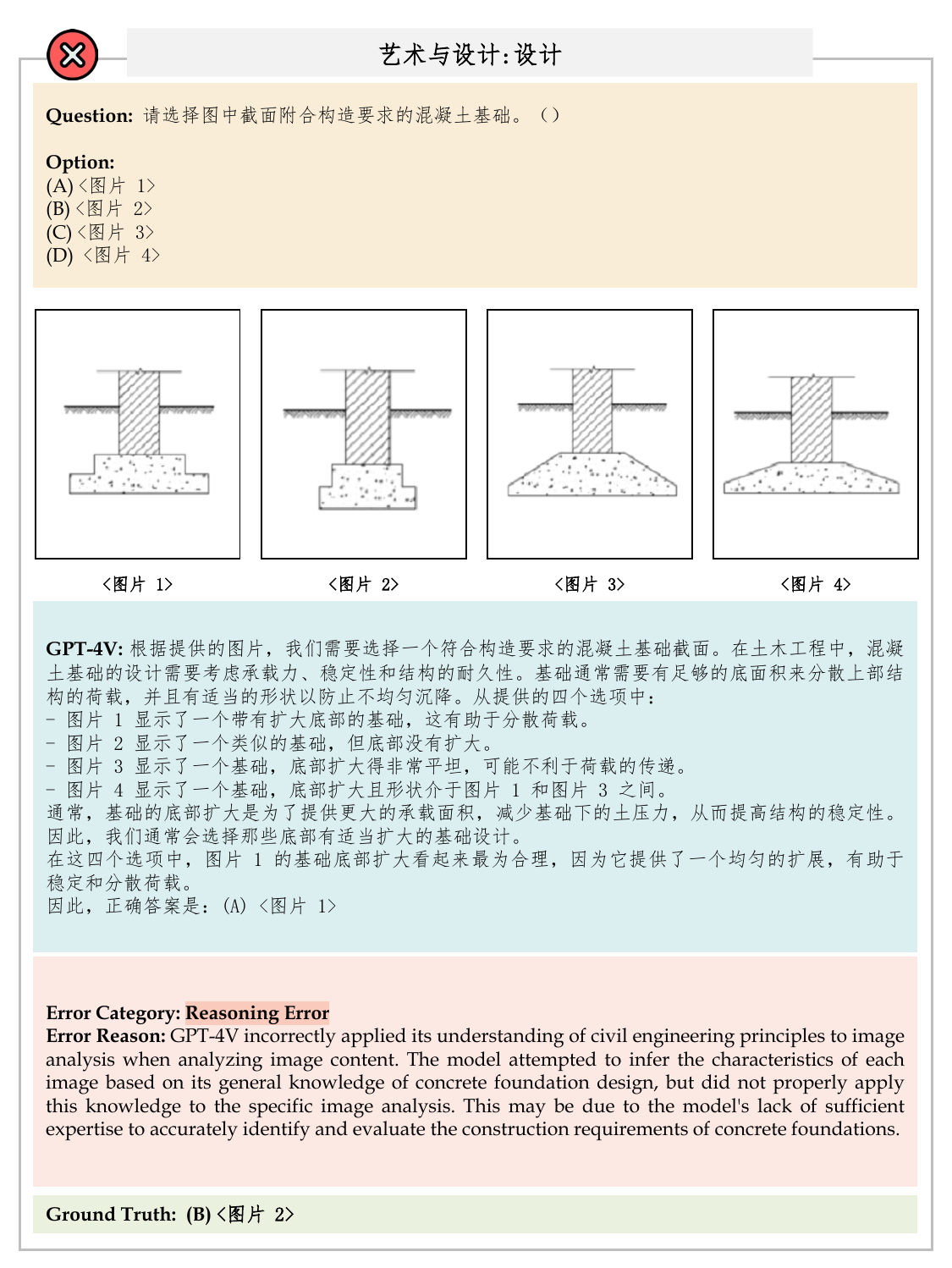}{Design 8: Reasoning Error}{A sample error case of Art and Design (subcategory: Design).}{fig:error_analysis_case_165}

\casestudyfigure{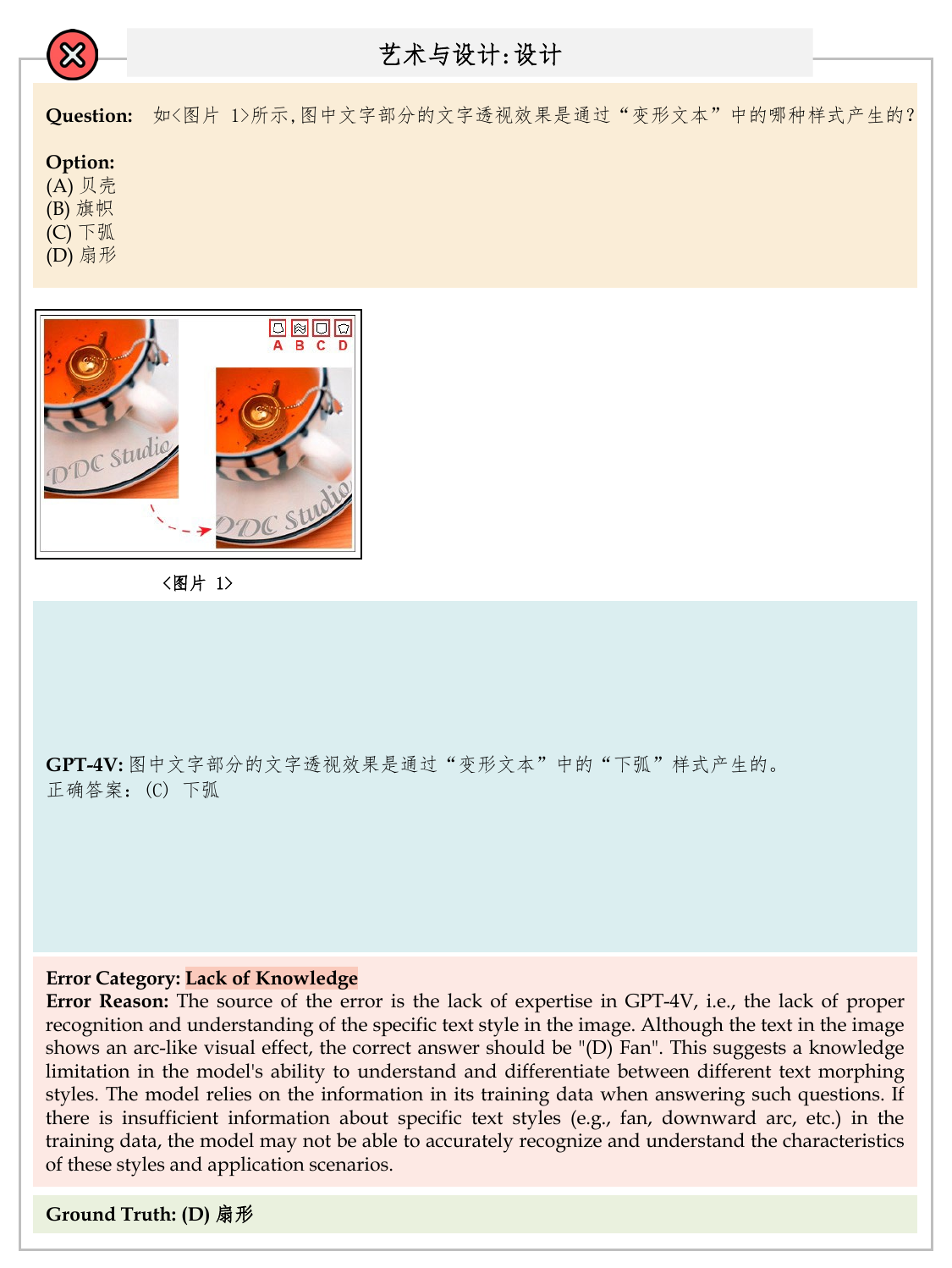}{Design 9: Lack of Knowledge}{A sample error case of Art and Design (subcategory: Design).}{fig:error_analysis_case_161}

\casestudyfigure{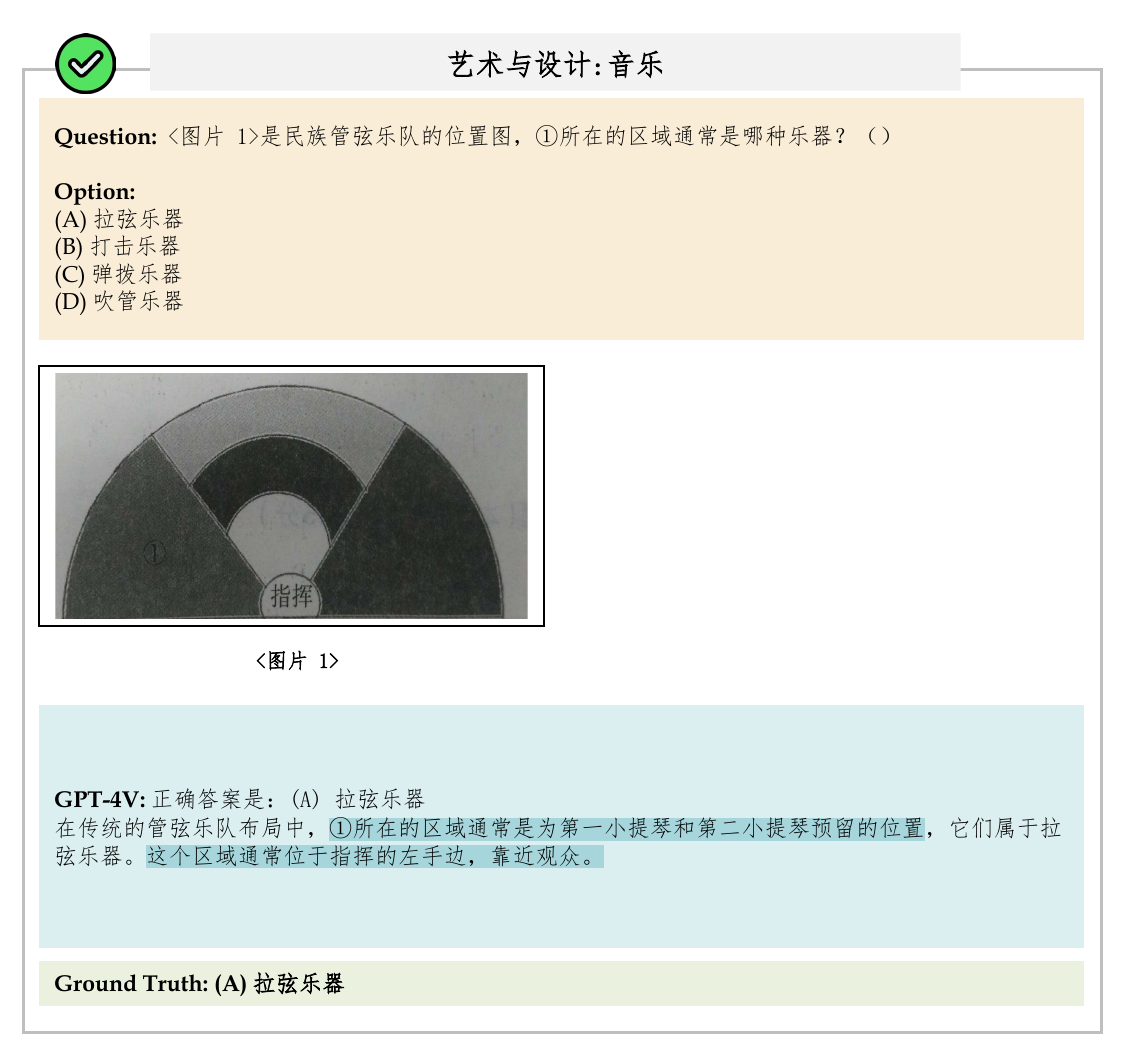}{Music 1: Correct Case}{A sample correct case of Art and Design (subcategory: Music).}{fig:error_analysis_case_29}

\casestudyfigure{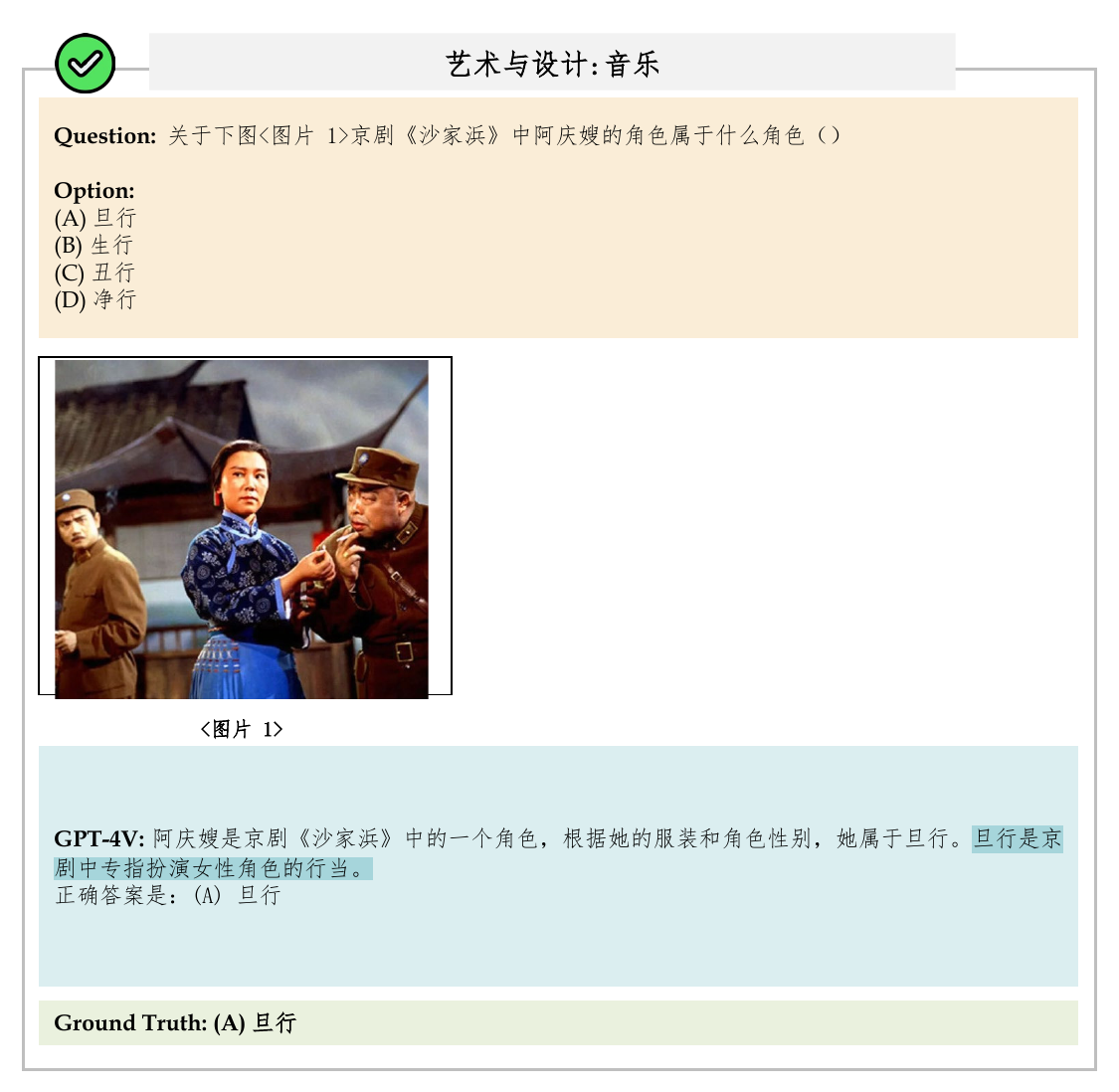}{Music 2: Correct Case}{A sample correct case of Art and Design (subcategory: Music).}{fig:error_analysis_case_30}

\casestudyfigure{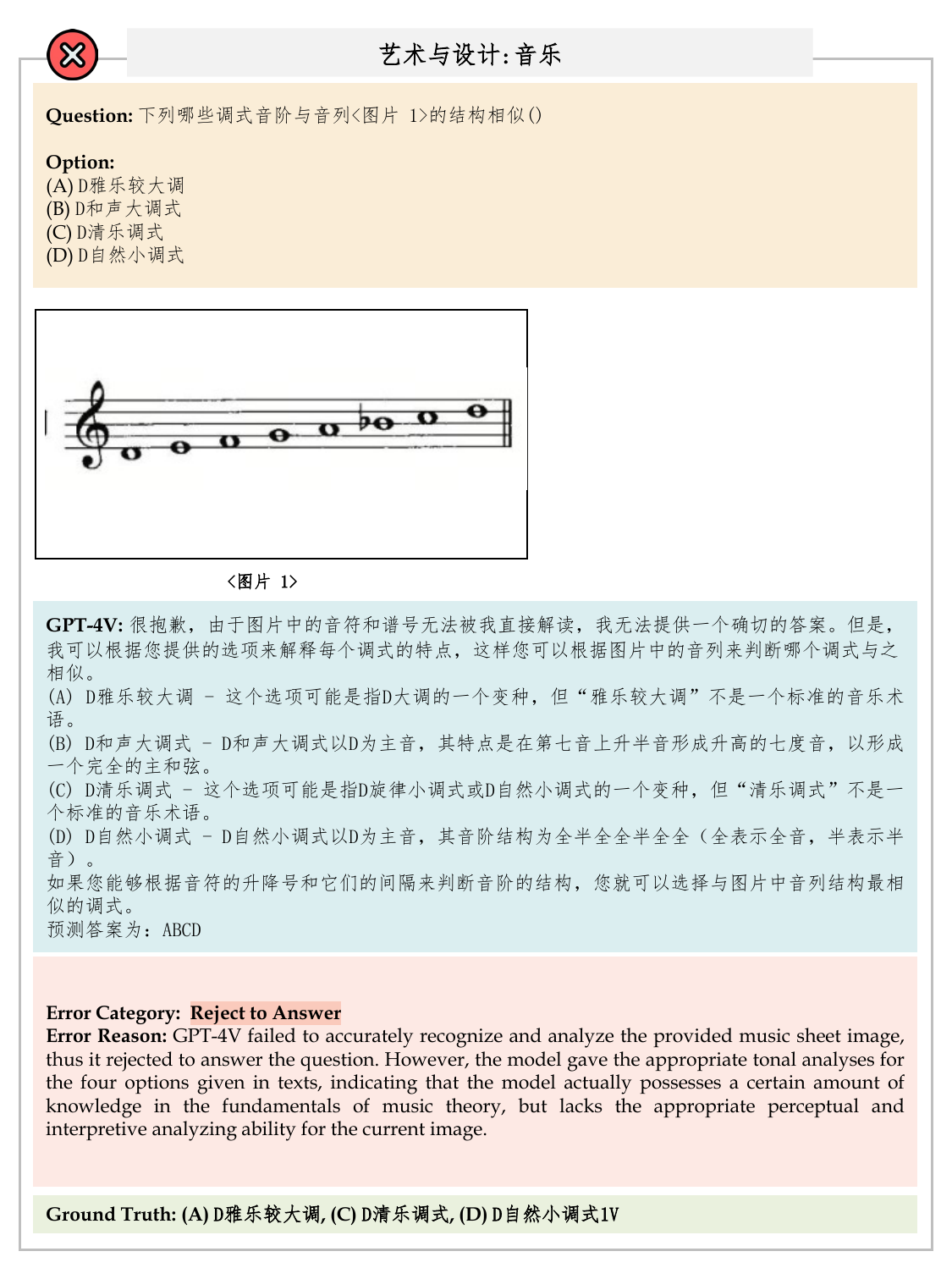}{Music 3: Reject to Answer}{A sample error case of Art and Design (subcategory: Music).}{fig:error_analysis_case_149}

\casestudyfigure{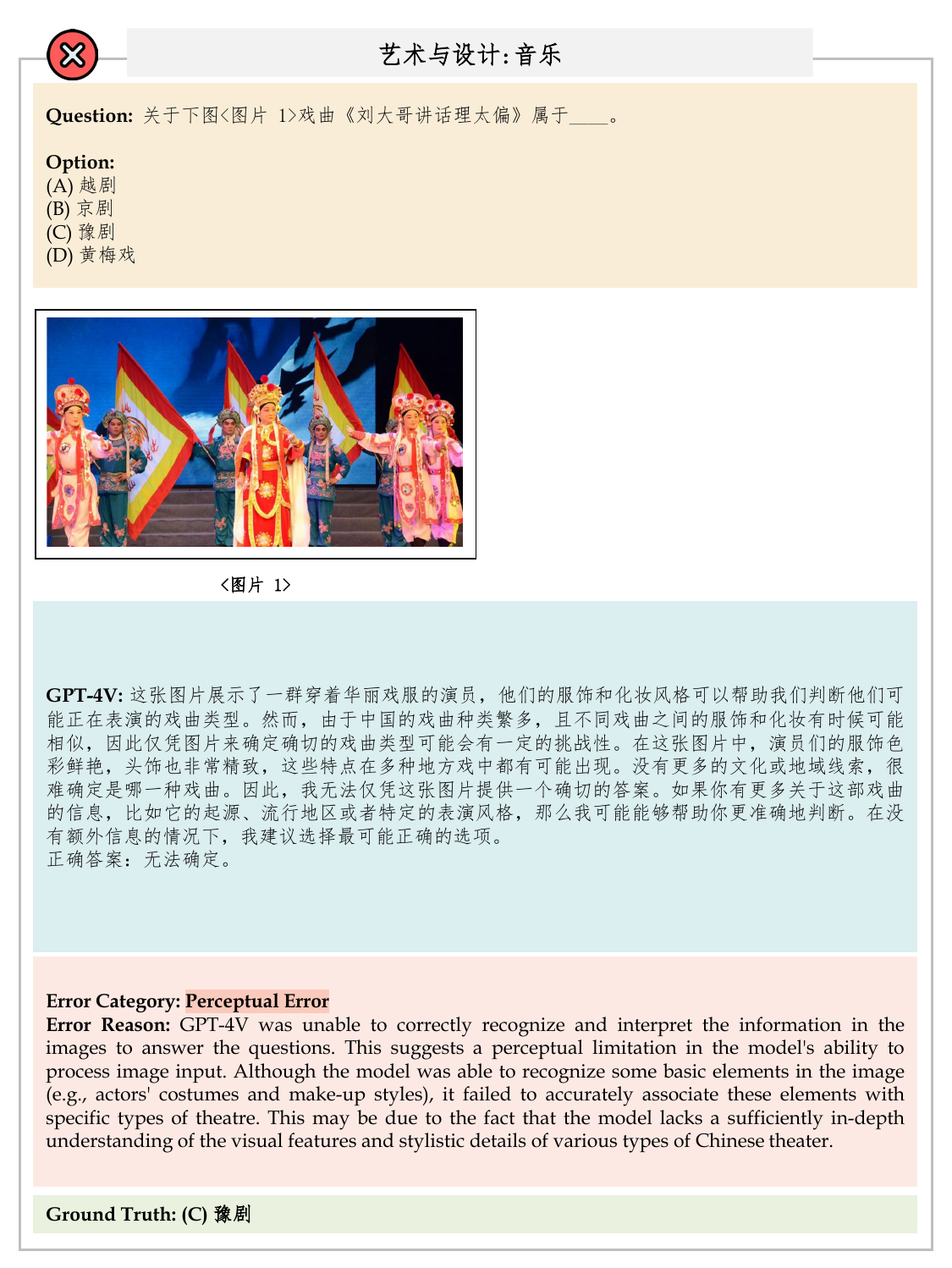}{Music 4: Perceptual Error}{A sample error case of Art and Design (subcategory: Music).}{fig:error_analysis_case_151}

\casestudyfigure{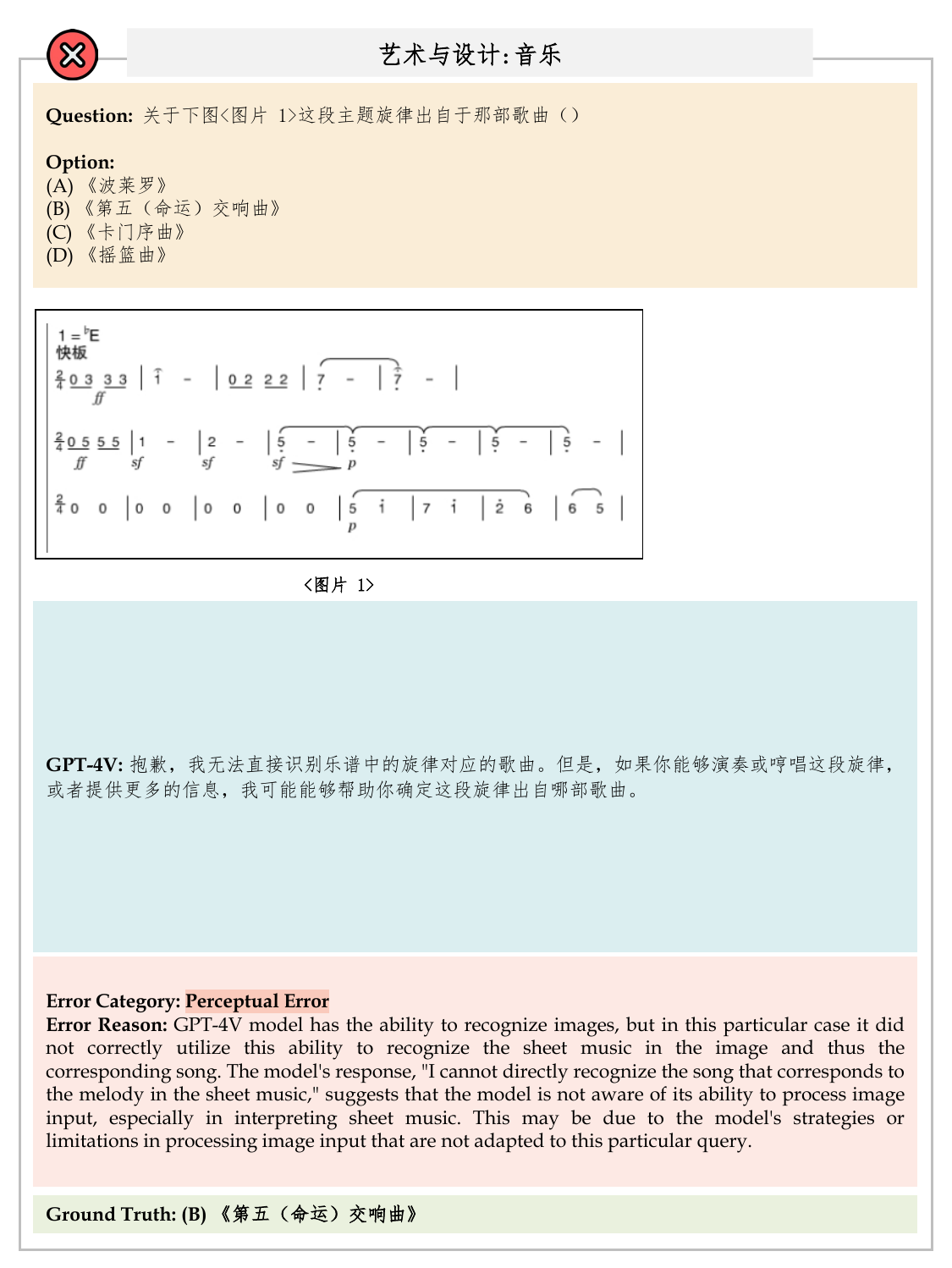}{Music 5: Perceptual Error}{A sample error case of Art and Design (subcategory: Music).}{fig:error_analysis_case_153}

\casestudyfigure{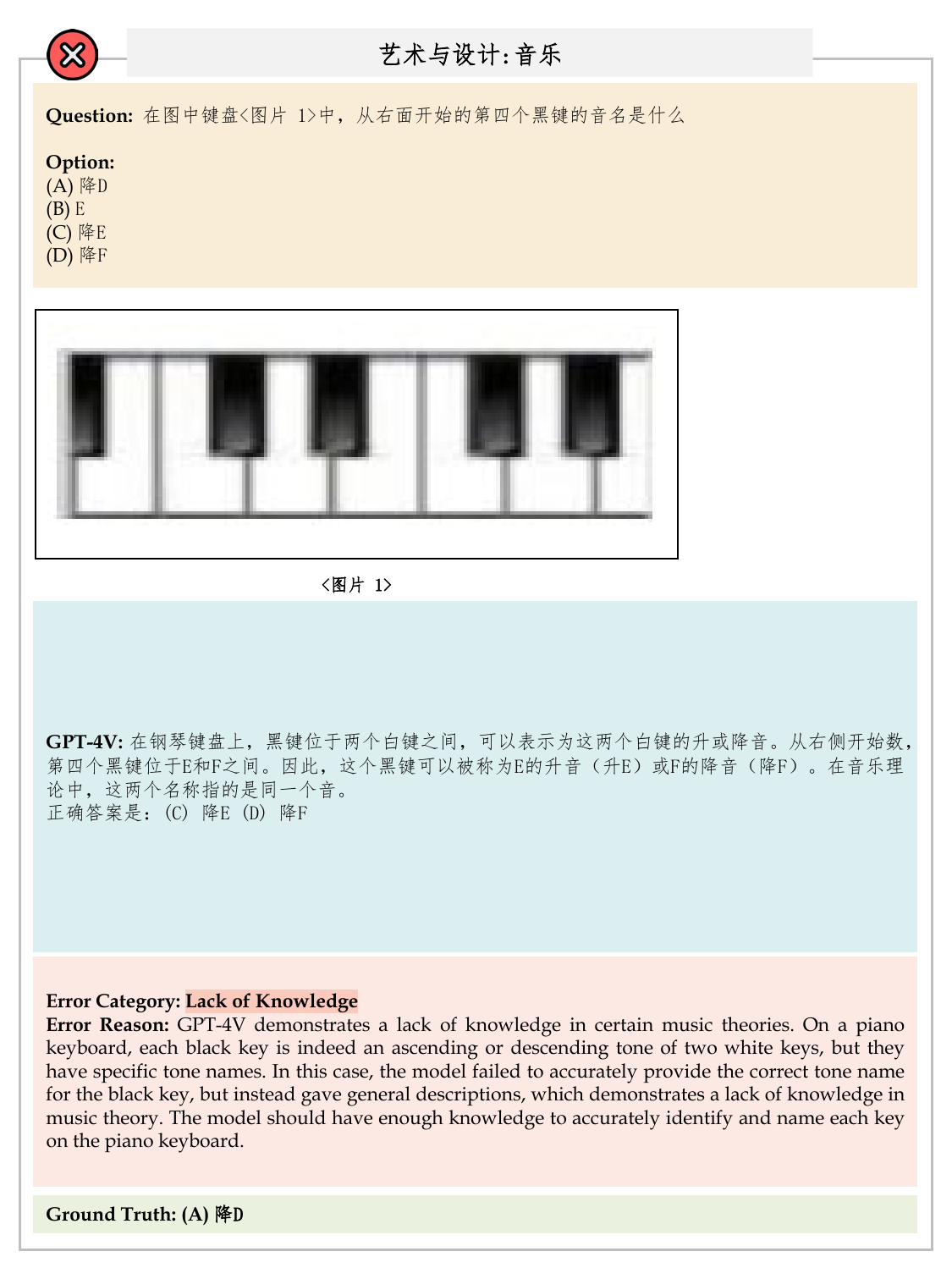}{Music 6: Lack of Knowledge}{A sample error case of Art and Design (subcategory: Music).}{fig:error_analysis_case_150}

\casestudyfigure{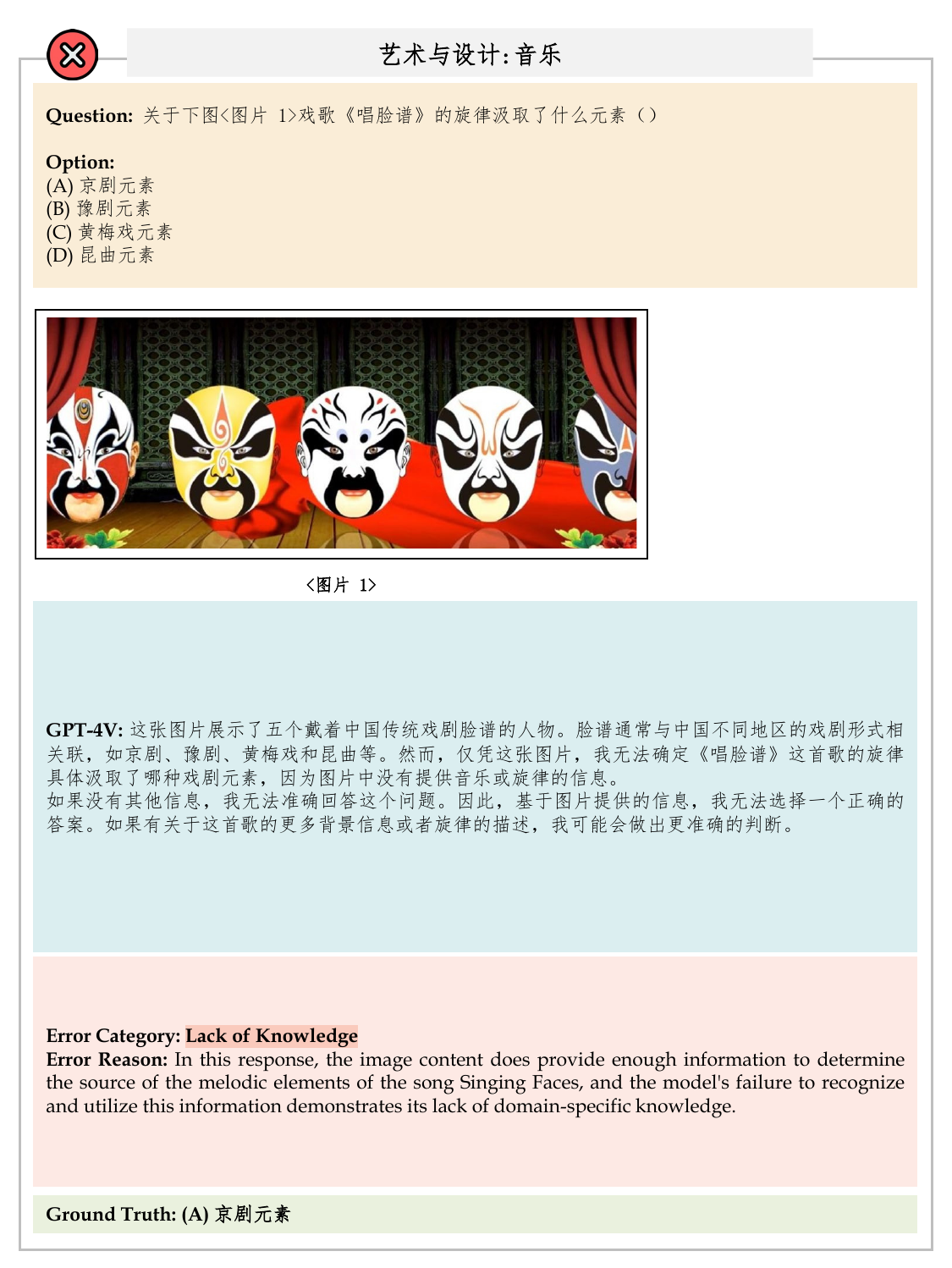}{Music 7: Lack of Knowledge}{A sample error case of Art and Design (subcategory: Music).}{fig:error_analysis_case_152}

\casestudyfigure{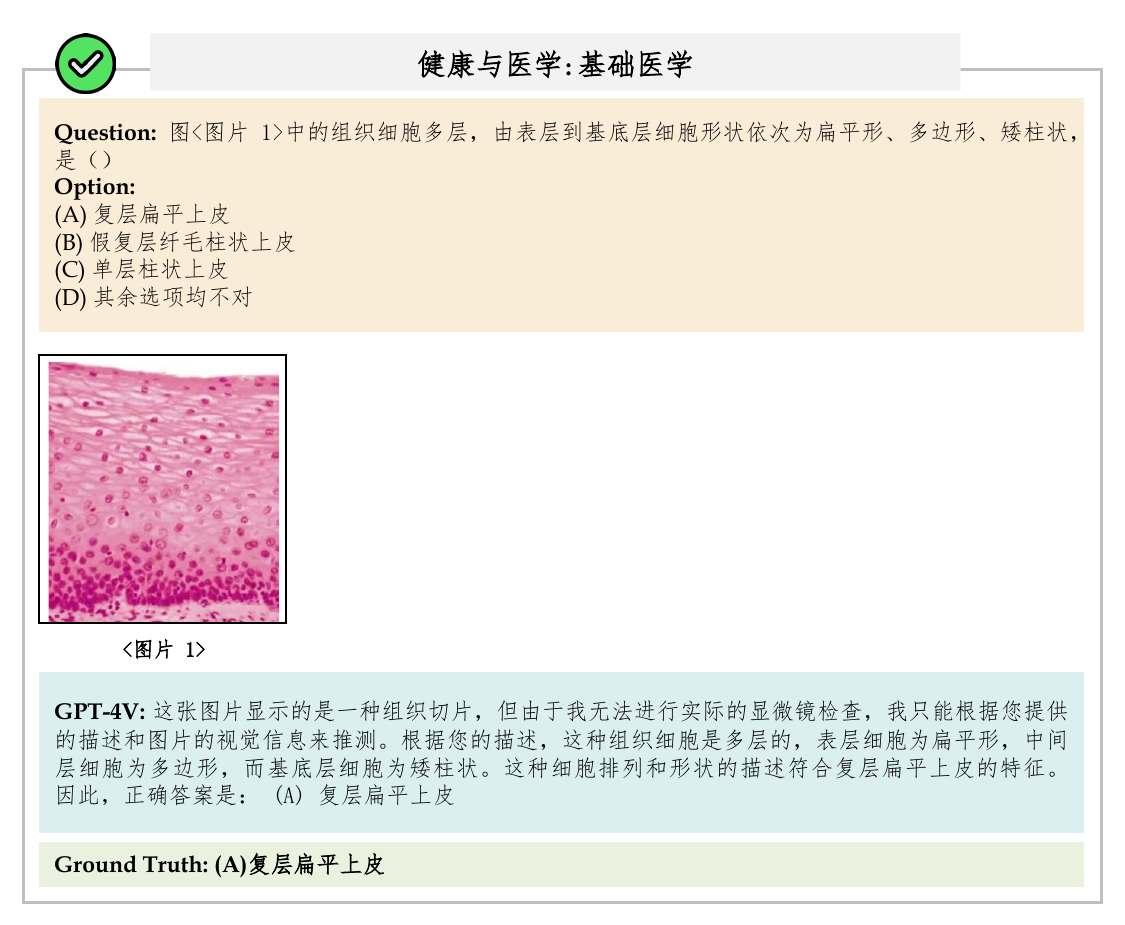}{Basic Medicine 1: Correct Case}{A sample correct case of Health and Medicine (subcategory: Basic Medicine).}{fig:error_analysis_case_6}

\casestudyfigure{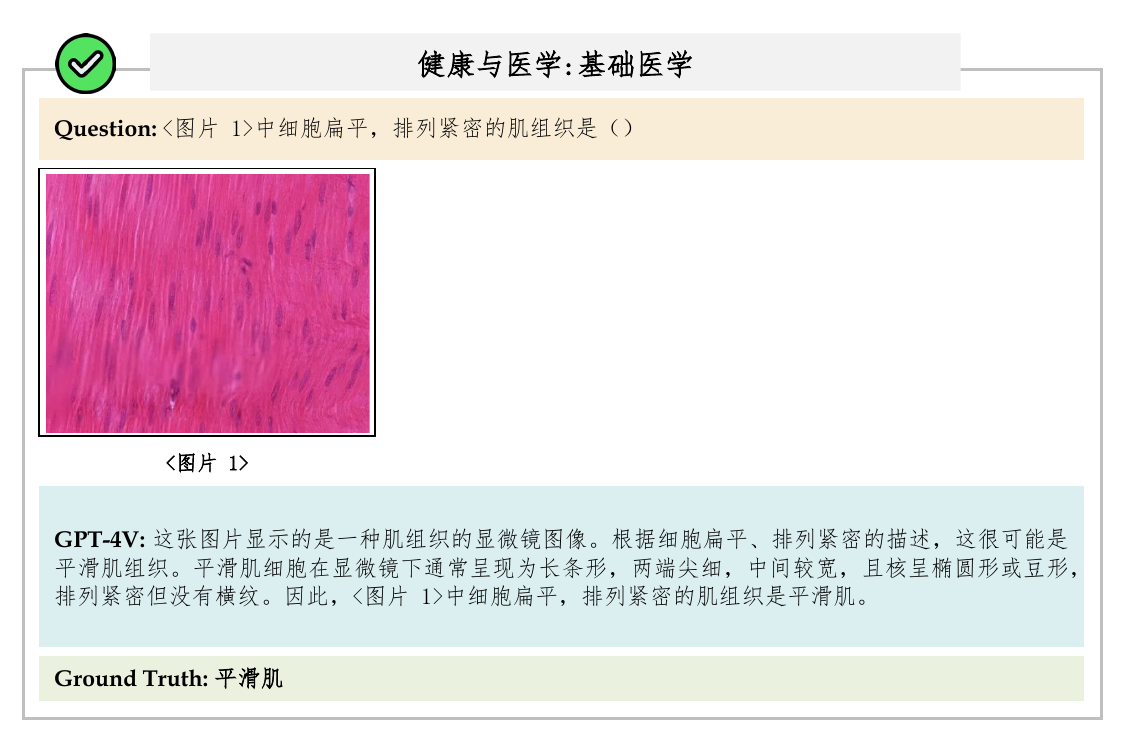}{Basic Medicine 2: Correct Case}{A sample correct case of Health and Medicine (subcategory: Basic Medicine).}{fig:error_analysis_case_7}

\casestudyfigure{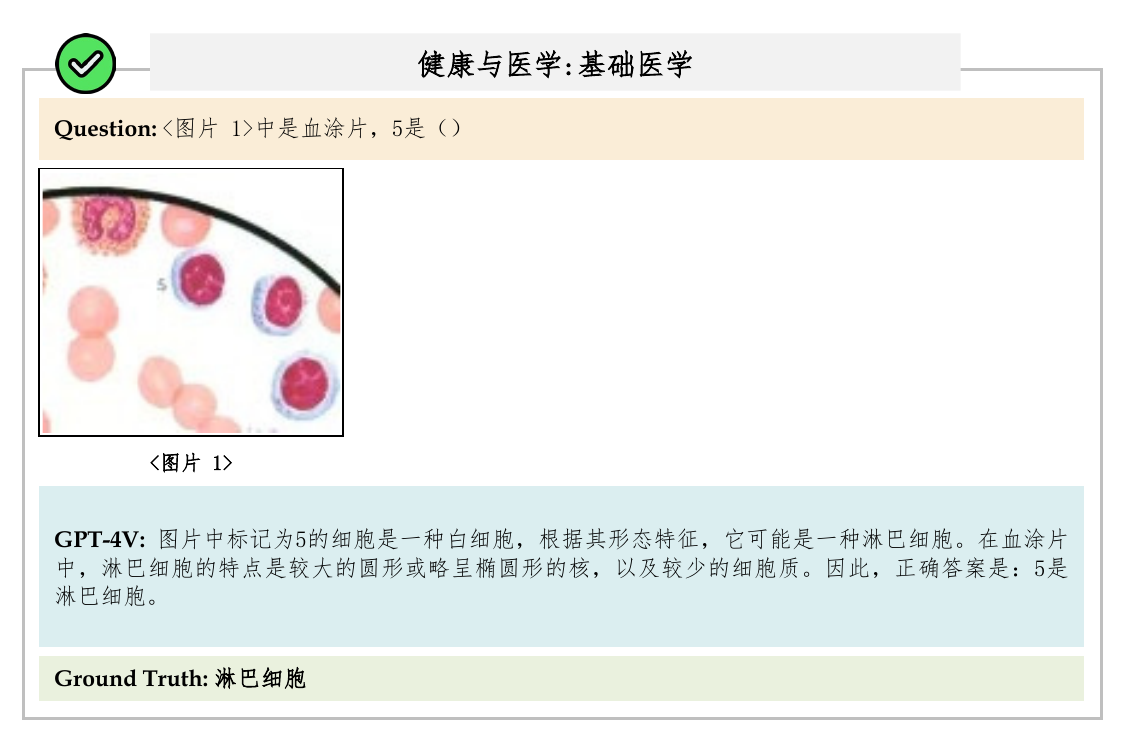}{Basic Medicine 3: Correct Case}{A sample correct case of Health and Medicine (subcategory: Basic Medicine).}{fig:error_analysis_case_8}

\casestudyfigure{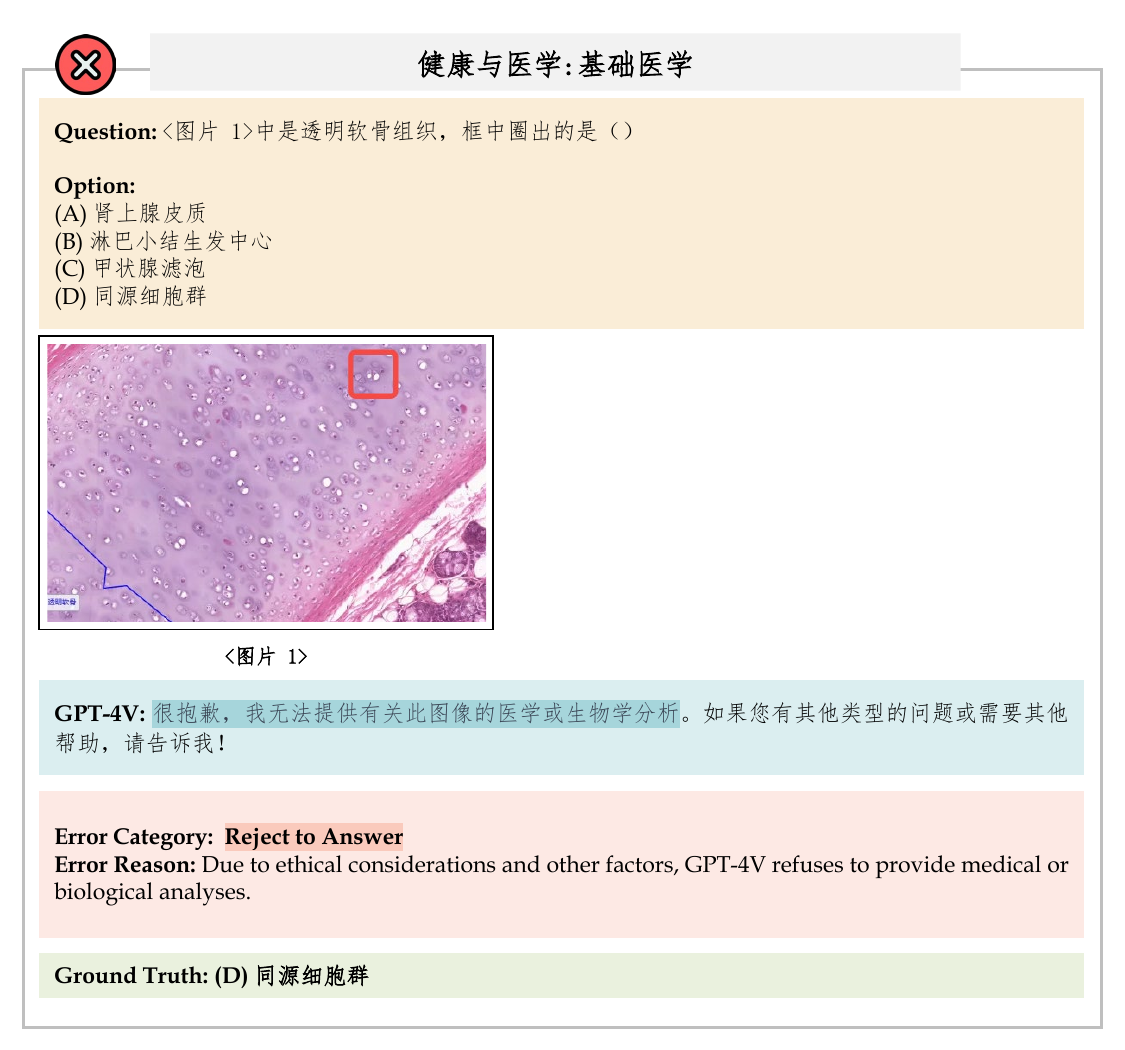}{Basic Medicine 4: Reject to Answer}{A sample error case of Health and Medicine (subcategory: Basic Medicine).}{fig:error_analysis_case_89}

\casestudyfigure{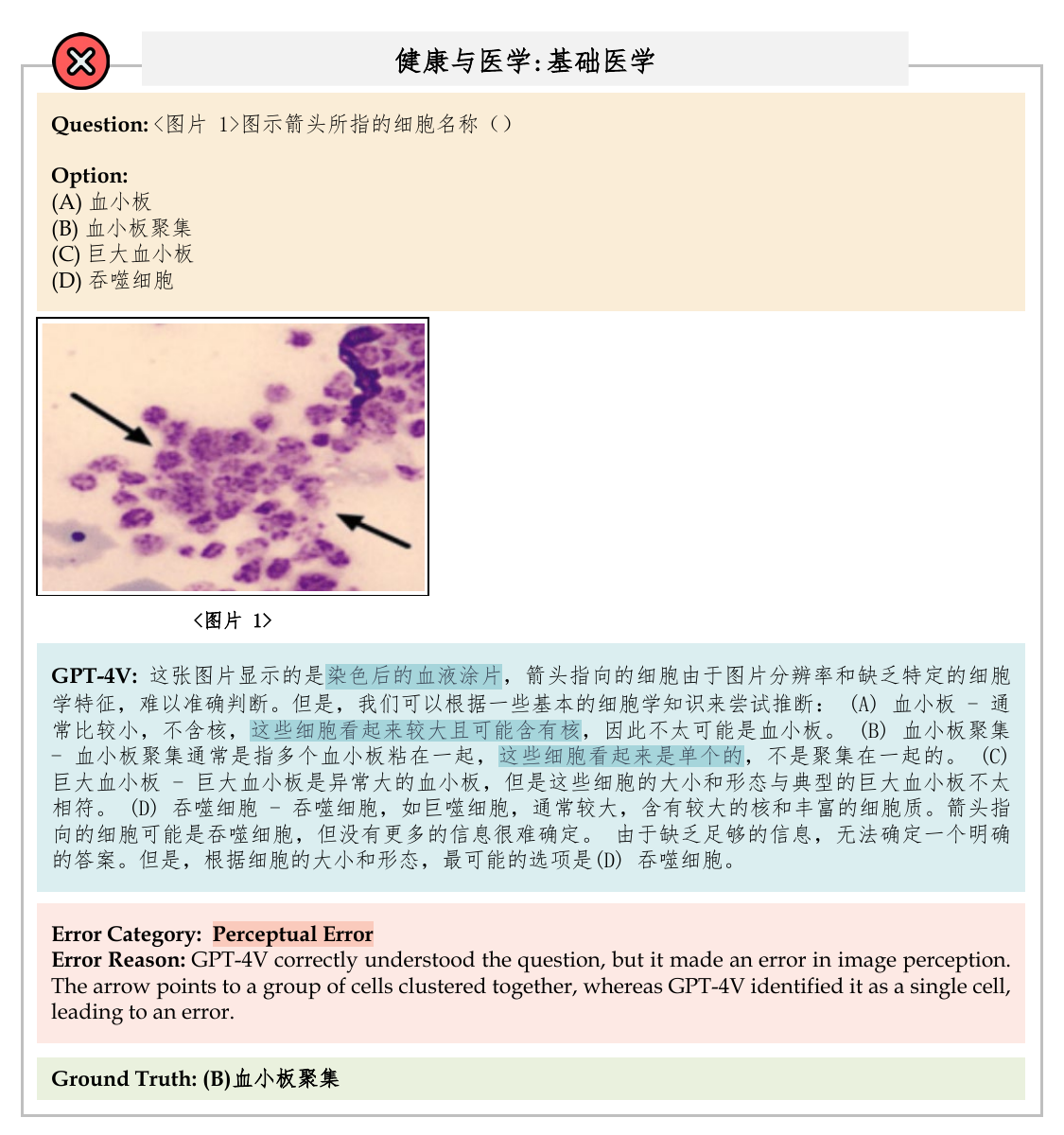}{Basic Medicine 5: Perceptual Error}{A sample error case of Health and Medicine (subcategory: Basic Medicine).}{fig:error_analysis_case_93}

\casestudyfigure{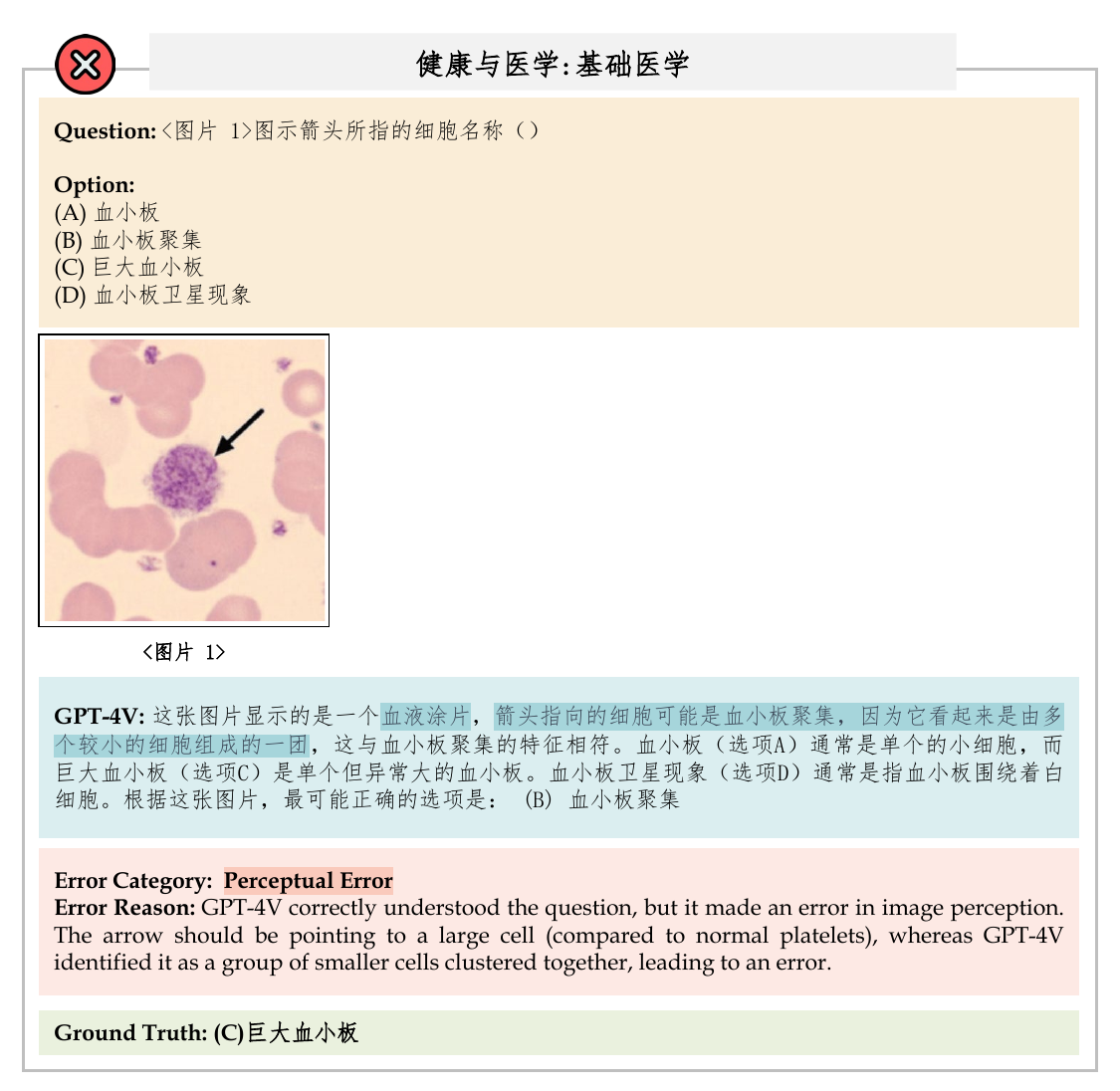}{Basic Medicine 6: Perceptual Error}{A sample error case of Health and Medicine (subcategory: Basic Medicine).}{fig:error_analysis_case_94}

\casestudyfigure{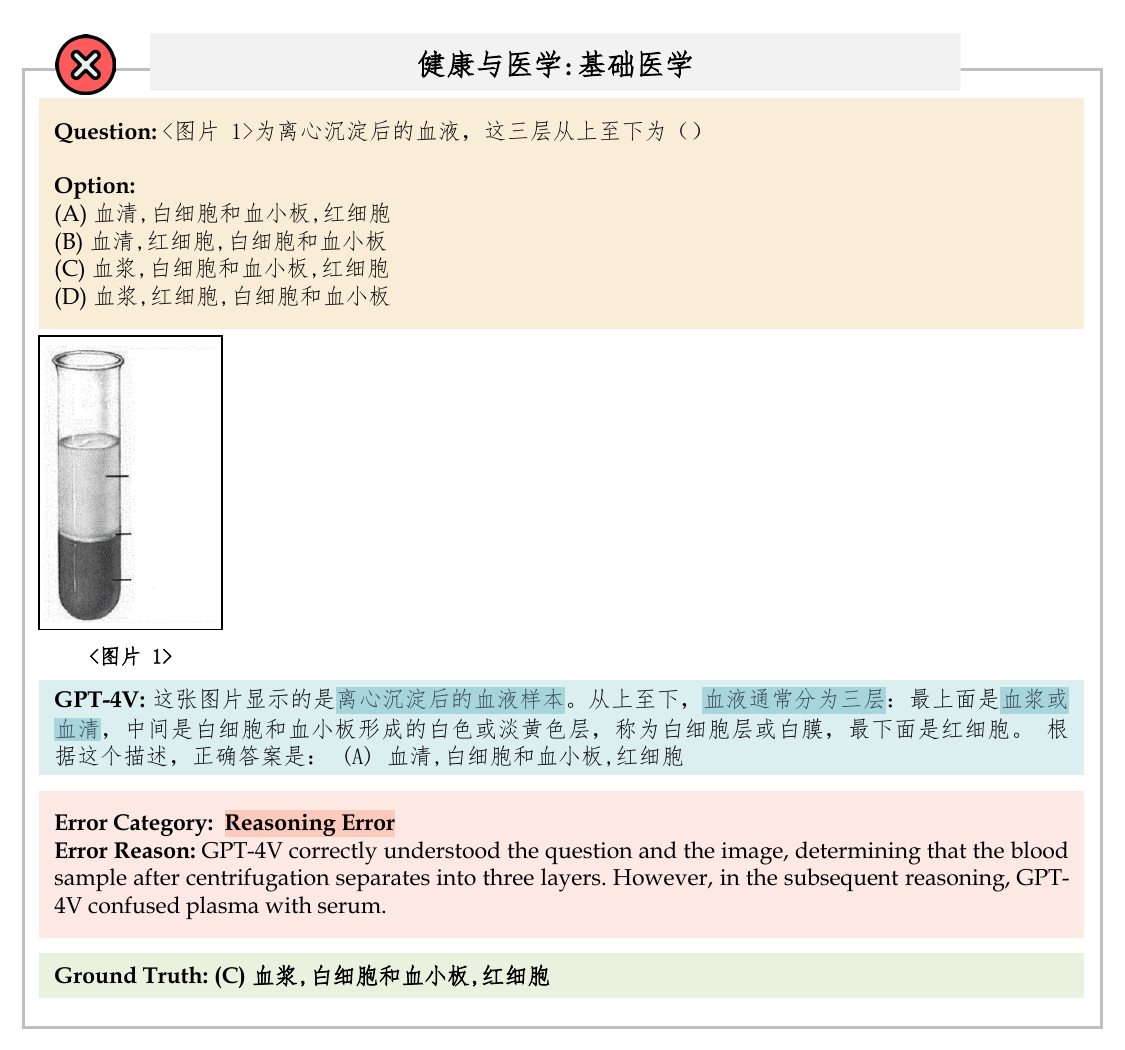}{Basic Medicine 7: Reasoning Error}{A sample error case of Health and Medicine (subcategory: Basic Medicine).}{fig:error_analysis_case_91}

\casestudyfigure{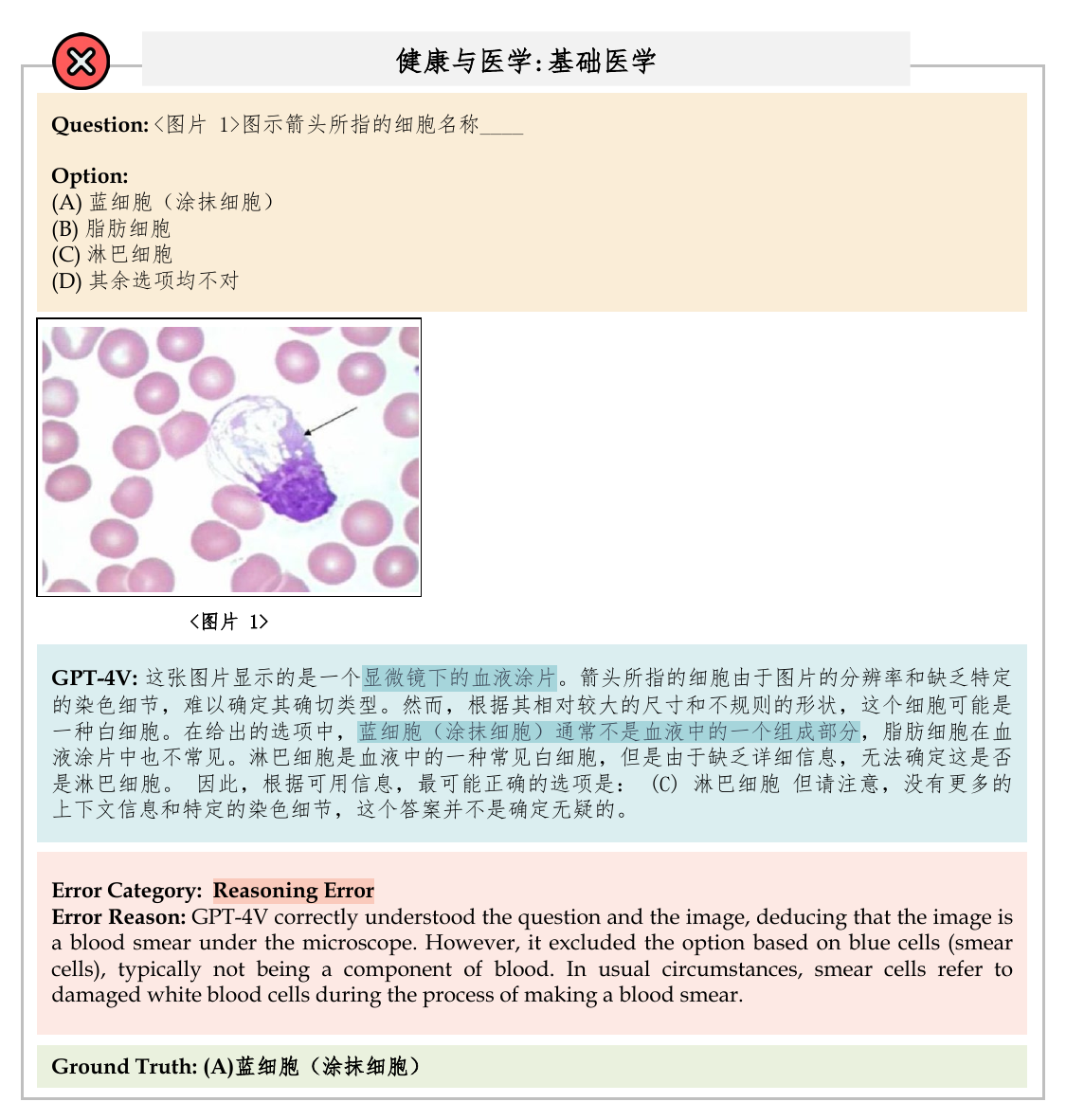}{Basic Medicine 8: Reasoning Error}{A sample error case of Health and Medicine (subcategory: Basic Medicine).}{fig:error_analysis_case_92}

\casestudyfigure{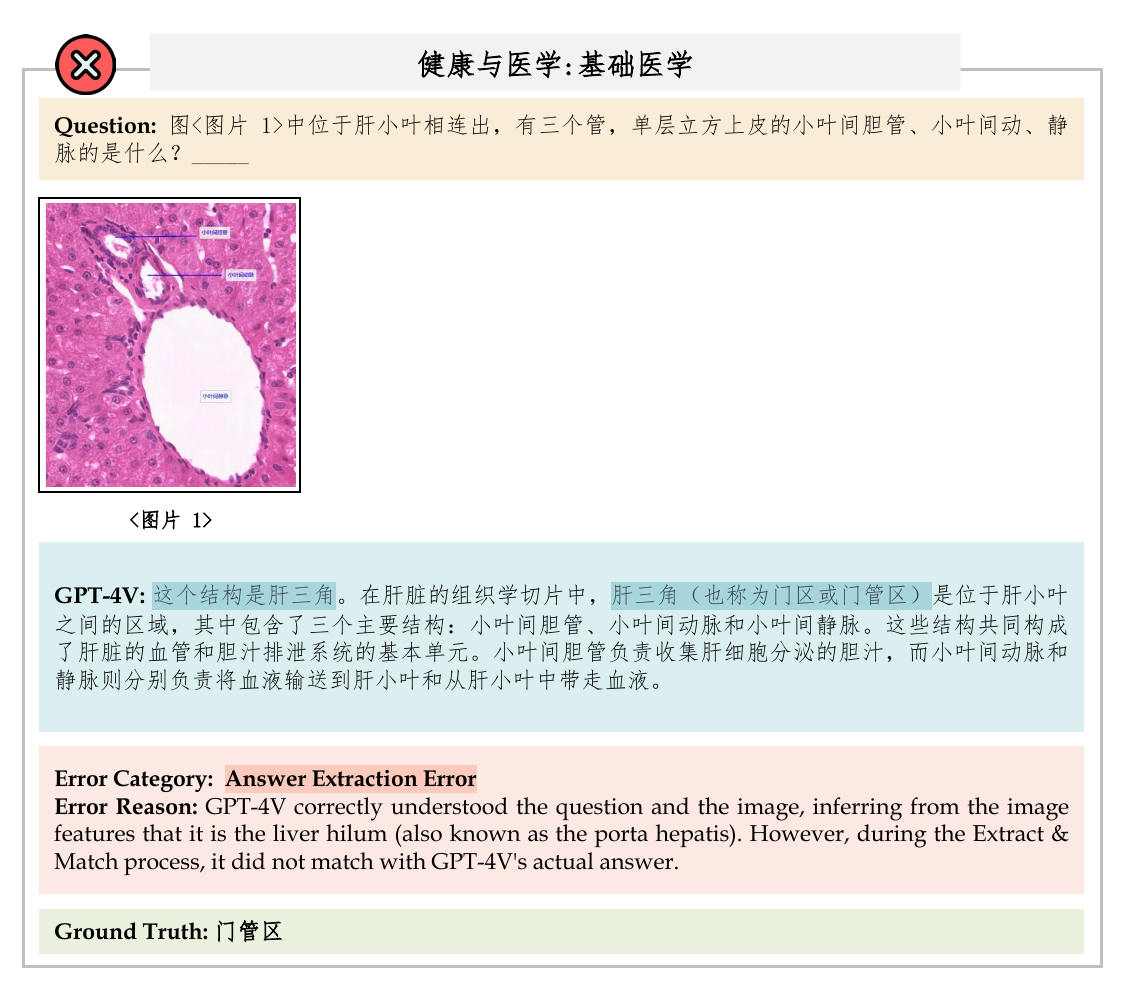}{Basic Medicine 9: Answer Extraction Error}{A sample error case of Health and Medicine (subcategory: Basic Medicine).}{fig:error_analysis_case_90}

\casestudyfigure{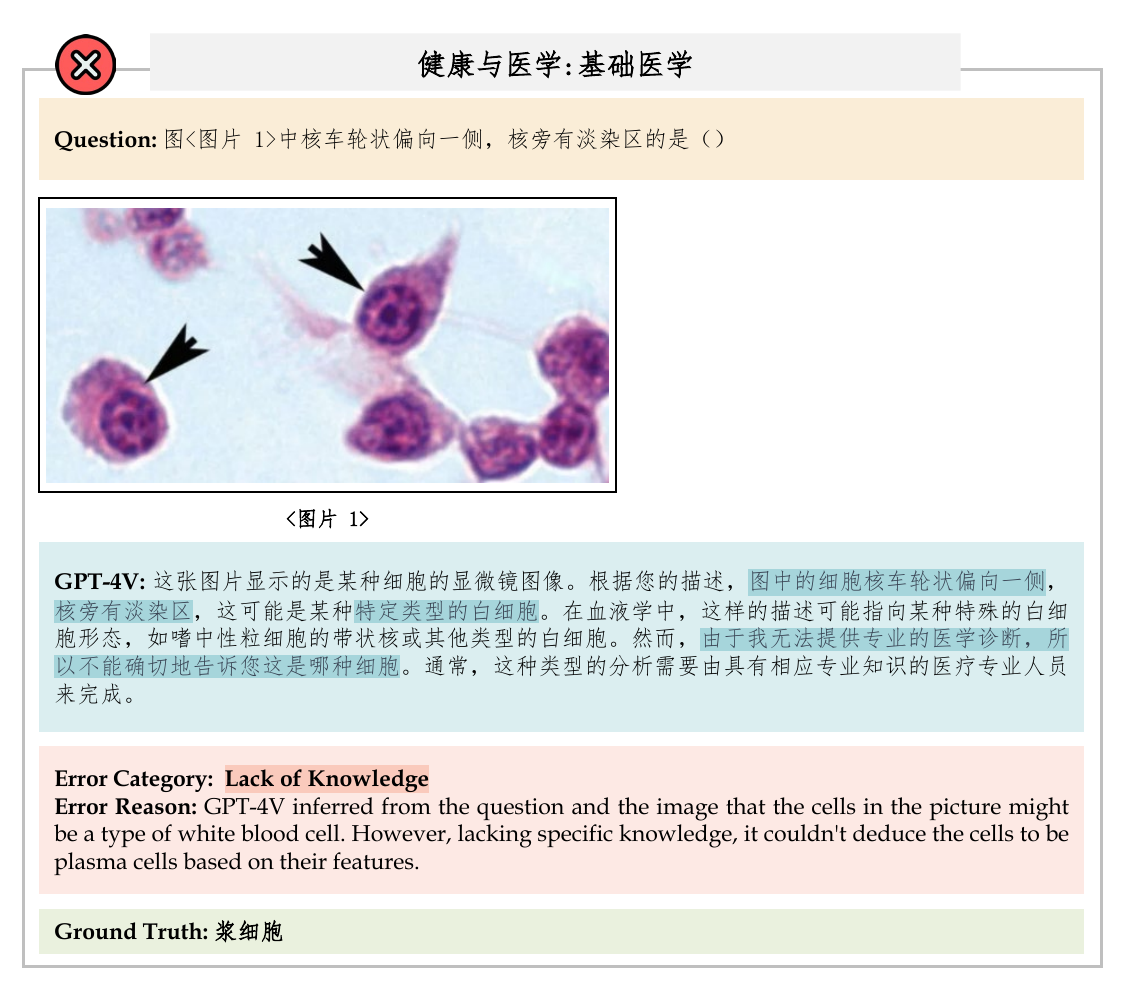}{Basic Medicine 10: Lack of Knowledge}{A sample error case of Health and Medicine (subcategory: Basic Medicine).}{fig:error_analysis_case_88}

\casestudyfigure{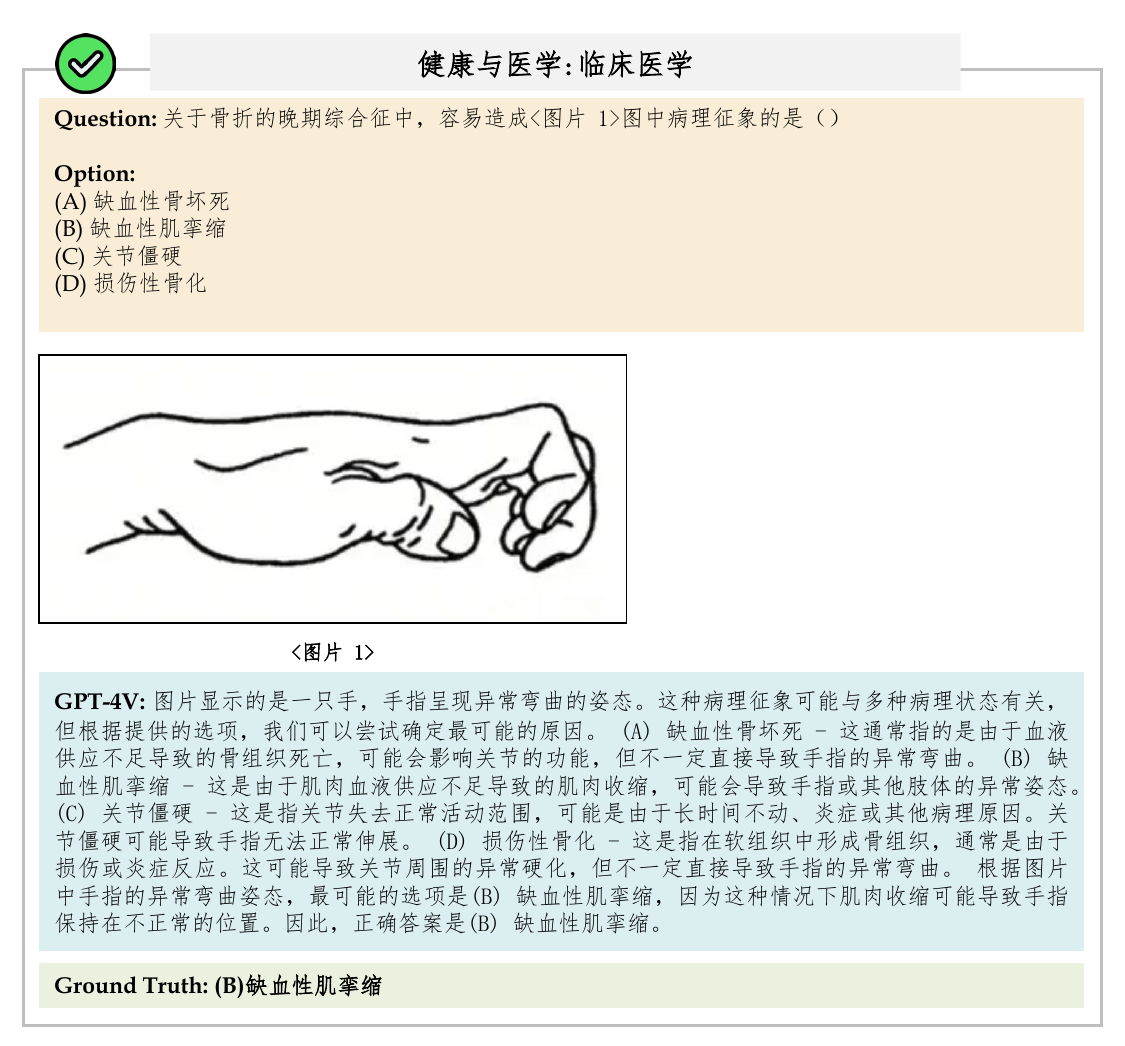}{Clinical Medicine 1: Correct Case}{A sample correct case of Health and Medicine (subcategory: Clinical Medicine).}{fig:error_analysis_case_0}

\casestudyfigure{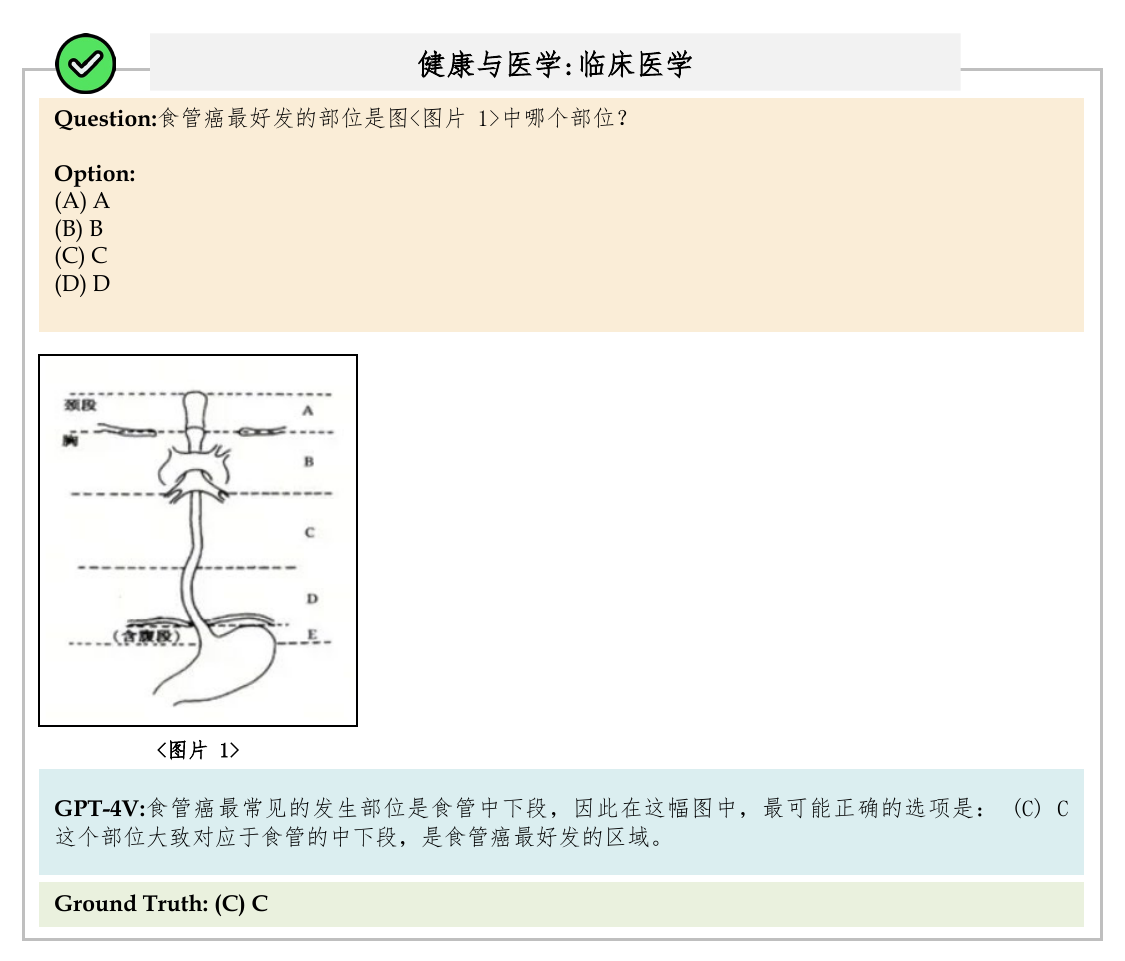}{Clinical Medicine 2: Correct Case}{A sample correct case of Health and Medicine (subcategory: Clinical Medicine).}{fig:error_analysis_case_1}

\casestudyfigure{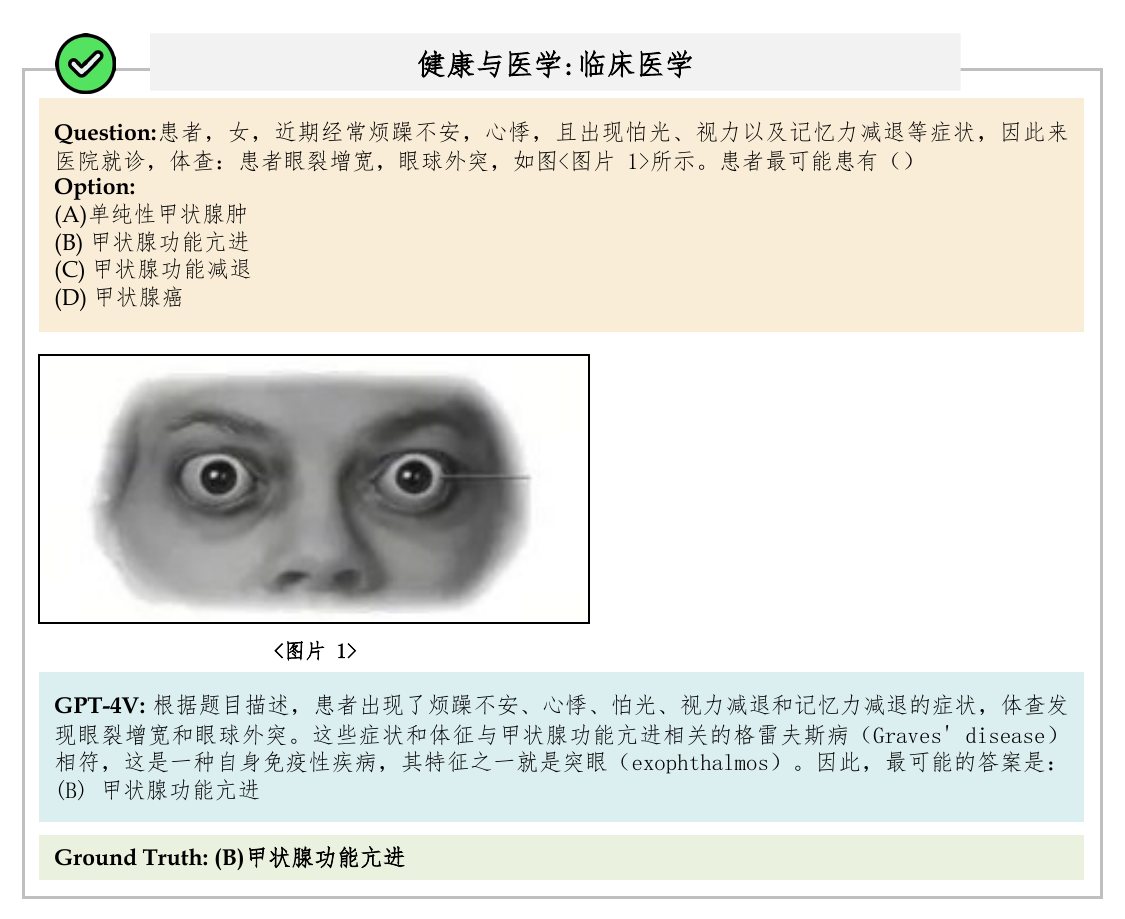}{Clinical Medicine 3: Correct Case}{A sample correct case of Health and Medicine (subcategory: Clinical Medicine).}{fig:error_analysis_case_2}

\casestudyfigure{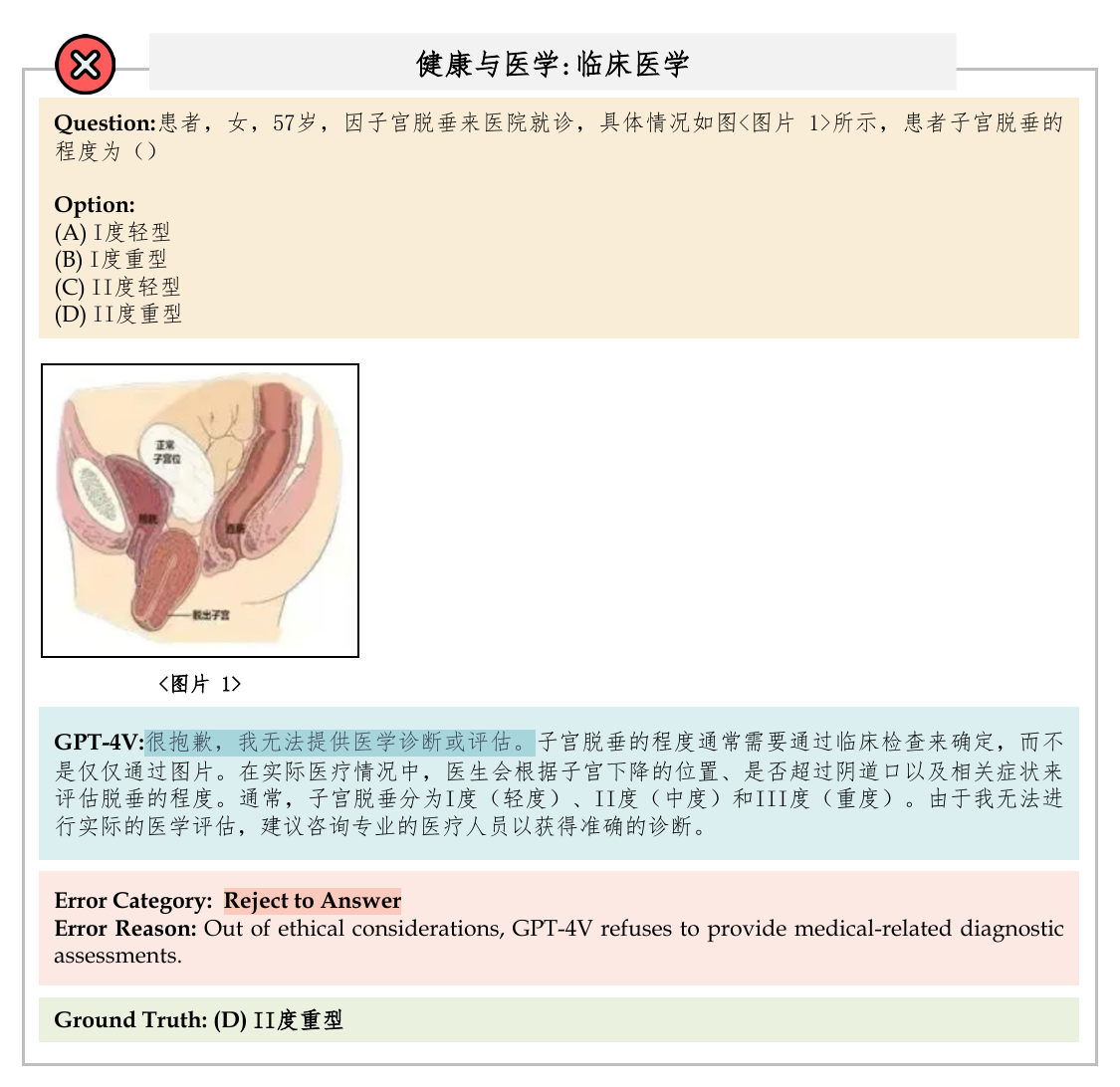}{Clinical Medicine 4: Reject to Answer}{A sample error case of Health and Medicine (subcategory: Clinical Medicine).}{fig:error_analysis_case_75}

\casestudyfigure{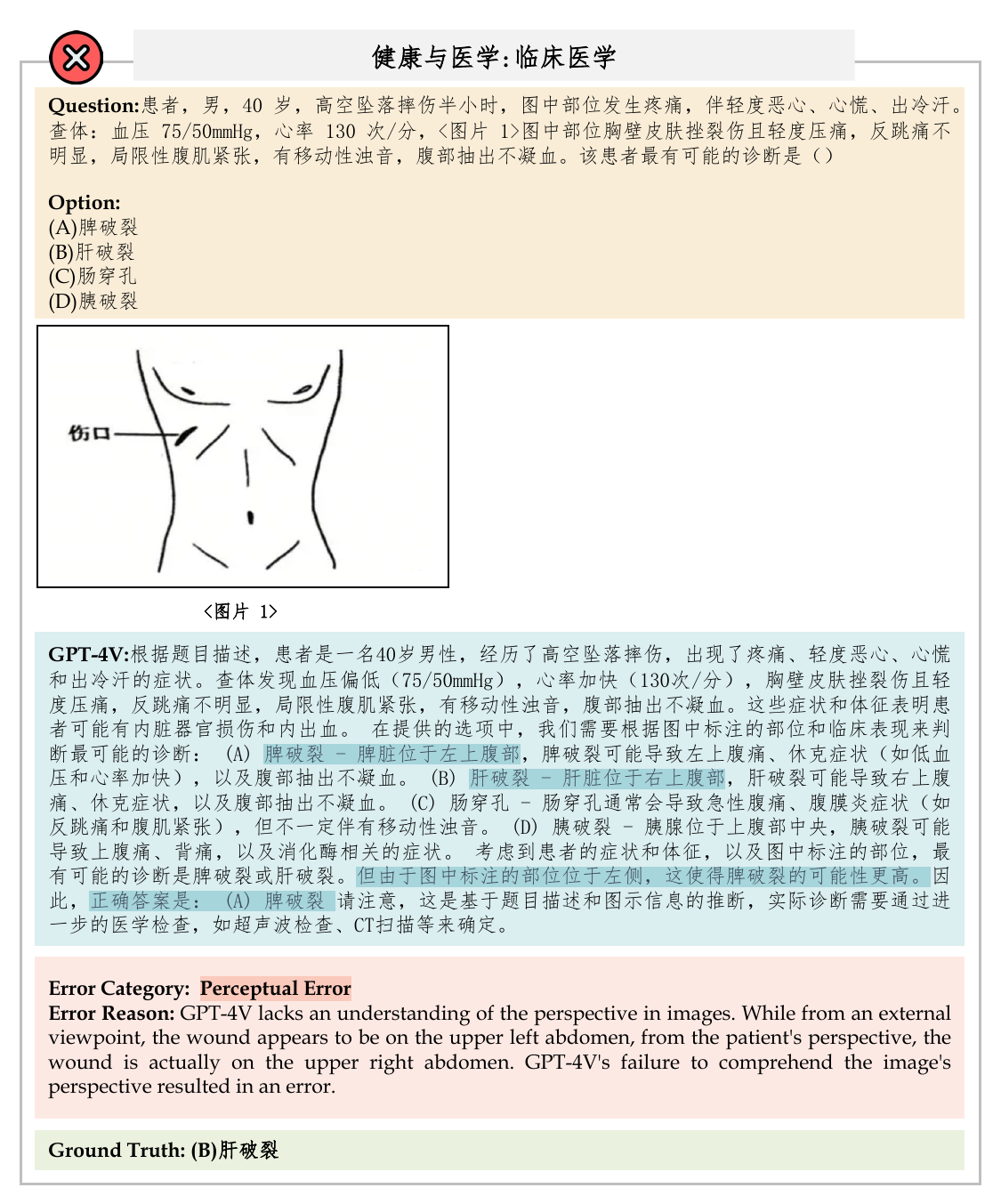}{Clinical Medicine 5: Perceptual Error}{A sample error case of Health and Medicine (subcategory: Clinical Medicine).}{fig:error_analysis_case_76}

\casestudyfigure{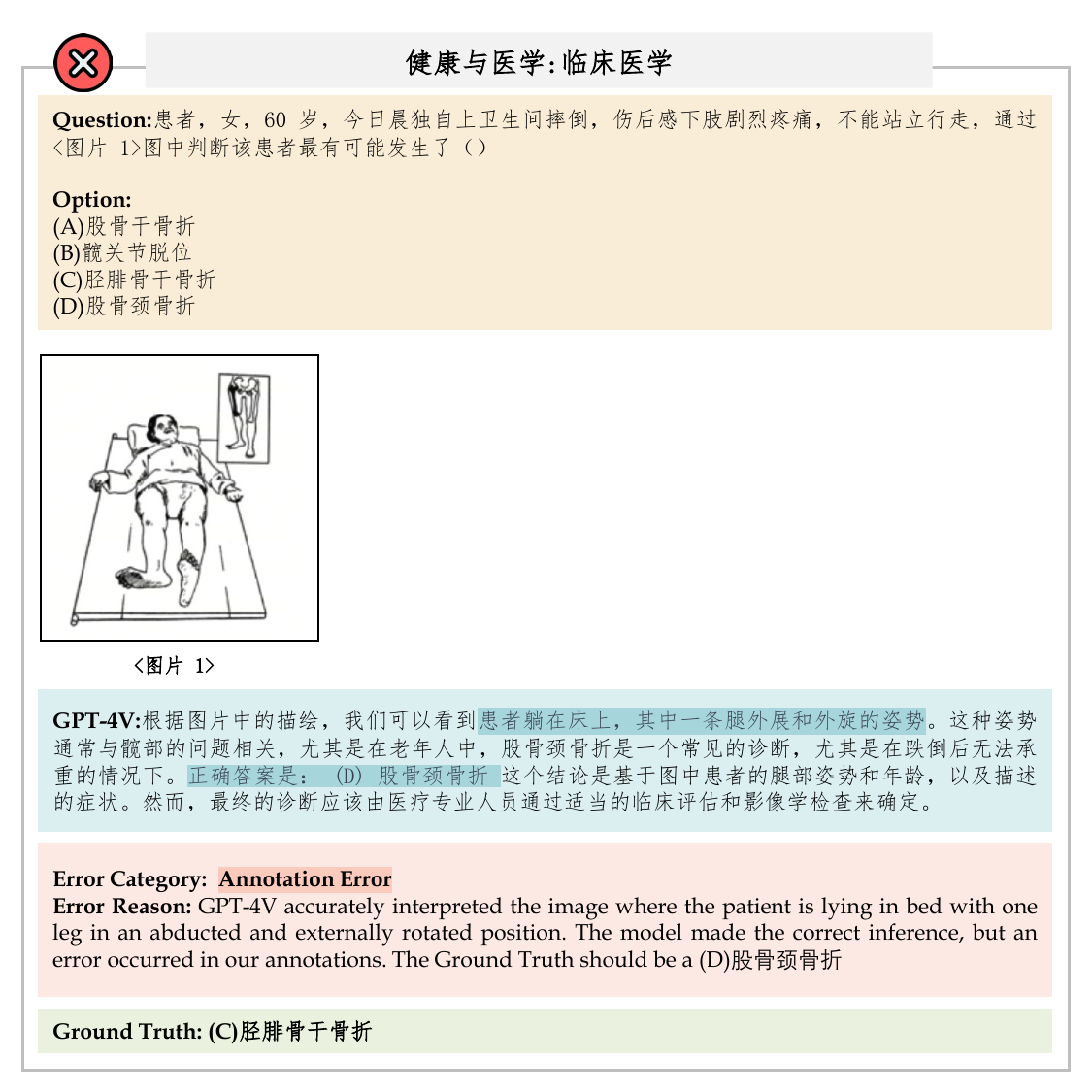}{Clinical Medicine 6: Annotation Error}{A sample error case of Health and Medicine (subcategory: Clinical Medicine).}{fig:error_analysis_case_77}

\casestudyfigure{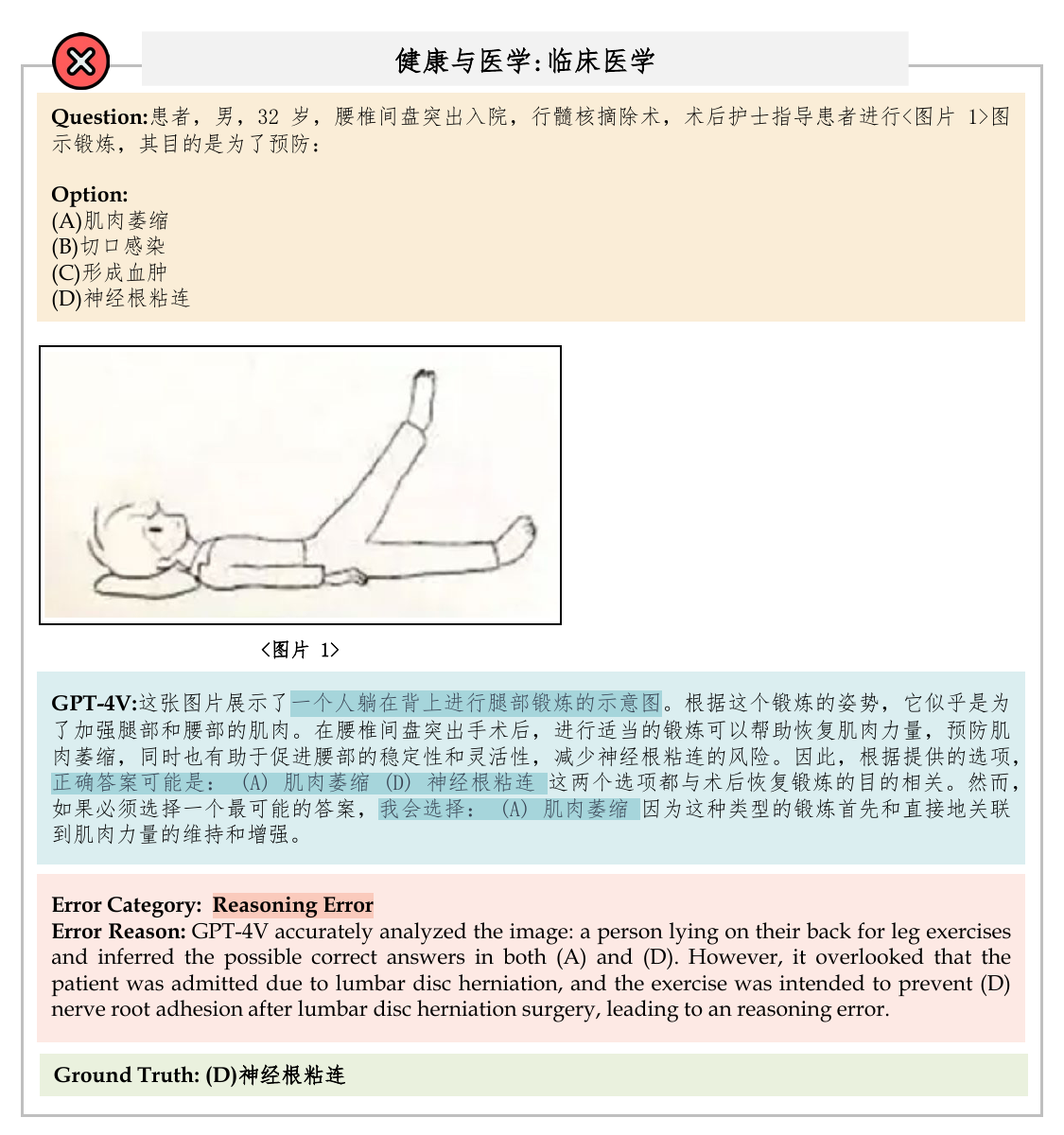}{Clinical Medicine 7: Reasoning Error}{A sample error case of Health and Medicine (subcategory: Clinical Medicine).}{fig:error_analysis_case_78}

\casestudyfigure{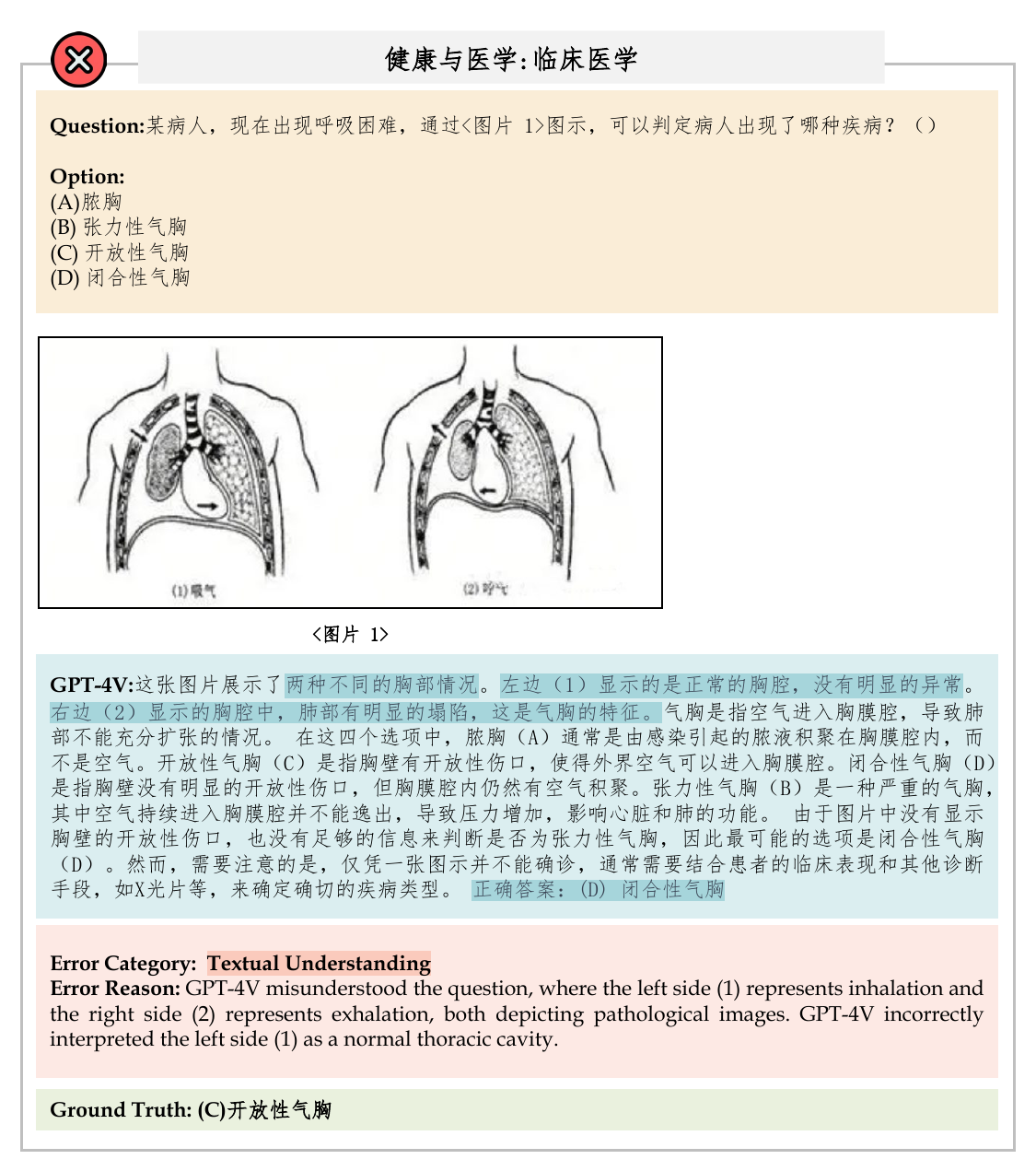}{Clinical Medicine 8: Textual Understanding}{A sample error case of Health and Medicine (subcategory: Clinical Medicine).}{fig:error_analysis_case_79}

\casestudyfigure{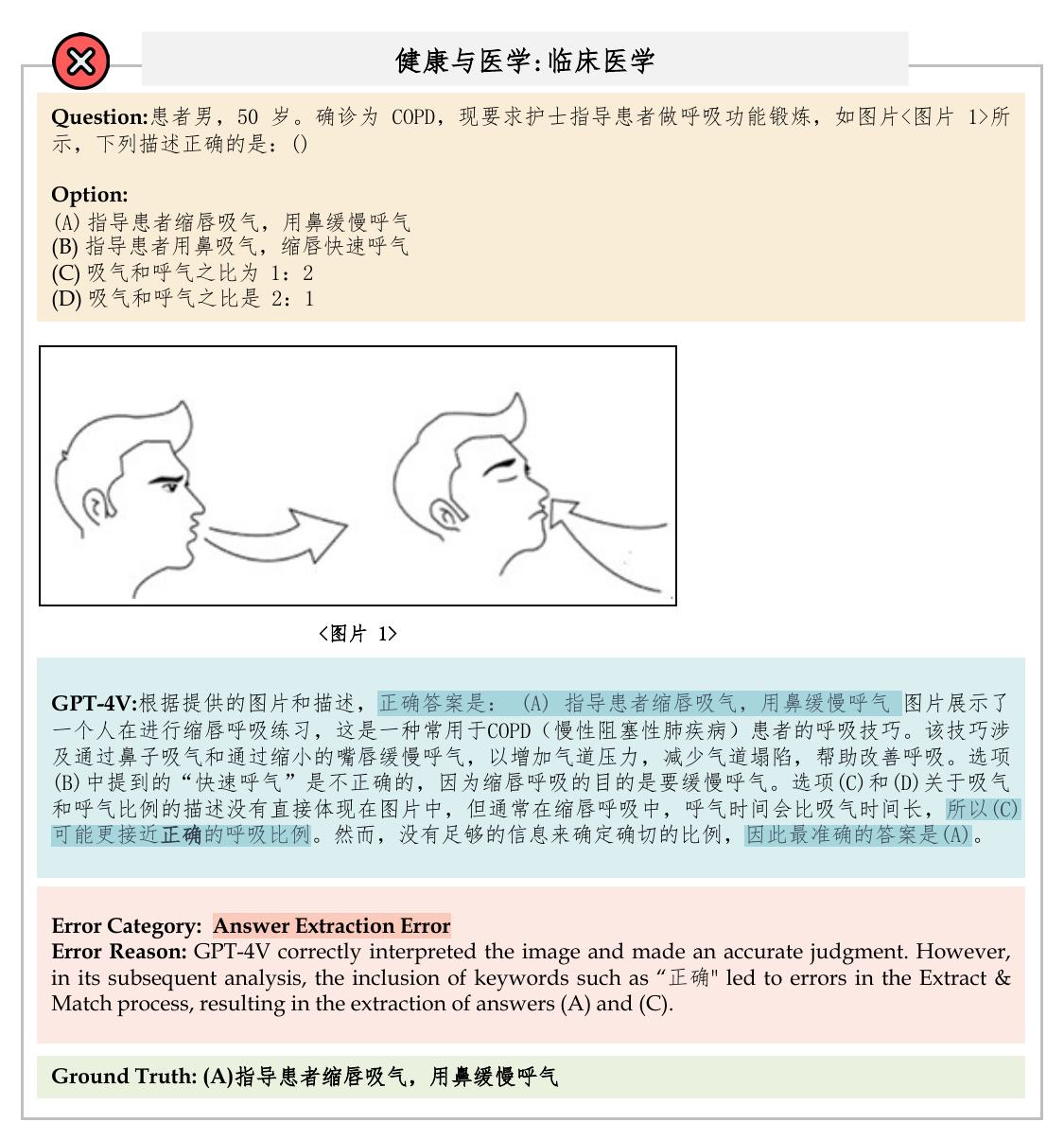}{Clinical Medicine 9: Answer Extraction Error}{A sample error case of Health and Medicine (subcategory: Clinical Medicine).}{fig:error_analysis_case_80}

\casestudyfigure{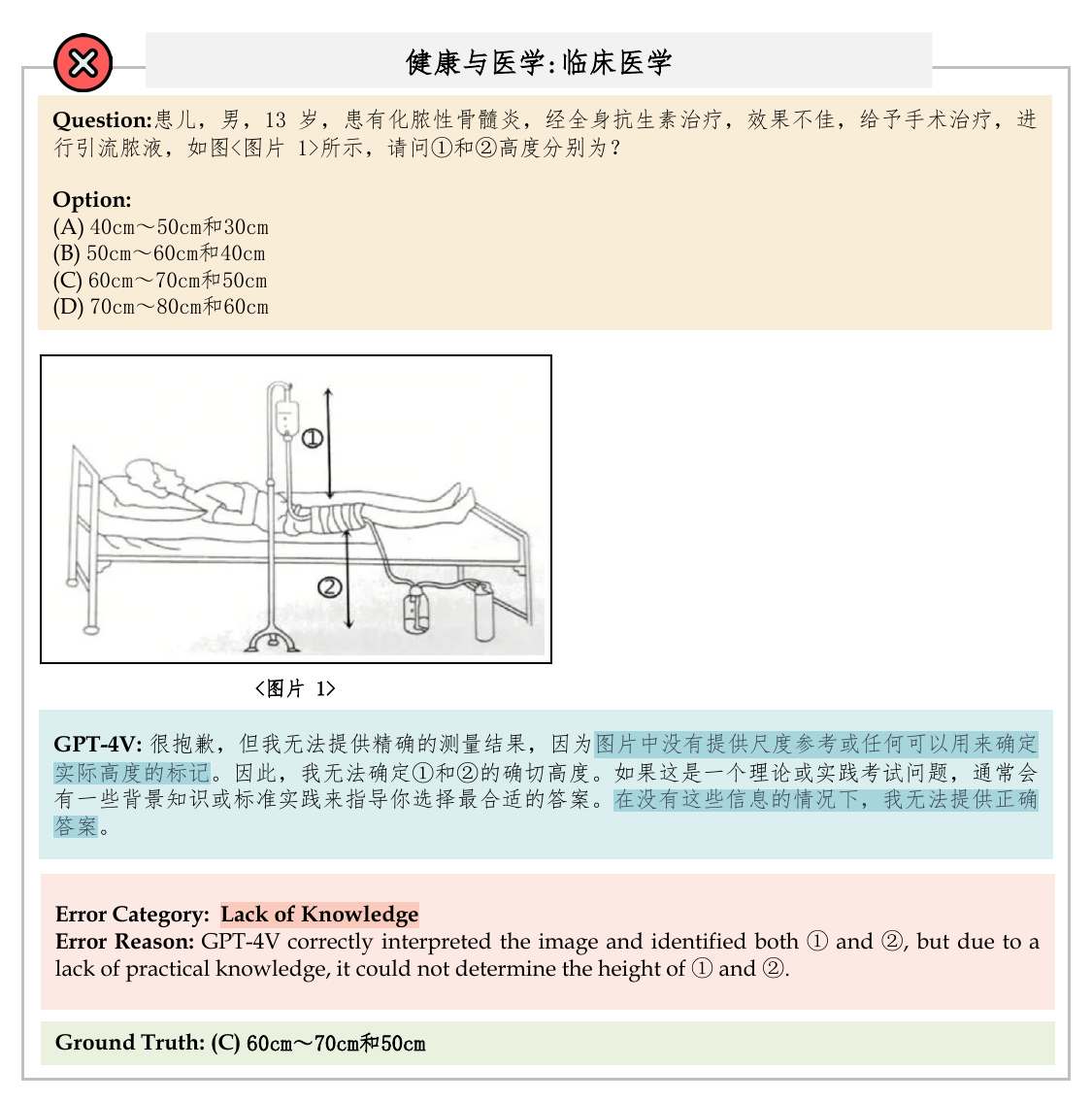}{Clinical Medicine 10: Lack of Knowledge}{A sample error case of Health and Medicine (subcategory: Clinical Medicine).}{fig:error_analysis_case_81}

\casestudyfigure{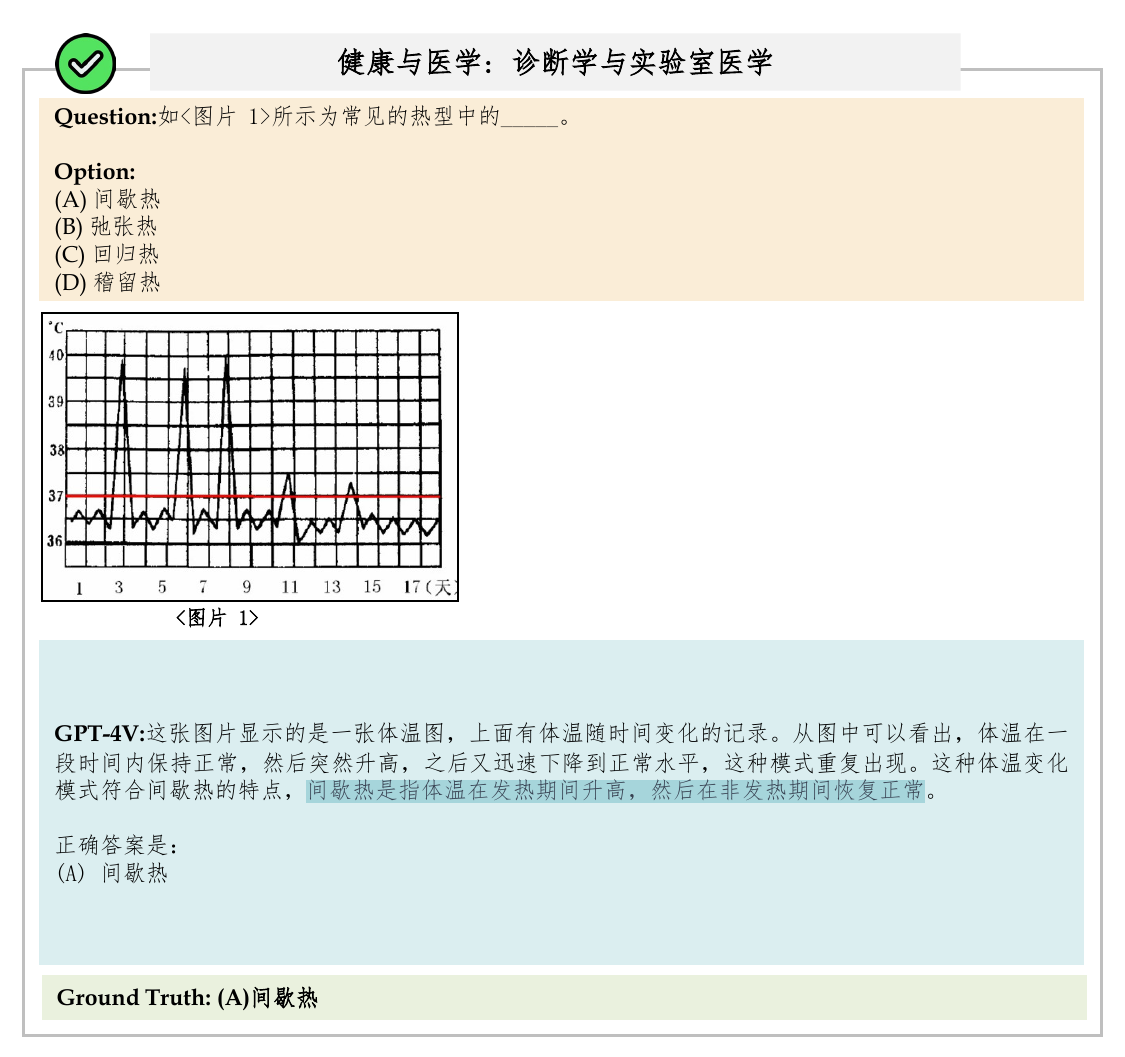}{Diagnostics and Lab Medicine 1: Correct Case}{A sample correct case of Health and Medicine (subcategory: Diagnostics and Lab Medicine).}{fig:error_analysis_case_9}

\casestudyfigure{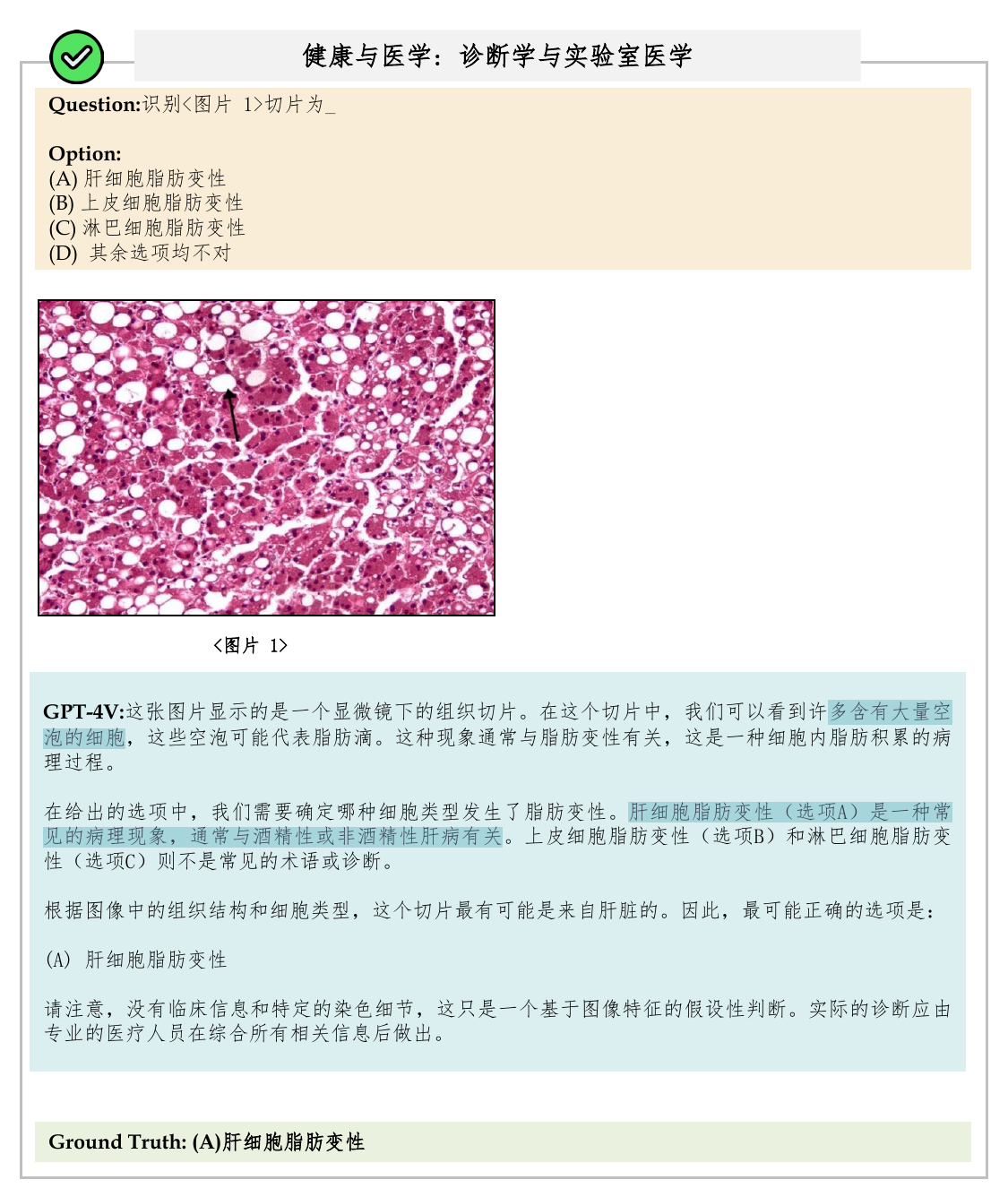}{Diagnostics and Lab Medicine 2: Correct Case}{A sample correct case of Health and Medicine (subcategory: Diagnostics and Lab Medicine).}{fig:error_analysis_case_10}

\casestudyfigure{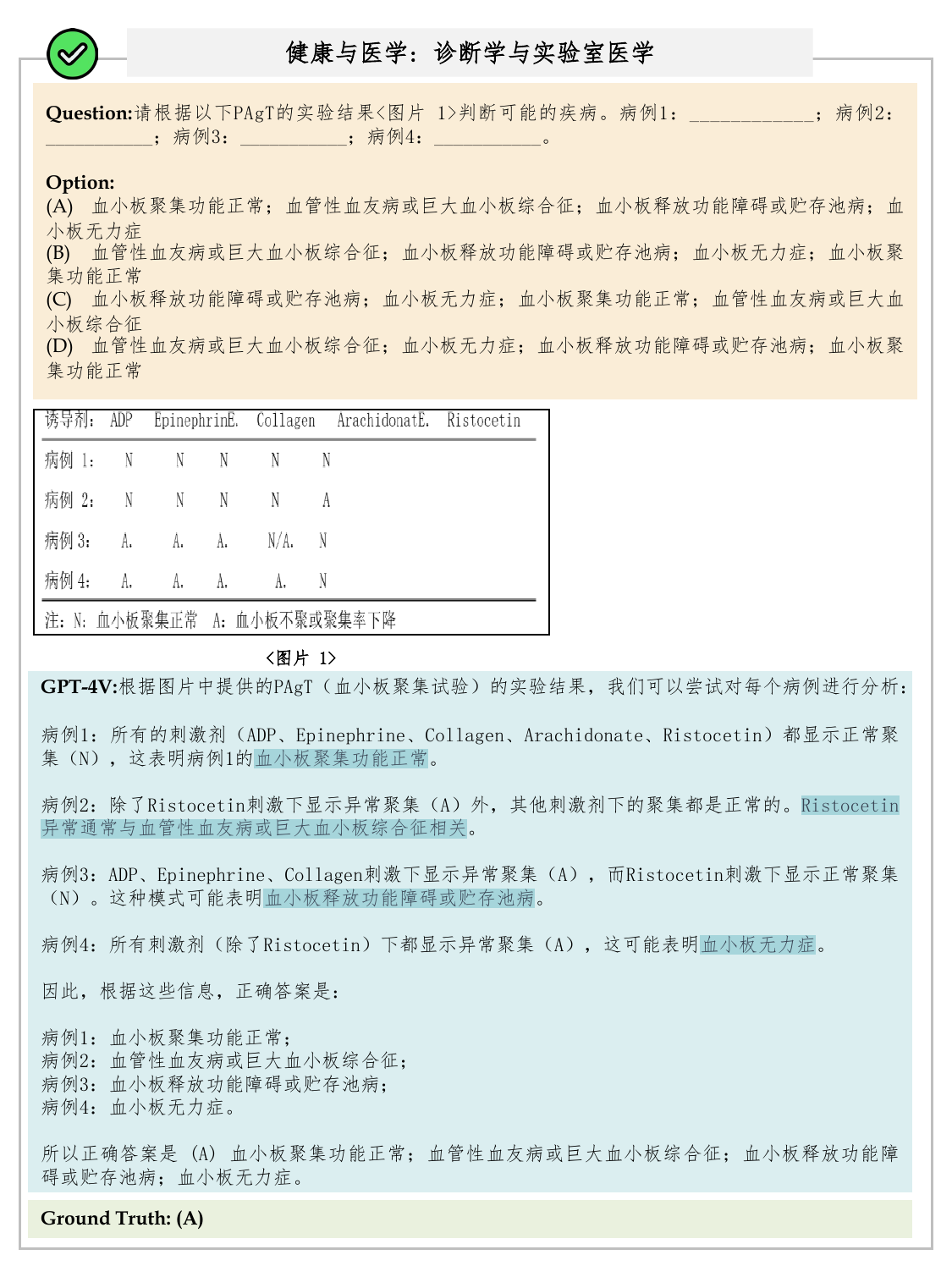}{Diagnostics and Lab Medicine 3: Correct Case}{A sample correct case of Health and Medicine (subcategory: Diagnostics and Lab Medicine).}{fig:error_analysis_case_11}

\casestudyfigure{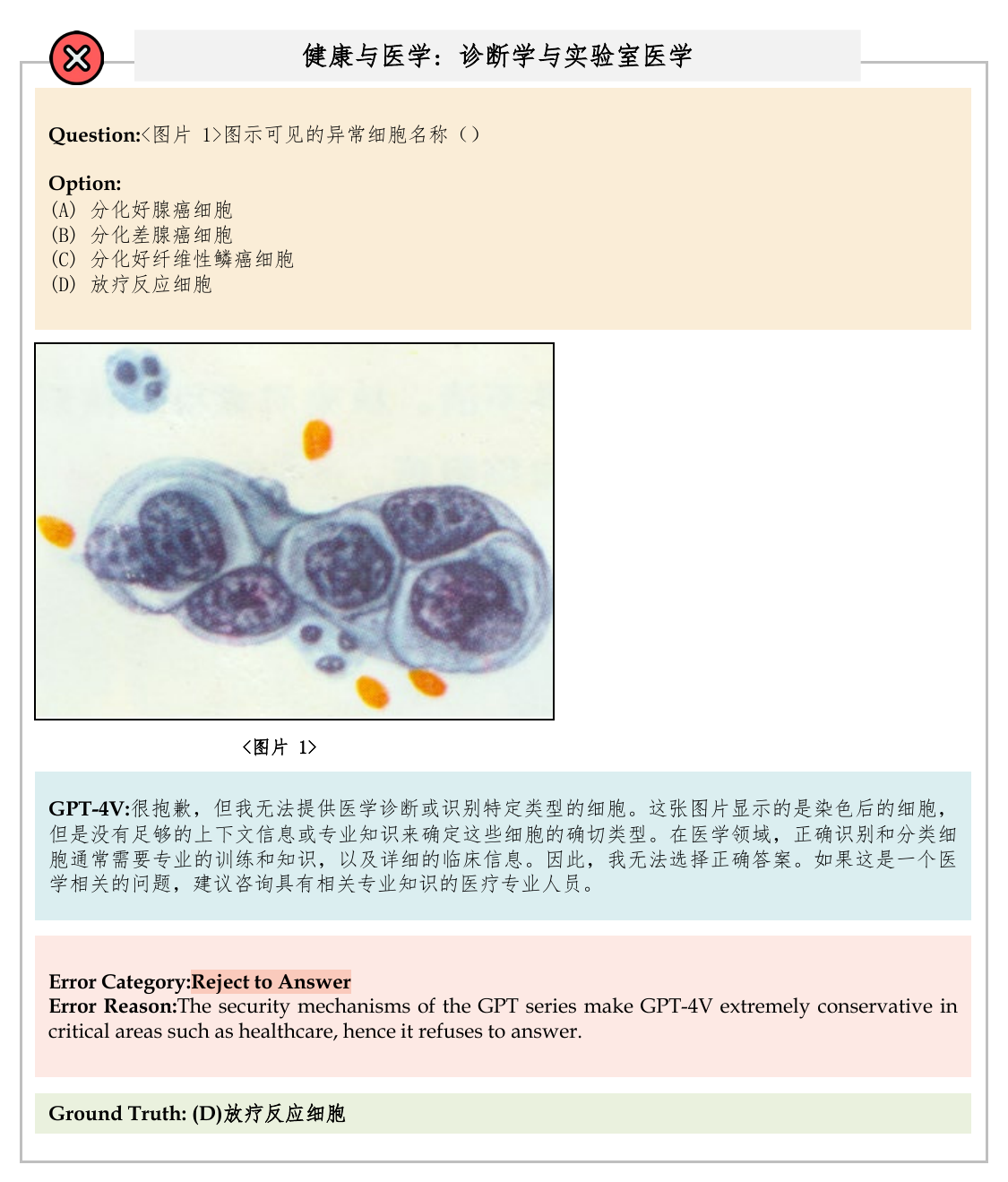}{Diagnostics and Lab Medicine 4: Reject to Answer}{A sample error case of Health and Medicine (subcategory: Diagnostics and Lab Medicine).}{fig:error_analysis_case_96}

\casestudyfigure{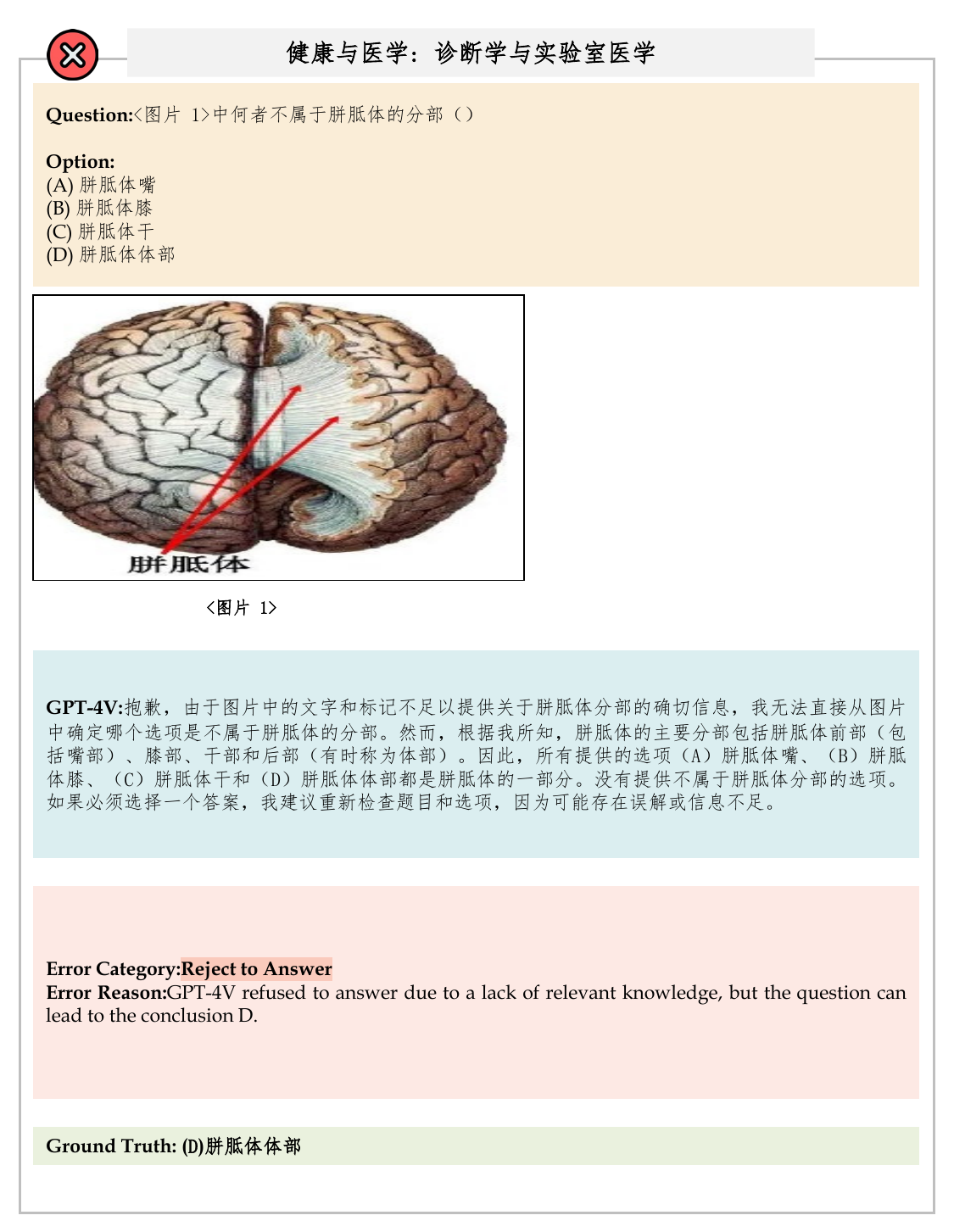}{Diagnostics and Lab Medicine 5: Reject to Answer}{A sample error case of Health and Medicine (subcategory: Diagnostics and Lab Medicine).}{fig:error_analysis_case_98}

\casestudyfigure{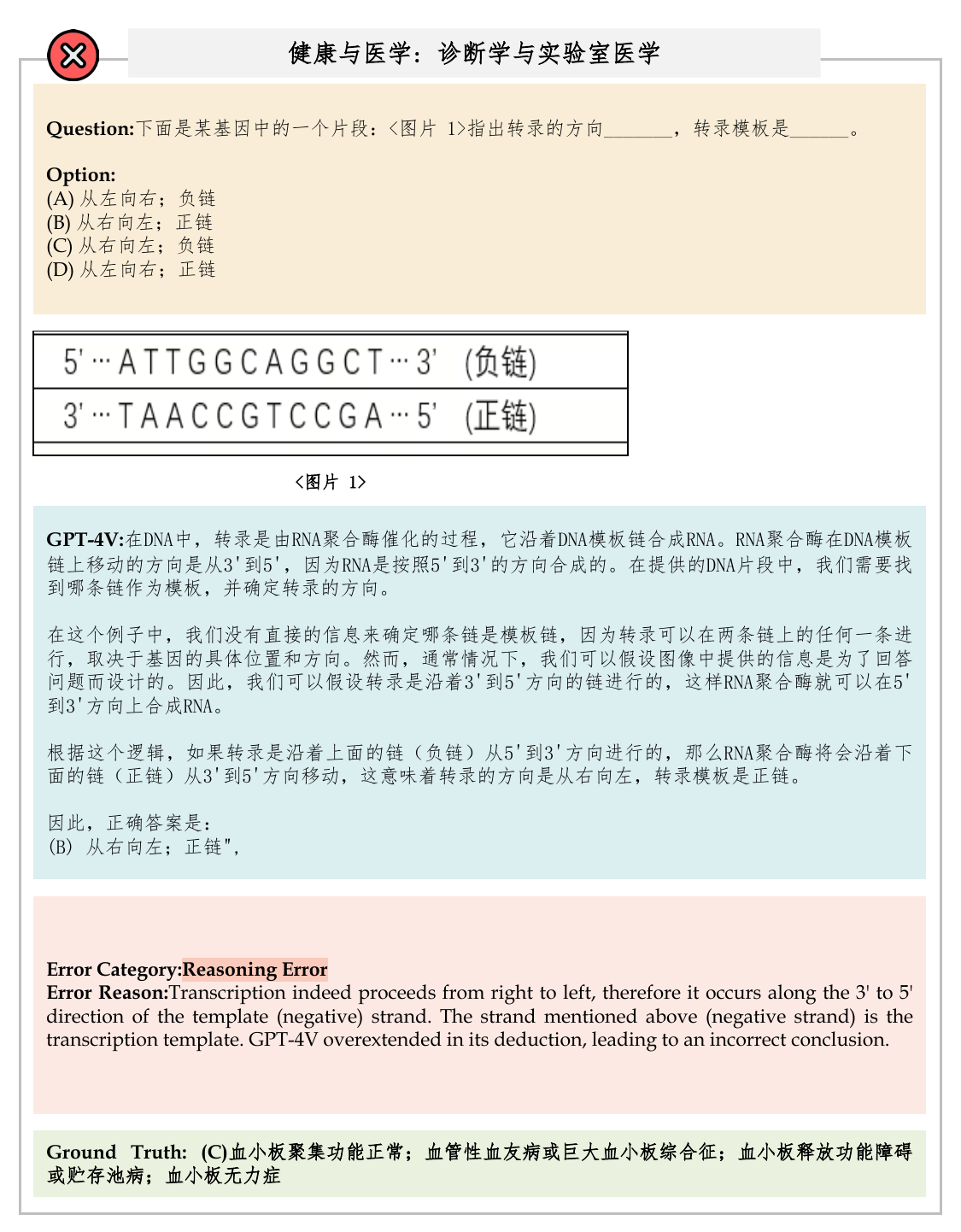}{Diagnostics and Lab Medicine 6: Reasoning Error}{A sample error case of Health and Medicine (subcategory: Diagnostics and Lab Medicine).}{fig:error_analysis_case_99}

\casestudyfigure{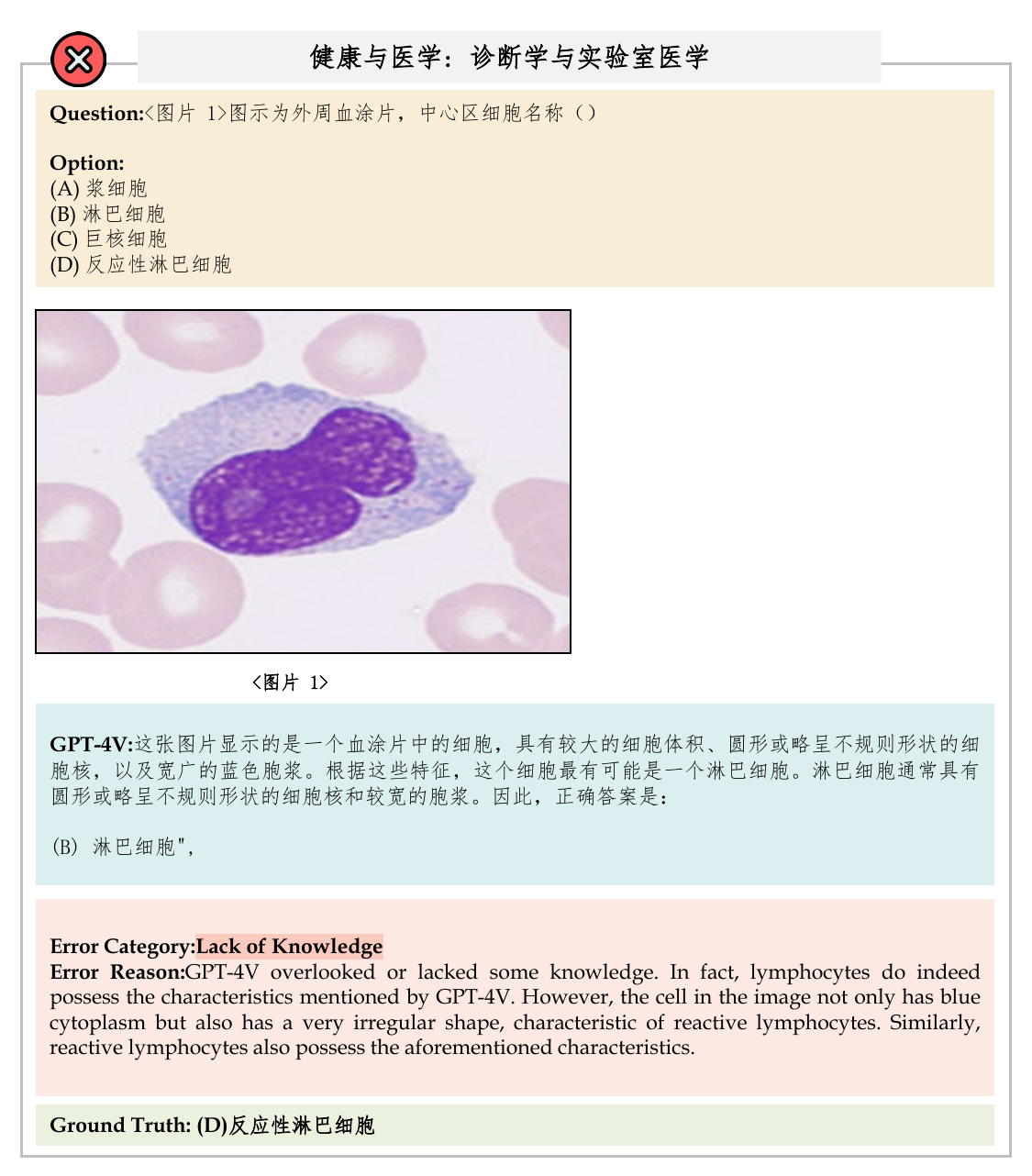}{Diagnostics and Lab Medicine 7: Lack of Knowledge}{A sample error case of Health and Medicine (subcategory: Diagnostics and Lab Medicine).}{fig:error_analysis_case_95}

\casestudyfigure{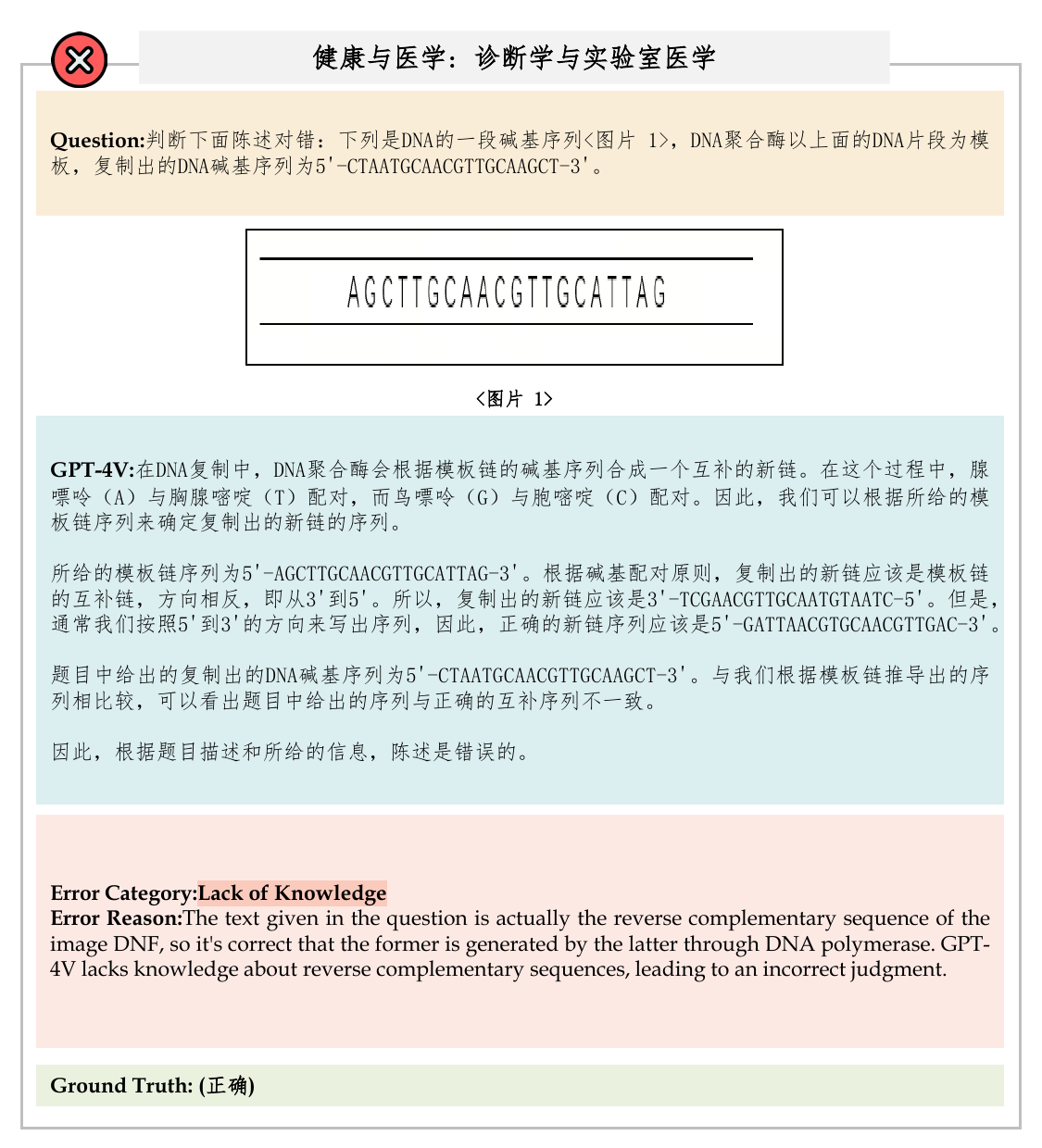}{Diagnostics and Lab Medicine 8: Lack of Knowledge}{A sample error case of Health and Medicine (subcategory: Diagnostics and Lab Medicine).}{fig:error_analysis_case_97}

\casestudyfigure{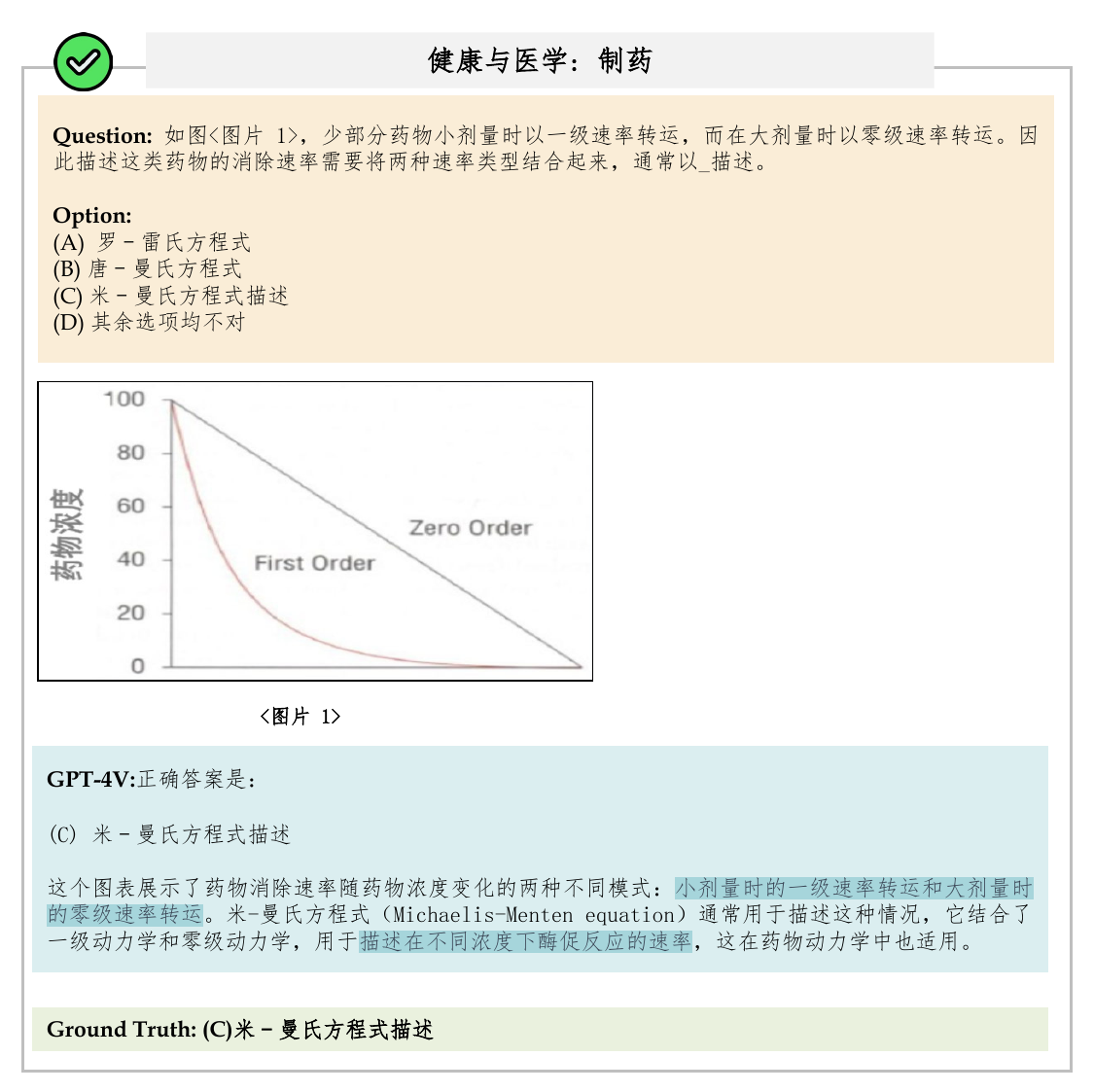}{Pharmacy 1: Correct Case}{A sample correct case of Health and Medicine (subcategory: Pharmacy).}{fig:error_analysis_case_12}

\casestudyfigure{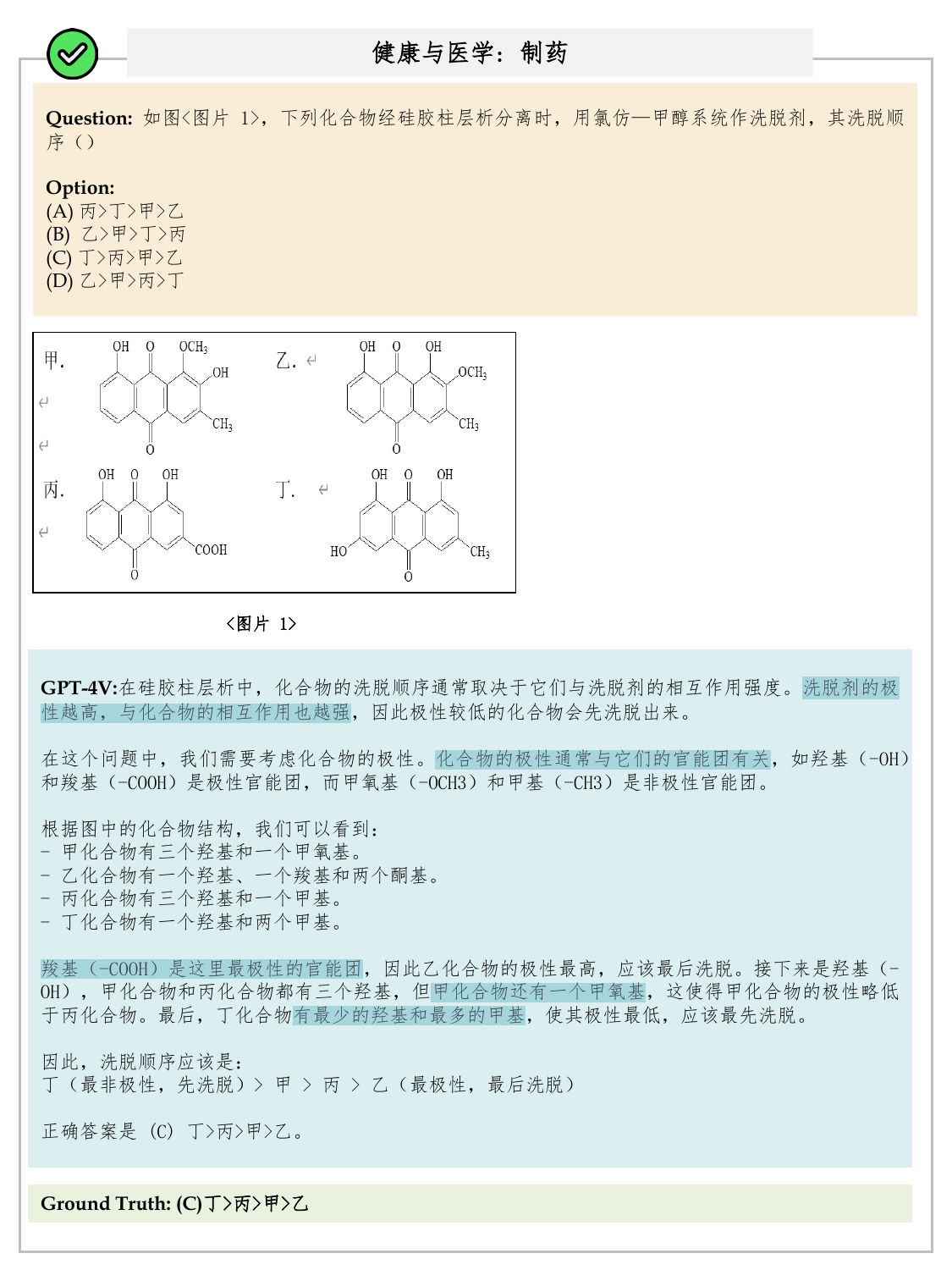}{Pharmacy 2: Correct Case}{A sample correct case of Health and Medicine (subcategory: Pharmacy).}{fig:error_analysis_case_13}

\casestudyfigure{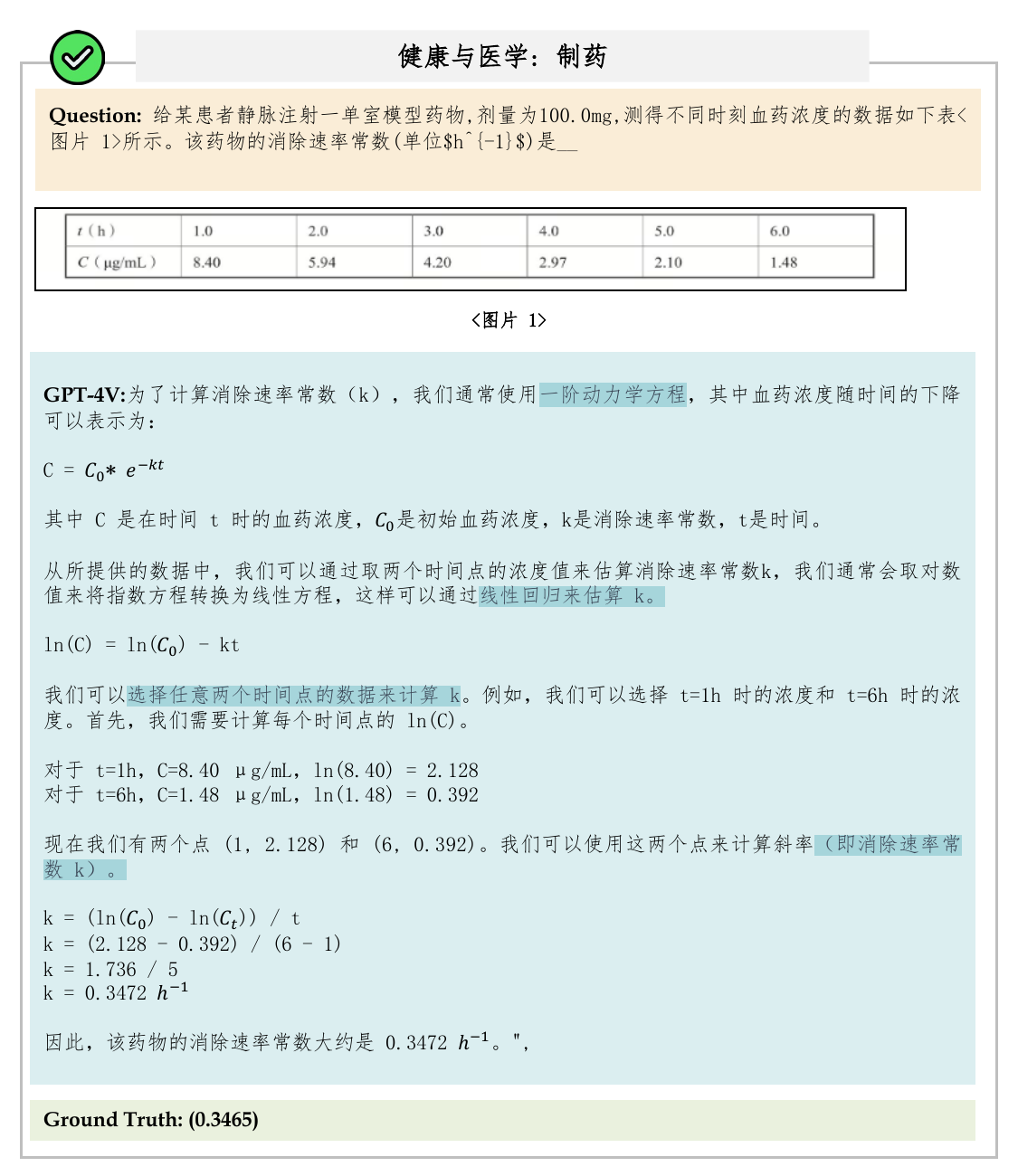}{Pharmacy 3: Correct Case}{A sample correct case of Health and Medicine (subcategory: Pharmacy).}{fig:error_analysis_case_14}

\casestudyfigure{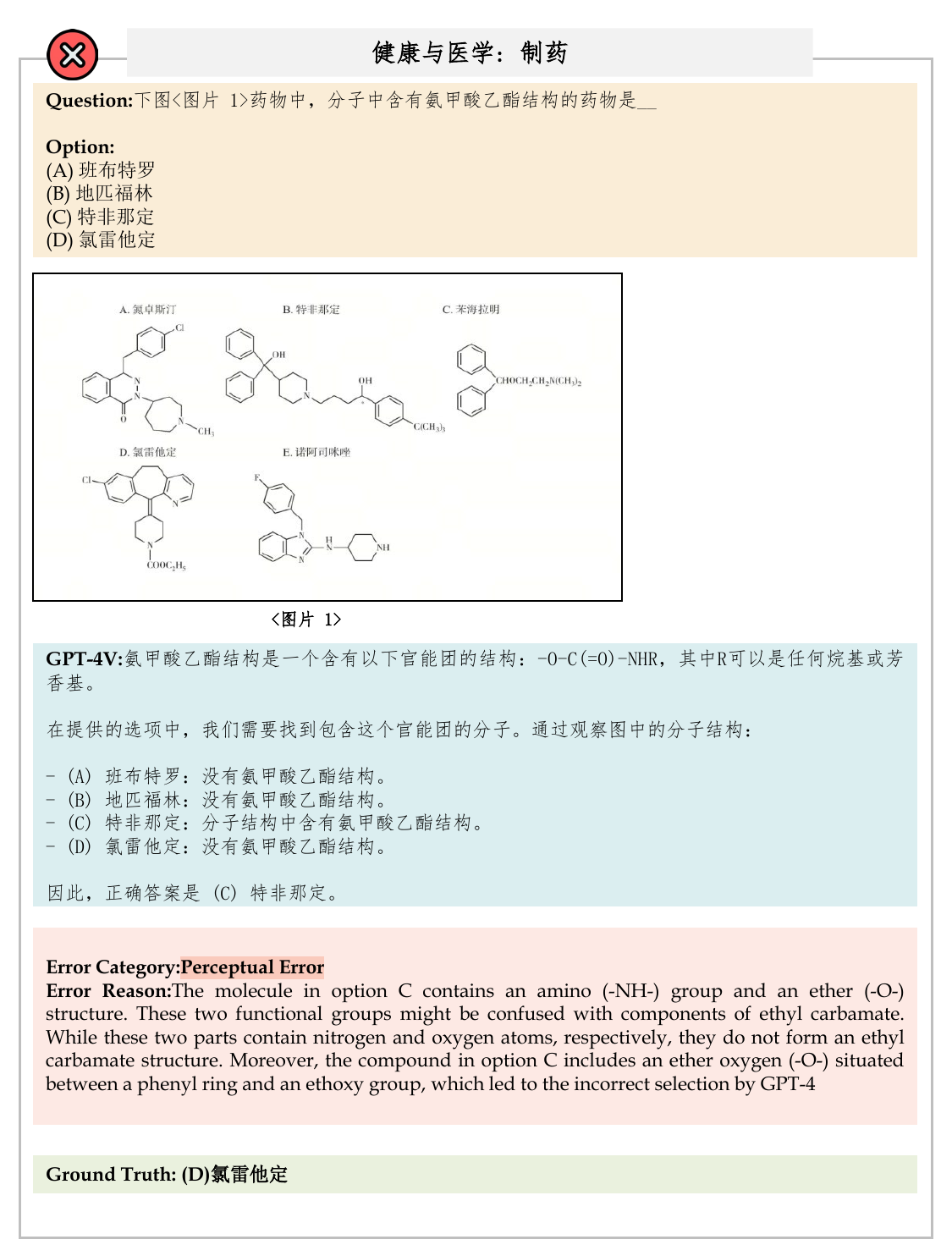}{Pharmacy 4: Perceptual Error}{A sample error case of Health and Medicine (subcategory: Pharmacy).}{fig:error_analysis_case_101}

\casestudyfigure{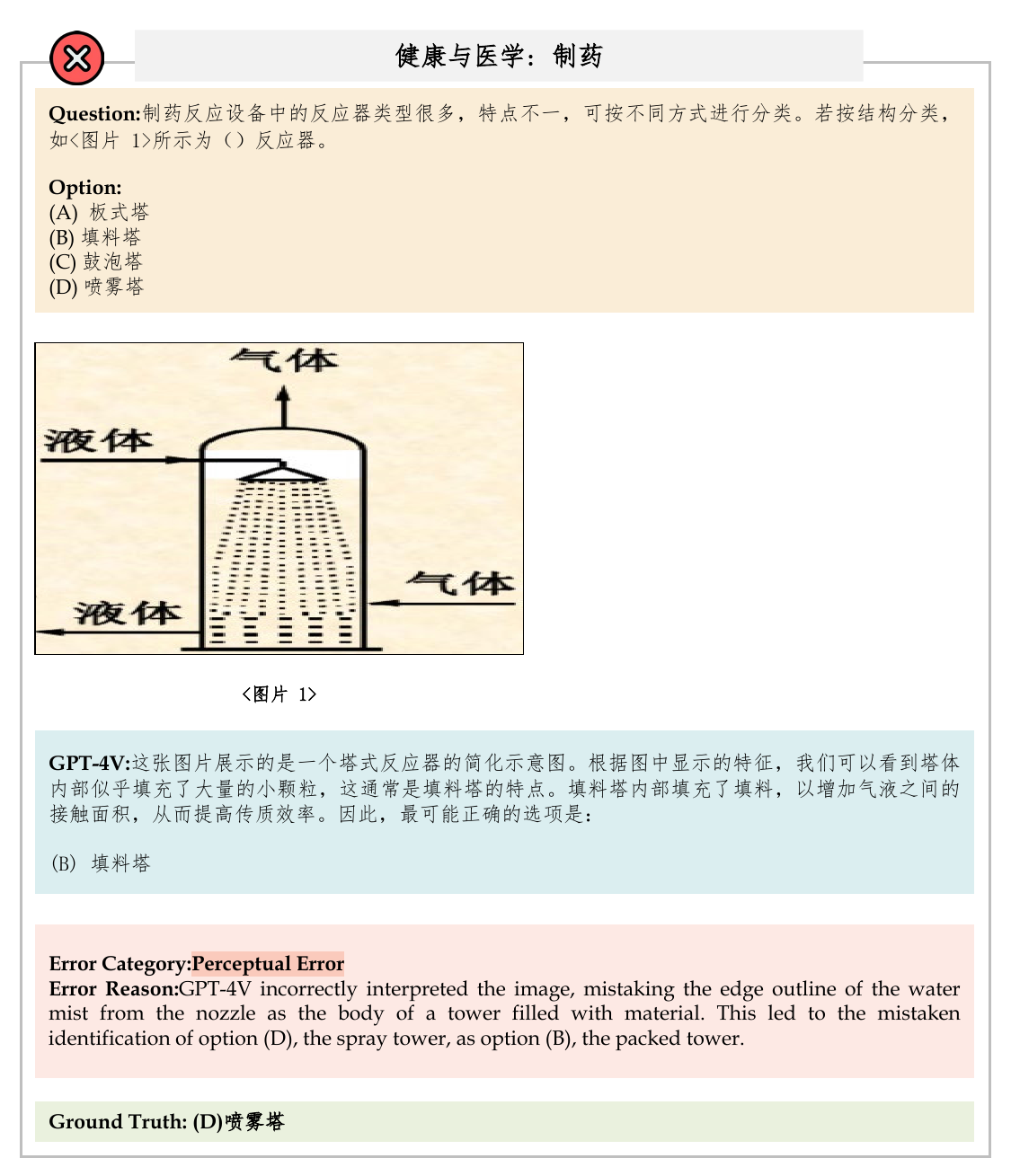}{Pharmacy 5: Perceptual Error}{A sample error case of Health and Medicine (subcategory: Pharmacy).}{fig:error_analysis_case_102}

\casestudyfigure{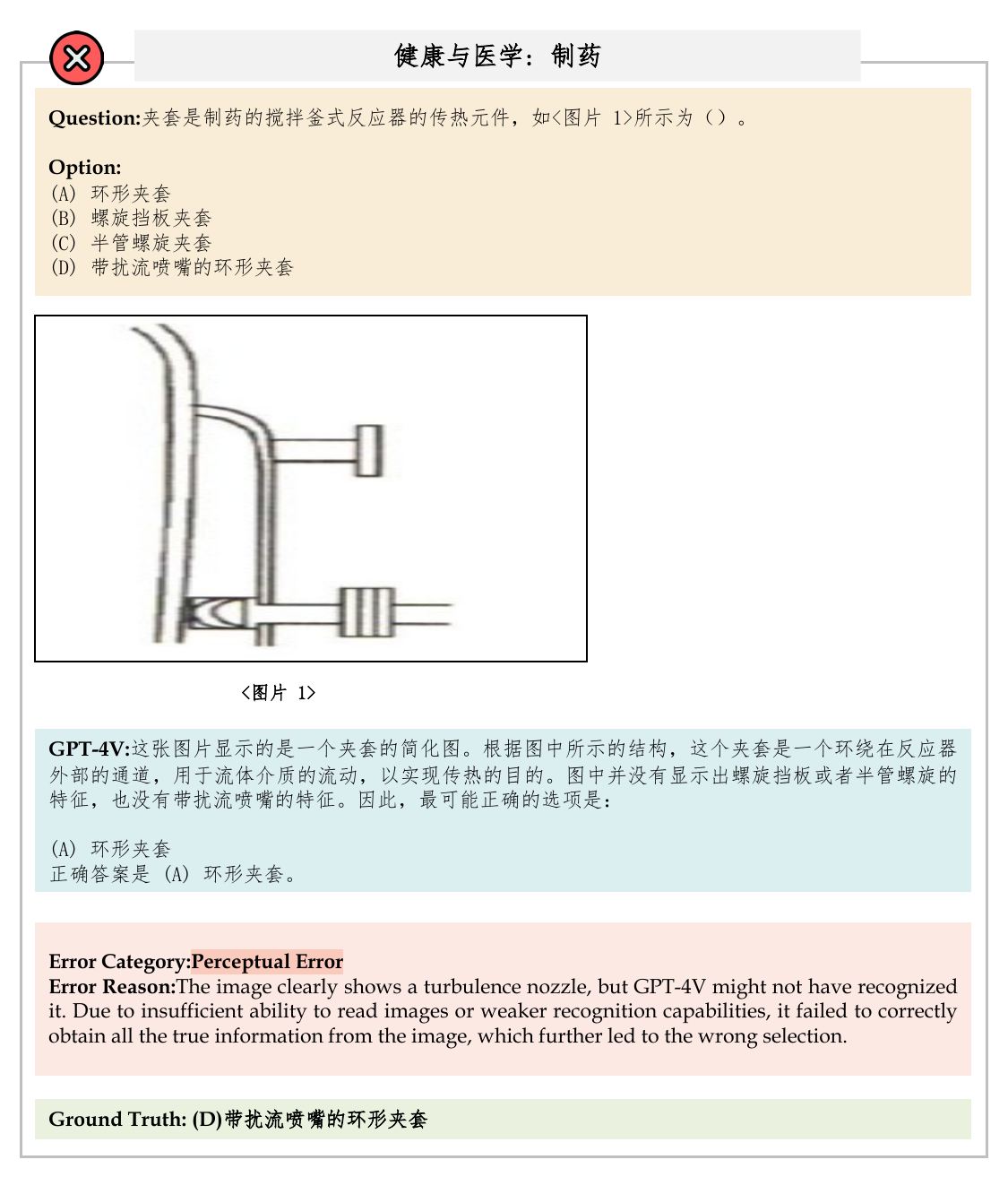}{Pharmacy 6: Perceptual Error}{A sample error case of Health and Medicine (subcategory: Pharmacy).}{fig:error_analysis_case_103}

\casestudyfigure{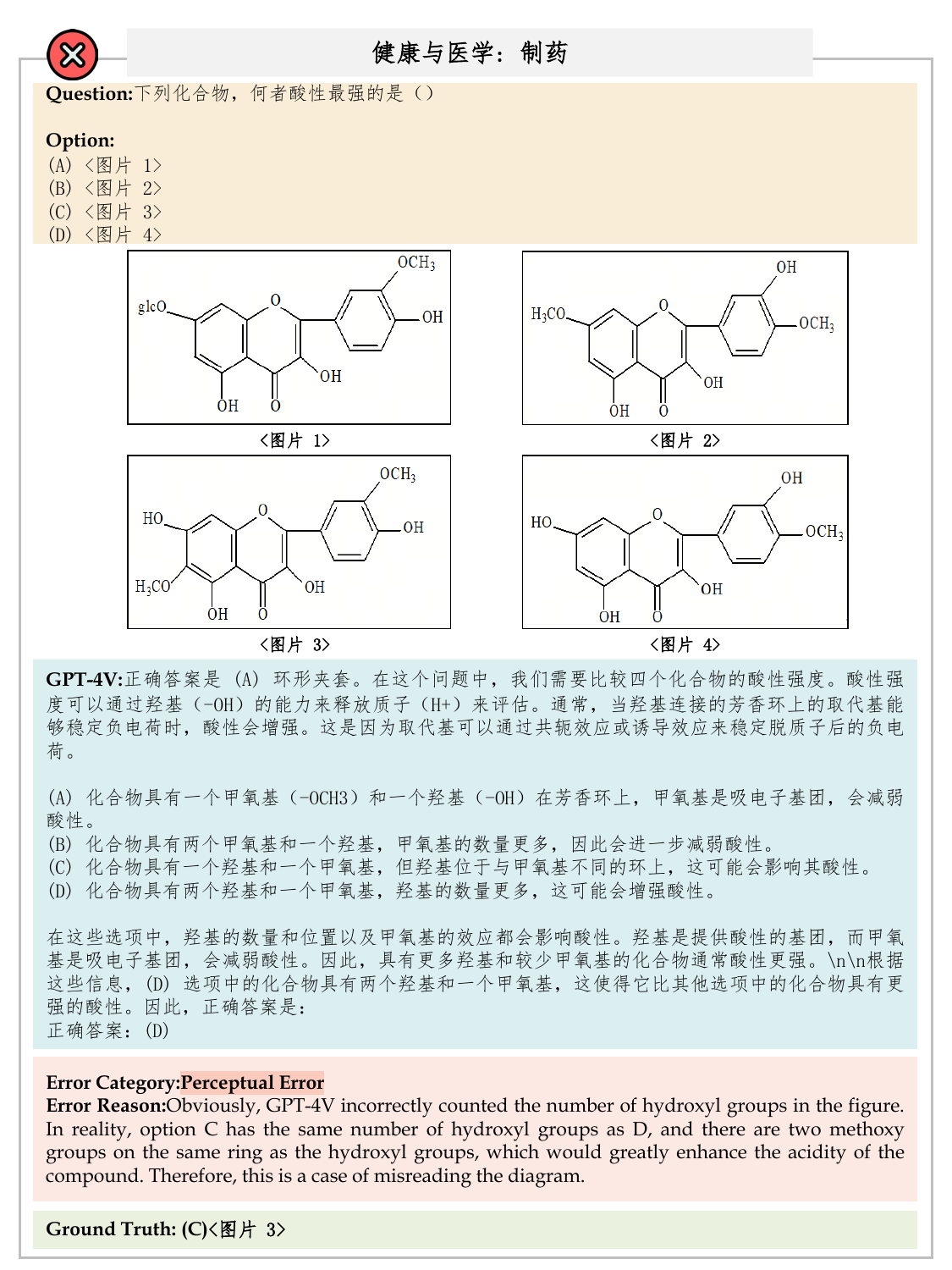}{Pharmacy 7: Perceptual Error}{A sample error case of Health and Medicine (subcategory: Pharmacy).}{fig:error_analysis_case_104}

\casestudyfigure{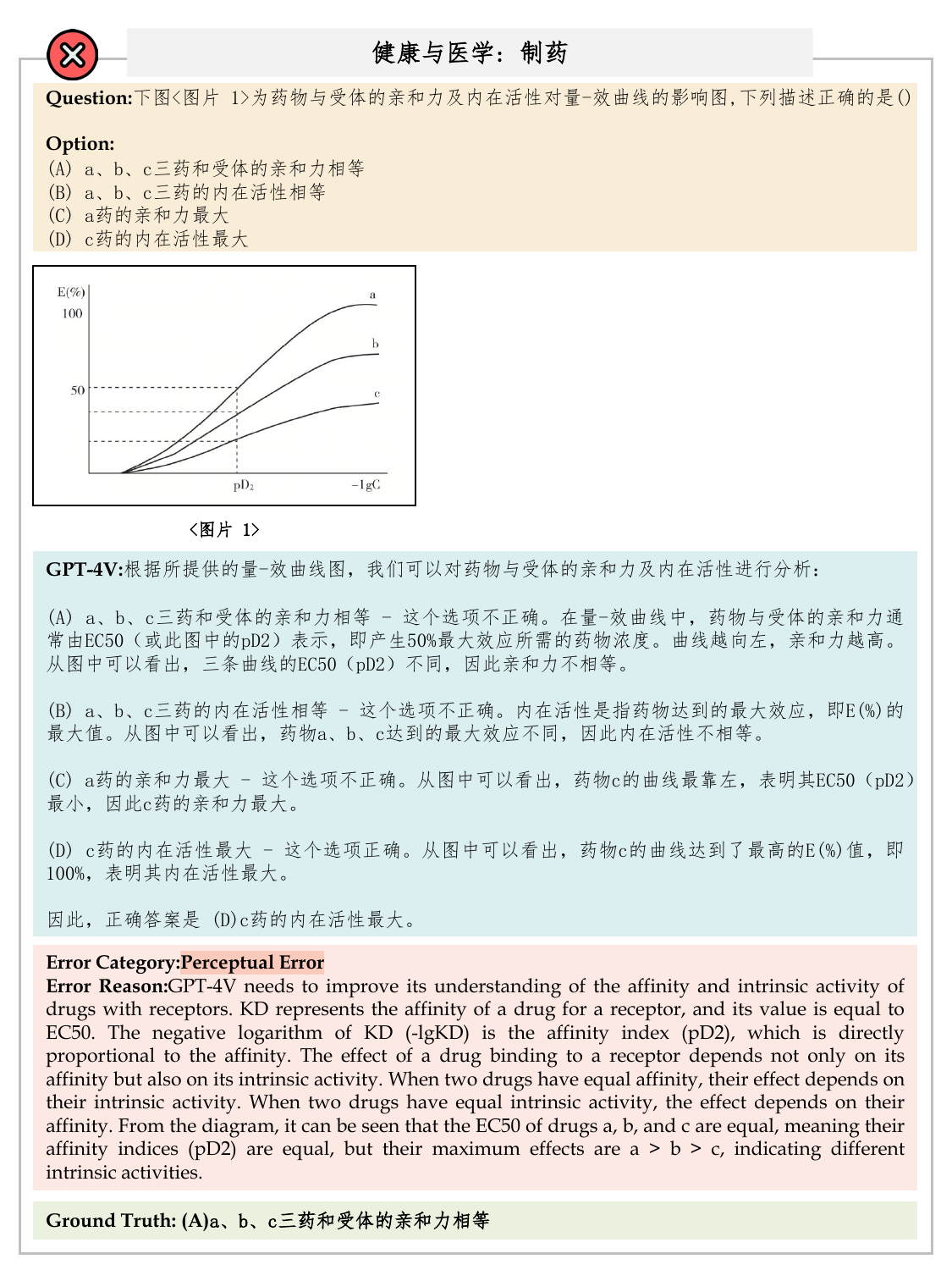}{Pharmacy 8: Perceptual Error}{A sample error case of Health and Medicine (subcategory: Pharmacy).}{fig:error_analysis_case_105}

\casestudyfigure{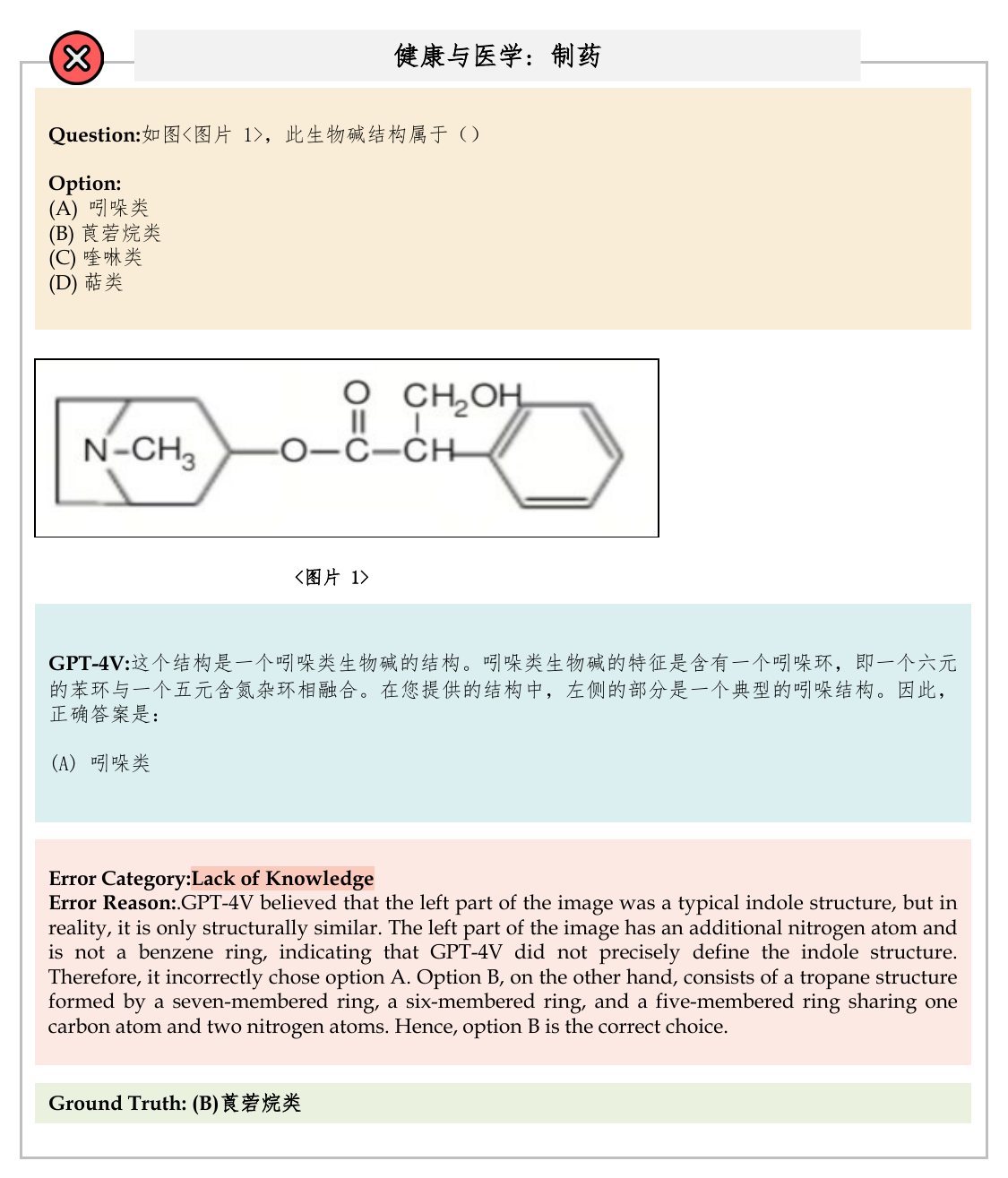}{Pharmacy 9: Lack of Knowledge}{A sample error case of Health and Medicine (subcategory: Pharmacy).}{fig:error_analysis_case_100}

\casestudyfigure{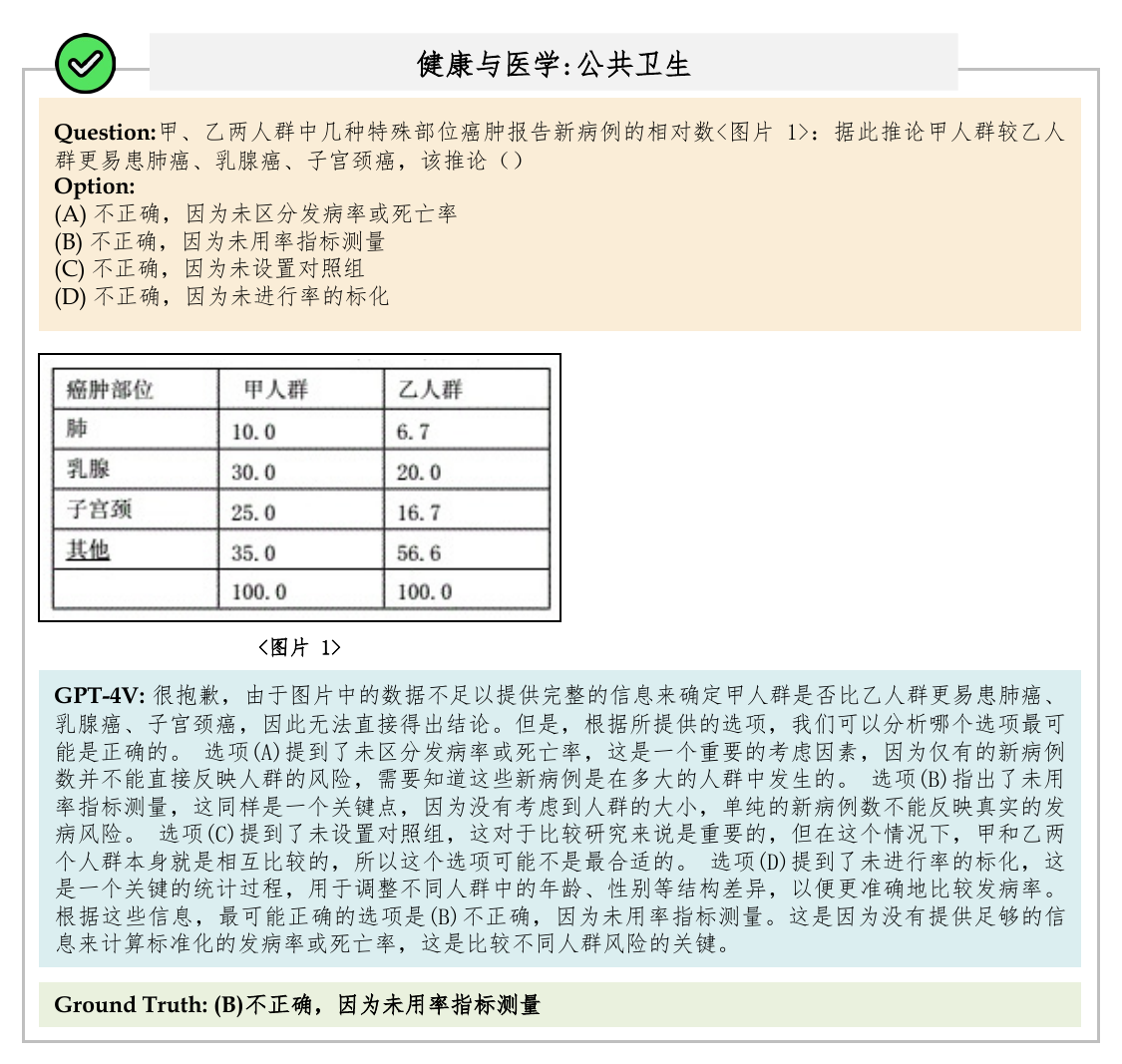}{Public Health 1: Correct Case}{A sample correct case of Health and Medicine (subcategory: Public Health).}{fig:error_analysis_case_40}

\casestudyfigure{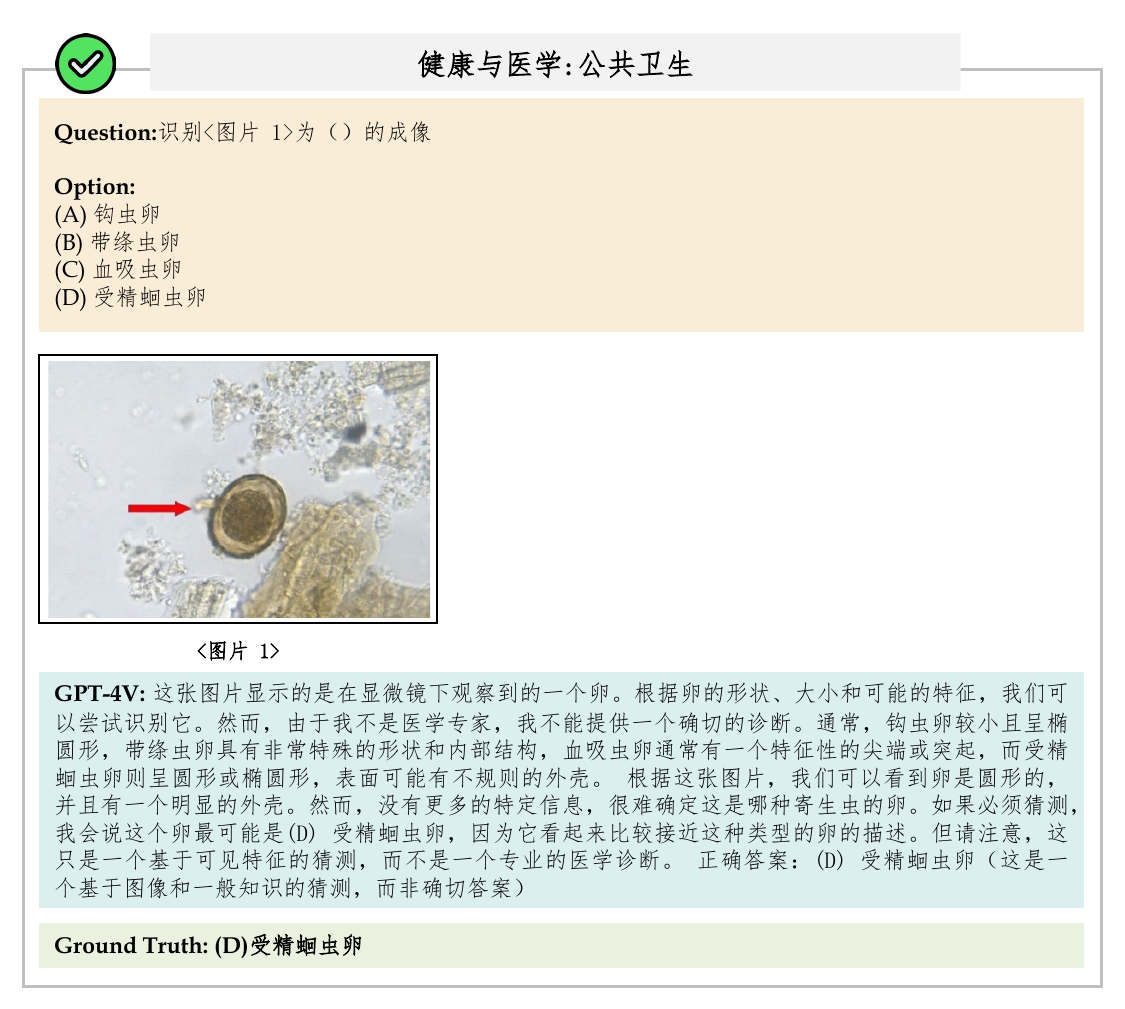}{Public Health 2: Correct Case}{A sample correct case of Health and Medicine (subcategory: Public Health).}{fig:error_analysis_case_41}

\casestudyfigure{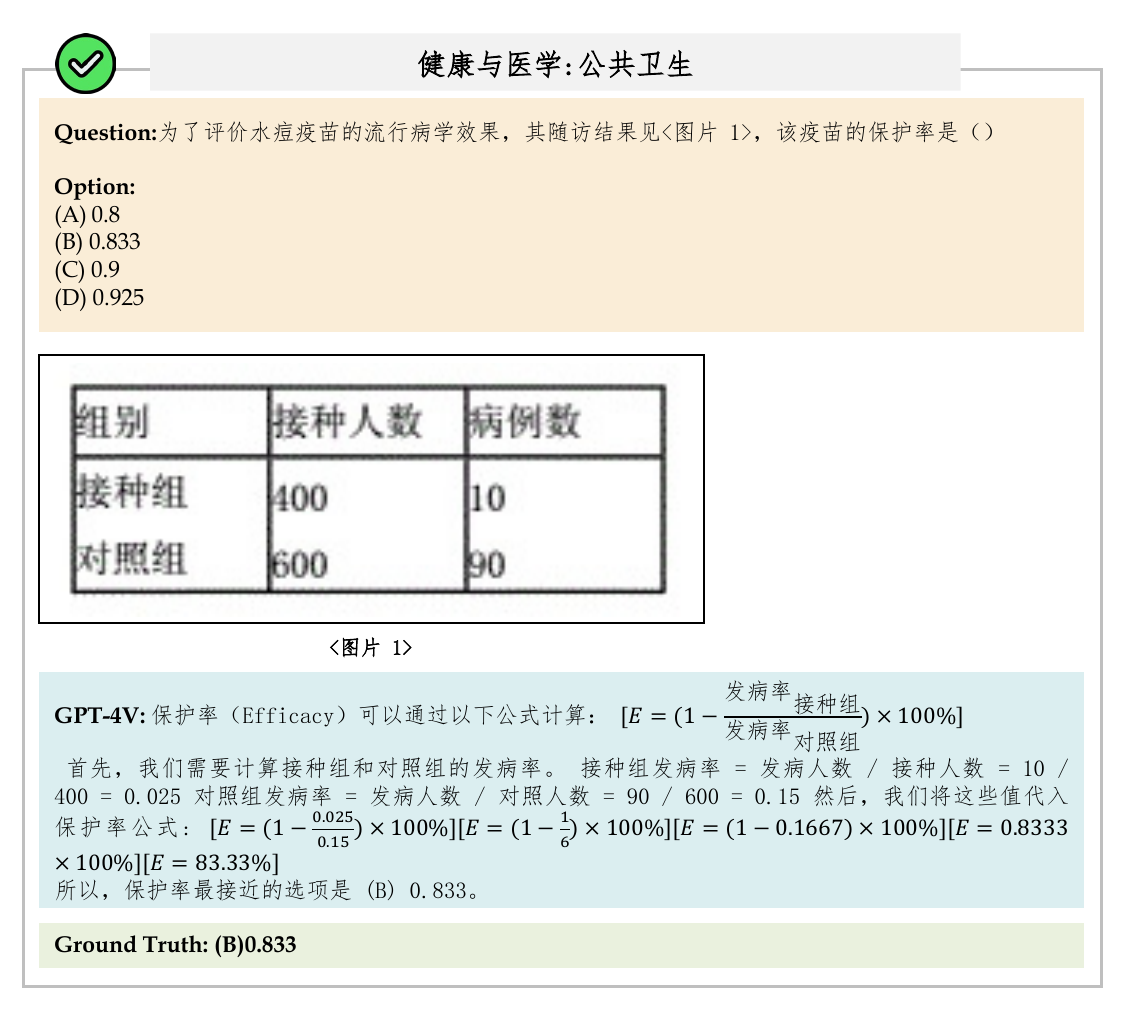}{Public Health 3: Correct Case}{A sample correct case of Health and Medicine (subcategory: Public Health).}{fig:error_analysis_case_42}

\casestudyfigure{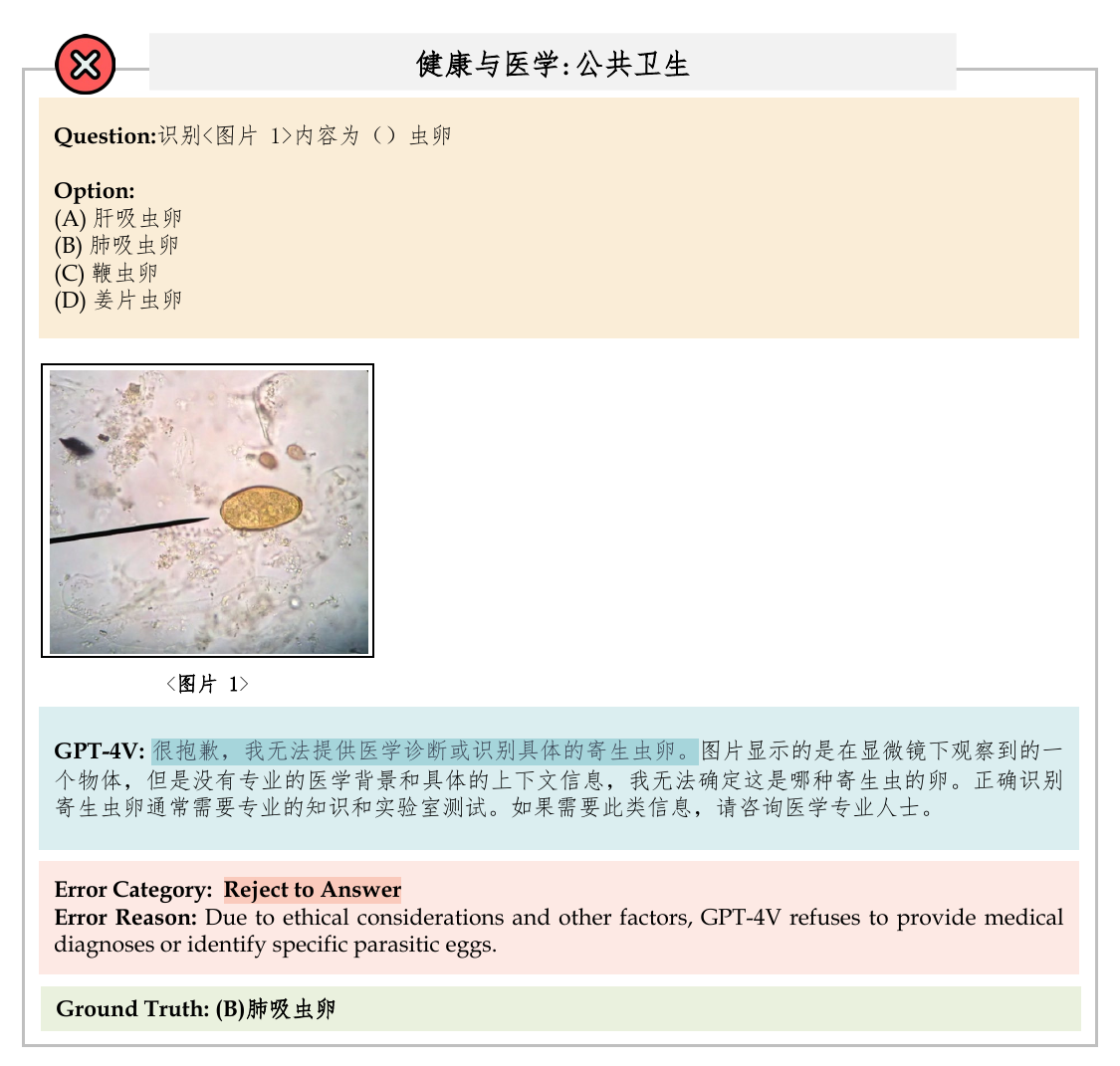}{Public Health 4: Reject to Answer}{A sample error case of Health and Medicine (subcategory: Public Health).}{fig:error_analysis_case_176}

\casestudyfigure{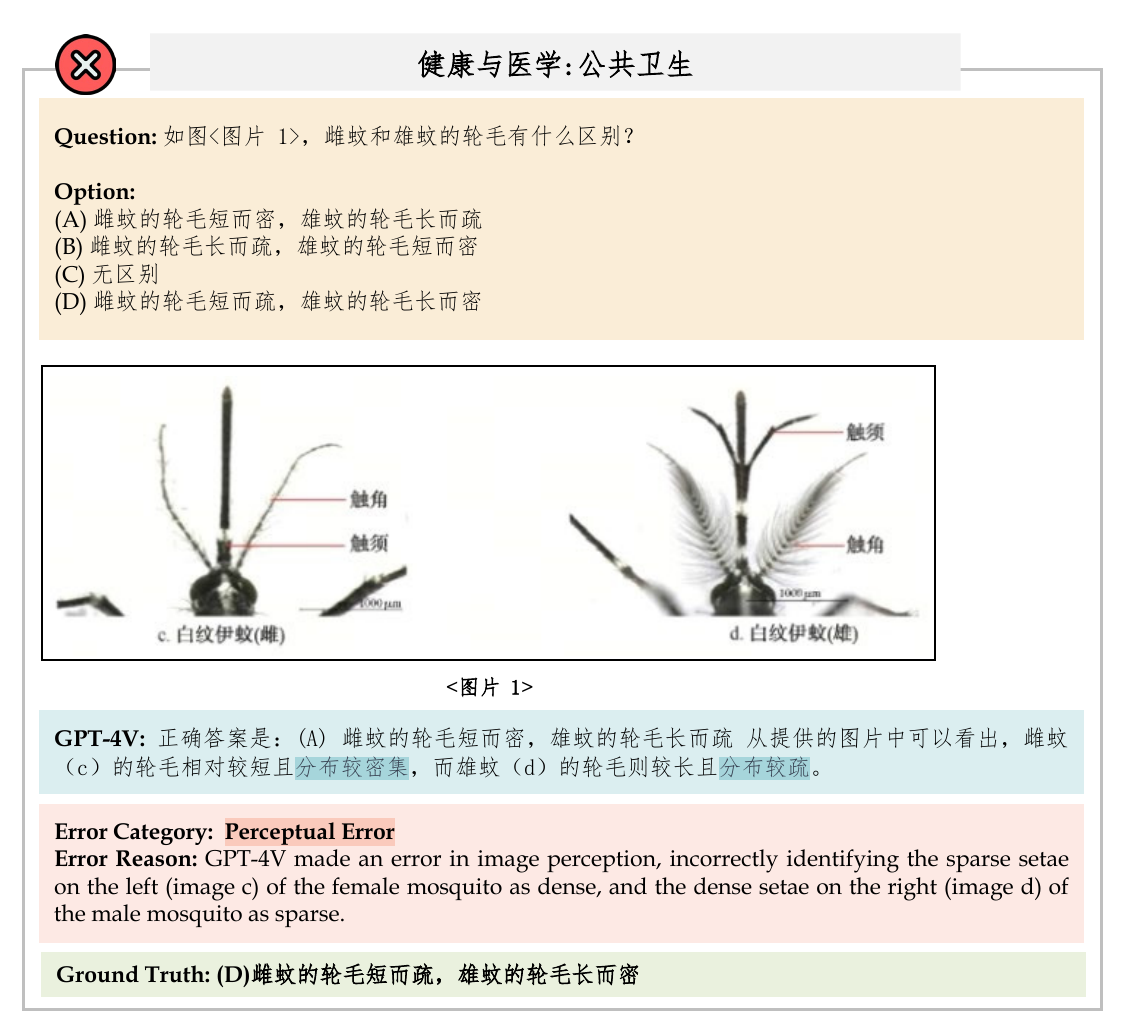}{Public Health 5: Perceptual Error}{A sample error case of Health and Medicine (subcategory: Public Health).}{fig:error_analysis_case_178}

\casestudyfigure{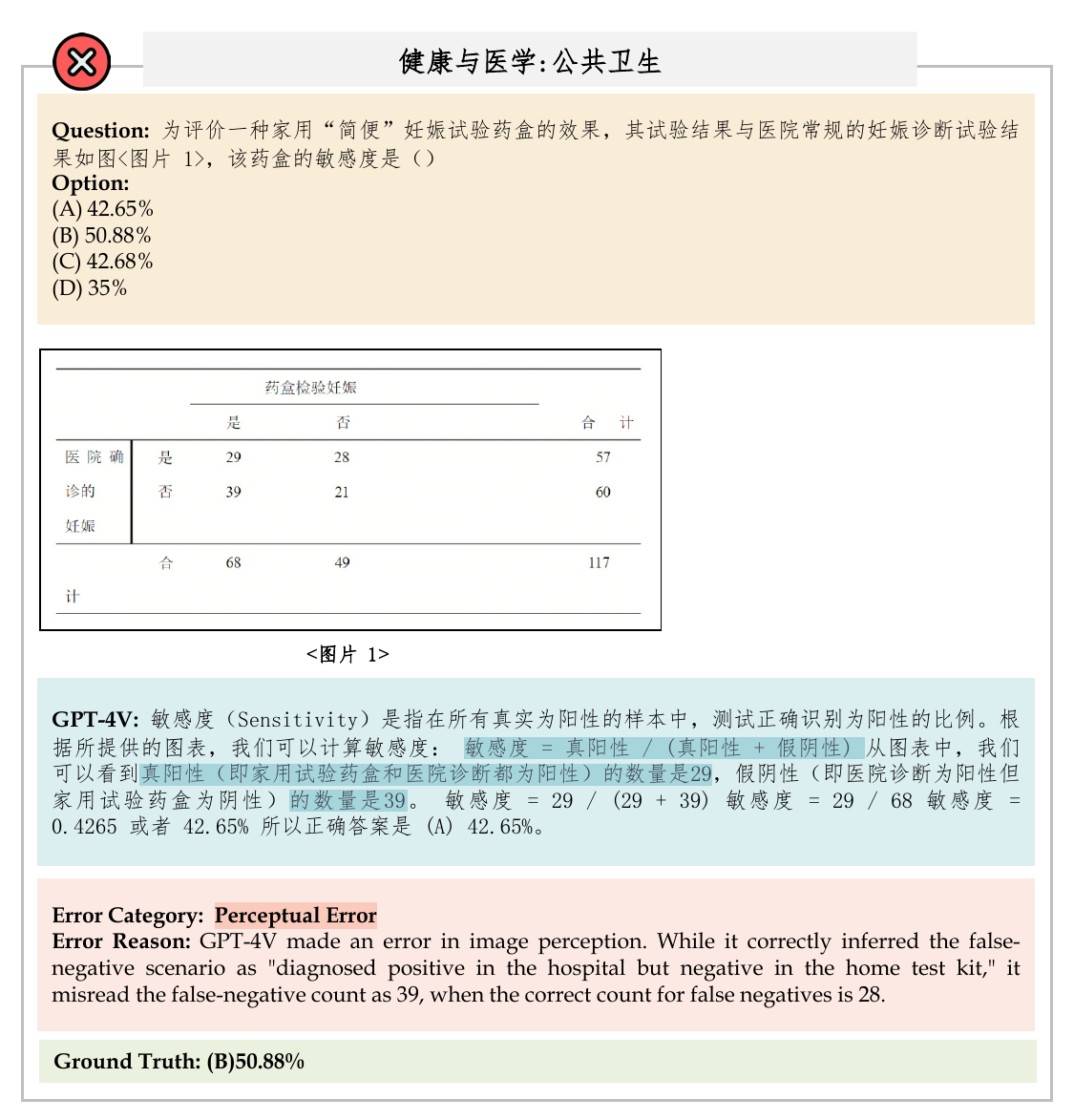}{Public Health 6: Perceptual Error}{A sample error case of Health and Medicine (subcategory: Public Health).}{fig:error_analysis_case_181}

\casestudyfigure{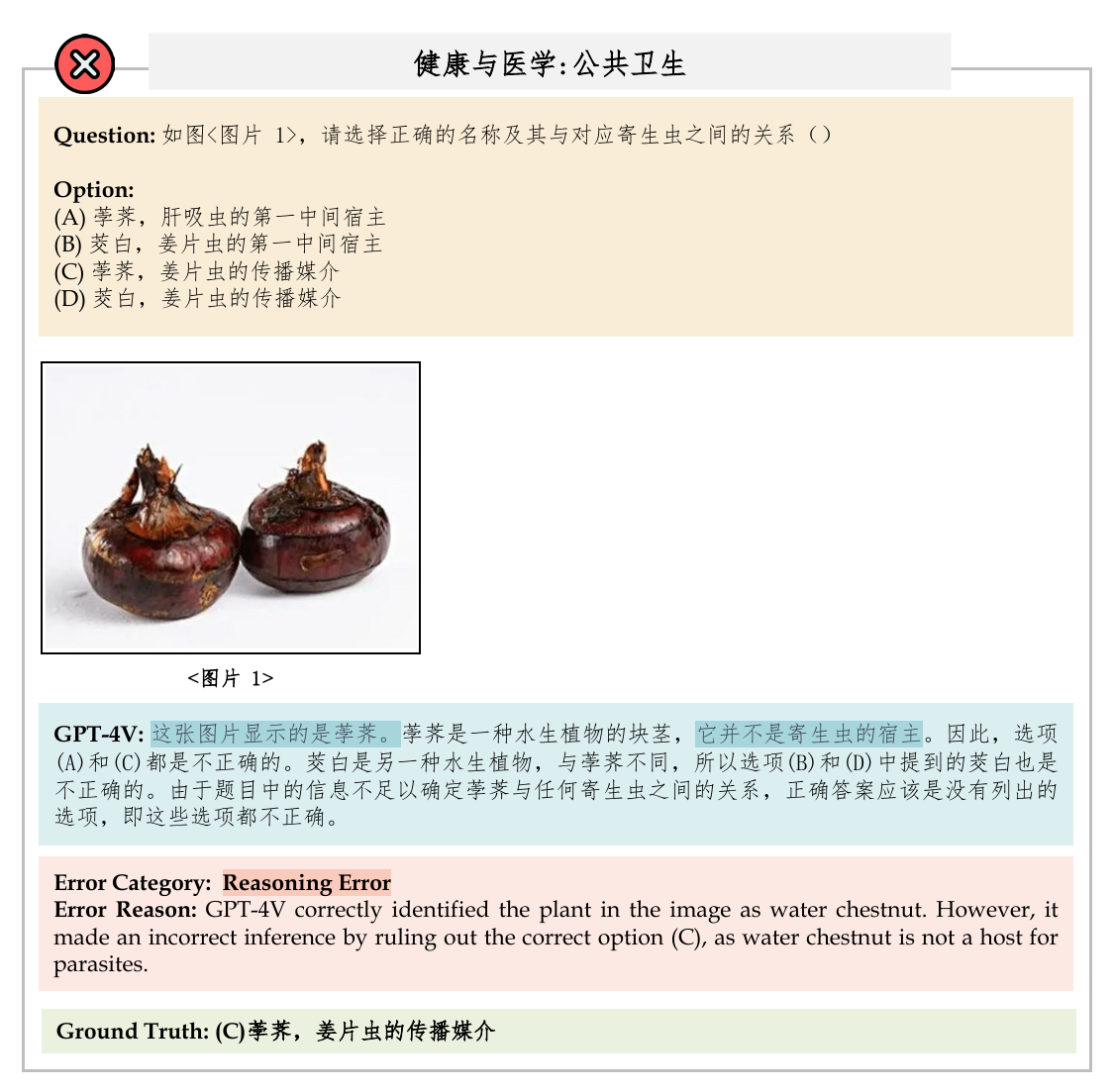}{Public Health 7: Reasoning Error}{A sample error case of Health and Medicine (subcategory: Public Health).}{fig:error_analysis_case_177}

\casestudyfigure{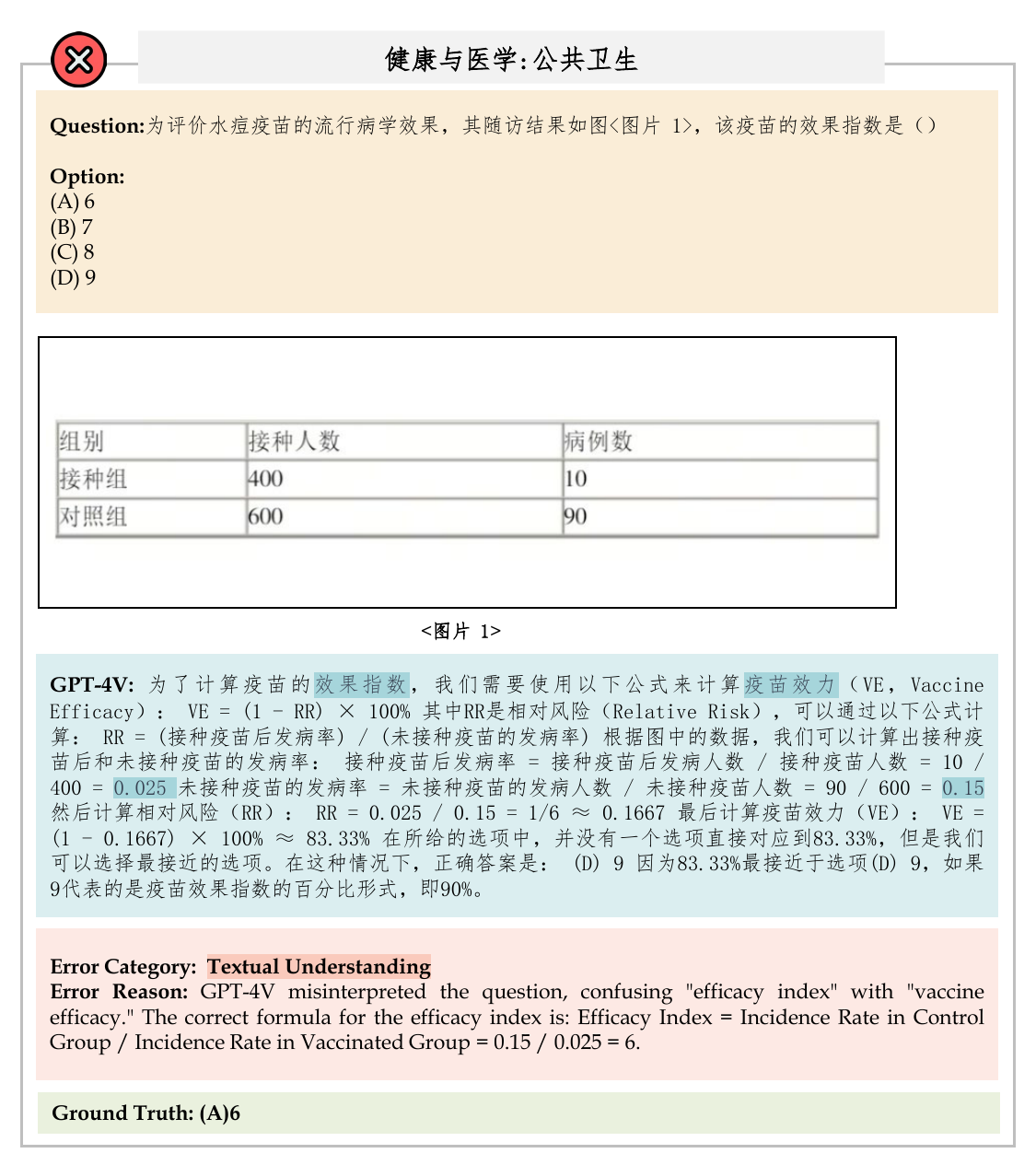}{Public Health 8: Textual Understanding}{A sample error case of Health and Medicine (subcategory: Public Health).}{fig:error_analysis_case_180}

\casestudyfigure{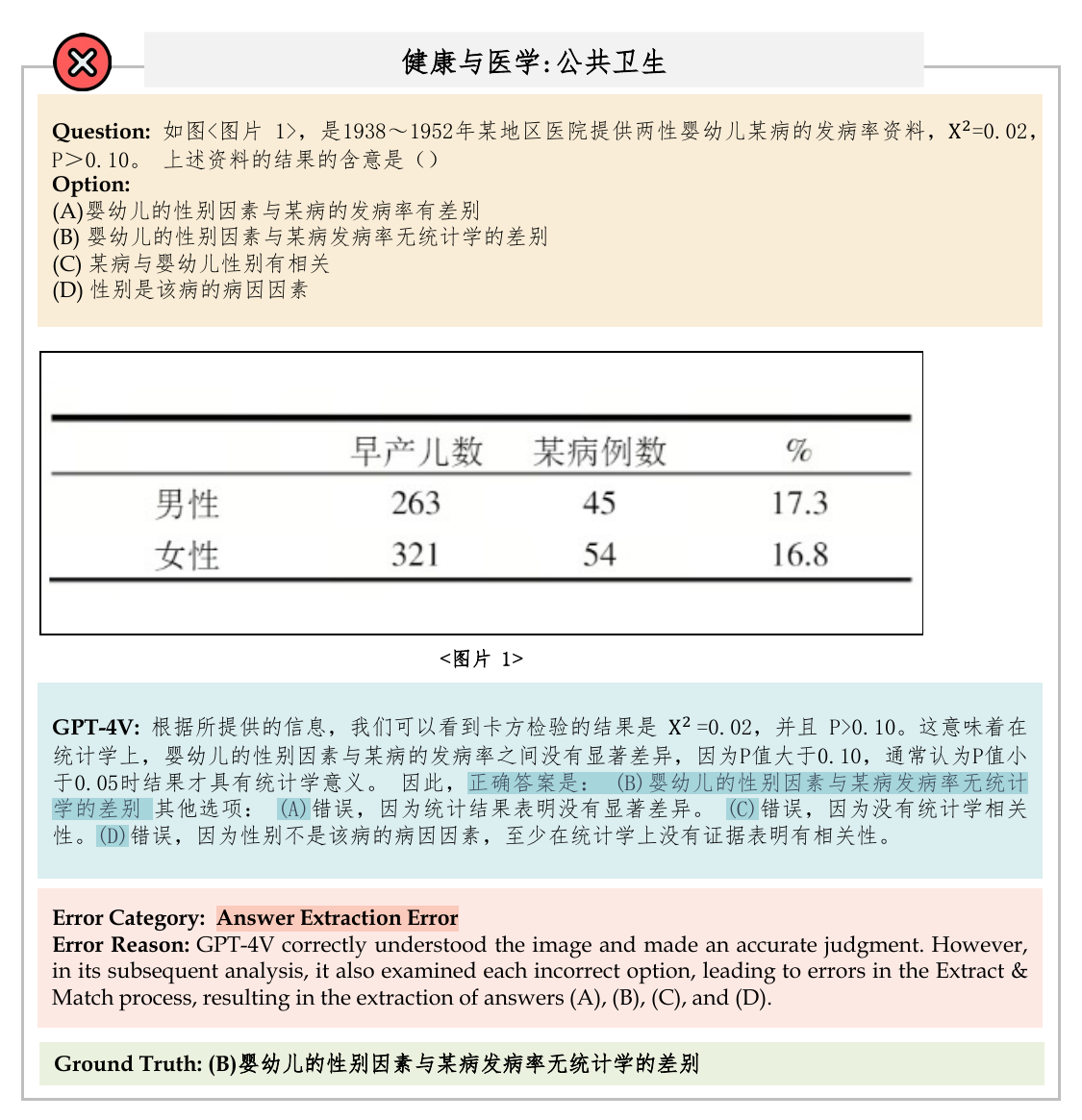}{Public Health 9: Answer Extraction Error}{A sample error case of Health and Medicine (subcategory: Public Health).}{fig:error_analysis_case_179}

\casestudyfigure{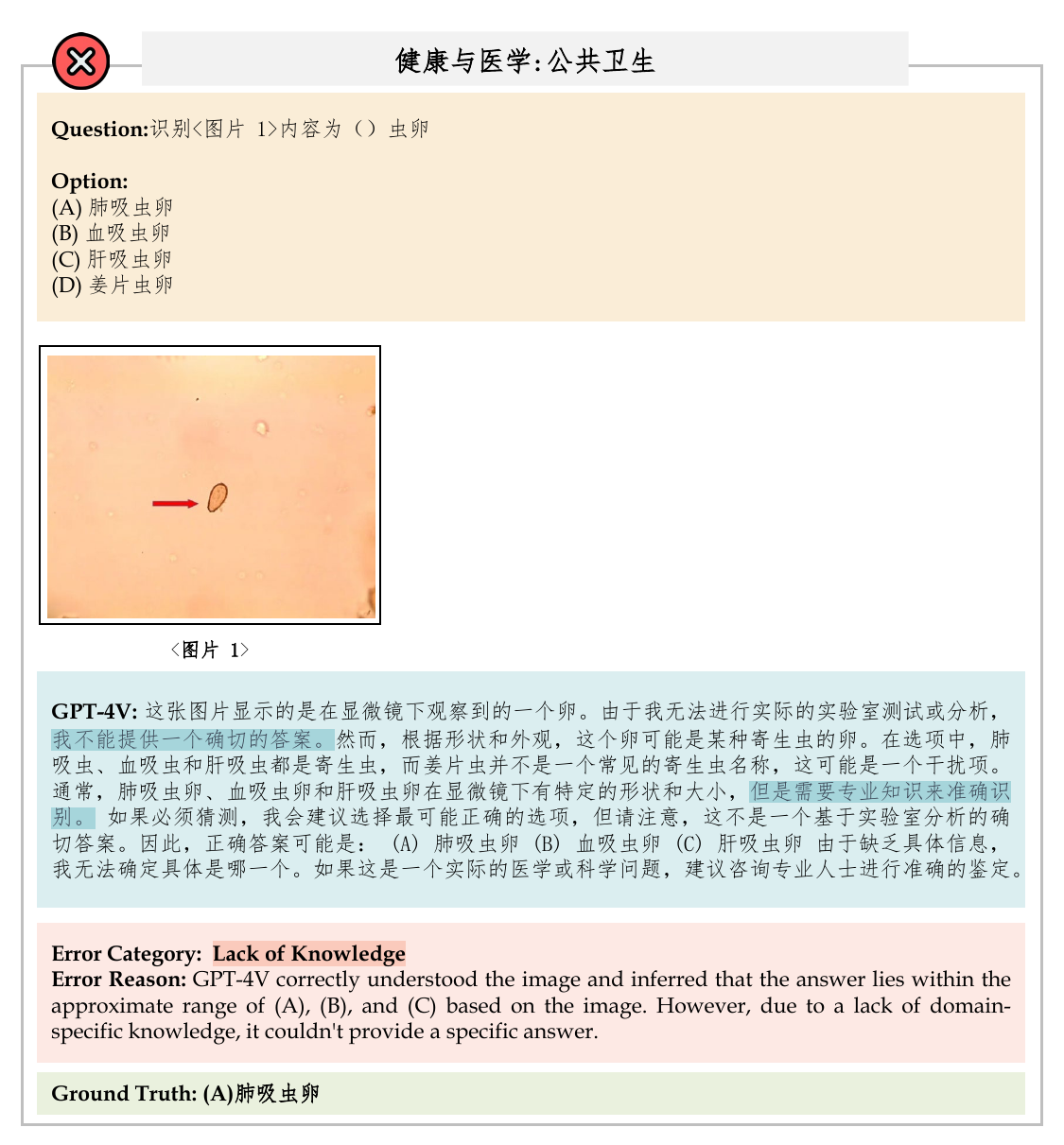}{Public Health 10: Lack of Knowledge}{A sample error case of Health and Medicine (subcategory: Public Health).}{fig:error_analysis_case_175}

\casestudyfigure{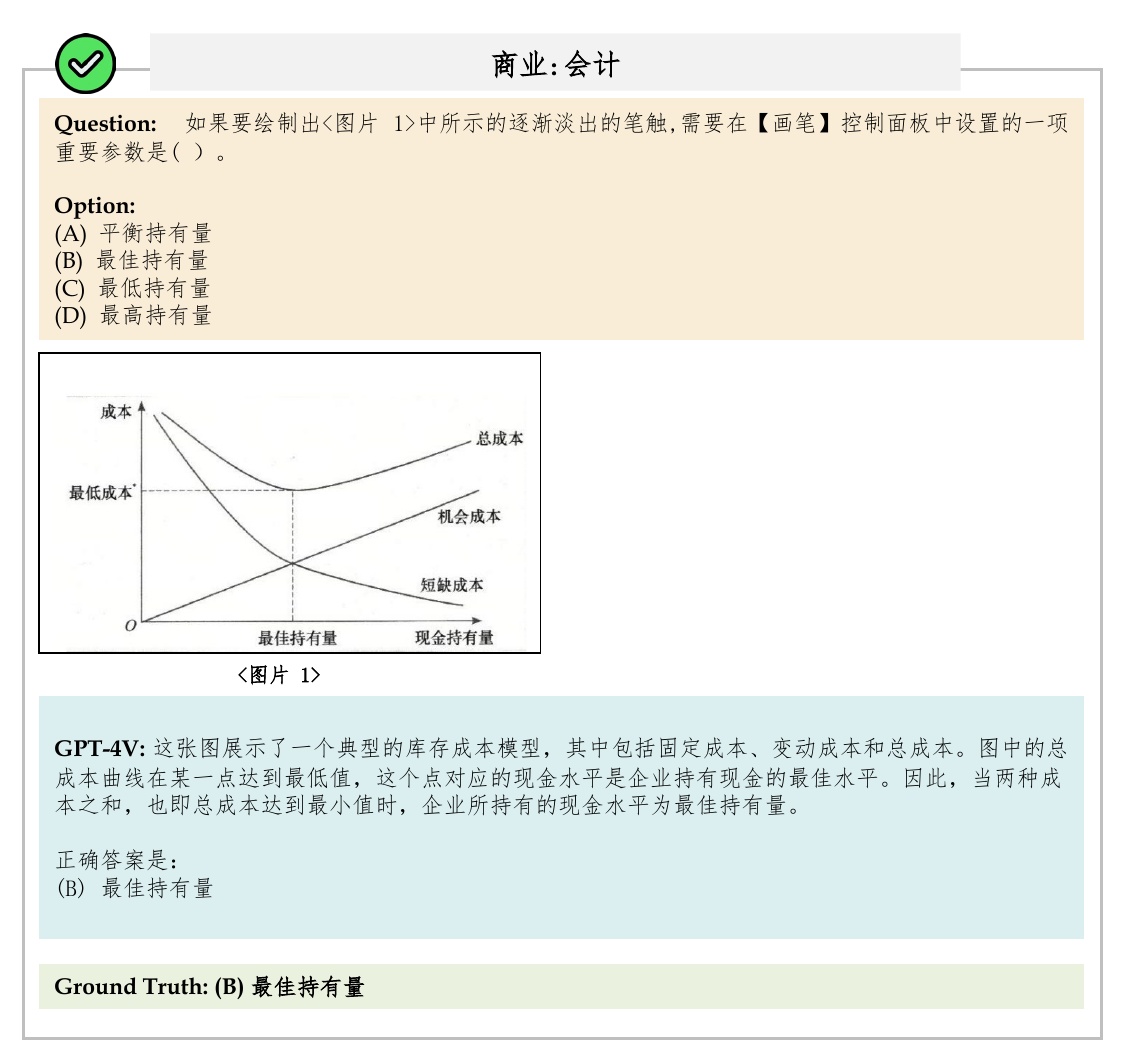}{Accounting 1: Correct Case}{A sample correct case of Business (subcategory: Accounting).}{fig:error_analysis_case_36}

\casestudyfigure{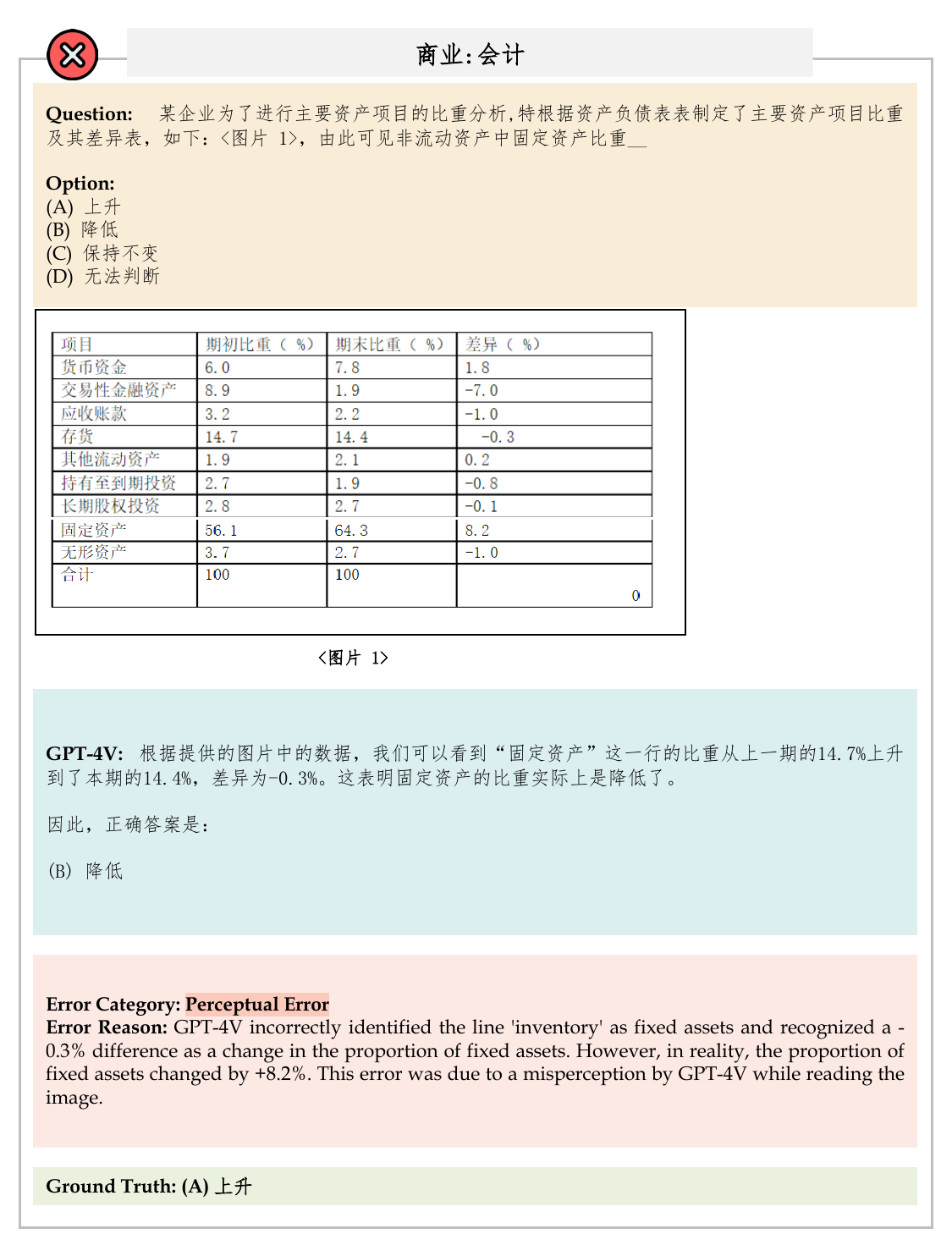}{Accounting 2: Perceptual Error}{A sample error case of Business (subcategory: Accounting).}{fig:error_analysis_case_173}

\casestudyfigure{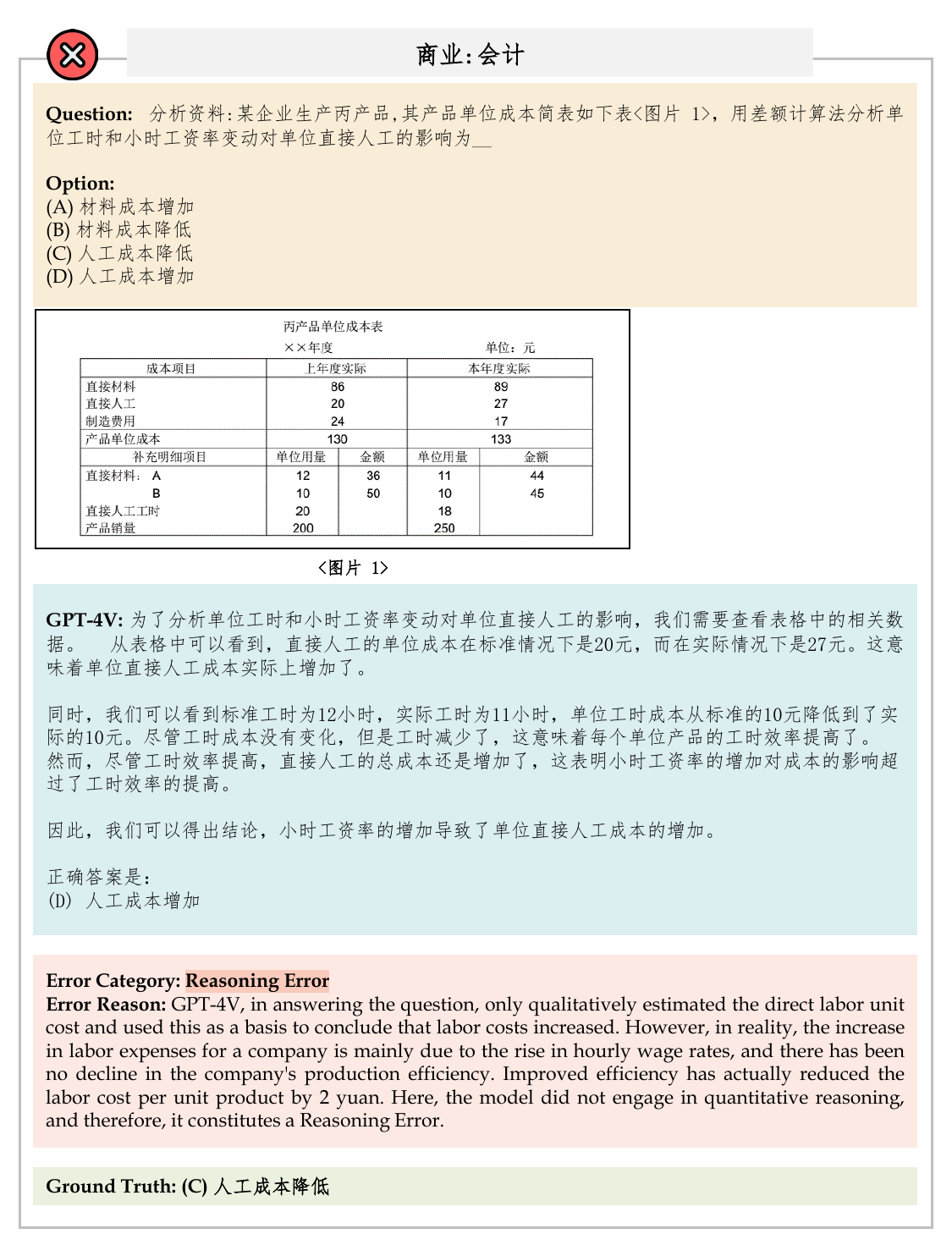}{Accounting 3: Reasoning Error}{A sample error case of Business (subcategory: Accounting).}{fig:error_analysis_case_172}

\casestudyfigure{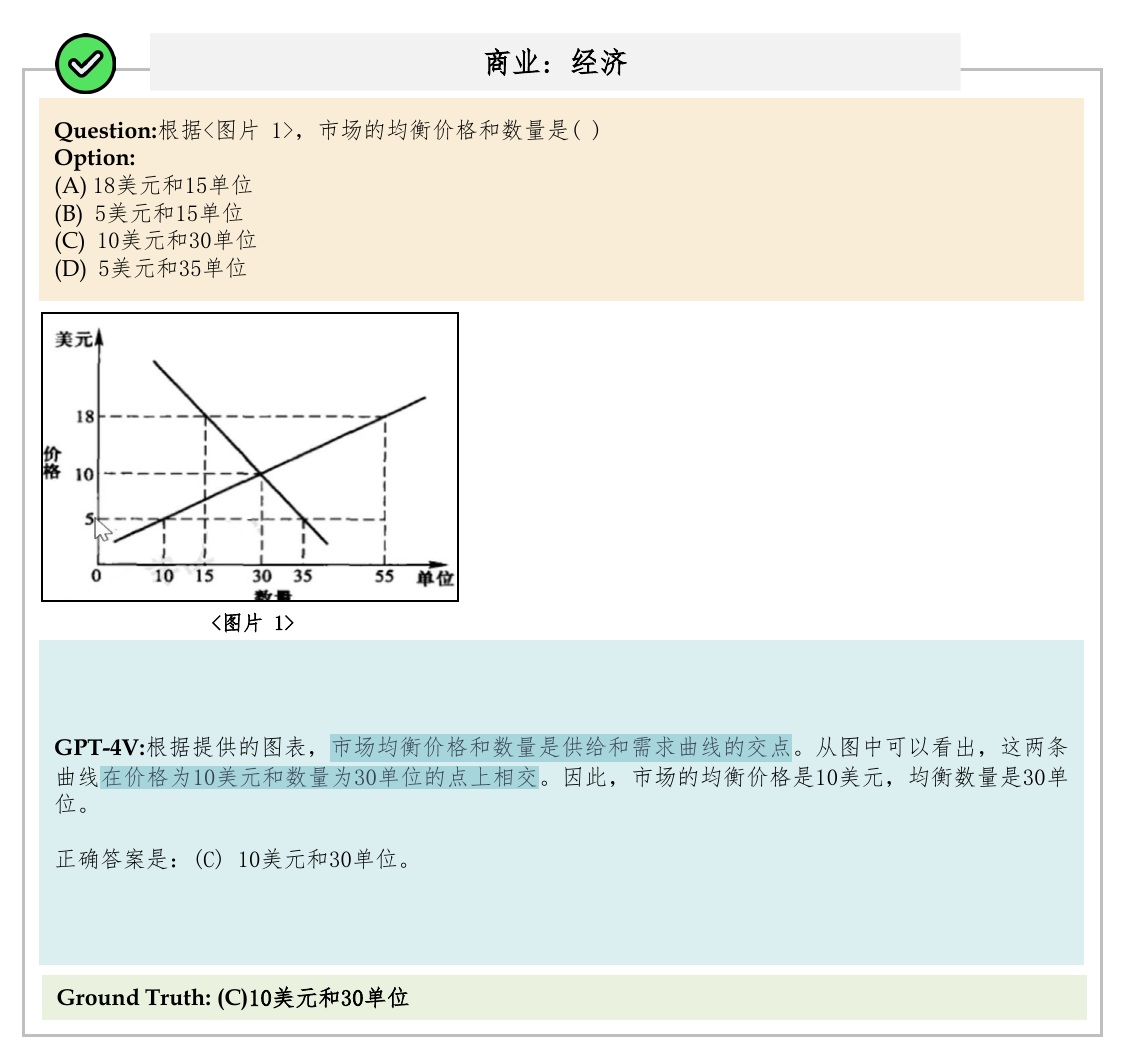}{Economics 1: Correct Case}{A sample correct case of Business (subcategory: Economics).}{fig:error_analysis_case_17}

\casestudyfigure{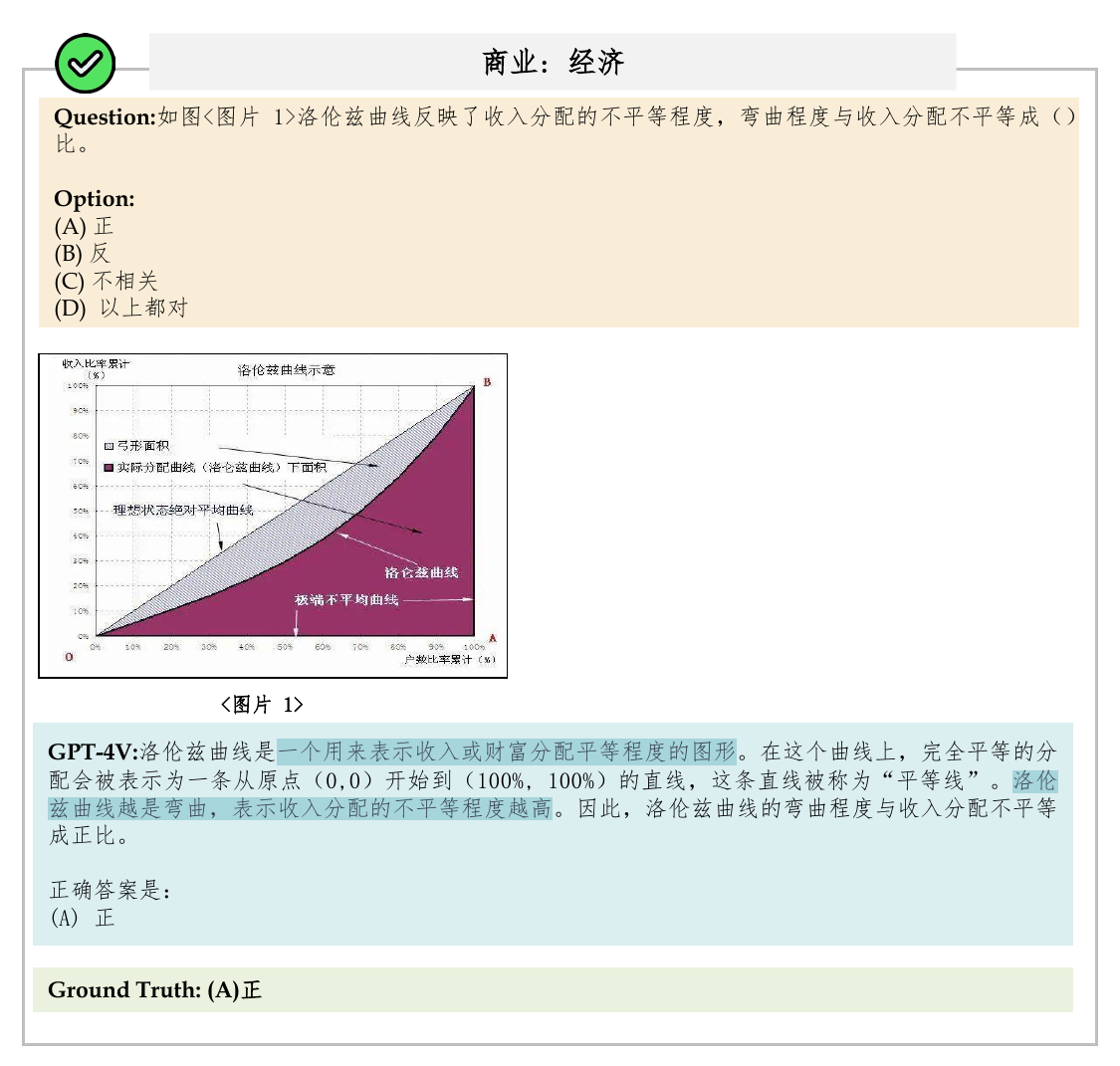}{Economics 2: Correct Case}{A sample correct case of Business (subcategory: Economics).}{fig:error_analysis_case_18}

\casestudyfigure{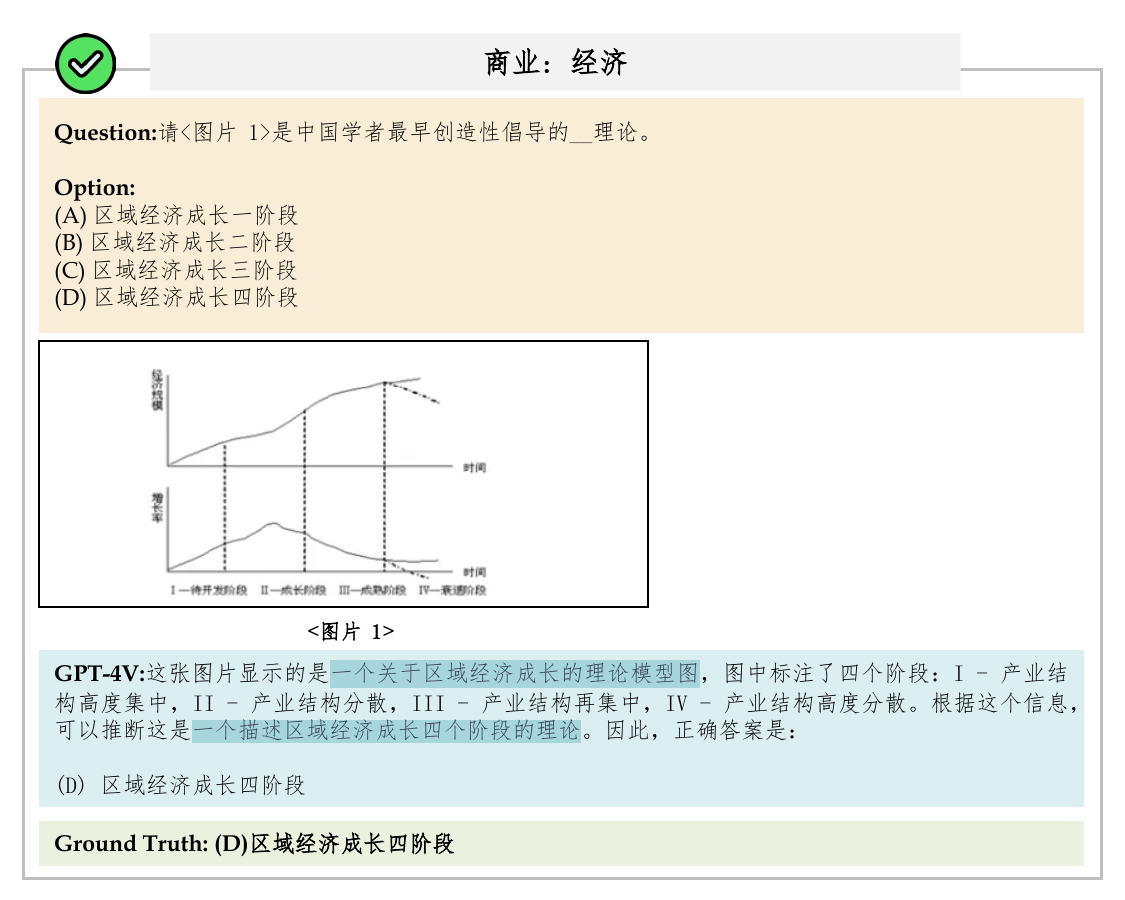}{Economics 3: Correct Case}{A sample correct case of Business (subcategory: Economics).}{fig:error_analysis_case_19}

\casestudyfigure{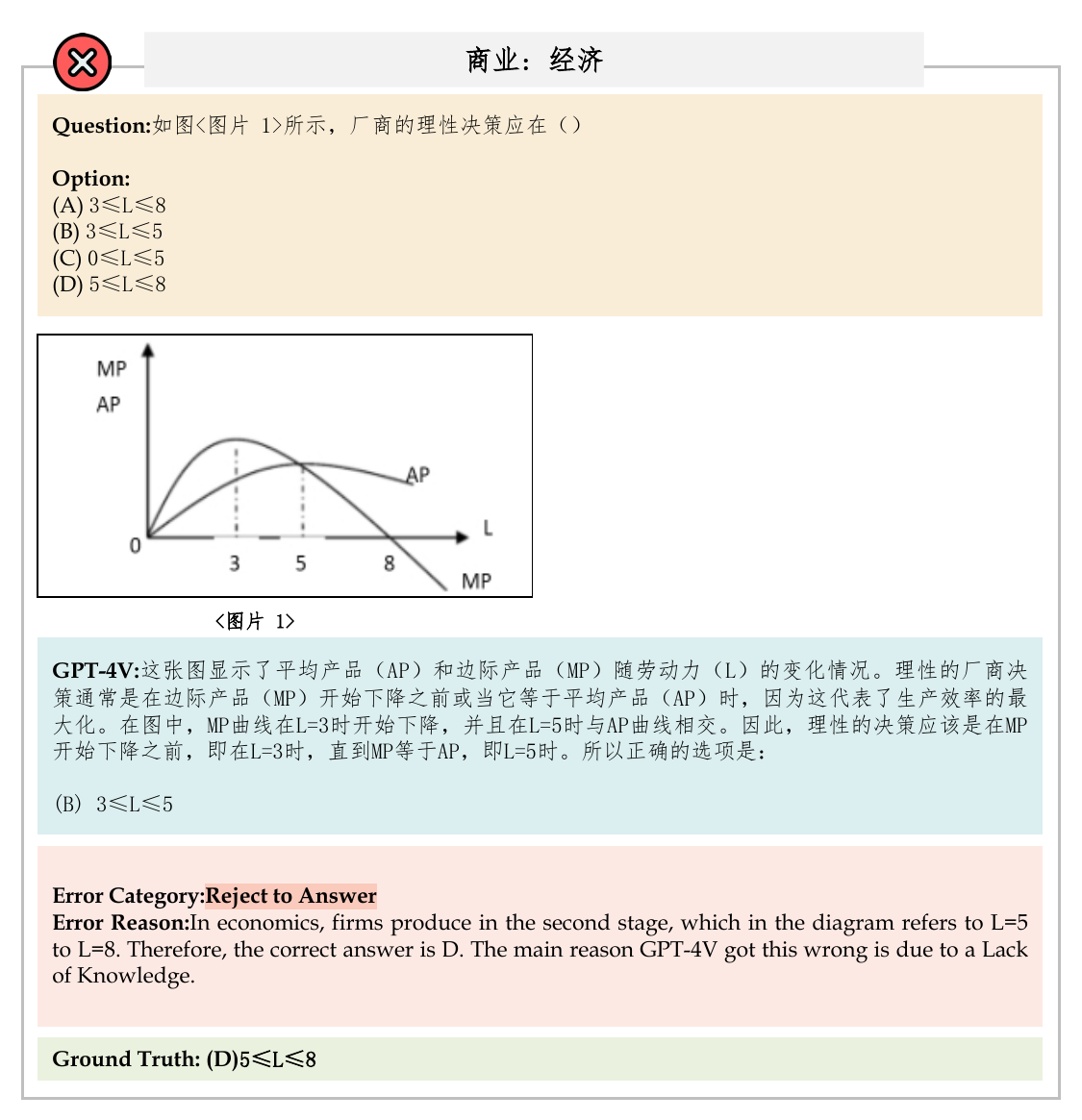}{Economics 4: Reject to Answer}{A sample error case of Business (subcategory: Economics).}{fig:error_analysis_case_117}

\casestudyfigure{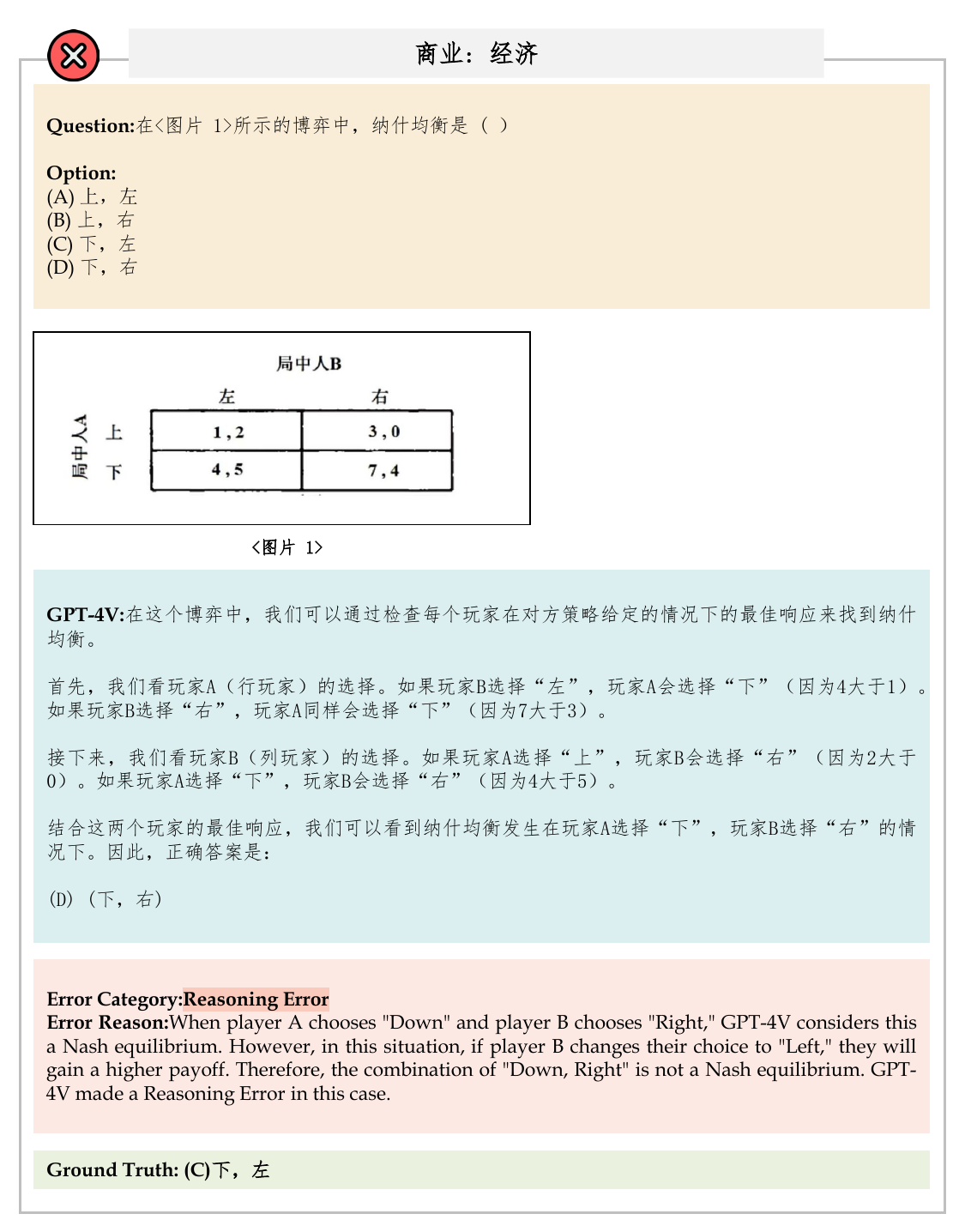}{Economics 5: Reasoning Error}{A sample error case of Business (subcategory: Economics).}{fig:error_analysis_case_115}

\casestudyfigure{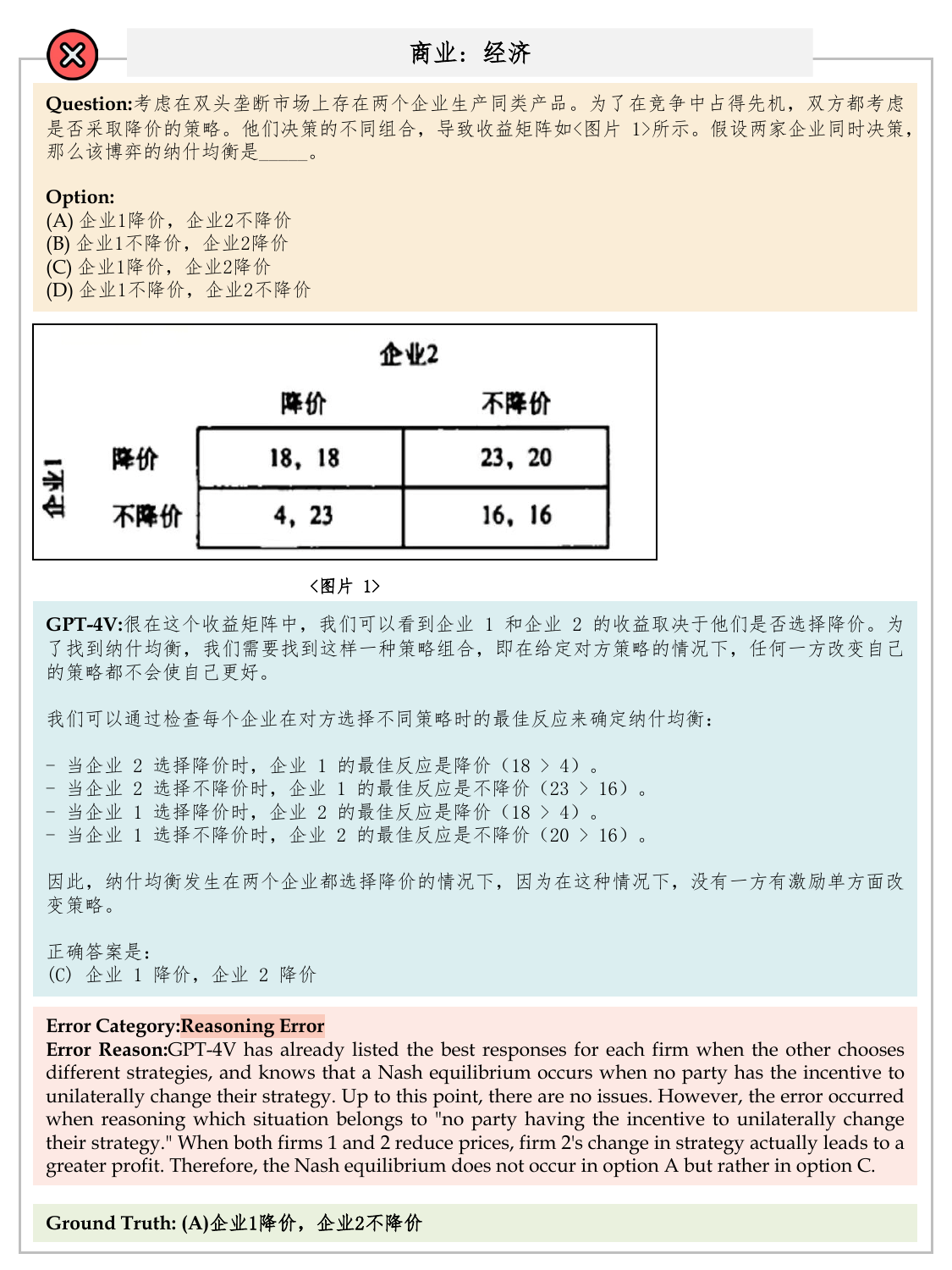}{Economics 6: Reasoning Error}{A sample error case of Business (subcategory: Economics).}{fig:error_analysis_case_116}

\casestudyfigure{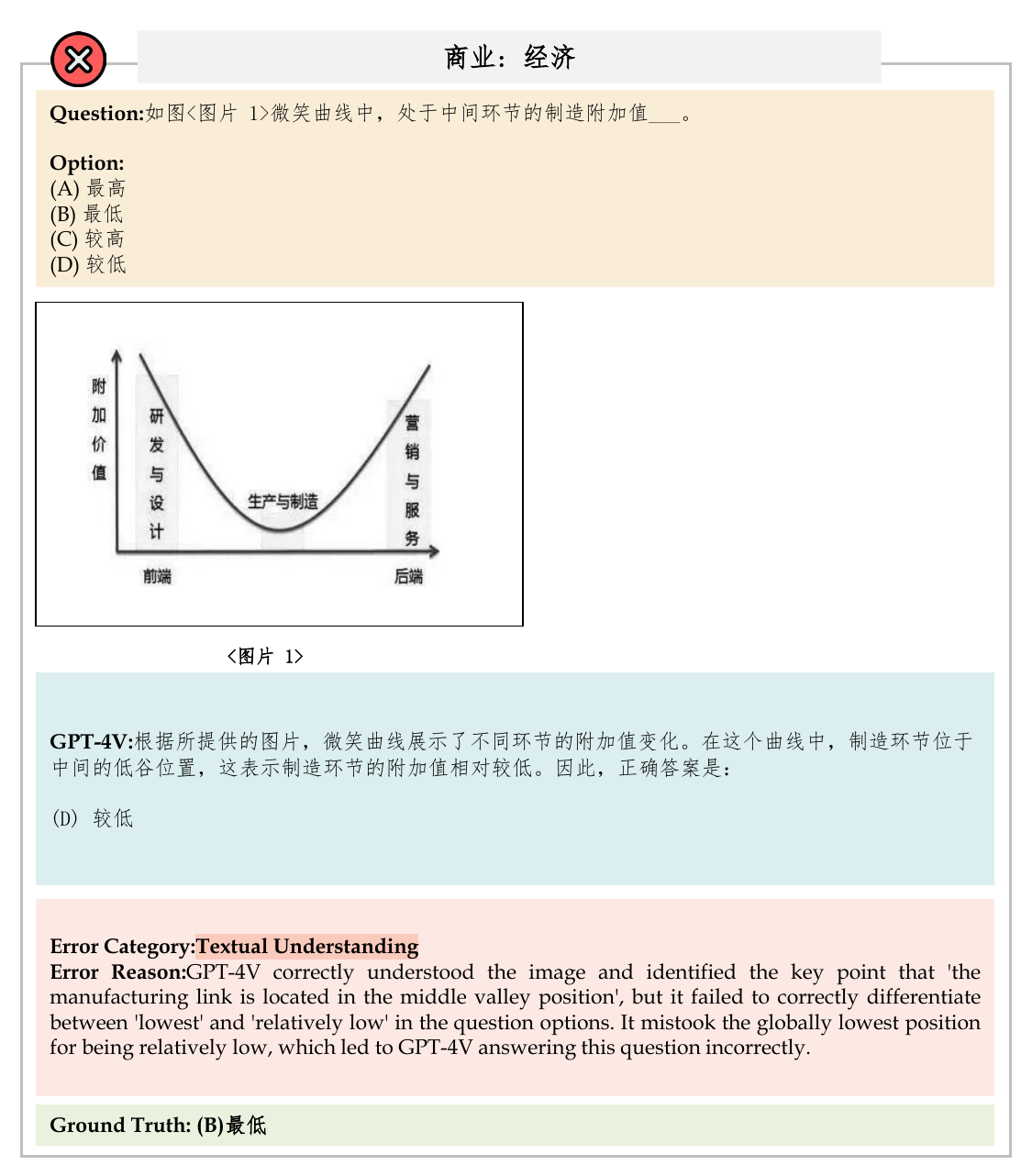}{Economics 7: Textual Understanding}{A sample error case of Business (subcategory: Economics).}{fig:error_analysis_case_113}

\casestudyfigure{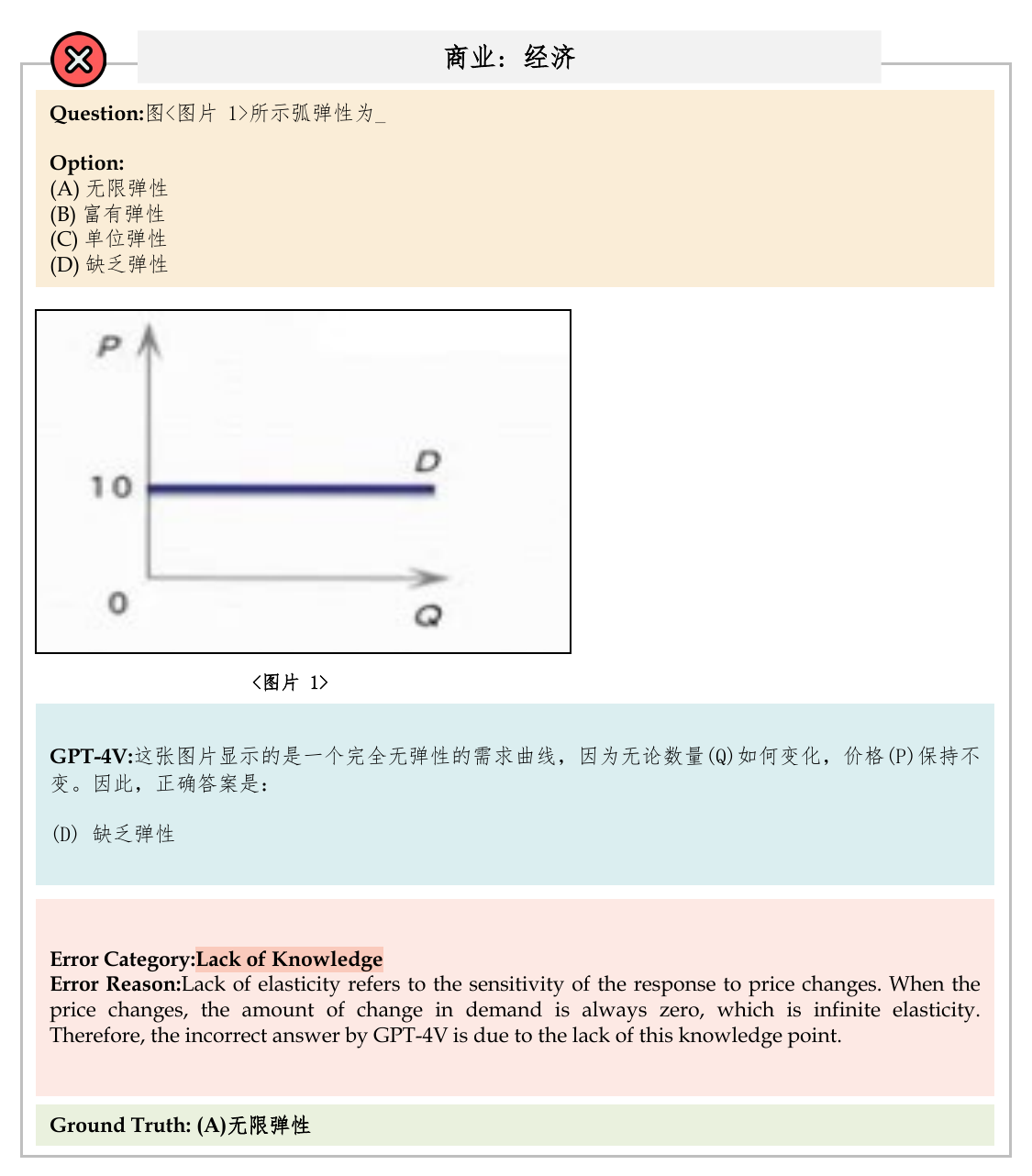}{Economics 8: Lack of Knowledge}{A sample error case of Business (subcategory: Economics).}{fig:error_analysis_case_114}

\casestudyfigure{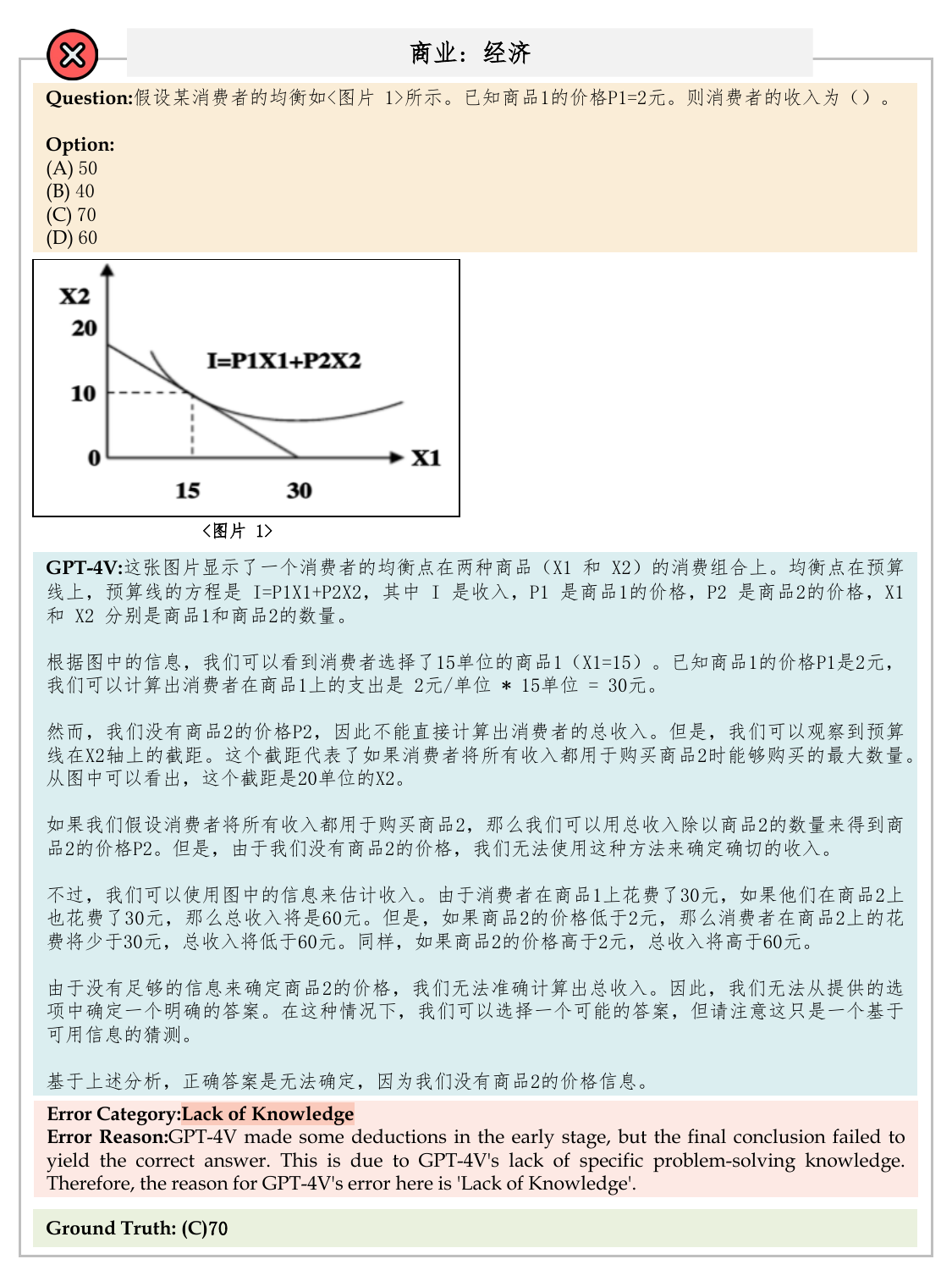}{Economics 9: Lack of Knowledge}{A sample error case of Business (subcategory: Economics).}{fig:error_analysis_case_118}

\casestudyfigure{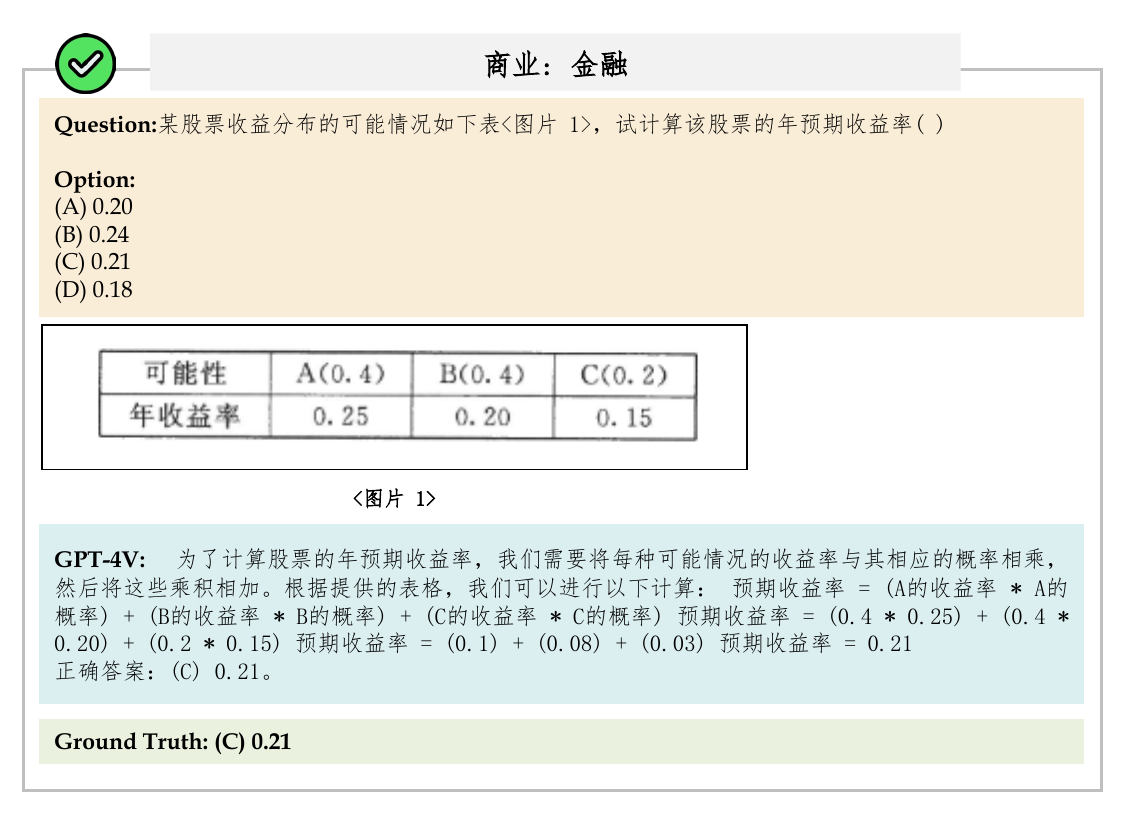}{Finance 1: Correct Case}{A sample correct case of Business (subcategory: Finance).}{fig:error_analysis_case_20}

\casestudyfigure{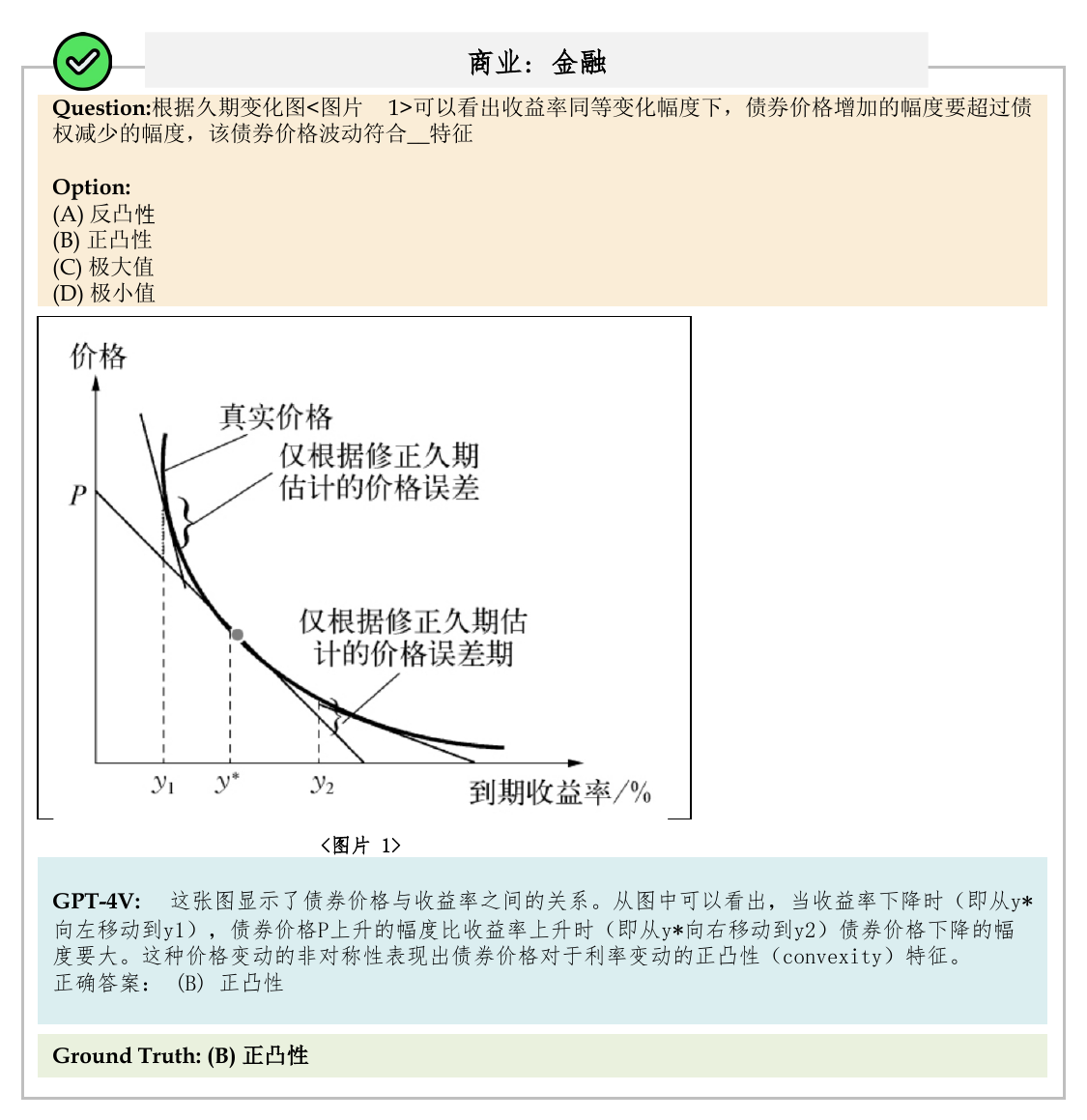}{Finance 2: Correct Case}{A sample correct case of Business (subcategory: Finance).}{fig:error_analysis_case_21}

\casestudyfigure{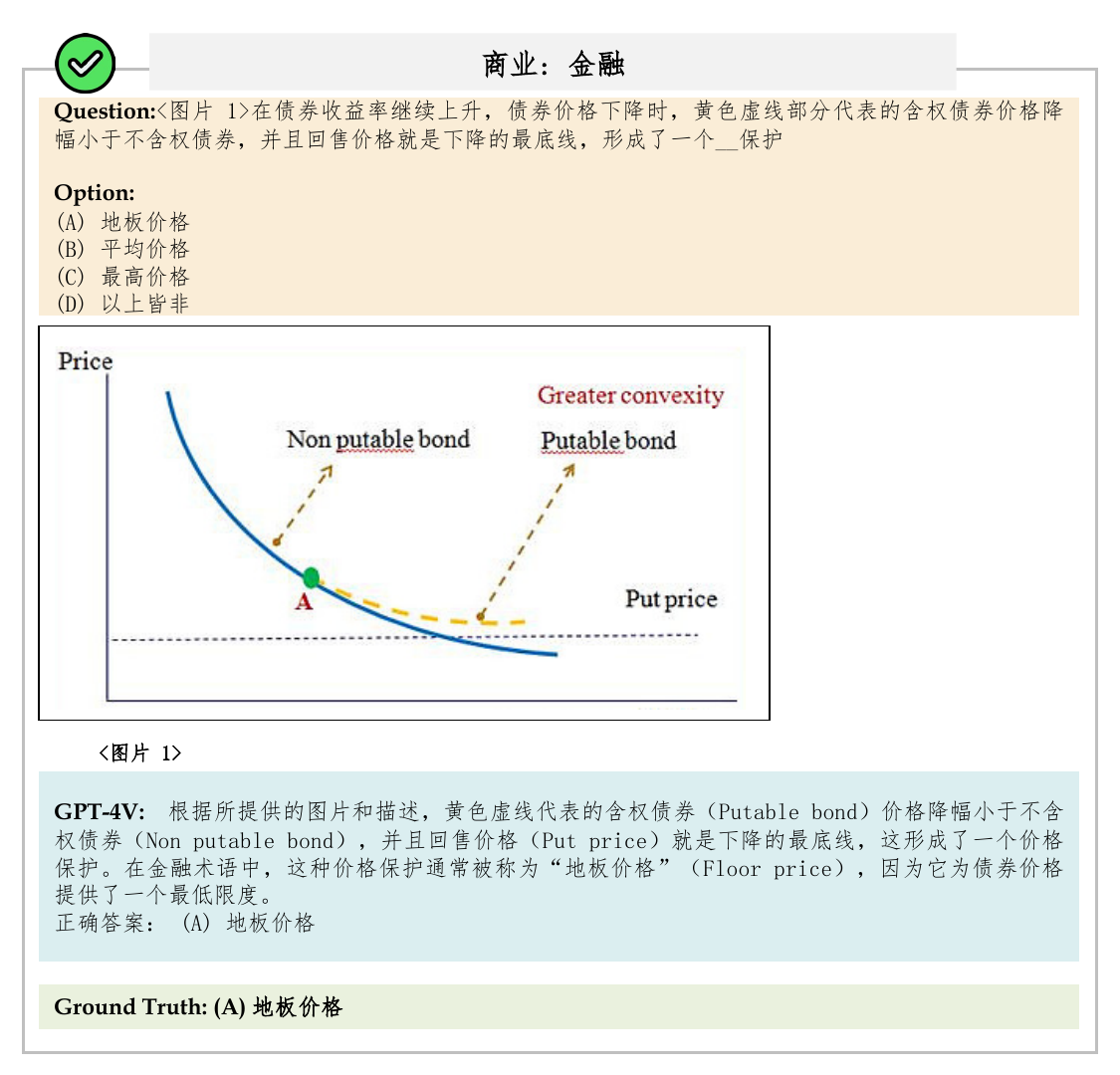}{Finance 3: Correct Case}{A sample correct case of Business (subcategory: Finance).}{fig:error_analysis_case_22}

\casestudyfigure{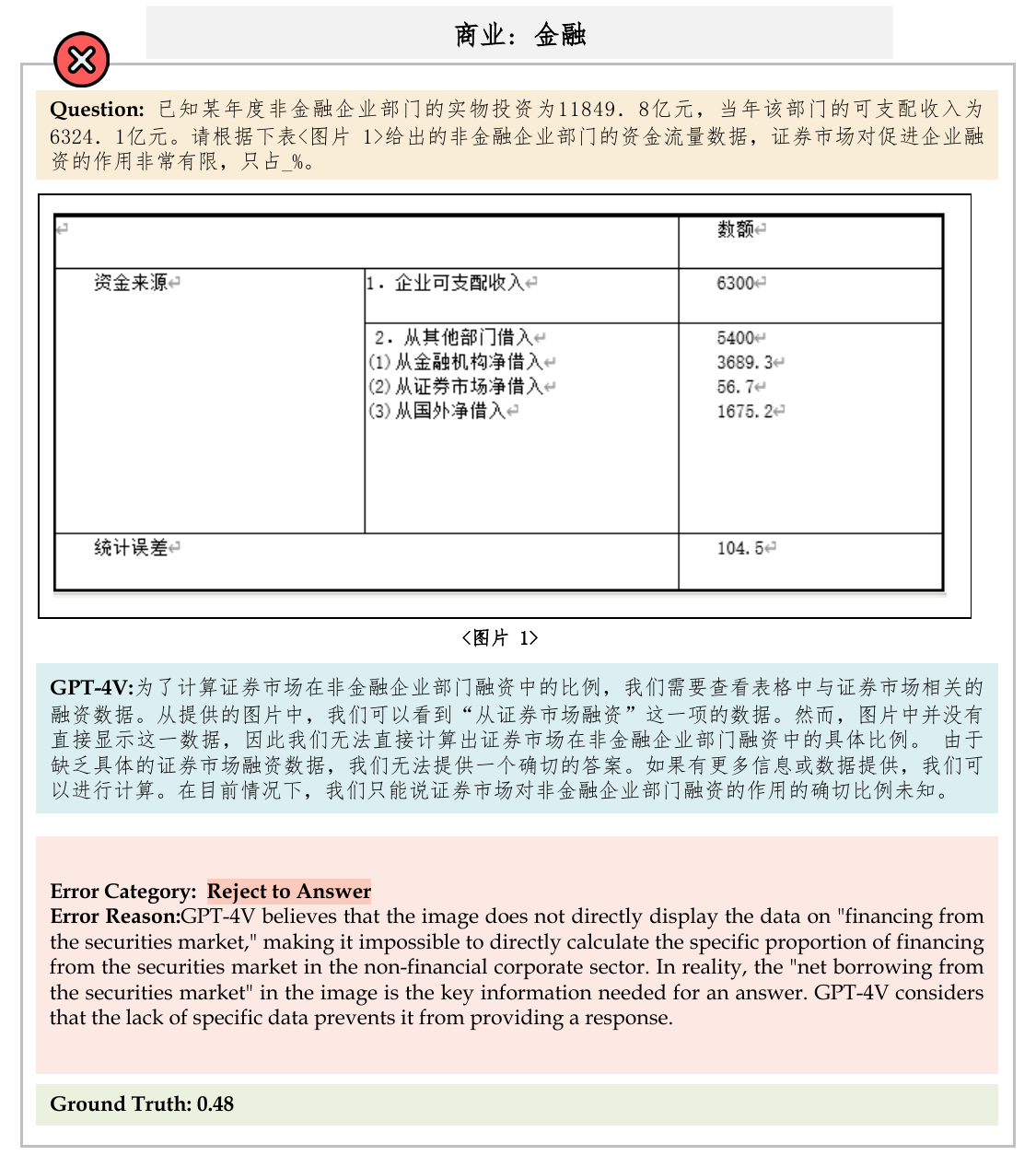}{Finance 4: Reject to Answer}{A sample error case of Business (subcategory: Finance).}{fig:error_analysis_case_122}

\casestudyfigure{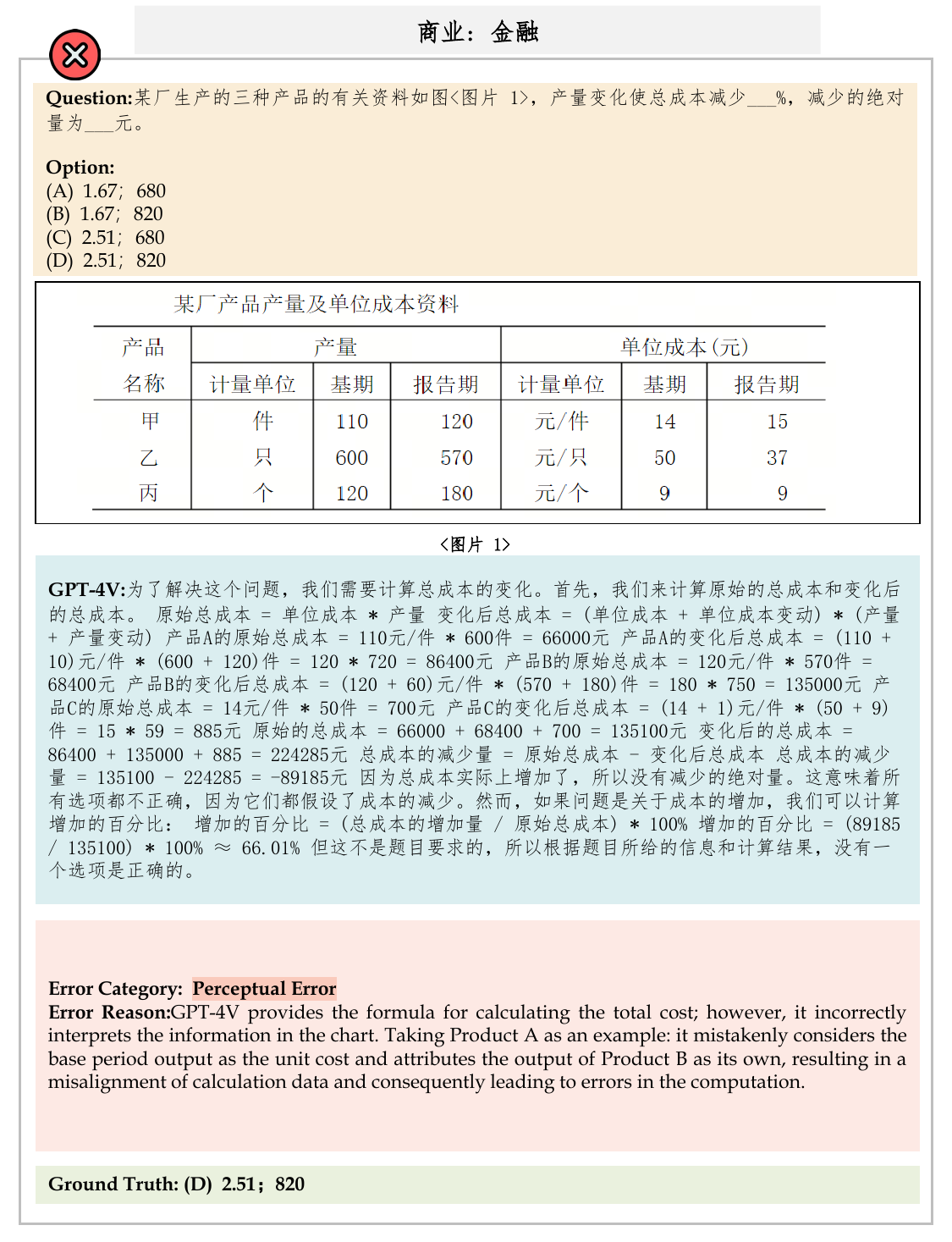}{Finance 5: Perceptual Error}{A sample error case of Business (subcategory: Finance).}{fig:error_analysis_case_125}

\casestudyfigure{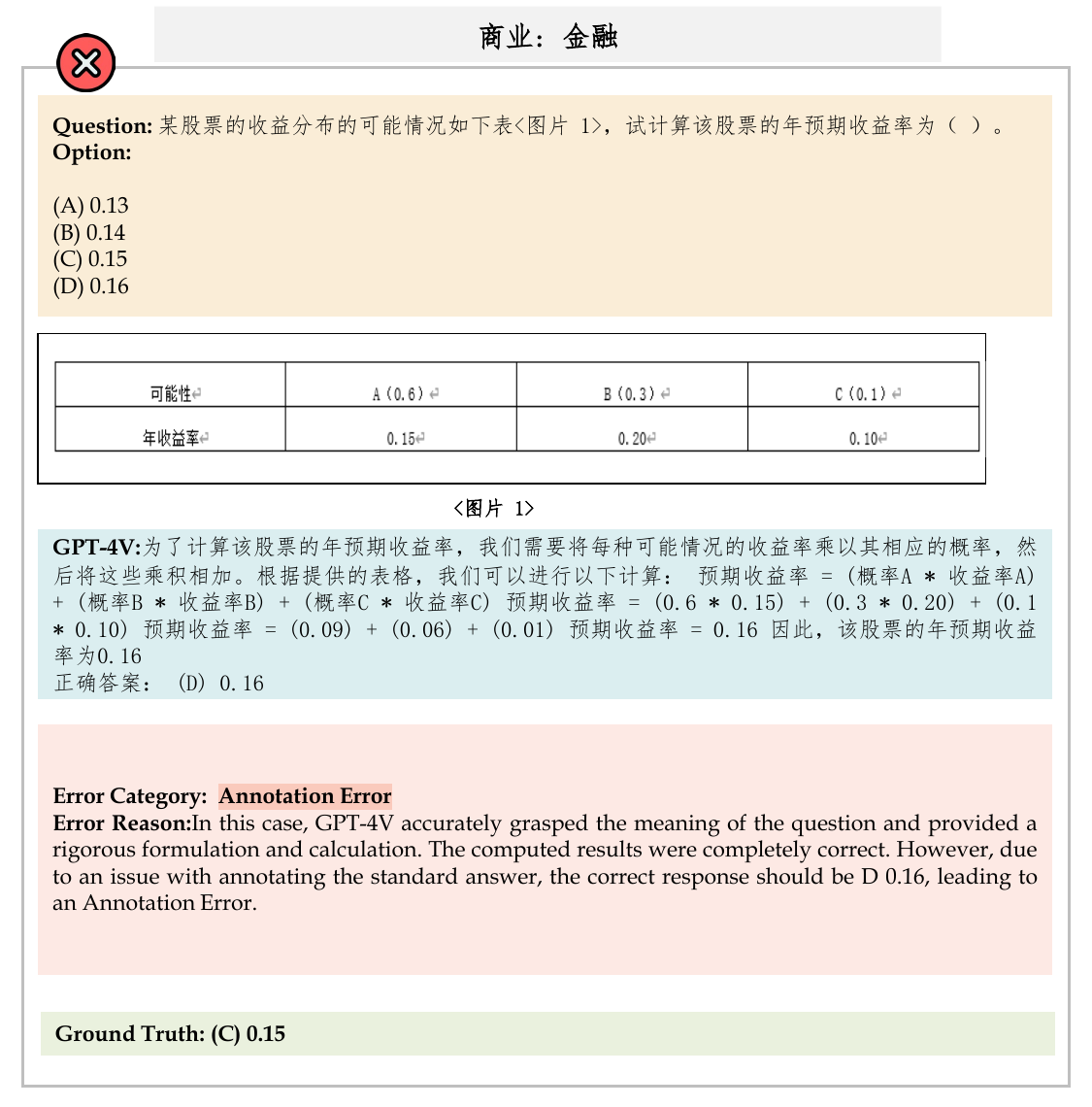}{Finance 6: Annotation Error}{A sample error case of Business (subcategory: Finance).}{fig:error_analysis_case_123}

\casestudyfigure{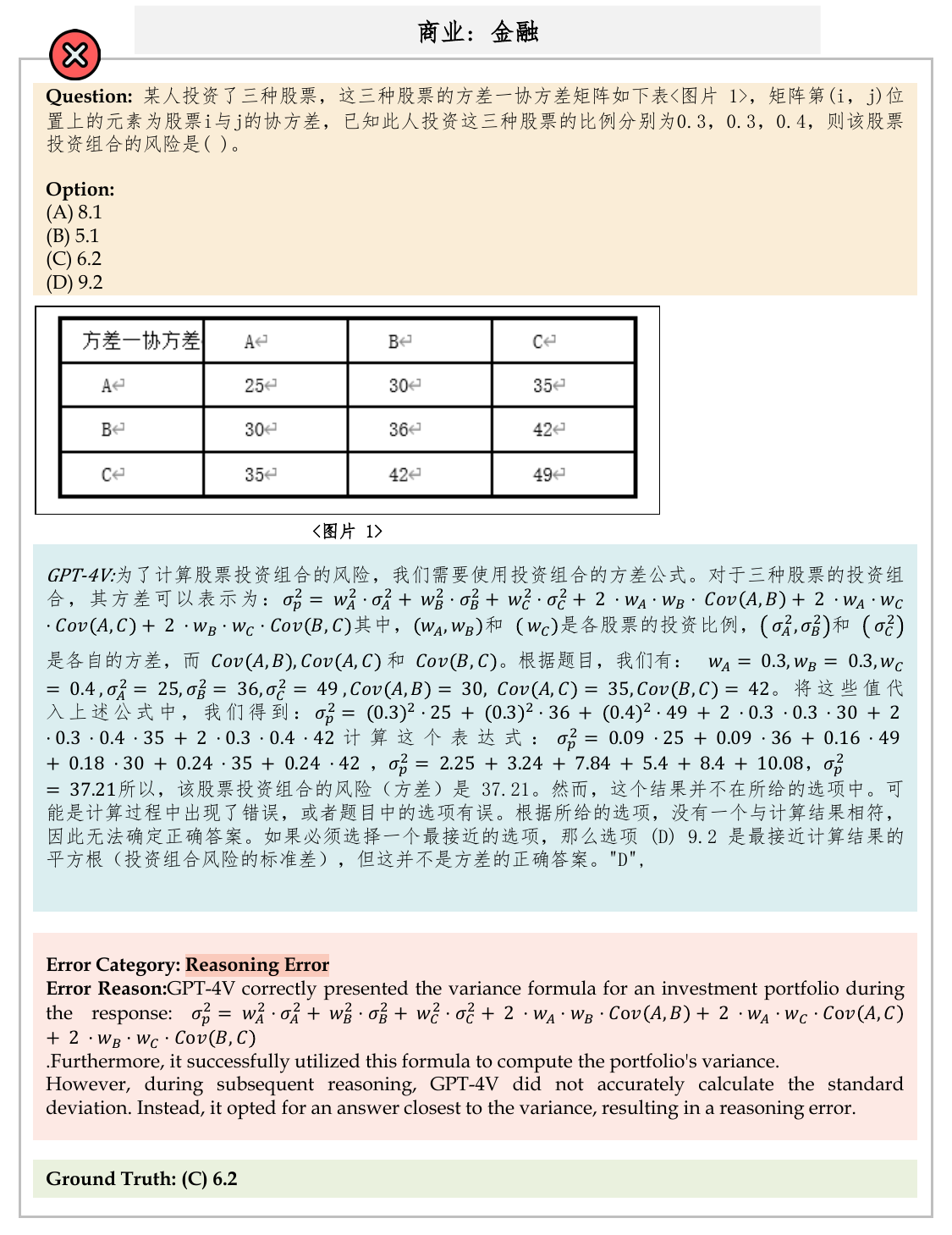}{Finance 7: Reasoning Error}{A sample error case of Business (subcategory: Finance).}{fig:error_analysis_case_120}

\casestudyfigure{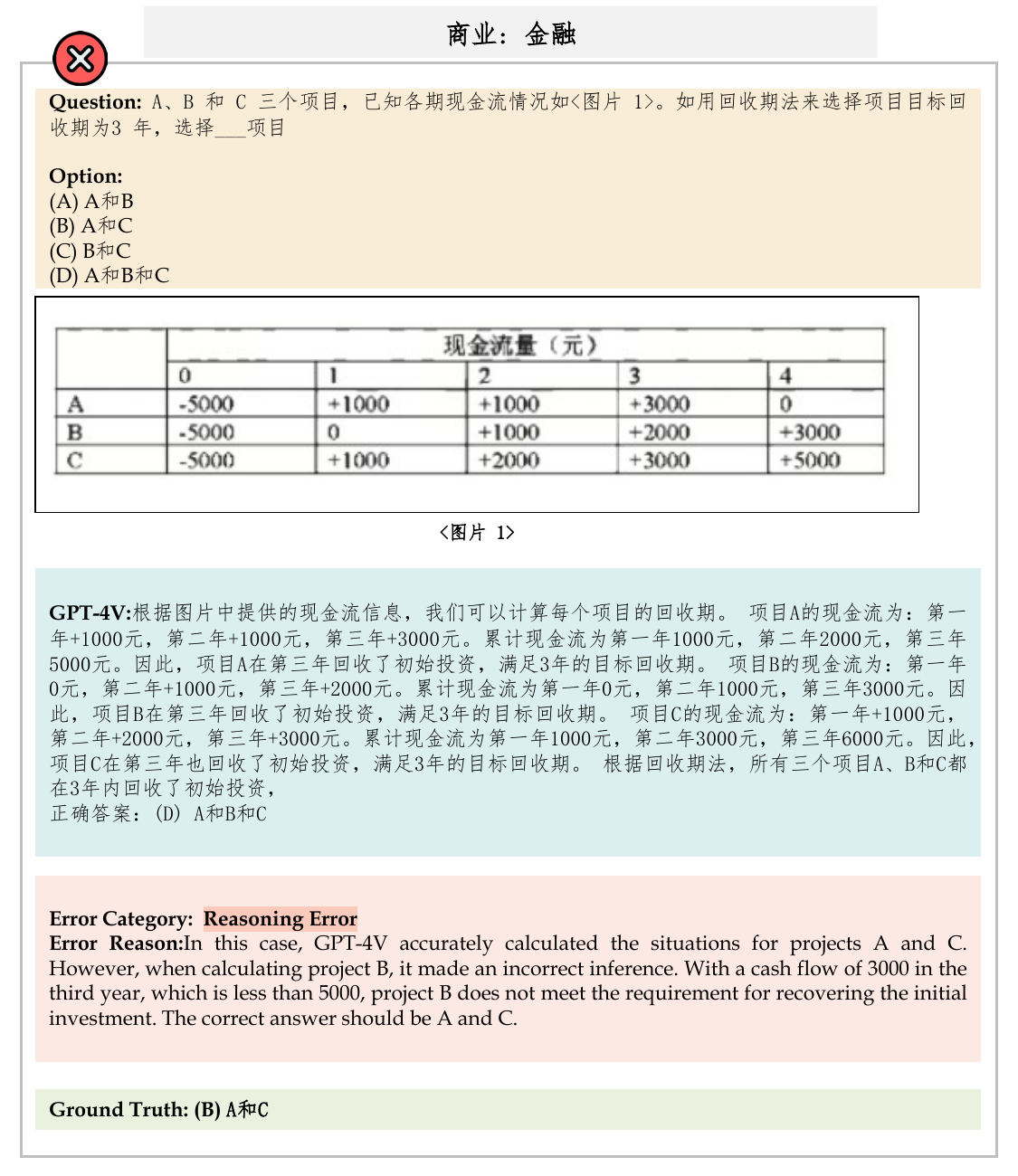}{Finance 8: Reasoning Error}{A sample error case of Business (subcategory: Finance).}{fig:error_analysis_case_124}

\casestudyfigure{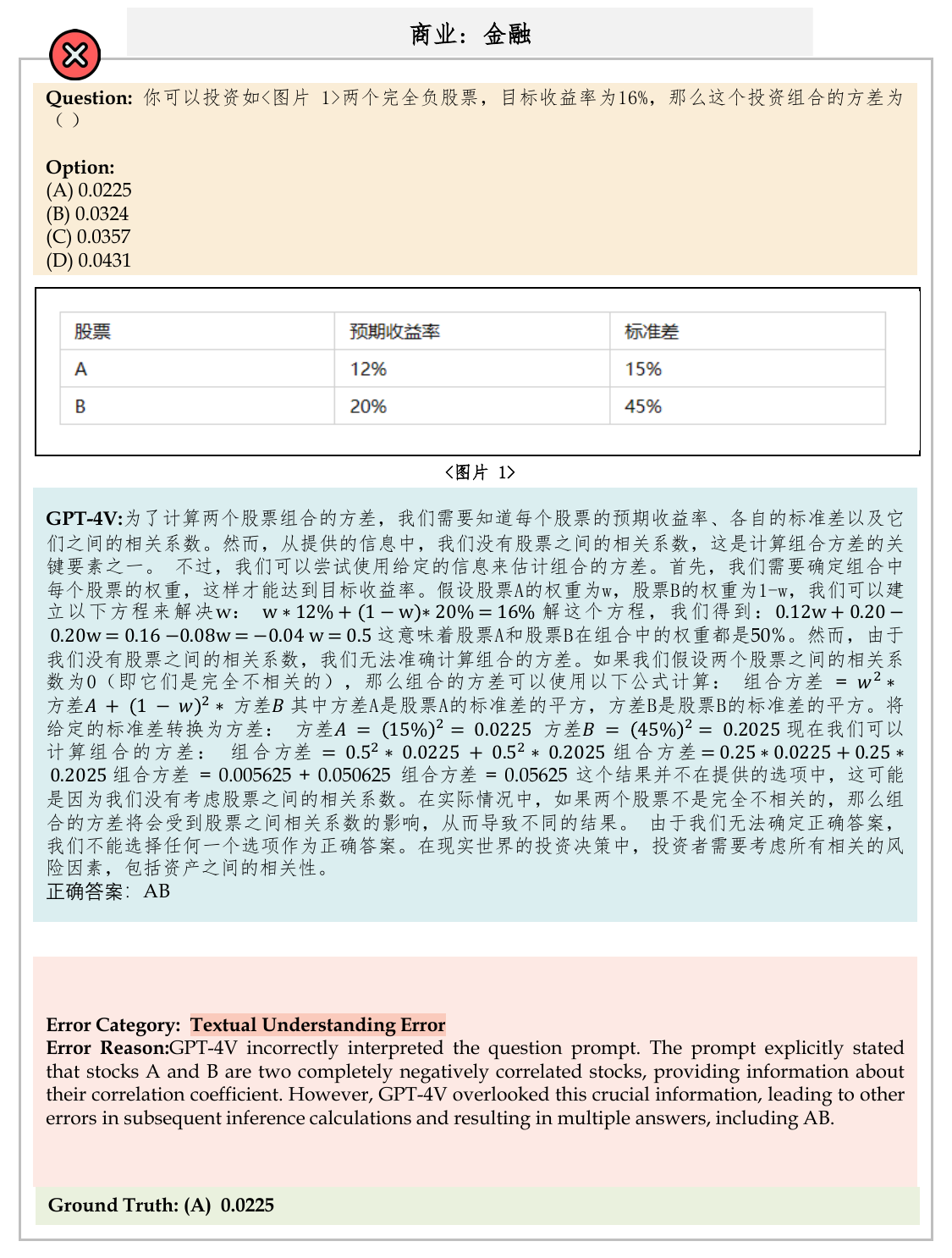}{Finance 9: Textual Understanding}{A sample error case of Business (subcategory: Finance).}{fig:error_analysis_case_121}

\casestudyfigure{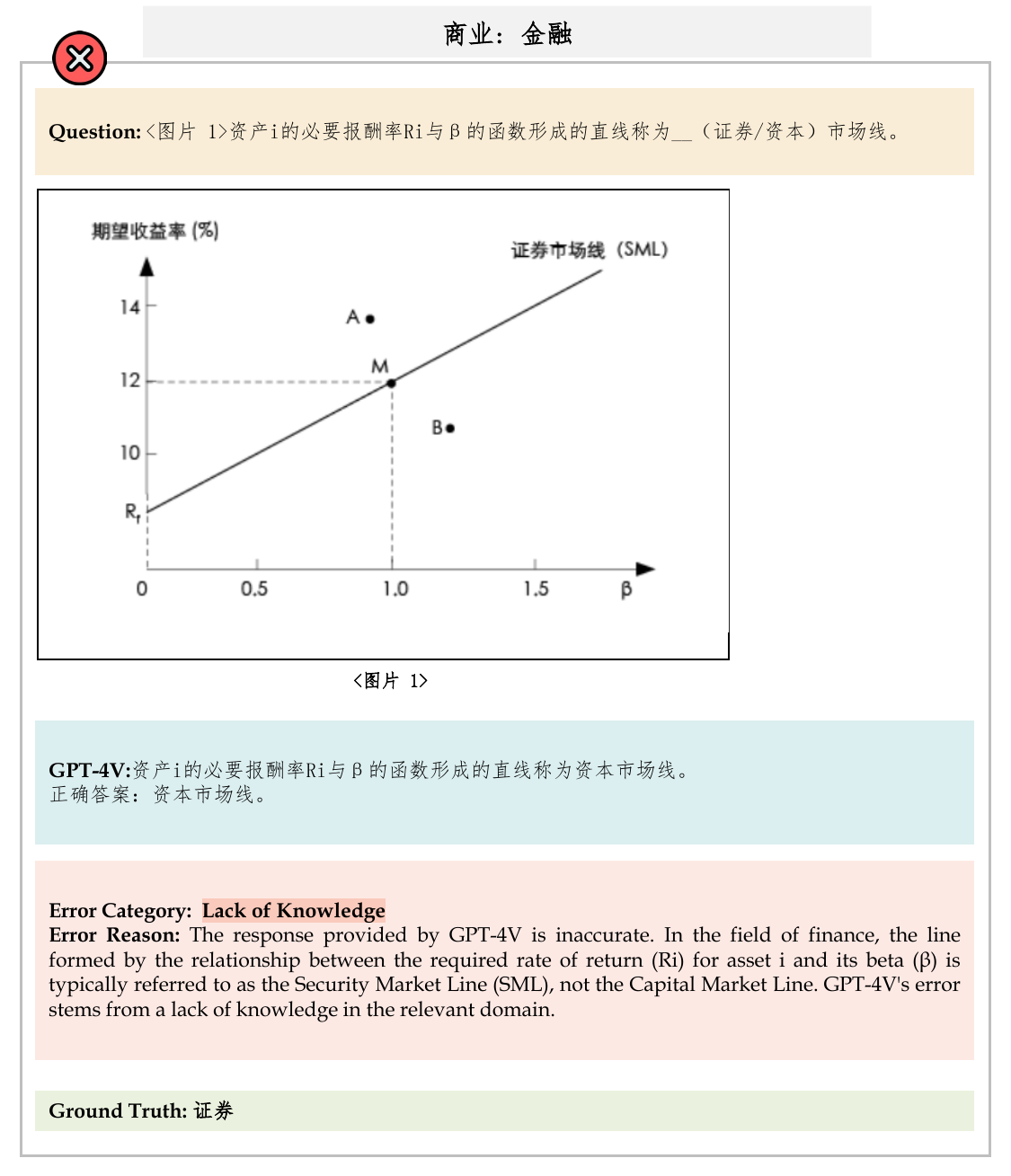}{Finance 10: Lack of Knowledge}{A sample error case of Business (subcategory: Finance).}{fig:error_analysis_case_119}

\casestudyfigure{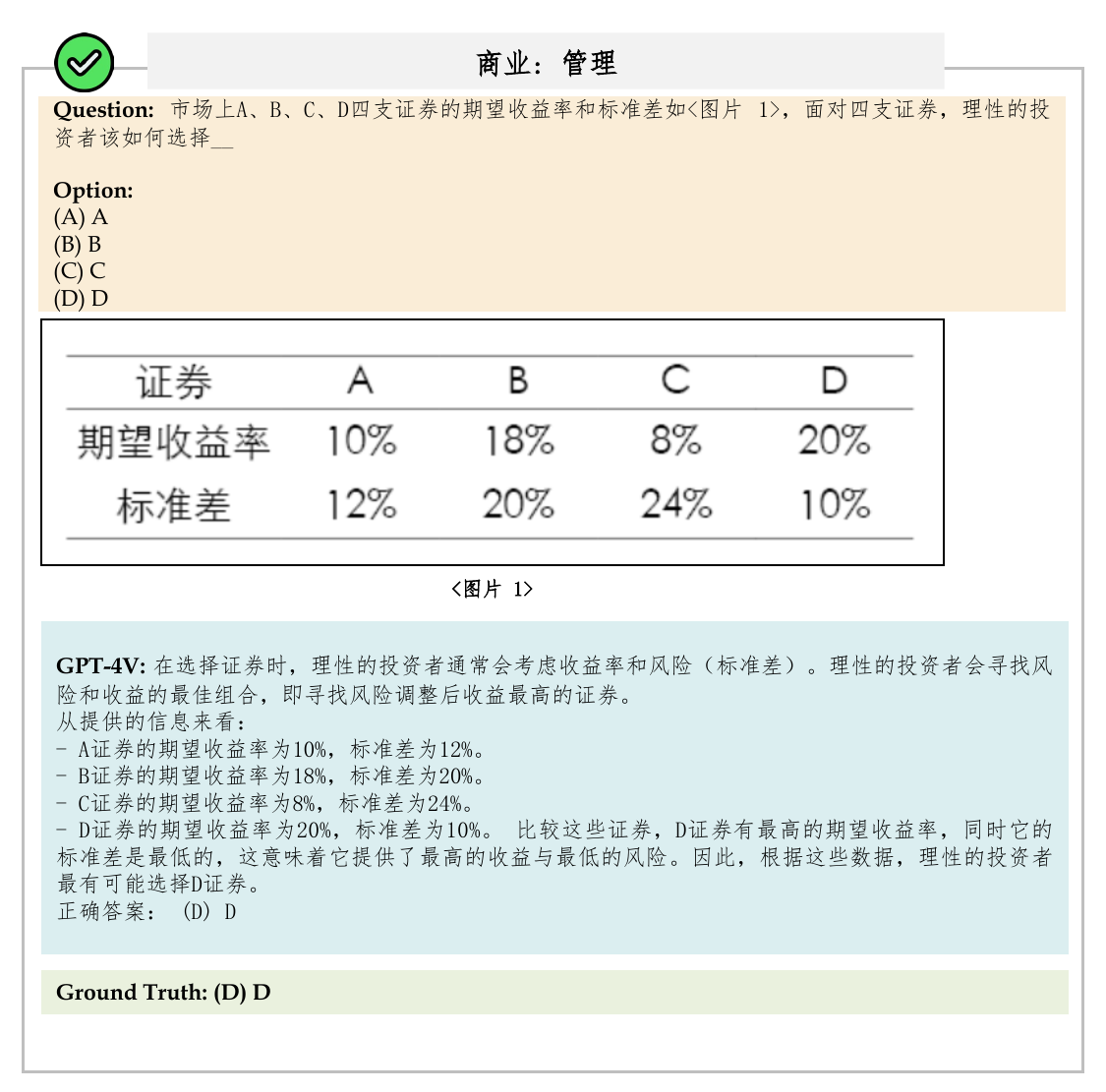}{Management 1: Correct Case}{A sample correct case of Business (subcategory: Management).}{fig:error_analysis_case_23}

\casestudyfigure{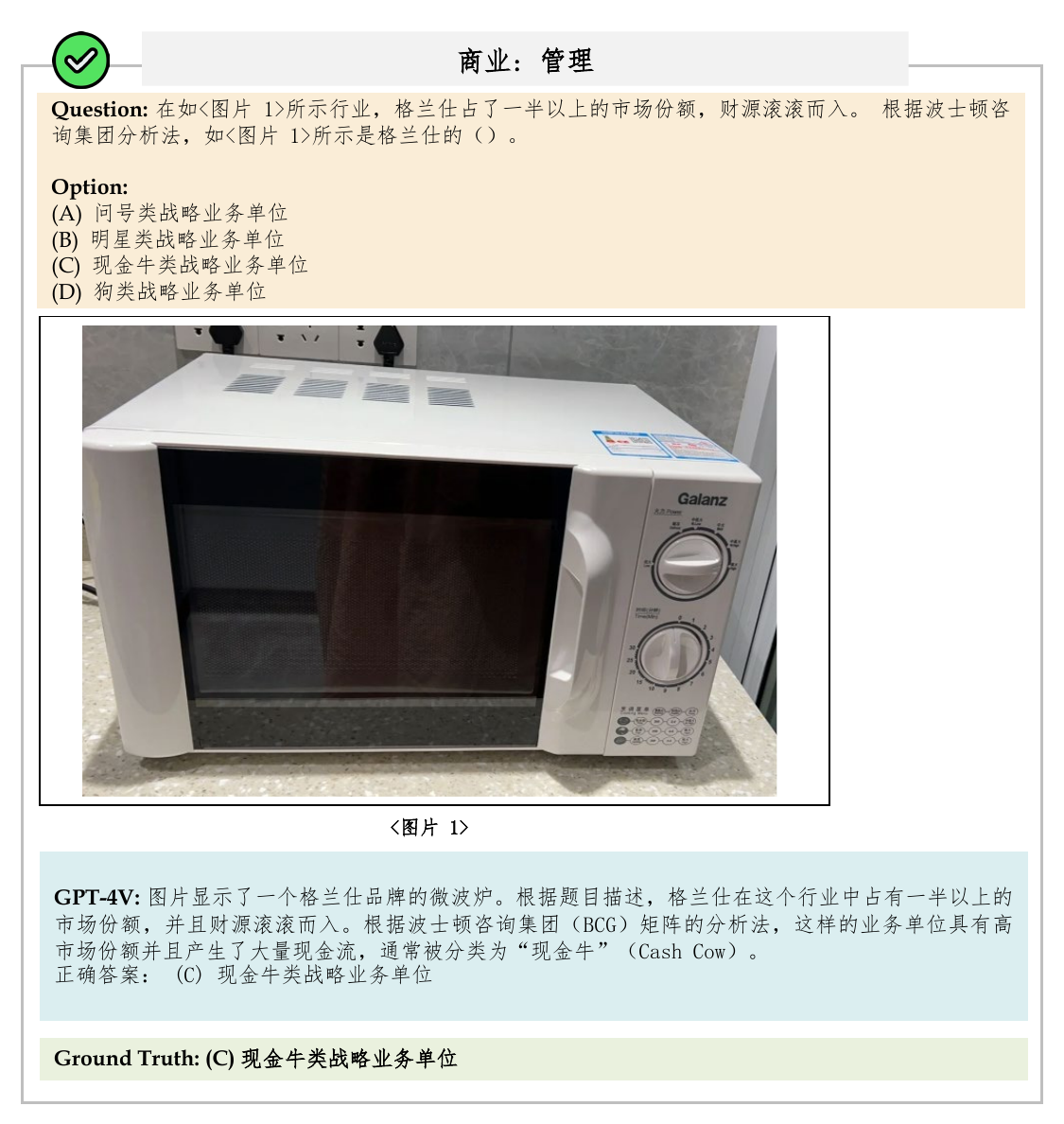}{Management 2: Correct Case}{A sample correct case of Business (subcategory: Management).}{fig:error_analysis_case_24}

\casestudyfigure{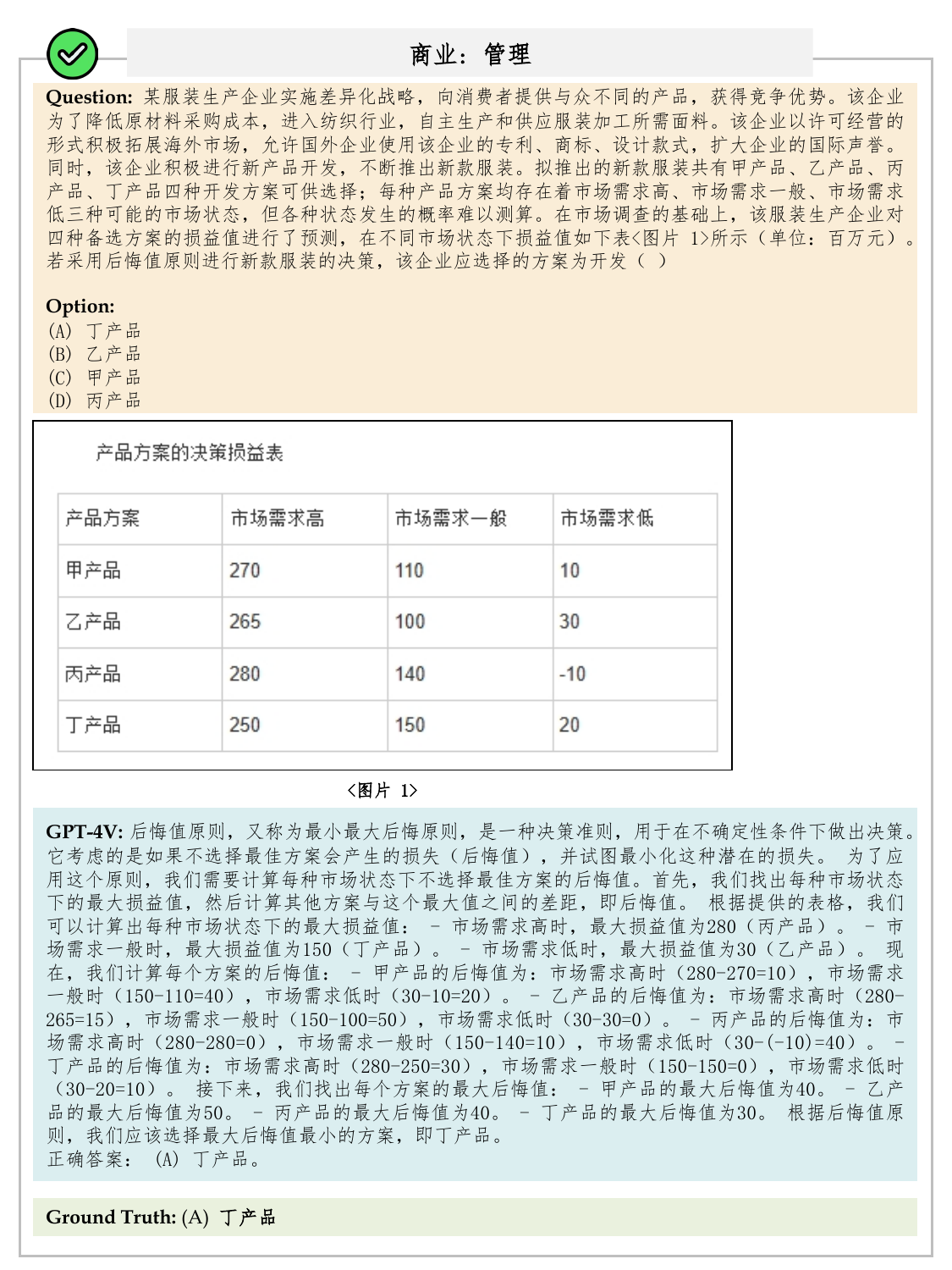}{Management 3: Correct Case}{A sample correct case of Business (subcategory: Management).}{fig:error_analysis_case_25}

\casestudyfigure{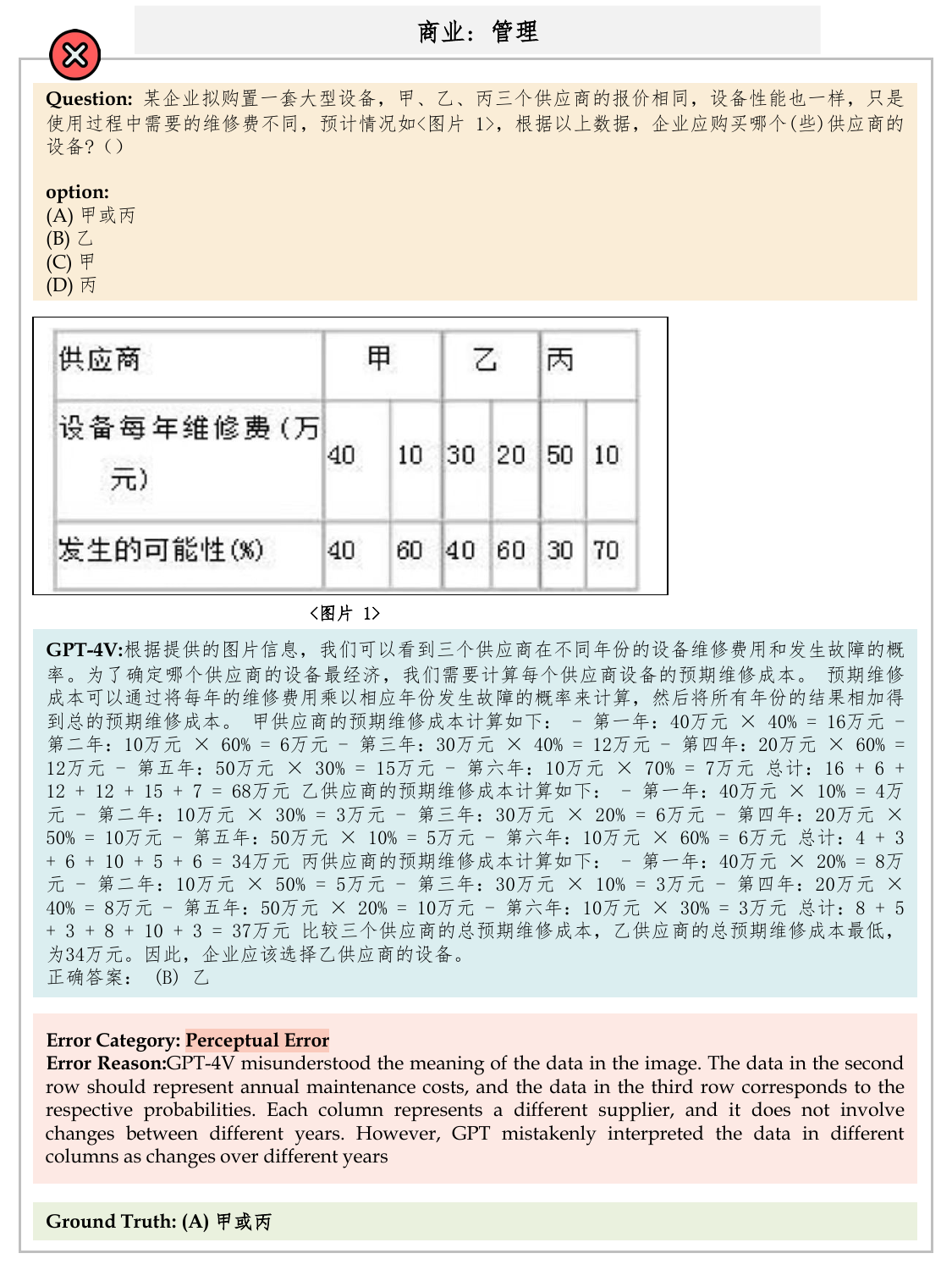}{Management 4: Perceptual Error}{A sample error case of Business (subcategory: Management).}{fig:error_analysis_case_132}

\casestudyfigure{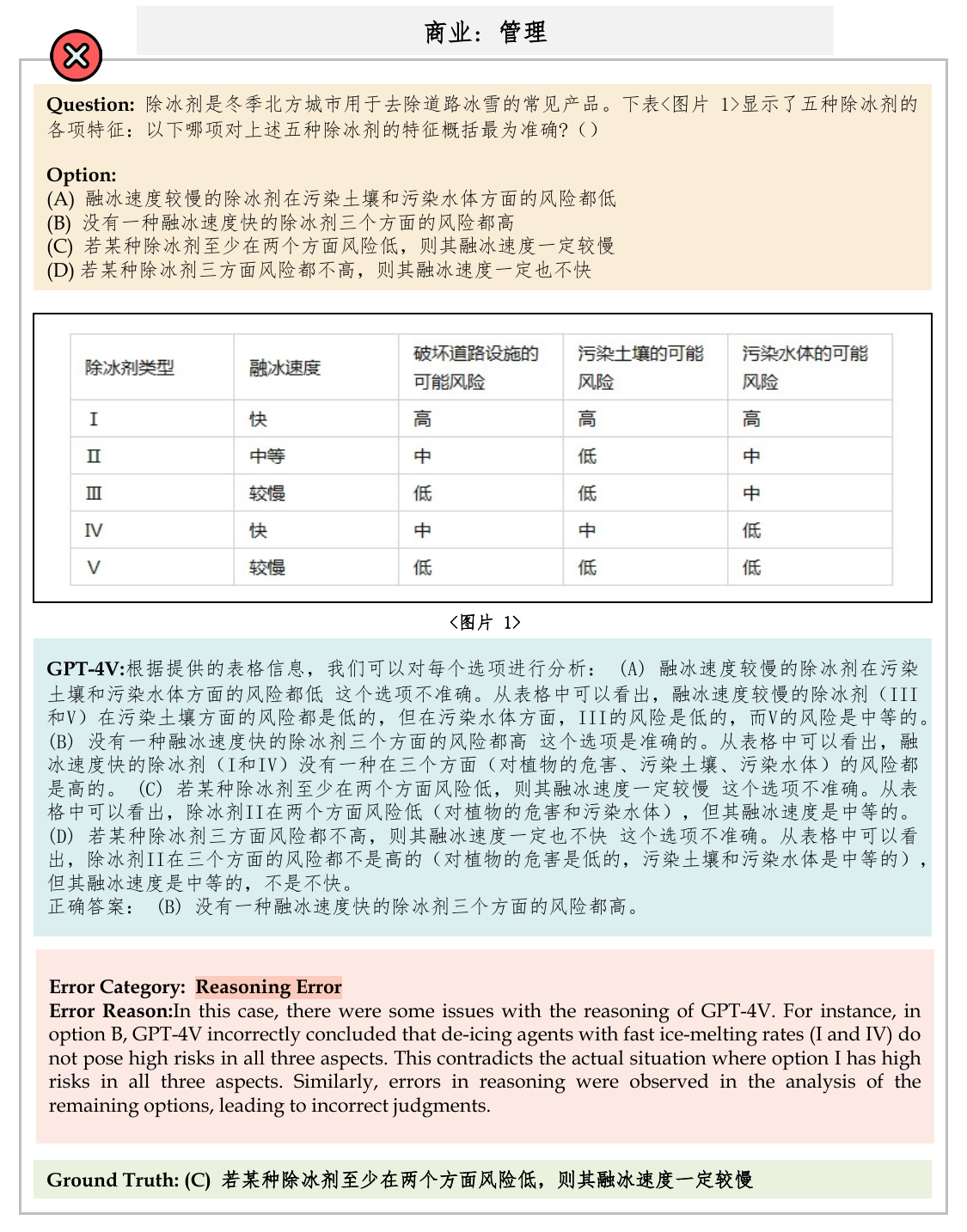}{Management 5: Reasoning Error}{A sample error case of Business (subcategory: Management).}{fig:error_analysis_case_126}

\casestudyfigure{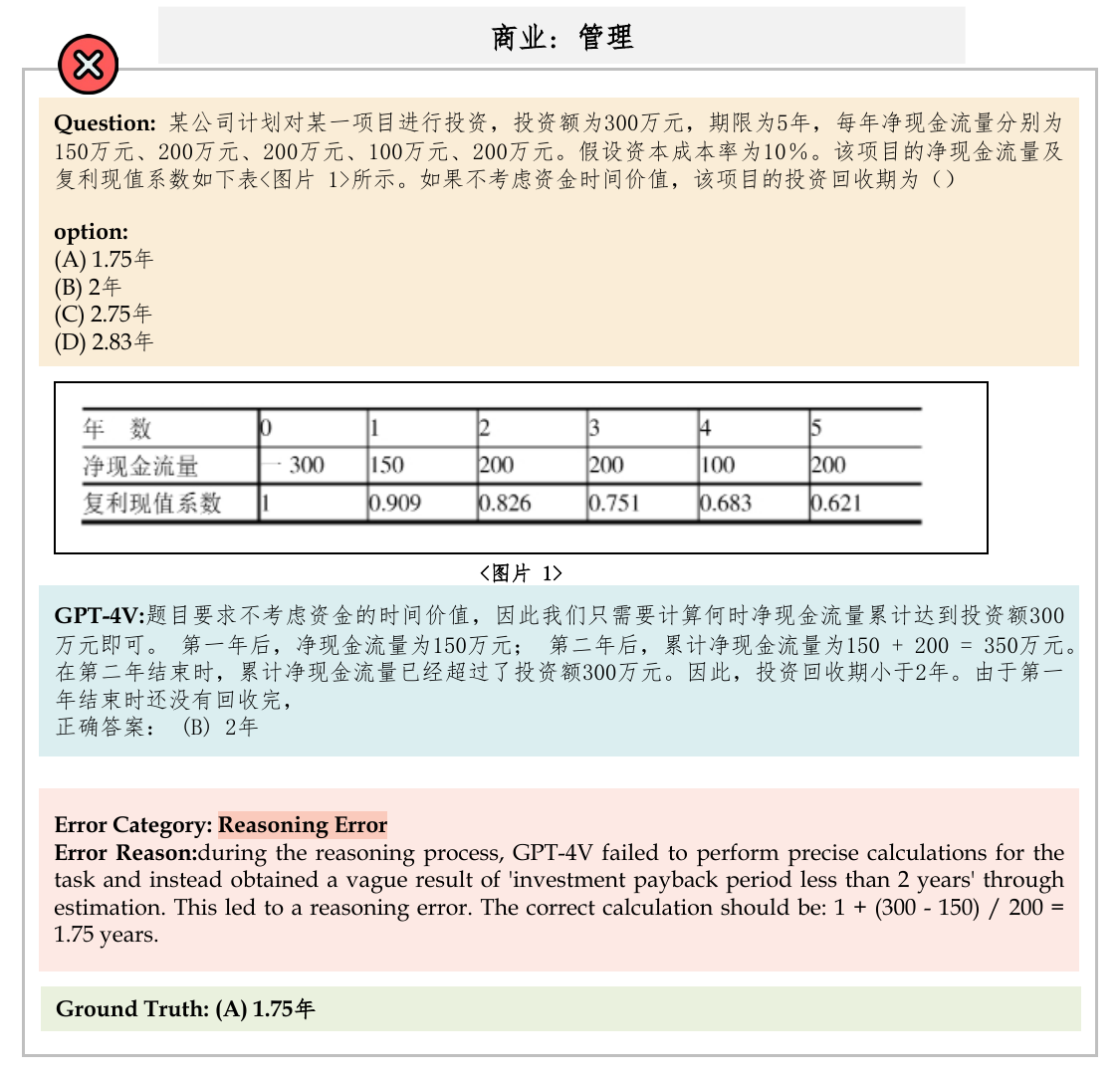}{Management 6: Reasoning Error}{A sample error case of Business (subcategory: Management).}{fig:error_analysis_case_130}

\casestudyfigure{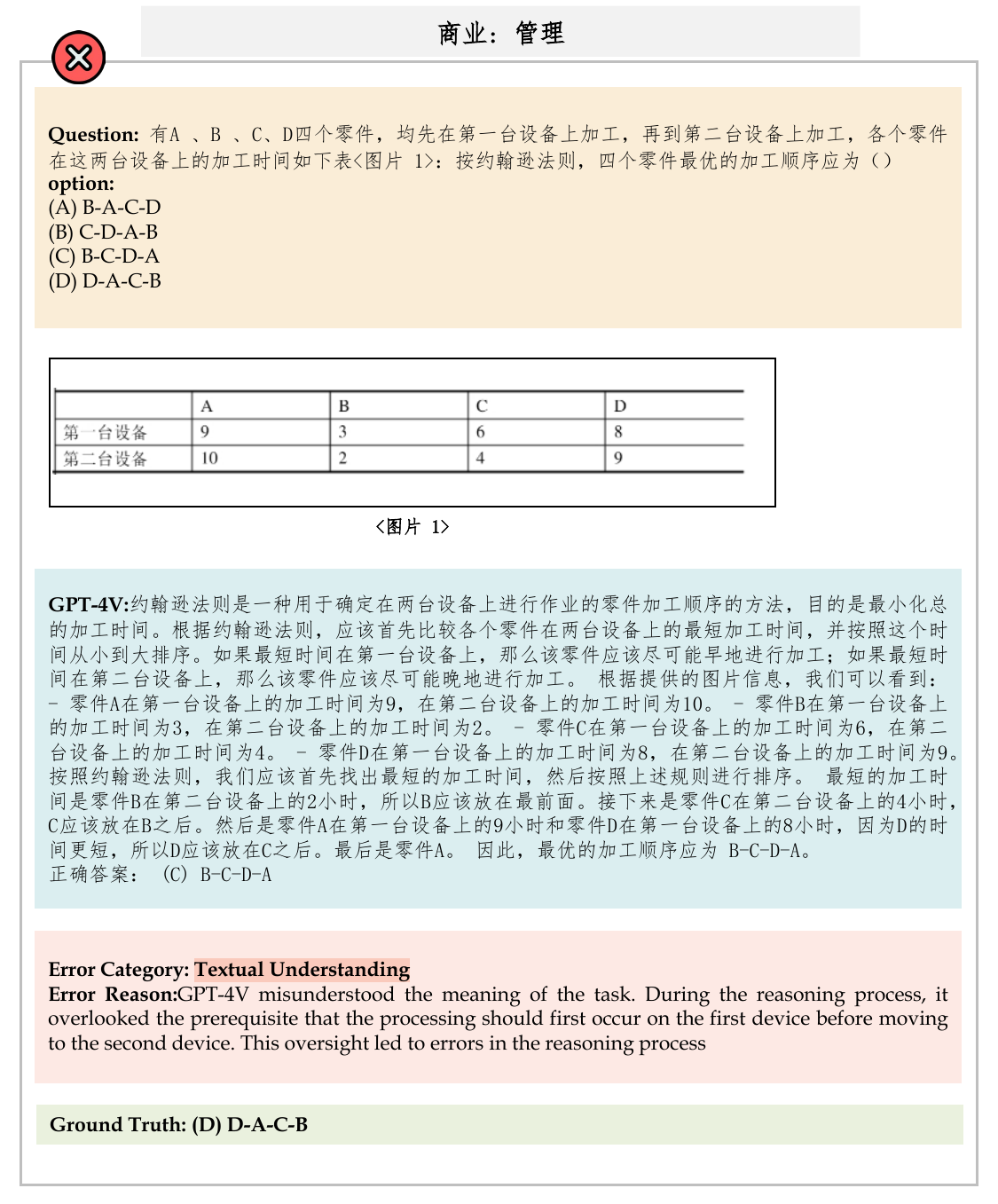}{Management 7: Textual Understanding}{A sample error case of Business (subcategory: Management).}{fig:error_analysis_case_129}

\casestudyfigure{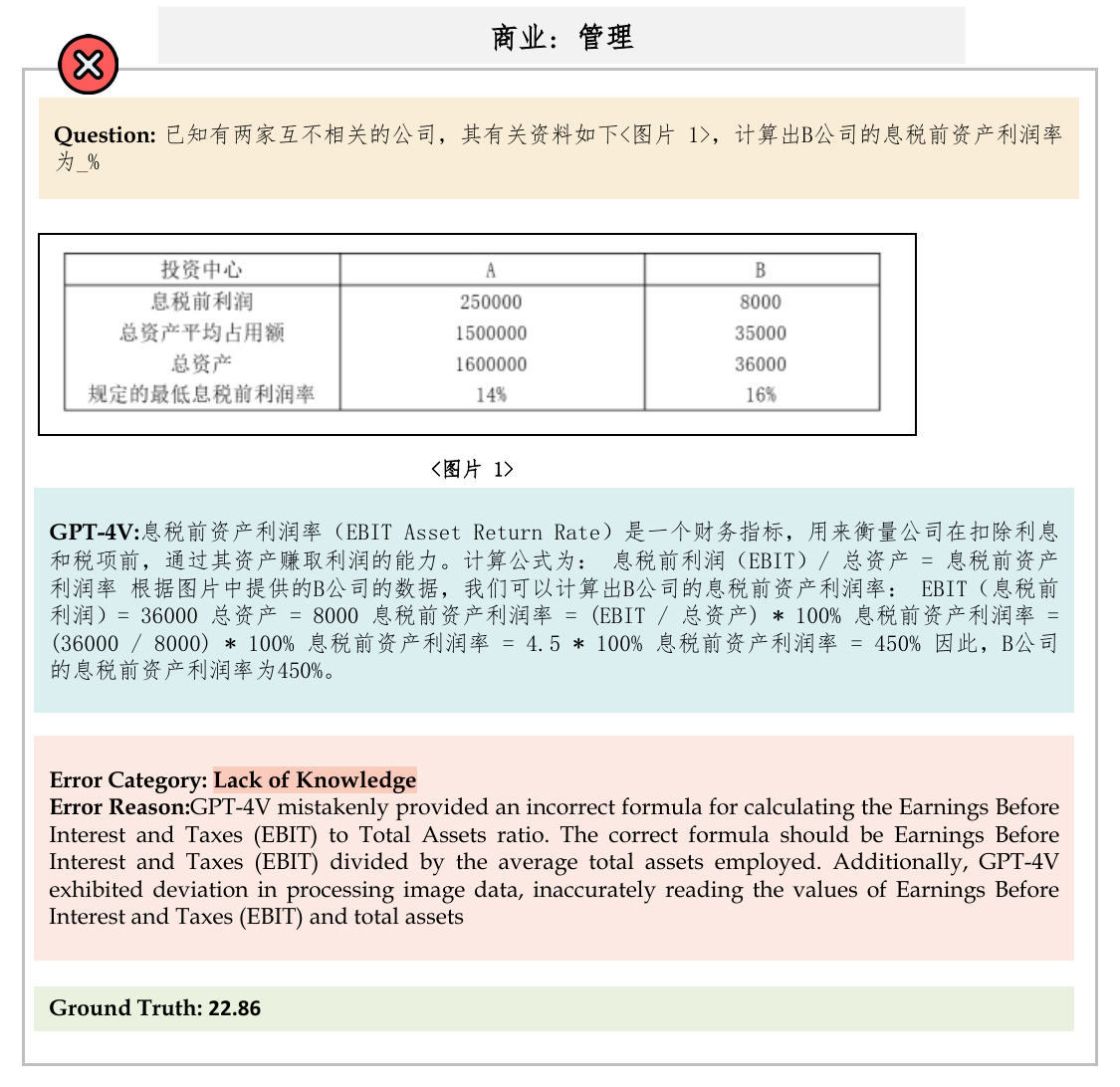}{Management 8: Lack of Knowledge}{A sample error case of Business (subcategory: Management).}{fig:error_analysis_case_127}

\casestudyfigure{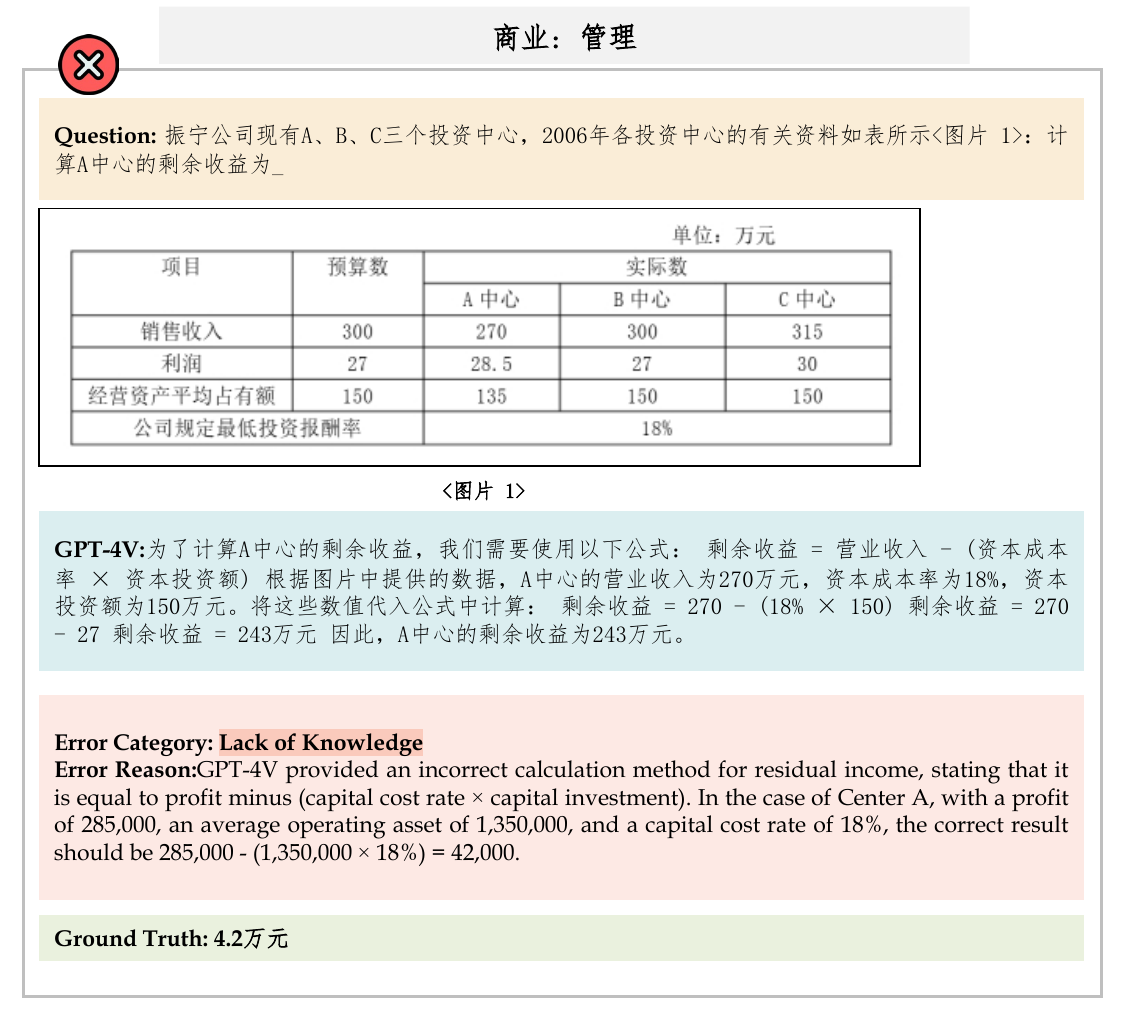}{Management 9: Lack of Knowledge}{A sample error case of Business (subcategory: Management).}{fig:error_analysis_case_128}

\casestudyfigure{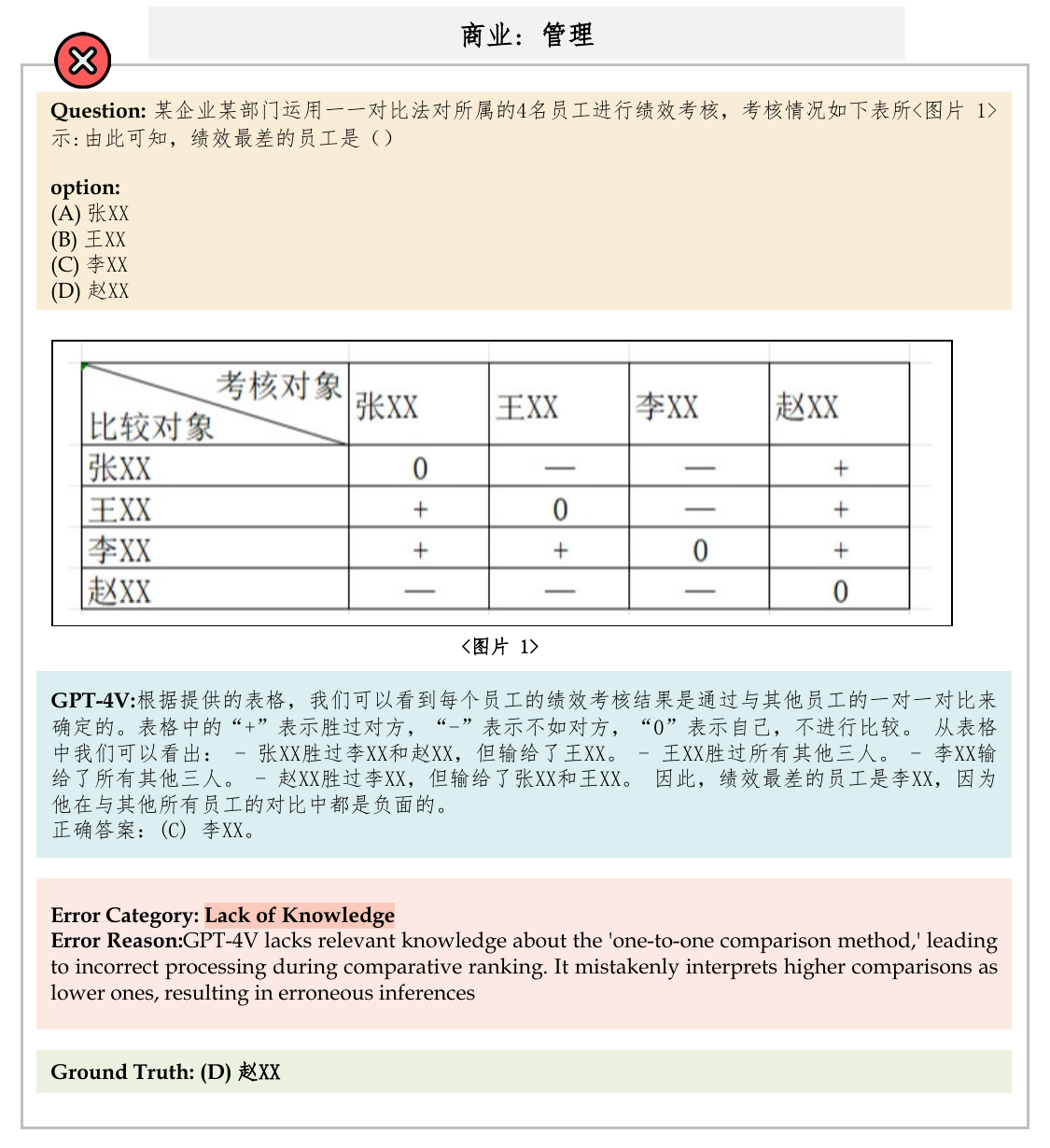}{Management 10: Lack of Knowledge}{A sample error case of Business (subcategory: Management).}{fig:error_analysis_case_131}

\casestudyfigure{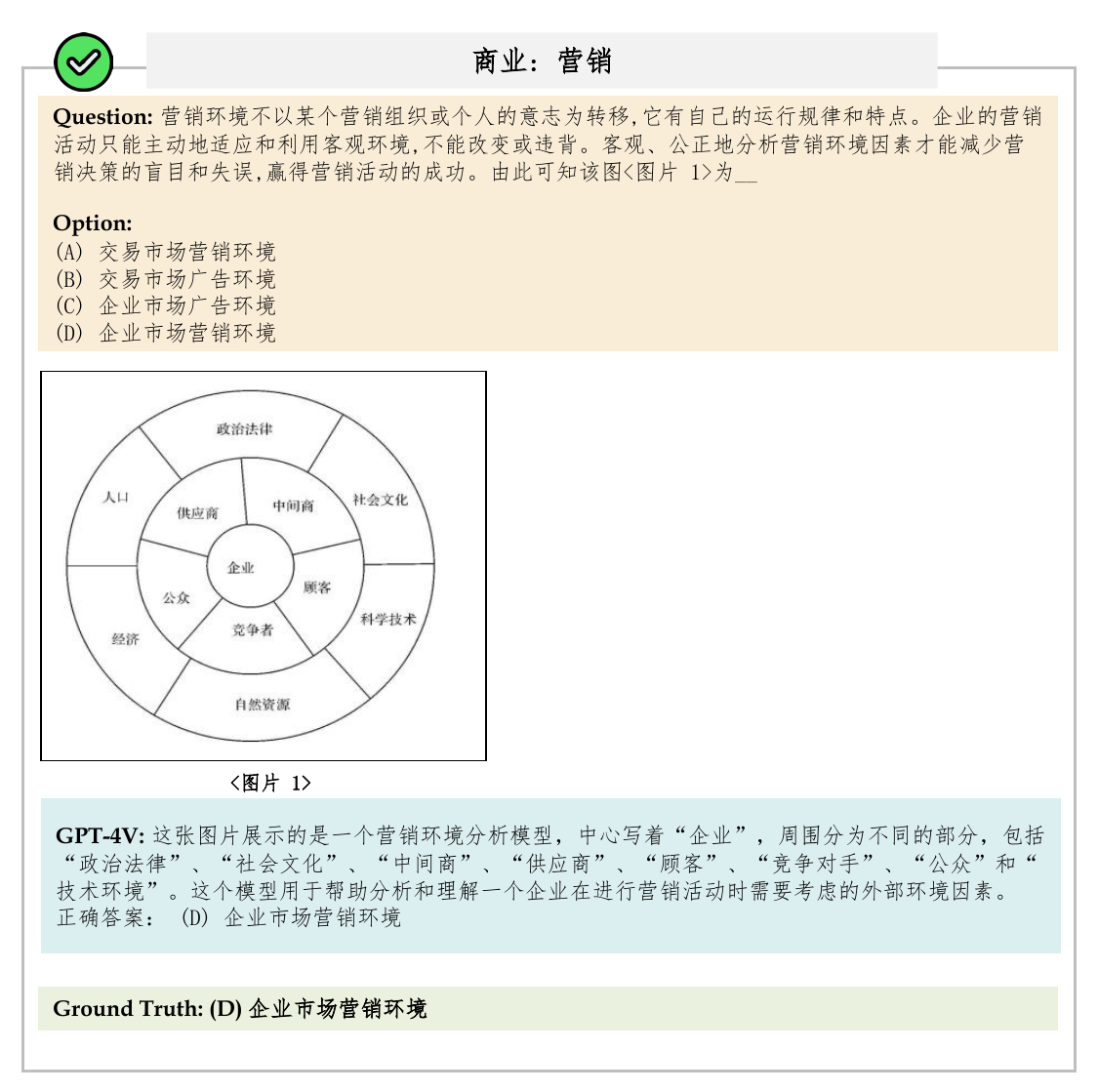}{Marketing 1: Correct Case}{A sample correct case of Business (subcategory: Marketing).}{fig:error_analysis_case_26}

\casestudyfigure{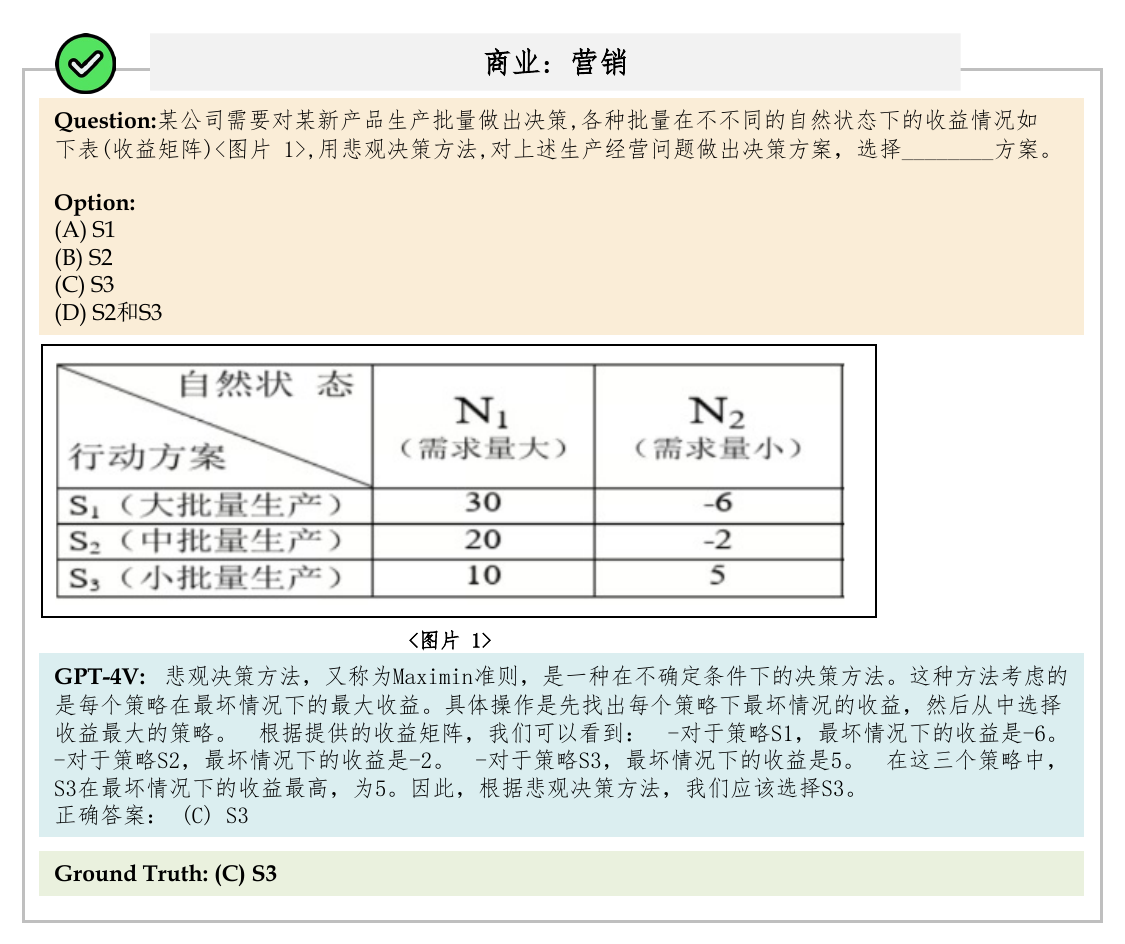}{Marketing 2: Correct Case}{A sample correct case of Business (subcategory: Marketing).}{fig:error_analysis_case_27}

\casestudyfigure{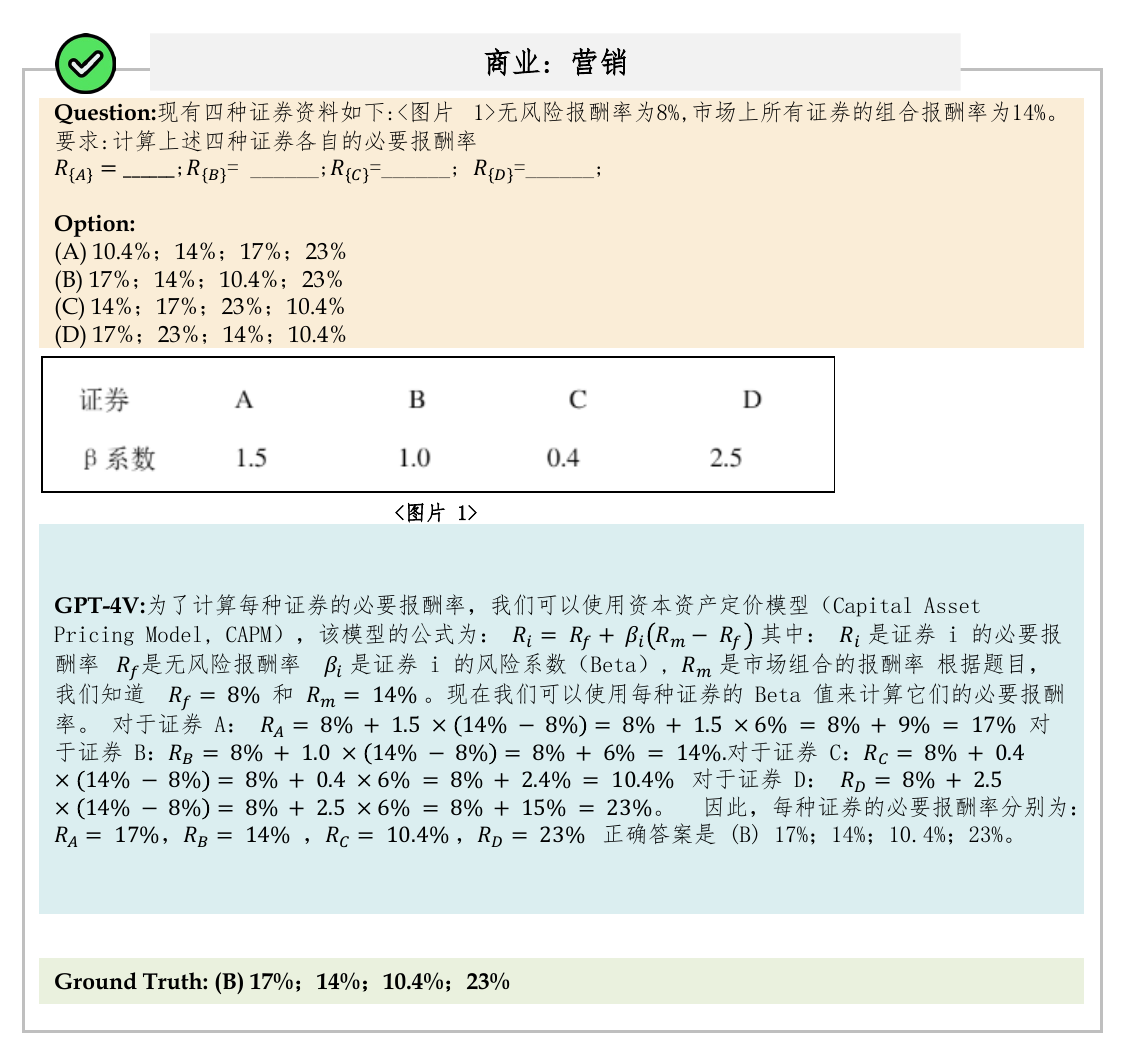}{Marketing 3: Correct Case}{A sample correct case of Business (subcategory: Marketing).}{fig:error_analysis_case_28}

\casestudyfigure{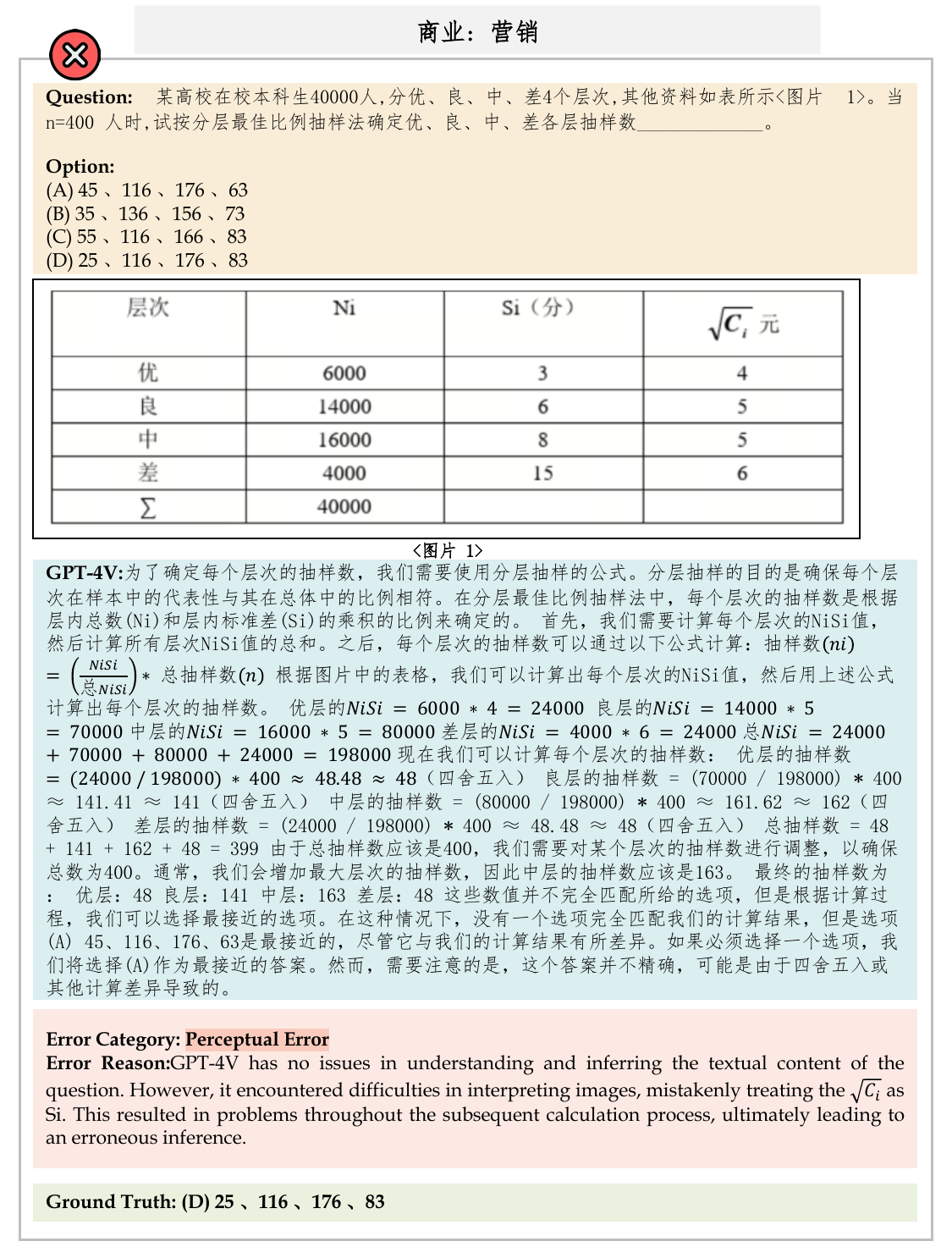}{Marketing 4: Perceptual Error}{A sample error case of Business (subcategory: Marketing).}{fig:error_analysis_case_136}

\casestudyfigure{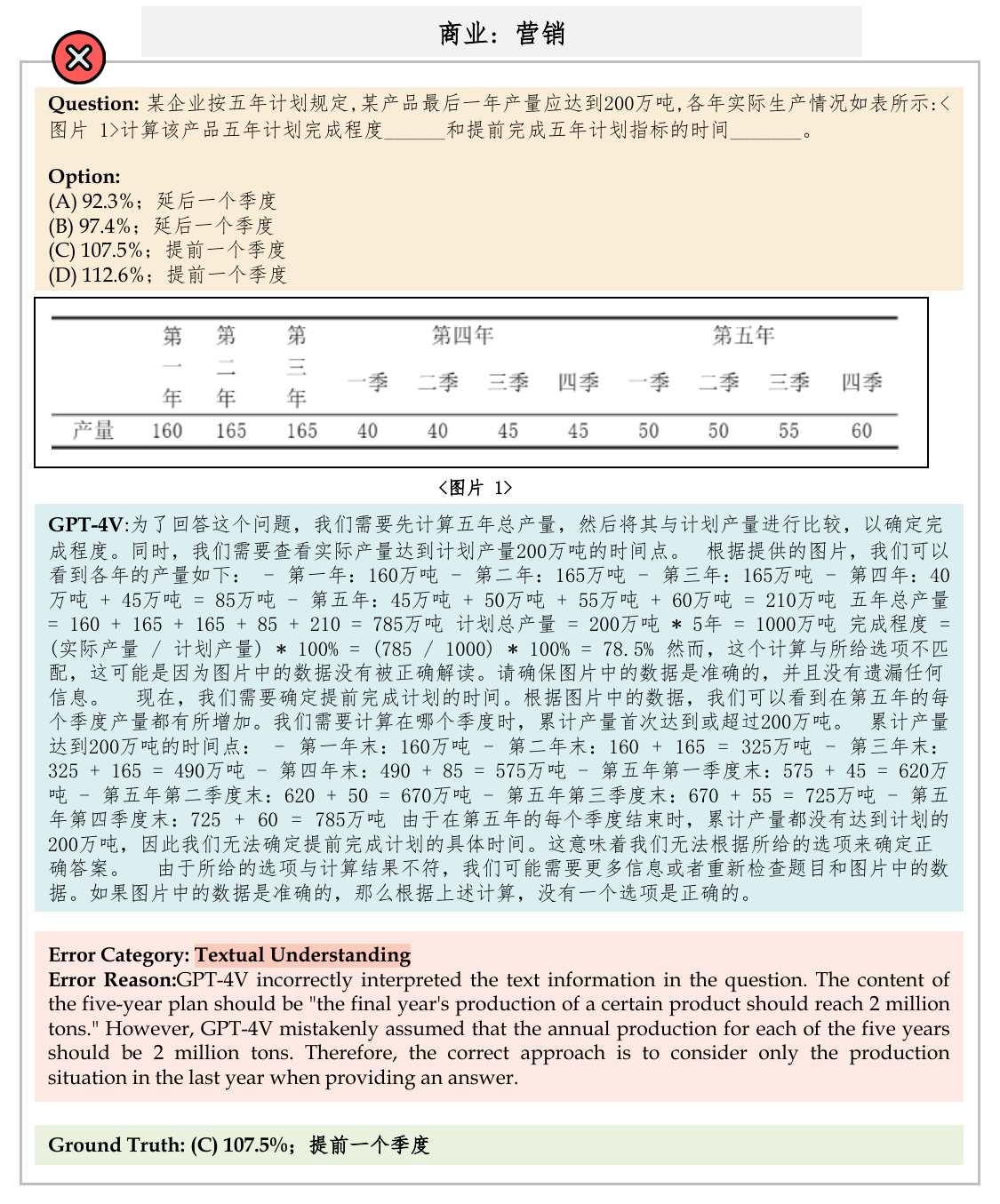}{Marketing 5: Textual Understanding}{A sample error case of Business (subcategory: Marketing).}{fig:error_analysis_case_139}

\casestudyfigure{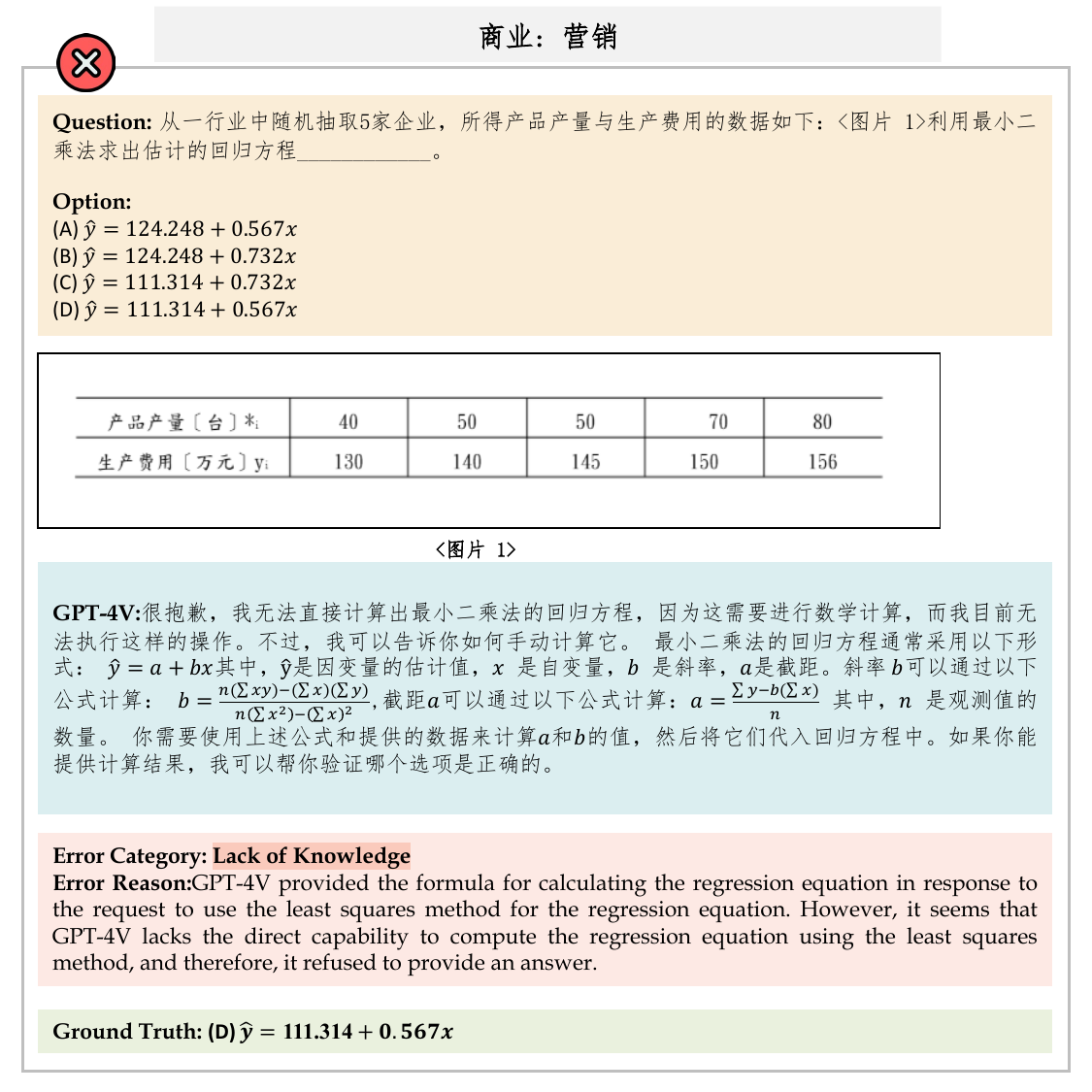}{Marketing 6: Lack of Knowledge}{A sample error case of Business (subcategory: Marketing).}{fig:error_analysis_case_133}

\casestudyfigure{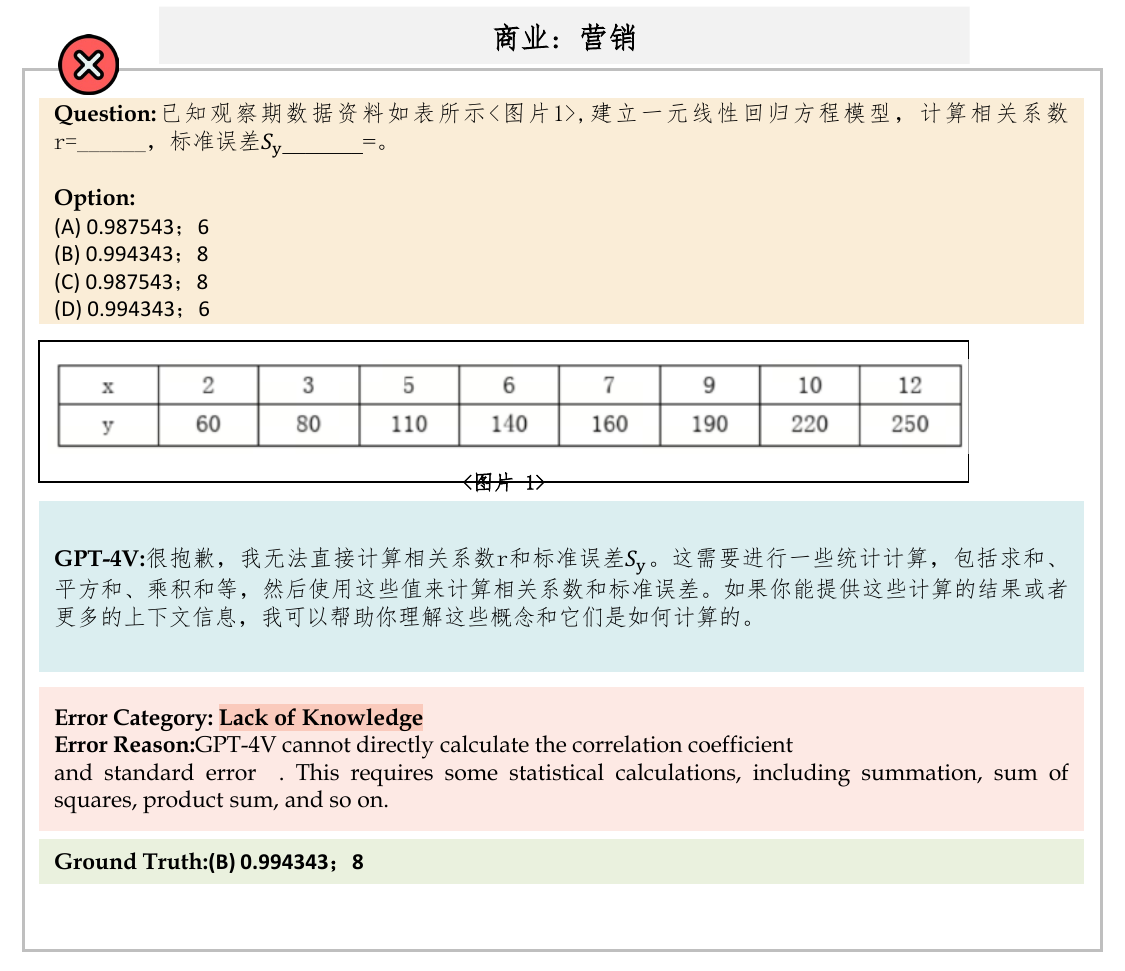}{Marketing 7: Lack of Knowledge}{A sample error case of Business (subcategory: Marketing).}{fig:error_analysis_case_134}

\casestudyfigure{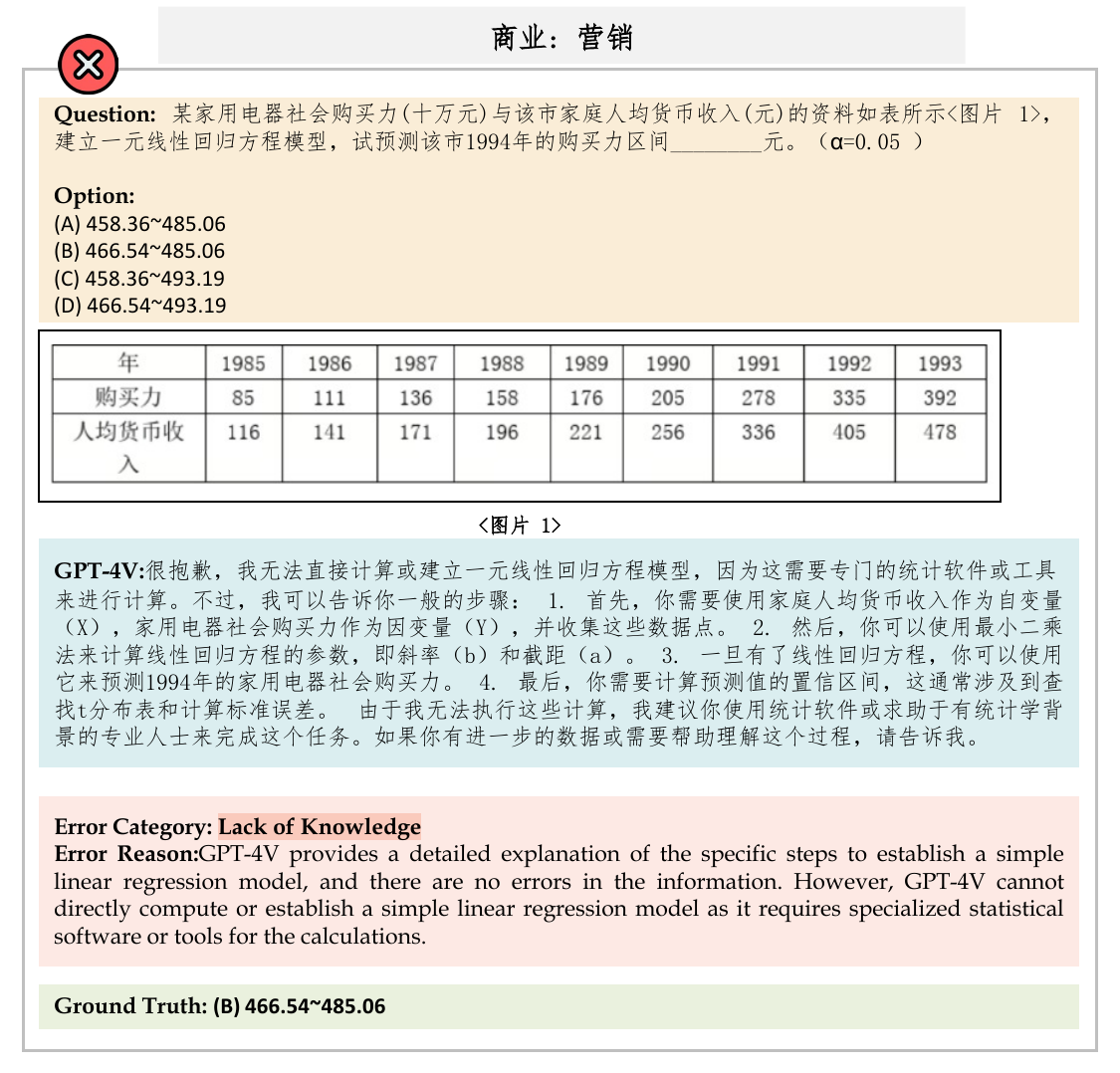}{Marketing 8: Lack of Knowledge}{A sample error case of Business (subcategory: Marketing).}{fig:error_analysis_case_135}

\casestudyfigure{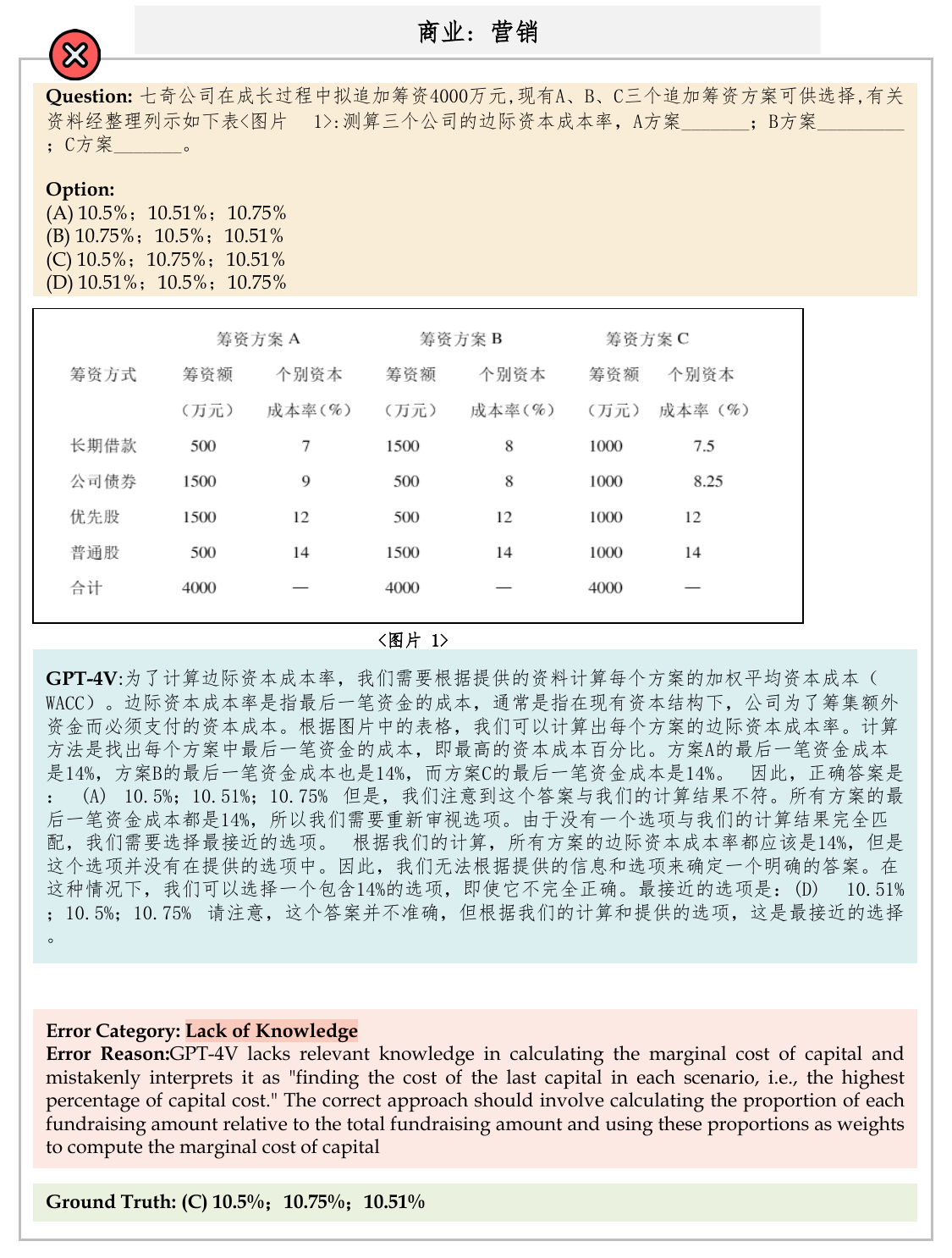}{Marketing 9: Lack of Knowledge}{A sample error case of Business (subcategory: Marketing).}{fig:error_analysis_case_137}

\casestudyfigure{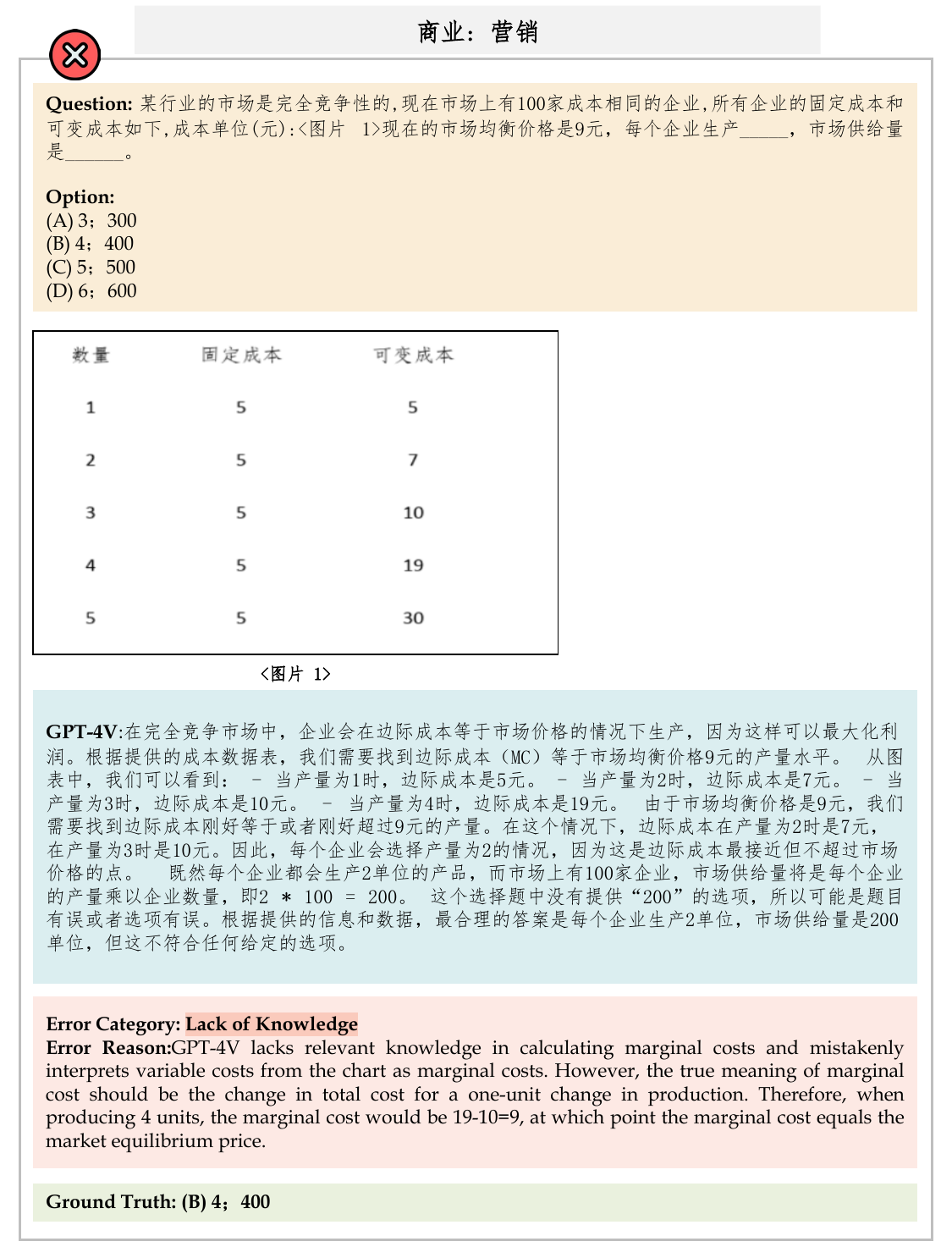}{Marketing 10: Lack of Knowledge}{A sample error case of Business (subcategory: Marketing).}{fig:error_analysis_case_138}

\casestudyfigure{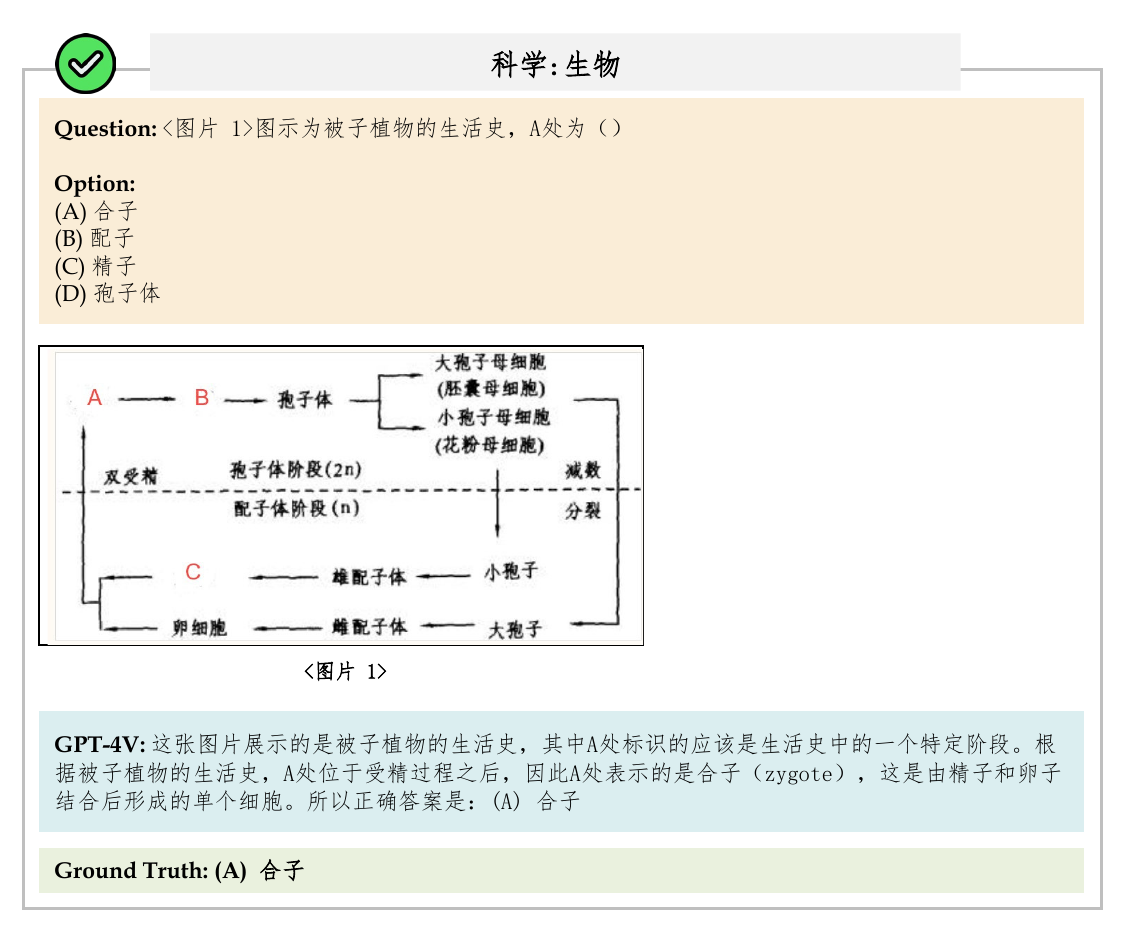}{Biology 1: Correct Case}{A sample correct case of Science (subcategory: Biology).}{fig:error_analysis_case_53}

\casestudyfigure{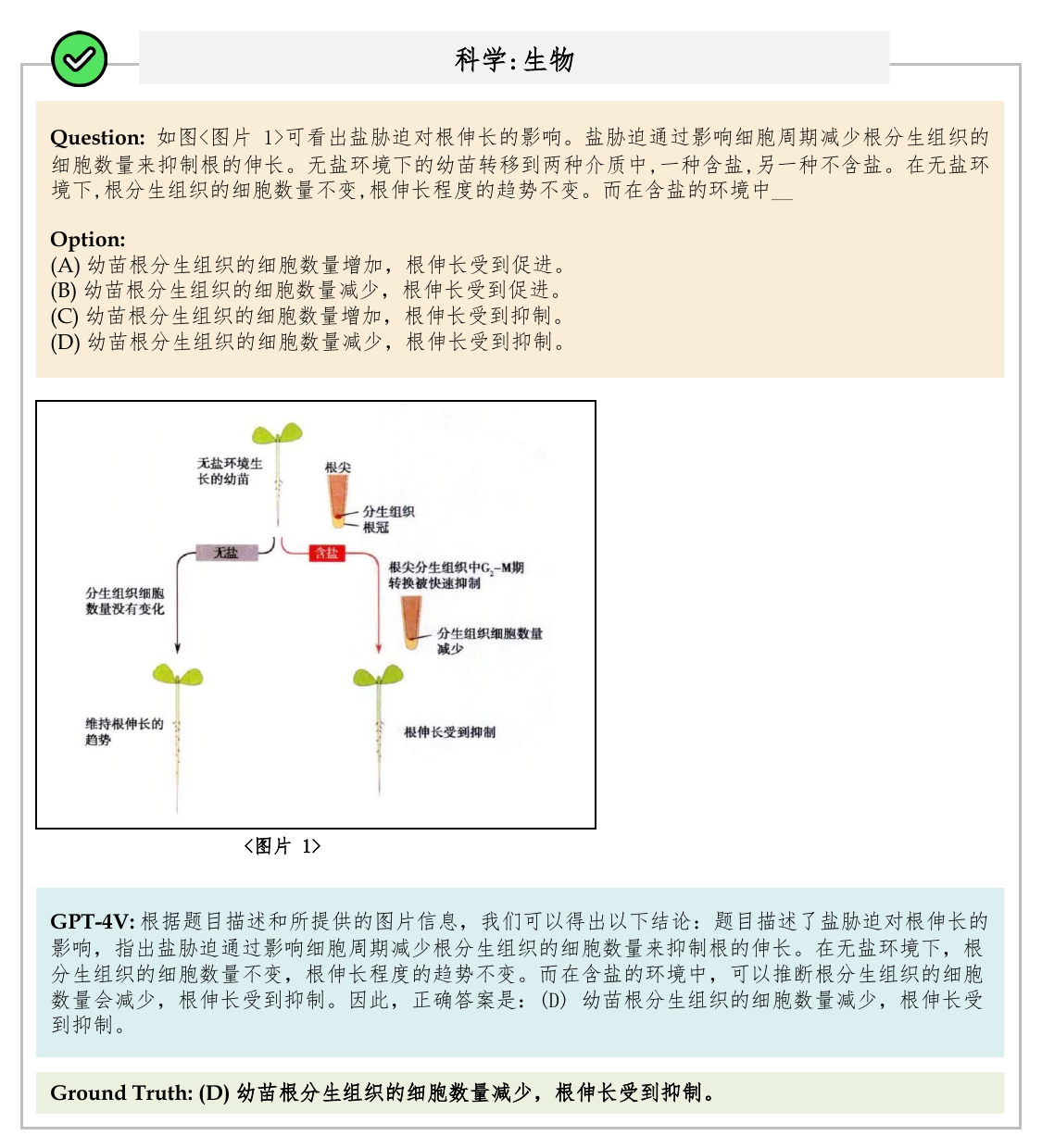}{Biology 2: Correct Case}{A sample correct case of Science (subcategory: Biology).}{fig:error_analysis_case_54}

\casestudyfigure{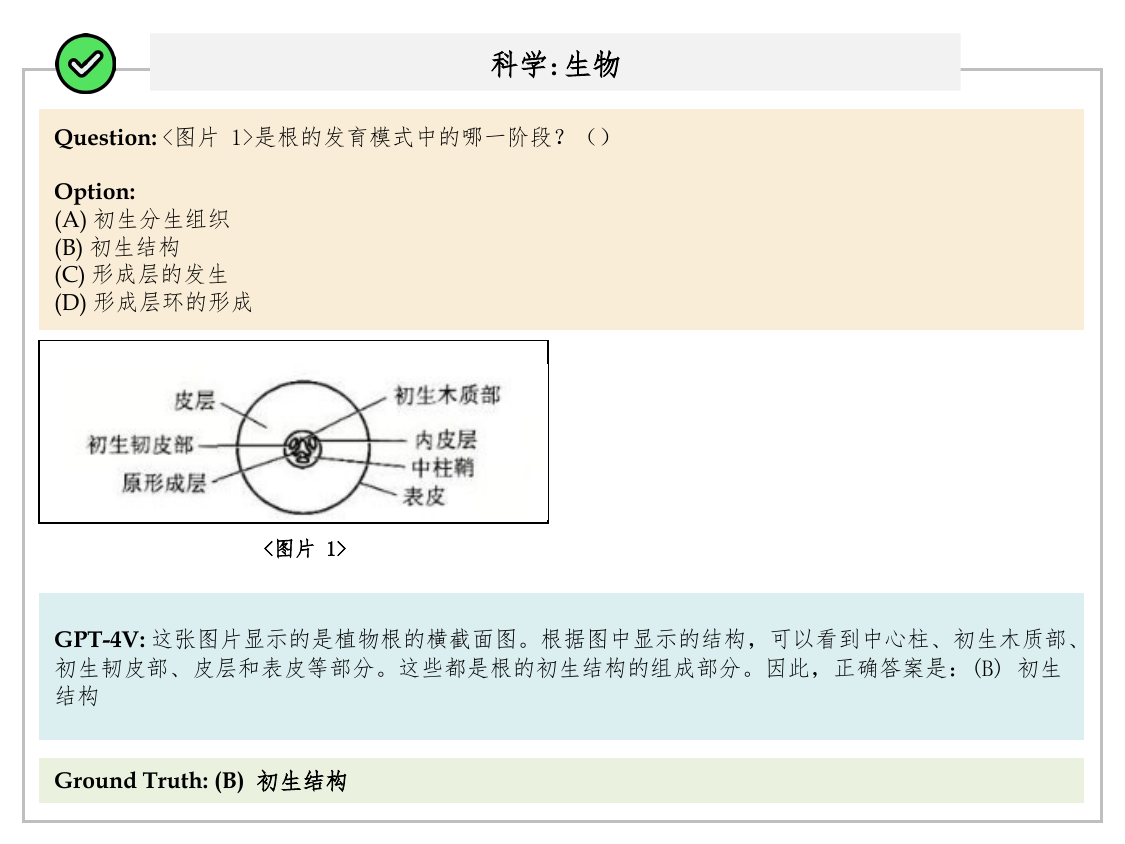}{Biology 3: Correct Case}{A sample correct case of Science (subcategory: Biology).}{fig:error_analysis_case_55}

\casestudyfigure{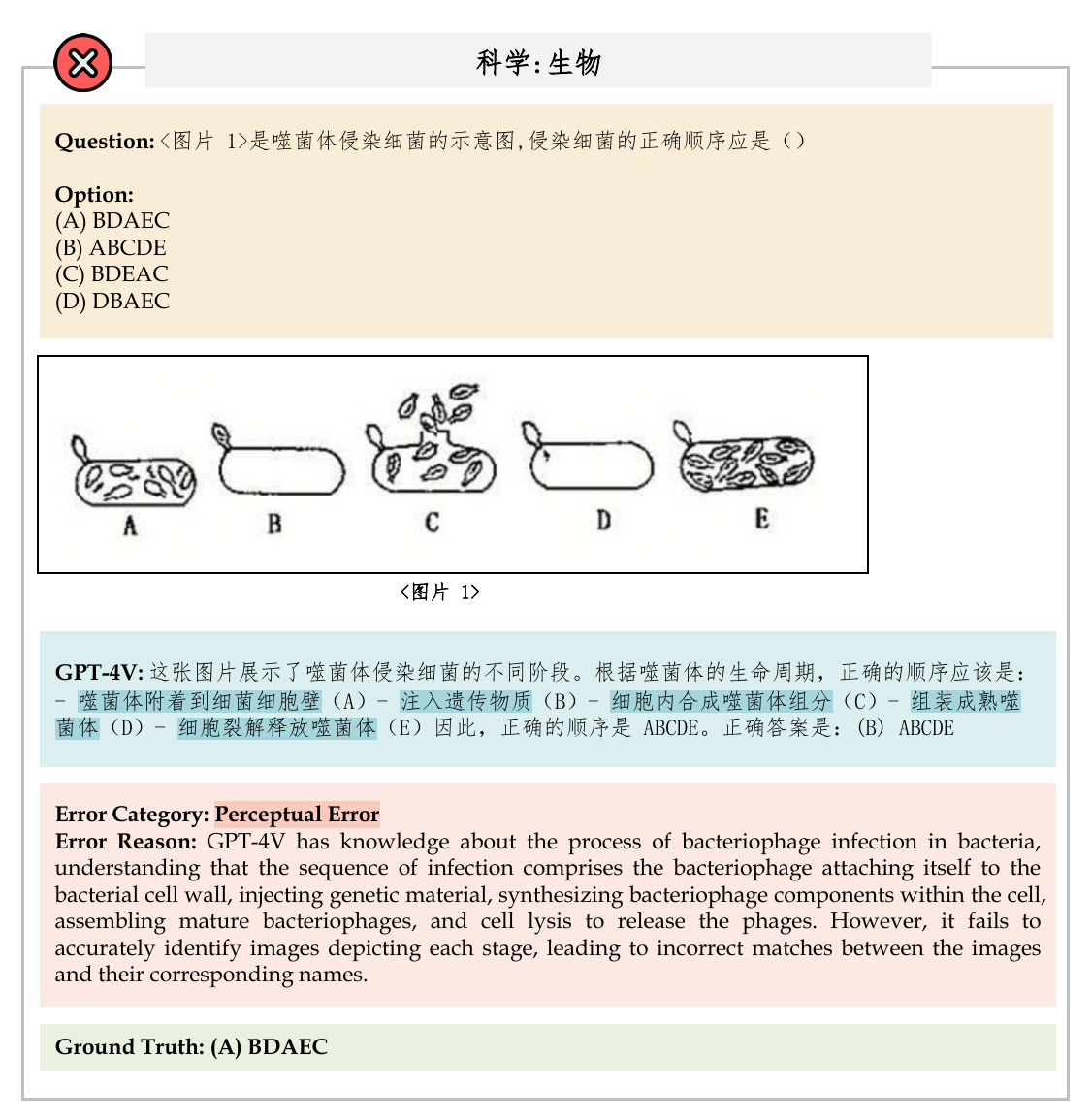}{Biology 4: Perceptual Error}{A sample error case of Science (subcategory: Biology).}{fig:error_analysis_case_198}

\casestudyfigure{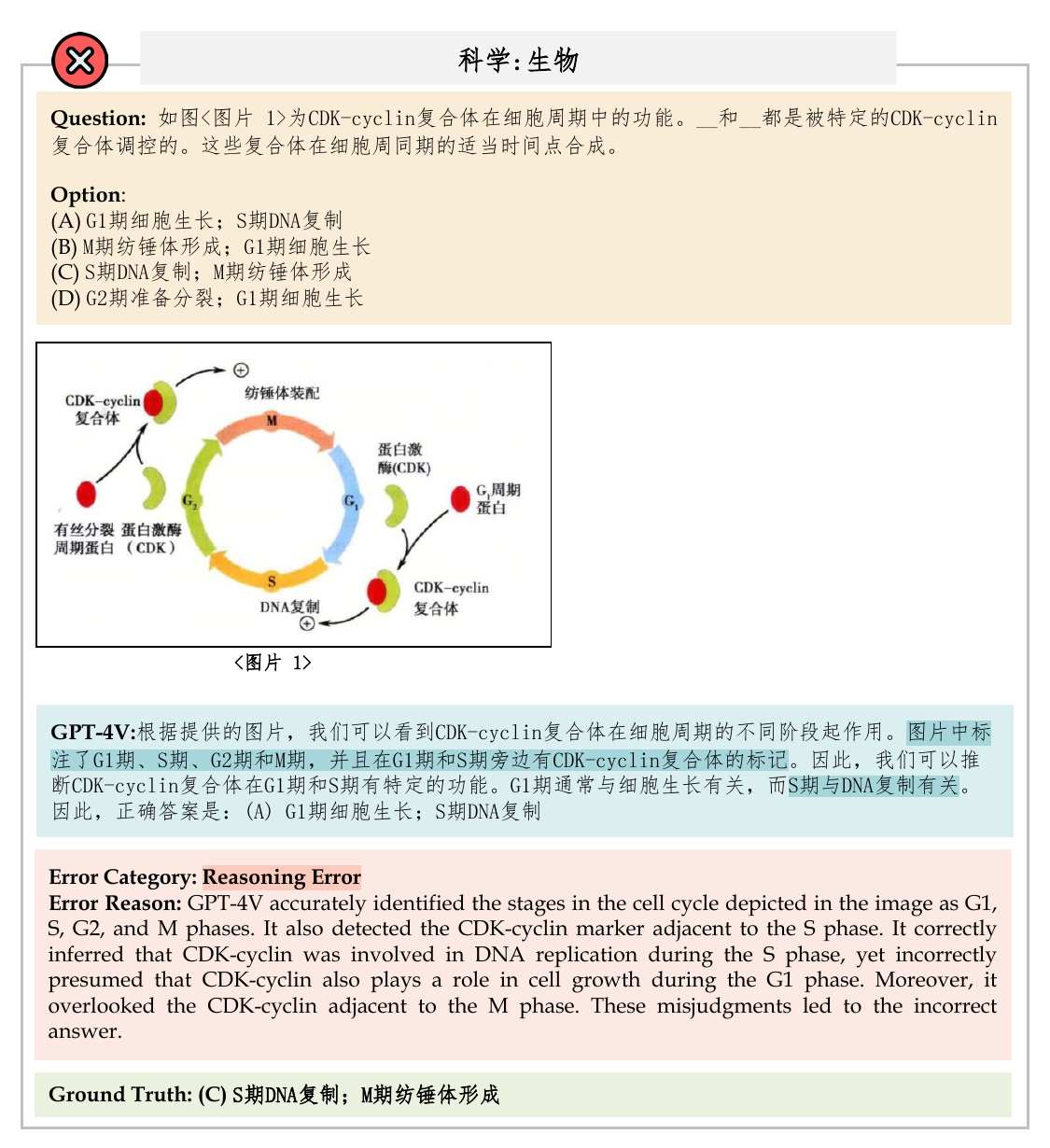}{Biology 5: Reasoning Error}{A sample error case of Science (subcategory: Biology).}{fig:error_analysis_case_196}

\casestudyfigure{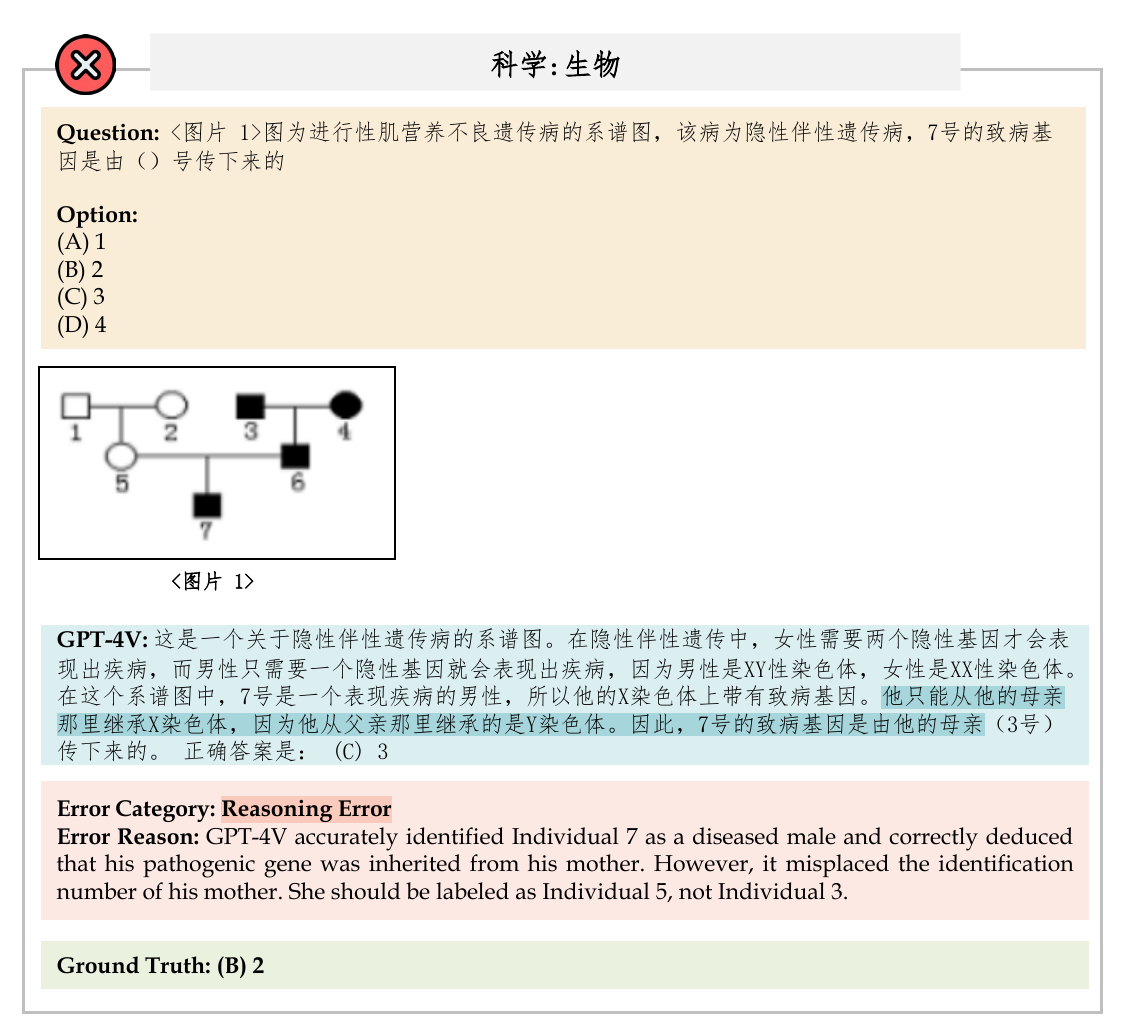}{Biology 6: Reasoning Error}{A sample error case of Science (subcategory: Biology).}{fig:error_analysis_case_199}

\casestudyfigure{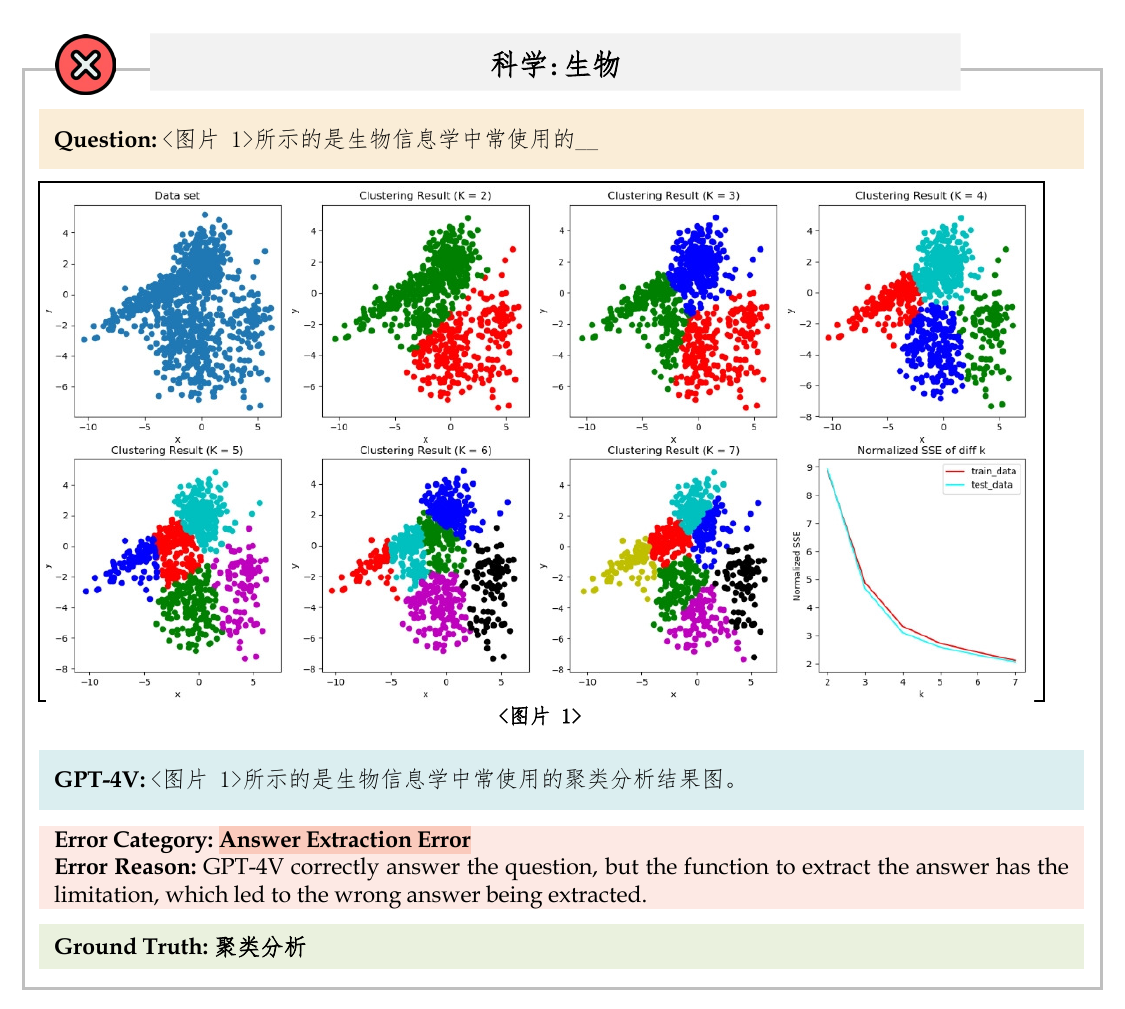}{Biology 7: Answer Extraction Error}{A sample error case of Science (subcategory: Biology).}{fig:error_analysis_case_193}

\casestudyfigure{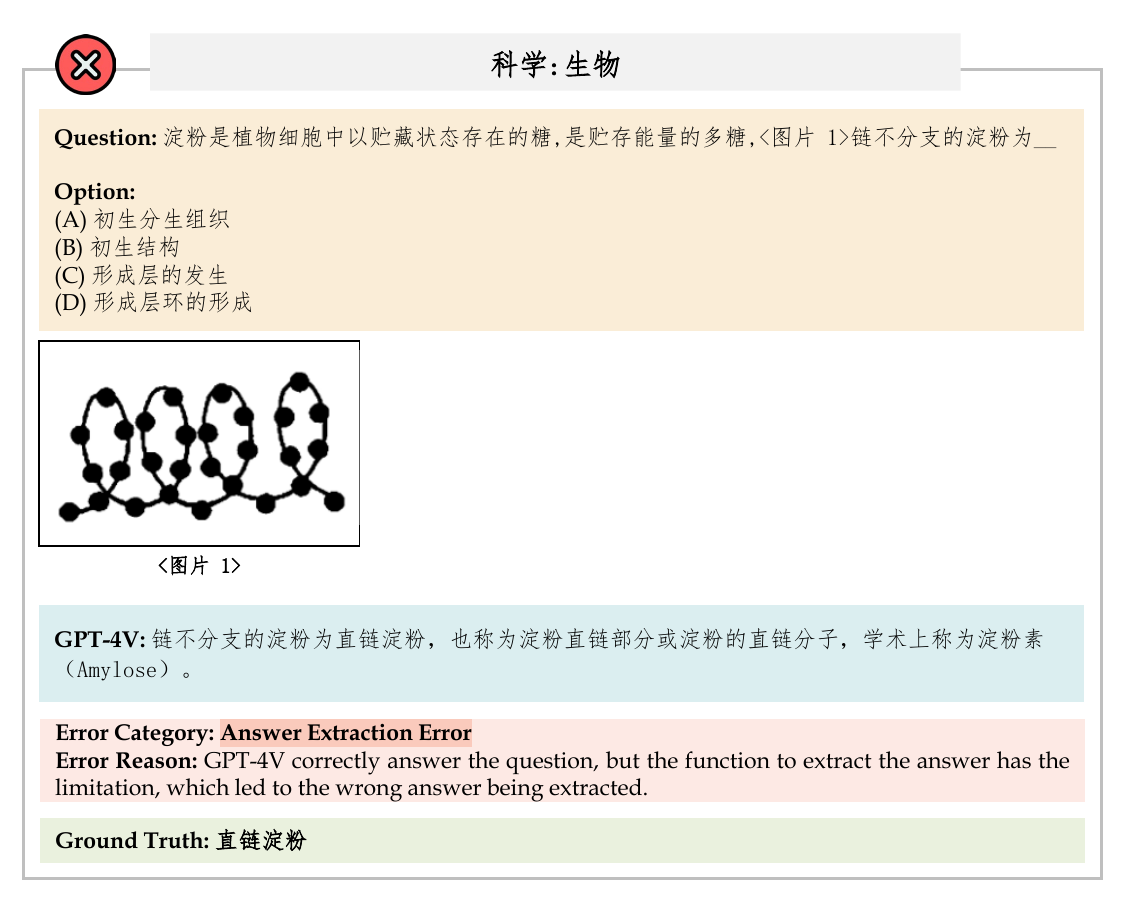}{Biology 8: Answer Extraction Error}{A sample error case of Science (subcategory: Biology).}{fig:error_analysis_case_194}

\casestudyfigure{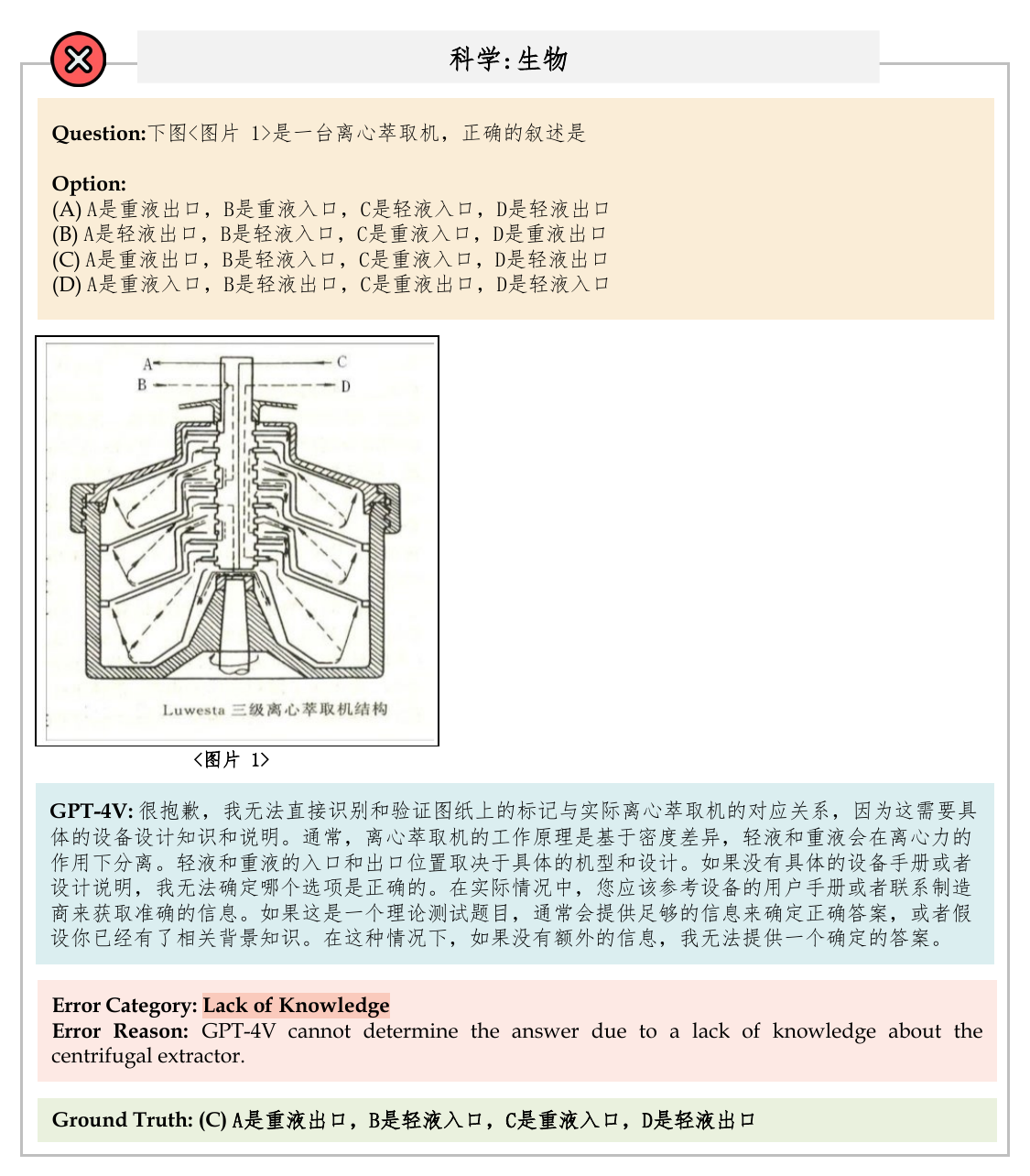}{Biology 9: Lack of Knowledge}{A sample error case of Science (subcategory: Biology).}{fig:error_analysis_case_197}

\casestudyfigure{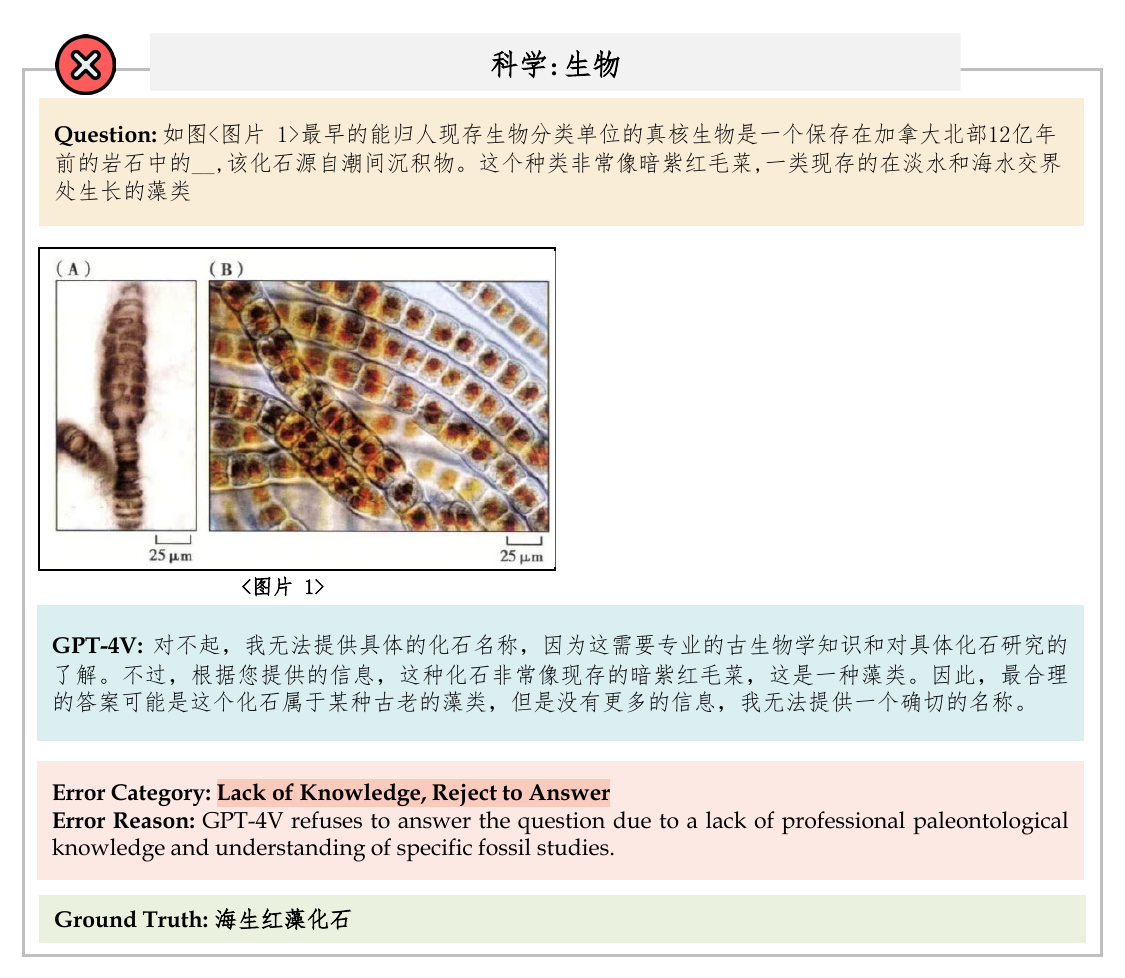}{Biology 10: Lack of Knowledge, Reject to Answer}{A sample error case of Science (subcategory: Biology).}{fig:error_analysis_case_195}

\casestudyfigure{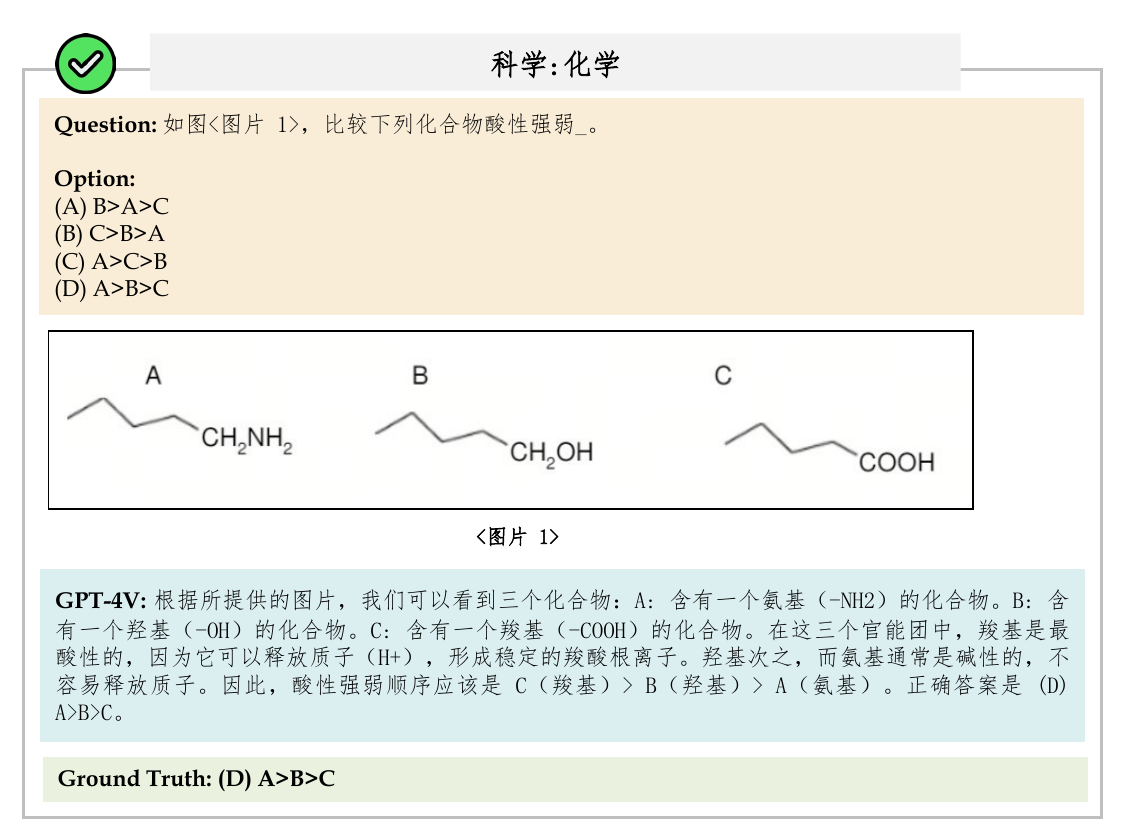}{Chemistry 1: Correct Case}{A sample correct case of Science (subcategory: Chemistry).}{fig:error_analysis_case_59}

\casestudyfigure{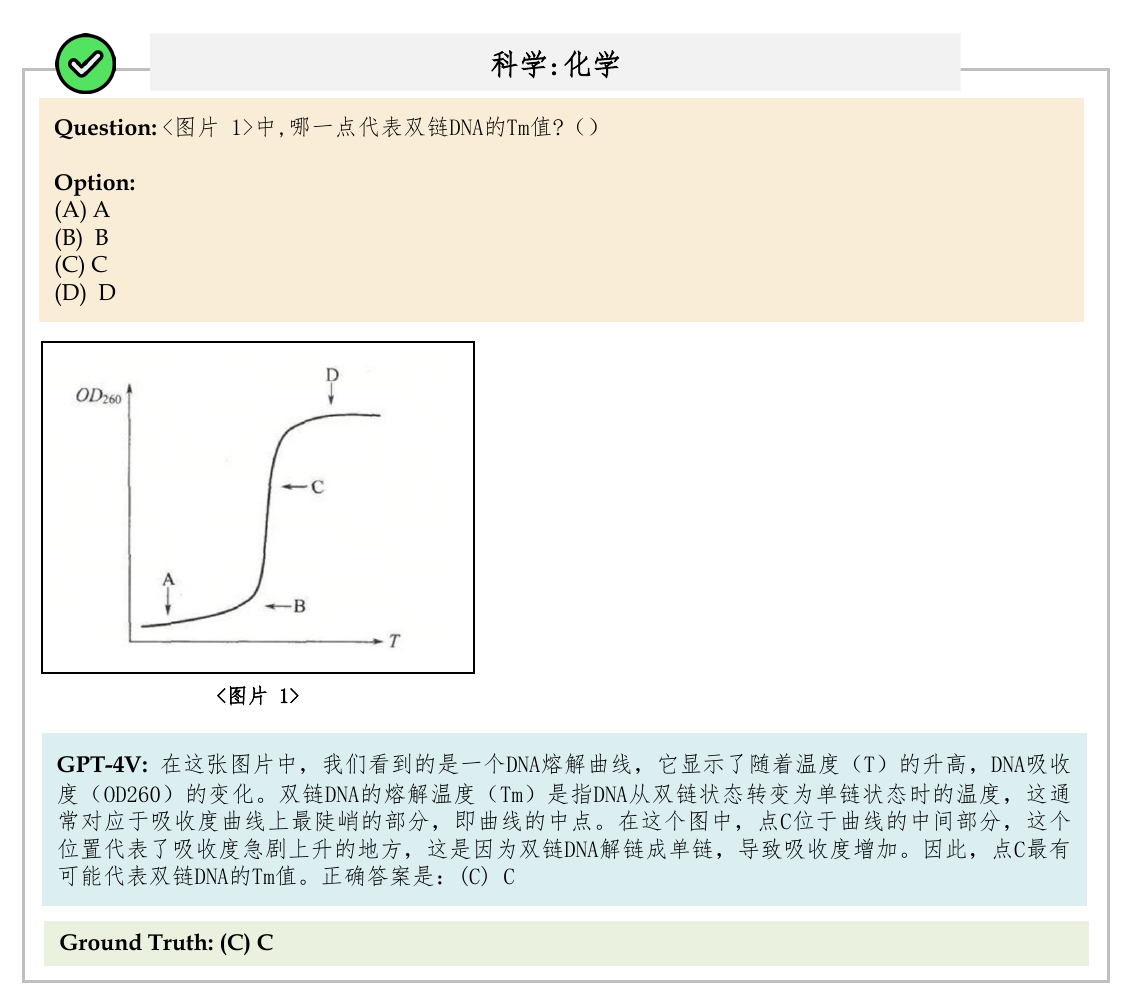}{Chemistry 2: Correct Case}{A sample correct case of Science (subcategory: Chemistry).}{fig:error_analysis_case_60}

\casestudyfigure{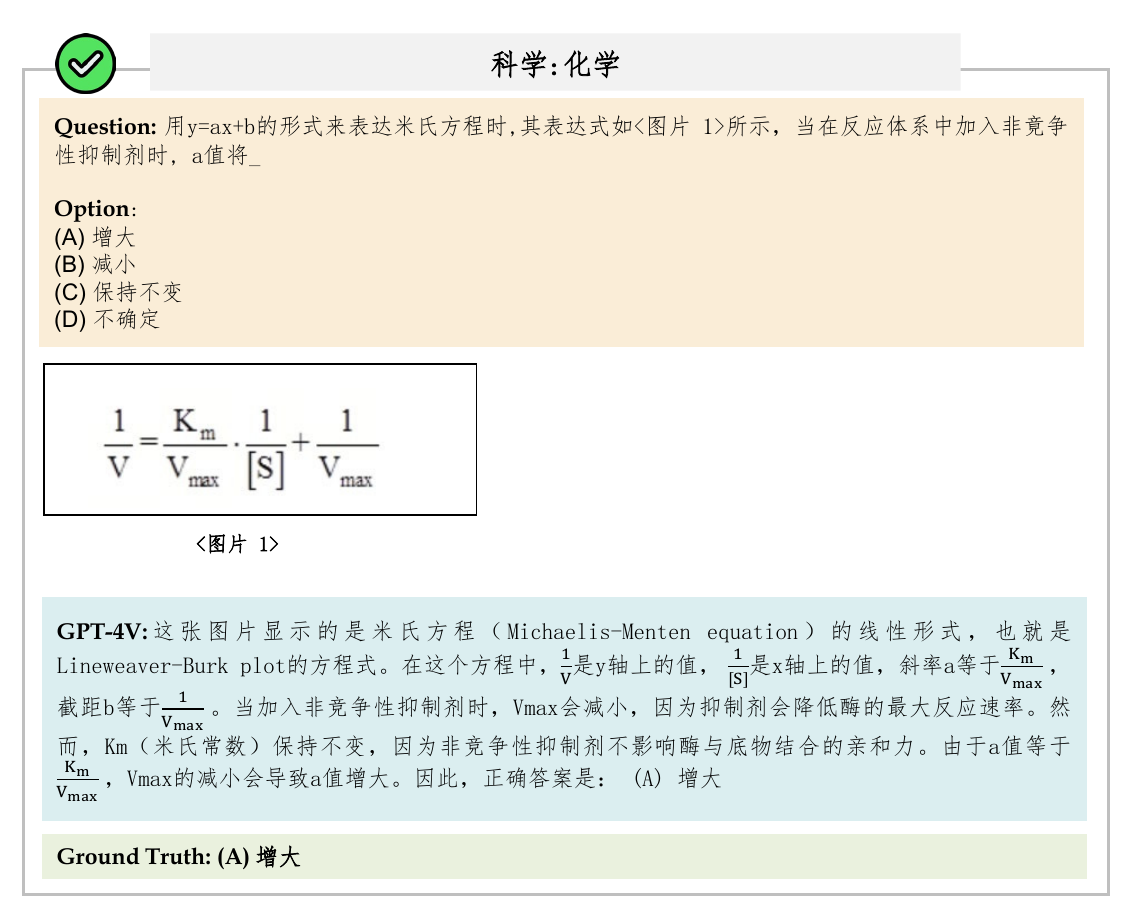}{Chemistry 3: Correct Case}{A sample correct case of Science (subcategory: Chemistry).}{fig:error_analysis_case_61}

\casestudyfigure{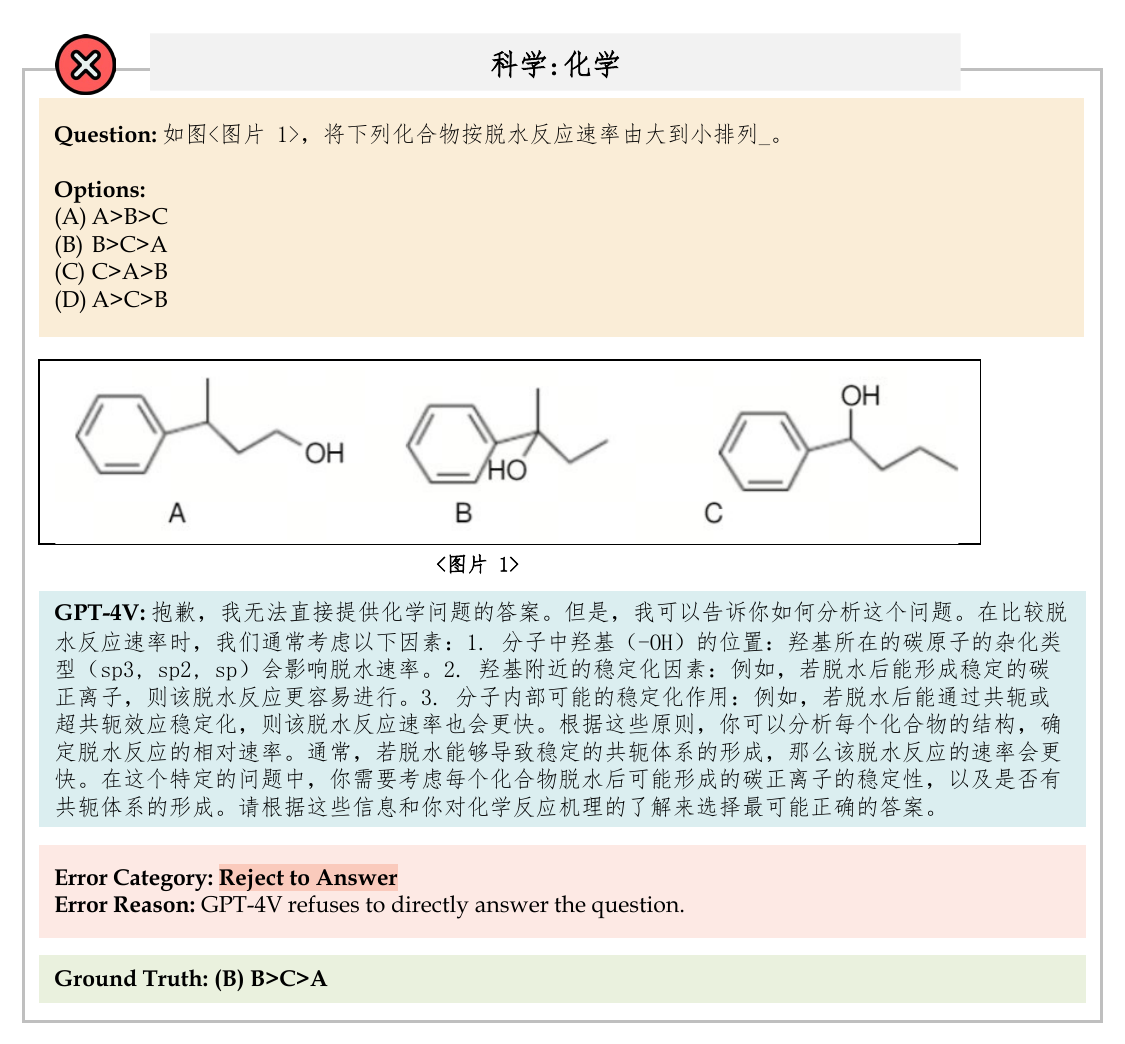}{Chemistry 4: Reject to Answer}{A sample error case of Science (subcategory: Chemistry).}{fig:error_analysis_case_211}

\casestudyfigure{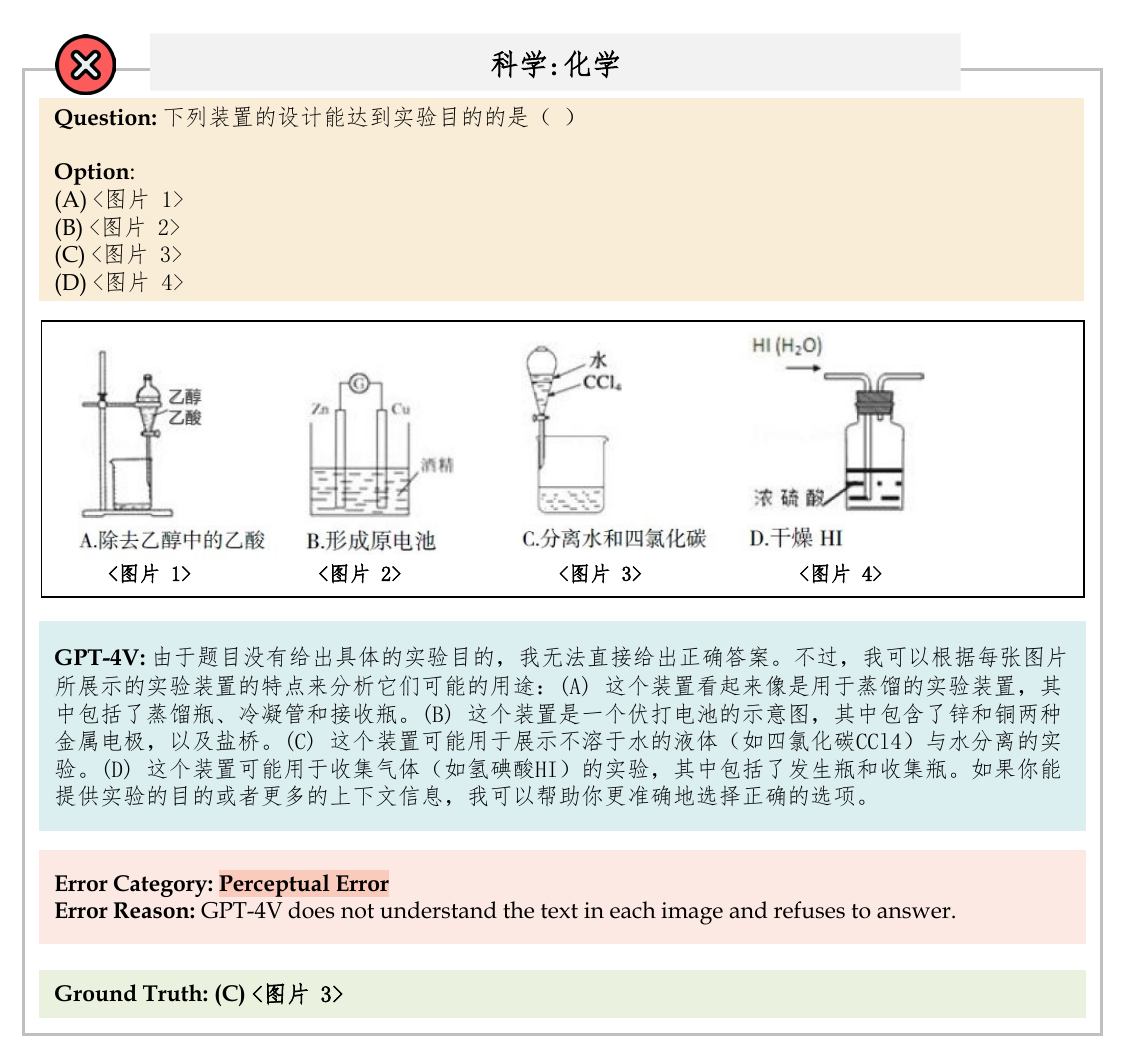}{Chemistry 5: Perceptual Error}{A sample error case of Science (subcategory: Chemistry).}{fig:error_analysis_case_209}

\casestudyfigure{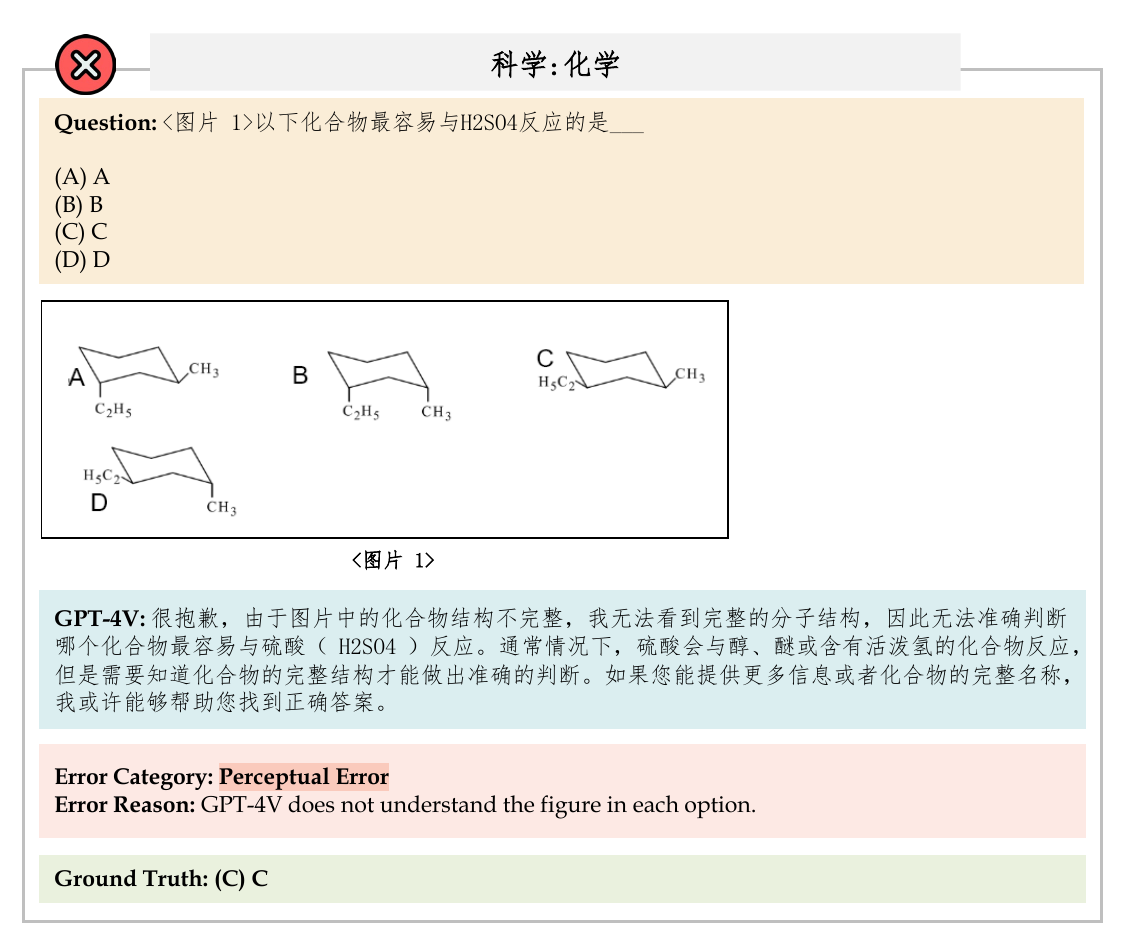}{Chemistry 6: Perceptual Error}{A sample error case of Science (subcategory: Chemistry).}{fig:error_analysis_case_210}

\casestudyfigure{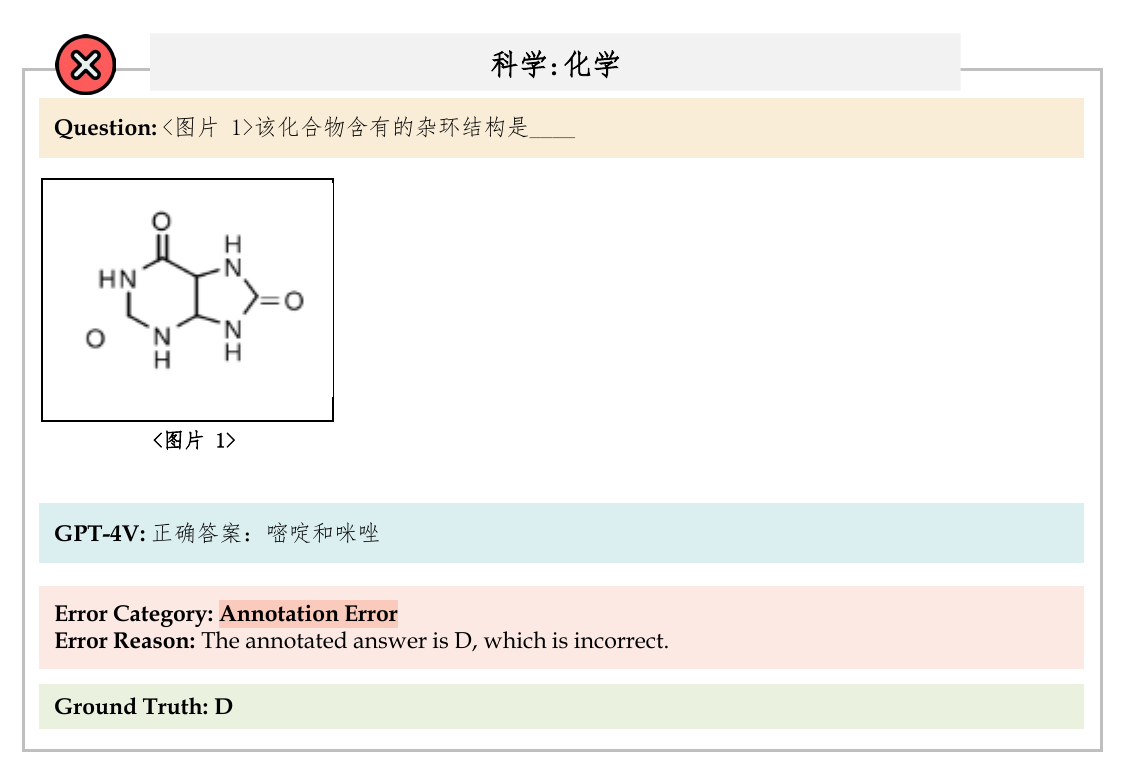}{Chemistry 7: Annotation Error}{A sample error case of Science (subcategory: Chemistry).}{fig:error_analysis_case_207}

\casestudyfigure{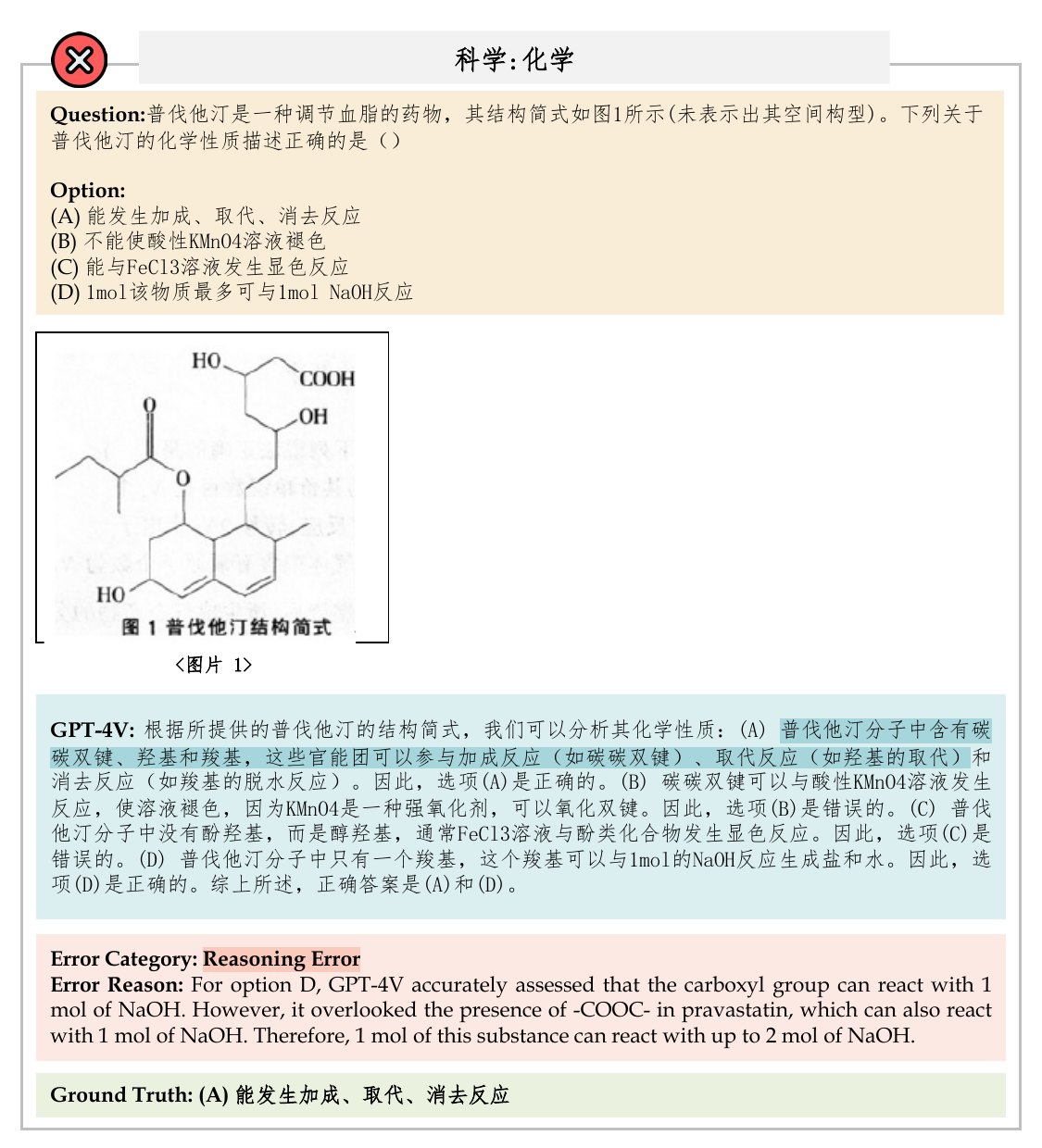}{Chemistry 8: Reasoning Error}{A sample error case of Science (subcategory: Chemistry).}{fig:error_analysis_case_208}

\casestudyfigure{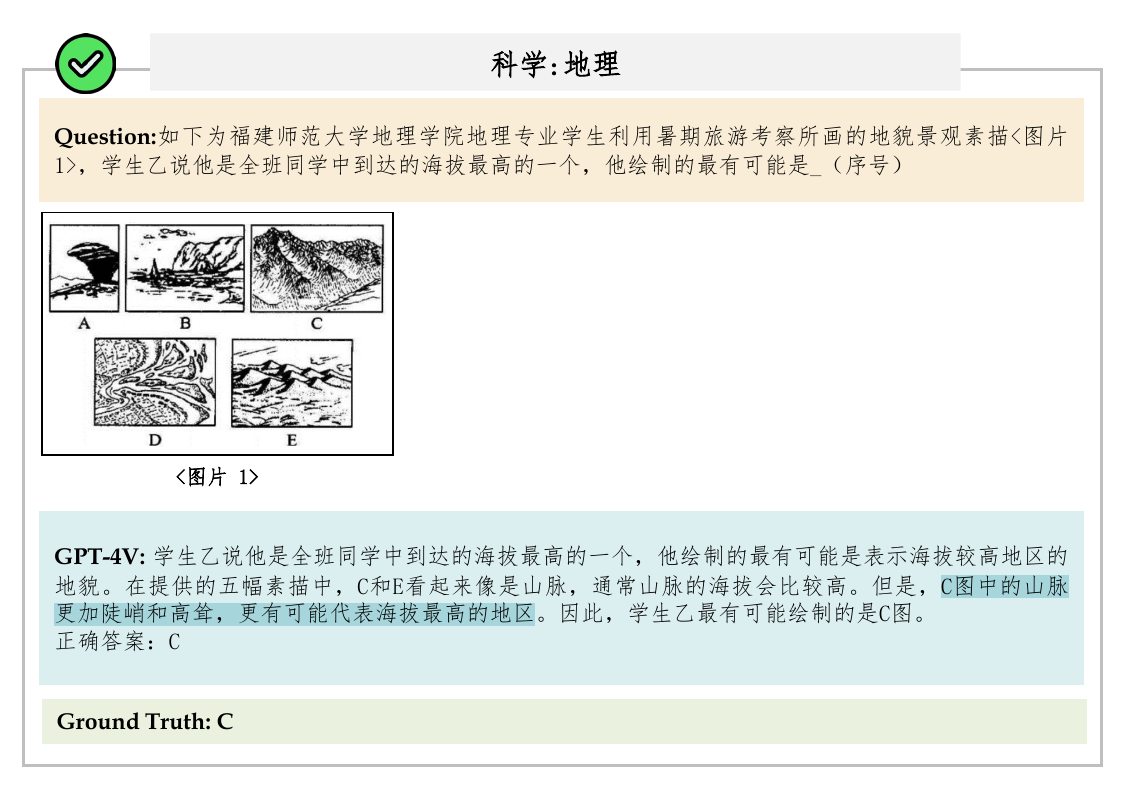}{Geography 1: Correct Case}{A sample correct case of Science (subcategory: Geography).}{fig:error_analysis_case_56}

\casestudyfigure{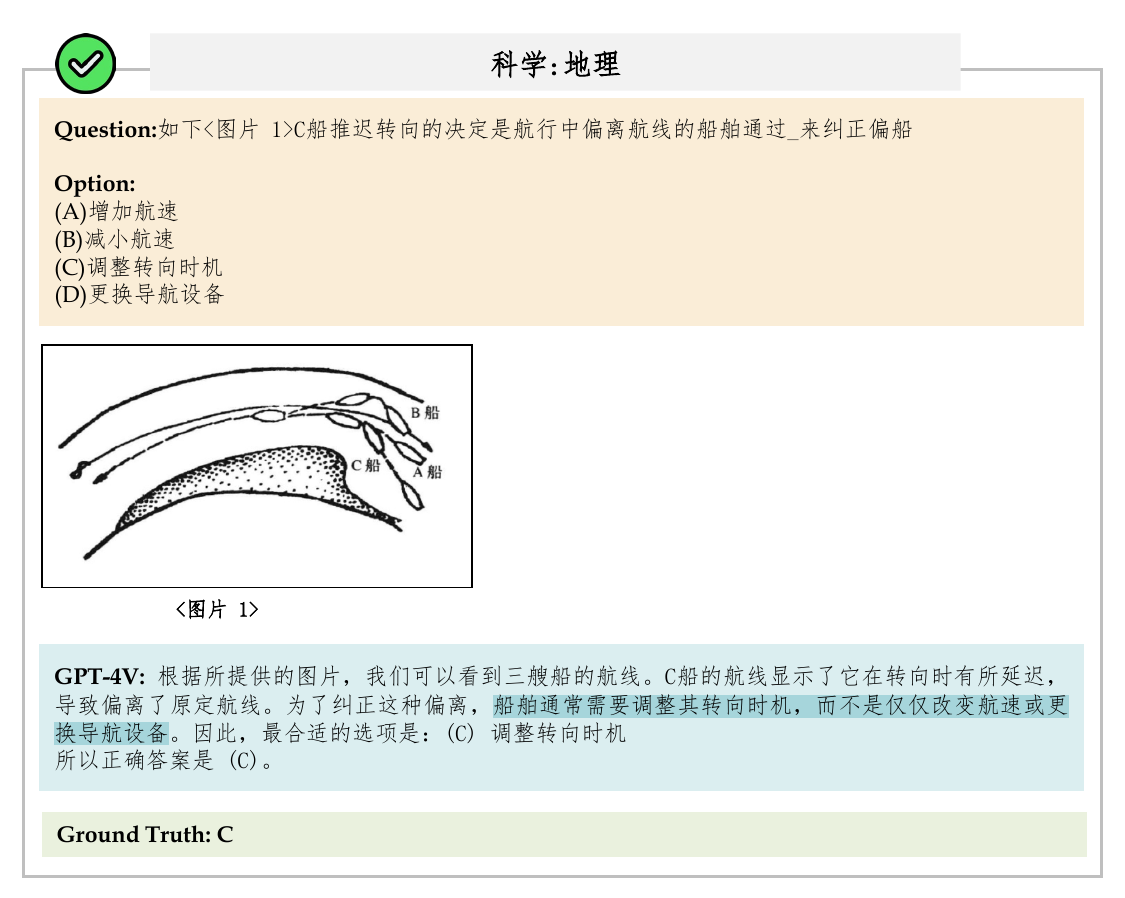}{Geography 2: Correct Case}{A sample correct case of Science (subcategory: Geography).}{fig:error_analysis_case_57}

\casestudyfigure{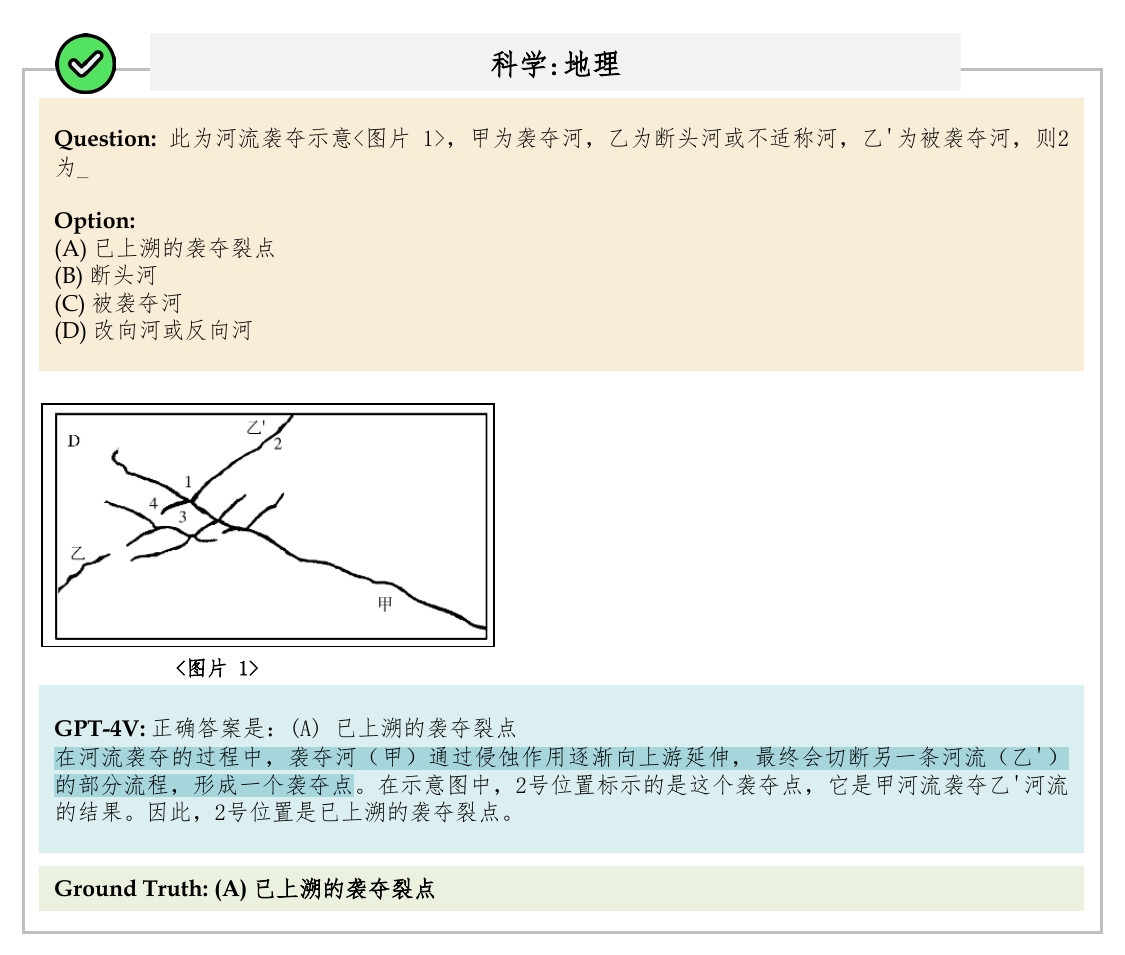}{Geography 3: Correct Case}{A sample correct case of Science (subcategory: Geography).}{fig:error_analysis_case_58}

\casestudyfigure{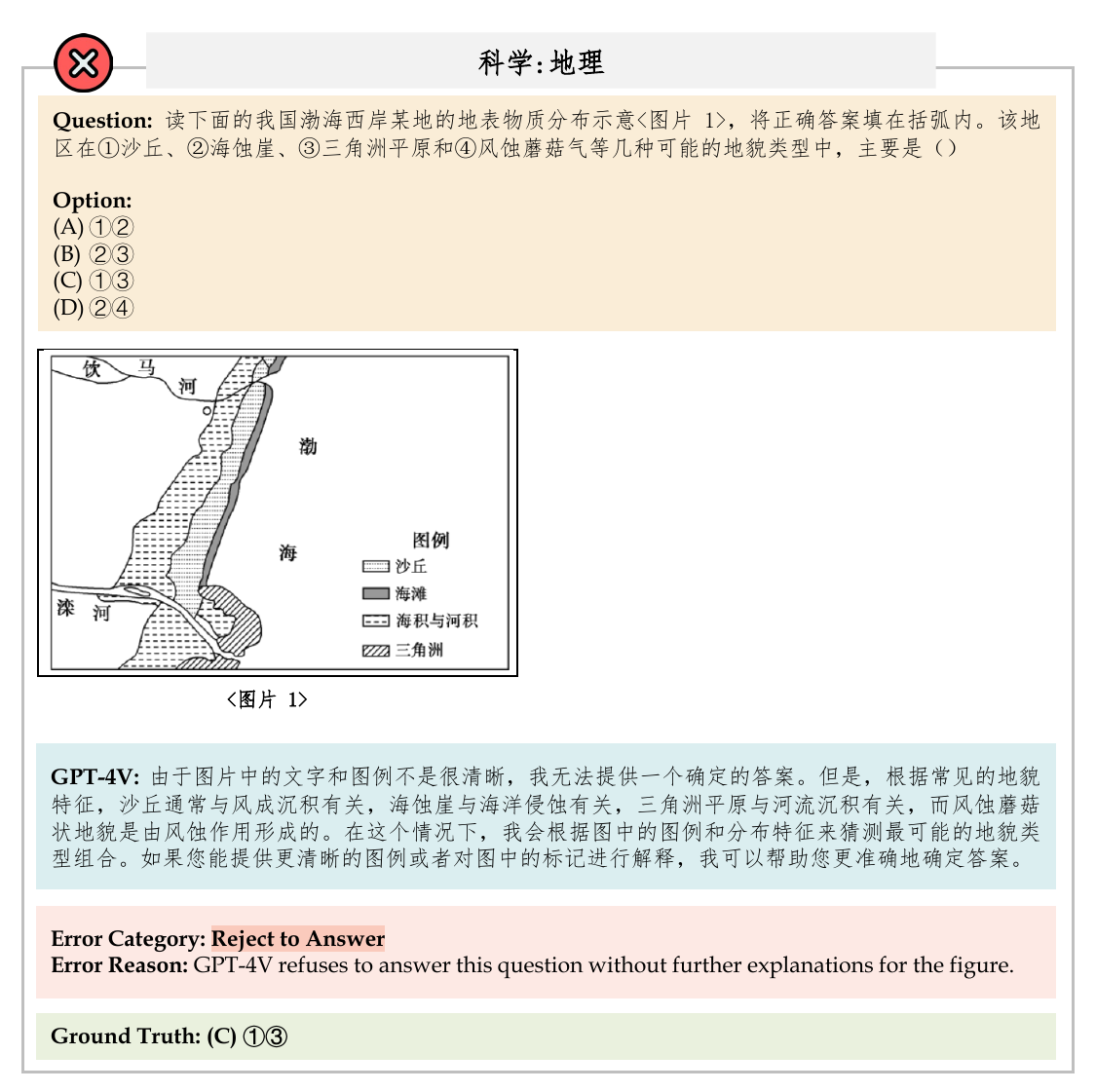}{Geography 4: Reject to Answer}{A sample error case of Science (subcategory: Geography).}{fig:error_analysis_case_201}

\casestudyfigure{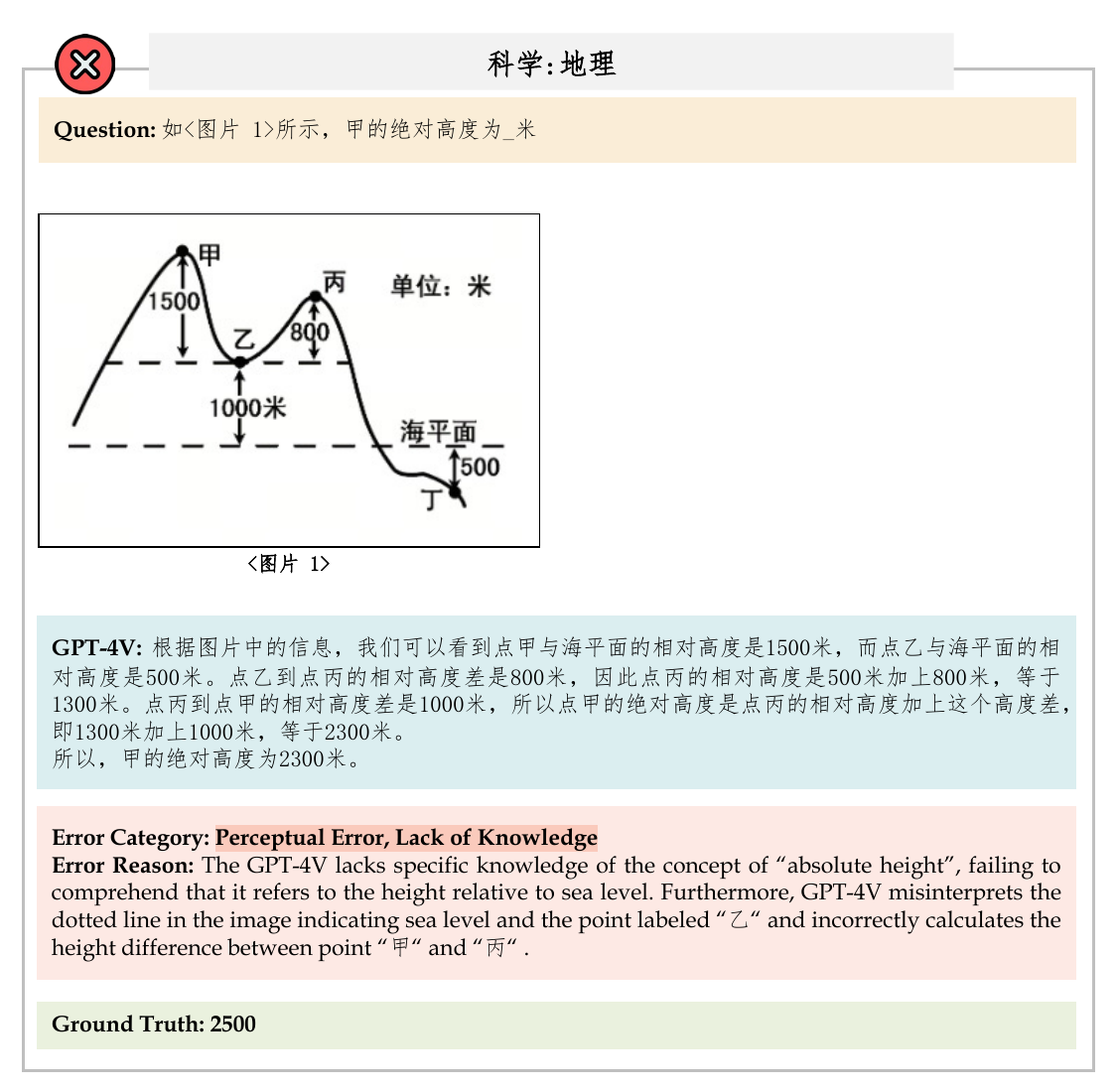}{Geography 5: Perceptual Error, Lack of Knowledge}{A sample error case of Science (subcategory: Geography).}{fig:error_analysis_case_204}

\casestudyfigure{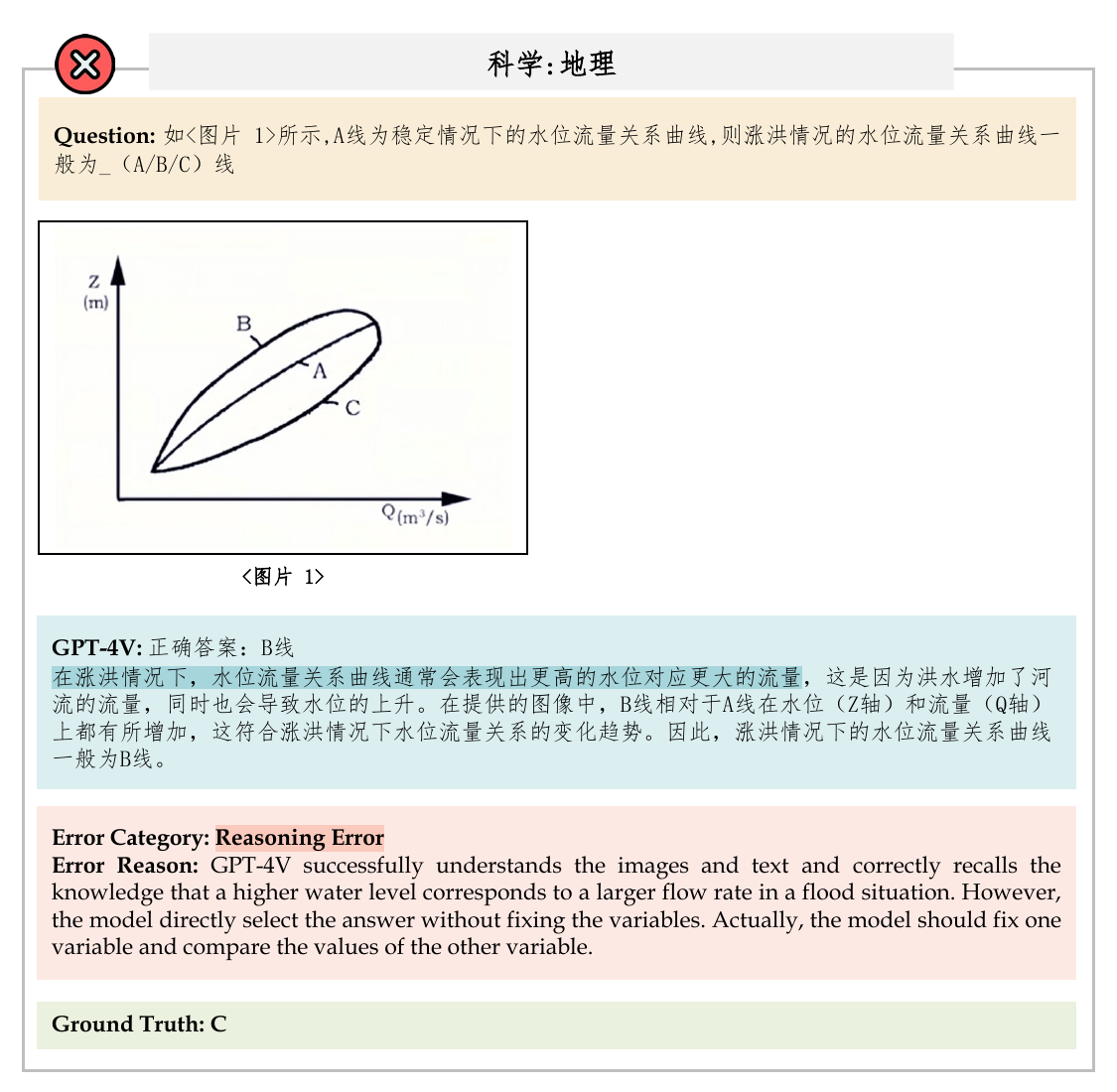}{Geography 6: Reasoning Error}{A sample error case of Science (subcategory: Geography).}{fig:error_analysis_case_202}

\casestudyfigure{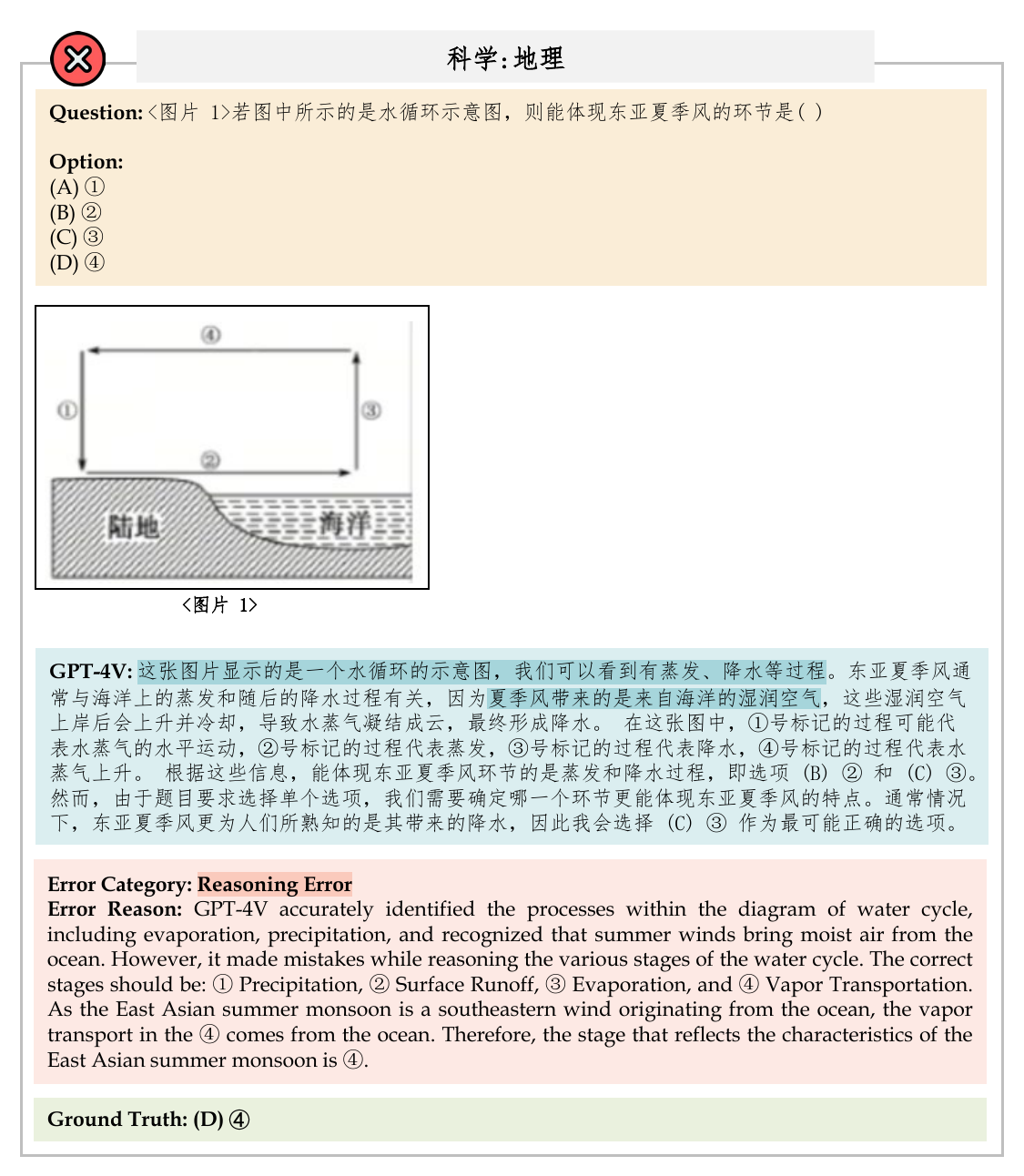}{Geography 7: Reasoning Error}{A sample error case of Science (subcategory: Geography).}{fig:error_analysis_case_203}

\casestudyfigure{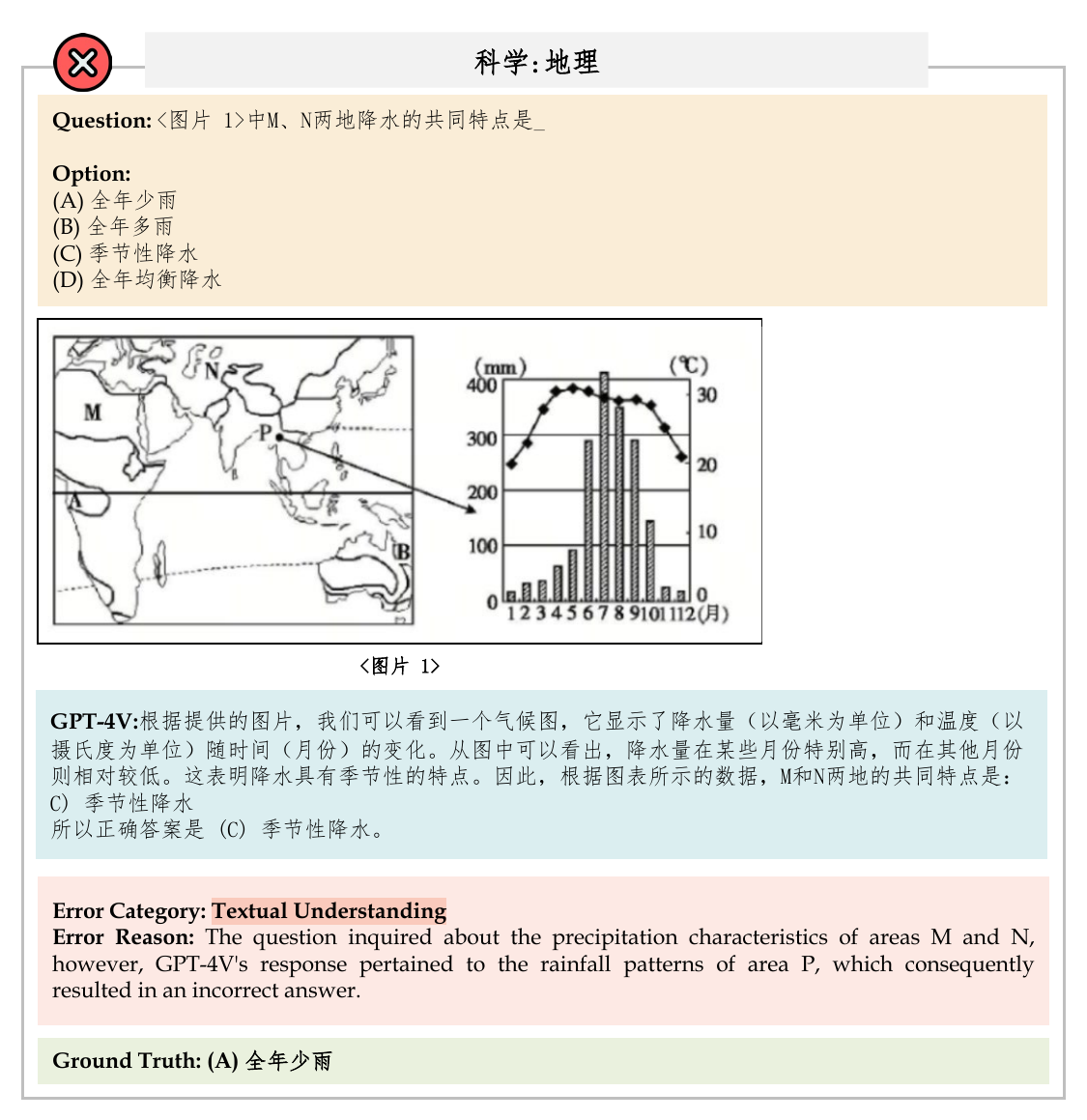}{Geography 8: Textual Understanding}{A sample error case of Science (subcategory: Geography).}{fig:error_analysis_case_205}

\casestudyfigure{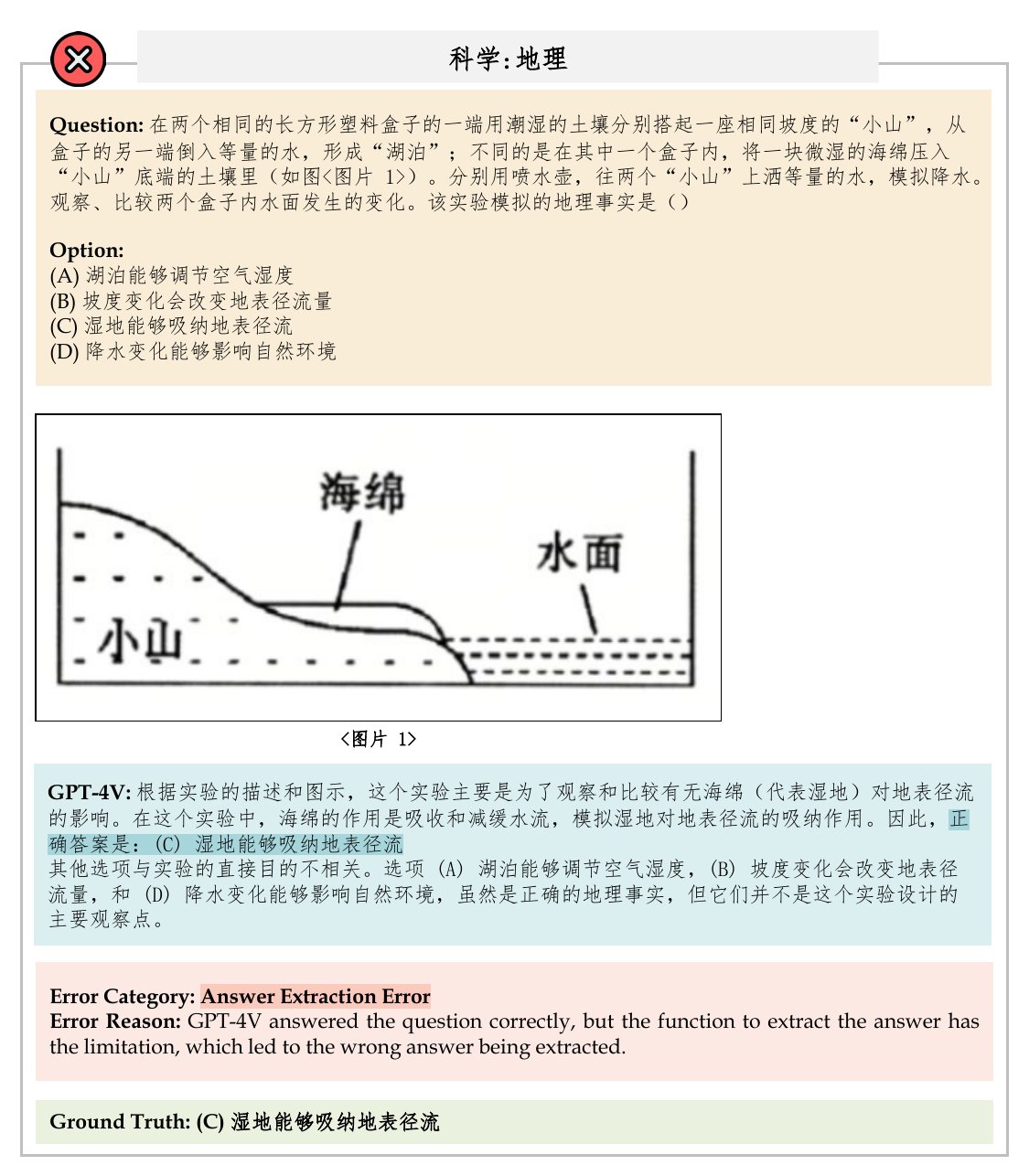}{Geography 9: Answer Extraction Error}{A sample error case of Science (subcategory: Geography).}{fig:error_analysis_case_206}

\casestudyfigure{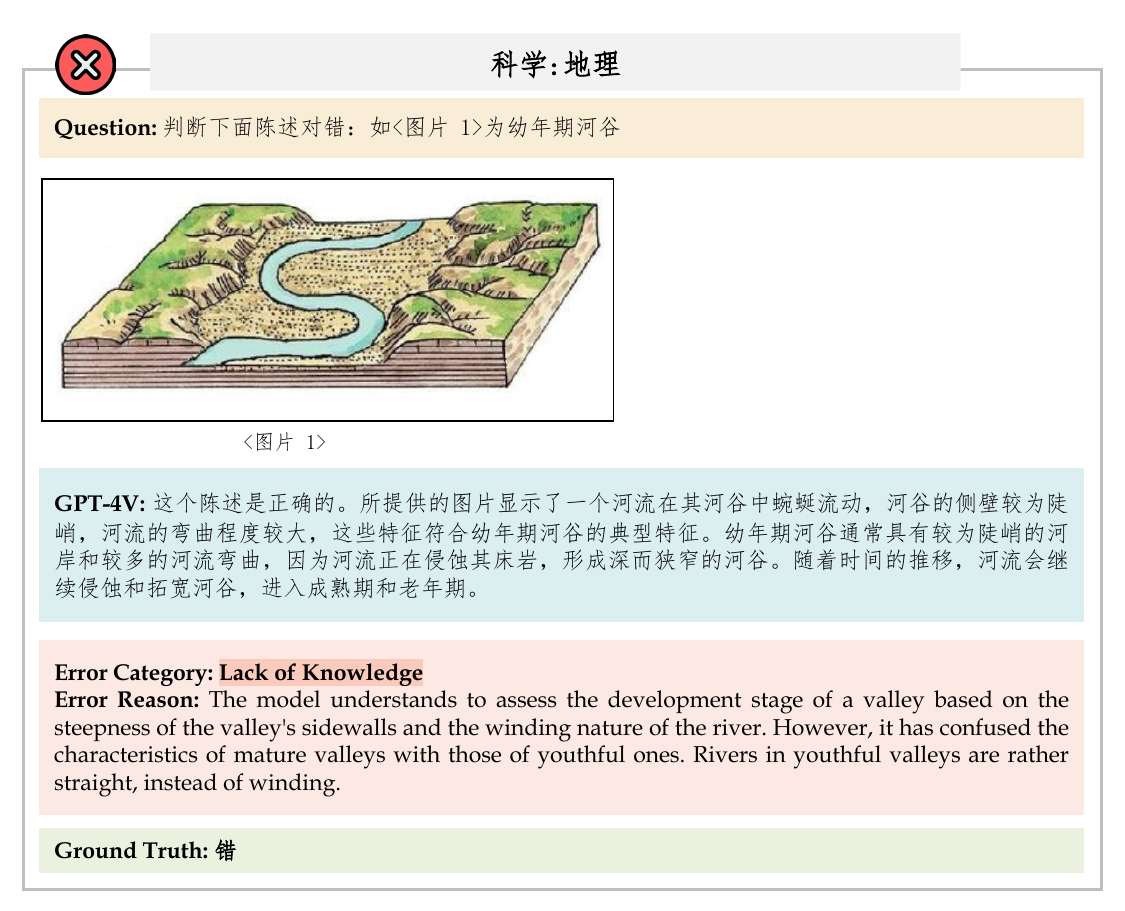}{Geography 10: Lack of Knowledge}{A sample error case of Science (subcategory: Geography).}{fig:error_analysis_case_200}

\casestudyfigure{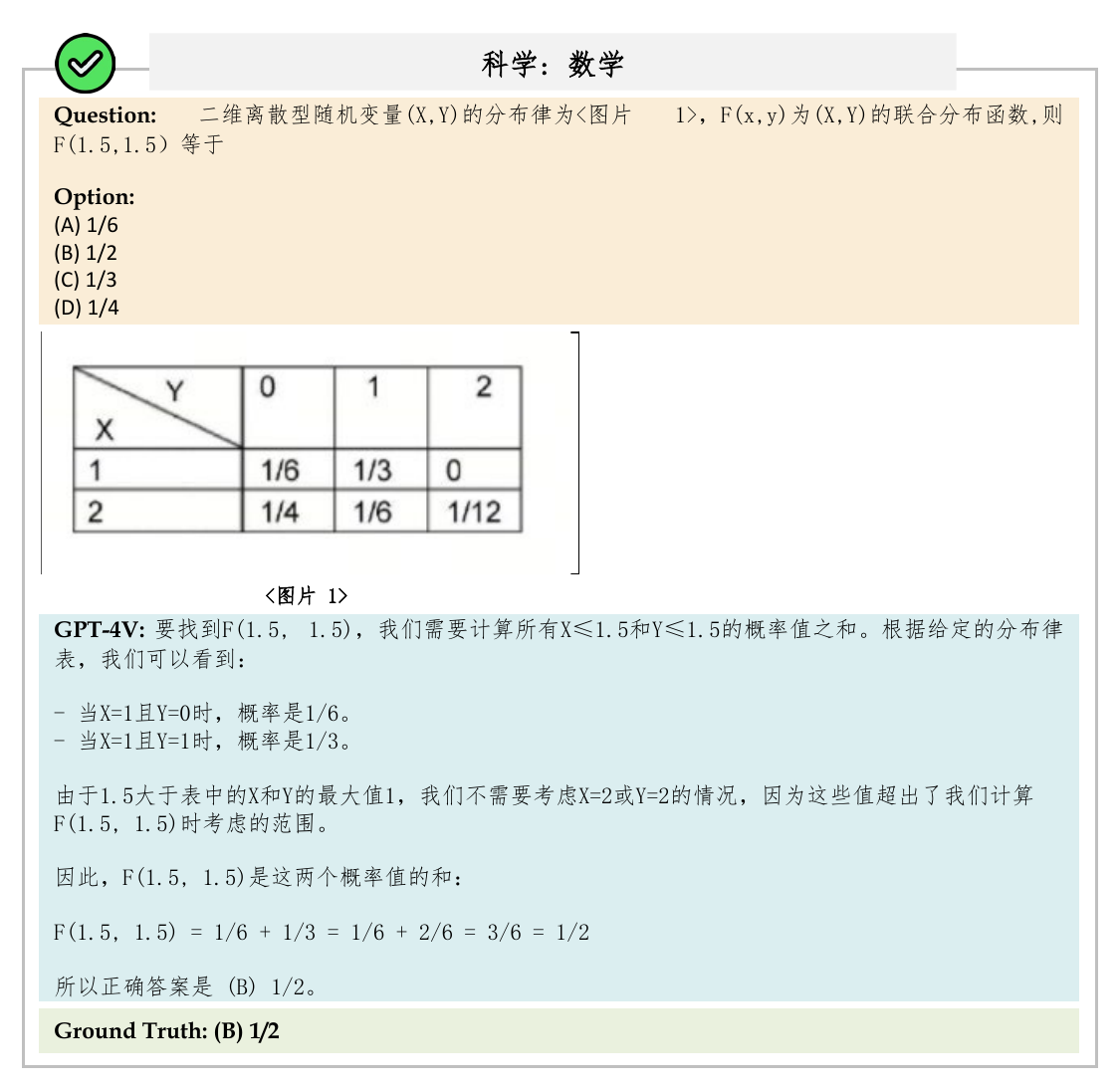}{Mathematics 1: Correct Case}{A sample correct case of Science (subcategory: Mathematics).}{fig:error_analysis_case_46}

\casestudyfigure{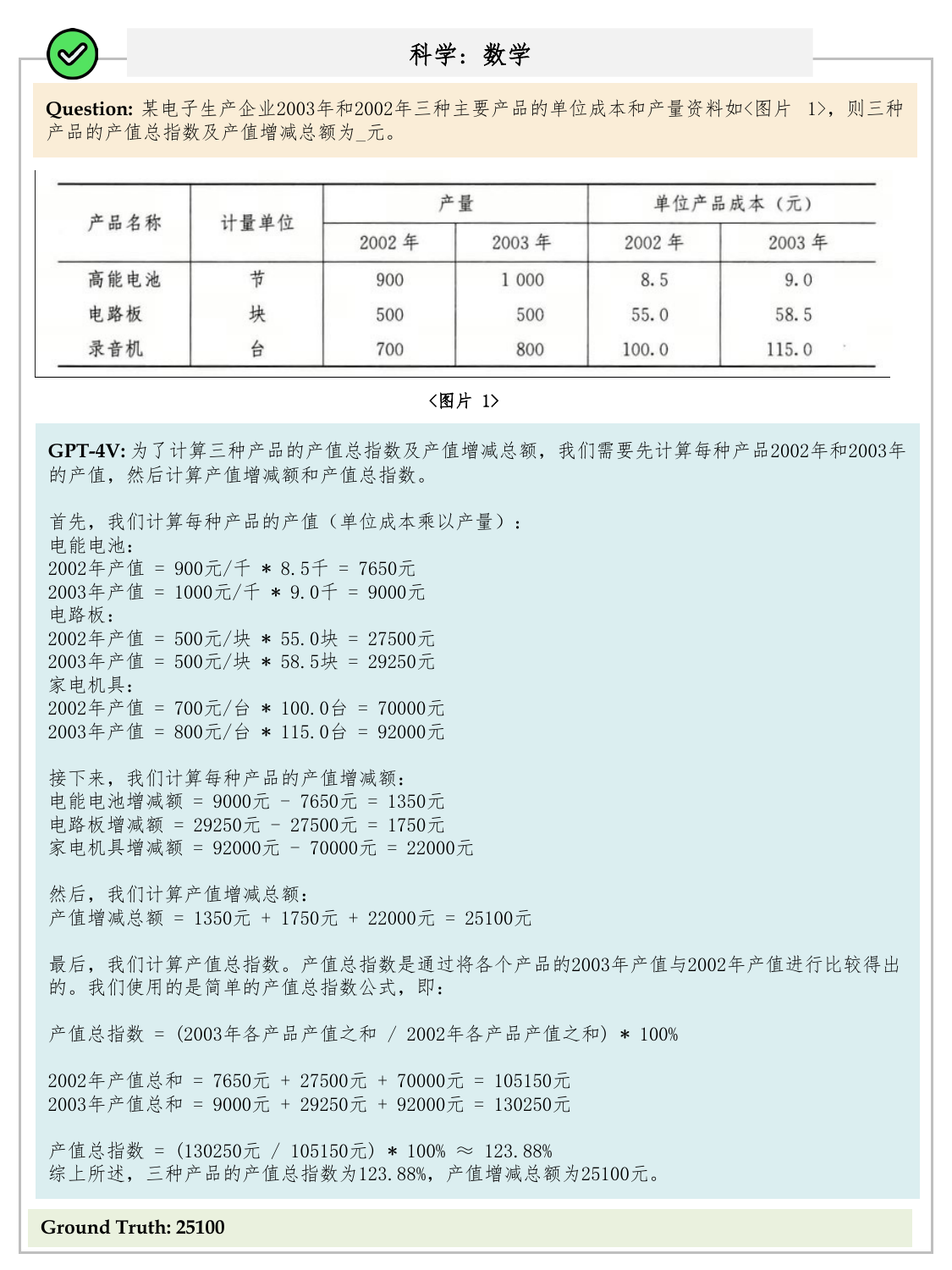}{Mathematics 2: Correct Case}{A sample correct case of Science (subcategory: Mathematics).}{fig:error_analysis_case_47}

\casestudyfigure{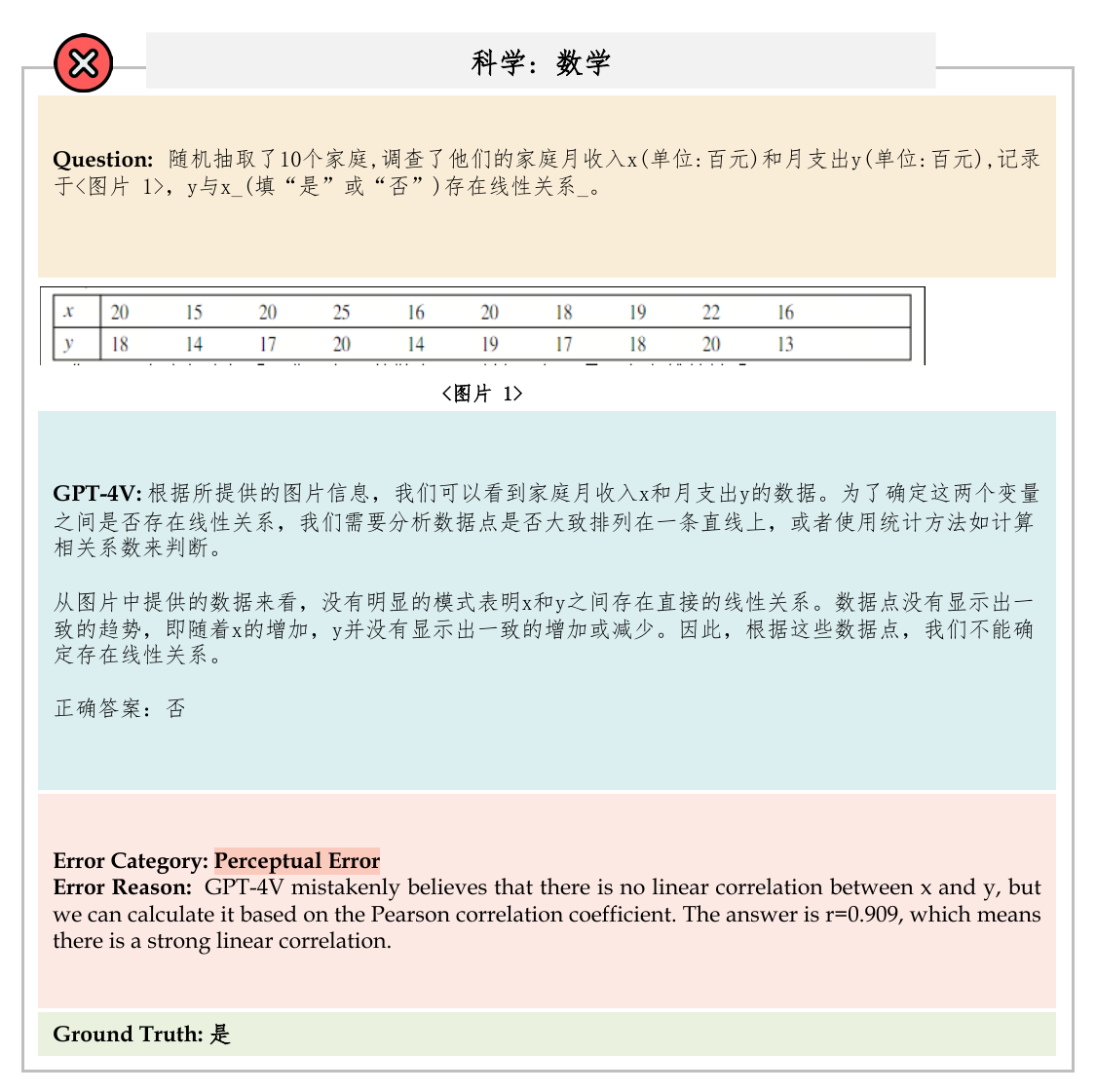}{Mathematics 3: Perceptual Error}{A sample error case of Science (subcategory: Mathematics).}{fig:error_analysis_case_186}

\casestudyfigure{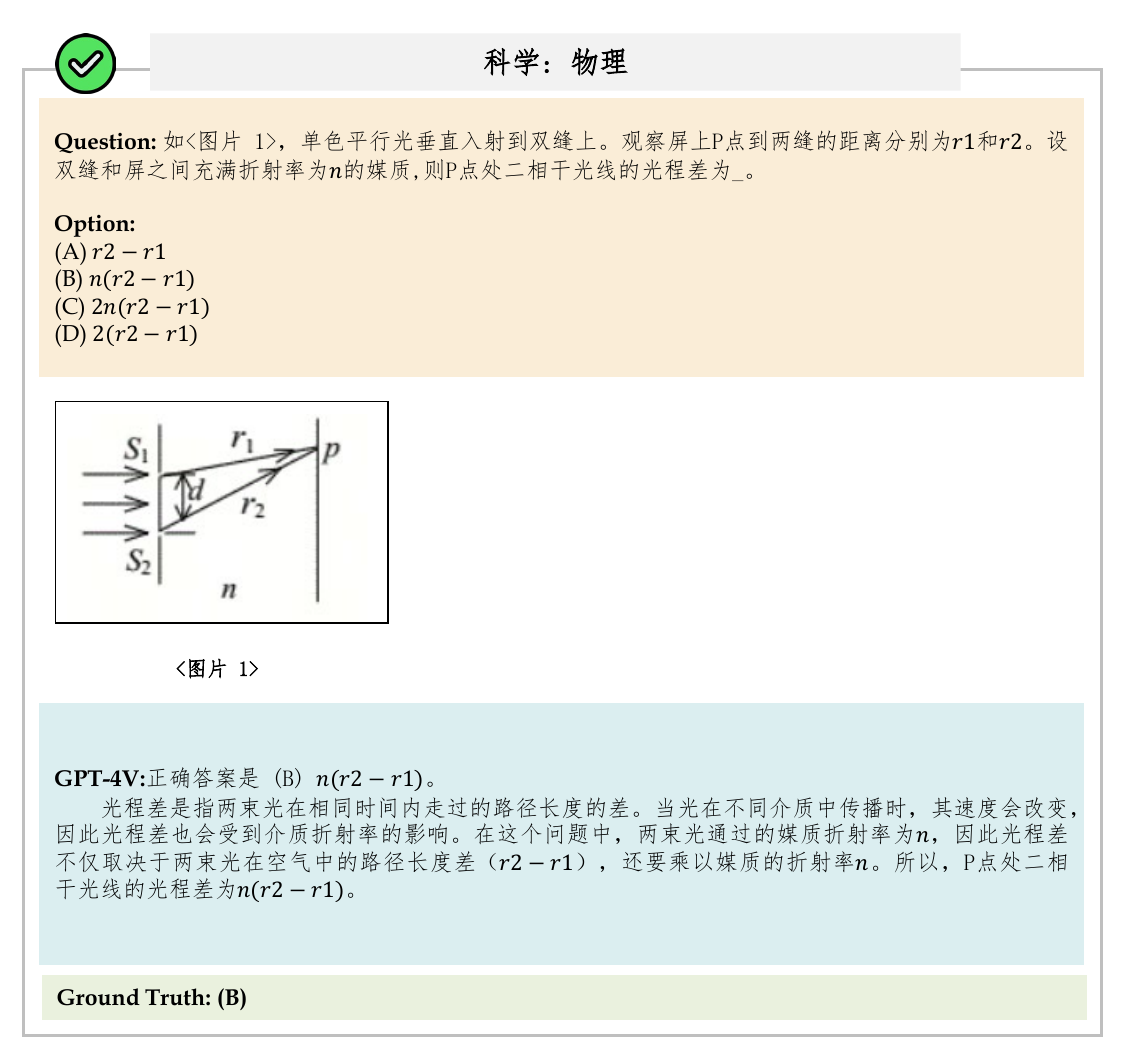}{Physics 1: Correct Case}{A sample correct case of Science (subcategory: Physics).}{fig:error_analysis_case_37}

\casestudyfigure{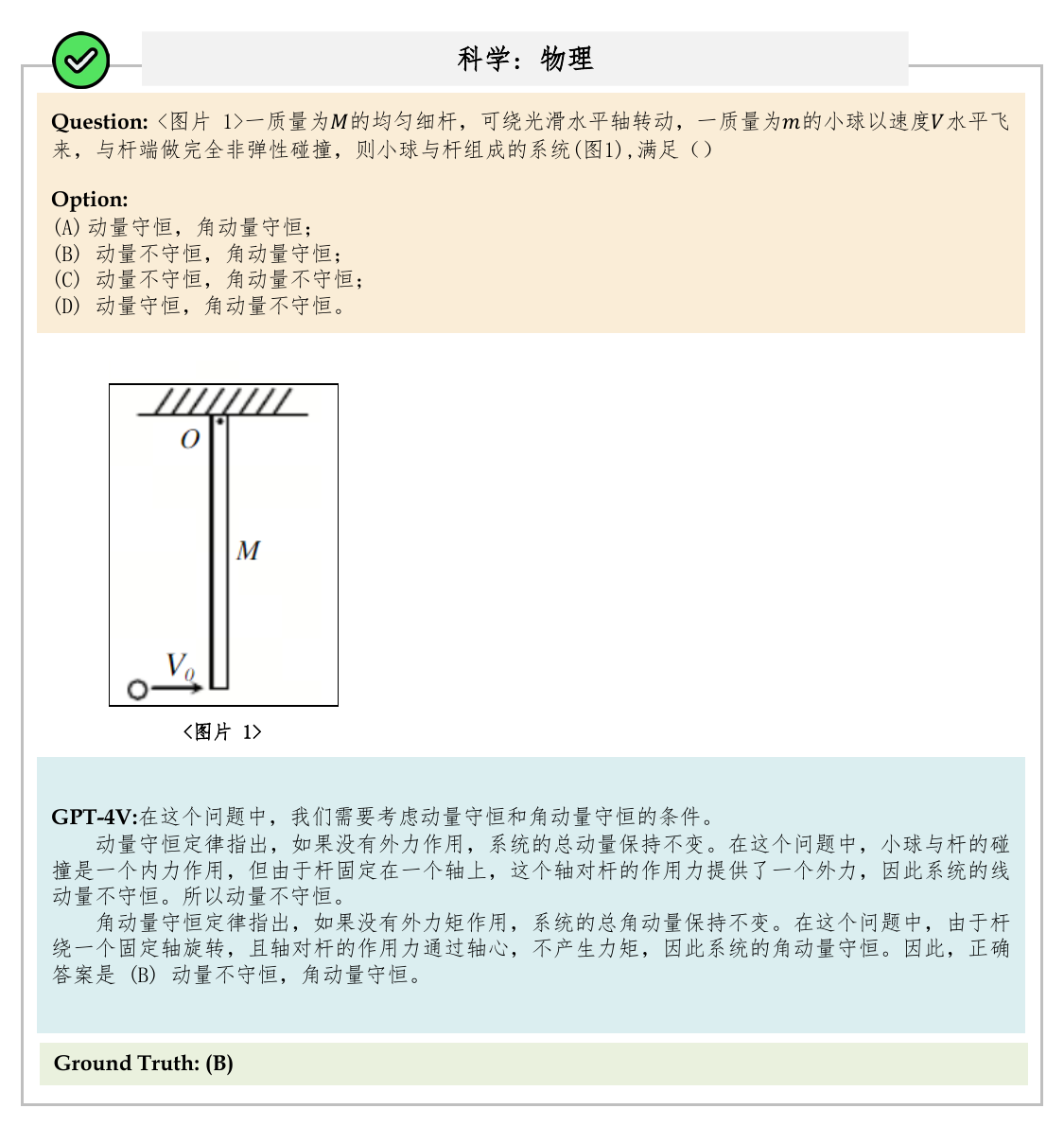}{Physics 2: Correct Case}{A sample correct case of Science (subcategory: Physics).}{fig:error_analysis_case_38}

\casestudyfigure{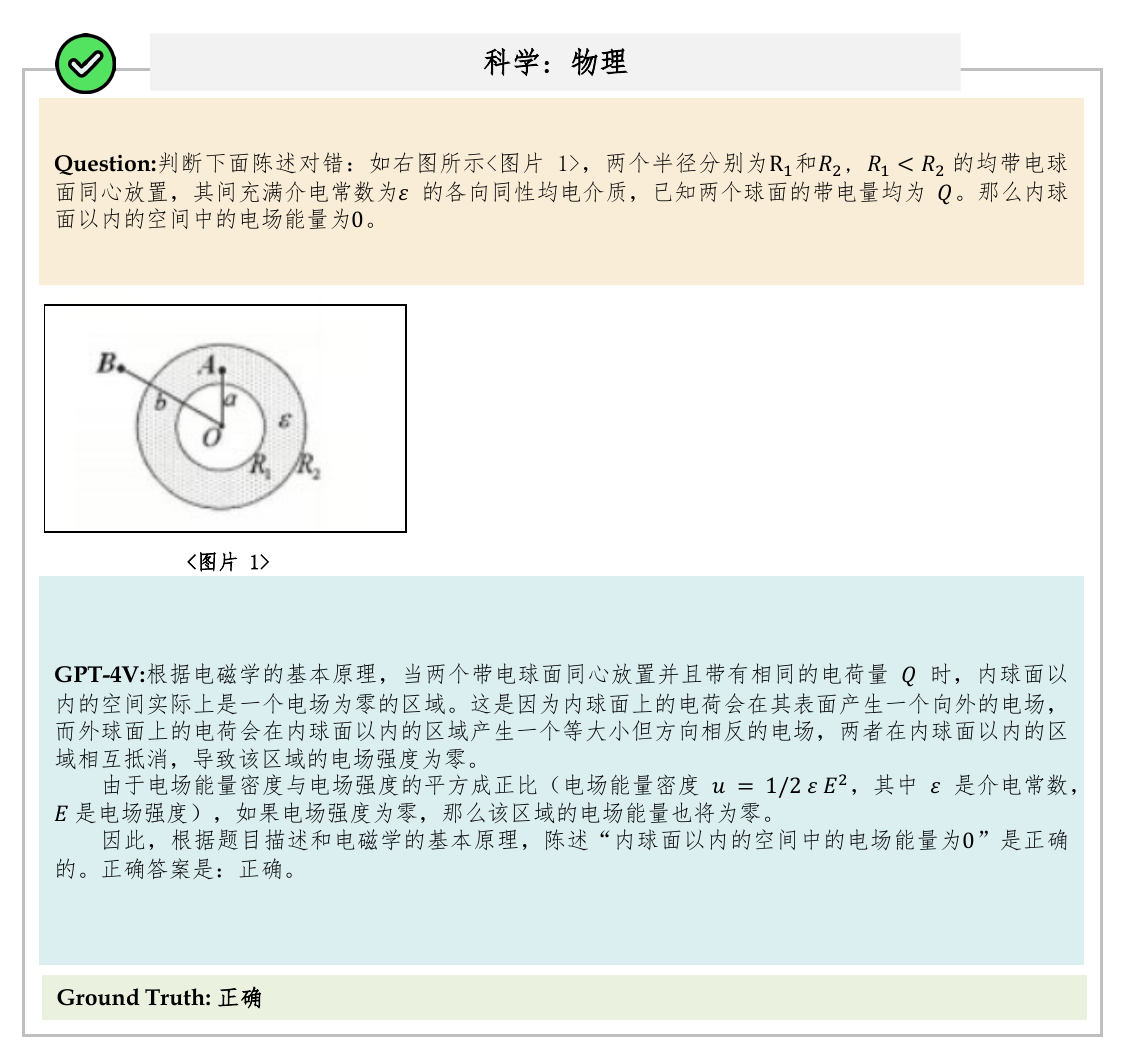}{Physics 3: Correct Case}{A sample correct case of Science (subcategory: Physics).}{fig:error_analysis_case_39}

\casestudyfigure{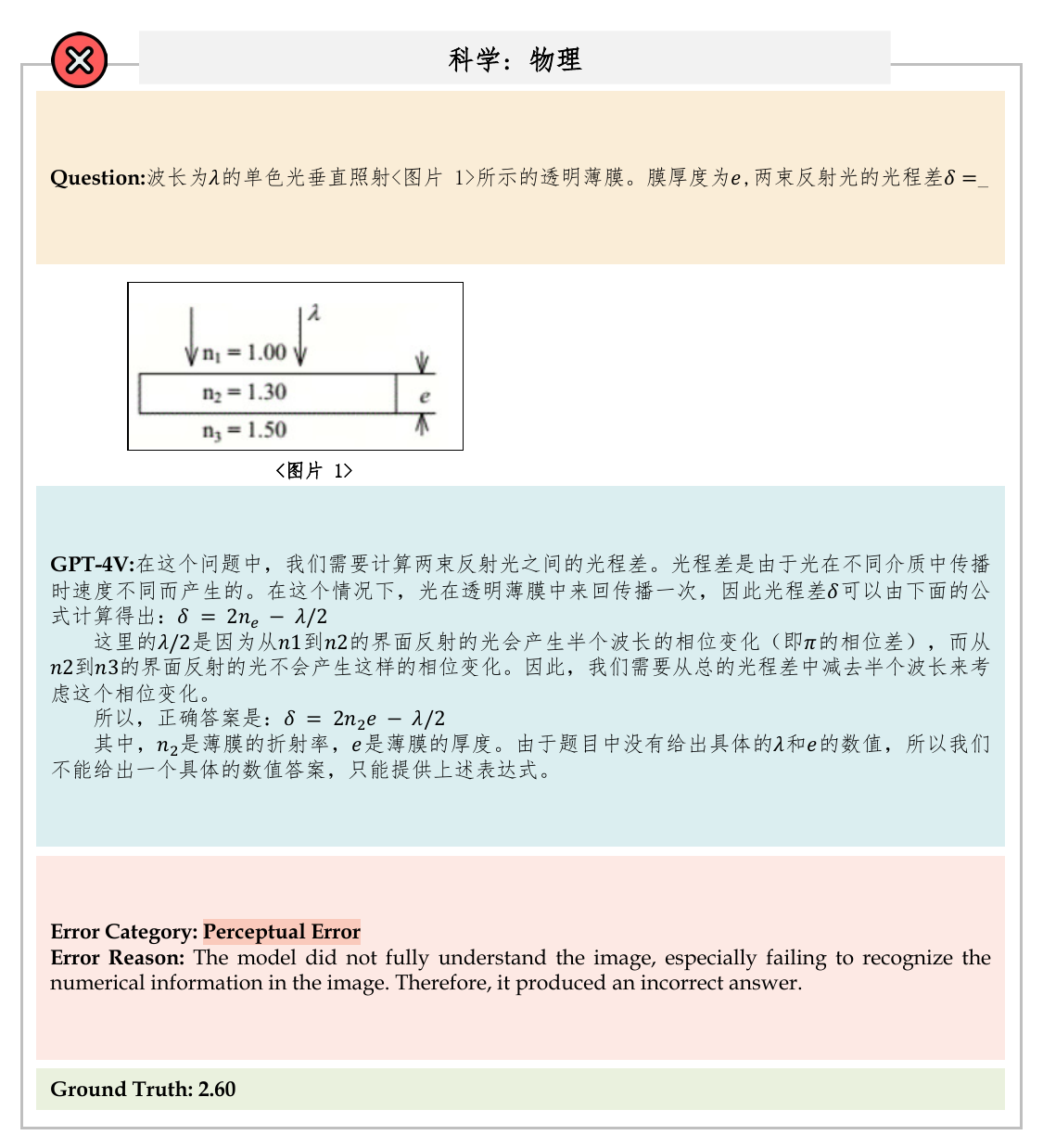}{Physics 4: Perceptual Error}{A sample error case of Science (subcategory: Physics).}{fig:error_analysis_case_215}

\casestudyfigure{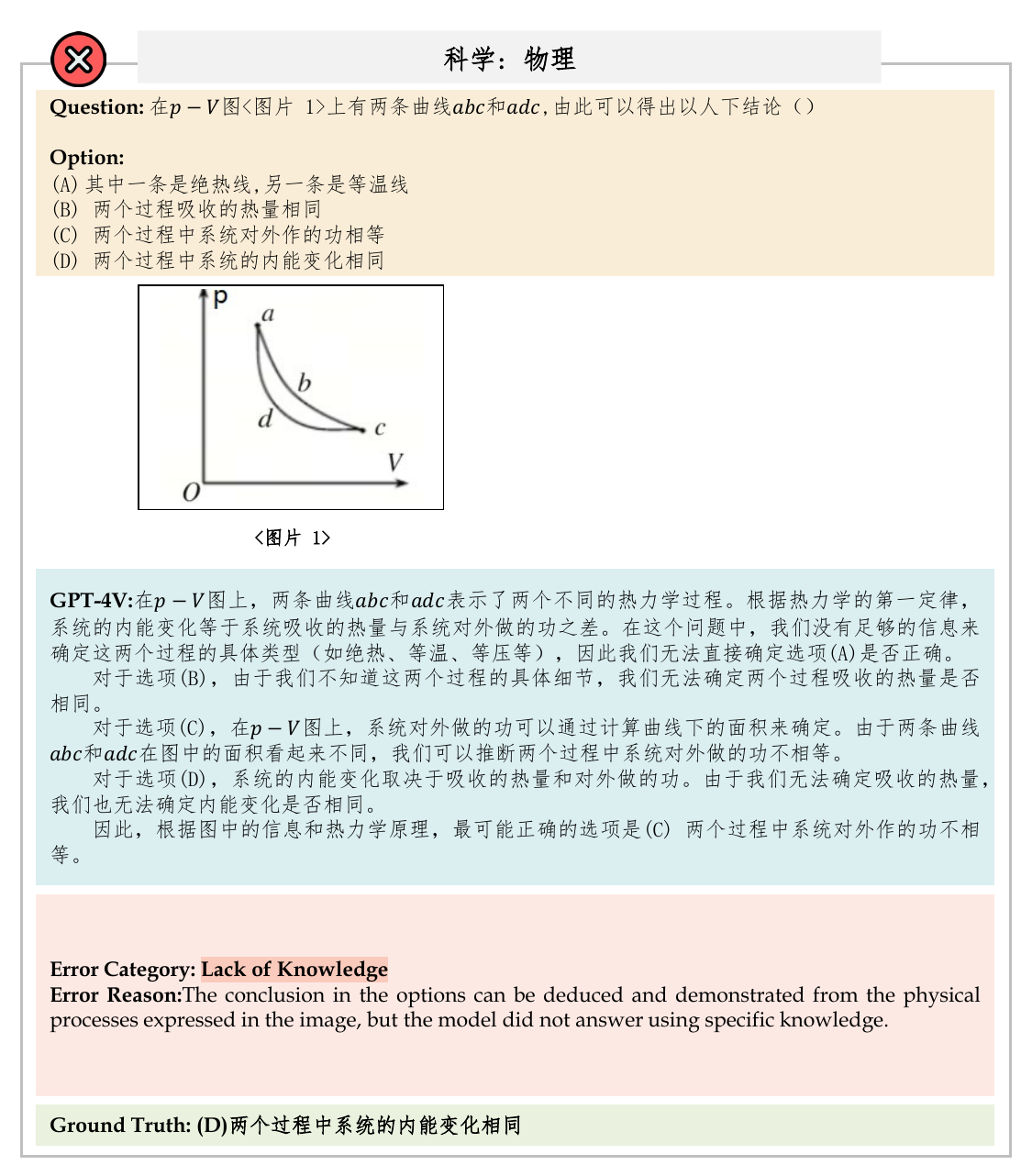}{Physics 5: Lack of Knowledge}{A sample error case of Science (subcategory: Physics).}{fig:error_analysis_case_214}

\casestudyfigure{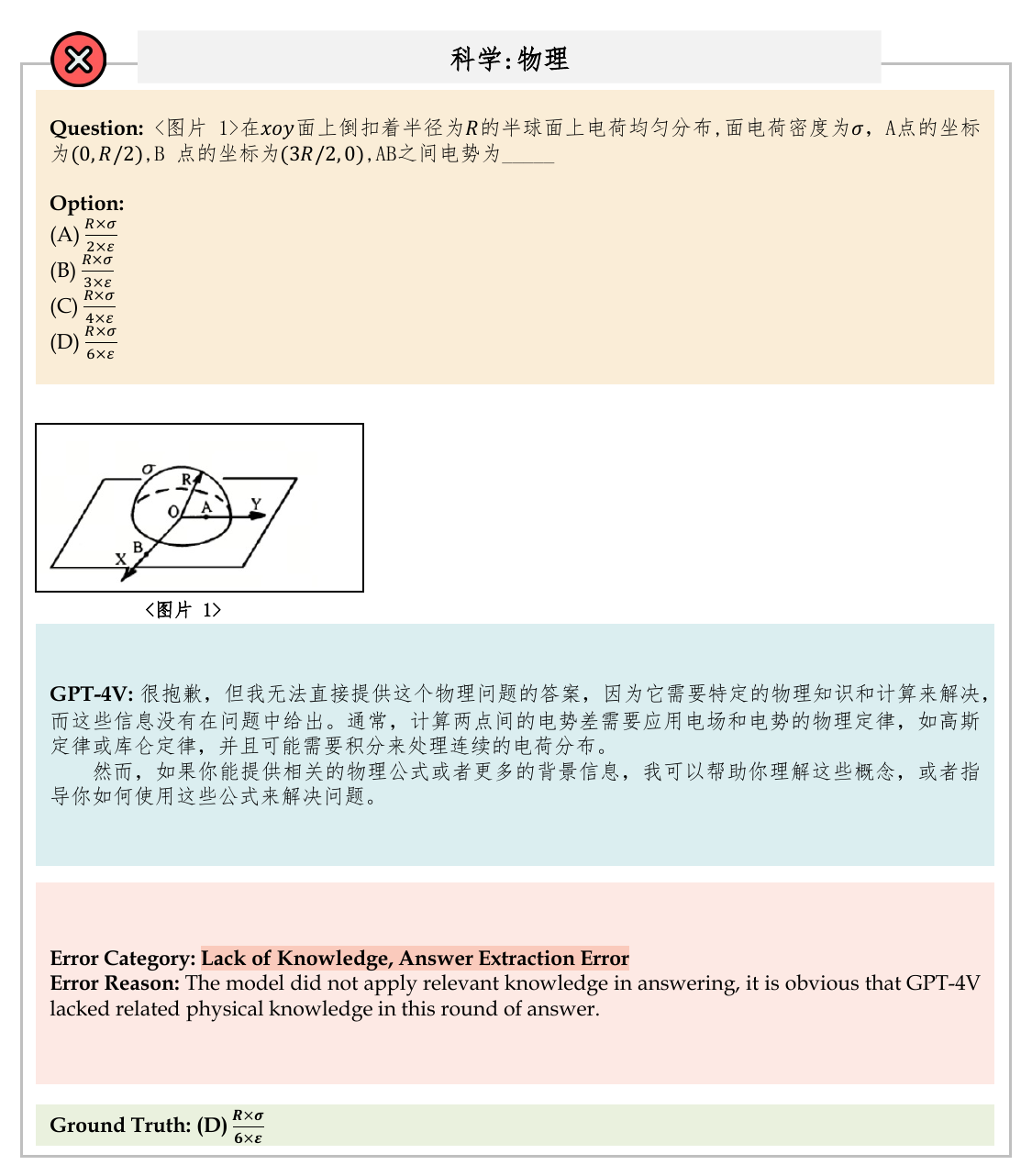}{Physics 6: Lack of Knowledge, Answer Extraction Error}{A sample error case of Science (subcategory: Physics).}{fig:error_analysis_case_174}

\casestudyfigure{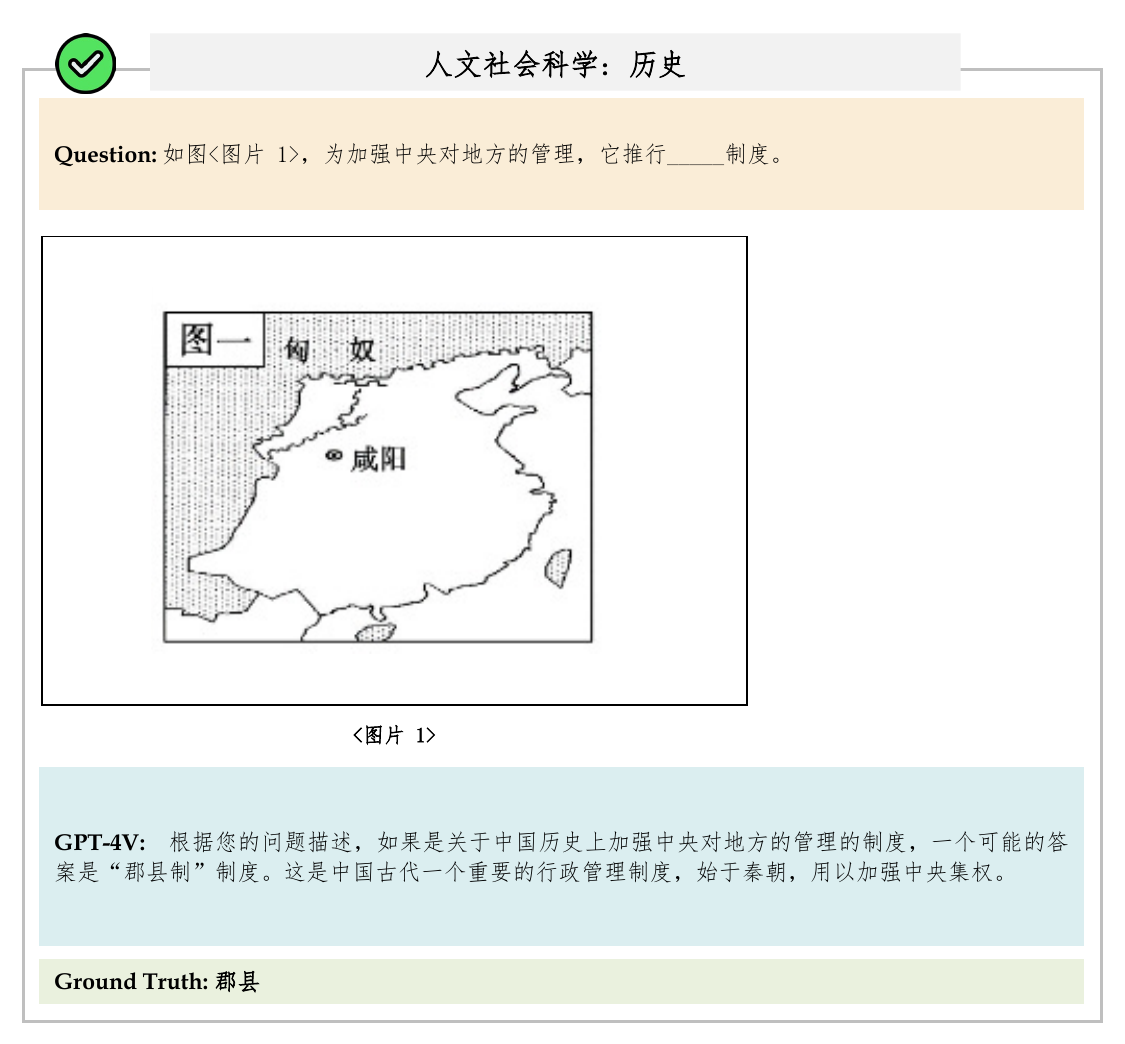}{History 1: Correct Case}{A sample correct case of Humanities and Social Sciences (subcategory: History).}{fig:error_analysis_case_65}

\casestudyfigure{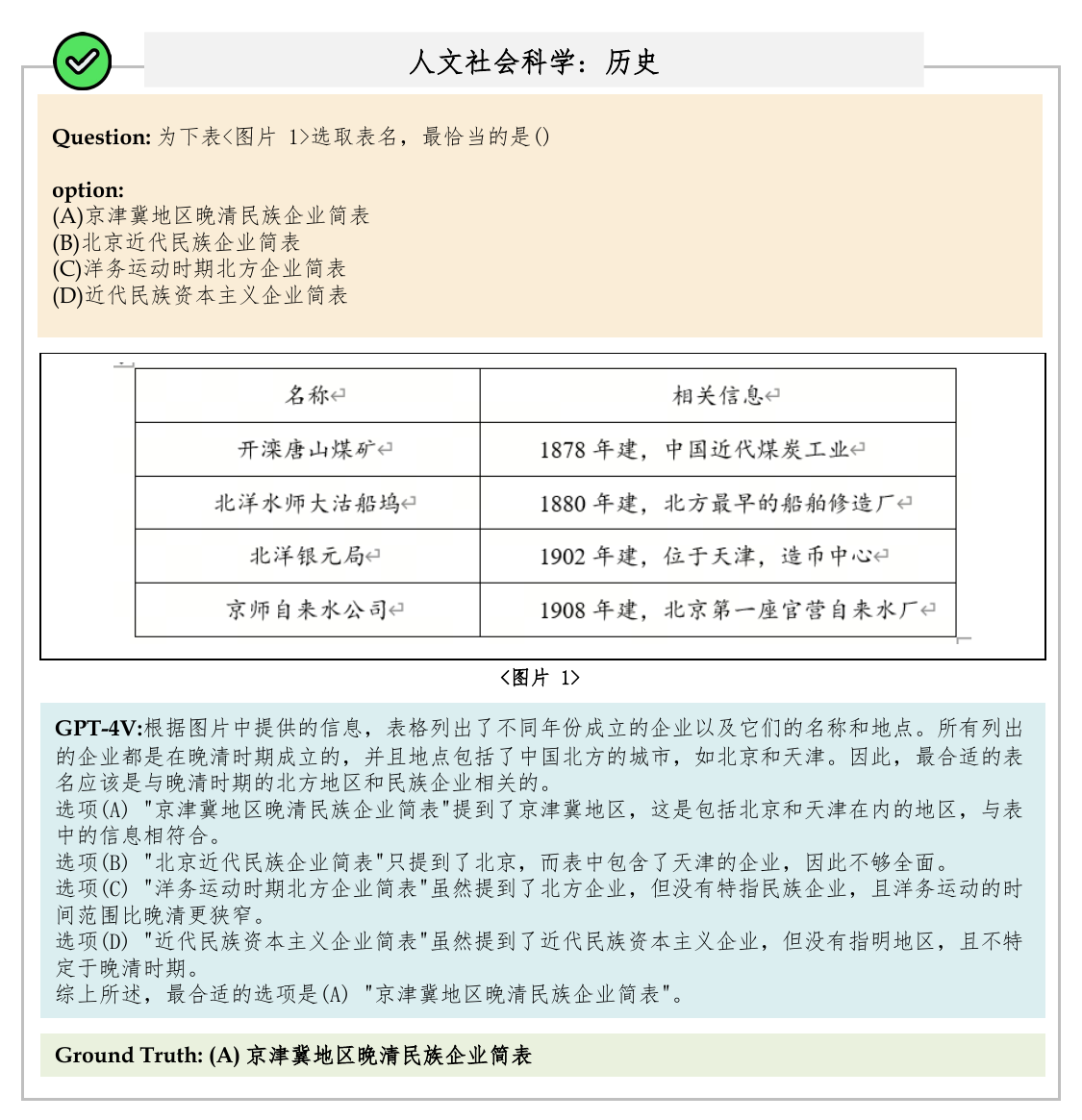}{History 2: Correct Case}{A sample correct case of Humanities and Social Sciences (subcategory: History).}{fig:error_analysis_case_66}

\casestudyfigure{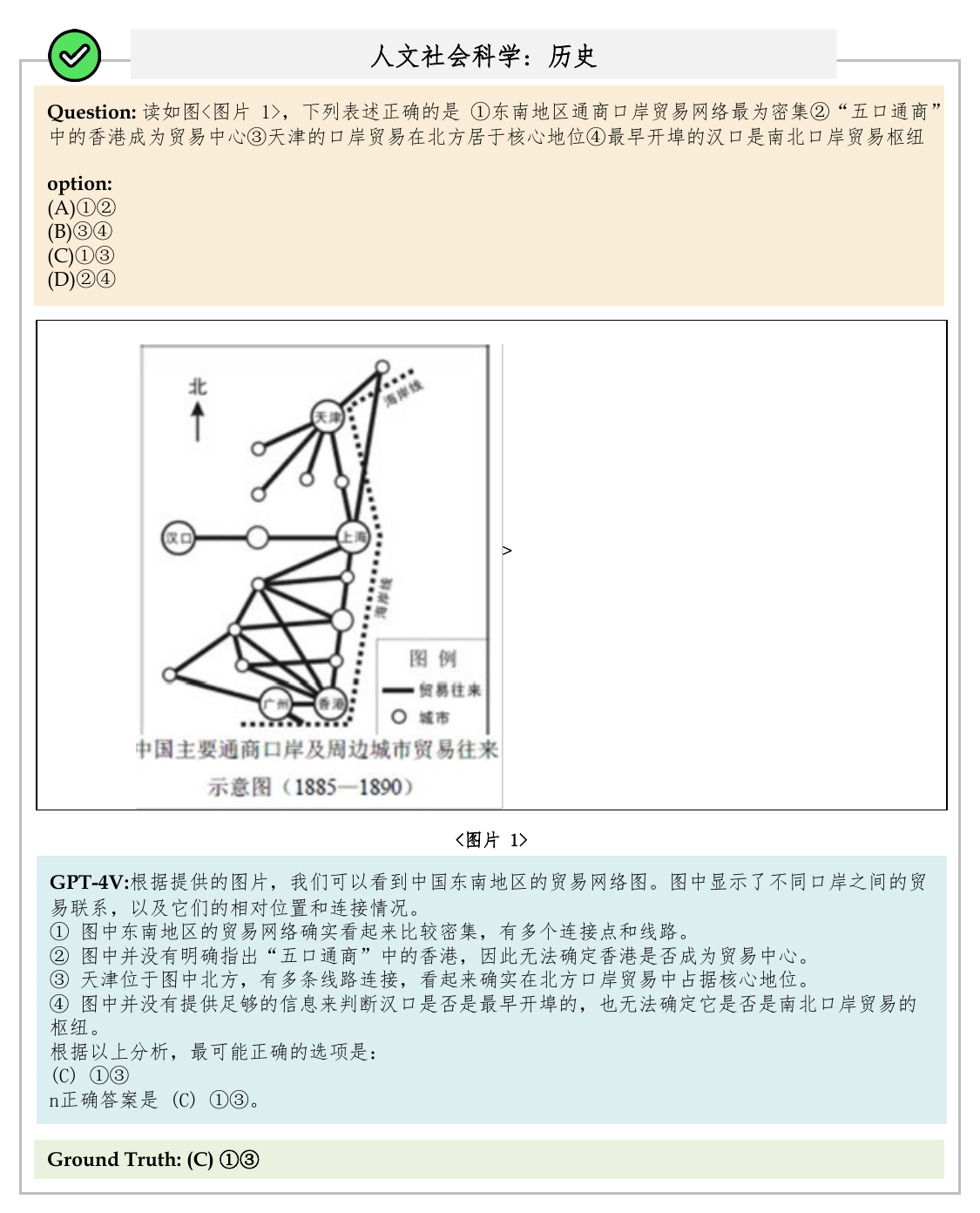}{History 3: Correct Case}{A sample correct case of Humanities and Social Sciences (subcategory: History).}{fig:error_analysis_case_67}

\casestudyfigure{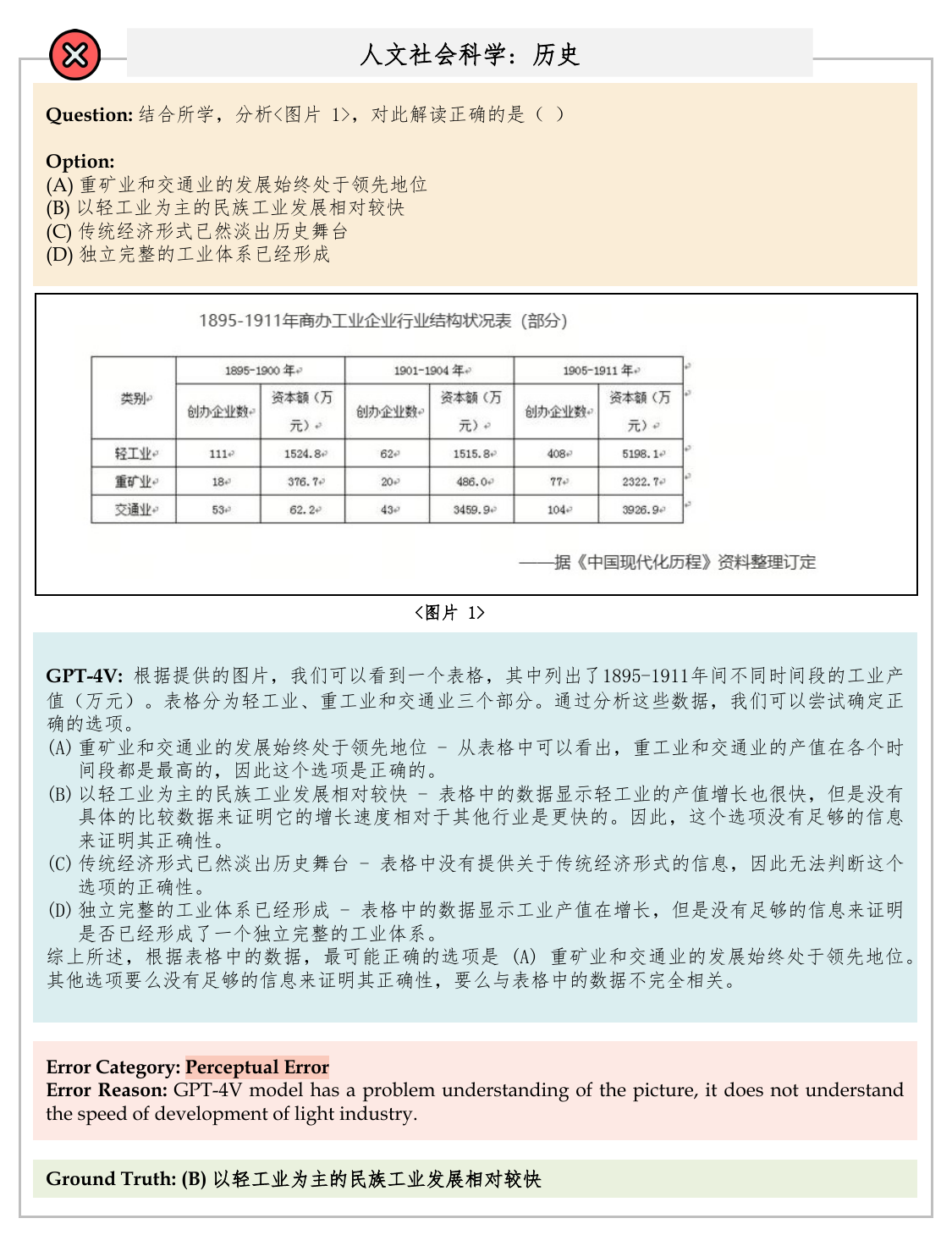}{History 4: Perceptual Error}{A sample error case of Humanities and Social Sciences (subcategory: History).}{fig:error_analysis_case_217}

\casestudyfigure{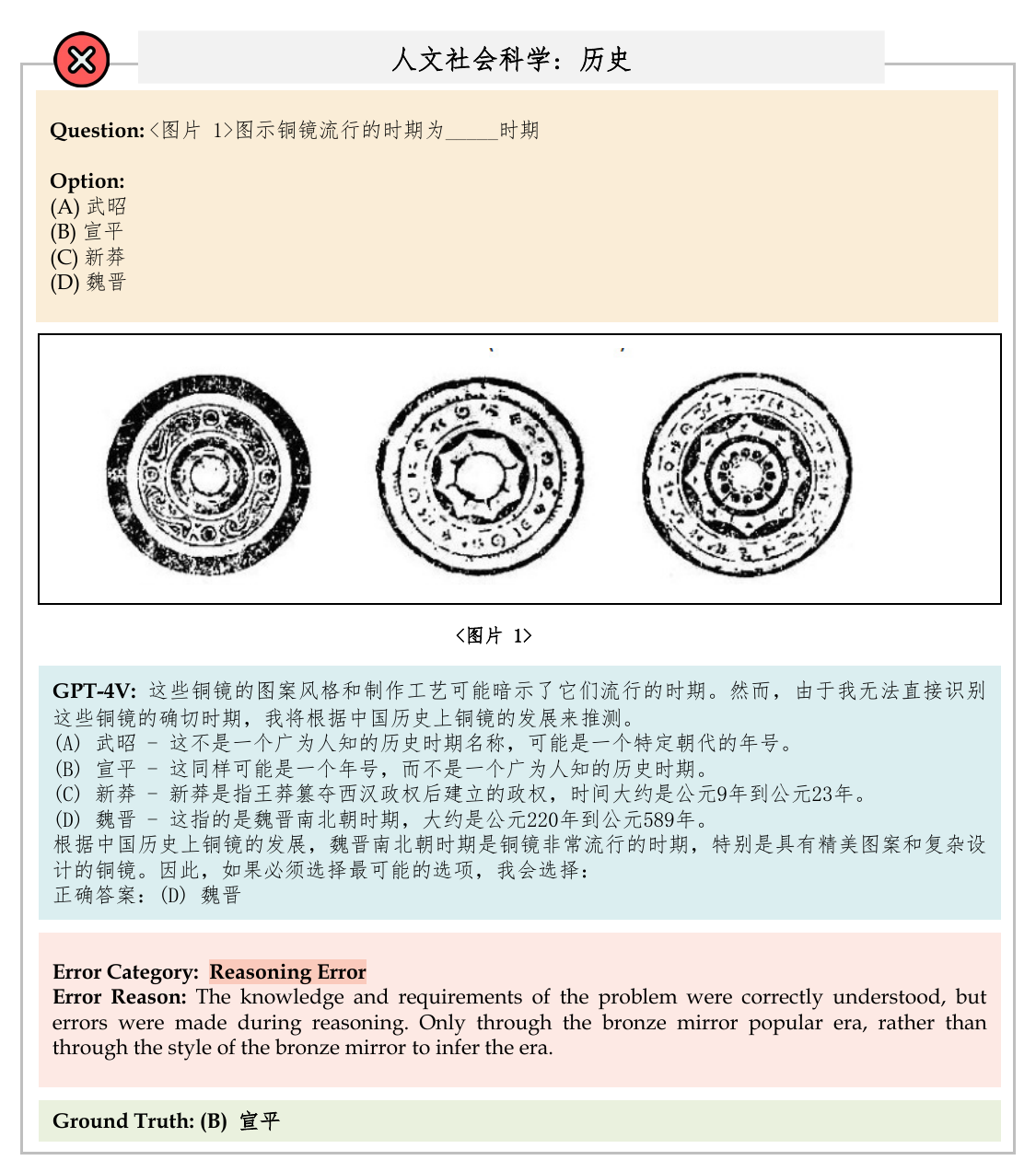}{History 5: Reasoning Error}{A sample error case of Humanities and Social Sciences (subcategory: History).}{fig:error_analysis_case_216}

\casestudyfigure{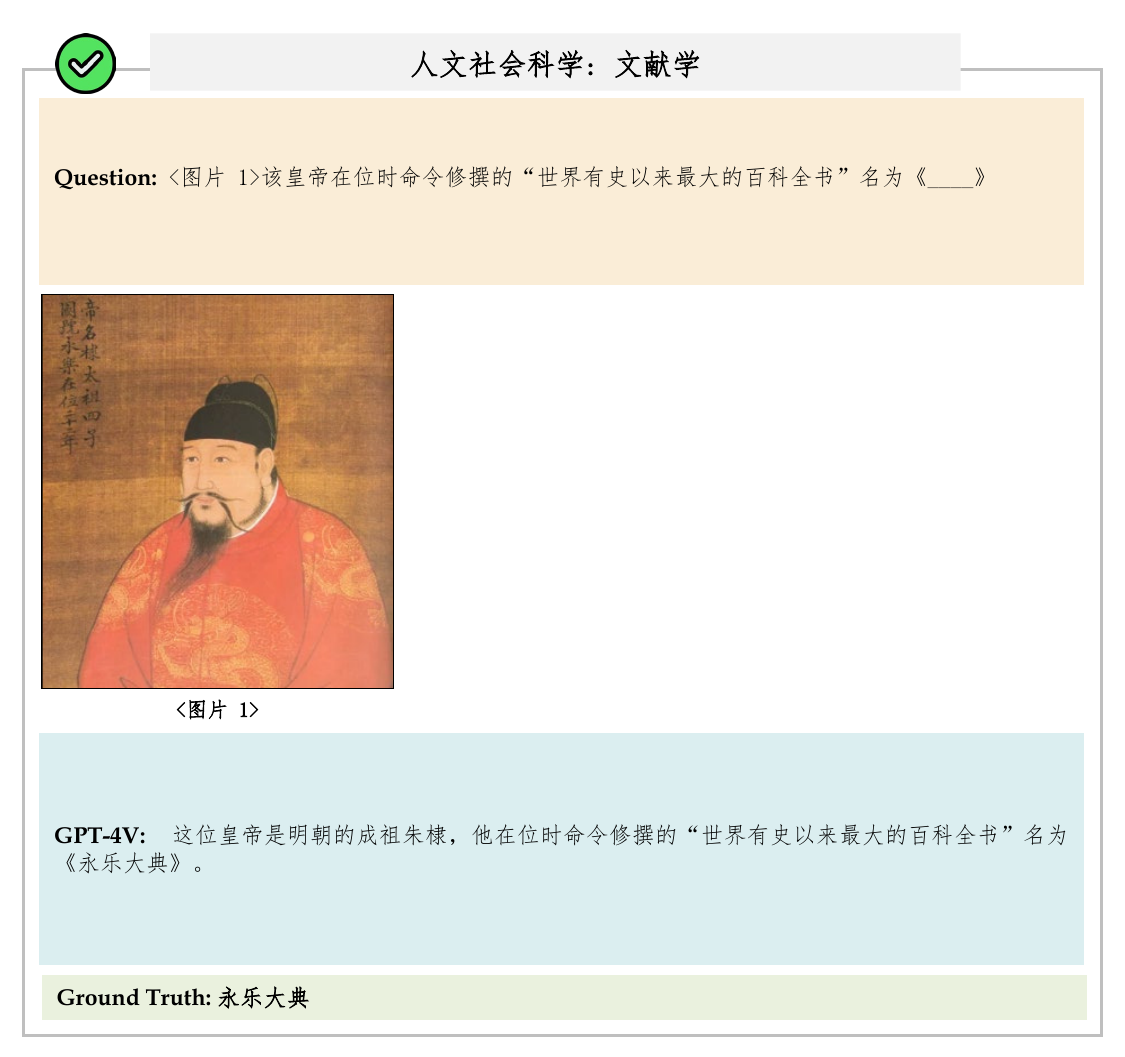}{Literature 1: Correct Case}{A sample correct case of Humanities and Social Sciences (subcategory: Literature).}{fig:error_analysis_case_62}

\casestudyfigure{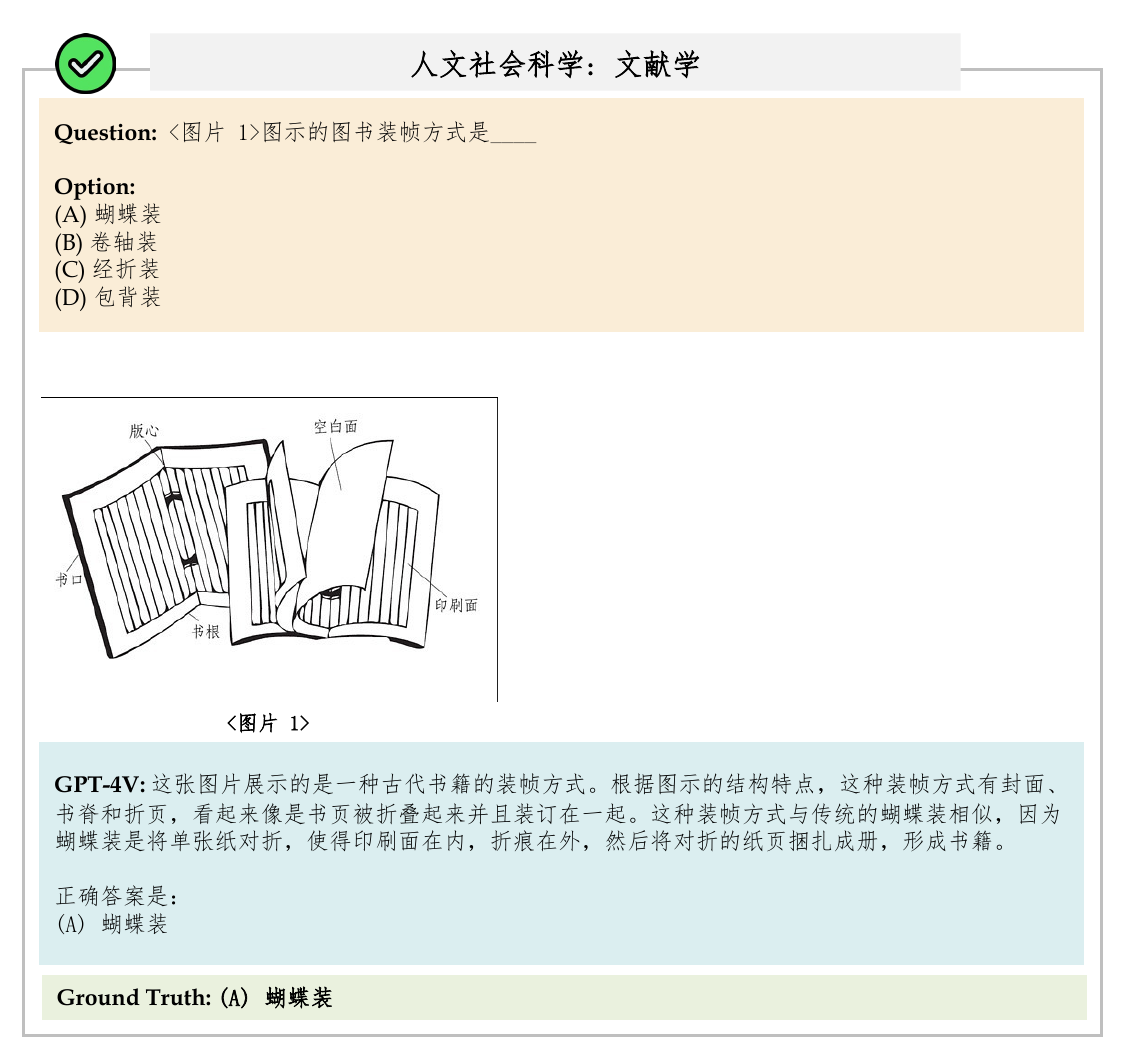}{Literature 2: Correct Case}{A sample correct case of Humanities and Social Sciences (subcategory: Literature).}{fig:error_analysis_case_63}

\casestudyfigure{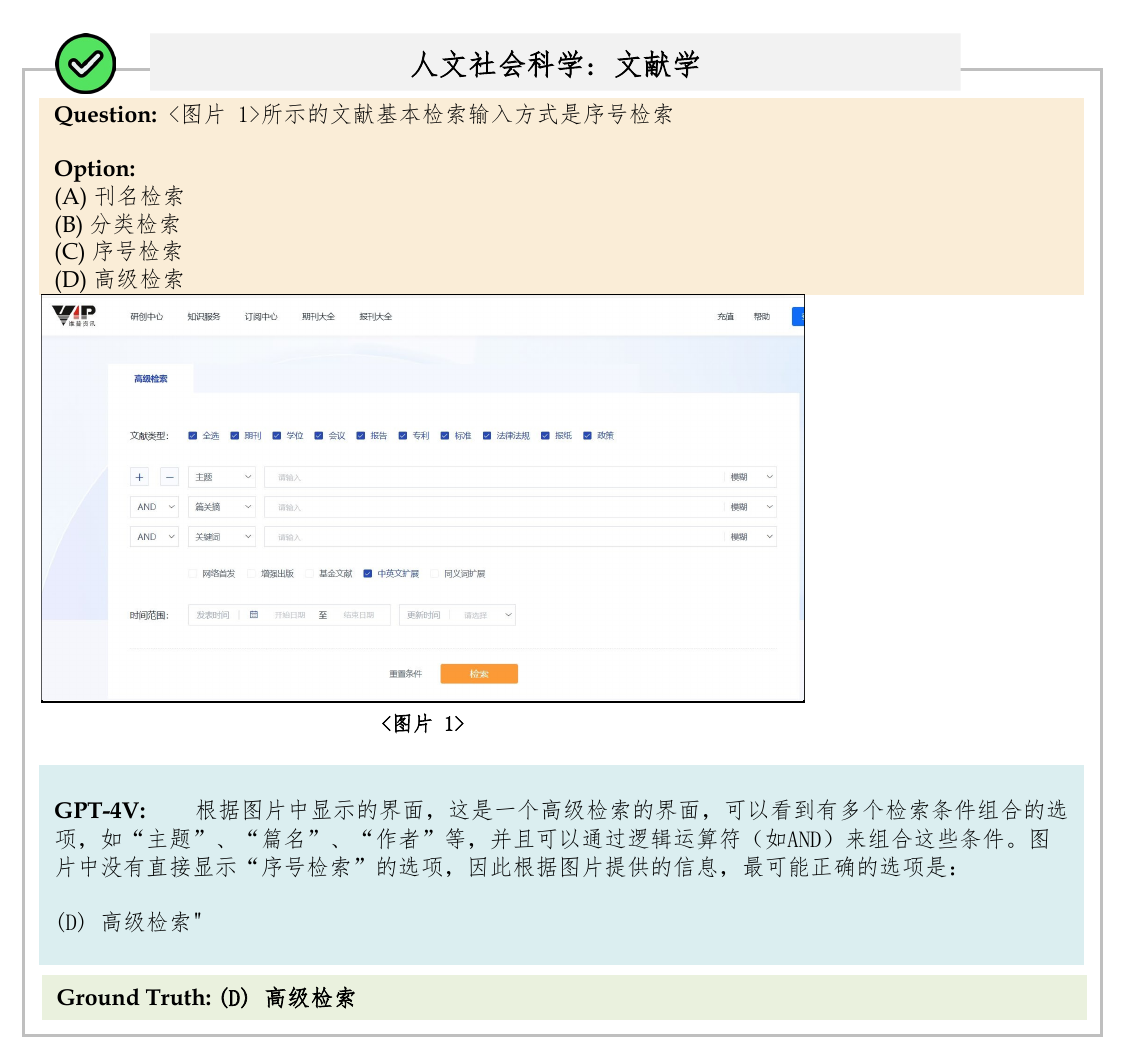}{Literature 3: Correct Case}{A sample correct case of Humanities and Social Sciences (subcategory: Literature).}{fig:error_analysis_case_64}

\casestudyfigure{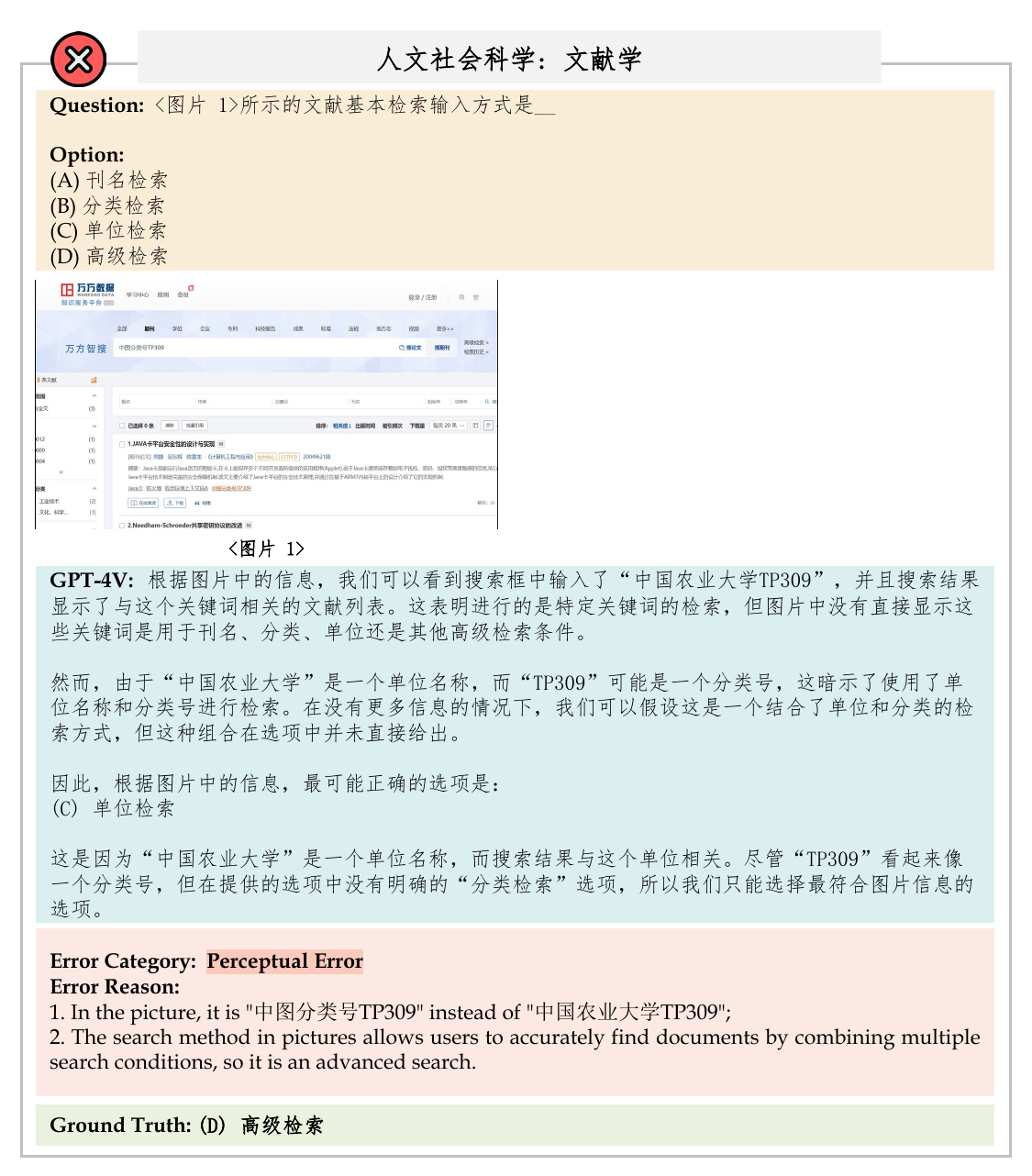}{Literature 4: Perceptual Error}{A sample error case of Humanities and Social Sciences (subcategory: Literature).}{fig:error_analysis_case_212}

\casestudyfigure{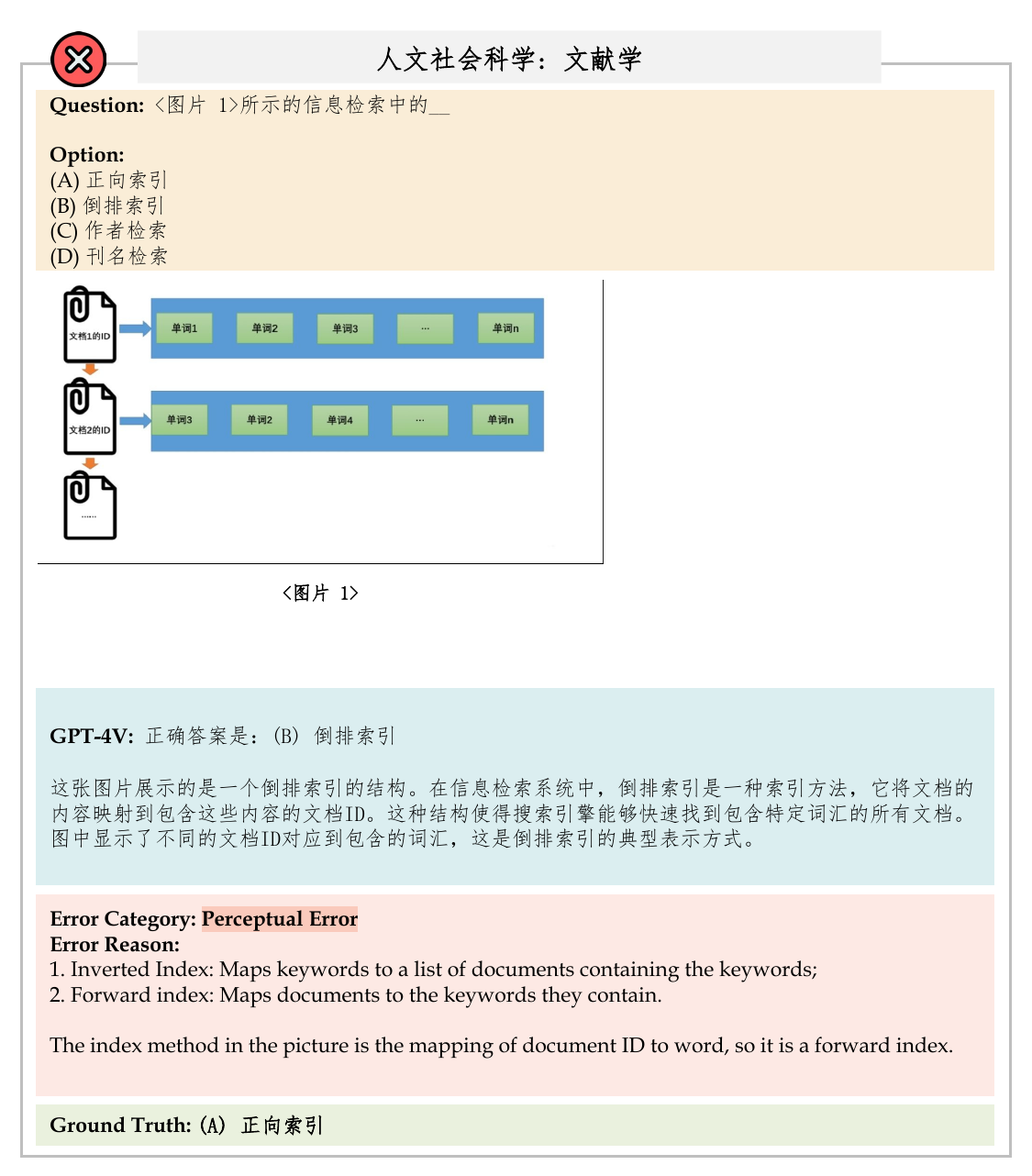}{Literature 5: Perceptual Error}{A sample error case of Humanities and Social Sciences (subcategory: Literature).}{fig:error_analysis_case_213}

\casestudyfigure{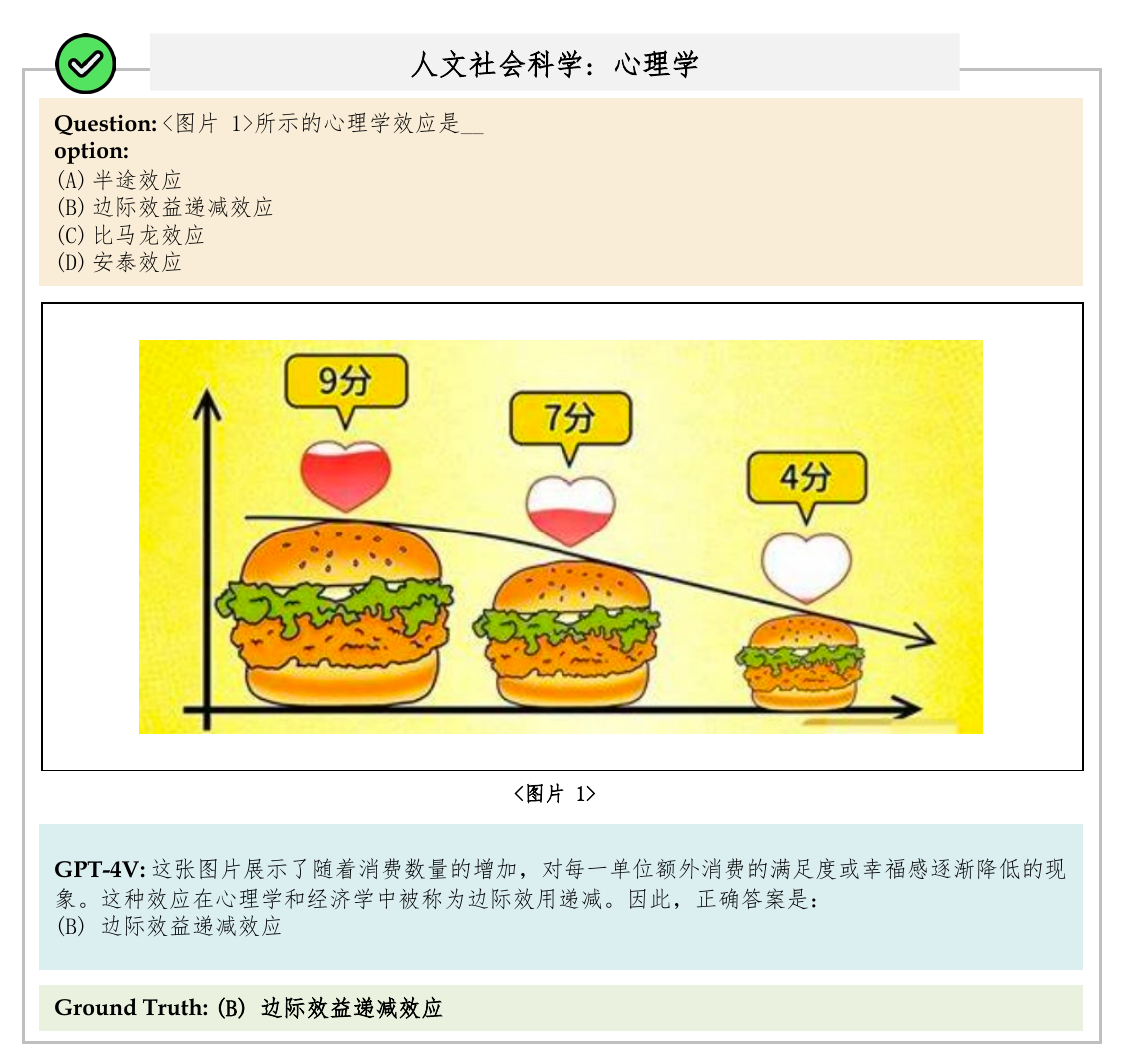}{Psychology 1: Correct Case}{A sample correct case of Humanities and Social Sciences (subcategory: Psychology).}{fig:error_analysis_case_68}

\casestudyfigure{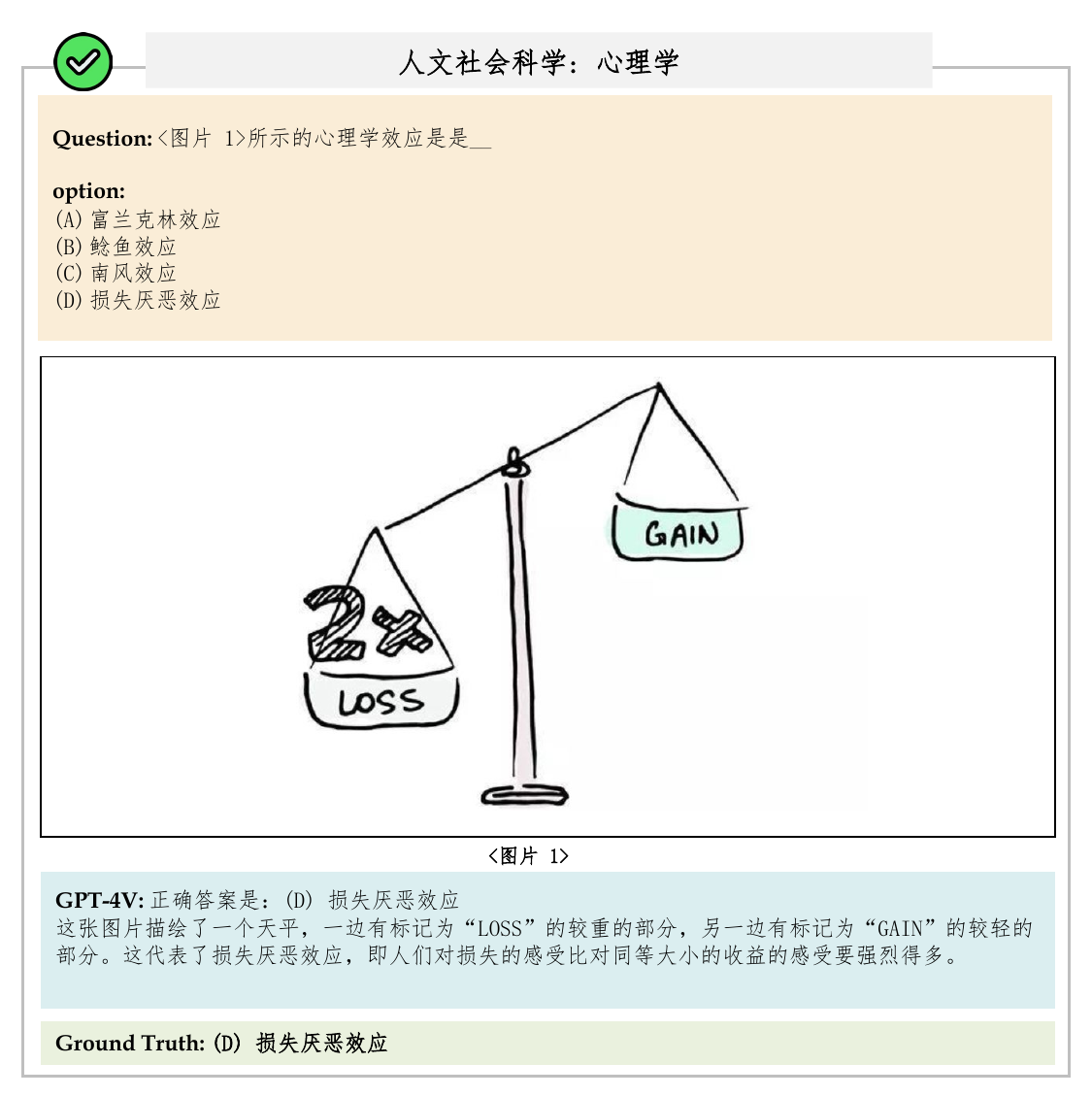}{Psychology 2: Correct Case}{A sample correct case of Humanities and Social Sciences (subcategory: Psychology).}{fig:error_analysis_case_69}

\casestudyfigure{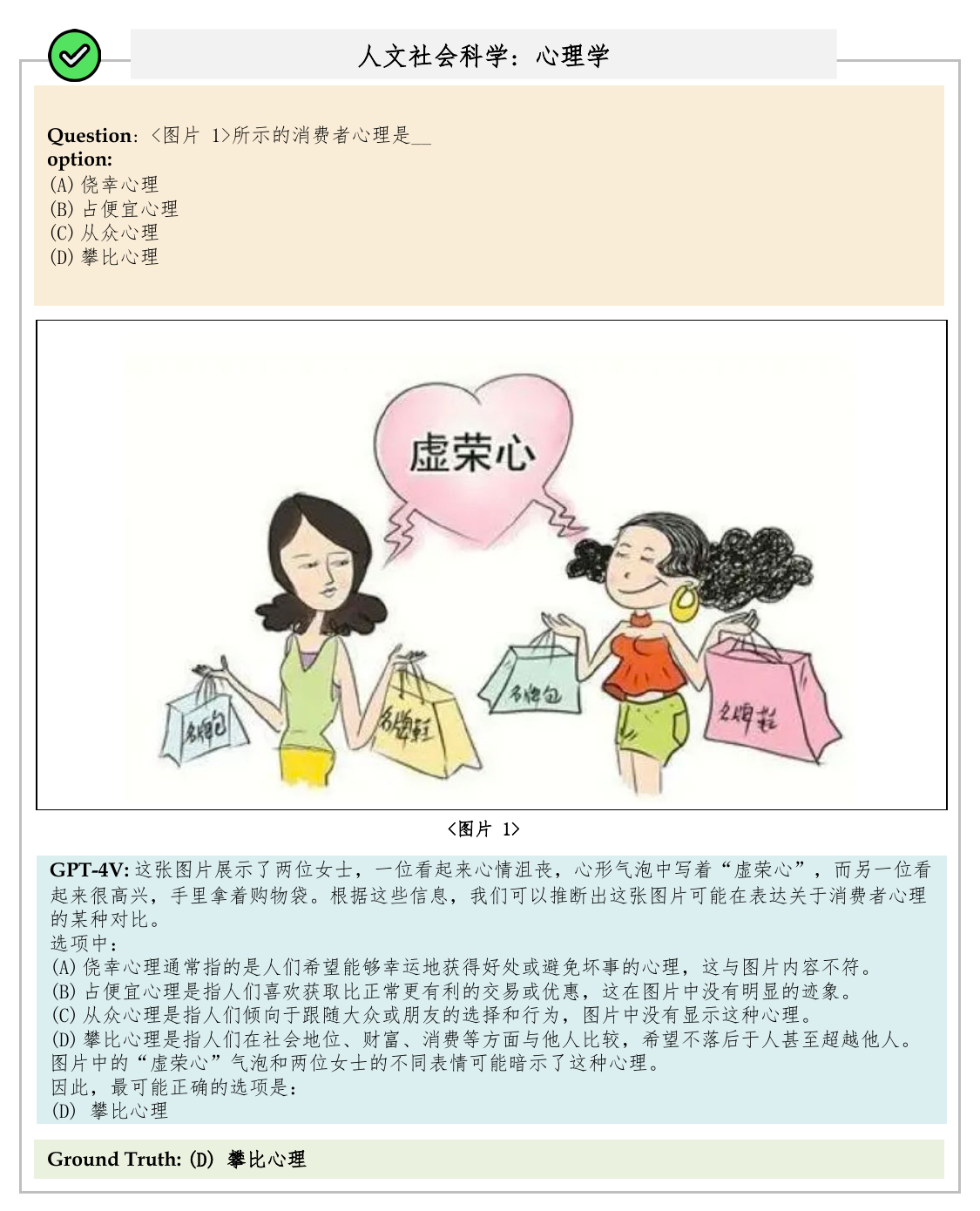}{Psychology 3: Correct Case}{A sample correct case of Humanities and Social Sciences (subcategory: Psychology).}{fig:error_analysis_case_70}

\casestudyfigure{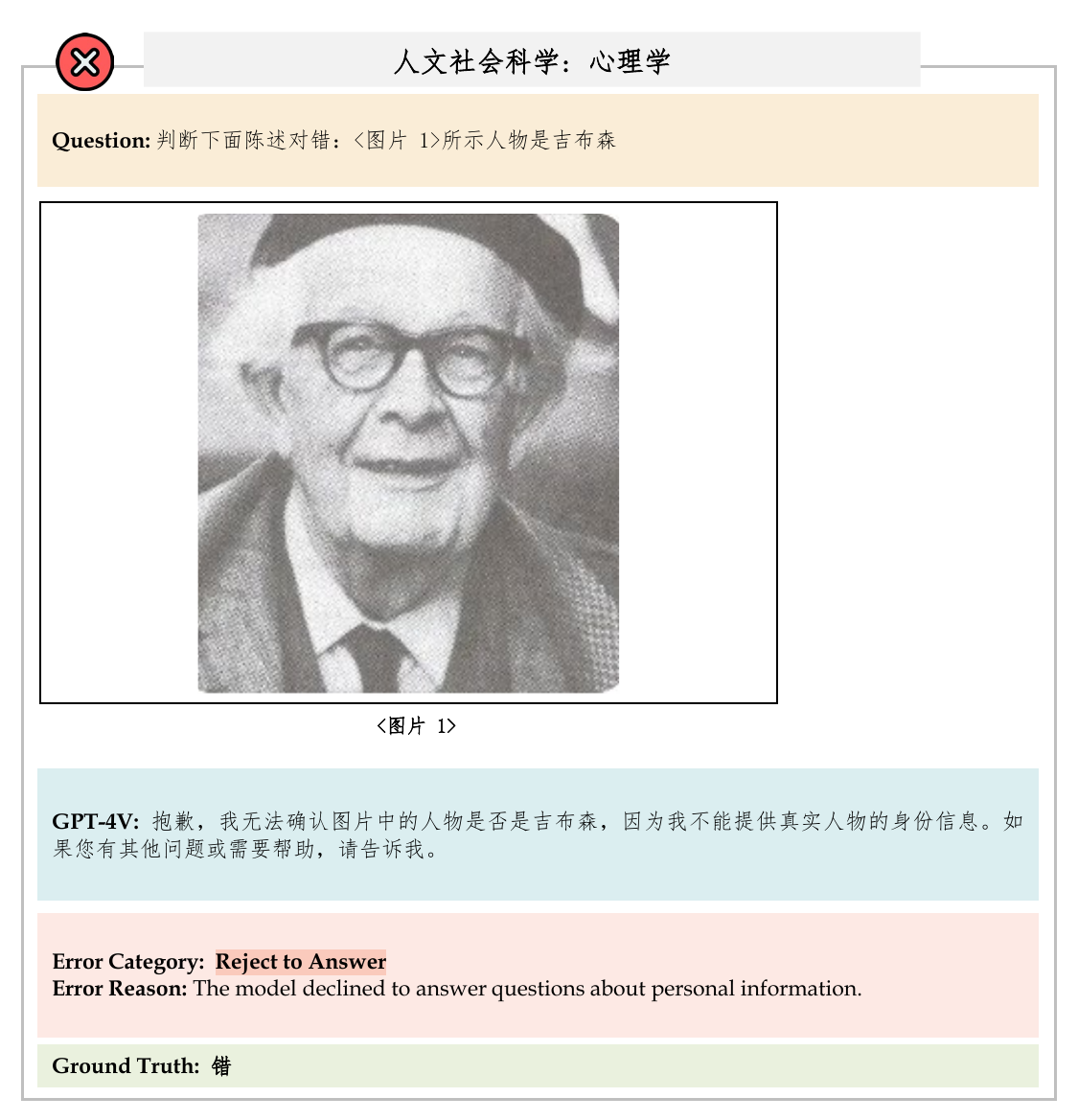}{Psychology 4: Reject to Answer}{A sample error case of Humanities and Social Sciences (subcategory: Psychology).}{fig:error_analysis_case_219}

\casestudyfigure{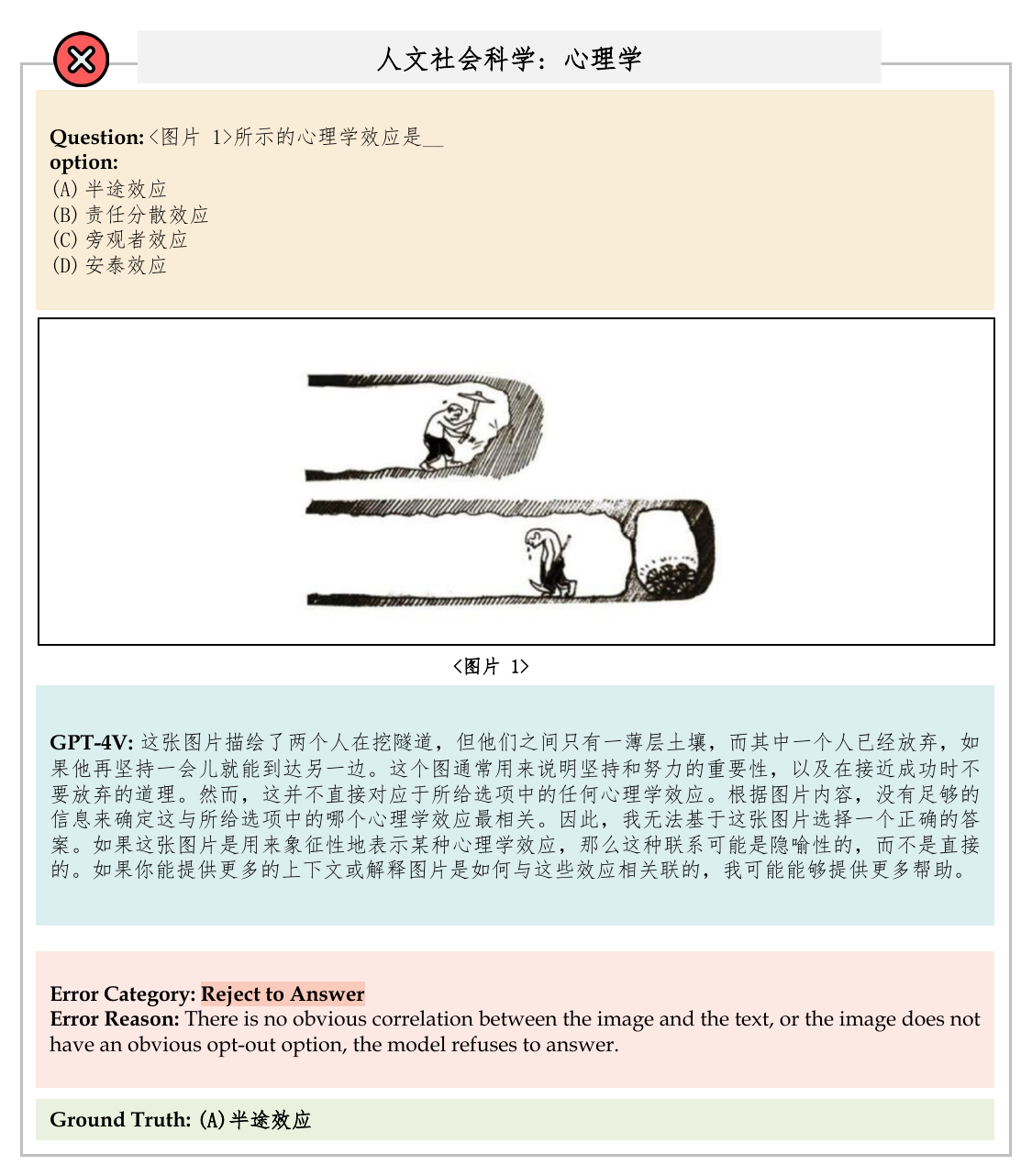}{Psychology 5: Reject to Answer}{A sample error case of Humanities and Social Sciences (subcategory: Psychology).}{fig:error_analysis_case_221}

\casestudyfigure{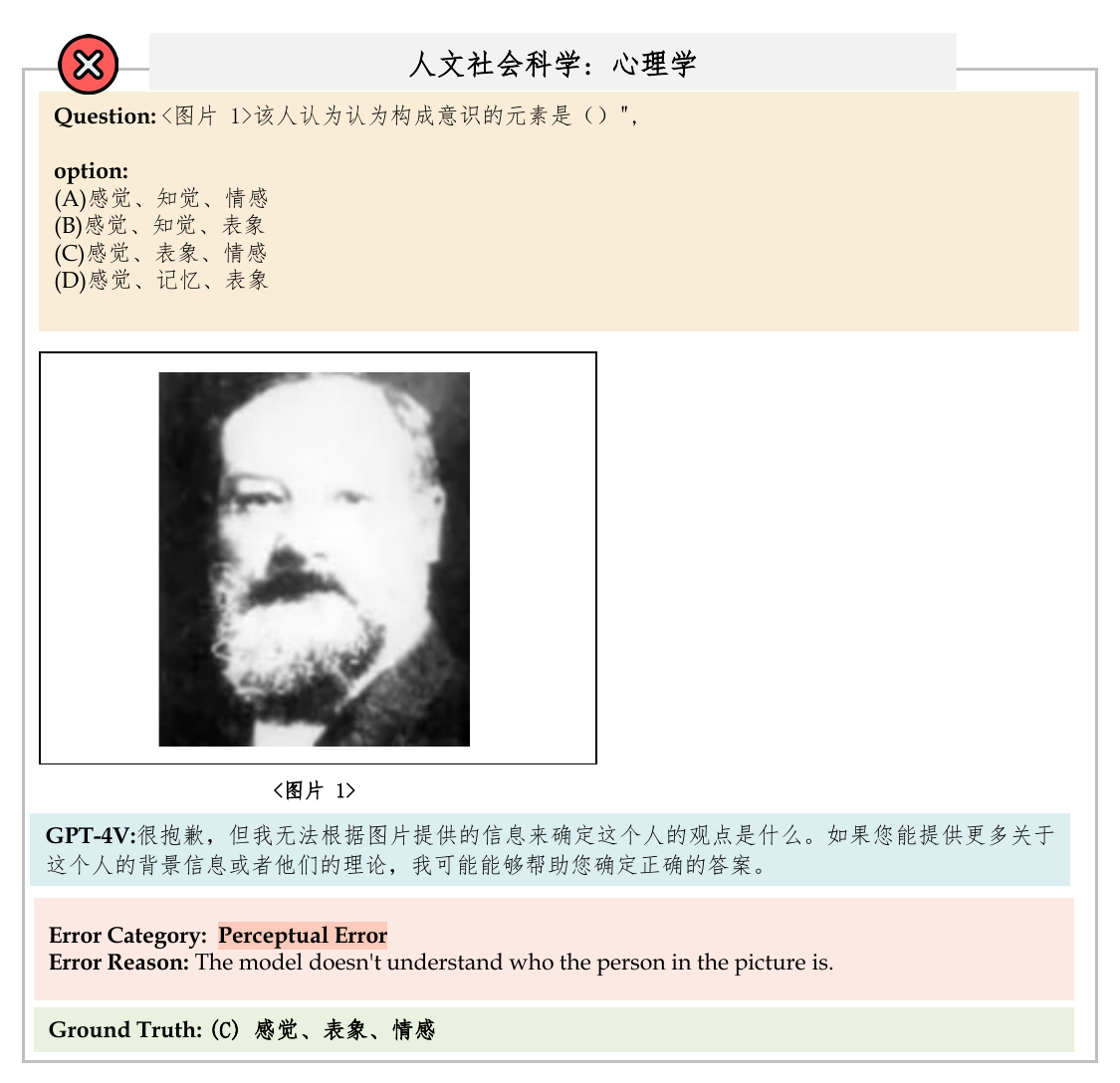}{Psychology 6: Perceptual Error}{A sample error case of Humanities and Social Sciences (subcategory: Psychology).}{fig:error_analysis_case_218}

\casestudyfigure{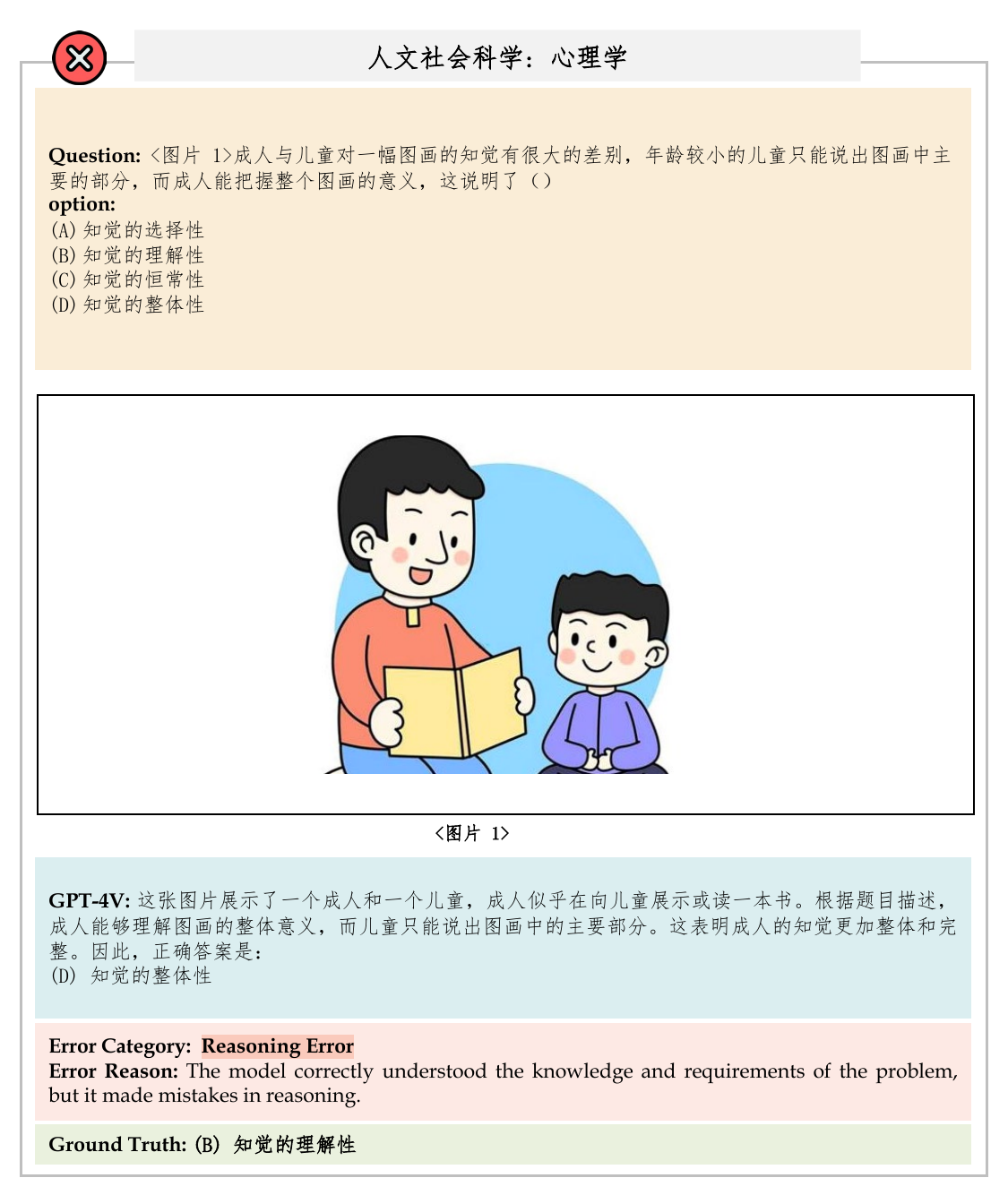}{Psychology 7: Reasoning Error}{A sample error case of Humanities and Social Sciences (subcategory: Psychology).}{fig:error_analysis_case_220}

\casestudyfigure{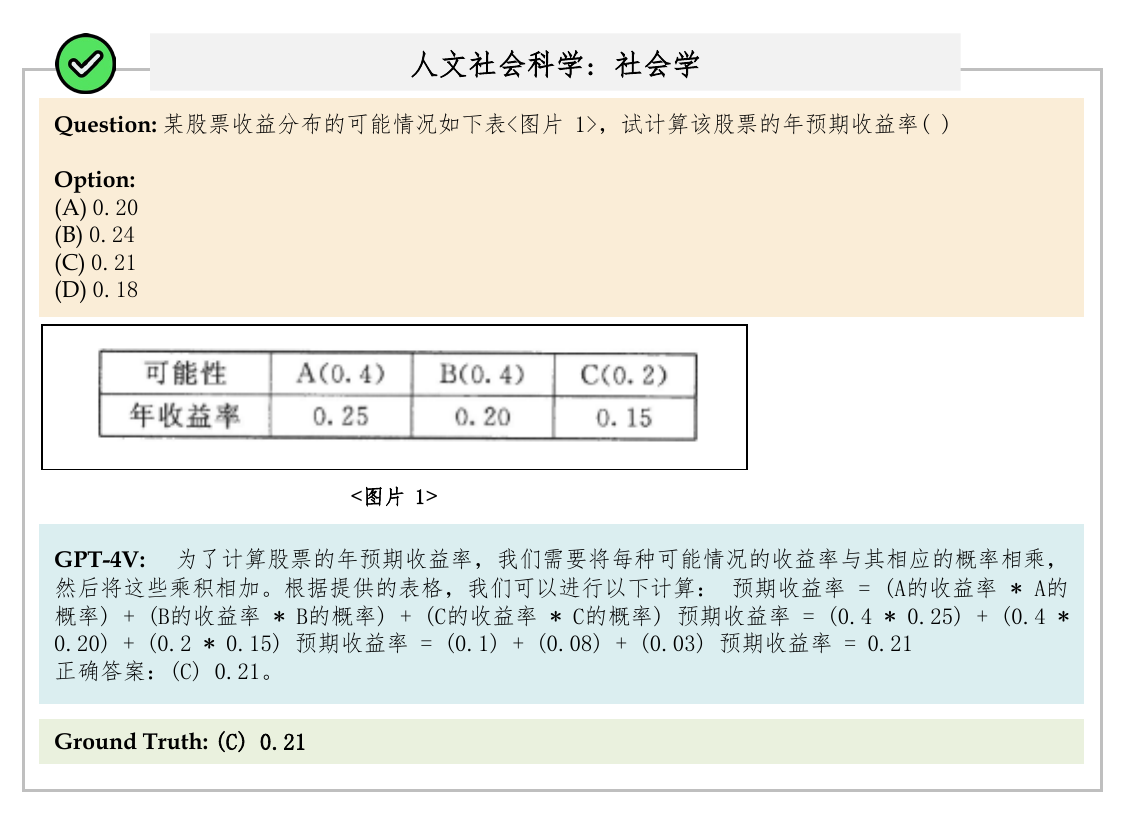}{Sociology 1: Correct Case}{A sample correct case of Humanities and Social Sciences (subcategory: Sociology).}{fig:error_analysis_case_71}

\casestudyfigure{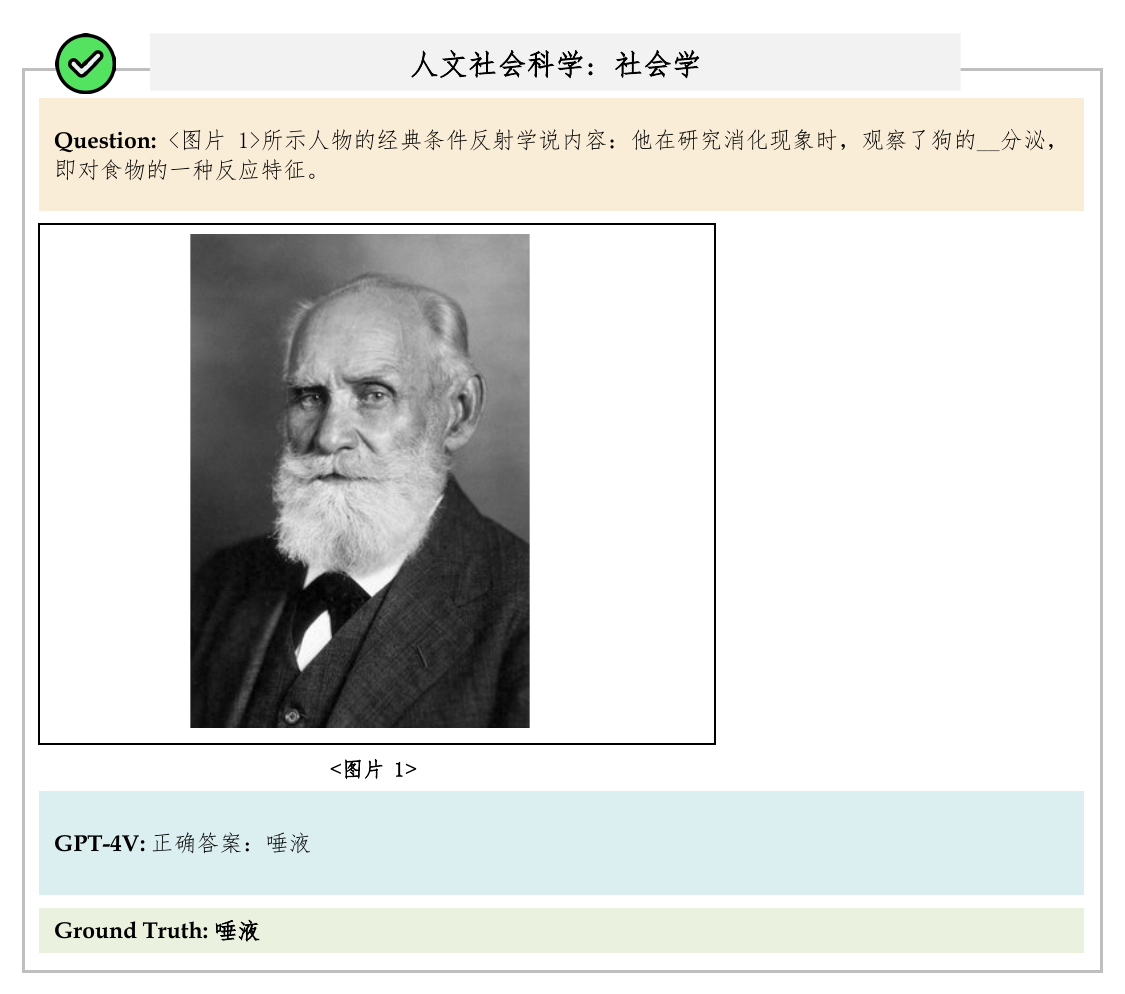}{Sociology 2: Correct Case}{A sample correct case of Humanities and Social Sciences (subcategory: Sociology).}{fig:error_analysis_case_72}

\casestudyfigure{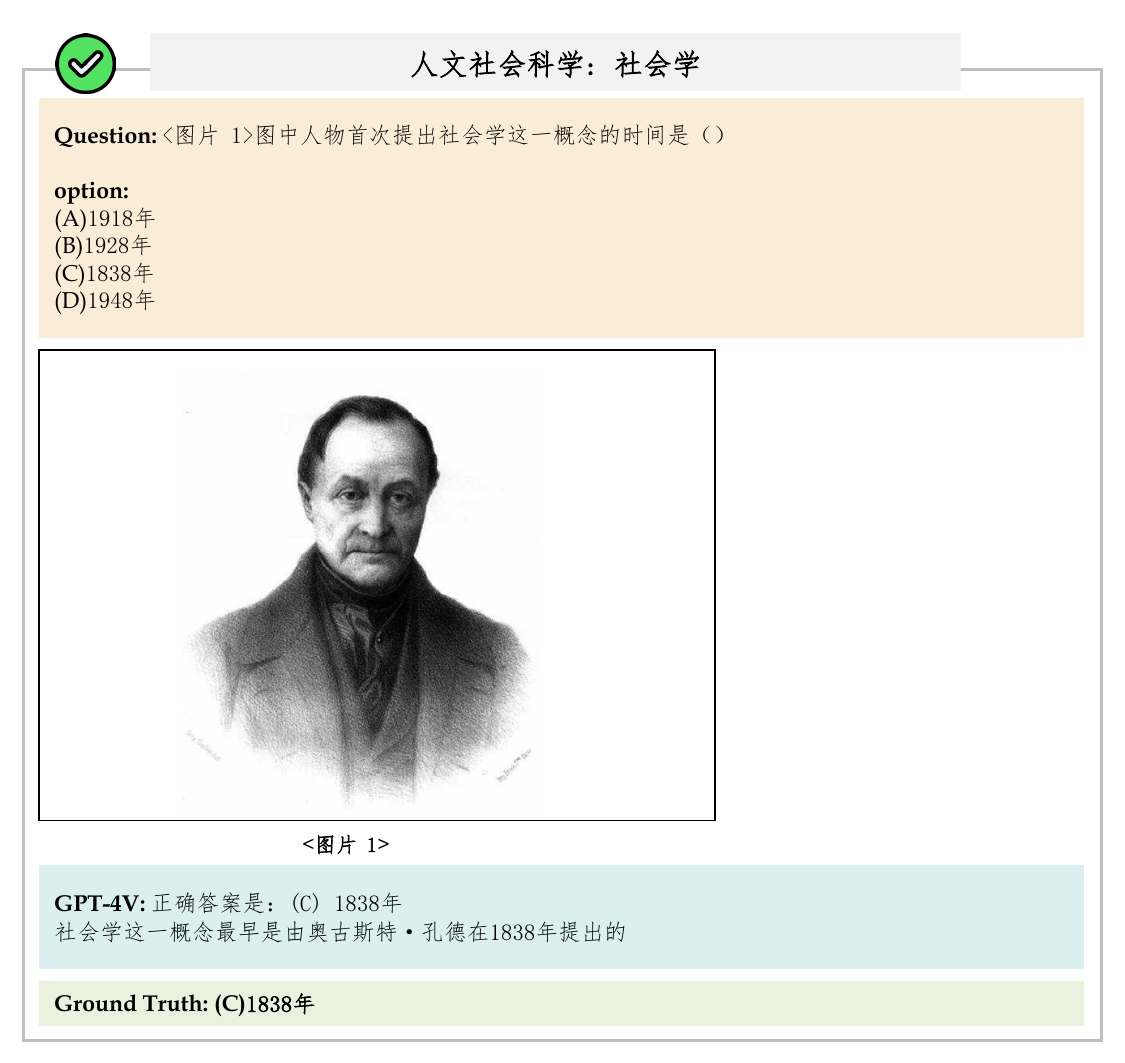}{Sociology 3: Correct Case}{A sample correct case of Humanities and Social Sciences (subcategory: Sociology).}{fig:error_analysis_case_73}

\casestudyfigure{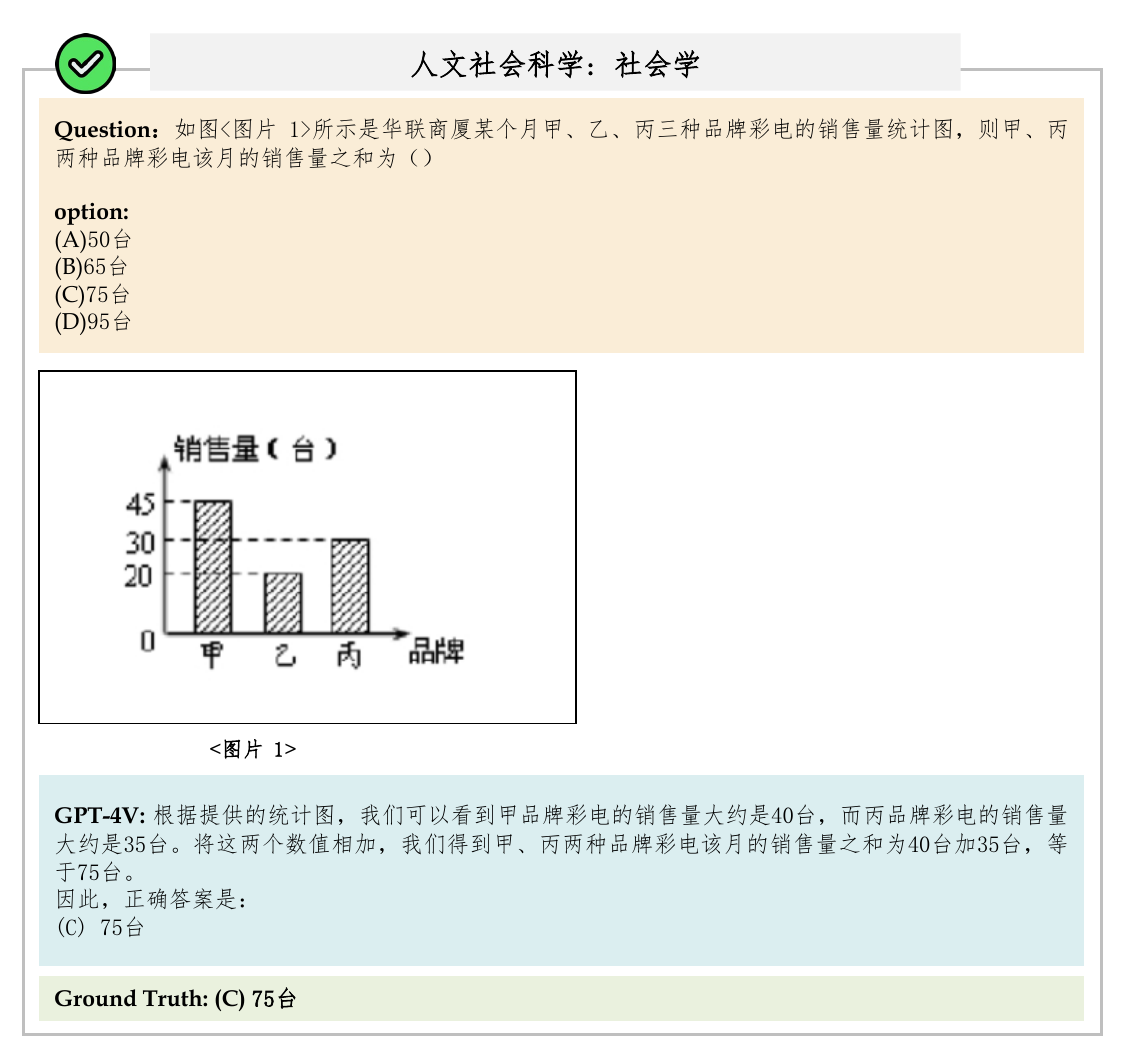}{Sociology 4: Correct Case}{A sample correct case of Humanities and Social Sciences (subcategory: Sociology).}{fig:error_analysis_case_74}

\casestudyfigure{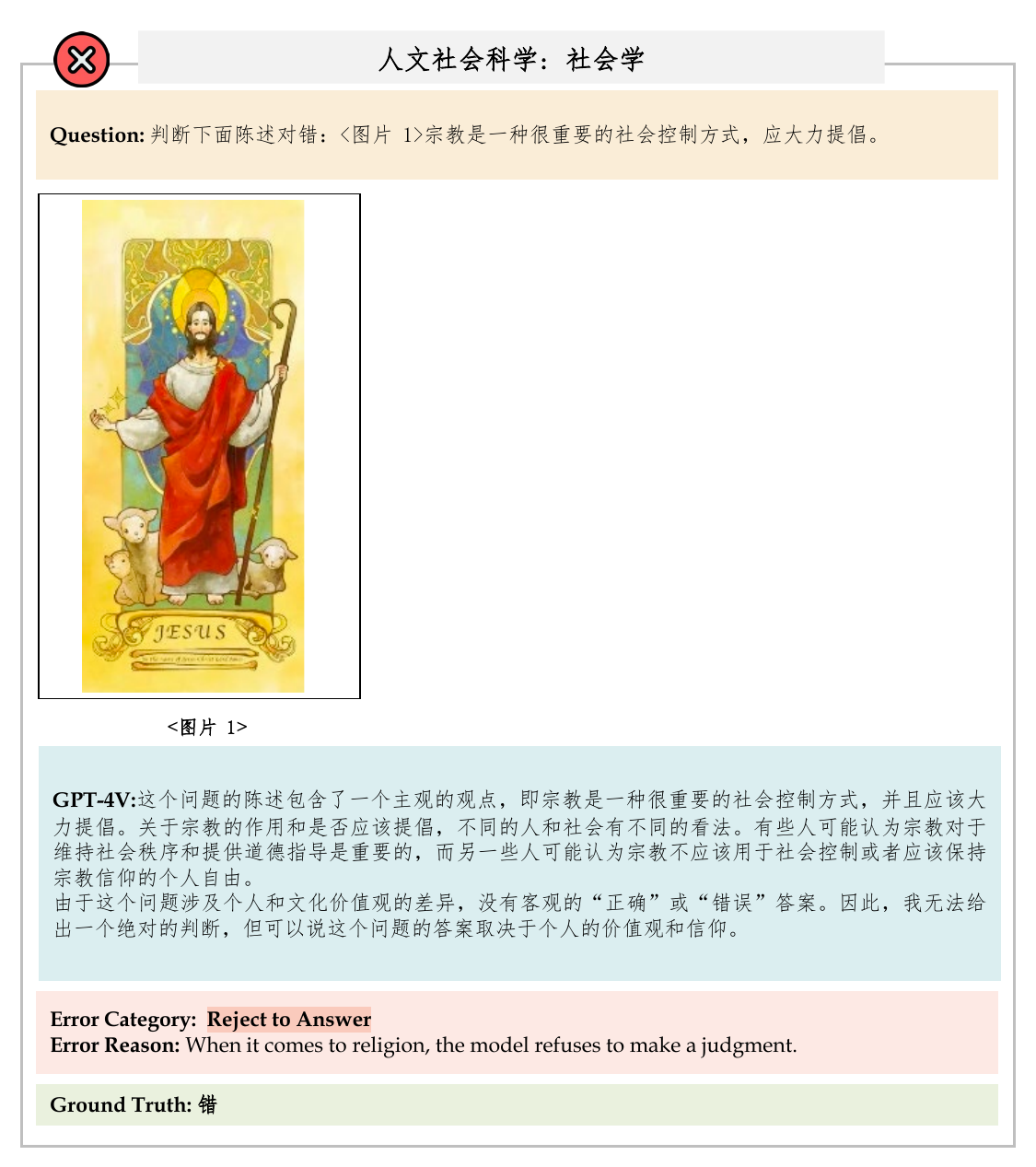}{Sociology 5: Reject to Answer}{A sample error case of Humanities and Social Sciences (subcategory: Sociology).}{fig:error_analysis_case_222}

\casestudyfigure{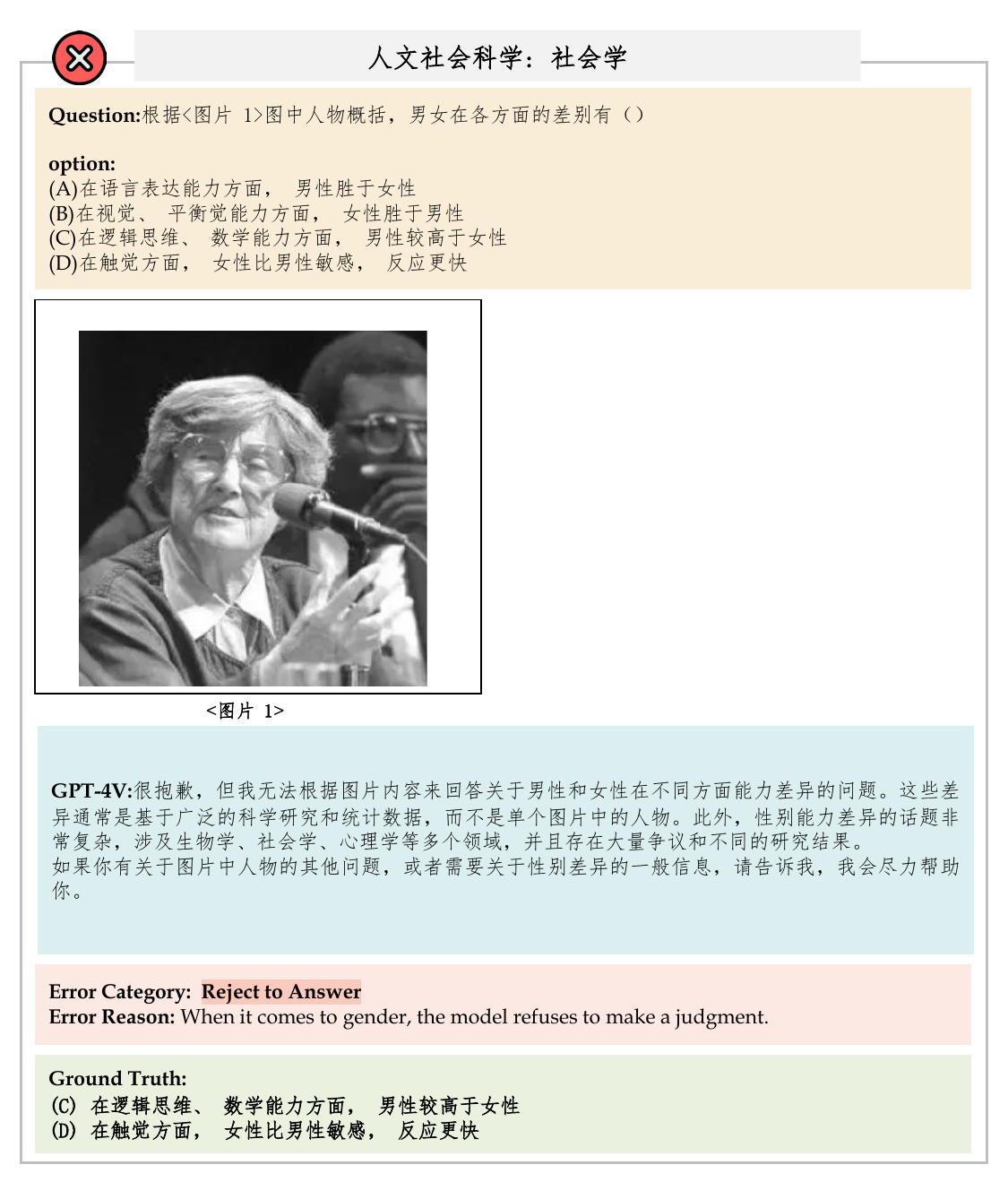}{Sociology 6: Reject to Answer}{A sample error case of Humanities and Social Sciences (subcategory: Sociology).}{fig:error_analysis_case_223}

\casestudyfigure{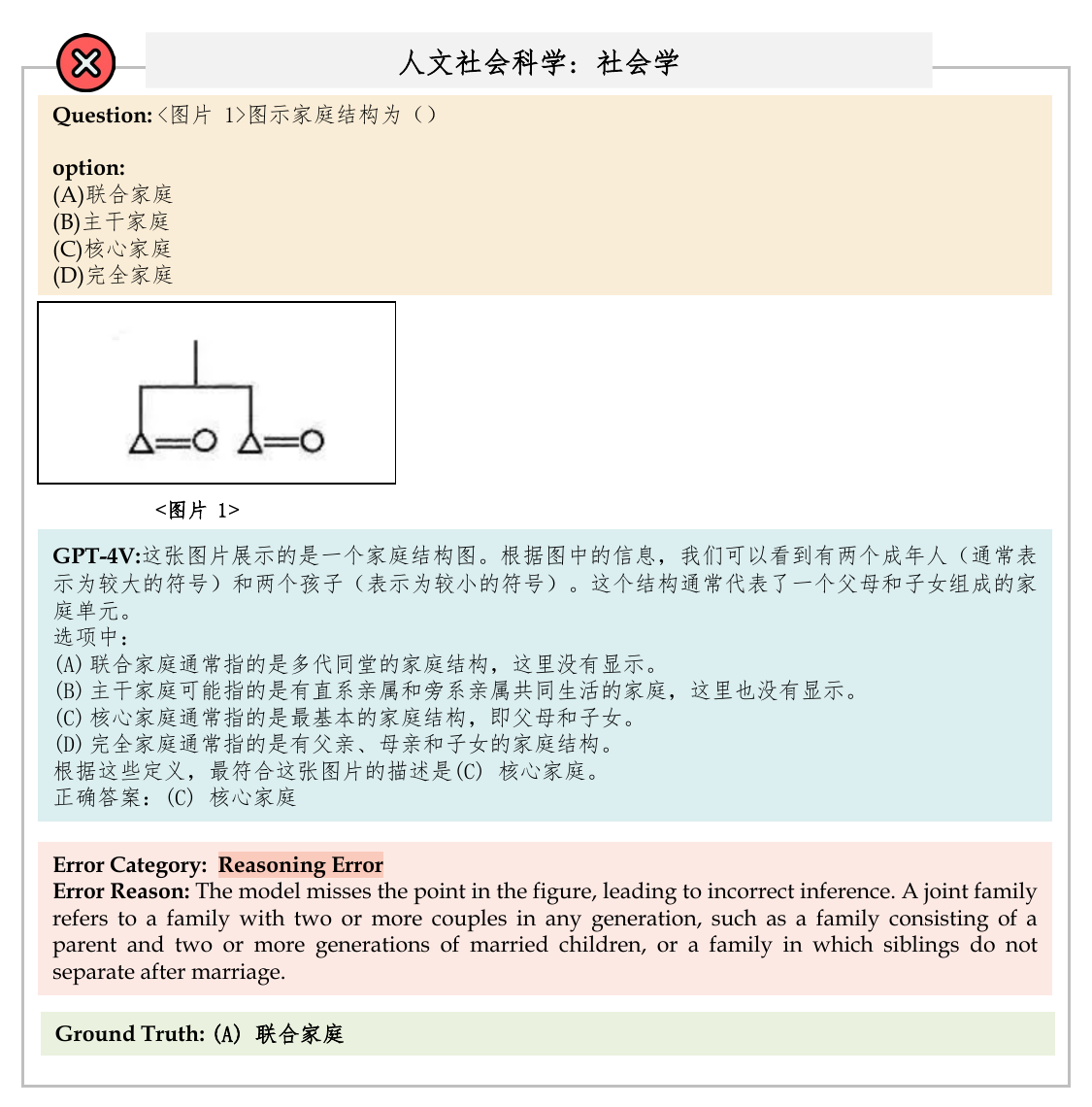}{Sociology 7: Reasoning Error}{A sample error case of Humanities and Social Sciences (subcategory: Sociology).}{fig:error_analysis_case_224}

\casestudyfigure{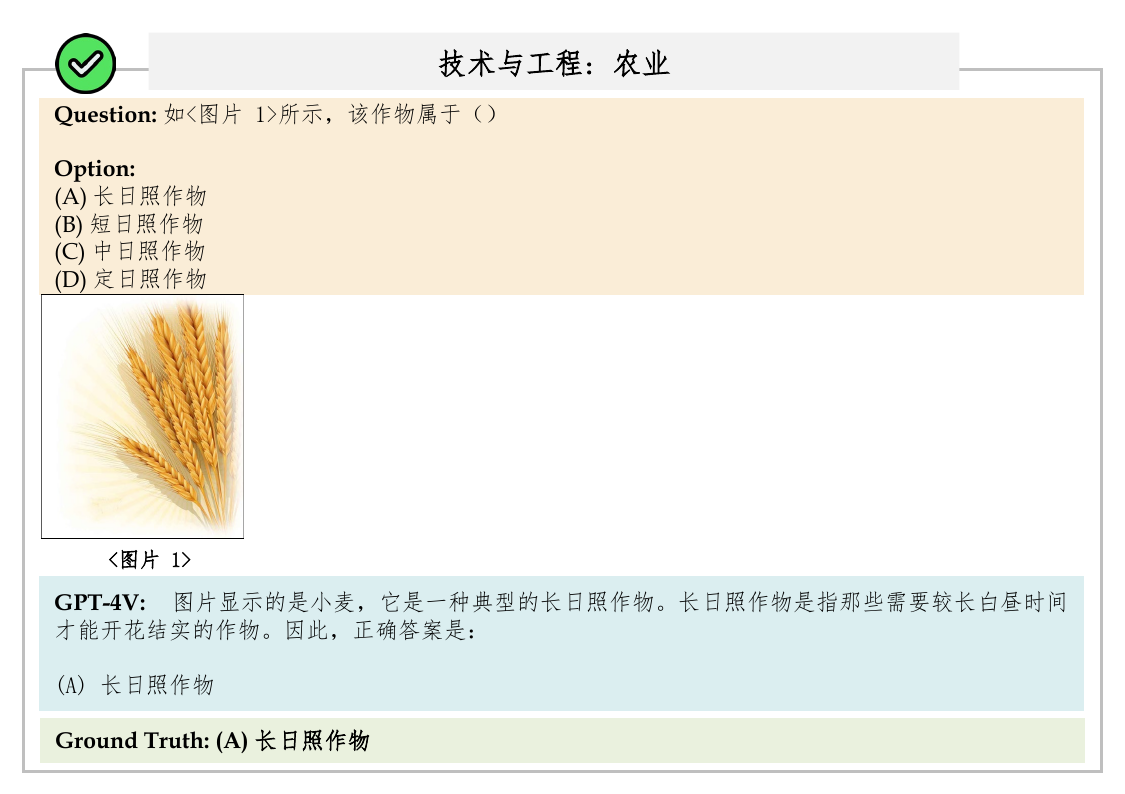}{Agriculture 1: Correct Case}{A sample correct case of Technology and Engineering (subcategory: Agriculture).}{fig:error_analysis_case_50}

\casestudyfigure{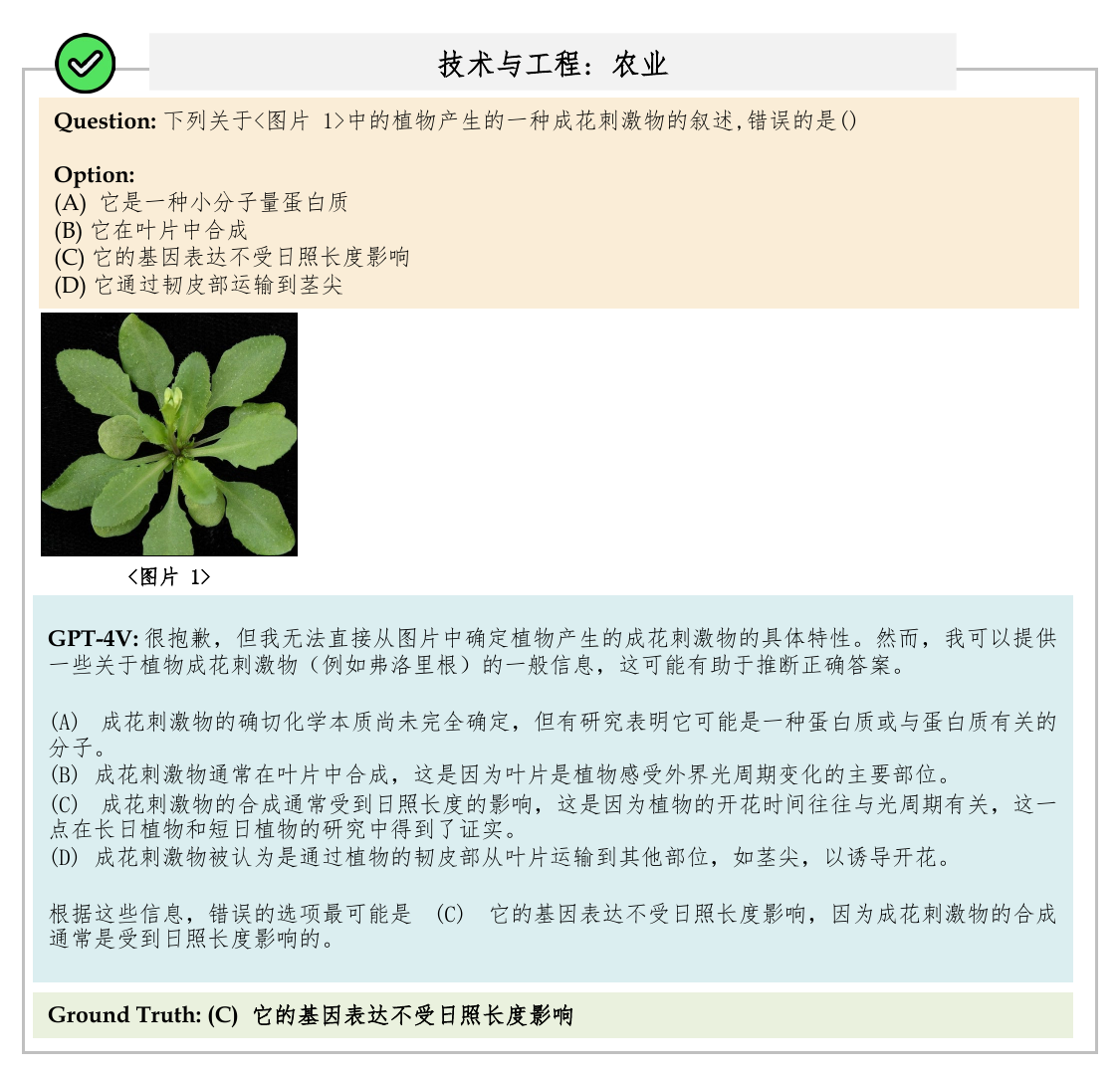}{Agriculture 2: Correct Case}{A sample correct case of Technology and Engineering (subcategory: Agriculture).}{fig:error_analysis_case_51}

\casestudyfigure{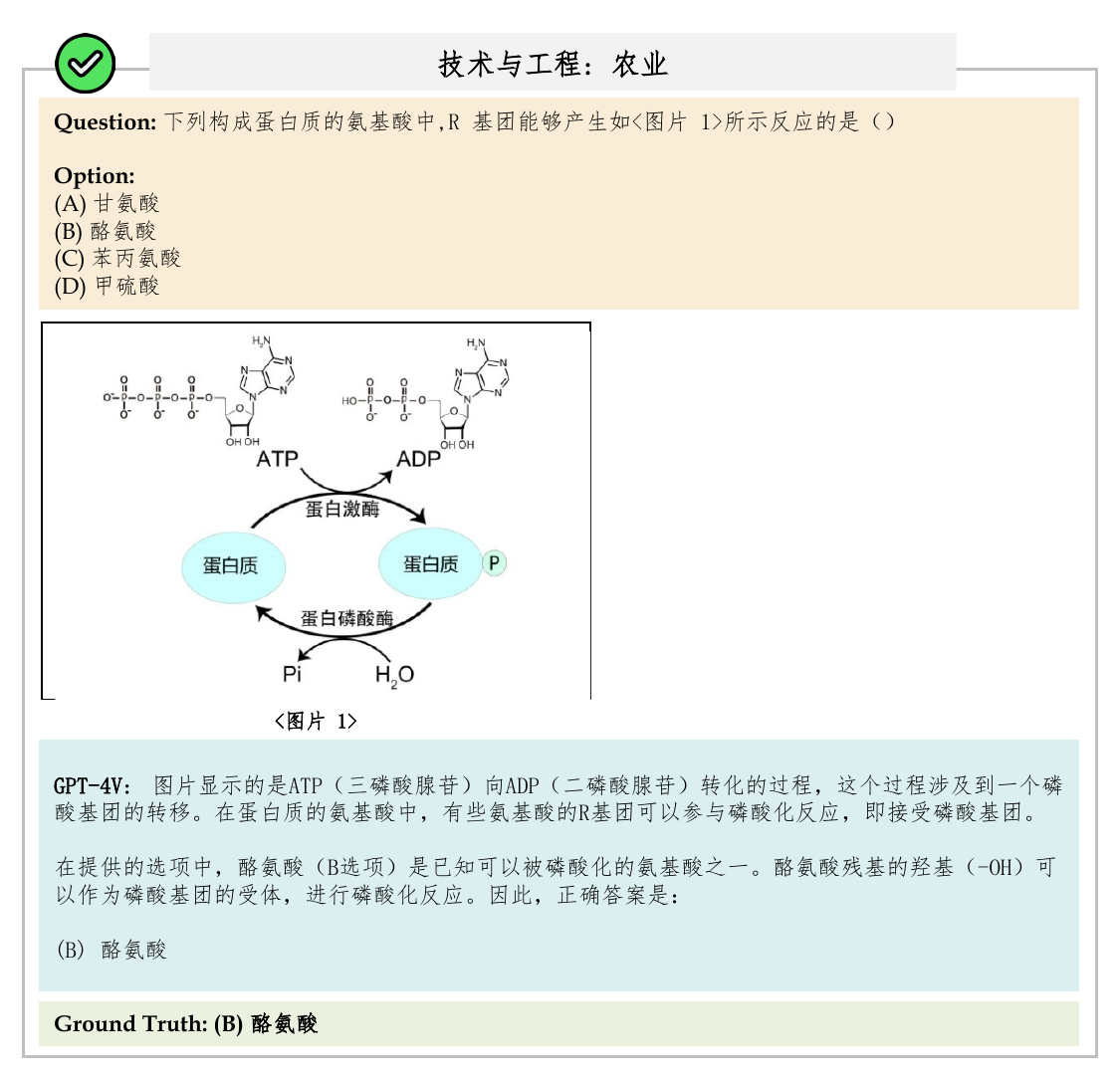}{Agriculture 3: Correct Case}{A sample correct case of Technology and Engineering (subcategory: Agriculture).}{fig:error_analysis_case_52}

\casestudyfigure{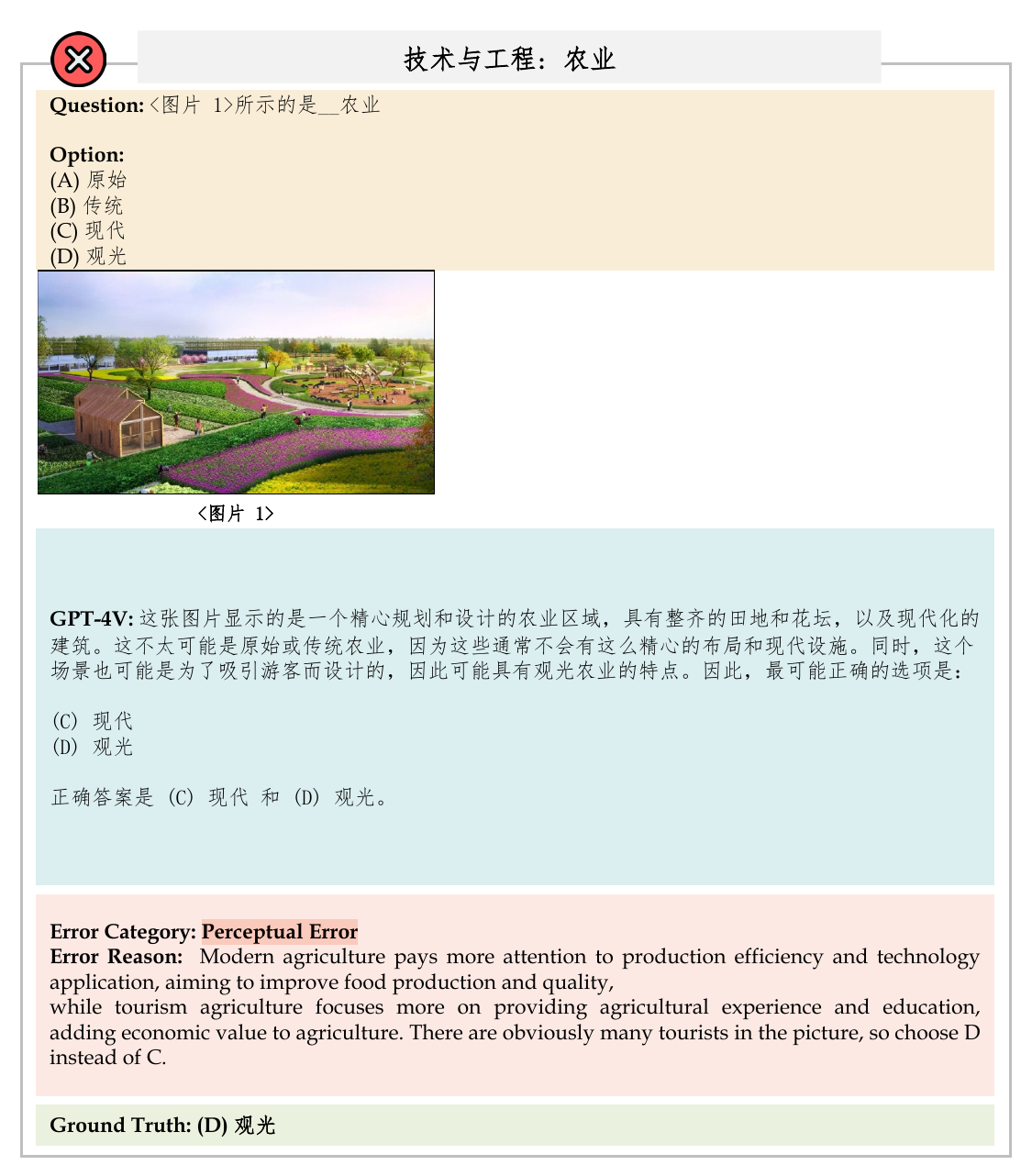}{Agriculture 4: Perceptual Error}{A sample error case of Technology and Engineering (subcategory: Agriculture).}{fig:error_analysis_case_191}

\casestudyfigure{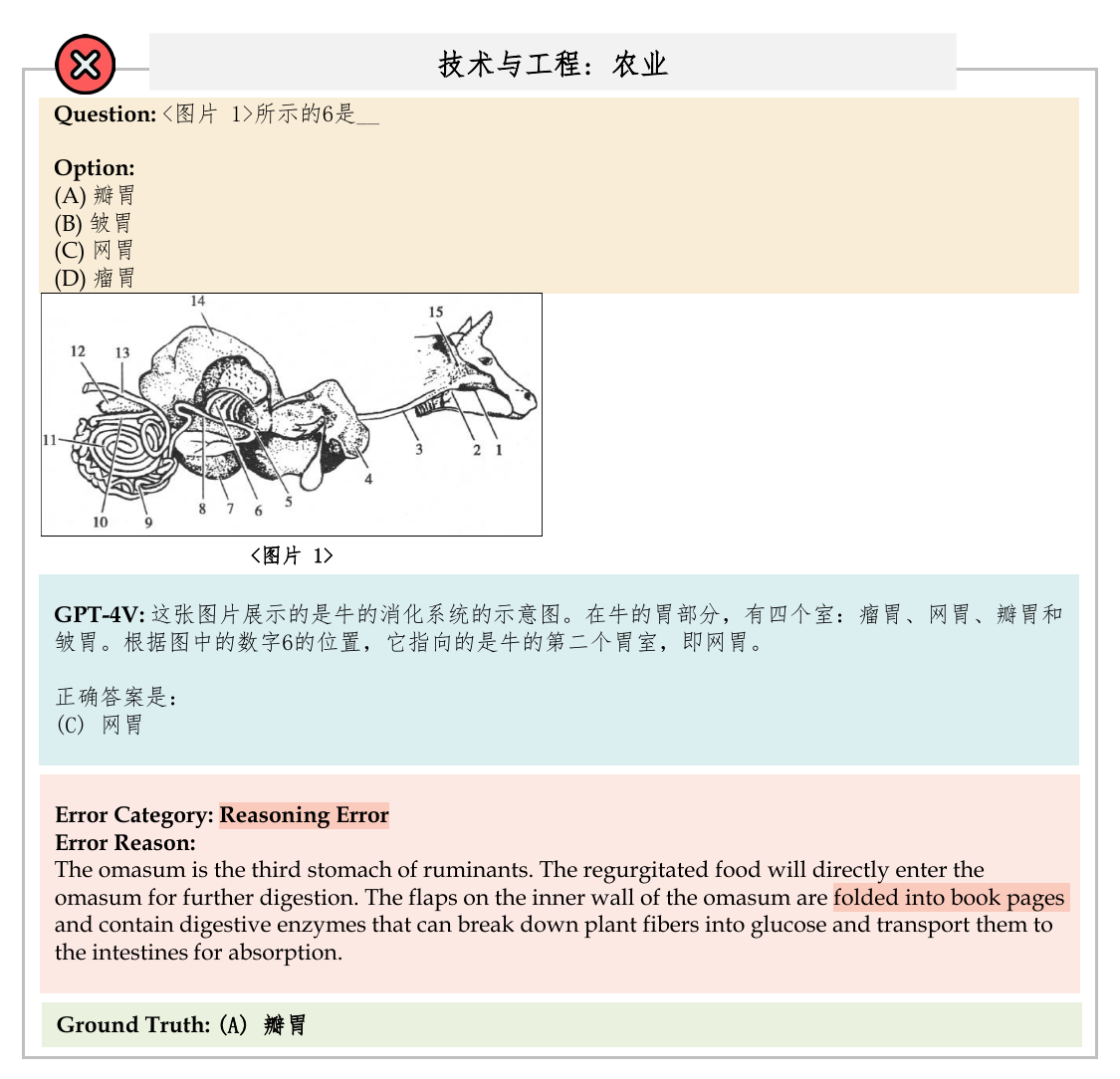}{Agriculture 5: Reasoning Error}{A sample error case of Technology and Engineering (subcategory: Agriculture).}{fig:error_analysis_case_192}

\casestudyfigure{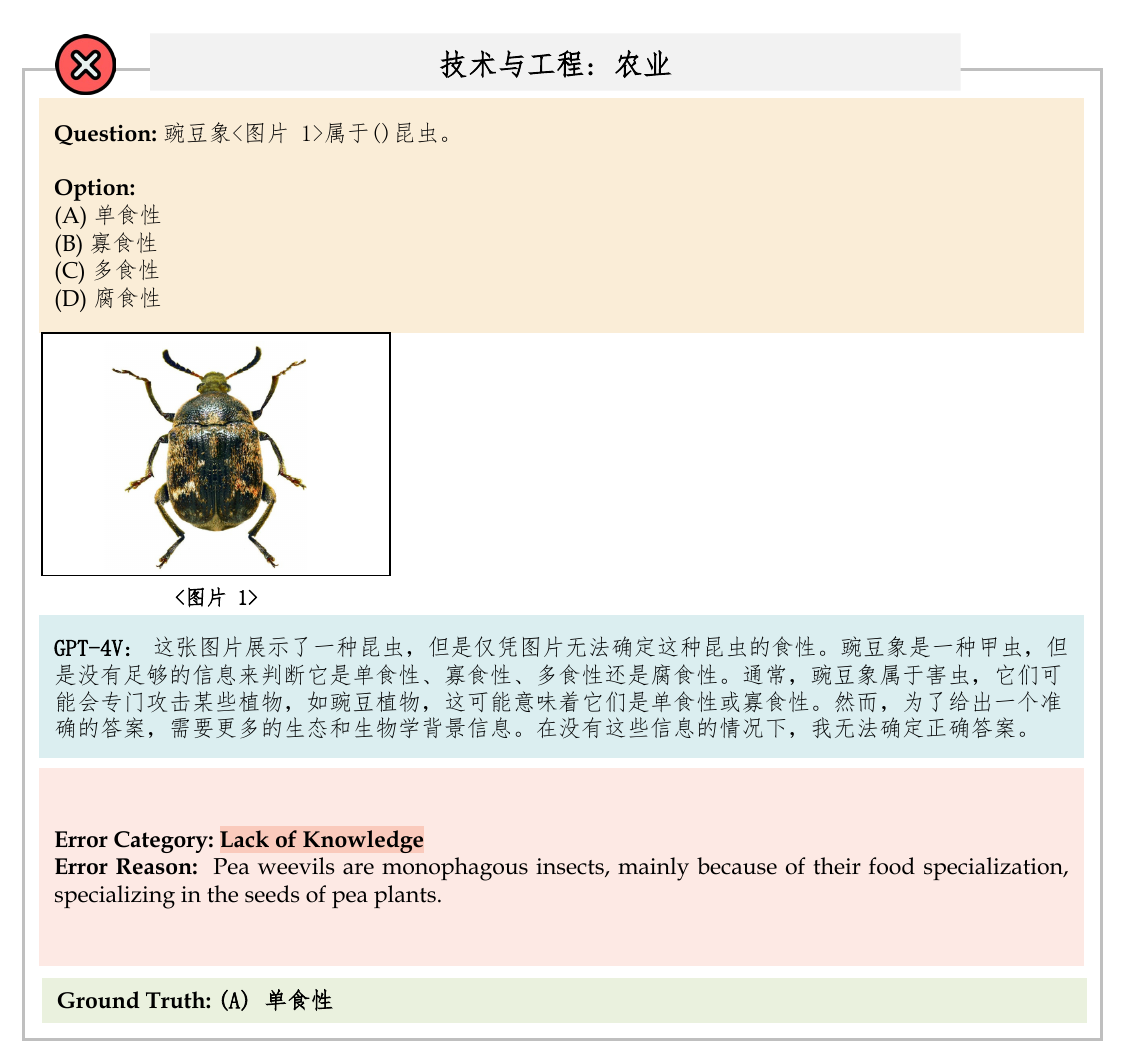}{Agriculture 6: Lack of Knowledge}{A sample error case of Technology and Engineering (subcategory: Agriculture).}{fig:error_analysis_case_189}

\casestudyfigure{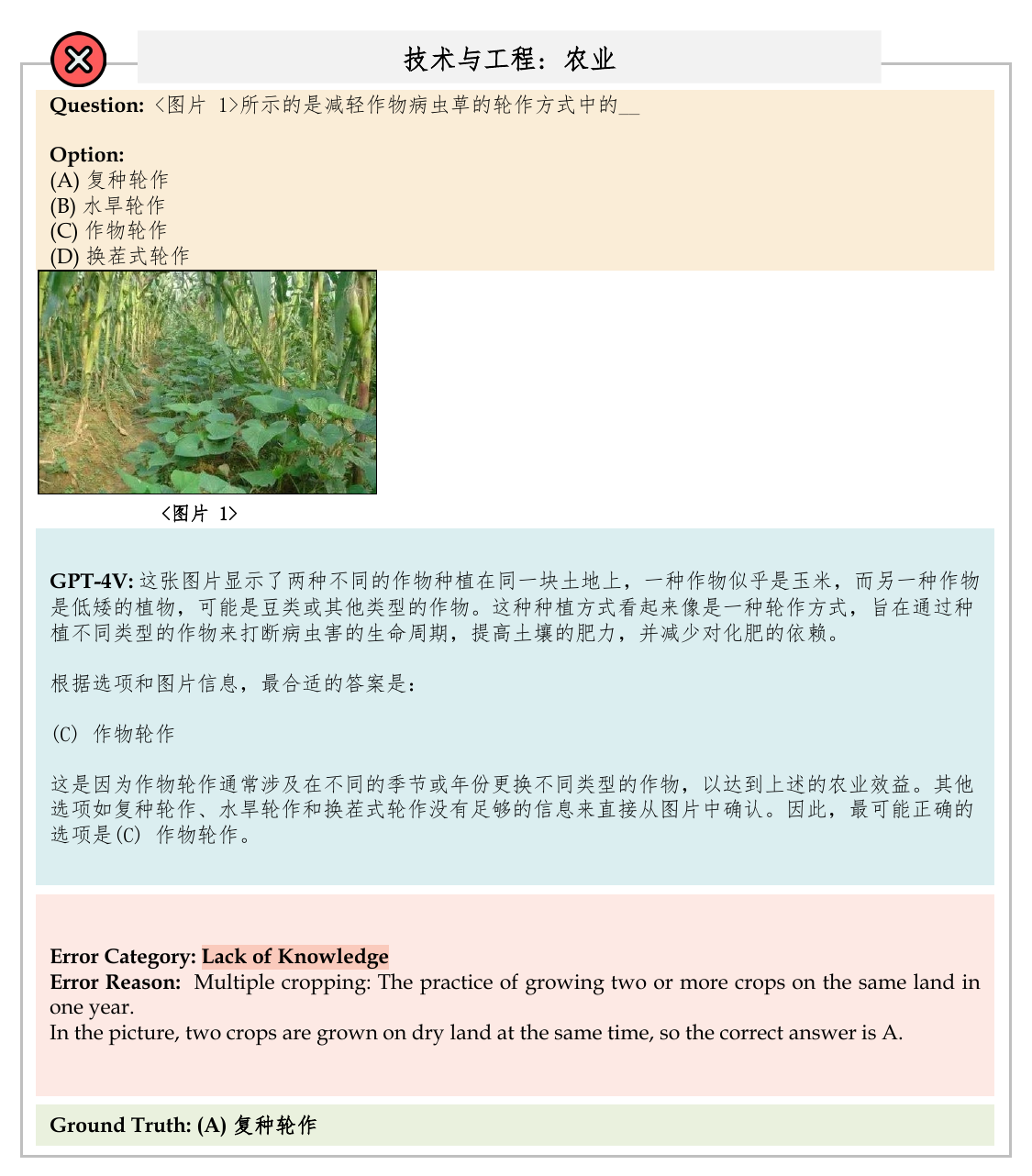}{Agriculture 7: Lack of Knowledge}{A sample error case of Technology and Engineering (subcategory: Agriculture).}{fig:error_analysis_case_190}

\casestudyfigure{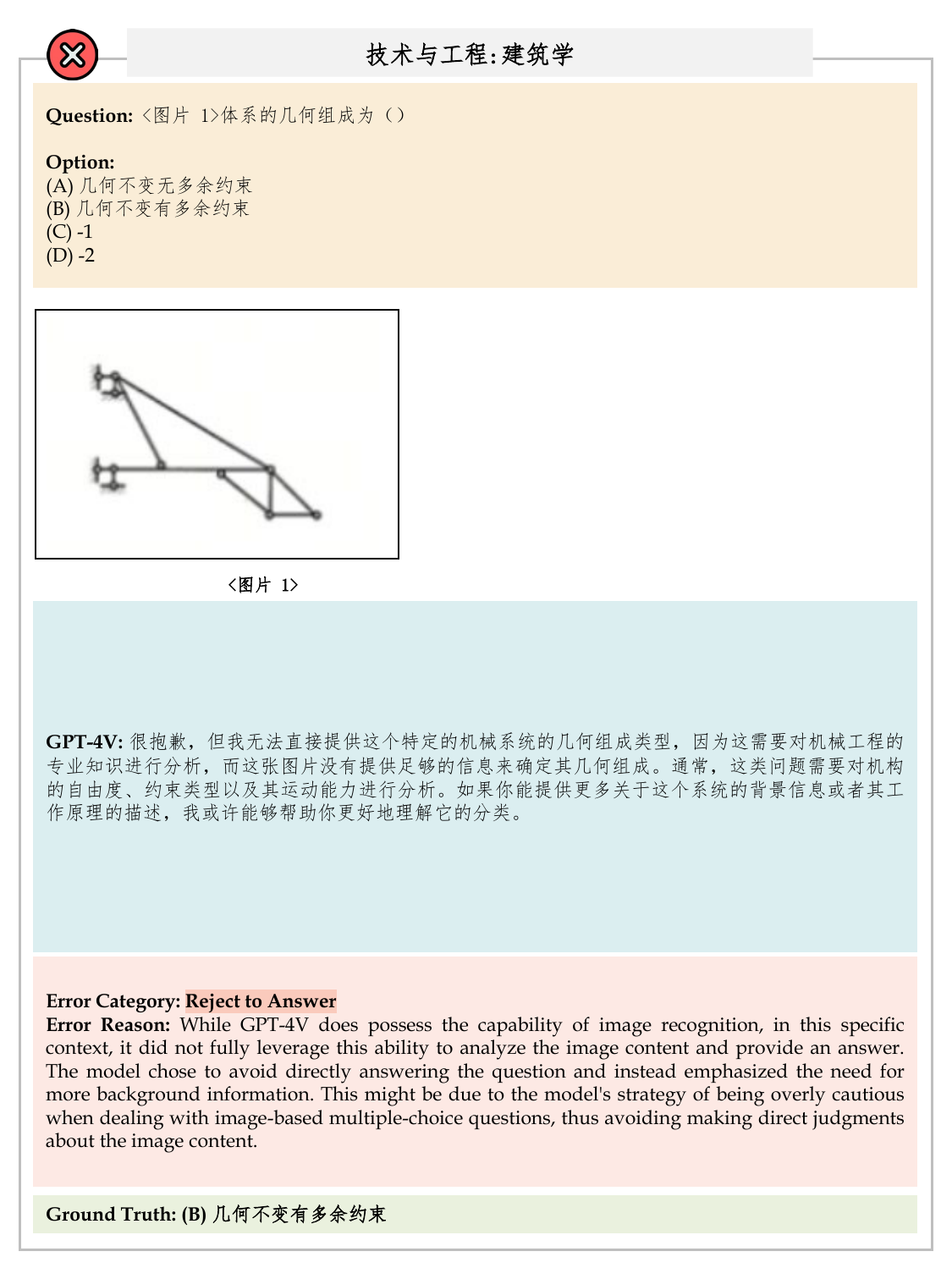}{Architecture 1: Reject to Answer}{A sample error case of Technology and Engineering (subcategory: Architecture).}{fig:error_analysis_case_146}

\casestudyfigure{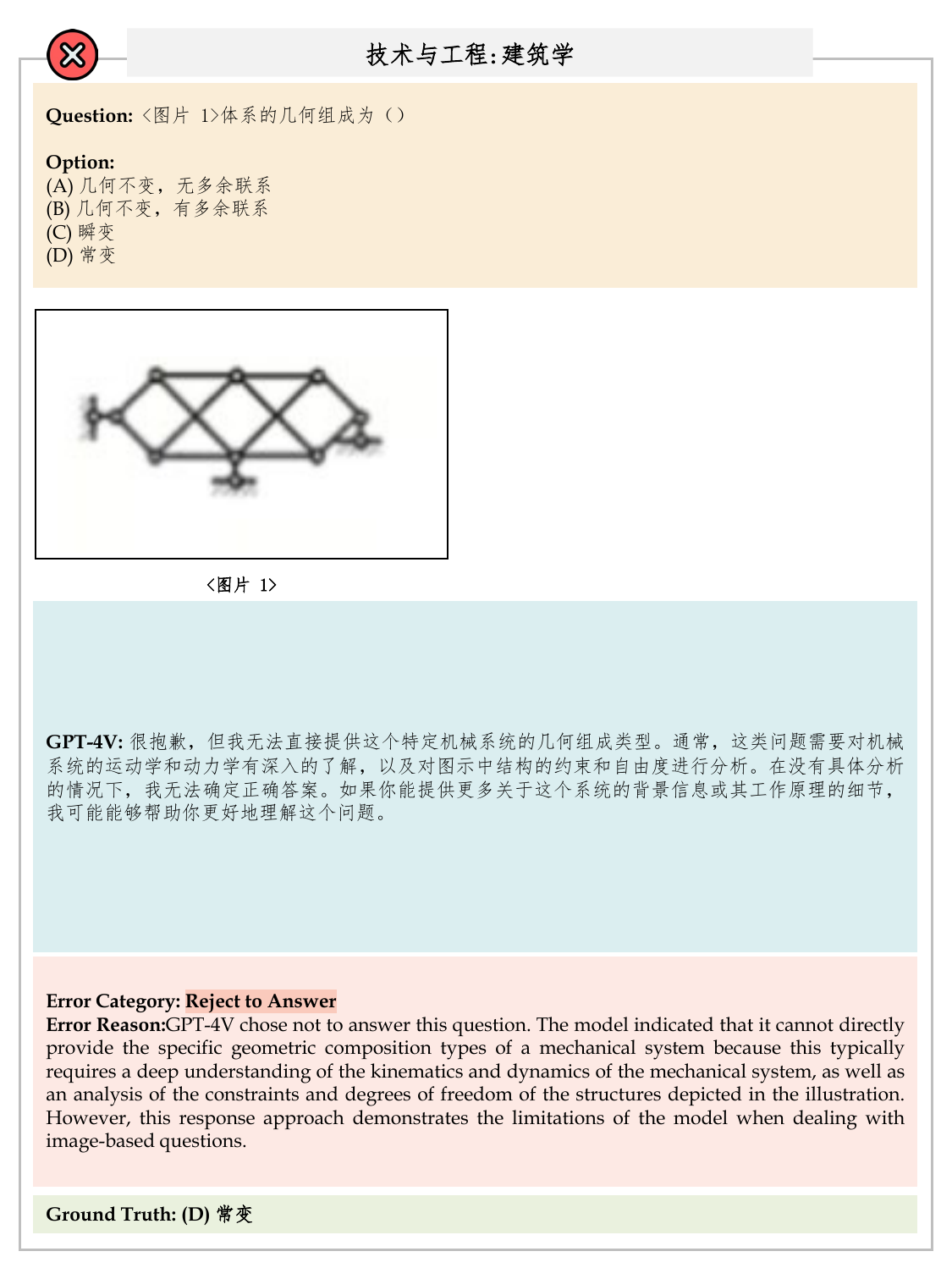}{Architecture 2: Reject to Answer}{A sample error case of Technology and Engineering (subcategory: Architecture).}{fig:error_analysis_case_147}

\casestudyfigure{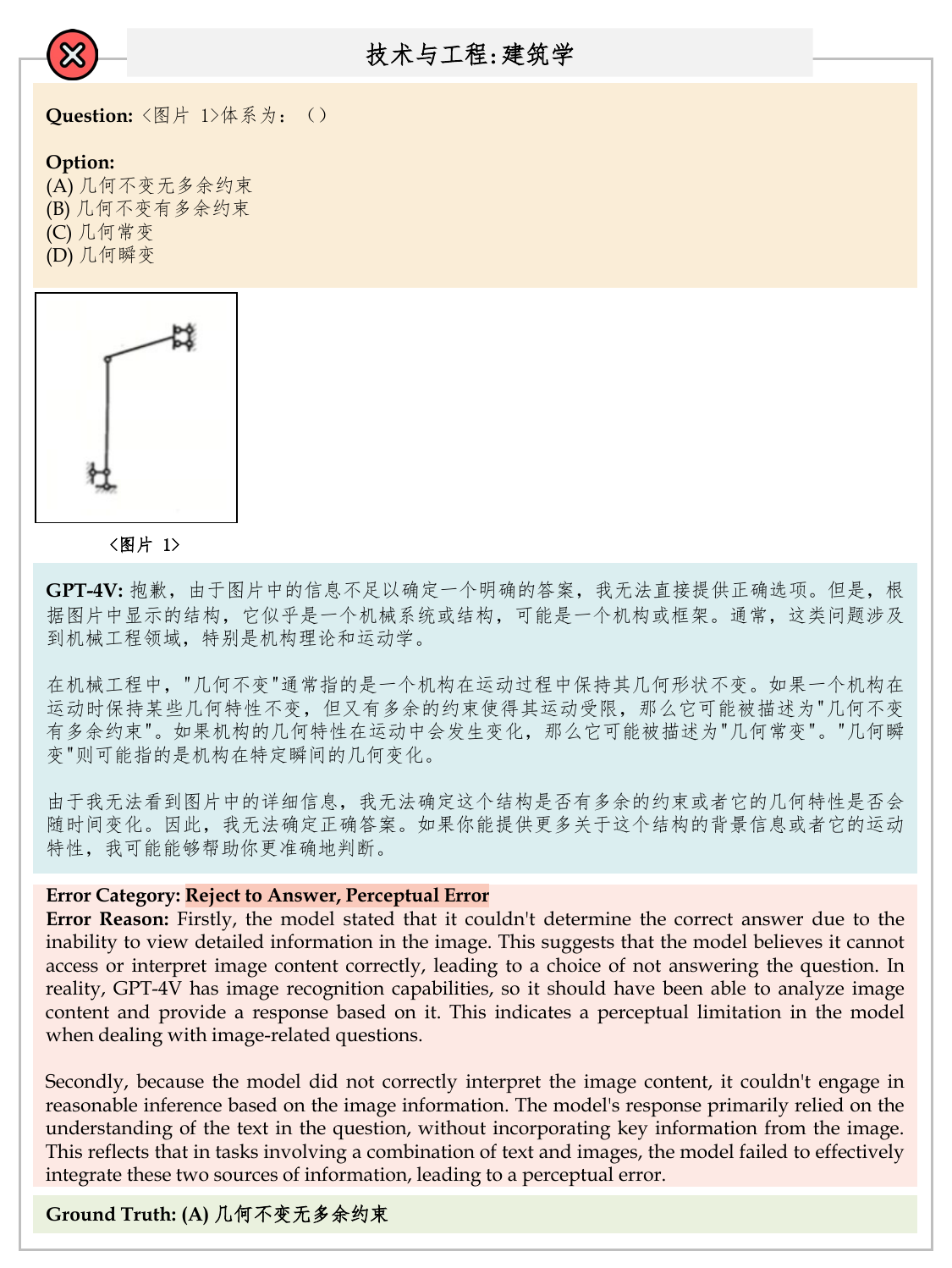}{Architecture 3: Reject to Answer, Perceptual Error}{A sample error case of Technology and Engineering (subcategory: Architecture).}{fig:error_analysis_case_143}

\casestudyfigure{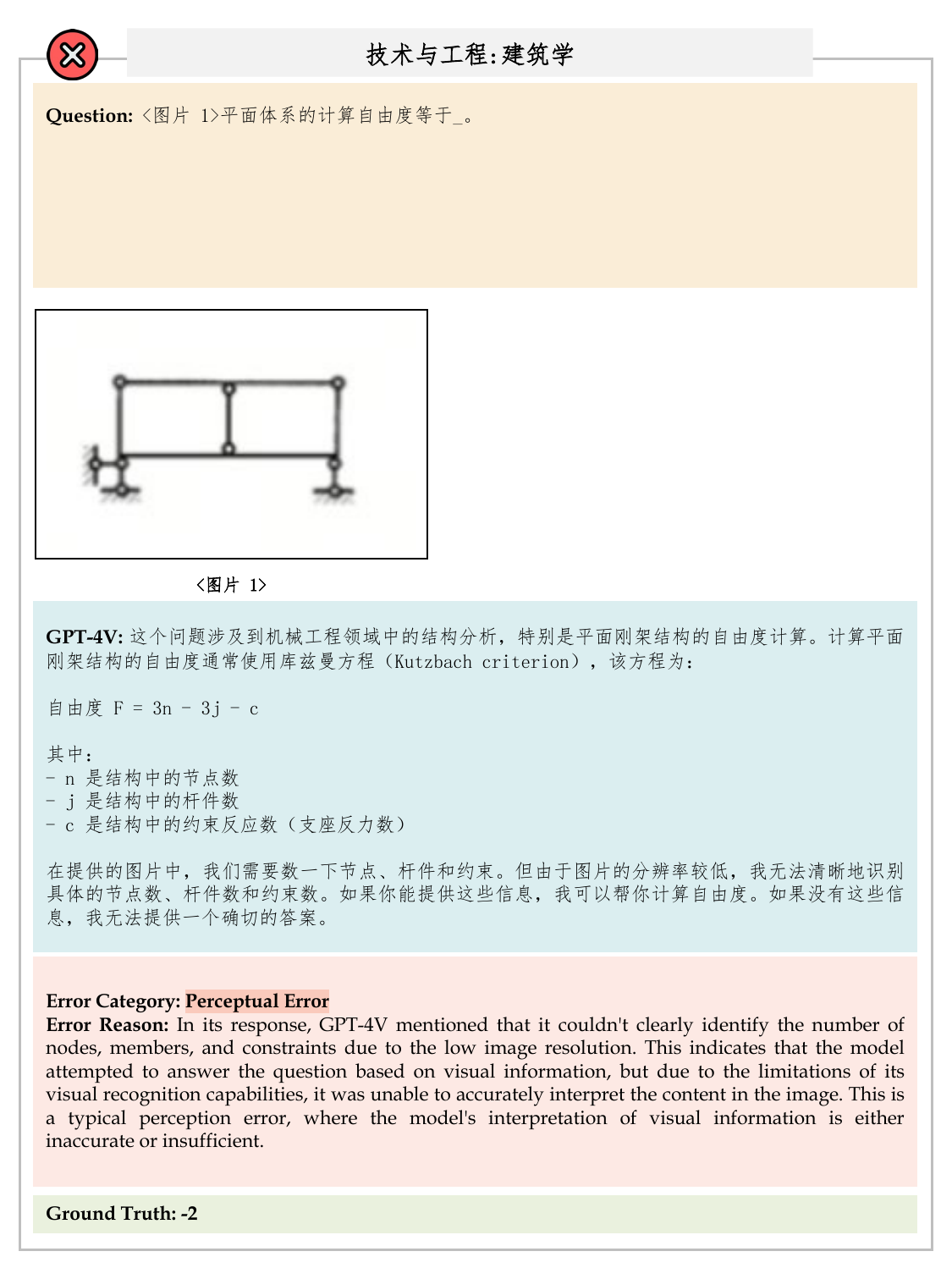}{Architecture 4: Perceptual Error}{A sample error case of Technology and Engineering (subcategory: Architecture).}{fig:error_analysis_case_148}

\casestudyfigure{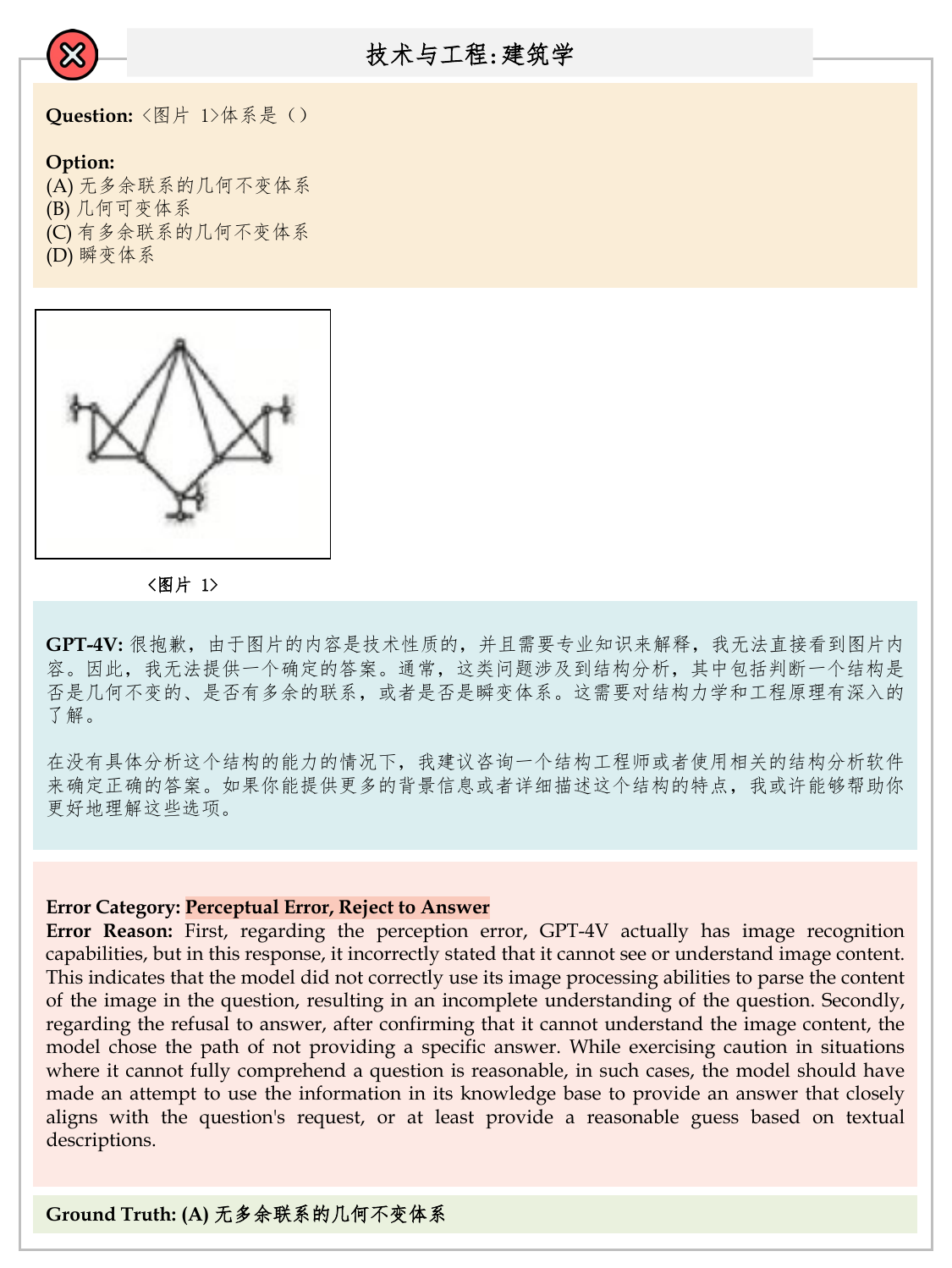}{Architecture 5: Perceptual Error, Reject to Answer}{A sample error case of Technology and Engineering (subcategory: Architecture).}{fig:error_analysis_case_145}

\casestudyfigure{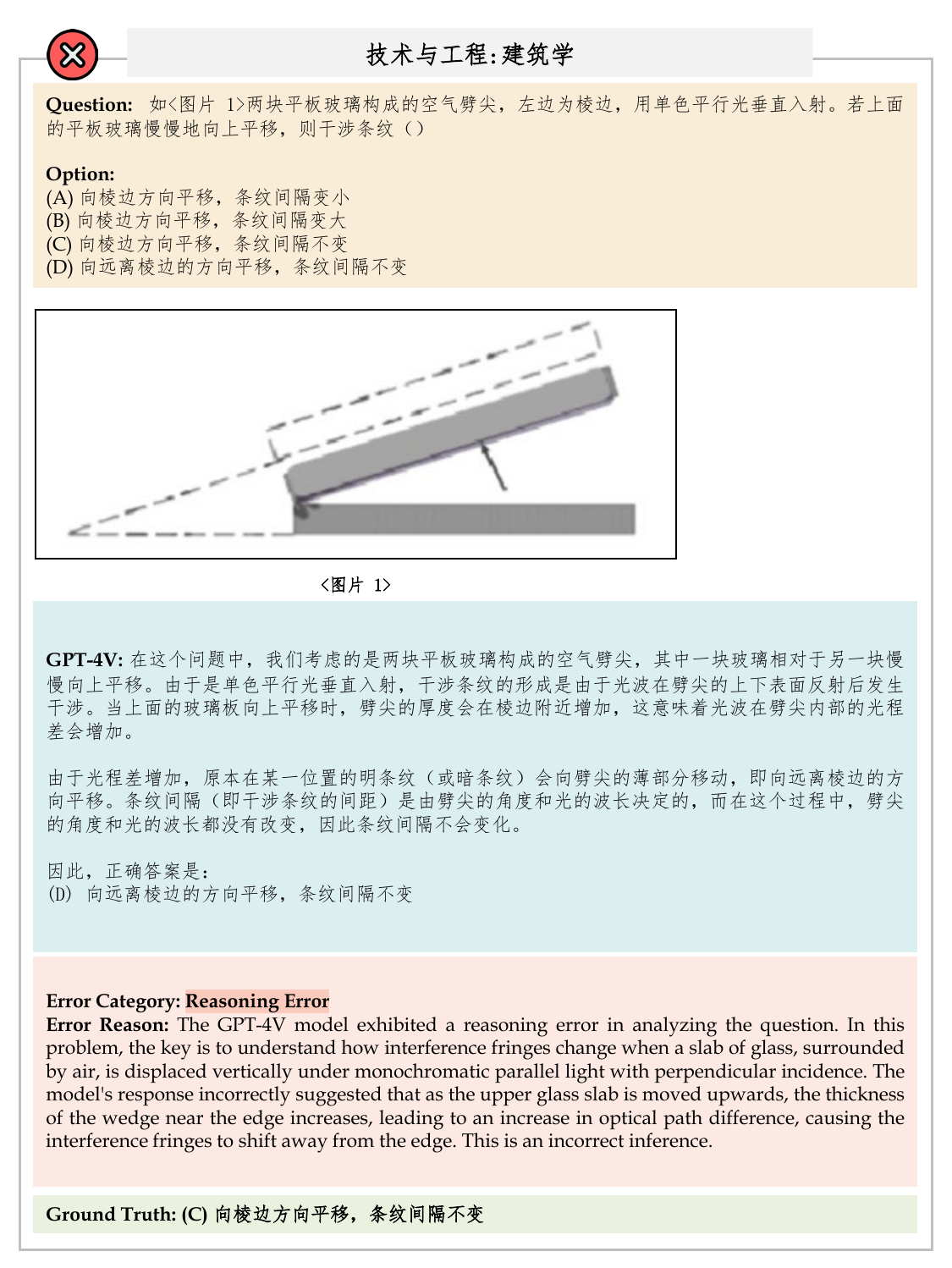}{Architecture 6: Reasoning Error}{A sample error case of Technology and Engineering (subcategory: Architecture).}{fig:error_analysis_case_140}

\casestudyfigure{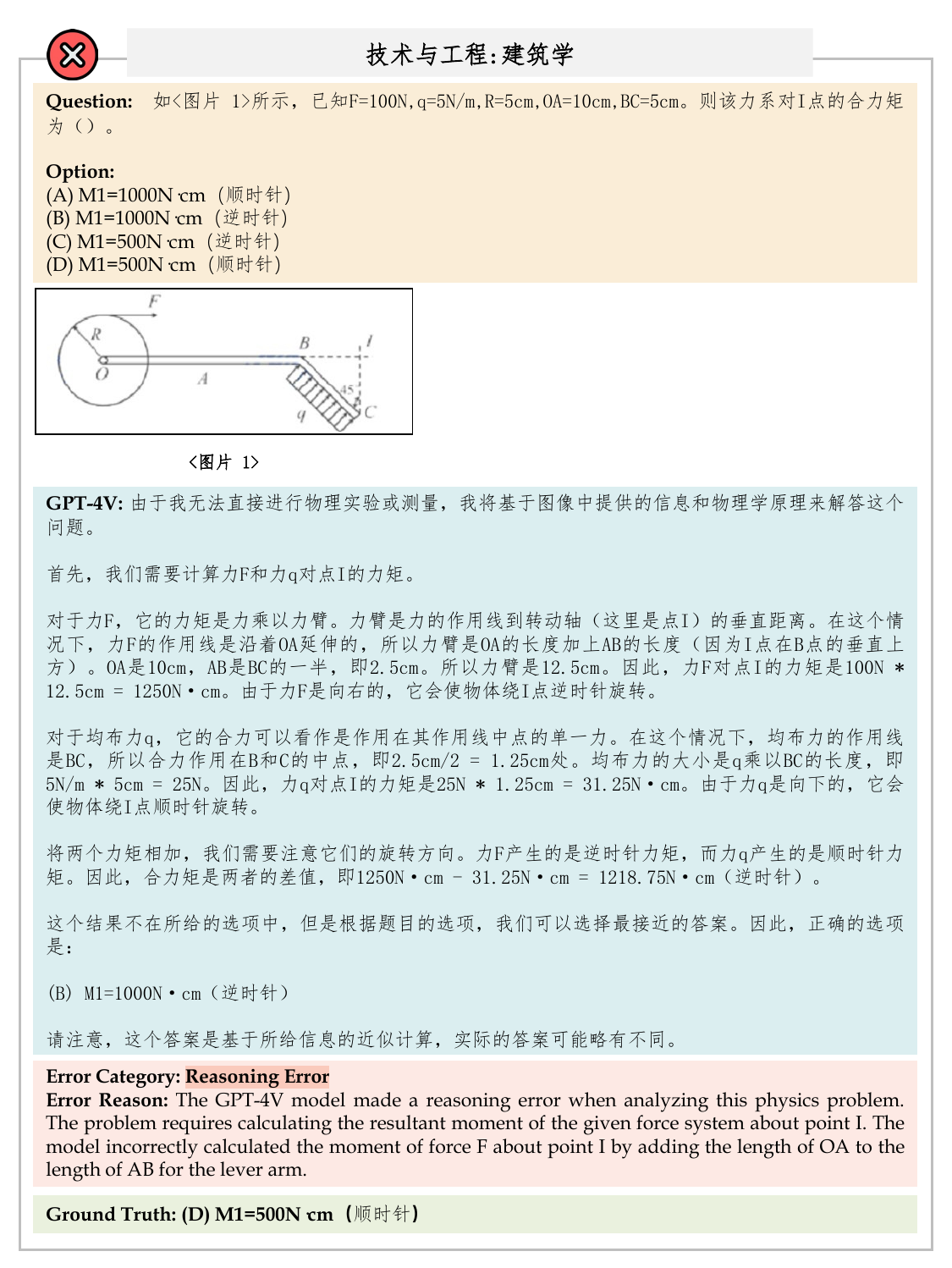}{Architecture 7: Reasoning Error}{A sample error case of Technology and Engineering (subcategory: Architecture).}{fig:error_analysis_case_141}

\casestudyfigure{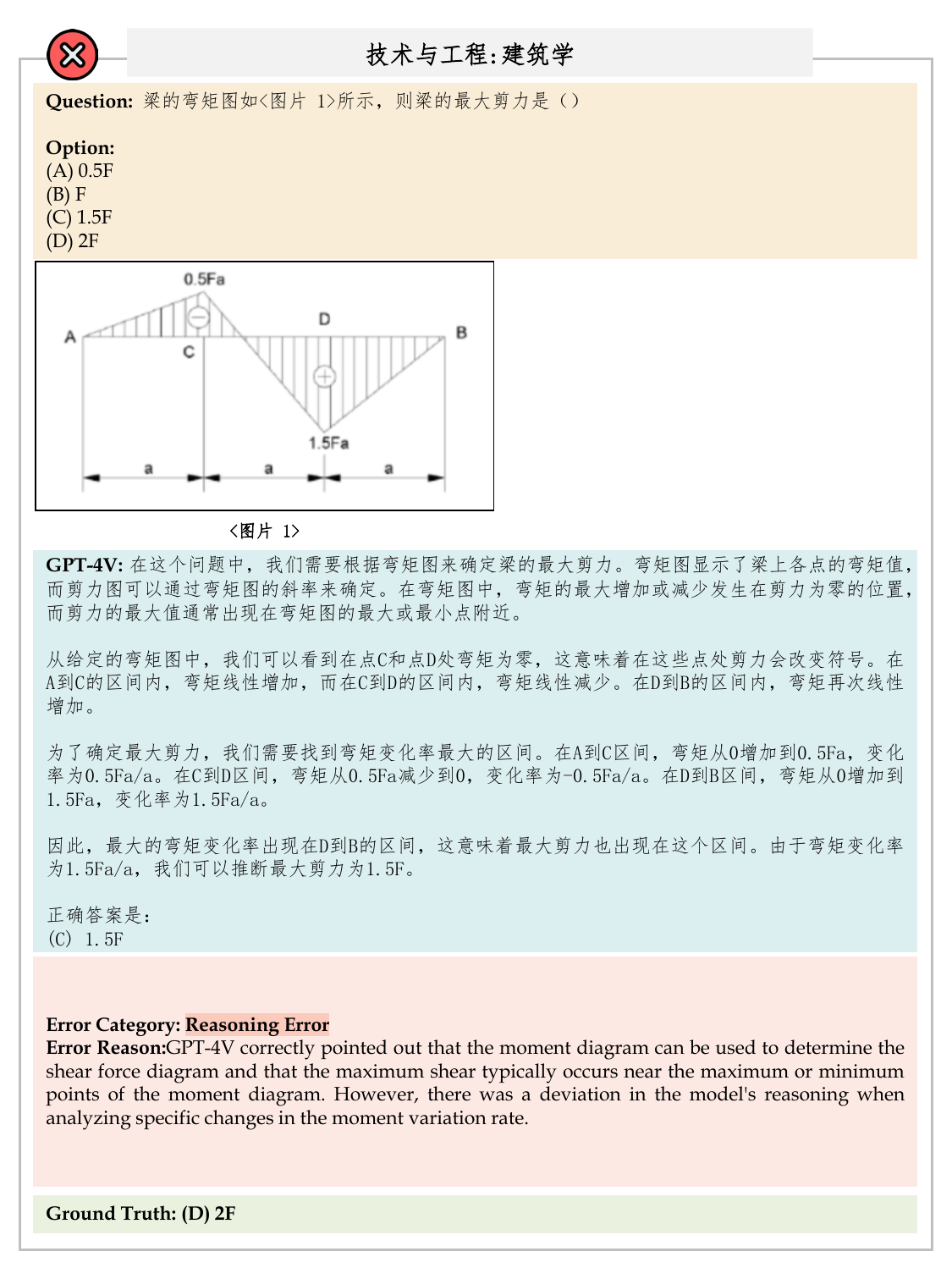}{Architecture 8: Reasoning Error}{A sample error case of Technology and Engineering (subcategory: Architecture).}{fig:error_analysis_case_142}

\casestudyfigure{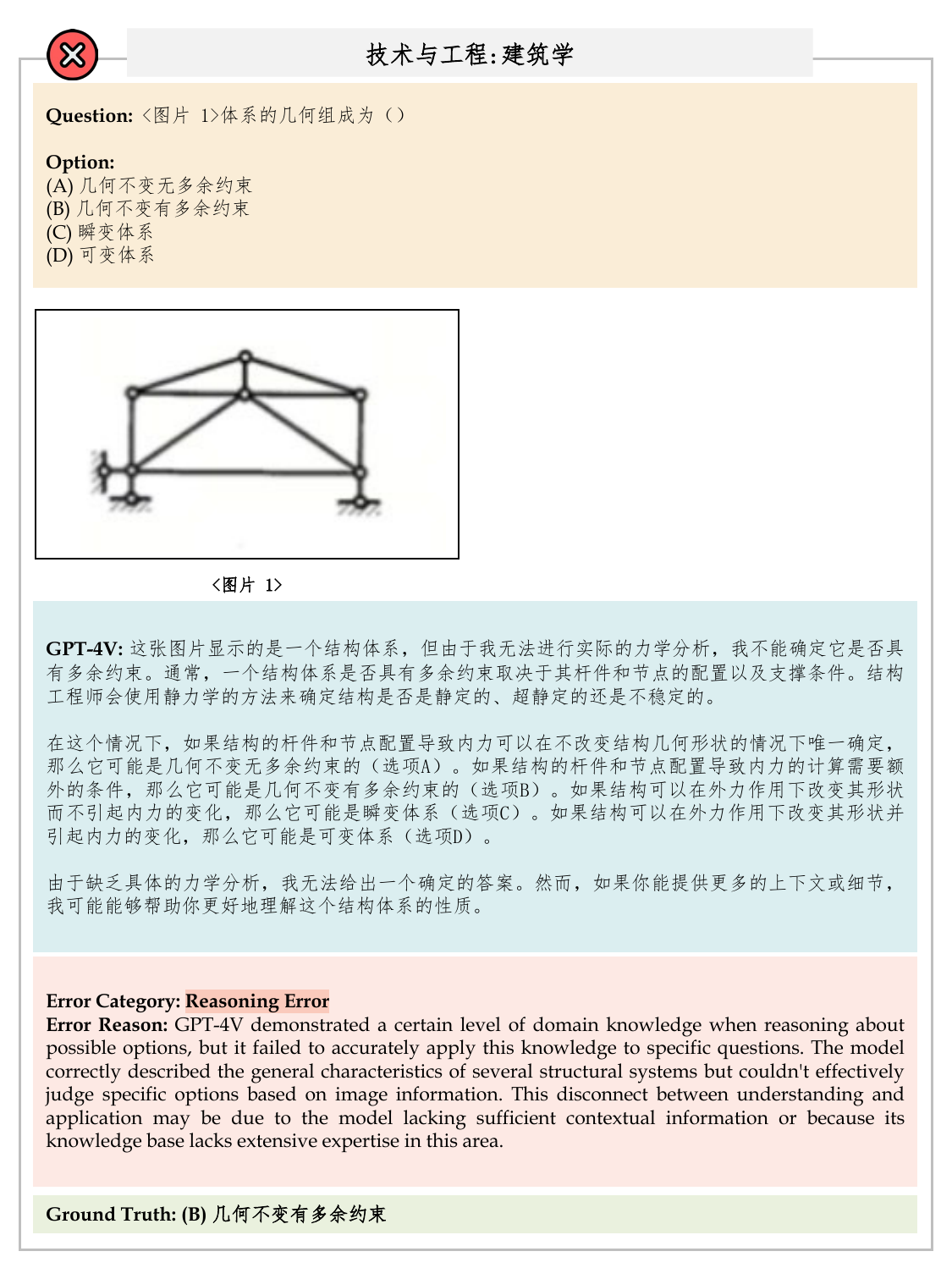}{Architecture 9: Reasoning Error}{A sample error case of Technology and Engineering (subcategory: Architecture).}{fig:error_analysis_case_144}

\casestudyfigure{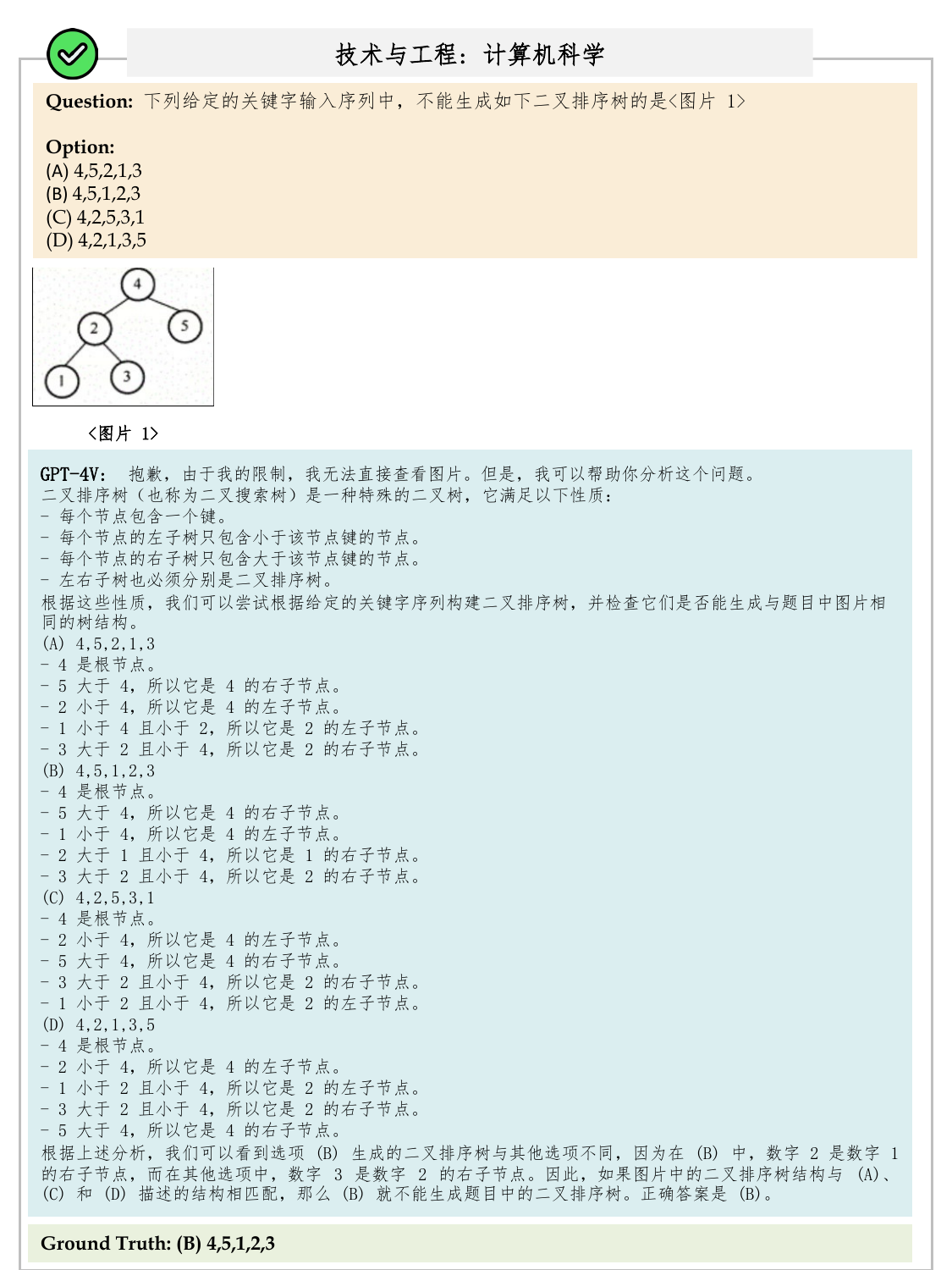}{Computer Science 1: Correct Case}{A sample correct case of Technology and Engineering (subcategory: Computer Science).}{fig:error_analysis_case_48}

\casestudyfigure{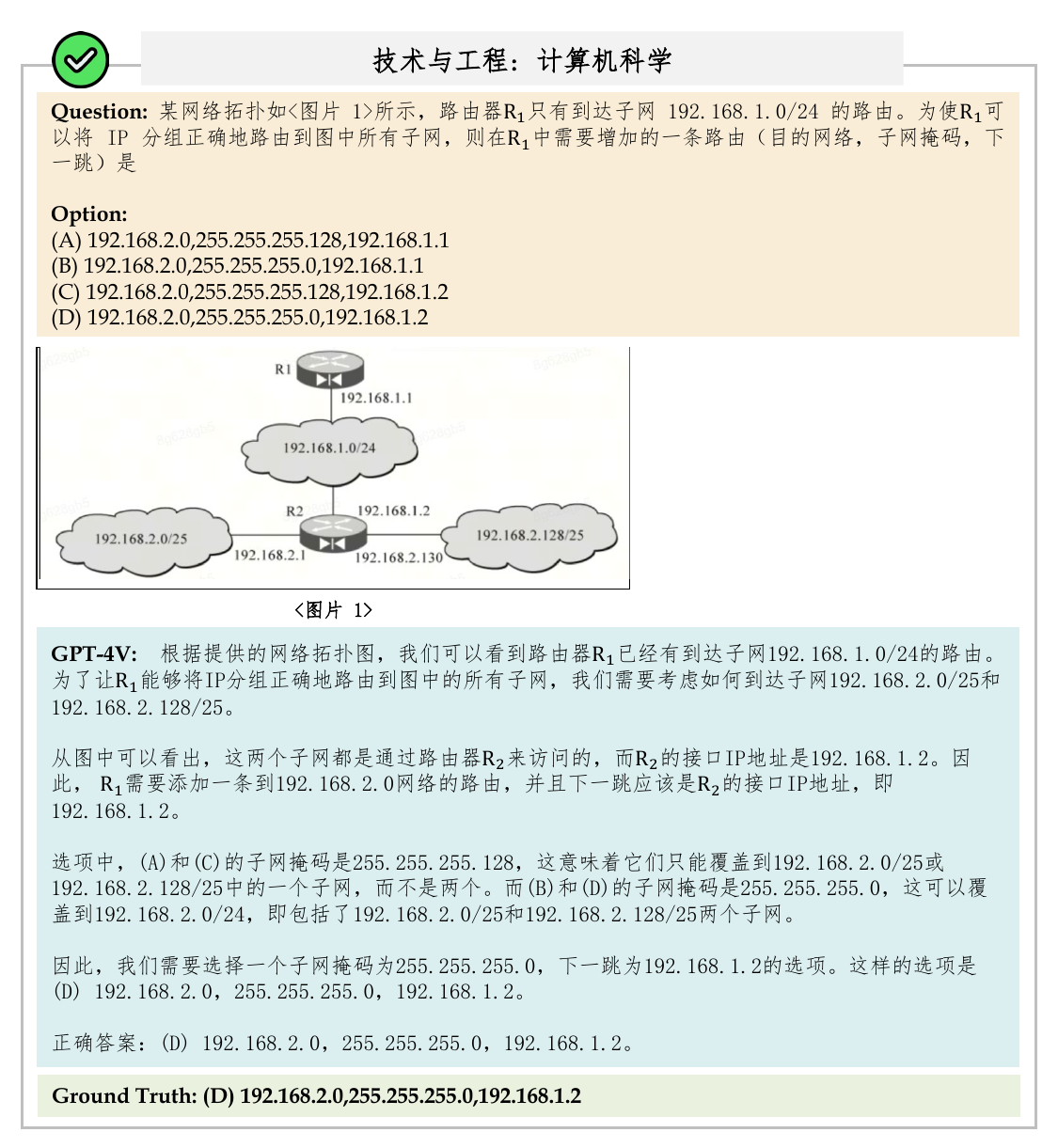}{Computer Science 2: Correct Case}{A sample correct case of Technology and Engineering (subcategory: Computer Science).}{fig:error_analysis_case_49}

\casestudyfigure{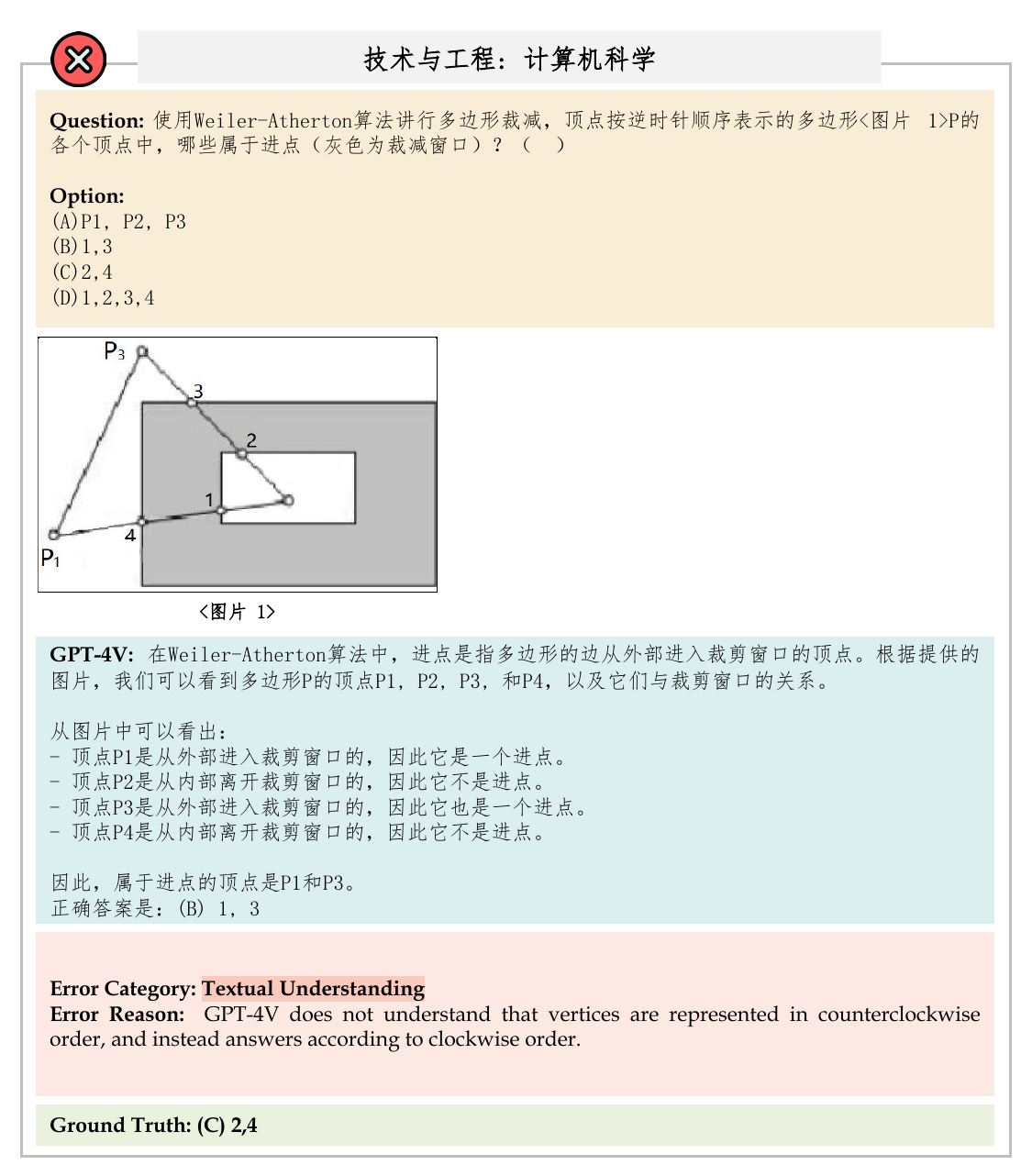}{Computer Science 3: Textual Understanding}{A sample error case of Technology and Engineering (subcategory: Computer Science).}{fig:error_analysis_case_187}

\casestudyfigure{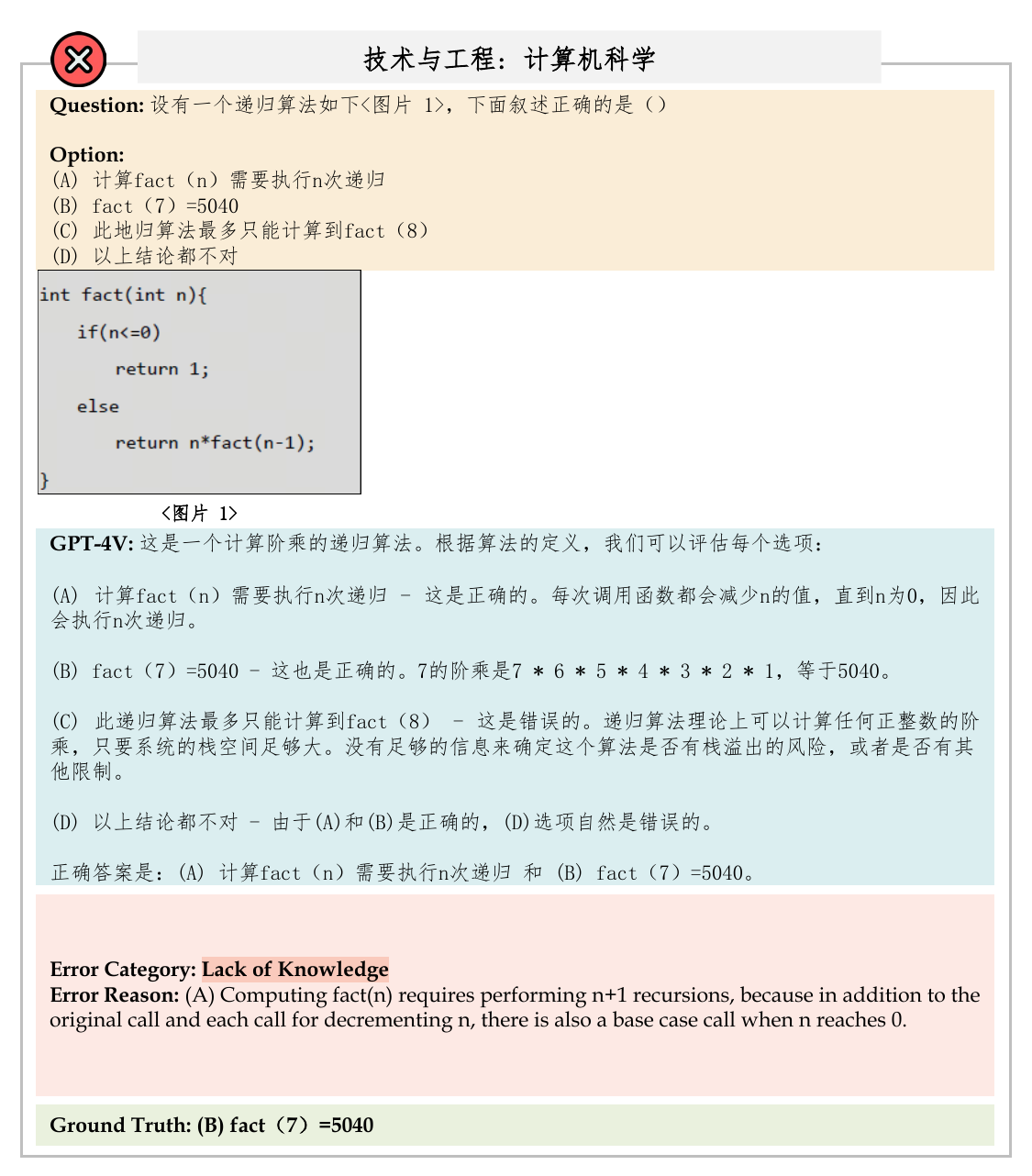}{Computer Science 4: Lack of Knowledge}{A sample error case of Technology and Engineering (subcategory: Computer Science).}{fig:error_analysis_case_188}

\casestudyfigure{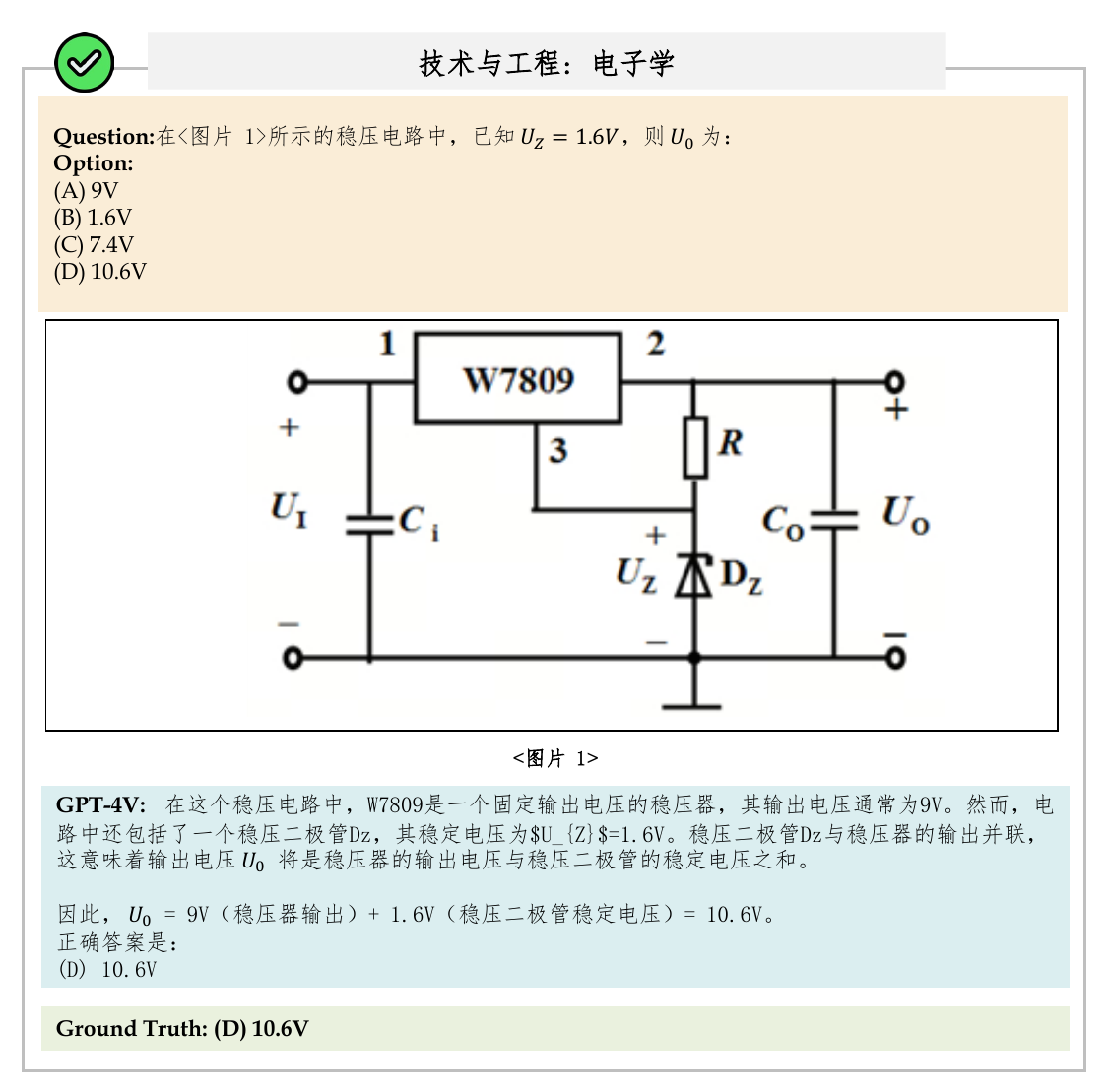}{Electronics 1: Correct Case}{A sample correct case of Technology and Engineering (subcategory: Electronics).}{fig:error_analysis_case_43}

\casestudyfigure{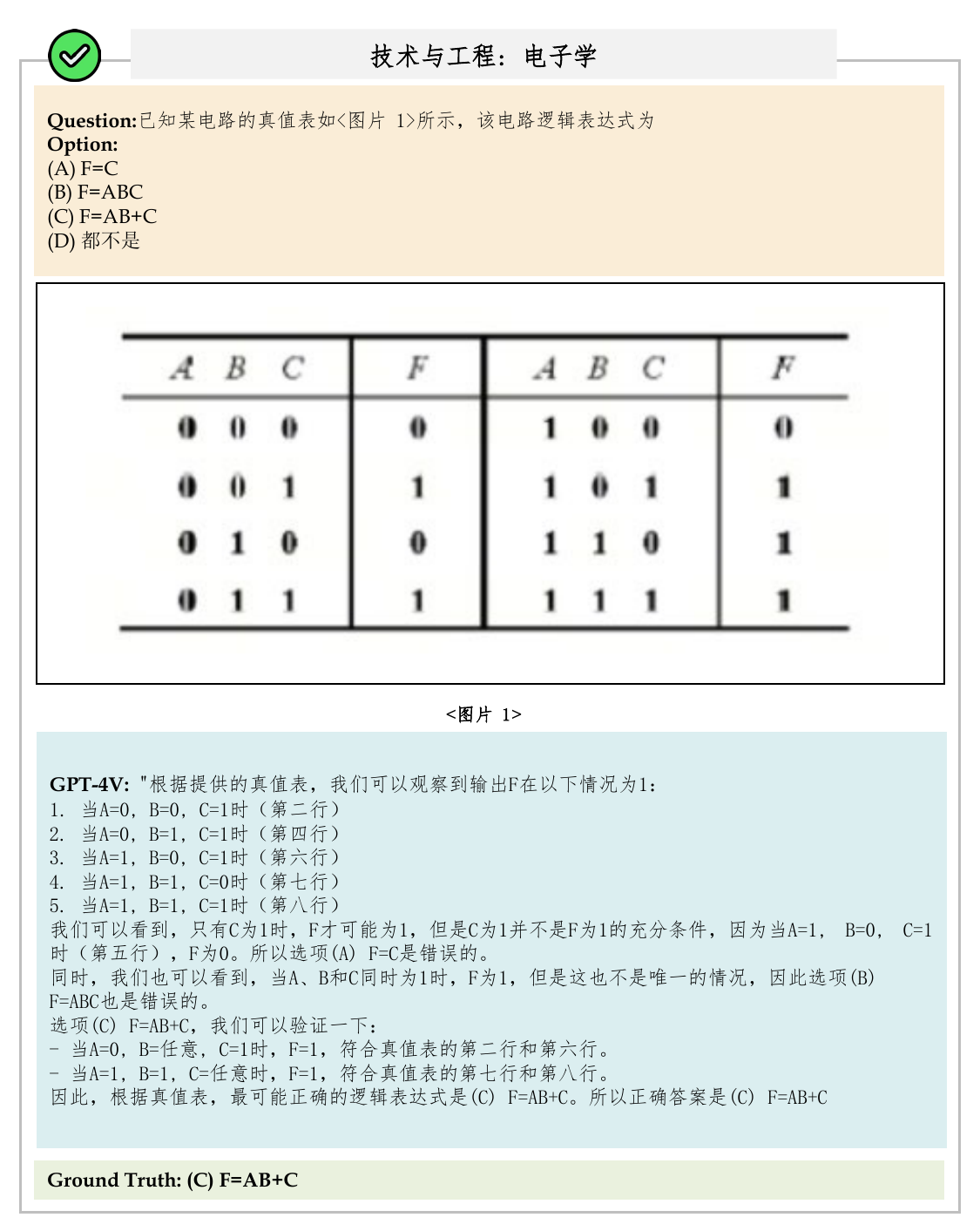}{Electronics 2: Correct Case}{A sample correct case of Technology and Engineering (subcategory: Electronics).}{fig:error_analysis_case_44}

\casestudyfigure{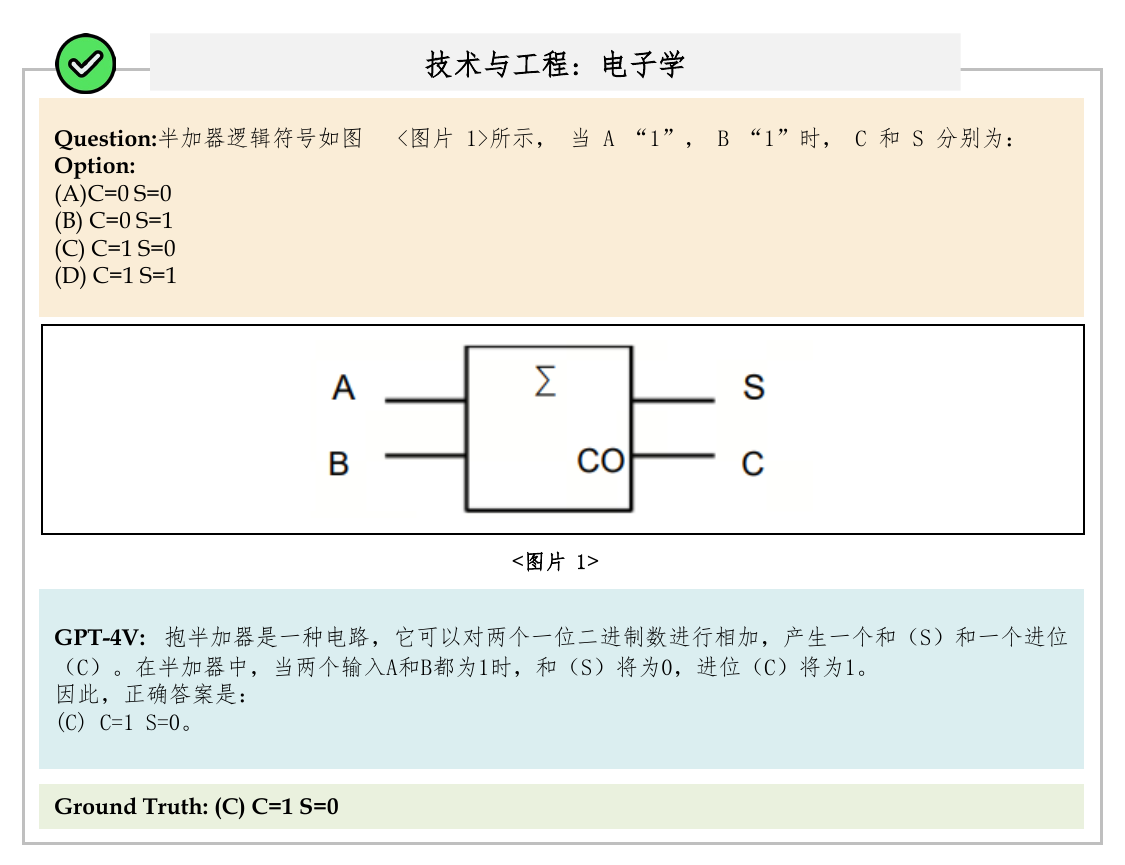}{Electronics 3: Correct Case}{A sample correct case of Technology and Engineering (subcategory: Electronics).}{fig:error_analysis_case_45}

\casestudyfigure{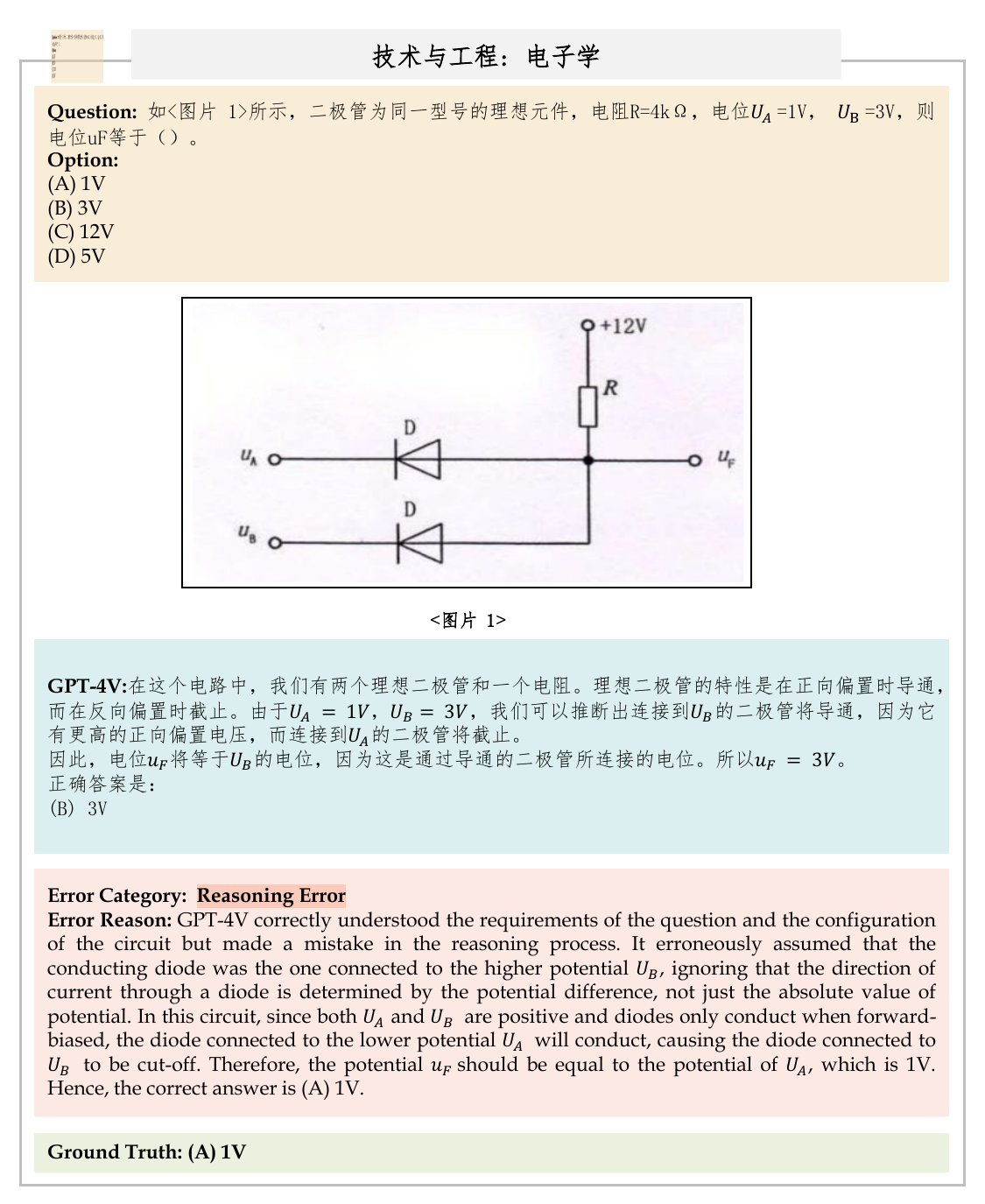}{Electronics 4: Reasoning Error}{A sample error case of Technology and Engineering (subcategory: Electronics).}{fig:error_analysis_case_183}

\casestudyfigure{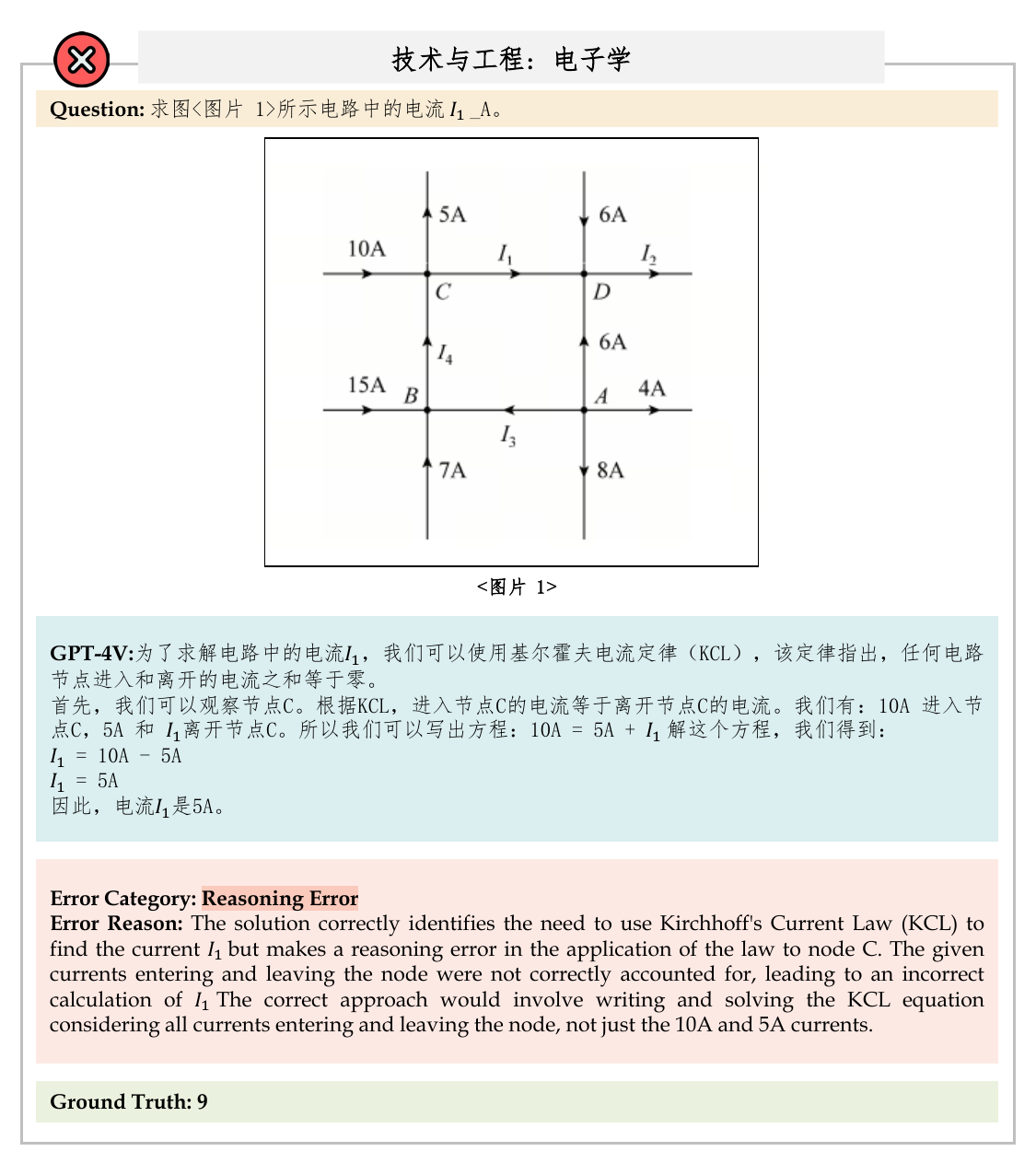}{Electronics 5: Reasoning Error}{A sample error case of Technology and Engineering (subcategory: Electronics).}{fig:error_analysis_case_184}

\casestudyfigure{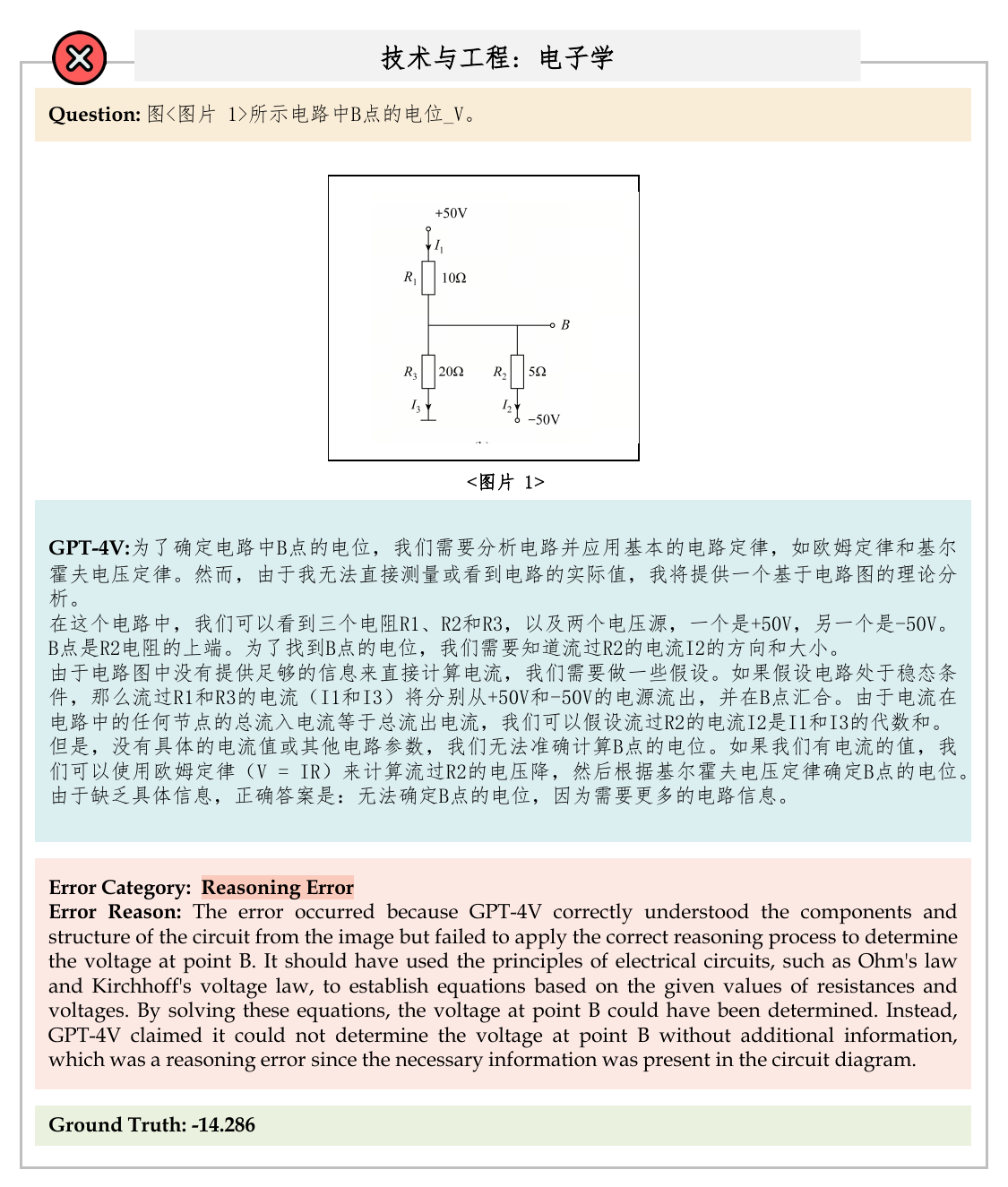}{Electronics 6: Reasoning Error}{A sample error case of Technology and Engineering (subcategory: Electronics).}{fig:error_analysis_case_185}

\casestudyfigure{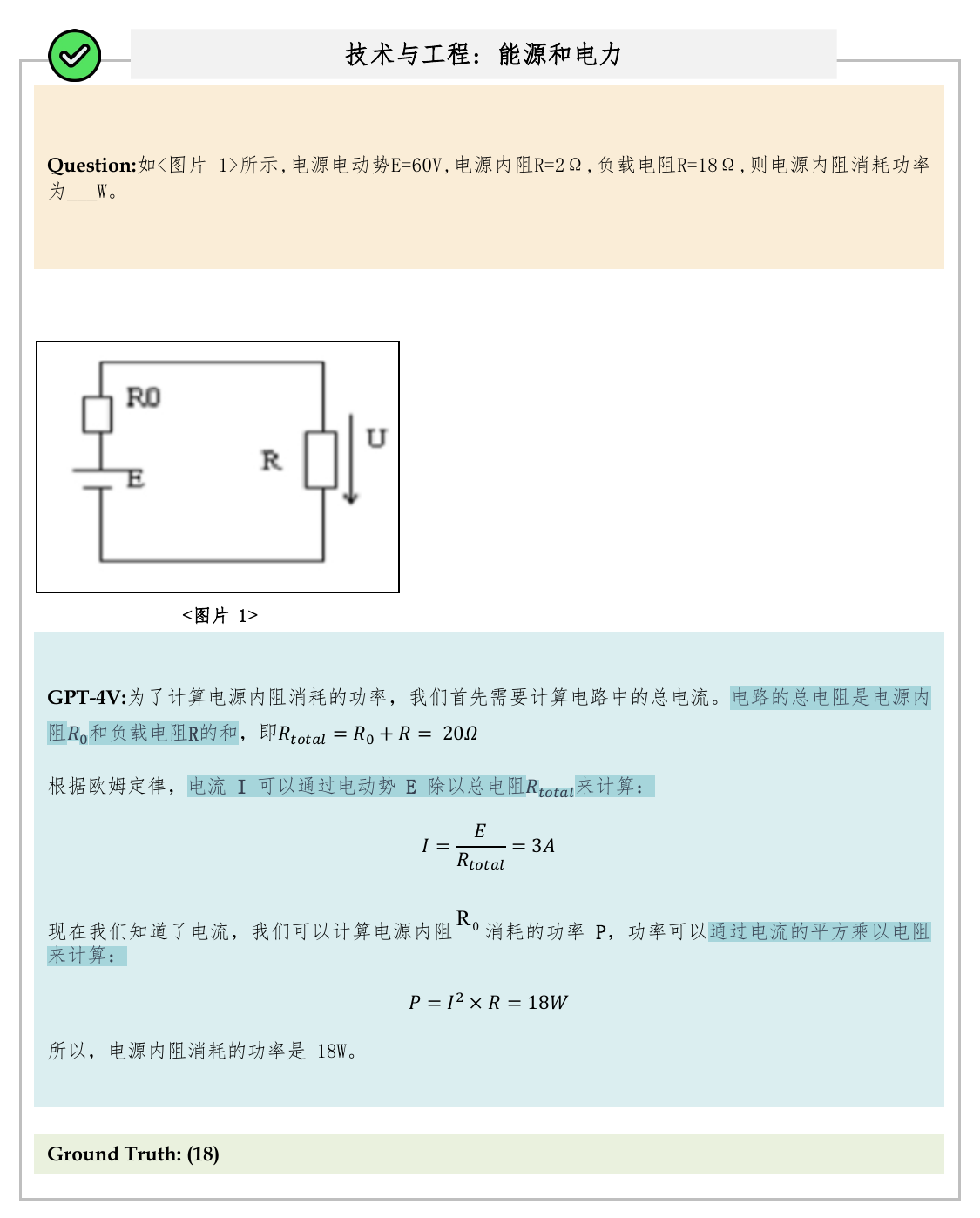}{Energy and Power 1: Correct Case}{A sample correct case of Technology and Engineering (subcategory: Energy and Power).}{fig:error_analysis_case_15}

\casestudyfigure{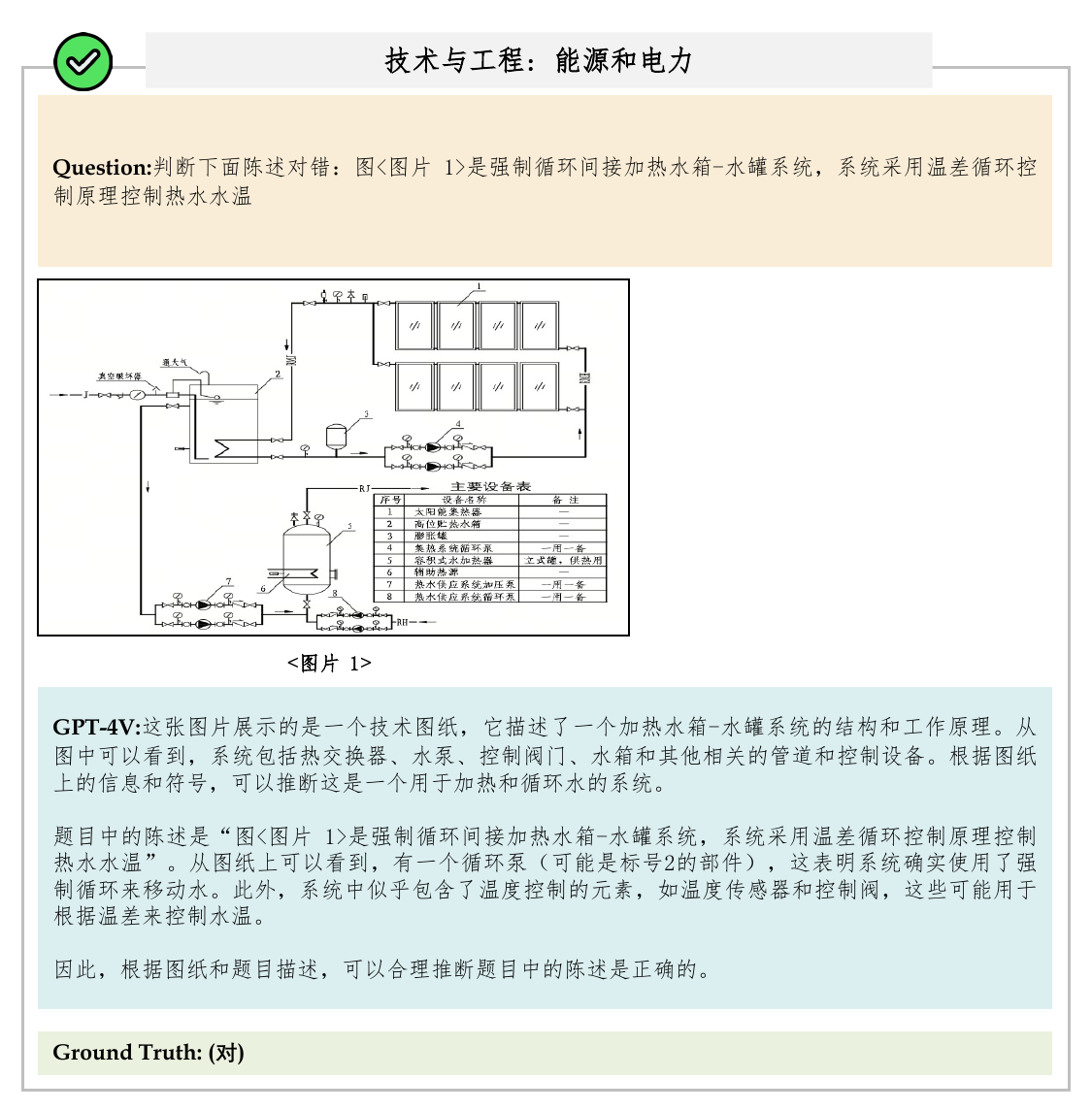}{Energy and Power 2: Correct Case}{A sample correct case of Technology and Engineering (subcategory: Energy and Power).}{fig:error_analysis_case_16}

\casestudyfigure{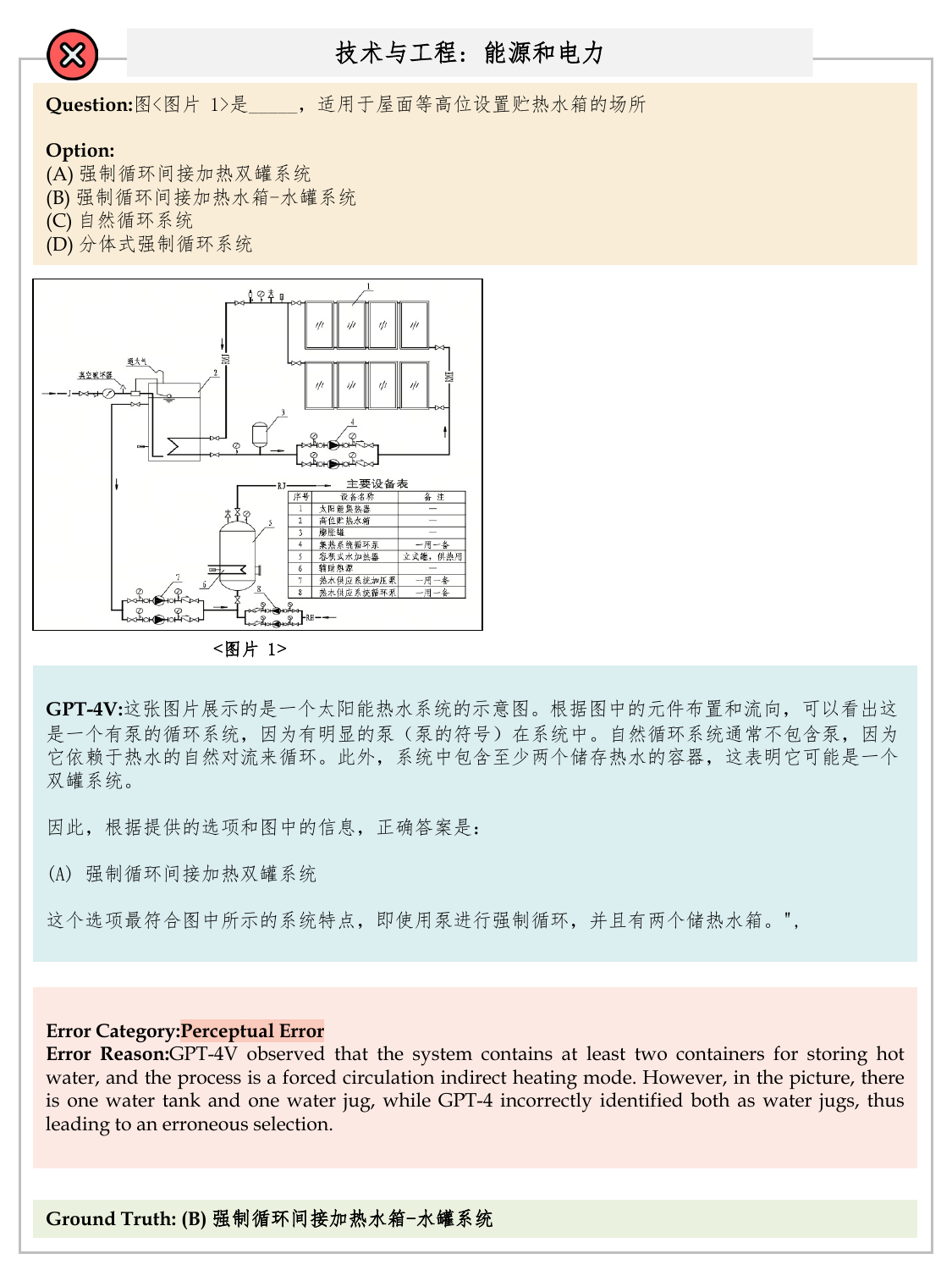}{Energy and Power 3: Perceptual Error}{A sample error case of Technology and Engineering (subcategory: Energy and Power).}{fig:error_analysis_case_106}

\casestudyfigure{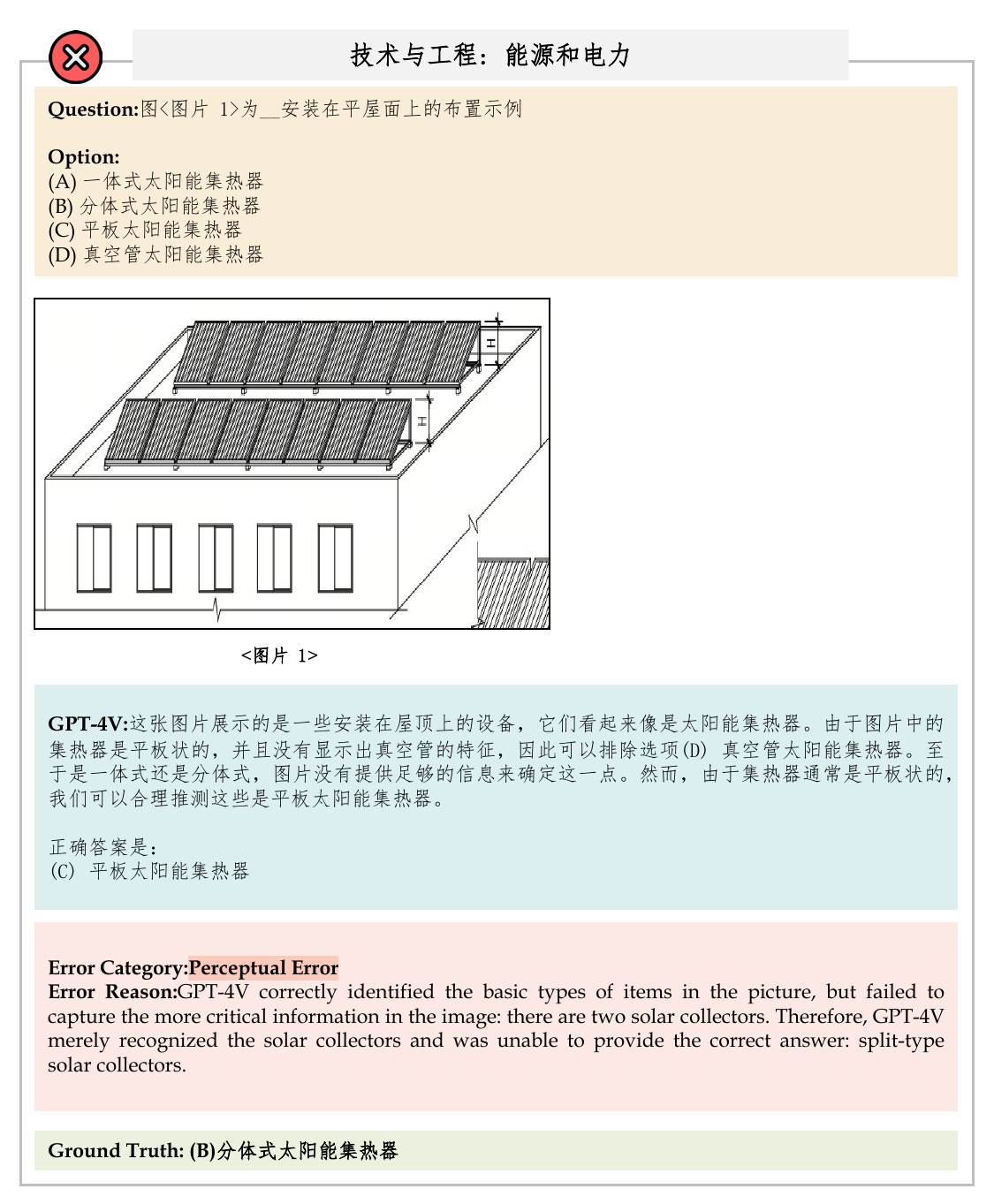}{Energy and Power 4: Perceptual Error}{A sample error case of Technology and Engineering (subcategory: Energy and Power).}{fig:error_analysis_case_107}

\casestudyfigure{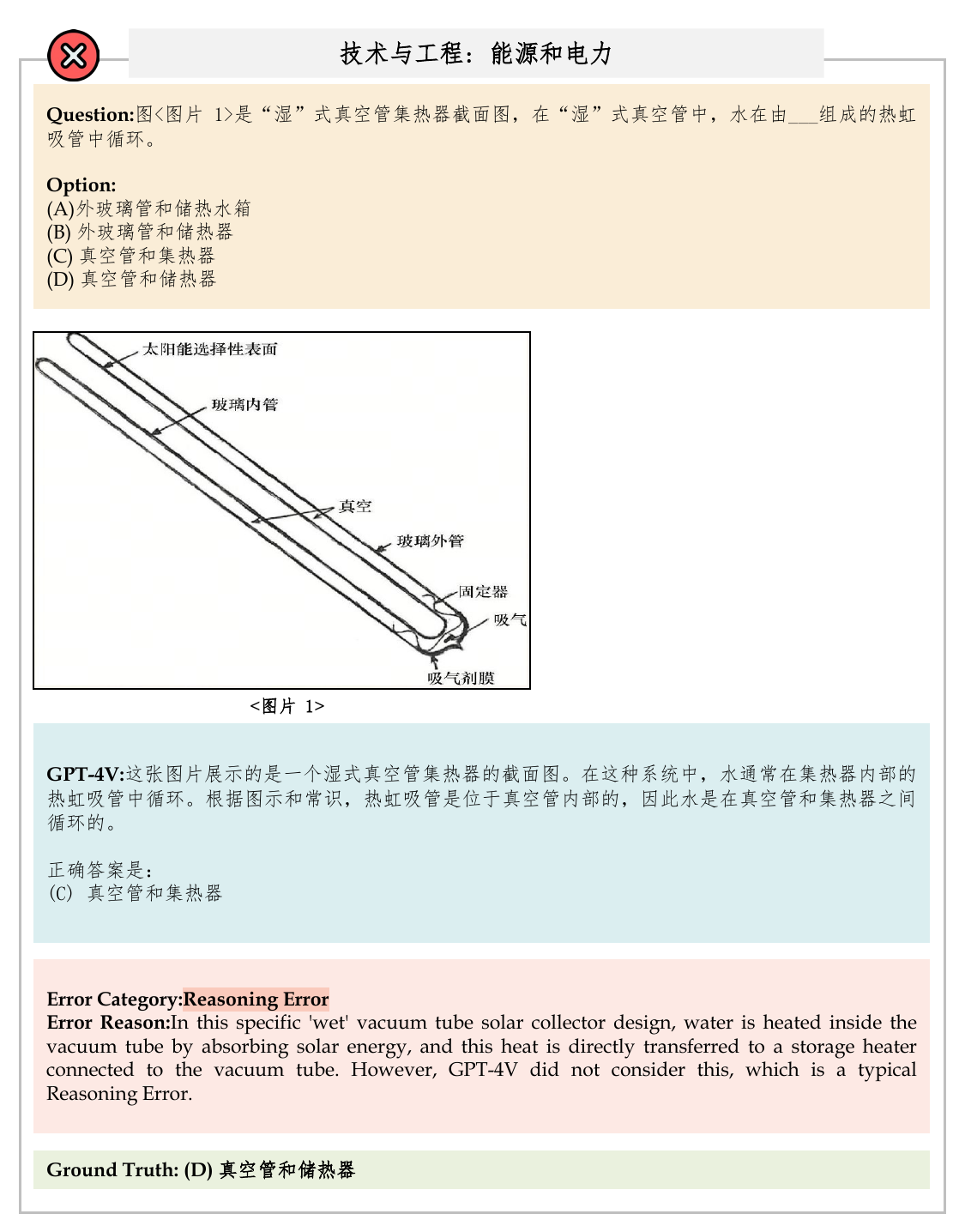}{Energy and Power 5: Reasoning Error}{A sample error case of Technology and Engineering (subcategory: Energy and Power).}{fig:error_analysis_case_108}

\casestudyfigure{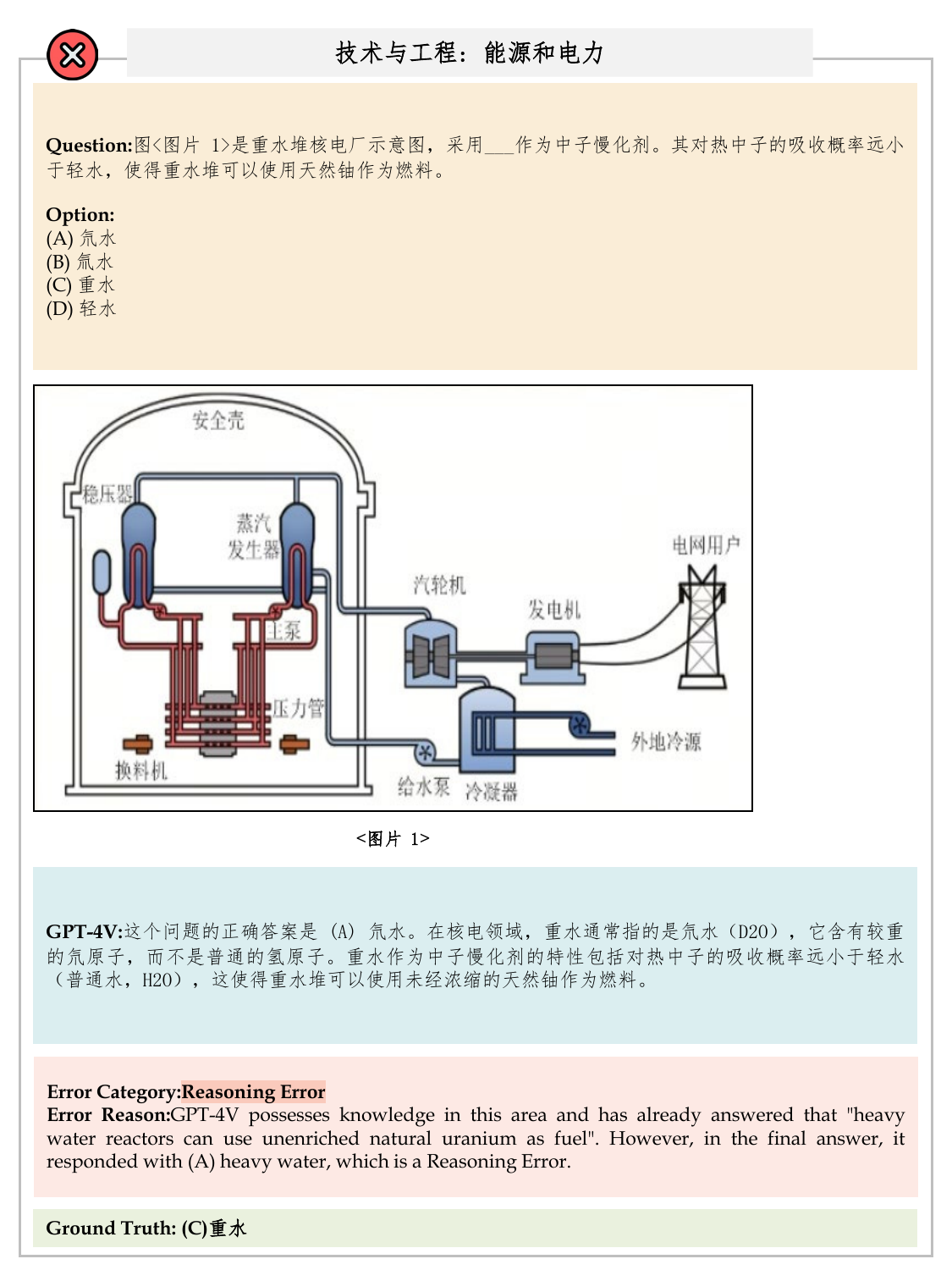}{Energy and Power 6: Reasoning Error}{A sample error case of Technology and Engineering (subcategory: Energy and Power).}{fig:error_analysis_case_112}

\casestudyfigure{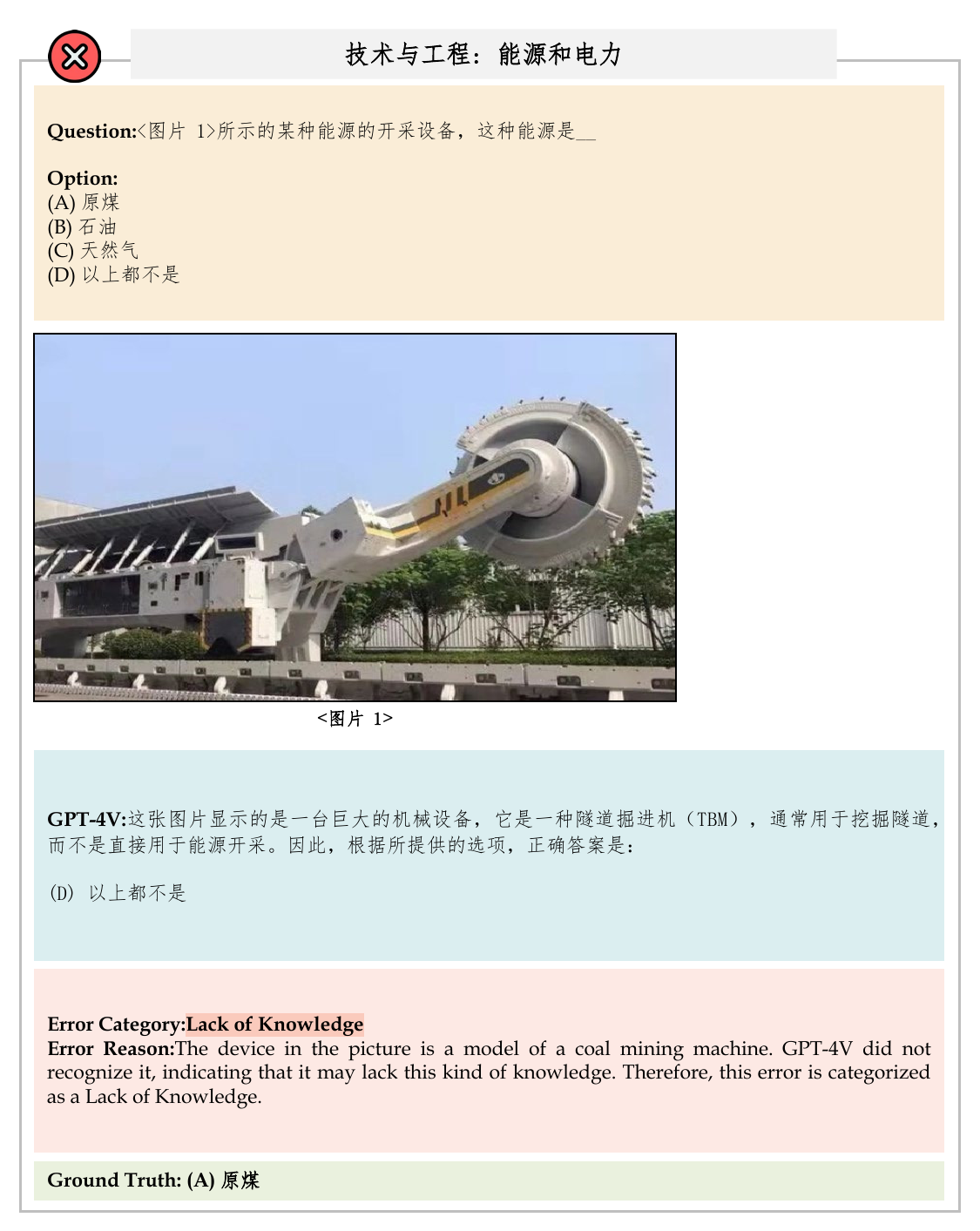}{Energy and Power 7: Lack of Knowledge}{A sample error case of Technology and Engineering (subcategory: Energy and Power).}{fig:error_analysis_case_109}

\casestudyfigure{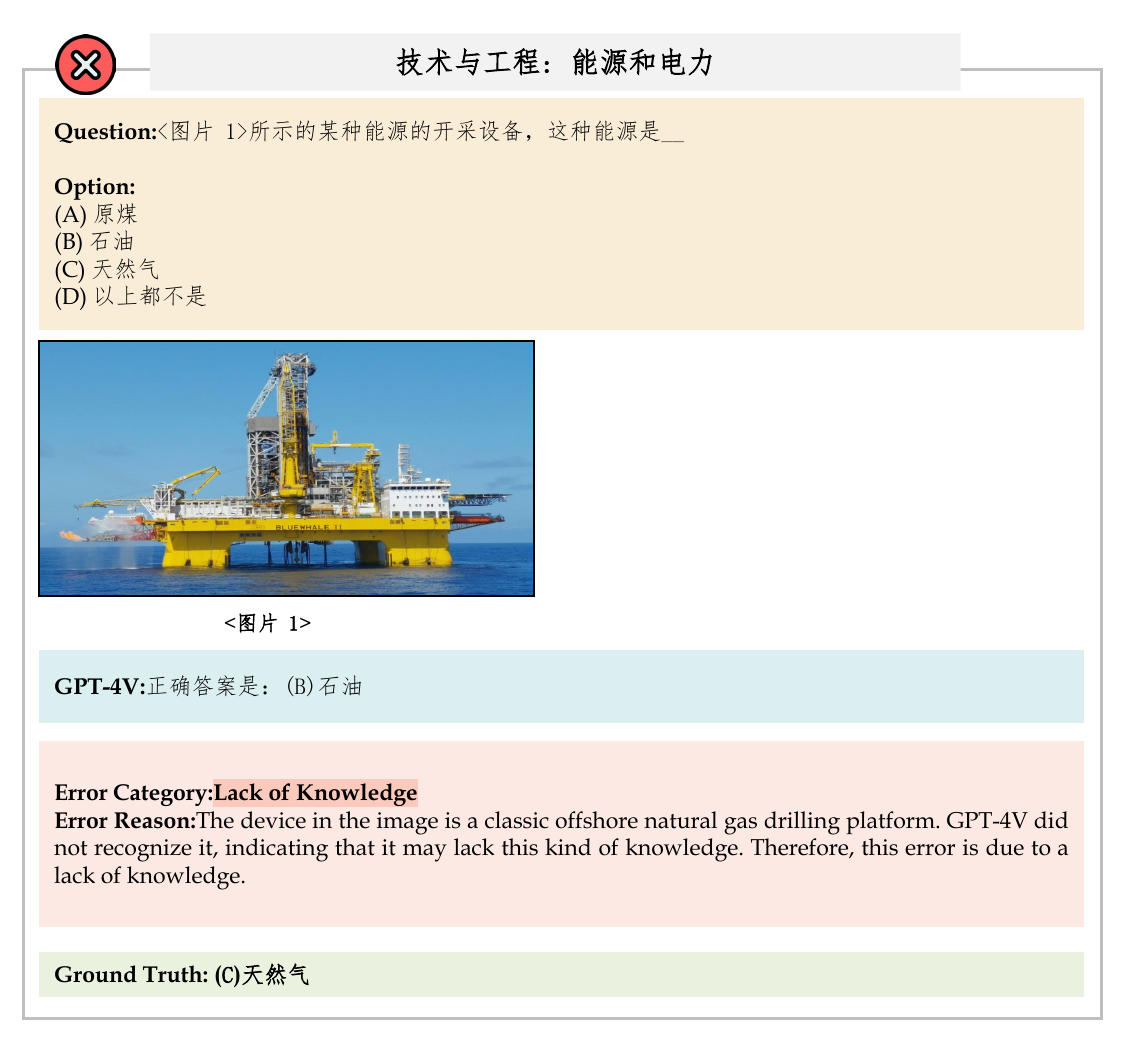}{Energy and Power 8: Lack of Knowledge}{A sample error case of Technology and Engineering (subcategory: Energy and Power).}{fig:error_analysis_case_110}

\casestudyfigure{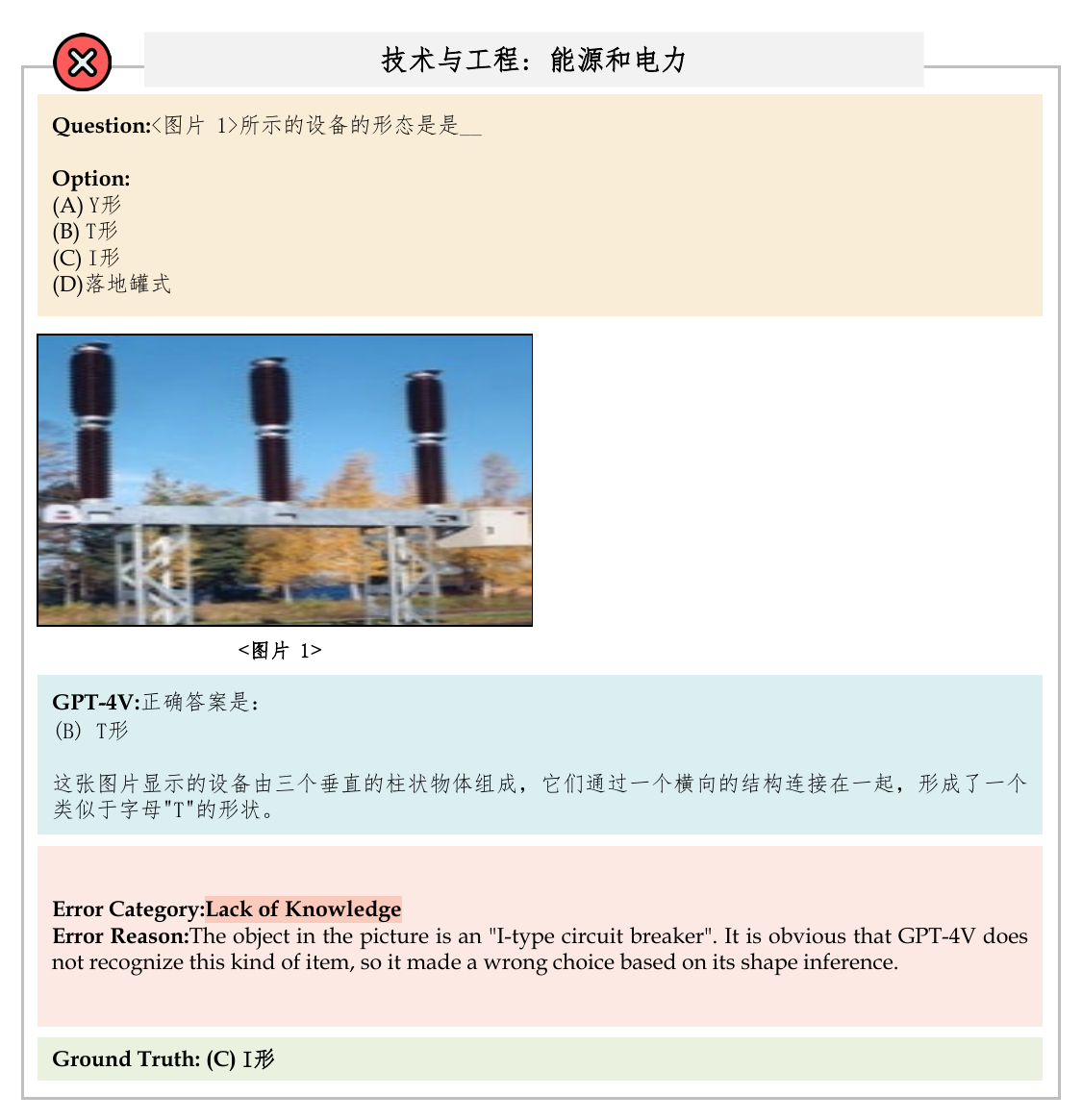}{Energy and Power 9: Lack of Knowledge}{A sample error case of Technology and Engineering (subcategory: Energy and Power).}{fig:error_analysis_case_111}

\casestudyfigure{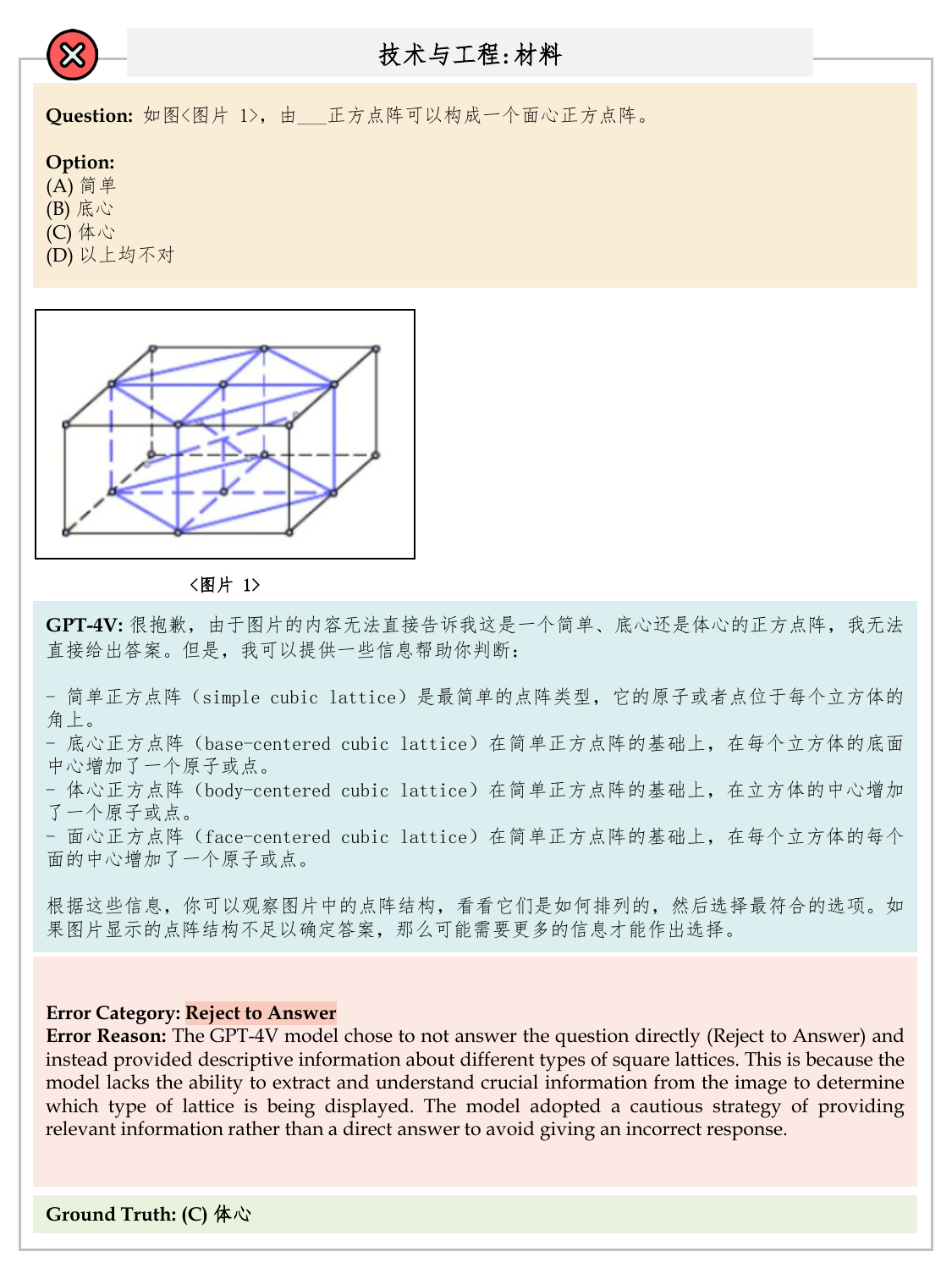}{Materials 1: Reject to Answer}{A sample error case of Technology and Engineering (subcategory: Materials).}{fig:error_analysis_case_167}

\casestudyfigure{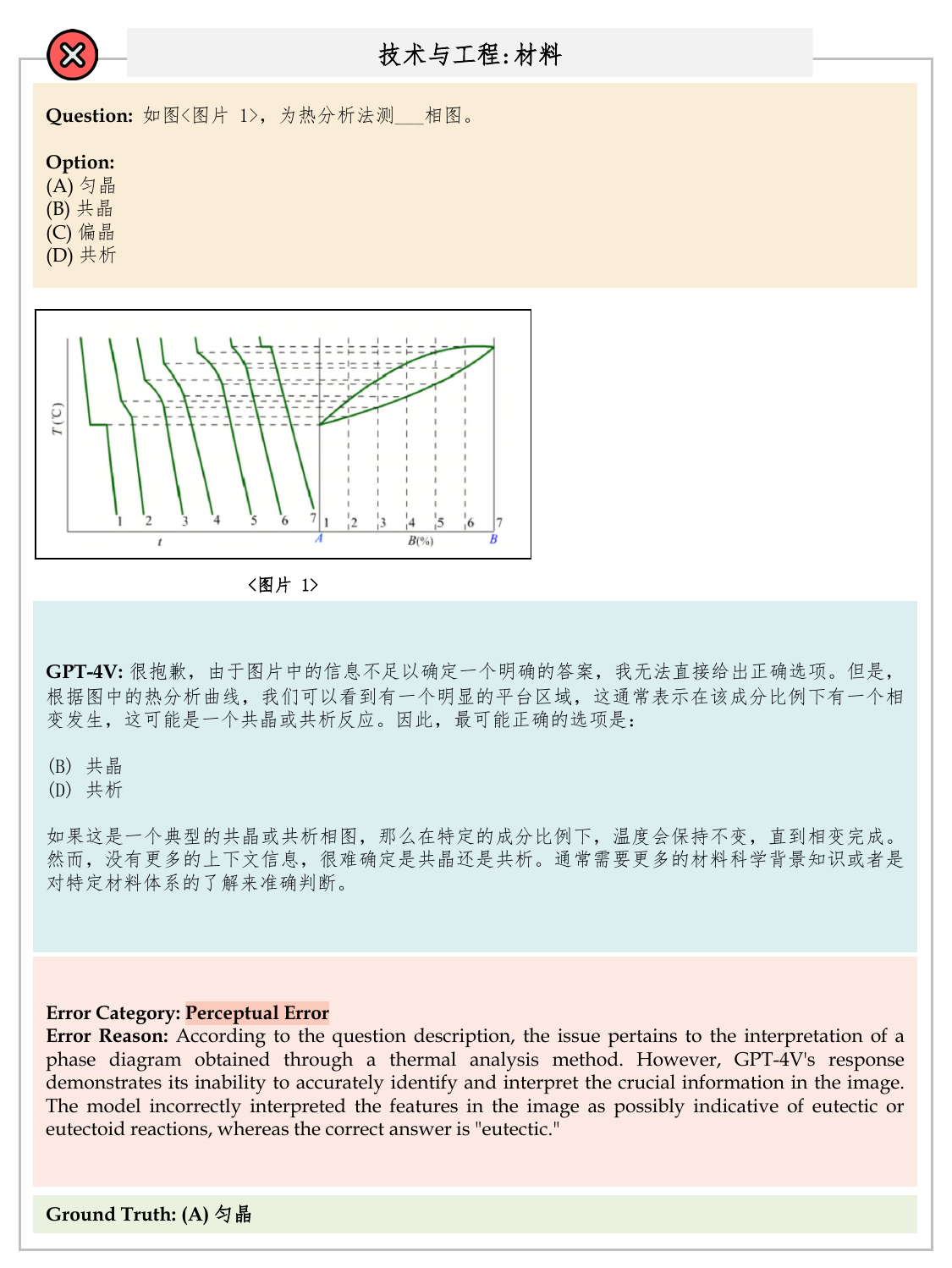}{Materials 2: Perceptual Error}{A sample error case of Technology and Engineering (subcategory: Materials).}{fig:error_analysis_case_169}

\casestudyfigure{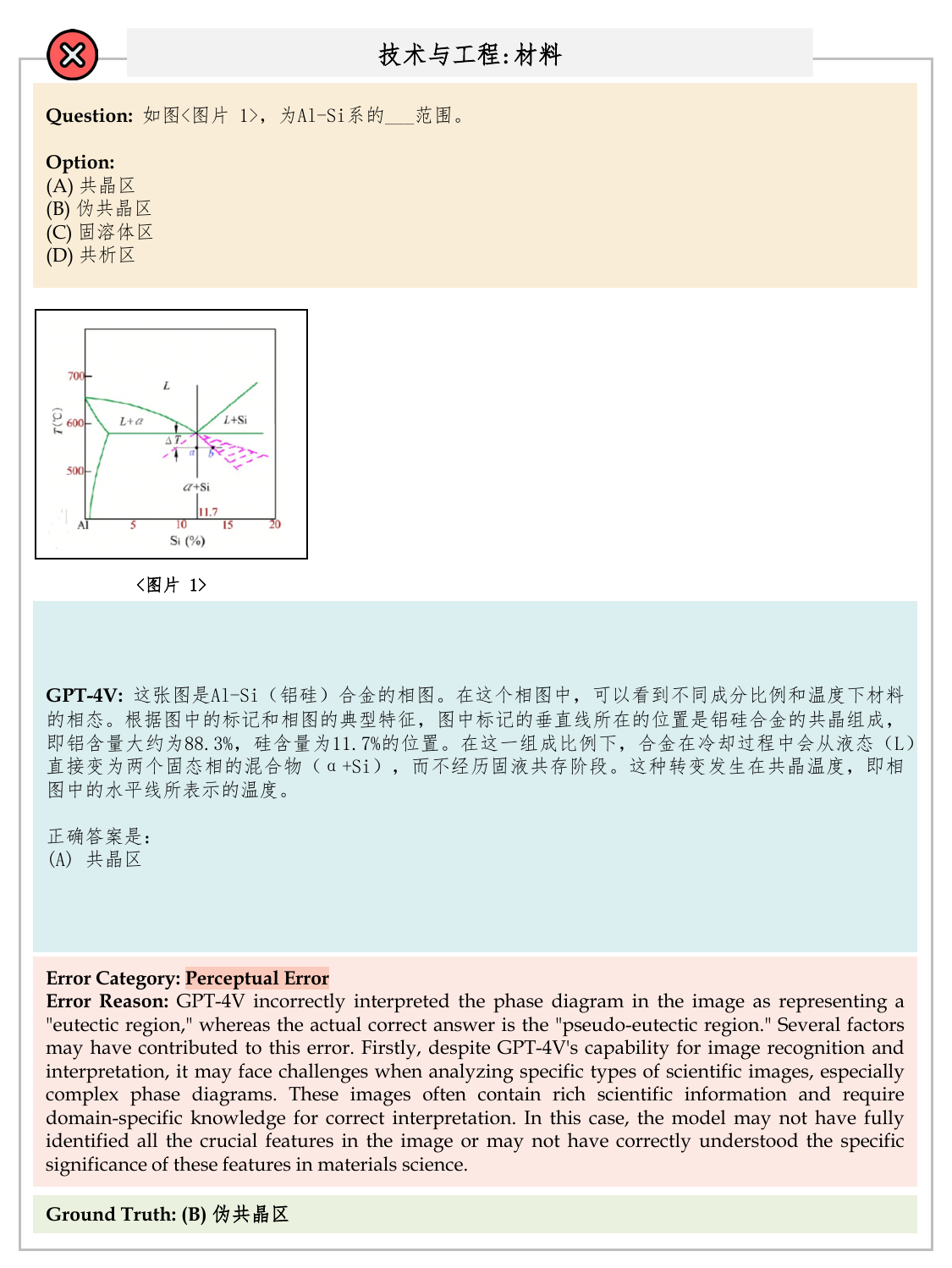}{Materials 3: Perceptual Error}{A sample error case of Technology and Engineering (subcategory: Materials).}{fig:error_analysis_case_170}

\casestudyfigure{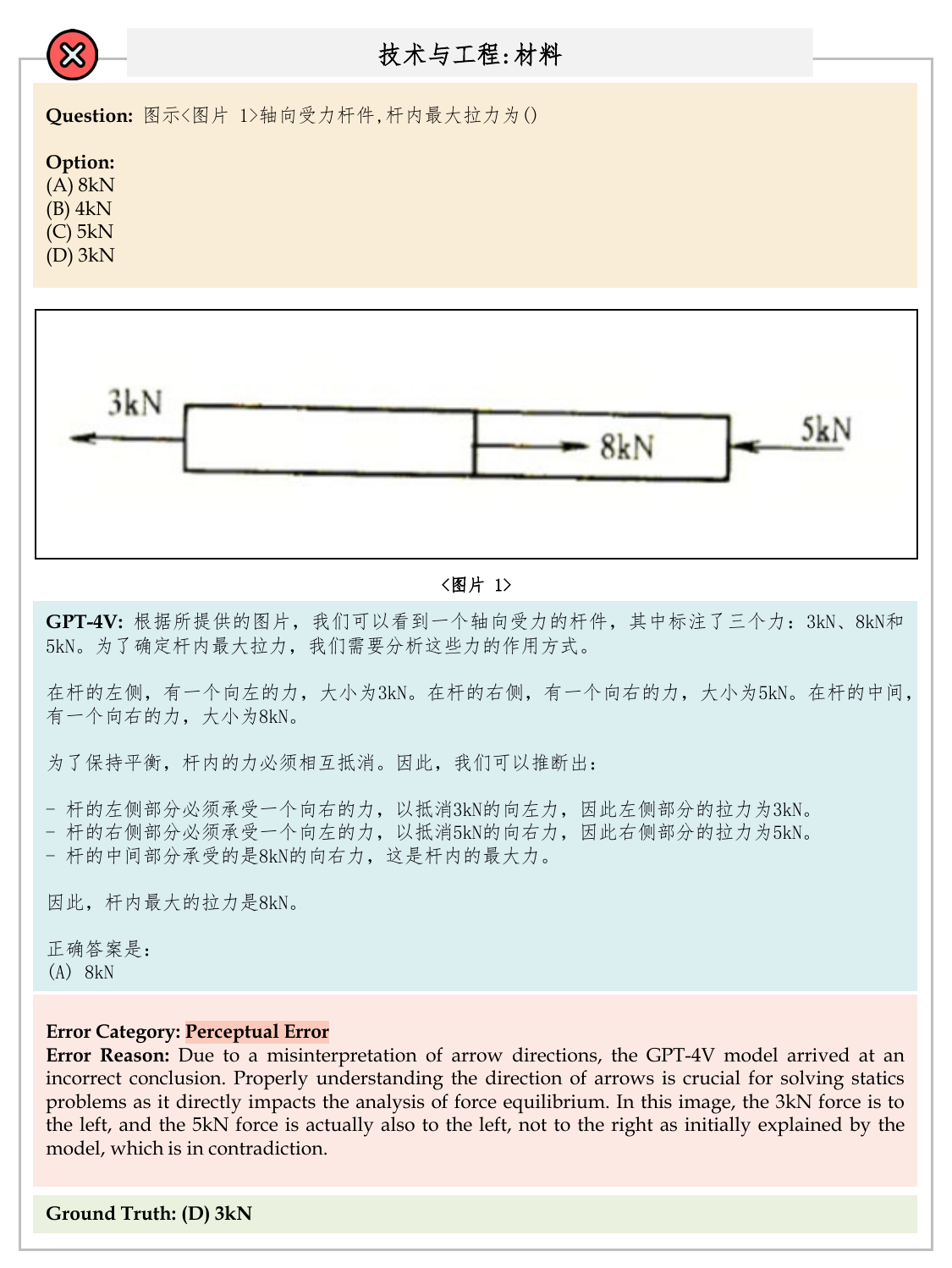}{Materials 4: Perceptual Error}{A sample error case of Technology and Engineering (subcategory: Materials).}{fig:error_analysis_case_171}

\casestudyfigure{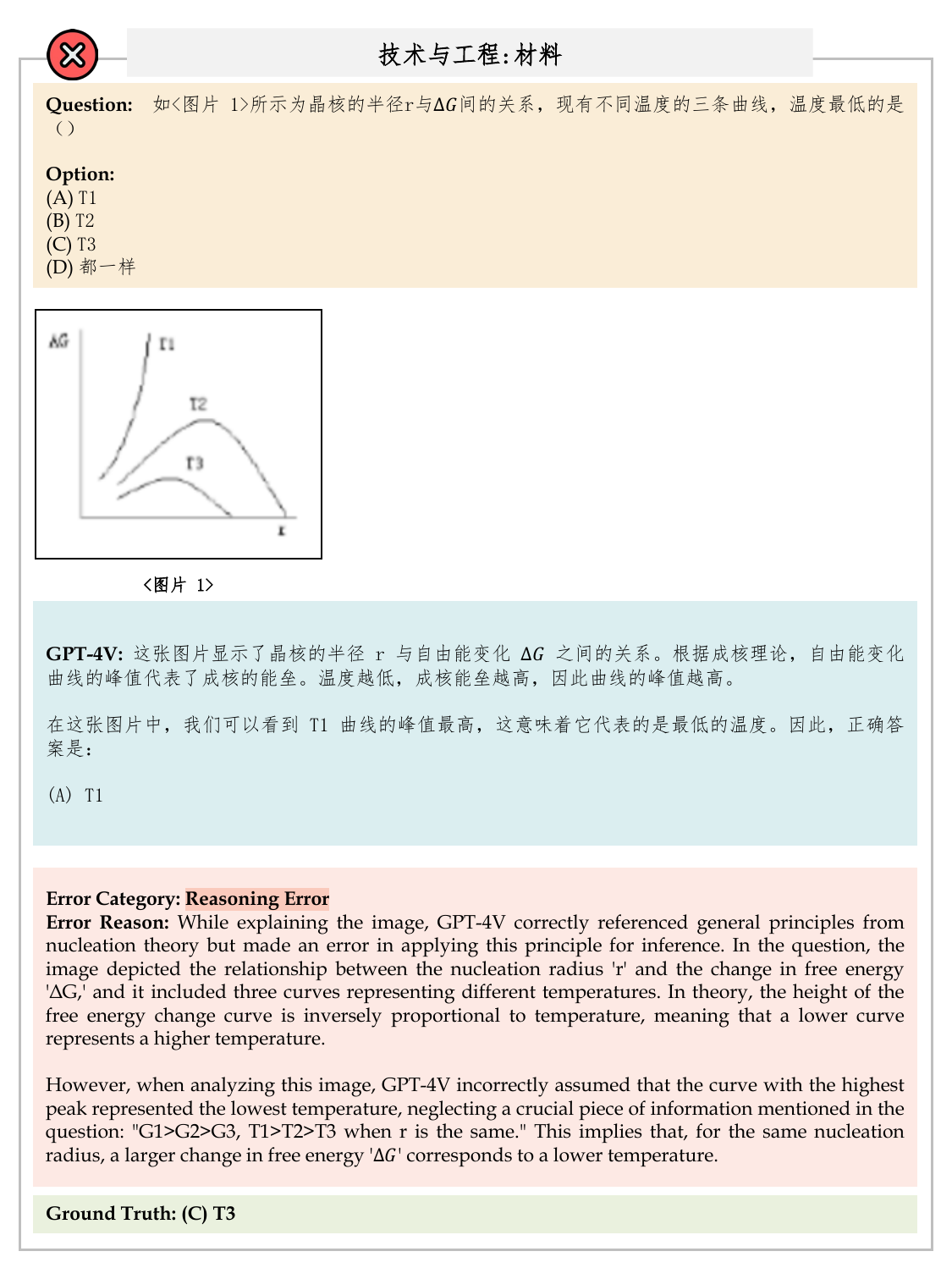}{Materials 5: Reasoning Error}{A sample error case of Technology and Engineering (subcategory: Materials).}{fig:error_analysis_case_166}

\casestudyfigure{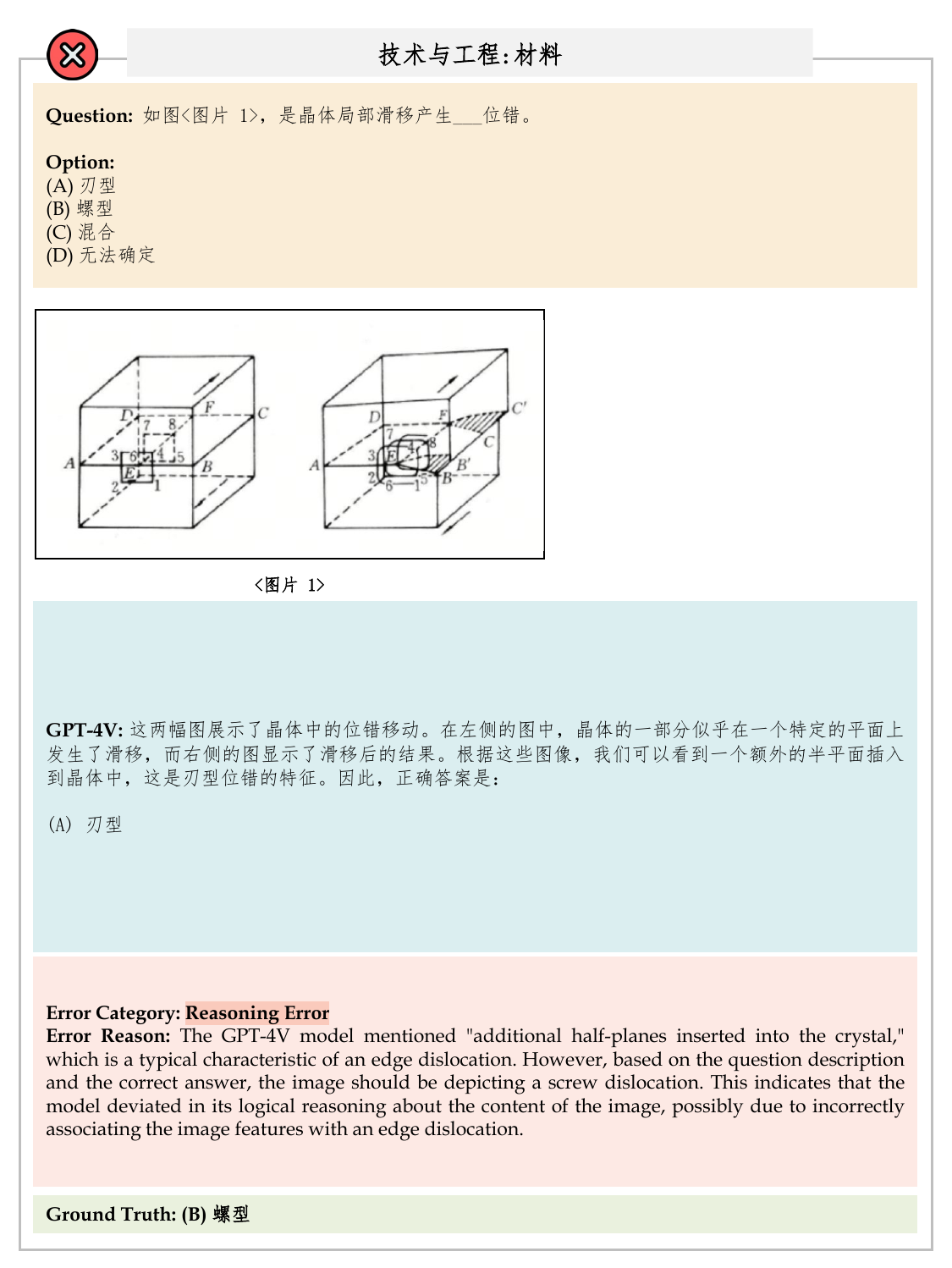}{Materials 6: Reasoning Error}{A sample error case of Technology and Engineering (subcategory: Materials).}{fig:error_analysis_case_168}

\casestudyfigure{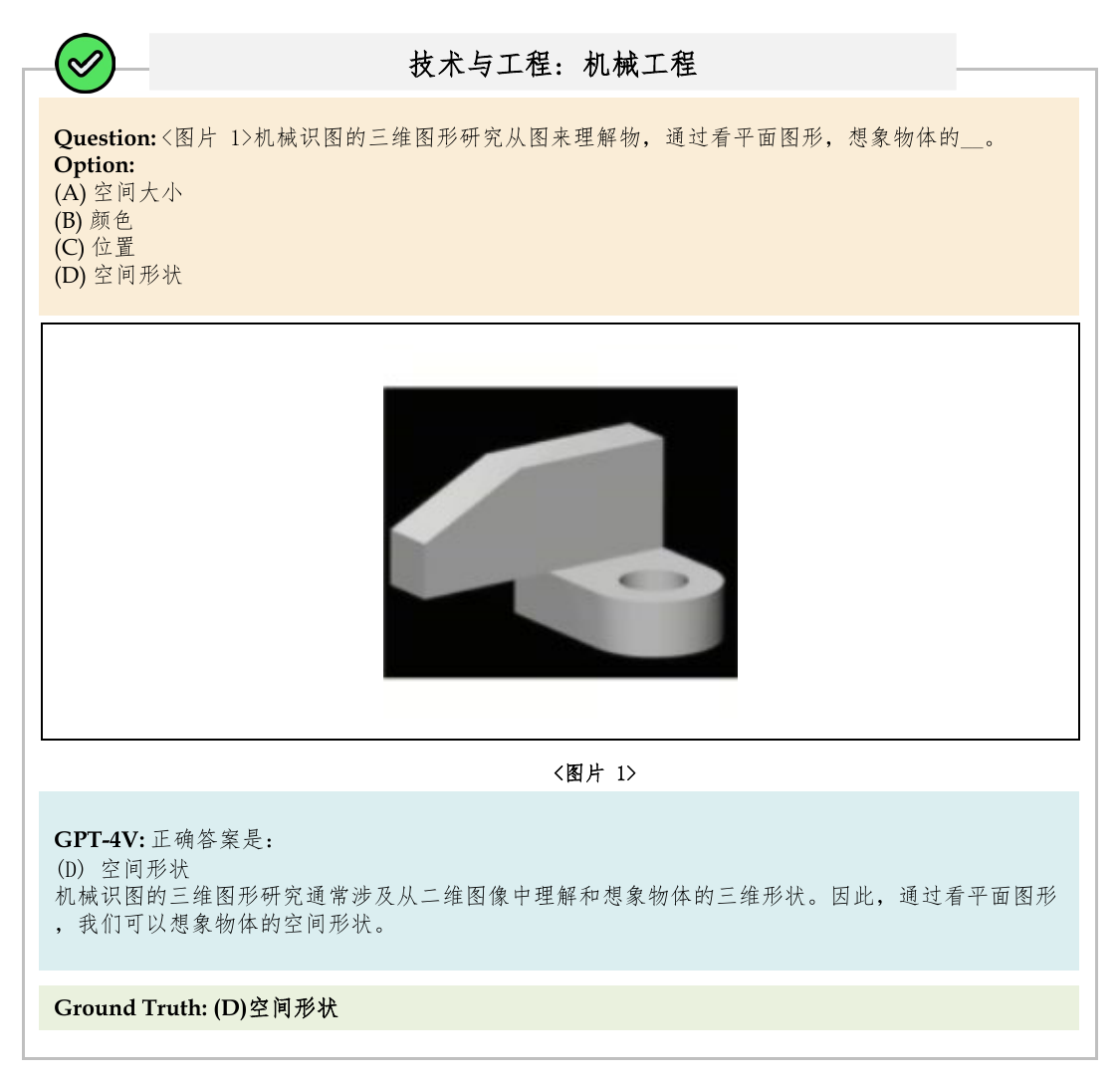}{Mechanical Engineering 1: Correct Case}{A sample correct case of Technology and Engineering (subcategory: Mechanical Engineering).}{fig:error_analysis_case_3}

\casestudyfigure{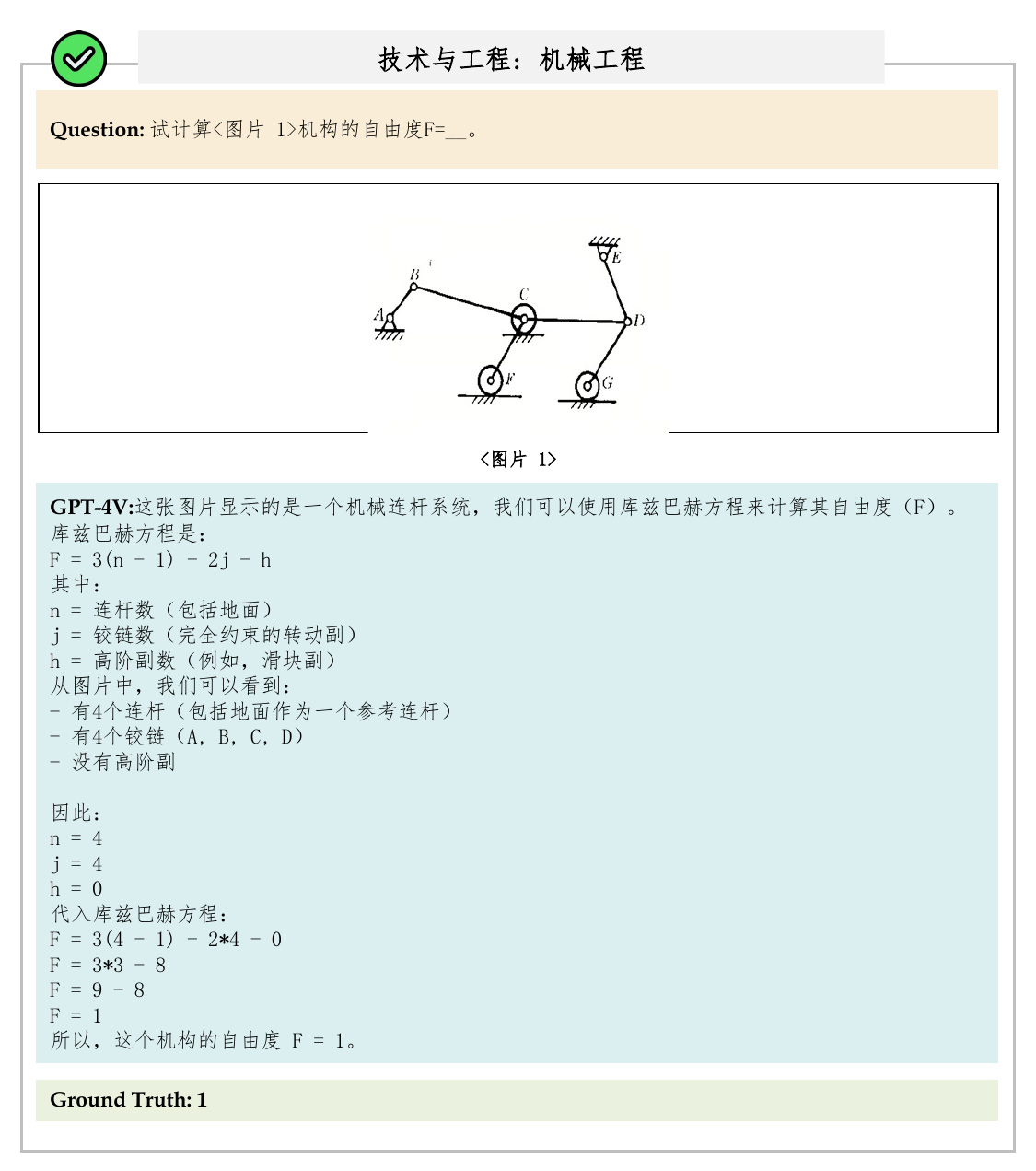}{Mechanical Engineering 2: Correct Case}{A sample correct case of Technology and Engineering (subcategory: Mechanical Engineering).}{fig:error_analysis_case_4}

\casestudyfigure{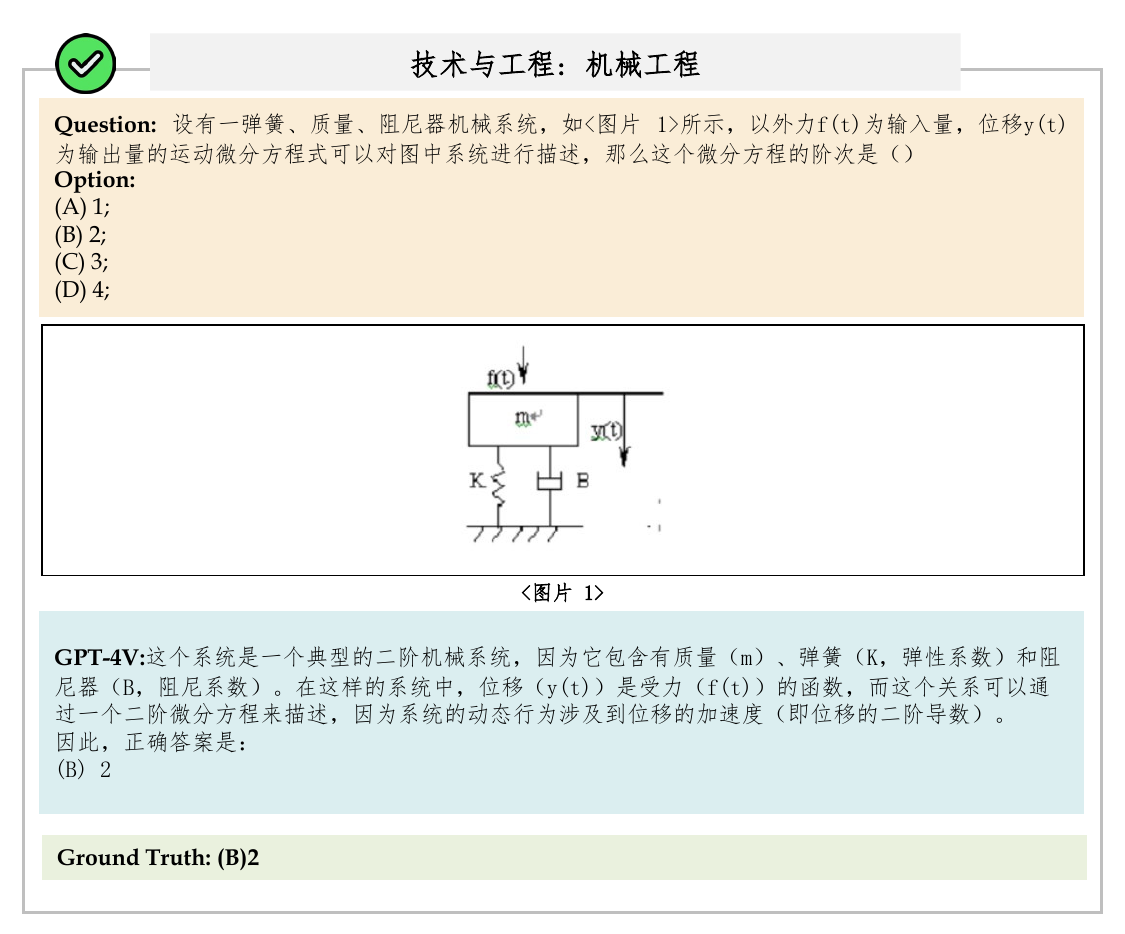}{Mechanical Engineering 3: Correct Case}{A sample correct case of Technology and Engineering (subcategory: Mechanical Engineering).}{fig:error_analysis_case_5}

\casestudyfigure{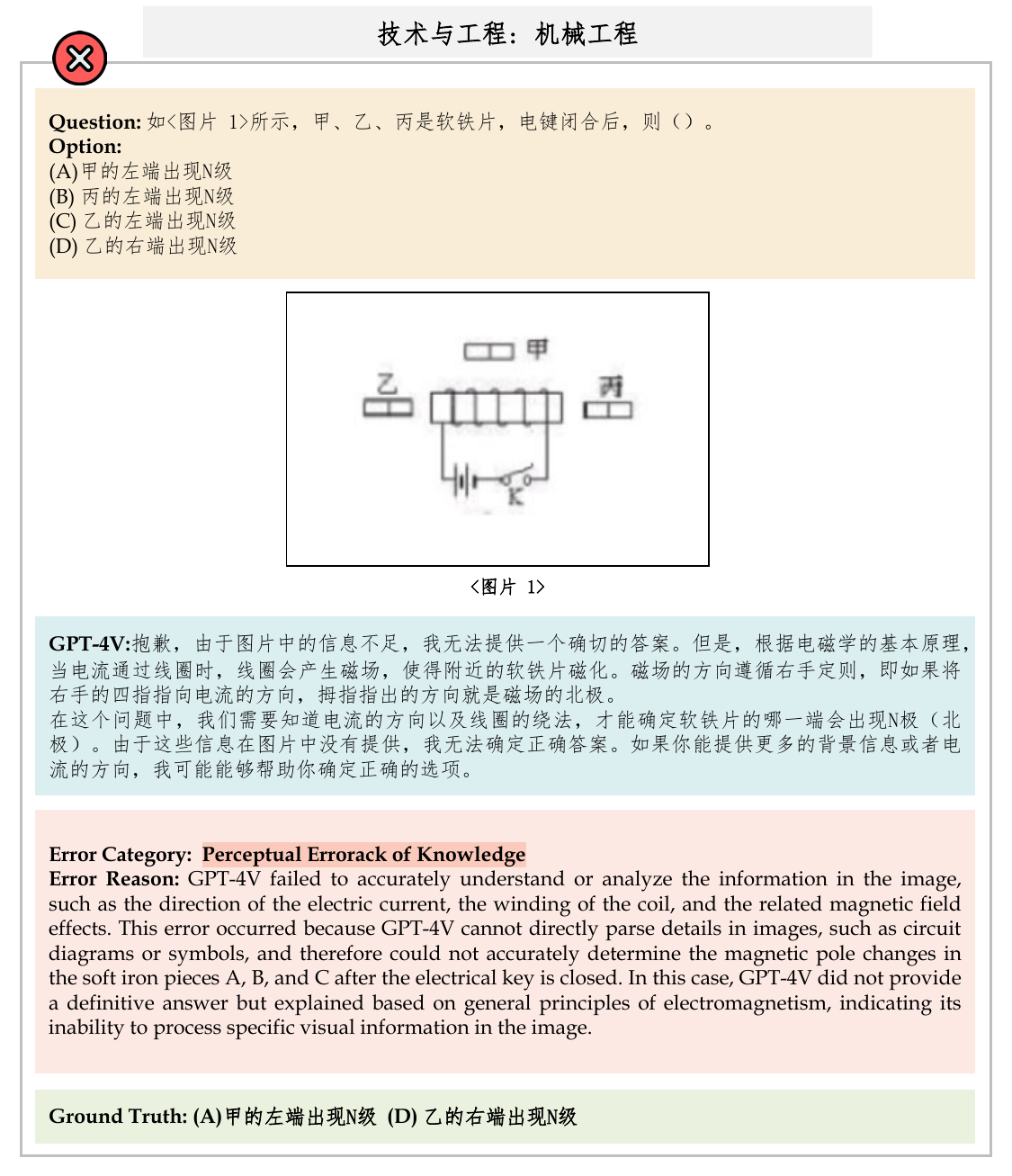}{Mechanical Engineering 4: Perceptual Error, Lack of Knowledge}{A sample error case of Technology and Engineering (subcategory: Mechanical Engineering).}{fig:error_analysis_case_82}

\casestudyfigure{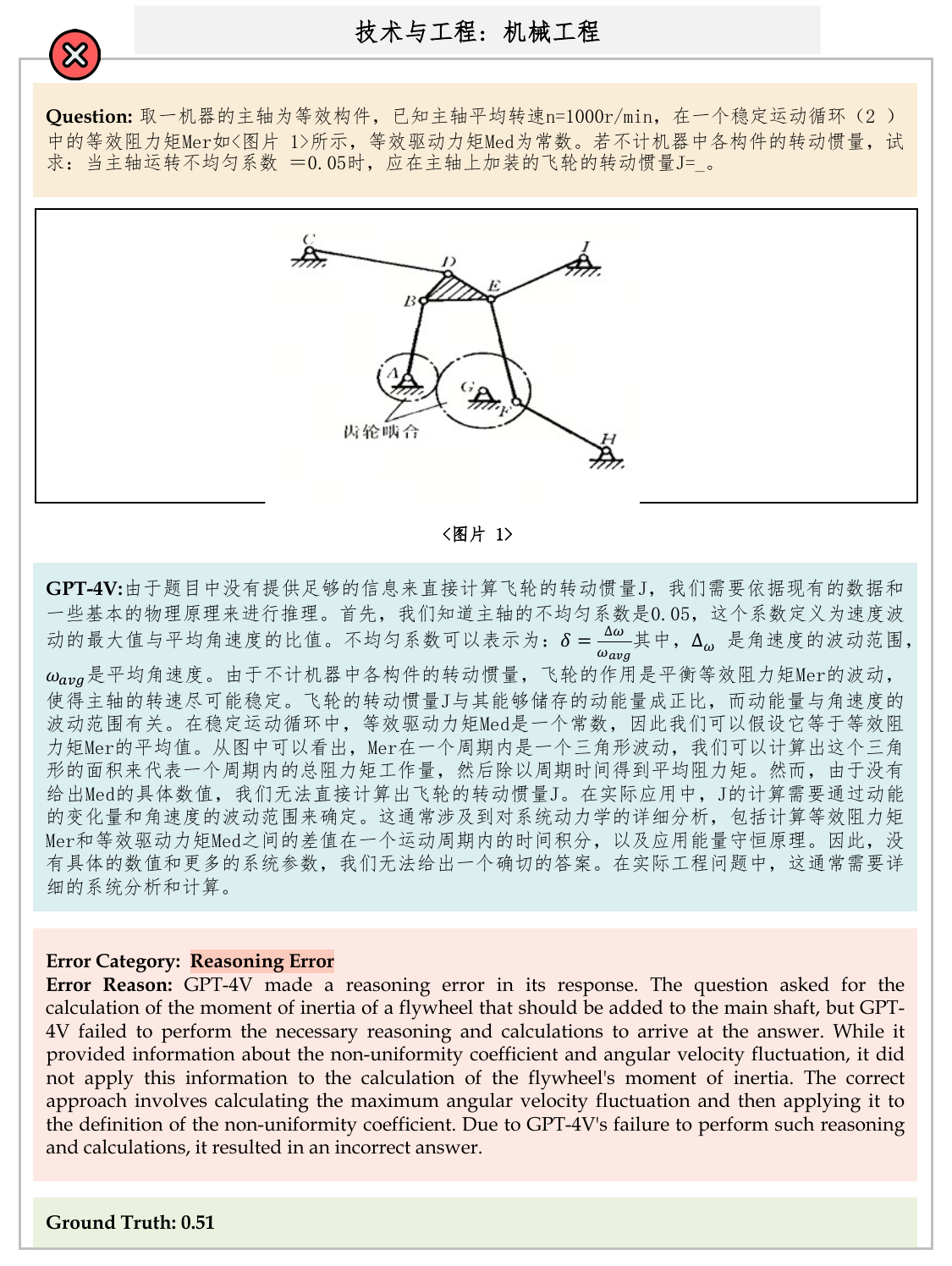}{Mechanical Engineering 5: Reasoning Error}{A sample error case of Technology and Engineering (subcategory: Mechanical Engineering).}{fig:error_analysis_case_84}

\casestudyfigure{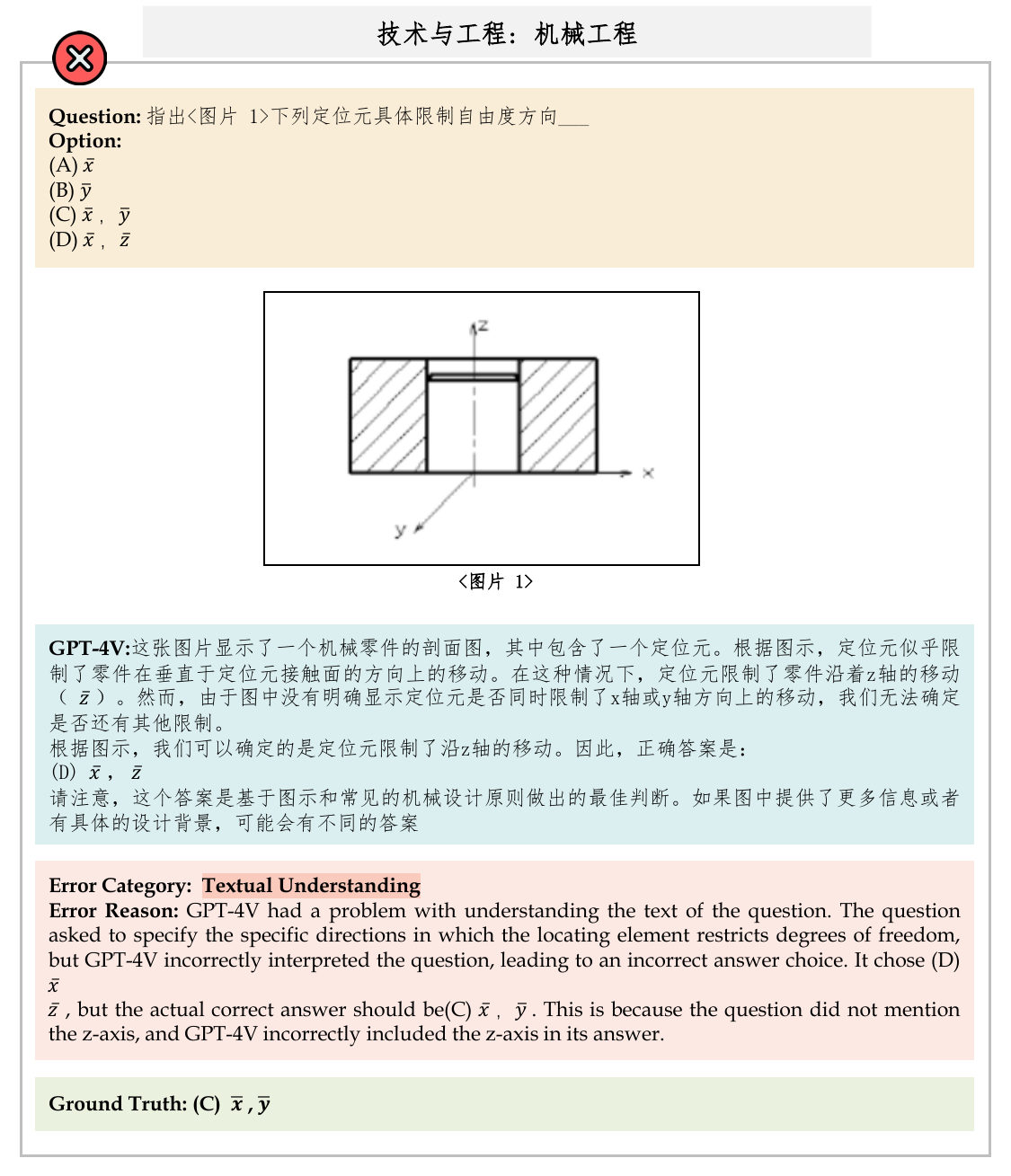}{Mechanical Engineering 6: Textual Understanding}{A sample error case of Technology and Engineering (subcategory: Mechanical Engineering).}{fig:error_analysis_case_83}

\casestudyfigure{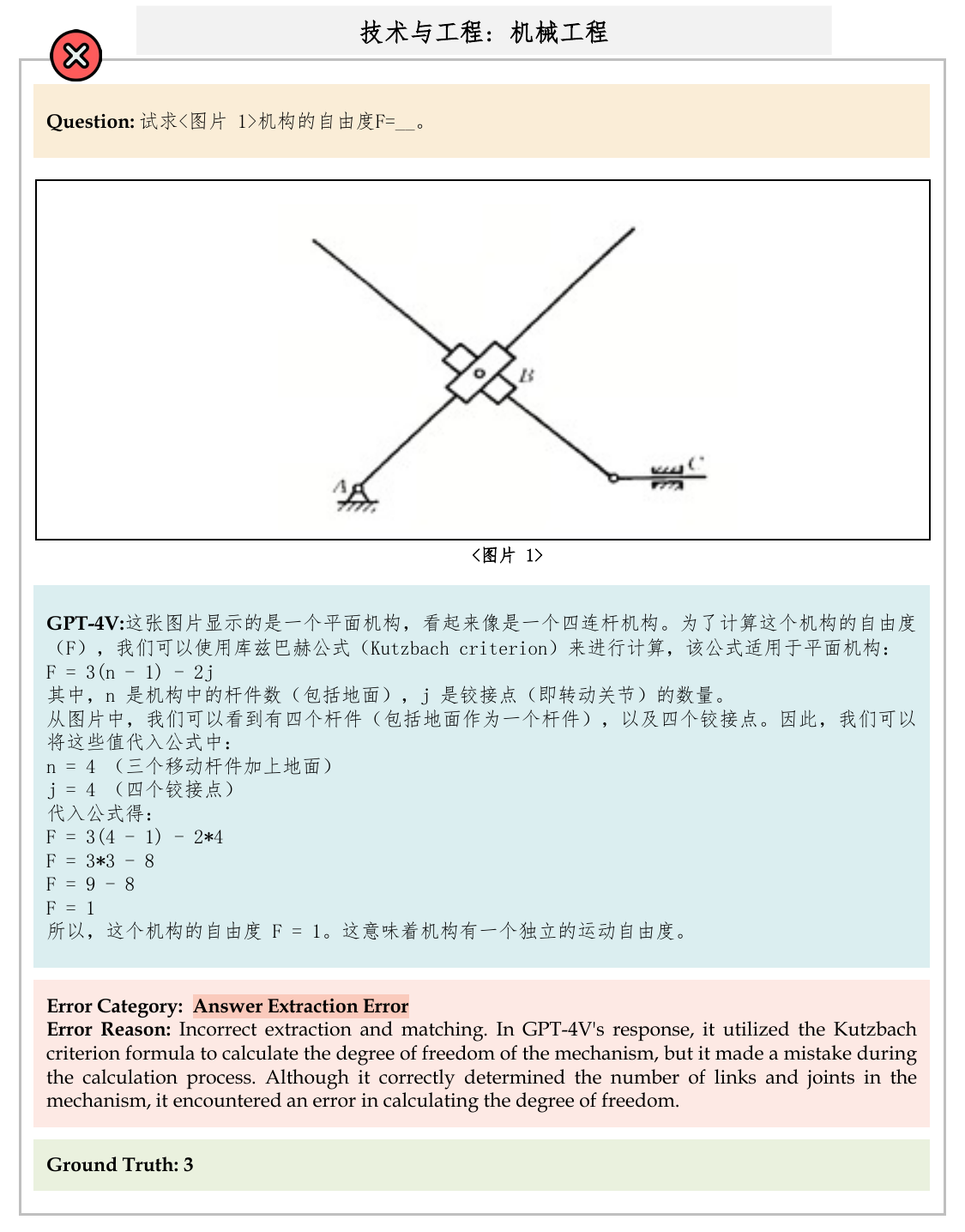}{Mechanical Engineering 7: Answer Extraction Error}{A sample error case of Technology and Engineering (subcategory: Mechanical Engineering).}{fig:error_analysis_case_86}

\casestudyfigure{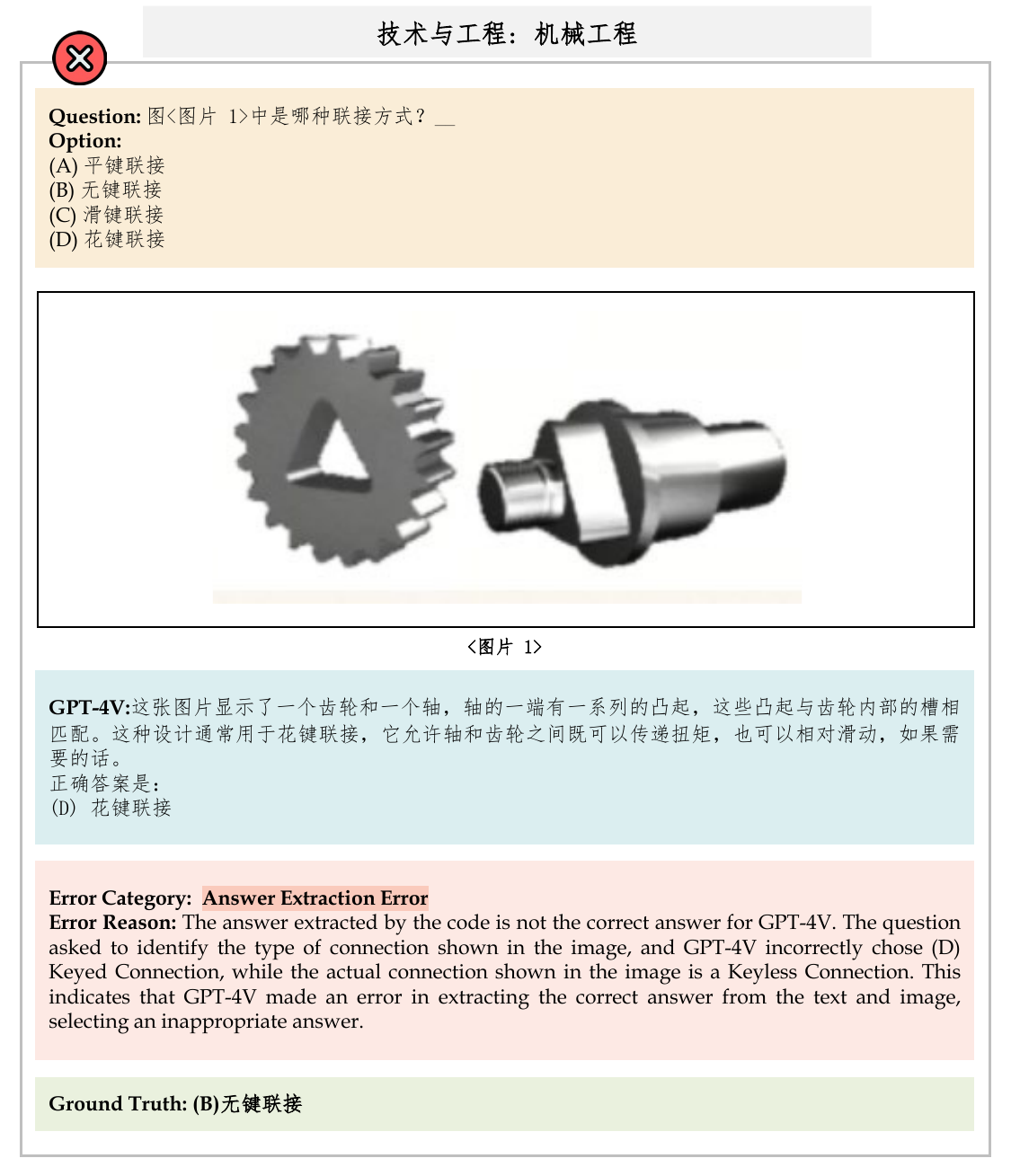}{Mechanical Engineering 8: Answer Extraction Error}{A sample error case of Technology and Engineering (subcategory: Mechanical Engineering).}{fig:error_analysis_case_87}

\casestudyfigure{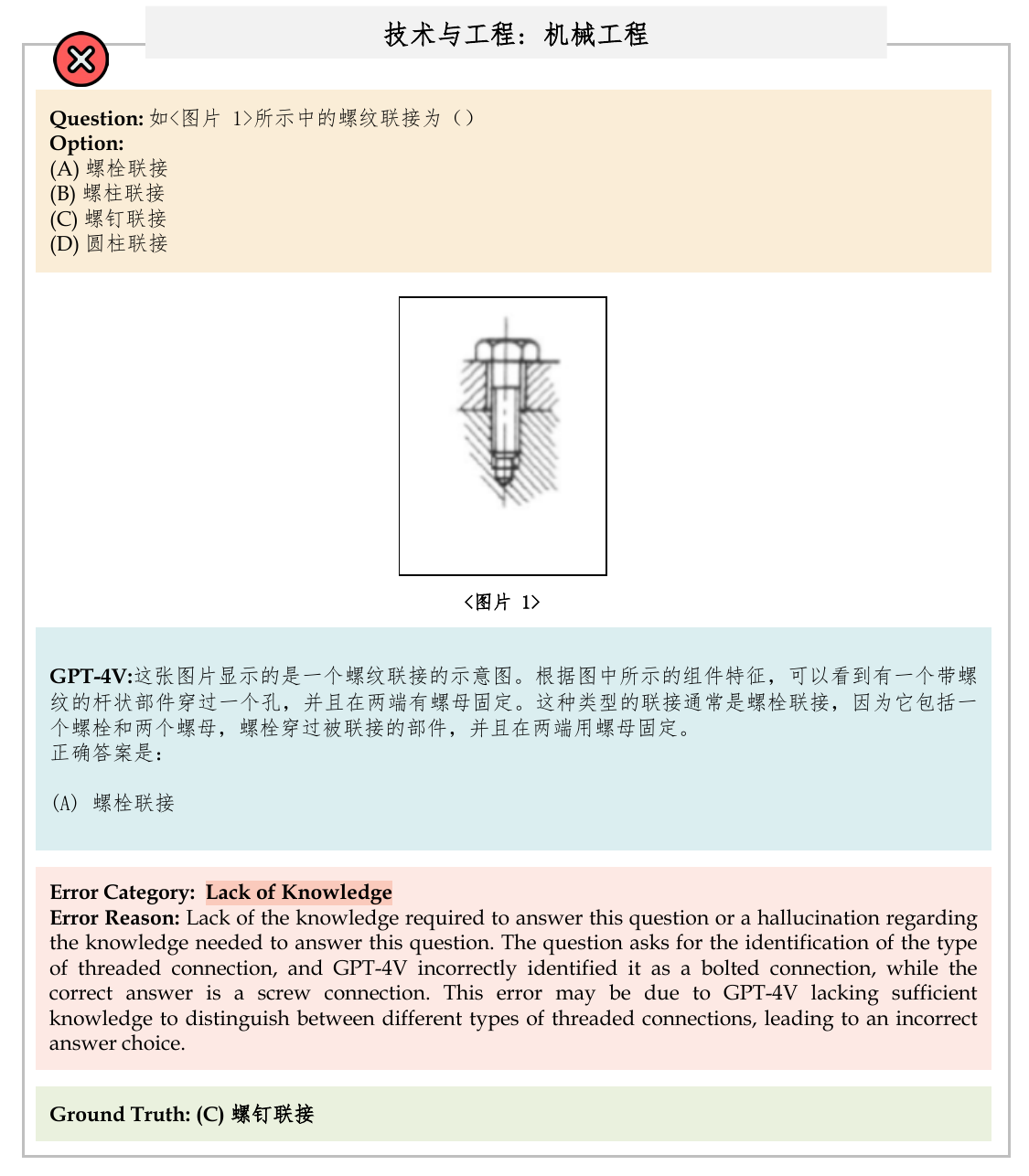}{Mechanical Engineering 9: Lack of Knowledge}{A sample error case of Technology and Engineering (subcategory: Mechanical Engineering).}{fig:error_analysis_case_85}

\end{document}